%% file: main.tex
\documentclass[preprint]{article}
\usepackage{neurips_2026}
\usepackage{amsmath}
\usepackage{amssymb,amsfonts,amsthm}
\usepackage{mathrsfs}
\usepackage{mathtools}
\usepackage{algorithm}
\usepackage{algpseudocode}
\usepackage{xcolor}
\usepackage{pgfplots}
\usepgfplotslibrary{groupplots}
\pgfplotsset{compat=1.18}
\usepackage{pgfplotstable}
\pgfplotsset{compat=1.18}
\usepackage{tikz}
\usetikzlibrary{arrows.meta, positioning,patterns}
\usepackage{dsfont}
\usepackage{thmtools}
\usepackage{thm-restate}
\usepackage{enumitem}
\usepackage{dsfont}
\usepackage{graphicx}
\usepackage{subcaption}
\usepackage{sidecap}
\usepackage{accents}
\usepackage{wrapfig}
\usepackage{booktabs} 
\usepackage{longtable}
\usepackage{siunitx}

\newtheorem{proposition}{Proposition}[]

\usepackage{hyperref}
\usepackage{xcolor}
\hypersetup{
  colorlinks   = true, 
  urlcolor     = blue, 
  linkcolor    = blue, 
  citecolor   = blue 
}
\usepackage[capitalize, nameinlink]{cleveref}

\usepackage[utf8]{inputenc} 
\usepackage[T1]{fontenc}    
\usepackage{hyperref}       
\usepackage{url}            
\usepackage{booktabs}       
\usepackage{amsfonts}       
\usepackage{nicefrac}       
\usepackage{microtype}      
\usepackage{xcolor}

\newcommand{\PMI}{\mathsf{K}_\mathsf{PMI}}
\newcommand{\rhsbox}[1]{\hfill\boxed{#1}} 
\newcommand{\corr}{\mathsf{corr}}

\title{Representation Alignment Rests on Linear Structure}
\author{%
  Kiril Bangachev\thanks{Author names appear in alphabetical order by last name.}\thanks{Supported by NAE Grand Challenge Vest Fellowship.} \\
  \texttt{kirilb@mit.edu} \\
  \And 
  Guy Bresler\thanks{Supported by NSF award 2428619.}\\
  \texttt{guy@mit.edu}\\
  \And
  Yury Polyanskiy\thanks{Supported by NSF Grant CCF-21-12665 and by generous gifts from Jane Street and Google Research.}\\
  \texttt{yp@mit.edu}\\ \\
  Department of Electrical Engineering and Computer Science\\
  Massachusetts Institute of Technology\\
  Cambridge, MA, 02139 \\}

\date{May 2025}

\begin{document}

\maketitle

\begin{abstract}
We investigate the Platonic Representation Hypothesis (PRH) through a tripartite statistical framework of representations: signal, bias, and noise. {1) Signal:} We propose that Platonic alignment arises from the universal relationship between objects and attributes, which is encoded linearly in representations according to the Linear Representation Hypothesis (LRH). We provide evidence that LRH helps explain PRH by extracting linear object-attribute features with sparse autoencoders and showing that these sparse representations often exhibit stronger cross-modal alignment than their dense counterparts. 
{2) Bias:} Models have different implicit biases due to the diverse architectures and training procedures used. We show that this difference can be partially mitigated. Centering and normalization consistently improve cross-model alignment. {3) Noise:} Finite-sample training leads to noise in representations. We provide evidence that representational noise is driven by data scarcity by revealing a strong and consistent positive correlation between word frequency and alignment in LLMs and text embedding models. Synthesizing signal, bias, and noise, we propose a statistical model that refines the Linear Representation Hypothesis and explains further phenomena related to the alignment of representations emerging from diverse modern AI architectures. 
All code is available at \href{https://github.com/BangachevKiril/L2PRH}{github/L2PRH}.
\end{abstract}

\input{sections/Introduction}

\input{sections/Prelim}

\input{sections/Experiments}

\input{sections/Limitations}

\section*{Acknowledgments}
We would like to thank Enric Boix-Adsera for a helpful conversation regarding Sparse Authoencoders and Sharut Gupta for many conversations on representation learning. We would also like to thank Victor Butoi and Michael Hla for helpful feedback on the exposition.

\newpage
\bibliography{ref}
\bibliographystyle{alpha}

\appendix
\onecolumn

\input{sections/StatModel}

\input{sections/PseudoCode}

\input{sections/FurtherExperiments}

\end{document}

%% file: sections/Introduction.tex
\section{Introduction}

\subsection{The Platonic Representation Hypothesis}
The Platonic Representation Hypothesis (PRH)
\cite{huh24platonic} posits that modern machine learning models learn surprisingly similar representations of data, even when trained with different architectures, datasets, modalities, and objectives. A central claim of the hypothesis is a striking trend: \emph{as models improve on downstream tasks, the representations they learn become increasingly similar} \cite{huh24platonic}. PRH is appealing because it hints at a universality in representation learning. Disparate models may be converging toward a shared, \emph{ground truth} (Platonic) representation of the world. The potential value of such representations is enormous. If learned features reliably captured what an object is and is not, they could provide a foundation for discrimination, detection, synthesis, causal reasoning, and generation. Put more boldly, if “ground truth” representations of objects exist at all, access to them would seem difficult to separate from general intelligence. From this perspective, alignment toward a Platonic representation is not merely an emergent curiosity but a learning objective in its own right.

\textbf{Puzzles About PRH.}
Pursuing this objective, however, requires a formal learning-theoretic framework that yields consistent, robust, and efficient algorithms for finding Platonic representations. To the best of our knowledge, the current PRH literature does not yet provide such a model. This gap persists in part due to several puzzles, central to the hypothesis:

\textbf{\textit{Question 1. Representations align, but toward what?}}
Empirical evidence \cite{huh24platonic,CIMAA25,SAC25,raugel2025disentanglingfactorsconvergencebrains,groger2026prharistotle} is already overwhelming for the convergence of representations of data between machine learning models (and even the human brain!) across modalities and architectures. However, there is no consensus yet on what ``Platonic object'' the representations converge to. 

\textbf{\emph{Question 2. How do model specifics bias representations?}}
Similarity of representations is shown to be strongly correlated with the performance of models on downstream tasks \cite{huh24platonic}. This is a compelling hypothesis that also carries a lot of philosophical weight -- greater understanding and capabilities (of the model) lead to greater alignment with the (Platonic) ground truth. Unfortunately, this trend of correlating performance and representation similarity is limited to the KNN overlap metric \cite{huh24platonic}. For other metrics such as CKA, the authors observe that ``alignment with CKA revealed a very
weak trend of alignment between models, even when comparing models within their own modality''. The recent work \cite{groger2026prharistotle} shows, in fact, that when appropriately controlling for width and depth, the trend in certain metrics, including CKA, altogether disappears. The \emph{inconsistency across similarity metrics} raises a central question: how do model-specific factors such as depth, width, and modality bias representation similarity?

\textbf{\textit{Question 3. How does inference data impact representation similarity?}}
Fully explaining the alignment of representations only through the models and their capacity is quite unlikely since a representation $f_\theta(X)$ is a combination of the trained model $f_\theta$ and the inference data $X.$ Missing from \cite{huh24platonic} and subsequent works is the basic question: what role does $X$ have? This role is, of course, non-trivial. In an extreme setting where $X$ is far out of distribution, we can hardly expect any non-trivial alignment. In fact, this idea is used to construct multimodal OOD detectors \cite{kim25ood}.   

\begin{figure}[htb!]
    \centering
    \begin{minipage}[t]{0.5\textwidth}
        \centering
  \centering
   \resizebox{.9\linewidth}{!}{\input{tikz/repdecomposition}}
    \caption{\small{Constituent Parts of Learned Representations. Most important to this work are the first and second levels. Further levels are omitted.}}
    \label{fig:repdecomposition}
    \end{minipage}
    \hfill
    \begin{minipage}[t]{0.42\textwidth}
        \centering
    \resizebox{.7\linewidth}{!}{\input{tikz/objectattribute}}
    \caption{\small{Sparse relationships between objects and attributes under LRH following an example in \cite{templeton2024scaling}. Encoding attribute-object relations in a bipartite graph is related to Formal Concept Analysis \cite{ganter2003formal,xiong2026lattice}.}
    }    
    \label{fig:objectattribute}
    \end{minipage}
    \hfill
\end{figure}

\subsection{Our Framework: Signal, Bias, and Noise in Representations}
\textbf{Methodology:}
Our aim is to put forward a statistical model of learned representations that elucidates Questions 1 - 3. To do so, we decompose data representations into their constituent parts (Figure~\ref{fig:repdecomposition}). This decomposition is our guide toward identifying \emph{signal, bias,} and \emph{noise} in representations.

\textbf{\emph{1. Signal: The Linear Representation Hypothesis.}} 
    Among $f, \theta, X,$ the only invariant between models is the inference data $X.$ But what about inference could be Platonic? We suggest that the Platonic signal is a rather old and well-known geometric structure of 
    learned representations, the Linear Representation Hypothesis (LRH). As stated in \cite{fel2025rabbithulltaskrelevantconcepts, garg2026featureslanguagemodelstore}, it posits that each representation is approximately a linear combination of representations of its constituent attributes (see Section~\ref{sec:lrhprior} for more background on LRH). More formally, representations linearly encode a ground truth sparse relationship between objects and their attributes as in Figure~\ref{fig:objectattribute}. 
    
    We confirm this hypothesis by training top-$k$ sparse autoencoders \cite{MakhzaniFrey2014kSparse} on features produced by a variety of different models. Our first contribution is to \textbf{\emph{demonstrate that sparse features are similar across models. In particular, the alignment between sparse features is typically higher than the alignment between the dense features from which they are extracted (Figure~\ref{fig:signalmain}).}}
    We propose that the Platonic signal is the sparse object-attribute structure posited by LRH: each object activates a small set of attributes, and learned representations encode these attributes approximately linearly. \textbf{\emph{This is the first link between the two hypotheses on representation geometries.}} 
    
\textbf{\emph{2. Bias: Architecture and Training.}} Unlike inference data -- which is common across representations -- architecture and weights are different. As a result, one may expect an \emph{approximation error} in the learned representations due to the concept class described by the architecture $f.$ Likewise, 
    there should be an \emph{optimization error} due to the training procedure resulting in weights $\theta.$ Hence, a statistical procedure for removing the bias of models should increase the alignment of representations. We confirm this hypothesis in Section~\ref{sec:biasexperiment} via a simple debiasing procedure that subtracts the population mean (over the entire inference dataset for the same model).  Our second main contribution is to \emph{\textbf{demonstrate that the simple debiasing procedure of subtracting the population mean and normalizing consistently increases alignment of representations.}} 

\begin{figure}[!htb]
    \centering
    \includegraphics[width = .9\linewidth]{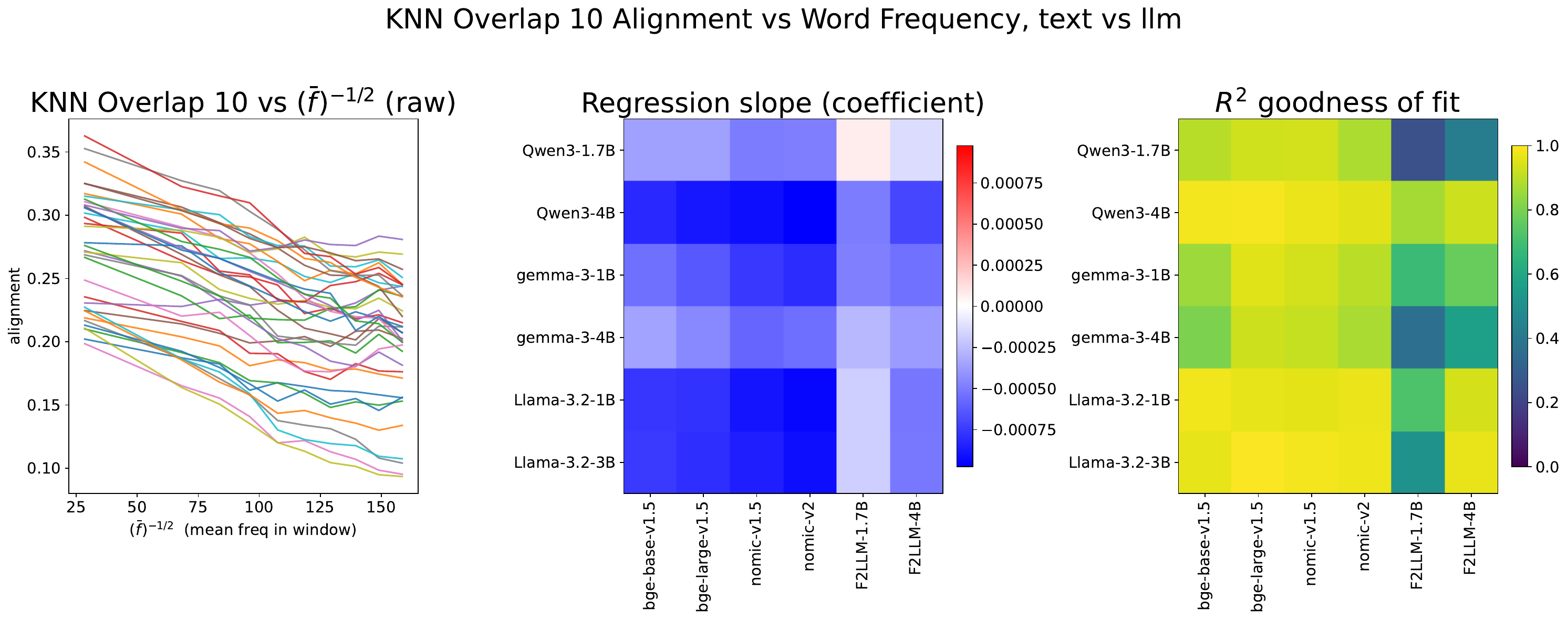}
    \caption{\small{Experiment on \textsf{wordfreq} dataset. Each line corresponds to the alignment of one pair of text embedding and LLM models. On the $y$-axis we have the alignment value and on the $x$-axis, the scaled relative frequency as $f^{-1/2}$ where $f$ is the frequency (so, more frequent words are to the left). }}
    \label{fig:noise_experiment_main}
    \vspace*{.1cm}
\end{figure}

\textbf{\emph{3. Noise: Frequency of Inference Data.}} To address Question 3 and go beyond the extreme setting of lack of alignment in far-OOD samples, we recall Statistics-101. More samples lead to lower estimation errors and, hence, better signal-to-noise ratio. This yields the natural hypothesis that alignment is greater for more frequent data points. We test this hypothesis over a dataset where data frequency can be reliably and accurately measured -- common English words. We \emph{\textbf{demonstrate the trend that alignment is greater for more frequent words between LLMs and text embedding models.}} Our experiments furthermore suggest a noise rate proportional to inverse square root of data frequency consistent with classical central limit theorem behavior of estimators, see Figure~\ref{fig:noise_experiment_main}.

\subsection{Our Statistical Model of Representations and Implications}
\label{sec:statmodel}

Having identified the different components -- signal, bias, and noise -- in learned representations, we can write down a statistical model as follows. It refines the Linear Representation Hypothesis by explicitly attributing different components of the model to different components of the representations:
\begin{equation}
\label{eq:statmodel}
\tag{Statistical Model of Representations}
f_\theta(X)  = 
A(\theta, f) \times ( Z(X) \odot M(X,\theta, f))  + \eta_{X,\theta, f}.
\end{equation}
The different components are:

\emph{Platonic Signal: Sparse feature vector} $Z(X)\in \{0,1\}^{m}$ which is a function purely of the inference data $X.$ It encodes which attributes are relevant to object $X.$

\emph{Bias 1: Signal Magnitude} $M(X,\theta, f) \in \mathbb{R}_{\ge 0}^m$ which encodes the importance of each attribute for the object $X$ in the specific training dataset $\mathcal{D},$ learning objective of $f,$ or architecture capability.

\emph{Bias 2: Dictionary of Attributes} $A(\theta,f)\in \mathbb{R}^{d\times m}$ which depends on the architecture and training.

\emph{Estimation Error} $\eta_{X, \theta, f}$ which is determined by how well $X$ is learned during training $\theta$ with model $f.$ Model architecture is relevant, for instance due to its capacity. Training and weights are relevant as they determine how data is utilized (e.g., learning rate schedule).

\begin{wrapfigure}{r}{.4\textwidth} 
\vspace*{-.3cm}
      \includegraphics[width =\linewidth]{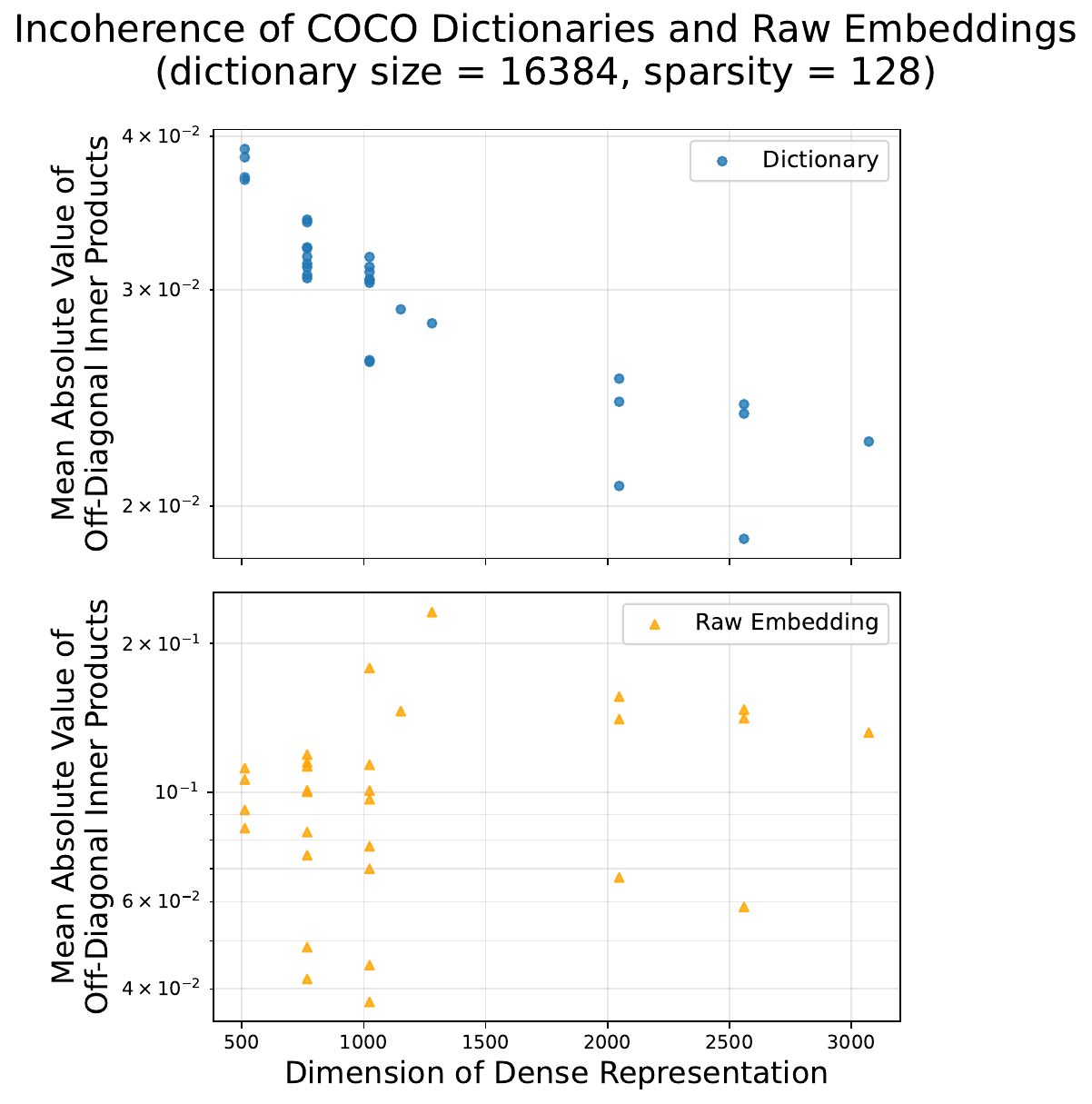}
    \caption{\small{Higher dimensions consistently lead to smaller in magnitude inner products of dictionary elements for the 30 models from Table~\ref{table:modelspecs_architecture}. However, this is not true for raw embeddings. In Appendix~\ref{appendix:dimensionandincoherence}, we 
    test other datasets, sparsities, and dimensionalities.}}
    \label{fig:mainincoherence}
    \vspace*{-.4cm}
\end{wrapfigure}

\textbf{Assumptions on Representation Components.} Our \ref{eq:statmodel} predicts several phenomena related to PRH observed in the literature under common assumptions on the different components of dictionary learning (see, for example, \cite{arora2015simple}). 
\begin{enumerate}
    \item [P.1)] \emph{Data Normalization: } $\|f_\theta(X)\|_2 =1.$ This property is a non-restrictive assumption as far as cosine similarities are concerned since we can always add a normalization layer. 
    \item [A.1)] \emph{Sparsity: }Each $Z(X)\in \mathbb{R}^m$ has sparsity $k = o(m).$ This common assumption can be interpreted as the fact that each object has a limited number of attributes. The small residuals of SAEs when fitting sparse autoencoders support A.1), see Appendix~\ref{appendix:saeresiduals}.
    \item [A.2)] \emph{Dictionary Incoherence: }For every $i \in [m],$ the feature $A(\theta, f)_i$ has unit norm, $\|A(\theta, f)_i\|_2 = 1,$ and for every $1\le i < j  \le m,$ $|\langle A(\theta, f)_i, A(\theta, f)_j\rangle|\le \epsilon_{\mathrm{dict}} = o(1/k).$ This assumption is common in theoretical works on sparse autoencoders such as \cite{arora2015simple,arora2014new}. It is also closely related to the proposition that ``causally separable'' concepts are represented as orthogonal vectors \cite{park24LRH}. When $m> d,$ an orthogonal frame does not exist, so the best one can do is a near-orthogonal frame with small $\epsilon_{\mathrm{dict}}.$ See Figure~\ref{fig:mainincoherence}, Appendix~\ref{appendix:dimensionandincoherence} for empirical support of the small incoherence of SAE dictionaries.
    \item [A.3)] \emph{Estimation with Non-Trivial Signal}. Non-zero coordinates are well-expressed. Namely, for some $0<\phi_{\text{sig}}<\Phi_{\text{sig}}$ and for any $X,i$ such that $Z(X)_i =1,$ it holds that $\frac{\phi_{\text{sig}}}{\sqrt{k}}\le M(X, \theta, f)_i\le \frac{\Phi_{\text{sig}}}{\sqrt{k}}.$ This assumption is common in theoretical works, see \cite[Assumption (2) of model]{arora2015simple}. See also Appendix~\ref{appendix:magntiduesofsparsefeaturs} for empirical support.
    \item [MA.1)] \emph{Estimation with Non-Trivial Noise}. With high probability over $\eta_{X, \theta, f},$ it holds that $\|\eta_{X, \theta, f}\|_2\le \epsilon_{\mathrm{noise}}$ for some $\epsilon_{\mathrm{noise}}\le 1.$ 
   This is our main assumption. It effectively states that the Linear Representation Hypothesis holds  \cite{arora2015simple,arora2014new,garg2026featureslanguagemodelstore}. 
\end{enumerate}

\textbf{Representation Similarity.} Let $X_1,X_2$ be two different objects and let $f_\theta$ be a model. For brevity, denote 
$A=  A(\theta, f), M_i =  M(X_i,\theta, f), Z_i = Z(X_i), \eta_i = \eta_{X_i,\theta, f}.$ Under the assumptions above:
\begin{equation*}
    \begin{split}
        & \langle f_{\theta}(X_1), f_{\theta}(X_2)\rangle\\
        & = \langle A(\theta, f) \times ( Z(X_1) \odot M(X_1,\theta, f))  + \eta_{X_1,\theta, f}, 
        A(\theta, f) \times ( Z(X_2) \odot M(X_2,\theta, f))  + \eta_{X_2,\theta, f}\rangle\\
        & \langle A  ( Z_1 \odot M_1), A  ( Z_2 \odot M_2)\rangle + 
        \langle f_{\theta}(X_1),  \eta_2\rangle +
        \langle  f_{\theta}(X_2),  \eta_1\rangle -
        \langle  \eta_1,  \eta_2\rangle\\
        & = 
        \langle \sum_{i \; : \; (Z_1)_i = 1} A_i (M_1)_i, 
        \sum_{i \; : \; (Z_2)_i = 1} A_i (M_2)_i\rangle + 
        \langle f_{\theta}(X_1),  \eta_2\rangle +
        \langle  f_{\theta}(X_2),  \eta_1\rangle -
        \langle  \eta_1,  \eta_2\rangle\\
    \end{split}
\end{equation*}
Recalling the assumptions, we rewrite as follows (the bounds are derived in Appendix~\ref{appendix:ComponentBounds}): 
\begin{equation}
\label{eq:innerproductfinalform}
\begin{aligned} 
& \sum_{i\; : \; (Z_1)_i = (Z_2)_i = 1} (M_1)_i(M_2)_i
&& \rhsbox{\text{Signal }\; \in \Big[\frac{\phi_{\mathrm{sig}}^2\langle Z_1, Z_2\rangle}{k},
								 \frac{\Phi_{\mathrm{sig}}^2\langle Z_1, Z_2\rangle}{k} \Big]}
\\
& + \sum_{i\neq j\; : \; (Z_1)_i = 1, (Z_2)_j = 1 } (M_1)_i(M_2)_j\langle A_i, A_j\rangle
&& \rhsbox{\text{Bias }\;\le k\epsilon_{\mathrm{dict}}\Phi^2_{\mathrm{sig}}}
\\
& + \langle f_{\theta}(X_1),  \eta_2\rangle +
    \langle f_{\theta}(X_2),  \eta_1\rangle -
    \langle \eta_1,  \eta_2\rangle
&& \rhsbox{\text{Noise }\; \le 3\epsilon_{\mathrm{noise}}}
\end{aligned}
\end{equation}

\textbf{Aristotelian View: Sparse Signal and The KNN overlap metric.} The original paper \cite{huh24platonic} suggests a strong positive correlation between model performance and alignment in the KNN overlap metric for small $k.$ The work \cite{groger2026prharistotle} further suggests that other metrics such as CKA and SVCCA suffer from biases arising due to depth and width of models. \cite{groger2026prharistotle} goes further to propose that Platonic is only the local neighborhood structure, but not the global geometry.

Our~\ref{eq:statmodel} explains why for strong models, similarity of local neighborhoods arises while global geometry may differ. We give a full proof in Appendix~\ref{appendix:OnTheStatisticalModel}.

\begin{proposition}
\label{prop:aristotlethm}
Suppose that $k\epsilon_{\mathrm{dict}}\Phi^2_{\textrm{sig}} + 3\epsilon_{\mathrm{noise}}\le 1.$ Then, for any $\gamma>0,$
\end{proposition}
\begin{enumerate}
\item If $\langle Z(X_1), Z(X_2)\rangle\ge \frac{2k}{\phi^2_{\mathrm{sig}}}(k\epsilon_{\mathrm{dict}}\Phi^2_{\textrm{sig}} + 3\epsilon_{\mathrm{noise}} + \gamma/2),$ 
then 
$\langle f_{\theta}(X_1), f_{\theta}(X_2)\rangle\ge k\epsilon_{\mathrm{dict}}\Phi^2_{\textrm{sig}} + 3\epsilon_{\mathrm{noise}} + \gamma.$
\item If $\langle f_{\theta}(X_1), f_{\theta}(X_2)\rangle \ge k\epsilon_{\mathrm{dict}}\Phi^2_{\textrm{sig}} + 3\epsilon_{\mathrm{noise}} + \gamma,$ 
then $\langle Z(X_1), Z(X_2)\rangle\ge\frac{k}{\Phi^2_{\mathrm{sig}}}\gamma.$
\end{enumerate}

In particular, for strong models (such that bias and noise are simultaneously small, $k\epsilon_{\mathrm{dict}}\Phi^2_{\textrm{sig}} + 3\epsilon_{\mathrm{noise}}\le 1$), the local neighborhood of $f_\theta(X_1)$ is nearly determined as the vectors $f_\theta(X_2)$ for which the number of shared attributes -- $\langle Z(X_1), Z(X_2)\rangle$ -- exceeds a certain threshold.

While the local neighborhoods are shared between representations, global geometry might be lost due to differences in magnitudes $M_i(X, \theta, f),$ dictionaries $A(\theta, f),$ and 
random noise terms $\eta_{X, \theta, f}$. 

\textbf{Bias and Similarity of Training Objectives.} \cite{CLMDKM24} demonstrates that similarity of training objectives leads to increased alignment. This view is consistent with our~\ref{eq:statmodel}. One may expect that distinct features are important for different objectives. As the sparse-feature weights $M(X, \theta, f)$ and the dictionary $A(\theta, f)$ depend on the objective, models with the same objective are more aligned.    

\textbf{Bias and Model Width.} It has been empirically observed that models in higher dimension tend to have higher alignment in metrics such as CKA \cite{groger2026prharistotle}. The work \cite{groger2026prharistotle} explains this via the fact that in high dimension, even two independent matrices, each with independent entries, have high CKA alignment since the spectrum of the corresponding Gram matrices converges weakly to a limiting measure \cite{wachter1978strong}. While illuminating, the intuition from independent-entry ensembles may not carry over to representations in modern AI systems. The latter have a highly non-random structure that allows for geometric properties such as LRH (see Section~\ref{sec:lrhprior}) or success in 
multimodal retrieval (see \cite{bangachev2025globalminimizerssigmoidcontrastive}).  

Our~\ref{eq:statmodel} explains the positive correlation between alignment and dimension even when correlations between the matrix entries due to common features of objects exist. Higher dimensionality enables better sphere packings of feature representations in the dictionary, thus driving $\epsilon_{\mathrm{dict}}$ down (see Figure~\ref{fig:mainincoherence} for evidence that this happens in practice). A lower value of $\epsilon_{\mathrm{dict}}$ in turn reduces the bias in  
\eqref{eq:innerproductfinalform} and makes representations more similar. Figure~\ref{fig:mainincoherence} also shows that the incoherence of raw representations does not decrease with dimension. This refutes an alternative explanation in the style of \cite{groger2026prharistotle} that the embeddings themselves are closer to orthogonal.

\textbf{Noise and Strong Models.} Stronger models are typically trained on larger or more carefully filtered datasets in order to decrease noise. 
Again, this leads to a smaller deviation of  $\langle f_{\theta}(X_1), f_{\theta}(X_2)\rangle$ from the Platonic signal $\frac{\langle Z(X_1), Z(X_2)\rangle}{k},$ which explains the trend between alignment and model performance in \cite{huh24platonic}. Our experiment in Figure~\ref{fig:regressionmain} additionally suggests that when both models have been trained on more data, alignment between them is greater. 
Likewise, 
the relationship between noise and strong models elucidates the observation that ``linear representations of truth emerge with scale'' from \cite{marksgeometry}. Representations from stronger models are less noisy, so are closer to perfectly satisfying the linear relationship between objects and attributes, i.e. $\eta_{X, \theta, f} = 0.$

%% file: tikz/repdecomposition.tex
  \resizebox{\linewidth}{!}{
\begin{tikzpicture}[
  box/.style={
    draw,
    rounded corners,
    align=center,
    inner sep=6pt,
    minimum width=3.2cm,
    minimum height=1.2cm
  },
  arrow/.style={-Latex, thick},
  node distance=12mm and 14mm
]

\node[box] (A1) {Representation\\$f_\theta(X)$};

\node[box, below left=of A1] (B1) {Architecture\\$f$};
\node[box, below=of A1]      (B2) {Training\\ $\theta$};
\node[box, below right=of A1](B3) {Inference Data\\$X$};

\draw[arrow] (B1) -- (A1);
\draw[arrow] (B2) -- (A1);
\draw[arrow] (B3) -- (A1);

\node[box, below=of B1, xshift=1.3cm
] (C0) {Aspect Ratio};
\node[box, below=of B1, xshift=-2.6cm] (C1) {Model size};
\node[box, below=of B1, yshift= -1.25cm, xshift = -.9cm] (C2) {Representation space};
\node[box, below=of B2]                (C3) {Training Dataset};
\node[box, below=of B2, yshift=-1.25cm, xshift = 2.9cm] (C4) {Training Dynamics};
\node[box, below=of B3] (C5) {Training Objective};

\draw[arrow] (C0) -- (B1);
\draw[arrow] (C1) -- (B1);
\draw[arrow] (C1) -- (B1);
\draw[arrow] (C2) -- (B1);
\draw[arrow] (C3) -- (B2);
\draw[arrow] (C4) -- (B2);
\draw[arrow] (C5) -- (B2);

\node[box, below=of C3, yshift = 0cm] (D1) {Modality};
\node[box, below=of C5] (D2) {Attributes};
\draw[arrow] (D1) -- (C3);
\draw[arrow] (D1) -- (B1);
\draw[arrow] (D2) -- (B3);
\end{tikzpicture}}

%% file: tikz/objectattribute.tex
\begin{tikzpicture}[x=1cm,y=1cm]

\fill[white] (-0.2,4) rectangle (16.2,16.5);

\definecolor{objorange}{RGB}{233,113,50}
\definecolor{attrgreen}{RGB}{78,167,46}
\definecolor{relblue}{RGB}{23,102,138}

\draw[objorange, line width=0.6pt] (0,4) rectangle (5.5,16);

\fill[objorange] (0,14) rectangle (5.5,16);
\node[white, font=\bfseries\fontsize{36}{38}\selectfont] at (2.75,15) {Objects};

\fill[objorange] (0,12) rectangle (5.5,14);
\fill[objorange] (0,10) rectangle (5.5,12);
\fill[objorange] (0,8) rectangle (5.5,10);
\fill[objorange] (0,6) rectangle (5.5,8);
\fill[objorange] (0,4) rectangle (5.5,6);

\draw[white, line width=1pt] (0,12) -- (5.5,12);
\draw[white, line width=1pt] (0,10) -- (5.5,10);
\draw[white, line width=1pt] (0,8) -- (5.5,8);
\draw[white, line width=1pt] (0,6) -- (5.5,6);

\node[anchor=west, white, font=\fontsize{22}{24}\selectfont] at (0.28,13.35) {Golden};
\node[anchor=west, white, font=\fontsize{22}{24}\selectfont] at (0.28,11.35) {Gate};
\node[anchor=west, white, font=\fontsize{22}{24}\selectfont] at (0.28,9.35) {Bridge};
\node[anchor=west, white, font=\fontsize{22}{24}\selectfont] at (0.28,7.35) {Chapel};
\node[anchor=west, white, font=\fontsize{22}{24}\selectfont] at (0.28,5.35) {Tunnel};

\draw[attrgreen, line width=0.6pt] (10.3,8.05) rectangle (15.7,16);

\fill[attrgreen] (10.3,14) rectangle (15.7,16);
\node[white, font=\bfseries\fontsize{36}{38}\selectfont] at (13,15) {Attributes};

\fill[attrgreen] (10.3,12.05) rectangle (15.7,14);
\fill[attrgreen] (10.3,10.05) rectangle (15.7,12.05);
\fill[attrgreen] (10.3,8.05) rectangle (15.7,10.05);

\draw[white, line width=1pt] (10.3,12.05) -- (15.7,12.05);
\draw[white, line width=1pt] (10.3,10.05) -- (15.7,10.05);

\node[anchor=west, align=left, white, font=\fontsize{20}{22}\selectfont] at (10.55,13.28)
    {Related to\\Golden Gate};
\node[anchor=west, align=left, white, font=\fontsize{20}{22}\selectfont] at (10.55,11.22)
    {Popular tourist\\attraction};
\node[anchor=west, align=left, white, font=\fontsize{20}{22}\selectfont] at (10.55,9.2)
    {Transit\\Infrastructure};

\fill[relblue] (5.5,13.25) rectangle (10.3,16);
\node[white, align=center, font=\fontsize{34}{36}\selectfont] at (7.9,14.525)
    {Sparse\\Relations};

\draw[relblue, line width=1.2mm]
    (5.5,13.05) -- (10.3,13.05);

\draw[relblue, line width=1.2mm]
    (5.5,11.0) -- (10.3,13.05);

\draw[relblue, line width=1.2mm]
    (5.5,8.95) -- (10.3,13.05);

\draw[relblue, line width=1.2mm]
    (5.5,8.95) -- (10.3,9.0);

\draw[relblue, line width=1.2mm]
    (5.5,7.05) -- (10.3,11.0);

\draw[relblue, line width=1.2mm]
    (5.5,5.0) -- (10.3,9.0);

\end{tikzpicture}

%% file: sections/Prelim.tex
\section{Prior Work and Preliminaries}
Modern AI systems represent various forms of data as vectors, either explicitly in representation models such as CLIP, DINO, or SigLIP, or implicitly via neuron activations in intermediate layers. We will abstract all of these as real-valued vectors $f_\theta(X)$ where $f$ is the model architecture, $\theta$ the trained weights, and $X$ the data to be represented. We scale all features to be unit norm.

\subsection{Platonic Representation Hypothesis}
\label{sec:prhprior}
Since the publication of \cite{huh24platonic}, there has been extensive interest in the PRH including alignment beyond image-text models \cite{CIMAA25,WLDGW25,murugaboopathy2025platonicpovertymapping,gupta2026canonicalizingmultimodalcontrastiverepresentation}, a learning-theoretic formalization \cite{insulla2025towards},
applications 
\cite{GWSJ25,subramaniam2024training,groger2025structureguidance,lin2024dreaming,YYK25jam}, possible explanations 
\cite{li2025exploring,ziyin2025proof,lobashev2025information}, refinements \cite{groger2026prharistotle}, and even some objections \cite{thomas2025towards}.

\textbf{Comparing Representations.} To make PRH formal, we need a formal notion of representation similarity. Let $(X_i, Y_i)_{i = 1}^N$ be a dataset of paired objects. Using two models $f_\theta, g_\phi,$ form the representations $F = (f_\theta(X_1), \ldots, f_\theta(X_N))\in \mathbb{R}^{d_1\times N}$, $G = (g_\phi(Y_1), \ldots, g_\phi(Y_N))\in \mathbb{R}^{d_2\times N}.$ The comparison is over the two Gramians $F^TF, G^TG\in \mathbb{R}^{N\times N}$ which contain the cosine similarities of unit-norm representations. The metrics that are used to compare $F^TF, G^TG$ in \cite{huh24platonic} and related works fall into two categories. \emph{Cardinal:} Important is the magnitude of distances; examples include CKA, Unbiased CKA, and SVCCA. \emph{Ordinal:} Important is the relative order of distances; examples include overlap and edit distance of KNN neighborhoods. See \cite{kriegeskorte2008representational,raghu2017svcca,kornblith2019similarity,huh24platonic,insulla2025towards,groger2026prharistotle} for definitions and interpretations of the metrics.

\subsection{Linear Representation Hypothesis}
\label{sec:lrhprior}
LRH predicts the existence of sparse linear relations between data representations.
Early works on the Linear Representation Hypothesis analyze relationships of the form $f_\theta(\textsf{man}) - f_\theta(\textsf{woman}) \approx f_\theta(\textsf{king}) - f_\theta(\textsf{queen}),$ demonstrating linearity of concepts (\textsf{man-to-woman} in the specific case) in models such as Word2Vec \cite{mikolov2013distributed,mikolov2013efficientestimationwordrepresentations,mikolov2013linguistic} and GloVe \cite{pennington2014glove}.

\textbf{Sparse Dictionary Learning Formulation.}
Recent work on sparse linear relations between data representations has focused on a different but related formulation of the Linear Representation Hypothesis -- each representation is a superposition of the representations of different attributes \cite{park24LRH,fel2025rabbithulltaskrelevantconcepts,garg2026featureslanguagemodelstore,venhoff2025visual}. This view can be formalized in the framework of sparse dictionary learning, where each feature $f_\theta(X)$ is a sparse linear combination of some representations of attributes $(D_a)_{a\in \mathcal{A}}$ for a set of global attributes $\mathcal{A}.$ Namely, $f_\theta(X) = \sum_{b \in B}c_bD_b$ where $B$ is a set of relevant attributes and the $c_b$ are \emph{non-negative} coefficients. This formulation bridges LRH with the literature on mechanistic interpretability, where sparse autoencoders are trained on top of dense representations and sparse interpretable features are extracted \cite{bricken2023towards,templeton2024scaling,openai2024scaling,tang2025theoretical,Huben2024SparseAutoencoders,pach2025sparse,xu25multimodalsparse}. 

\subsection{Comparison with Prior Work}

\textbf{Learning-Theoretic Perspective.} \cite{insulla2025towards} also views representation alignment as a learning-theoretic task. They propose that there is a common reality $\Xi$ and different models are different observations of it. In our work, we make both the common reality and observation models more explicit. First, we postulate that the reality $\Xi$ is a sparse relationship between features and attributes. Second, we make our model fit into a standard formulation with signal, bias, and noise. The concreteness of our approach opens the door to empirical predictions which we verify experimentally. While illuminating, the abstract formulation of \cite{insulla2025towards} does not yield any concrete experimental predictions.

\textbf{Alignment emerging from training.} \cite{ziyin2025proof} shows that representation similarity arises via the inductive bias of (stochastic) gradient descent on linear networks. While training dynamics certainly play a role (note that training $\theta$ is part of our~\ref{eq:statmodel}), the linear set-up is perhaps too simplistic to explain the alignment of features of practical models on real datasets. For instance, one byproduct of their main theorem is that the features at each layer of the network are the same up to a rotation and scalar multiplication. In particular, the trained network is equivalent to a single-layer network and the complexities of depth and architecture are not captured. \cite{guth24rainbow} takes a related approach and explains PRH via the assumption that the weights of trained models are independent samples from the same distribution.  

\textbf{Alignment emerging from a shared PMI matrix.} \cite{huh24platonic} outlines that one possible cause of the Platonic Representation Hypothesis is that the Gram matrices of representations converge to the same Pointwise Mutual Information (PMI) matrix $\PMI$. Convergence to the $\PMI$ matrix has also been used to explain the Linear Representation Hypothesis in the form $f_\theta(\textsf{man}) - f_\theta(\textsf{woman}) \approx f_\theta(\textsf{king}) - f_\theta(\textsf{queen})$ \cite{levy2014sgns,park24LRH,arora2016latent,arora19classification,gupta2026canonicalizingmultimodalcontrastiverepresentation,jiang2024origins,korchinski2025emergence}. Concretely, the PMI matrix is defined as follows. The entry corresponding to the pair of data points $(X_1,X_2)$ is given by $\PMI(X_1,X_2) = \log\frac{P(X_1, X_2)}{P(X_1)P(X_2)}.$ Here, $X_1, X_2$ are two data points, corresponding to object-concept or object-object pairs. The key identity is that when a fully unconstrained representation model $f_\theta$ is minimized using the InfoNCE or sigmoid contrastive objectives, it holds that $\langle f_\theta(X_1), f_\theta(X_2)\rangle = \PMI(X_1,X_2) + C$ where $C$ is some constant independent of $X_1, X_2$ (in the multimodal or object-context setting, $\langle f^1_\theta(X_1), f^2_\theta(X_2)\rangle = \PMI(X_1,X_2) + C$ for different encoders $f^1, f^2$).  

There are, however, several challenges in explaining PRH as convergence to the $\PMI$ matrix: \emph{1. Distributional:}  
The PMI matrix may not be similar across training datasets and modalities.
\emph{2. Geometric:} In practice, representations are low-dimensional ($d\le 4096$) while the $\PMI$ matrix captures  co-occurrences between billions of data-points. Thus, for $\langle f^1_\theta(X_1), f^2_\theta(X_2)\rangle = \PMI(X_1,X_2) + C$ to hold even approximately, $\PMI$ has to be approximately low-rank. We don't know of any evidence for this. In the object-object setting used to explain PRH in \cite{huh24platonic}, $\langle f_\theta(X_1), f_\theta(X_2)\rangle = \PMI(X_1,X_2) + C$ can only hold if $\PMI$ is positive semi-definite in the space orthogonal to the all-ones vector, which is again unjustified. \emph{3. Optimization-Related:} In practice, the (inverse) temperature parameter is also trainable for InfoNCE-based models like CLIP \cite{radford21clip} and sigmoid-loss-based models like SigLIP \cite{zhai23siglip}. When the temperature parameter is trainable rather than fixed, the geometry of global minimizers changes and does not correspond any longer to $\langle f^1_\theta(X_1), f^2_\theta(X_2)\rangle = \PMI(X_1,X_2) + C,$ see \cite{bangachev2025globalminimizerssigmoidcontrastive}. \emph{4. Generality:} While the PMI matrix has been shown to arise in certain contrastive learners, it is less clear what the relevance of the PMI matrix is to other models such as LLMs and image foundation models.

%% file: sections/Experiments.tex
\section{Experiments}
\label{sec:experiments}
\textbf{Metrics.} We focus on the KNN overlap metric with a small value of $k = 10$ as in \cite{huh24platonic}. This metric captures similarity of local neighborhoods of representations. The recent work \cite{groger2026prharistotle} suggests that other spectral and geometric metrics suffer from undesirable biases. This is especially important for us when considering sparse features produced by SAEs since they only have non-negative coordinates, which leads, for instance, to the strong bias that all representations belong to the same orthant. Nevertheless, in Appendix~\ref{appendix:experimentaldetails}, we reproduce all experiments with other metrics including CKA, Unbiased CKA, KNN overlap with $k = 100,$ 
and KNN edit distance with $k = 10, 100.$

\begin{wrapfigure}{r}{.5\textwidth}
    \vspace*{-.3cm}
    \includegraphics[width=\linewidth]{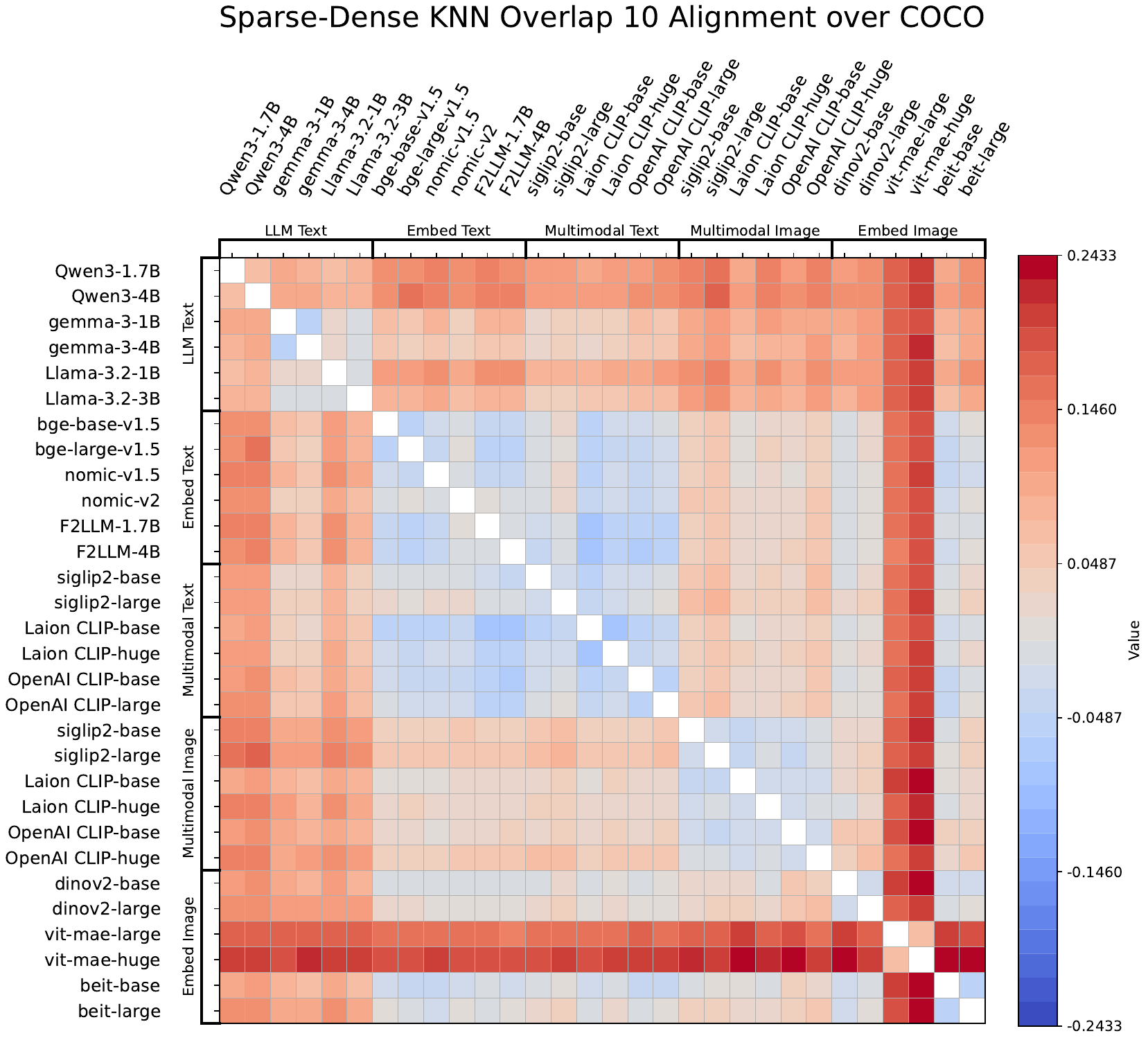}
    \caption{\small{The difference of model alignment in the KNN-10 metric between sparse features and dense features over COCO. Alignment is higher (red) with few weak exceptions, which are predominantly for models over the same modality. 
    }} 
    \label{fig:signalmain}
\end{wrapfigure}

\textbf{Models.} We perform experiments with 2 models from 3 families from each of the following 4 categories: Text Embedding, Image Embedding, Multimodal Contrastive, and LLMs. This leads to 30 different representations of each object (the text and image representations for each multimodal model are different). The models as well as their characteristics are given in Appendix~\ref{appendix:featuresofmodels}. Unlike in text, image, and multimodal embedding models, LLMs do not explicitly produce text embeddings. Instead, on a text input of $T$ tokens, we find the $T\times D$ state of the residual stream of the middle layer and mean-pool it into a vector of dimension $D,$ which we then normalize to have unit norm. While not necessary, the choices of mean-pooling and mid-layer are consistent with prior mechanistic interpretability literature such as \cite{templeton2024scaling,kelly2020sentence}. By choosing a concrete layer, we avoid the confounder of depth arising from optimizing over layers \cite{groger2026prharistotle}.

\textbf{Datasets.} The experiments in Sections~\ref{sec:signalexperiment} and \ref{sec:biasexperiment} are over COCO \cite{cocodataset}. In Appendix~\ref{appendix:experimentaldetails}, we reproduce the experiments over CC3M \cite{sharma2018cc3m} and Visual Genome
\cite{krishna2017visual}. The experiments in Section~\ref{sec:noiseexperiment} are over English words together with their frequencies in the \textsf{wordfreq} library \cite{robyn_speer_2022_7199437}.

\subsection{Signal: Sparse Features Increase Similarity}
\label{sec:signalexperiment}

We show that the alignment between sparse features is typically higher than the alignment between the dense features from which they are extracted. 
This trend is especially prevalent for the alignment between two models with different objectives and modalities.


\textbf{Sparse Features.} To derive the sparse features, for each model $f_\theta$ resulting in an embedding matrix $F\in \mathbb{R}^{d\times N}$ we train top-$k$ sparse autoencoders on $F$ \cite{MakhzaniFrey2014kSparse}. The concrete parameters we choose are $D = 16384$ and $k = 32$ for text models, $k = 64$ for multimodal models, $k = 128$ for image models (we use bigger $k$ for models that more strongly rely on image data due to the intuitive fact that images contain more information than single-sentence captions). As is common, we additionally resample dead neurons (at every 2500 steps) to mitigate the issue of dead neurons \cite{bricken2023towards}.  Finally, we remove features that activate on more than $10\%$ of the data points since they are unlikely to be monosemantic and on less than $0.001\%$ since they are likely noise. We trained on a single H200 GPU. Each run took up to 1 hour per model per choice of dimension and sparsity. 

\textbf{Experiment 1: Alignment with Sparse Features.} We averaged the KNN overlap similarity metric over 10 different uniformly random samples of size 1000 for the dense and sparse models.

\textbf{Analysis.} Alignment in the KNN-10 metric is higher for sparse features, especially in the case of models with different modalities. We explain this via the fact that dictionaries $A(\theta, f)$ which depend on model architecture and training (in particular, modality of training dataset) should have a similar bias for models with similar architecture and training. Thus, the fact that alignment is sometimes lower between multimodal image models when sparse features instead of dense ones are considered does not contradict our~\ref{eq:statmodel}. 

We note the special case of the ViT-MAE models. In the original PRH paper~\cite{huh24platonic}, the alignment of the MAE (Figure 9 in the arXiv version) with other models is surprisingly low compared to the alignment between other models. This is also true in our experiments as far as dense features are concerned. The low alignment of MAE models is largely mitigated when passing to sparse features.

\textbf{Experiment 2: Correlation of Sparse Features.}
We additionally test the null hypothesis that sparse features between different models are independent. Concretely, suppose that the sparse features from two models $f^1_{\theta^1}$ and $f^2_{\theta^2}$ over paired data $(X_i,Y_i)_{i=1}^N$ are $Z_1, Z_2 \in \mathbb{R}_{\ge 0}^{N\times D}.$ Since the sparse features are undefined up to a global permutation, we compute the correlation between $Z_1, Z_2$ as $\corr(Z_1, Z_2) \coloneqq \max_{\Pi \in \mathcal{S}_{D}} \frac{\langle Z_1, Z_2\Pi\rangle}{\|Z_1\|_2\times \|Z_2\|_2}.$ This optimization problem over permutations is tractable since it is simply a maximal matching problem in a $D\times D$ weighted graph with weights $Z_1^TZ_2$.  
Following the methodology of \cite{groger2026prharistotle}, we compare against a baseline of randomly permuting the objects,
$\corr_\Phi(Z_1, Z_2)\coloneqq \max_{\Pi \in \mathcal{S}_{D}} \frac{\langle \Phi Z_1, Z_2\Pi\rangle}{\|Z_1\|_2\times \|Z_2\|_2}$ for $\Phi$ uniformly drawn from $\mathcal{S}_N.$

\textbf{Analysis.} The difference between $\corr(Z_1, Z_2)$ and the empirical mean of $\corr_\Phi(Z_1,Z_2)$ is on the order of thousands of standard deviations of $\corr_\Phi(Z_1,Z_2),$ see Appendix~\ref{appendix:signalcorrelation}. This strongly rejects the null that $\corr(Z_1, Z_2)$ and the permutation null 
$\corr_\Phi(Z_1,Z_2)$ come from the same distribution.

\subsection{Bias}
\label{sec:biasexperiment}

\begin{wrapfigure}{l}{.5\textwidth}
\vspace*{-.3cm}
    \includegraphics[width=\linewidth]{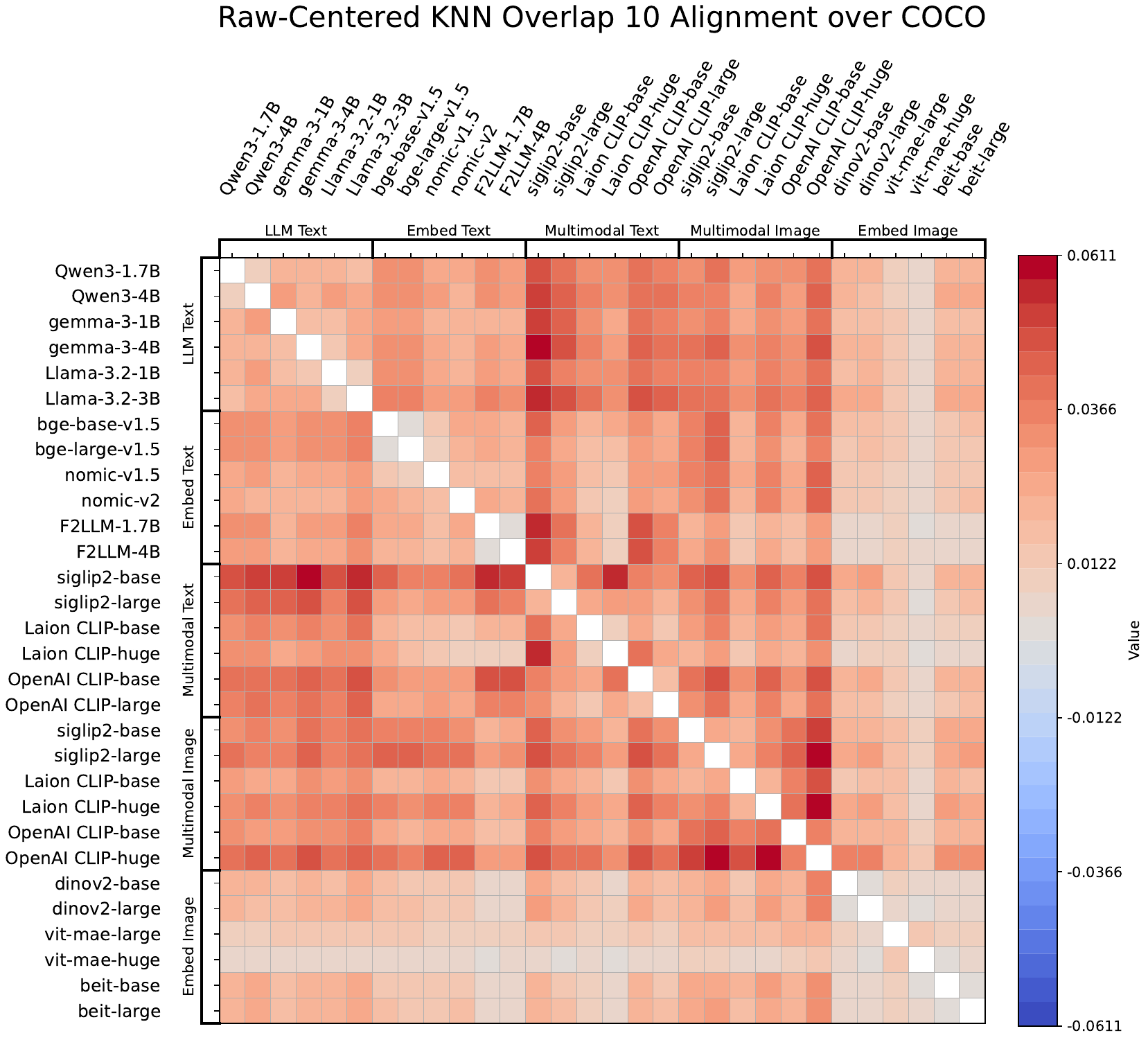}
    \caption{\small{Effect of centering on alignment.}}
    \vspace*{-.4cm}
    \label{fig:biasmain}
\end{wrapfigure}

\textbf{Experiment.} We  work with the same models and the COCO dataset, averaging over 10 random subsamples of size 1000. In Figure~\ref{fig:biasmain} we plot the difference of alignment for ``raw'' versus ``centered and re-normalized'' features. 

\textbf{Analysis.}
Centered features are more aligned across all pairs of models. We show that the trend is consistent across datasets and metrics in Appendix~\ref{appendix:biasexperiments}. That centered representations are closer to the Platonic signal is also consistent with the observation that centering improves representation strength \cite{mu2017all}.

We clarify that we do not claim that model bias is exactly the population mean. Rather, mean subtraction may remove a dominant first-order anisotropic component, while higher-order biases may remain.

\subsection{Noise: Signal-to-Noise Ratio}
\label{sec:noiseexperiment}
In our experiment demonstrating the importance of data frequency for alignment, we use text models (LLMs and text-embedding) to embed the most frequent English words paired with their relative frequencies from the \textsf{wordfreq} library \cite{robyn_speer_2022_7199437}. The utility of using words only is that for single words, there is a very simple, clear, and verifiable notion of what ``data frequency'' is. 

\textbf{Experimental Details.} We measure the alignment of models under the chosen metrics by sliding a window of size 500 with step size 250 from more common to less common English words.

\textbf{Analysis.} The leftmost plot in Figure~\ref{fig:noise_experiment_main} shows a consistent trend for higher alignment over windows with higher average frequency. This trend supports the work \cite{schick2020rare} which establishes that attention-based text embedding models have poor performance on rare words. From our statistical perspective, this is caused by the high noise level in the representations. We furthermore fit a 1-dimensional linear regression for $f^{-1/2}$ versus alignment. We consistently get a good fit (rightmost plot) and negative coefficient (middle plot). The strong linear relationship between $f^{-1/2}$ and alignment is consistent with standard CLT-like estimation error proportional to inverse square root of samples. 

We carry out the same experiment for different pairs of models and metrics in Appendix~\ref{appendix:noise}. We also perform two ablation studies to demonstrate that the negative linear trend between $f^{-1/2}$ and alignment does not arise due to plausible confounders. Concretely, we address the confounders of number of tokens, by controlling for only single-token words for all models, and word type, by controlling for a single part of speech 
(adjective, verb, noun). In both experiments, the trend persists.

\subsection{Other Experiments: Further Decomposing the Alignment}
Our tripartite framework explains \emph{what drives similarity of representations} -- the universal sparse relationships between objects and attributes. Of course, differences between models do exist due to their architecture and training procedure as explained in Figure~\ref{fig:repdecomposition}. Which of these differences of trained models are responsible for the similarity (or, lack thereof) of representations? 

\begin{wrapfigure}{r}{.6\textwidth}
    \includegraphics[width=\linewidth]{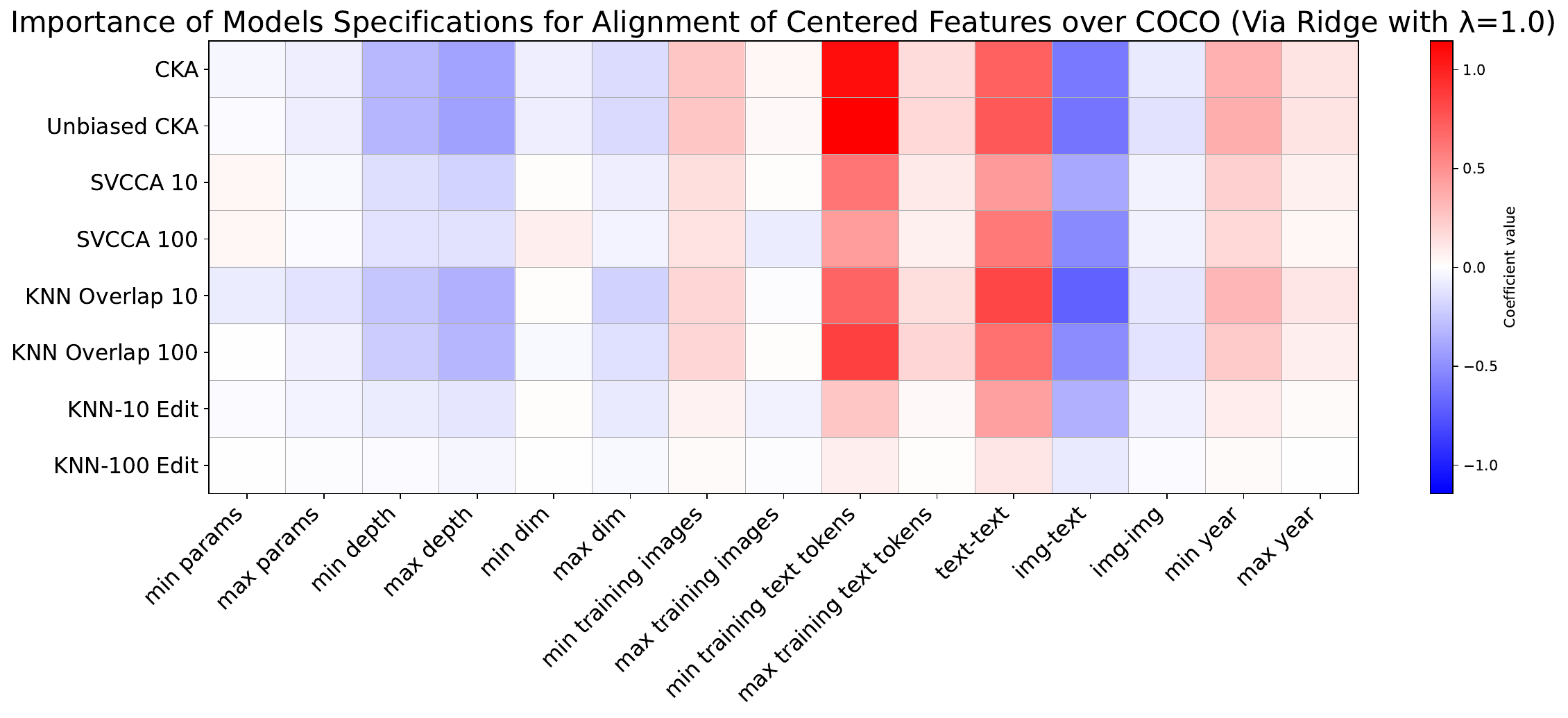}
    \caption{\small{Coefficients of ridge regression when regressing alignment on specifications of pairs of models. The same experiment with different datasets and values of $\lambda$ appears in Appendix~\ref{appendix:decomposingthealignment}. 
    }}
    \label{fig:regressionmain}
\end{wrapfigure}

\textbf{Experiment.}
To address this question, we set up the following experiment in which we \emph{regress} the alignment of models (we have 30 models, so these are $\binom{30}{2} = 435$ pairs) on 10 easy-to-quantify model properties. They are respectively: 1) \emph{Architectural}: model size, model depth, dimension of representations, objective. 2) \emph{Training-related:} number of text tokens and number of images used in training. 3) \emph{Mixed:} we also include the year of release of the model due to the innovation factor. More recent models tend to be better and, hence, can be expected to be more aligned to the Platonic ground truth.  Whenever the model specifics are numeric, we create features corresponding to the min and max over the two models. For the categorical feature of modality, we one-hot encode the three possibilities (image-image, image-text, text-text). This process gives us 15 features, which we normalize to have unit norm and zero mean. See more in Appendix~\ref{appendix:decomposingthealignment} about the feature-creation process. Due to the evident correlation of features (for example, depth and size, modality and size, etc), we apply a ridge penalty to obtain interpretable coefficients for the importance of different features. 

\textbf{Analysis.} Figure~\ref{fig:regressionmain} suggests that \emph{specifications of models have a consistent impact on alignment across metrics}. This statement supports the ordinal consistency of alignment across metrics made in \cite{huh24platonic}. Our decomposition also suggests the importance of the following model specifications. First, min training text tokens and min training images positively impact the alignment. This is consistent with our view that the signal-to-noise ratio in representations scales with training data size; according to this view, the model that has been trained on less data is noisier. Hence, a larger min value implies that both models are less noisy. Likewise, the interpretation of the consistent positive correlation with min year is similar -- if both models are more recent, we can expect them to both be closer to the Platonic ground truth. Also expected are the strong trends that text-text objectives strongly increase alignment, while different text-image objectives decrease alignment. 


%% file: sections/Limitations.tex
\section{Limitations and Future Directions}
\label{sec:limitations}
One limitation of our work is that the models in our experiments are up to 4B parameters,  
while frontier LLMs currently have more than a trillion parameters. A second limitation comes from the fact that the current literature on SAEs still has many gaps such as polysemanticity and feature absorption \cite{minegishi2025rethinking,bussmann2025learning}, the problem of non-canonical units \cite{leask2025sparse},
dead-neuron problem \cite{bricken2023towards}, the issue for top-$k$-based models that $k$ varies between objects in practice \cite{joseph2025steering}, and shrinkage problems on $L_1$-penalized SAEs \cite{rajamanoharan2024improving}. 


%% file: sections/StatModel.tex
\section{On The Statistical Model}
\label{appendix:OnTheStatisticalModel}
\subsection{Derivation of Bounds on Representation Components}
\label{appendix:ComponentBounds}

Following the~\ref{eq:statmodel}, $f_\theta(X) = A(\theta, f) \times ( Z(X) \odot M(X,\theta, f)) + \eta_{X,\theta, f},$ we derive the bounds for the inner product $\langle f_{\theta}(X_1), f_{\theta}(X_2)\rangle$ under the assumptions from Section~\ref{sec:statmodel}.

\textbf{Signal Bound.}
The signal component is defined as the interaction between shared attributes:
\begin{equation*}
    S = \sum_{i\; : \; (Z_1)_i = (Z_2)_i = 1} (M_1)_i(M_2)_i.
\end{equation*}
By Assumption A.3, for any active attribute $i$, the magnitude is bounded by $\frac{\phi_{\text{sig}}}{\sqrt{k}} \le M_i \le \frac{\Phi_{\text{sig}}}{\sqrt{k}}.$ 
Since the number of shared attributes is exactly the inner product of the binary vectors $\langle Z_1, Z_2 \rangle,$ the summation contains $\langle Z_1, Z_2 \rangle$ terms. 
Applying the magnitude bounds:
\begin{itemize}
    \item \textbf{Lower Bound}: $S \ge \langle Z_1, Z_2 \rangle \cdot \left(\frac{\phi_{\text{sig}}}{\sqrt{k}}\right)^2 = \frac{\phi_{\text{sig}}^2 \langle Z_1, Z_2 \rangle}{k}.$
    \item \textbf{Upper Bound}: $S \le \langle Z_1, Z_2 \rangle \cdot \left(\frac{\Phi_{\text{sig}}}{\sqrt{k}}\right)^2 = \frac{\Phi_{\text{sig}}^2 \langle Z_1, Z_2 \rangle}{k}.$
\end{itemize}

\textbf{Bias Bound.}
The bias component arises from the dictionary incoherence between different attribute representations:
\begin{equation*}
    B = \sum_{i \neq j \; : \; (Z_1)_i = 1, (Z_2)_j = 1} (M_1)_i (M_2)_j \langle A_i, A_j \rangle.
\end{equation*}
There are at most $k^2$ such pairs $(i, j)$ since each vector is $k$-sparse.
By Assumption A.2, $|\langle A_i, A_j \rangle| \le \epsilon_{\text{dict}}.$
Using the upper bound for magnitudes $M_i \le \frac{\Phi_{\text{sig}}}{\sqrt{k}},$
\begin{equation*}
    |B| \le k^2 \cdot \left(\frac{\Phi_{\text{sig}}}{\sqrt{k}}\right)^2 \cdot \epsilon_{\text{dict}} = k\times  \epsilon_{\text{dict}}\times \Phi_{\text{sig}}^2.
\end{equation*}

\textbf{Noise Bound.}
The noise component involves the interactions between the signal and the estimation error $\eta$ and can be written as:
\begin{equation*}
    N = \langle f_\theta(X_1), \eta_2 \rangle + \langle f_\theta(X_2), \eta_1 \rangle - \langle \eta_1, \eta_2 \rangle.
\end{equation*}
By Cauchy-Schwarz and the fact that $\|f_\theta(X_1)\|_2  = \|f_\theta(X_2)\|_2 = 1,$
\begin{itemize}
    \item $|\langle f_\theta(X_1), \eta_2 \rangle| \le \|f_\theta(X_1)\|_2 \|\eta_2\|_2 \le \epsilon_{\text{noise}}.$
    \item $|\langle f_\theta(X_2), \eta_1 \rangle| \le \epsilon_{\text{noise}}.$
    \item $|\langle \eta_1, \eta_2 \rangle| \le \|\eta_1\|_2 \|\eta_2\|_2 \le \epsilon_{\text{noise}}^2 < \epsilon_{\text{noise}}.$
\end{itemize}
Summing these terms and applying the triangle inequality yields the bound $|N| \le 3\epsilon_{\text{noise}}.$\qedhere

\subsection{Proof of Proposition 1}

Recall from the Statistical Model of Representations that the inner product of two representations can be decomposed into signal ($S$), bias ($B$), and noise ($N$) components: 
\[ \langle f_\theta(X_1), f_\theta(X_2) \rangle = S + B + N \]
Recall that these components are bounded as $|B| \le k \epsilon_{\mathrm{dict}} \Phi_{\mathrm{sig}}^2$ and $|N| \le 3\epsilon_{\mathrm{noise}}$.

\textbf{Proof of Part 1 (Signal Alignment)} 
Assume $\langle Z(X_1), Z(X_2) \rangle \ge \frac{2k}{\phi_{\mathrm{sig}}^2} \left( k \epsilon_{\mathrm{dict}} \Phi_{\mathrm{sig}}^2 + 3\epsilon_{\mathrm{noise}} + \gamma/2 \right)$. 
From the lower bound of the signal component, we have $S \ge \langle Z_1, Z_2 \rangle \frac{\phi_{\mathrm{sig}}^2}{k}$. Substituting the assumption:
\[ S \ge \frac{2k}{\phi_{\mathrm{sig}}^2} \left( k \epsilon_{\mathrm{dict}} \Phi_{\mathrm{sig}}^2 + 3\epsilon_{\mathrm{noise}} + \gamma/2 \right) \cdot \frac{\phi_{\mathrm{sig}}^2}{k} = 2(k \epsilon_{\mathrm{dict}} \Phi_{\mathrm{sig}}^2 + 3\epsilon_{\mathrm{noise}} + \gamma/2) \]
Using the triangle inequality $\langle f_\theta(X_1), f_\theta(X_2) \rangle \ge S - |B| - |N|$ and applying the upper bounds for bias and noise:
\begin{align*}
    \langle f_\theta(X_1), f_\theta(X_2) \rangle &\ge 2k \epsilon_{\mathrm{dict}} \Phi_{\mathrm{sig}}^2 + 6\epsilon_{\mathrm{noise}} + \gamma - k \epsilon_{\mathrm{dict}} \Phi_{\mathrm{sig}}^2 - 3\epsilon_{\mathrm{noise}} \\
    &= k \epsilon_{\mathrm{dict}} \Phi_{\mathrm{sig}}^2 + 3\epsilon_{\mathrm{noise}} + \gamma
\end{align*}

\textbf{Proof of Part 2 (Alignment Signal)} 
Assume $\langle f_\theta(X_1), f_\theta(X_2) \rangle \ge k \epsilon_{\mathrm{dict}} \Phi_{\mathrm{sig}}^2 + 3\epsilon_{\mathrm{noise}} + \gamma$. 
Isolating the signal component, we have $S = \langle f_\theta(X_1), f_\theta(X_2) \rangle - B - N$. By the triangle inequality:
\[ S \ge \langle f_\theta(X_1), f_\theta(X_2) \rangle - |B| - |N| \]
Substituting the hypothesis and the component bounds:
\[ S \ge (k \epsilon_{\mathrm{dict}} \Phi_{\mathrm{sig}}^2 + 3\epsilon_{\mathrm{noise}} + \gamma) - k \epsilon_{\mathrm{dict}} \Phi_{\mathrm{sig}}^2 - 3\epsilon_{\mathrm{noise}} = \gamma \]
From the signal upper bound $S \le \frac{\Phi_{\mathrm{sig}}^2 \langle Z_1, Z_2 \rangle}{k}$, we conclude:
\[ \langle Z(X_1), Z(X_2) \rangle \ge \frac{k}{\Phi_{\mathrm{sig}}^2} S \ge \frac{k}{\Phi_{\mathrm{sig}}^2} \gamma. \]
\qedhere

\clearpage

%% file: sections/PseudoCode.tex
\section{Pseudo-Code}
\subsection{Top-\texorpdfstring{$K$}{K} SAE}
\label{sec:saepseudo}

\begin{algorithm}[H]
\caption{Top-K Sparse Autoencoder with Dead Neuron Resampling}
\begin{algorithmic}[1]

\Require dataset $X \in \mathbb{R}^{N \times d_{\text{model}}}$, hidden size $d_{\text{sparse}}$, sparsity level $k$,
batch size $B$, number of steps $T$, learning rate $\eta$, weight decay $\lambda$,
resampling period $R$, print period $P$, optional decoder renormalization
\Ensure trained parameters $(W_e, b_e, W_d, b_{\text{dec}})$

\State Initialize encoder $W_e \in \mathbb{R}^{d_{\text{sparse}} \times d_{\text{model}}}$ and $b_e \in \mathbb{R}^{d_{\text{sparse}}}$
\State Initialize decoder $W_d \in \mathbb{R}^{d_{\text{model}} \times d_{\text{sparse}}}$ with Kaiming-uniform initialization
\State Initialize decoder bias $b_{\text{dec}} \in \mathbb{R}^{d_{\text{model}}}$ to $0$
\State Set $b_{\text{dec}} \gets \frac{1}{N} \sum_{i=1}^{N} X_i$ \Comment{data mean initialization}
\State Initialize AdamW optimizer on $(W_e, b_e, W_d, b_{\text{dec}})$ with learning rate $\eta$ and weight decay $\lambda$
\State Initialize activity counter $c \in \mathbb{R}^{d_{\text{sparse}}}$ to $0$

\For{$t = 1$ to $T$}
    \State Sample indices $\mathcal{I} \sim \mathrm{Unif}\{1,\dots,N\}$ of size $B$
    \State $x \gets X[\mathcal{I}]$

    \Statex
    \Comment{Forward pass}
    \State $\tilde{x} \gets x - b_{\text{dec}}$
    \State $f \gets \mathrm{ReLU}(W_e \tilde{x} + b_e)$
    \State For each row of $f$, keep only its top-$k$ values:
    \[
      z_{ij} \gets
      \begin{cases}
      f_{ij}, & \text{if } f_{ij} \text{ is among the top-$k$ entries in row } i \\
      0, & \text{otherwise}
      \end{cases}
    \]
    \State $\hat{x} \gets W_d z + b_{\text{dec}}$

    \Statex
    \Comment{Optimization step}
    \State $\mathcal{L}_{\mathrm{recon}} \gets \mathrm{MSE}(\hat{x}, x)$
    \State Take one AdamW step on $\mathcal{L}_{\mathrm{recon}}$

    \If{decoder renormalization is enabled}
        \For{$j = 1$ to $d_{\text{sparse}}$}
            \State $W_d[:,j] \gets W_d[:,j] / \max(\lVert W_d[:,j]\rVert_2, \varepsilon)$
        \EndFor
    \EndIf

    \Statex
    \Comment{Track feature activity}
    \State $c \gets c + \sum_{i=1}^{B} \mathbf{1}[z_i > 0]$

    \If{$t \bmod R = 0$ \textbf{and} $t < 0.8T$}
        \State $m \gets \Call{ResampleDeadNeurons}{X, c, W_e, b_e, W_d, b_{\text{dec}}, k}$
        \State $c \gets 0$
    \EndIf

\EndFor

\State \Return $(W_e, b_e, W_d, b_{\text{dec}})$

\end{algorithmic}
\end{algorithm}

\clearpage

\begin{algorithm}[H]
\caption{Dead Neuron Resampling}
\begin{algorithmic}[1]

\Function{ResampleDeadNeurons}{$X, c, W_e, b_e, W_d, b_{\text{dec}}, k$}
    \State $\mathcal{D} \gets \{j \in \{1,\dots,d_{\text{sparse}}\} : c_j = 0\}$ \Comment{dead neurons}
    \State $m \gets |\mathcal{D}|$
    \If{$m = 0$}
        \State \Return $0$
    \EndIf

    \State Sample indices $\mathcal{I} \sim \mathrm{Unif}\{1,\dots,N\}$ of size $m$
    \State $x \gets X[\mathcal{I}]$

    \Statex
    \Comment{Compute residuals under current model}
    \State $\tilde{x} \gets x - b_{\text{dec}}$
    \State $f \gets \mathrm{ReLU}(W_e \tilde{x} + b_e)$
    \State Form top-$k$ sparse codes $z$ by keeping only the top-$k$ entries per row of $f$
    \State $\hat{x} \gets W_d z + b_{\text{dec}}$
    \State $r \gets x - \hat{x}$

    \Statex
    \Comment{Normalize residuals and use them to reset dead features}
    \For{$q = 1$ to $m$}
        \State $u_q \gets r_q / (\|r_q\|_2 + 10^{-8})$
        \State Let $j = \mathcal{D}[q]$
        \State $W_d[:,j] \gets u_q$
        \State $W_e[j,:] \gets 0.1 \, u_q$
        \State $b_e[j] \gets 0$
    \EndFor

    \State \Return $m$
\EndFunction

\end{algorithmic}
\end{algorithm}

\clearpage

%% file: sections/FurtherExperiments.tex
\section{Further Experiments and Experimental Details}
\label{appendix:experimentaldetails}
We reproduce all experiments from the main paper for further metrics, namely CKA, Unbiased CKA, SVCCA in dimension 10, SVCCA in dimension 100, KNN-10 neighborhood Overlap, KNN-100 neighborhood Overlap, 
KNN-10 neighborhood edit distance, KNN-100 neighborhood edit distance. We note that for all metrics, larger values mean more similar except for the two edit distances where larger value means less similar.  
\subsection{Signal: Alignment of Sparse Features}
\label{appendix:signal}
\subsubsection{Experiments on COCO}
\label{appendix:cocosignal}

\begin{figure}[htbp]
    \centering

    \begin{minipage}[t]{0.3\textwidth}
        \centering
        \includegraphics[width = \linewidth]{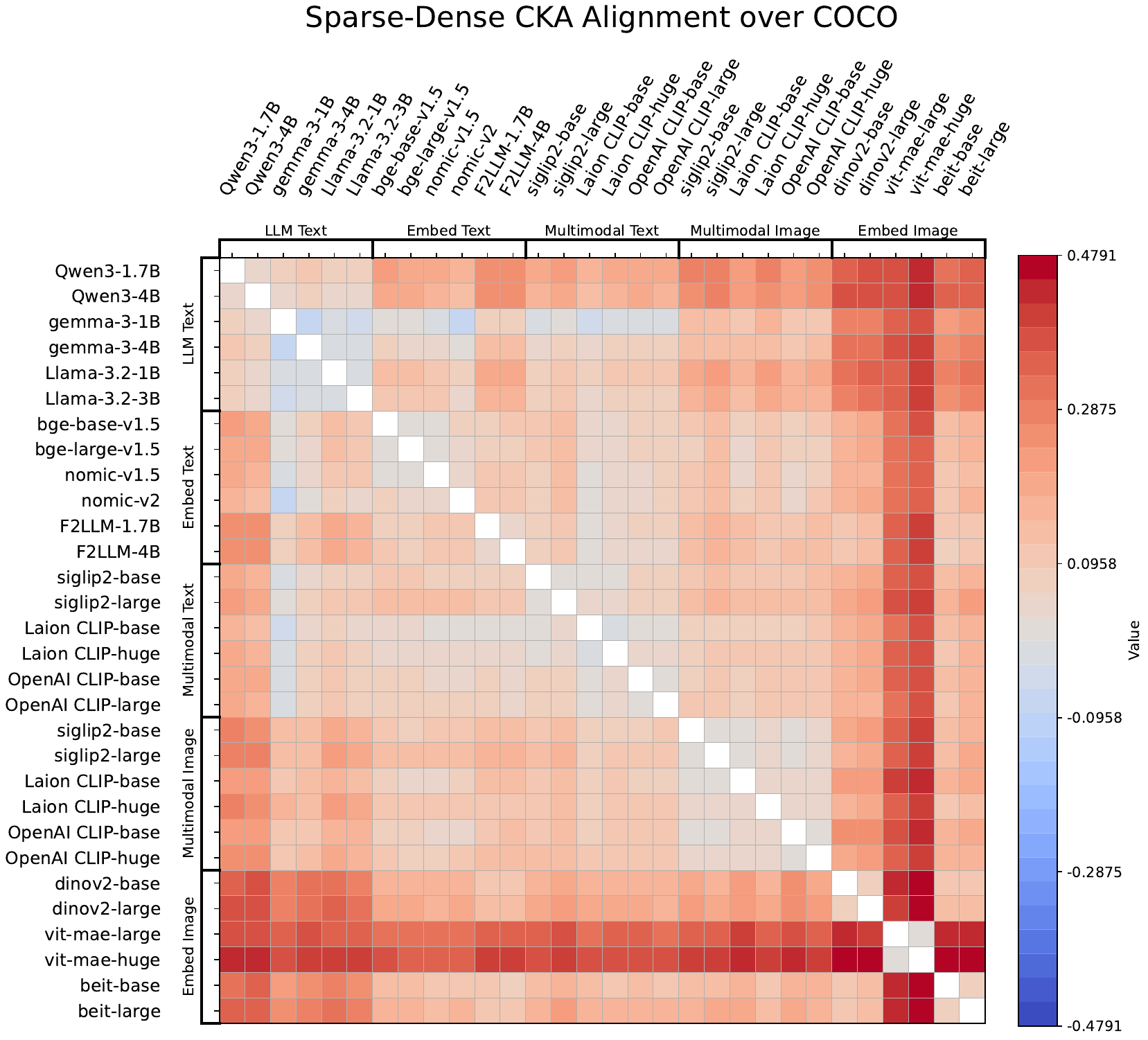}
    \end{minipage}
    \hfill
    \begin{minipage}[t]{0.3\textwidth}
        \centering
        \includegraphics[width = \linewidth]{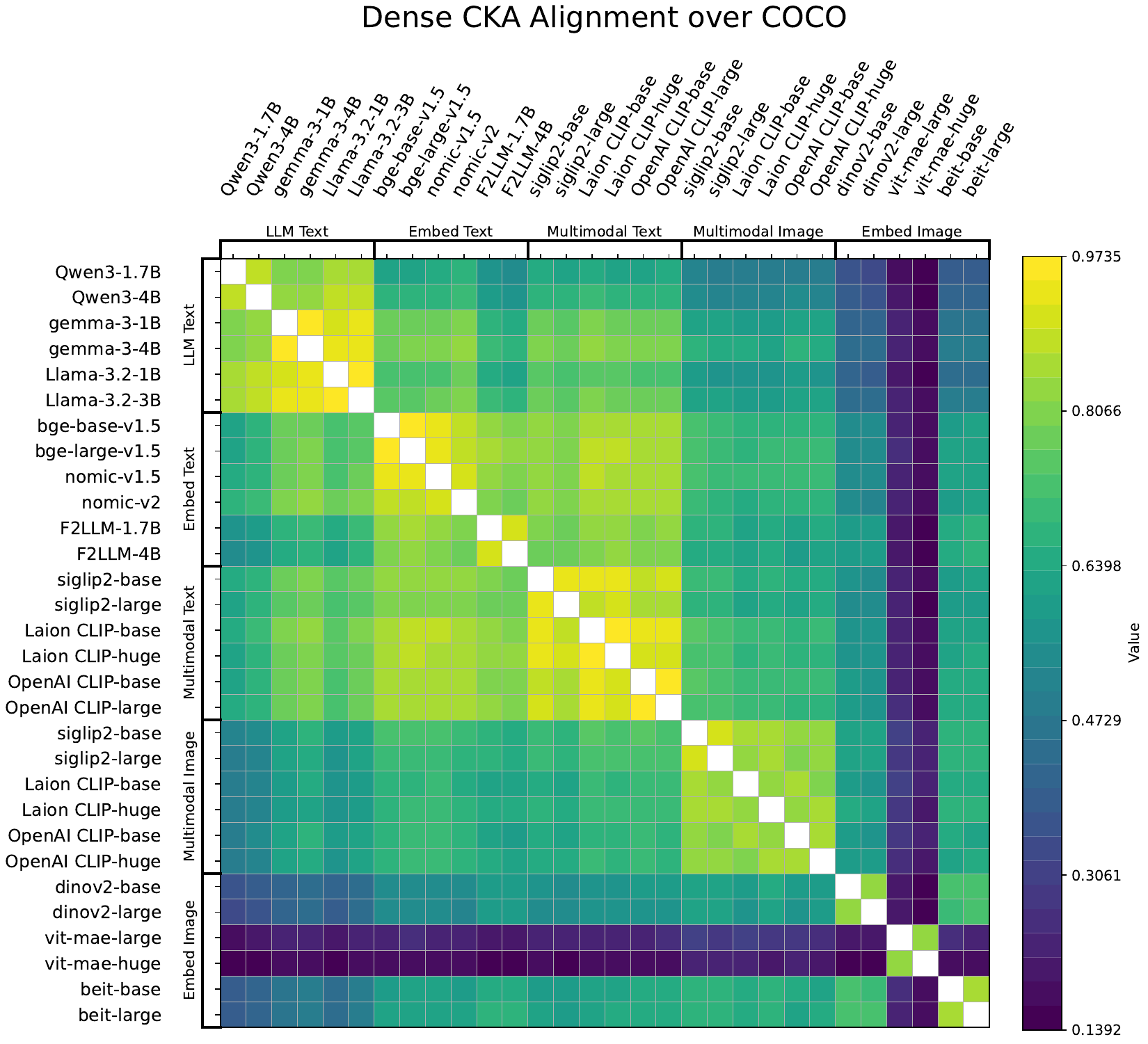}
    \end{minipage}
    \hfill
    \begin{minipage}[t]{0.3\textwidth}
        \centering
        \includegraphics[width = \linewidth]{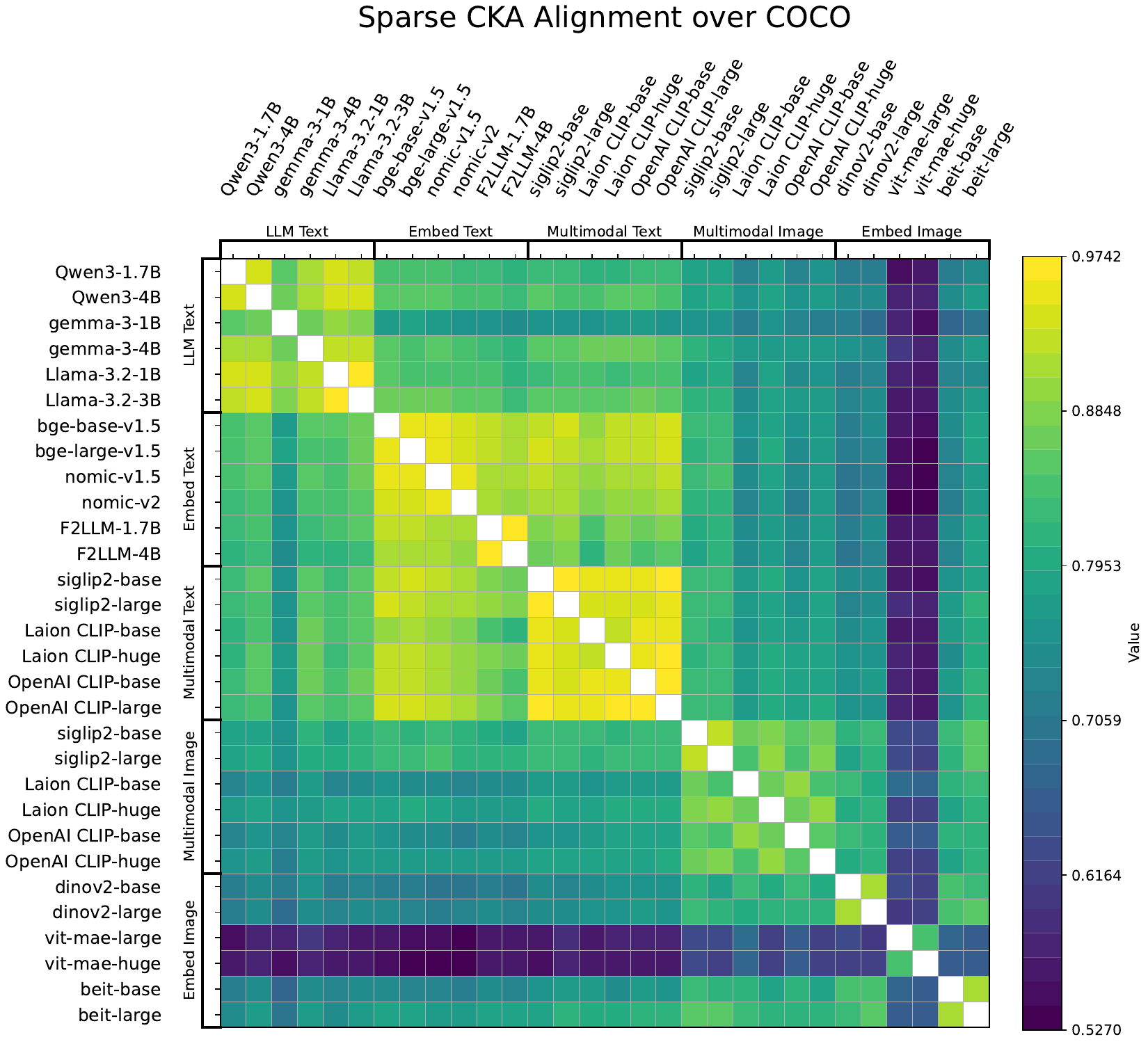}
    \end{minipage}
    
    \vspace{0.4cm}

        \begin{minipage}[t]{0.3\textwidth}
        \centering
        \includegraphics[width = \linewidth]{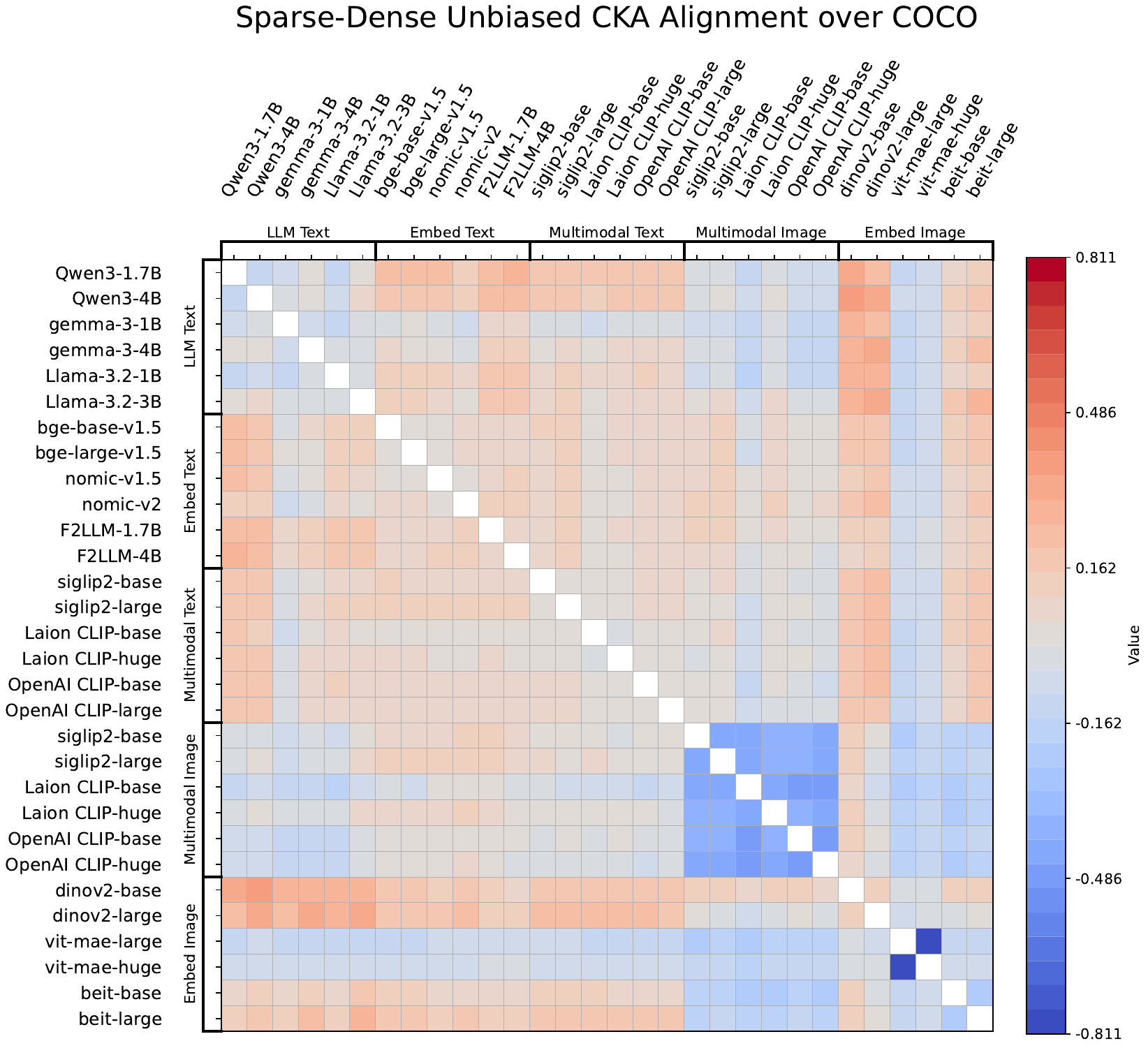}
    \end{minipage}
    \hfill
    \begin{minipage}[t]{0.3\textwidth}
        \centering
        \includegraphics[width = \linewidth]{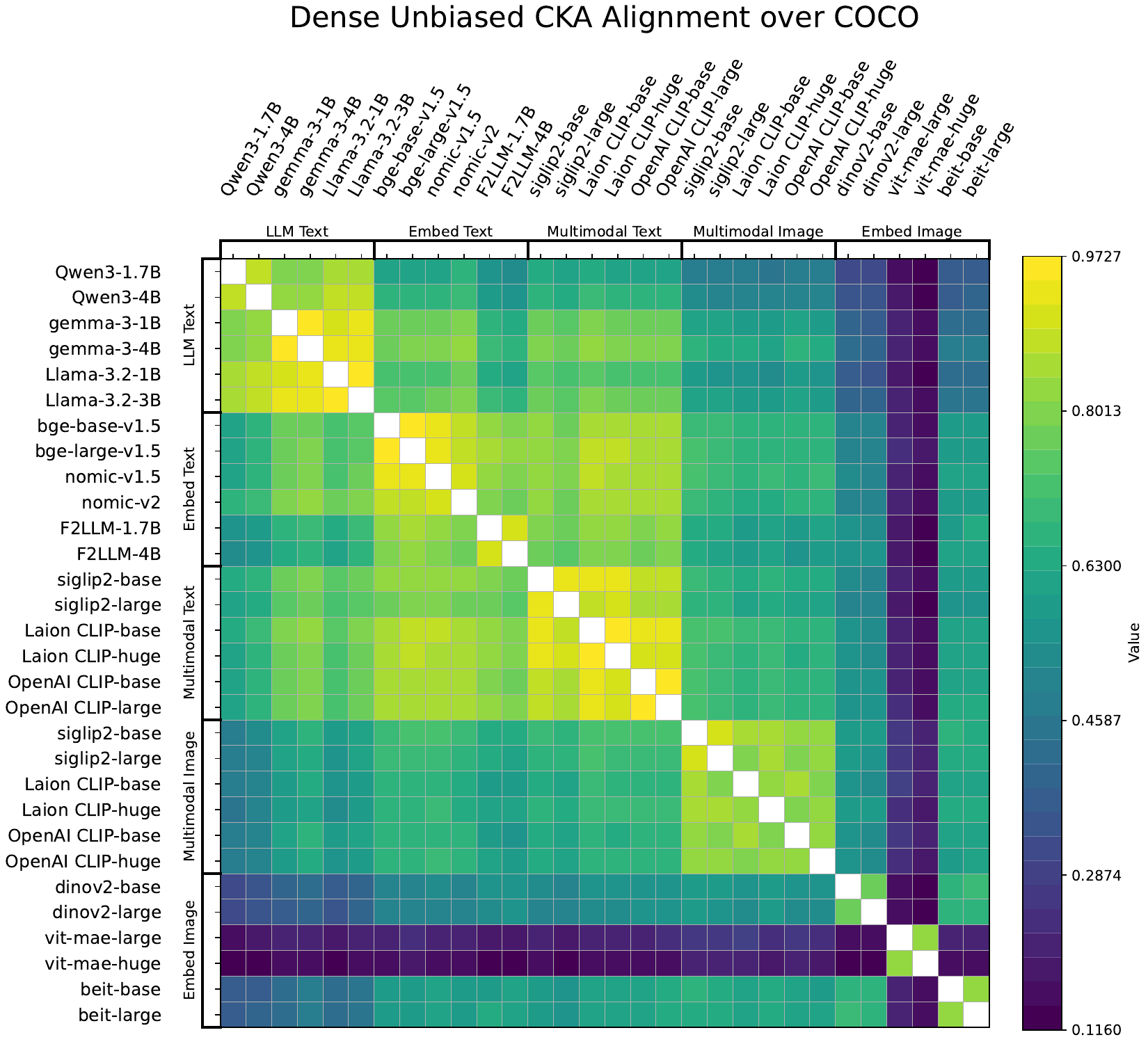}
    \end{minipage}
    \hfill
    \begin{minipage}[t]{0.3\textwidth}
        \centering
        \includegraphics[width = \linewidth]{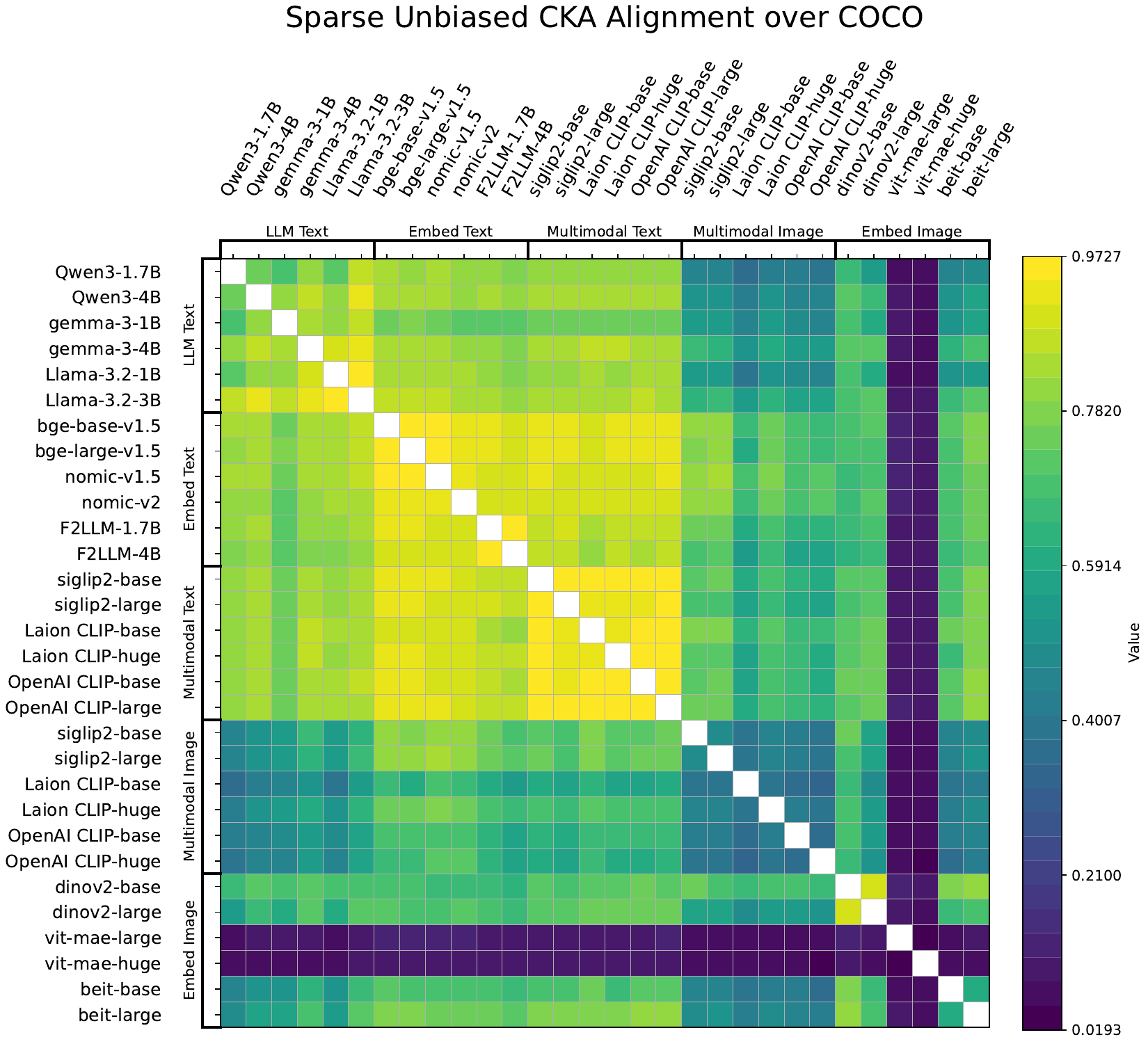}
    \end{minipage}

        \vspace{0.4cm}

        \begin{minipage}[t]{0.3\textwidth}
        \centering
        \includegraphics[width = \linewidth]{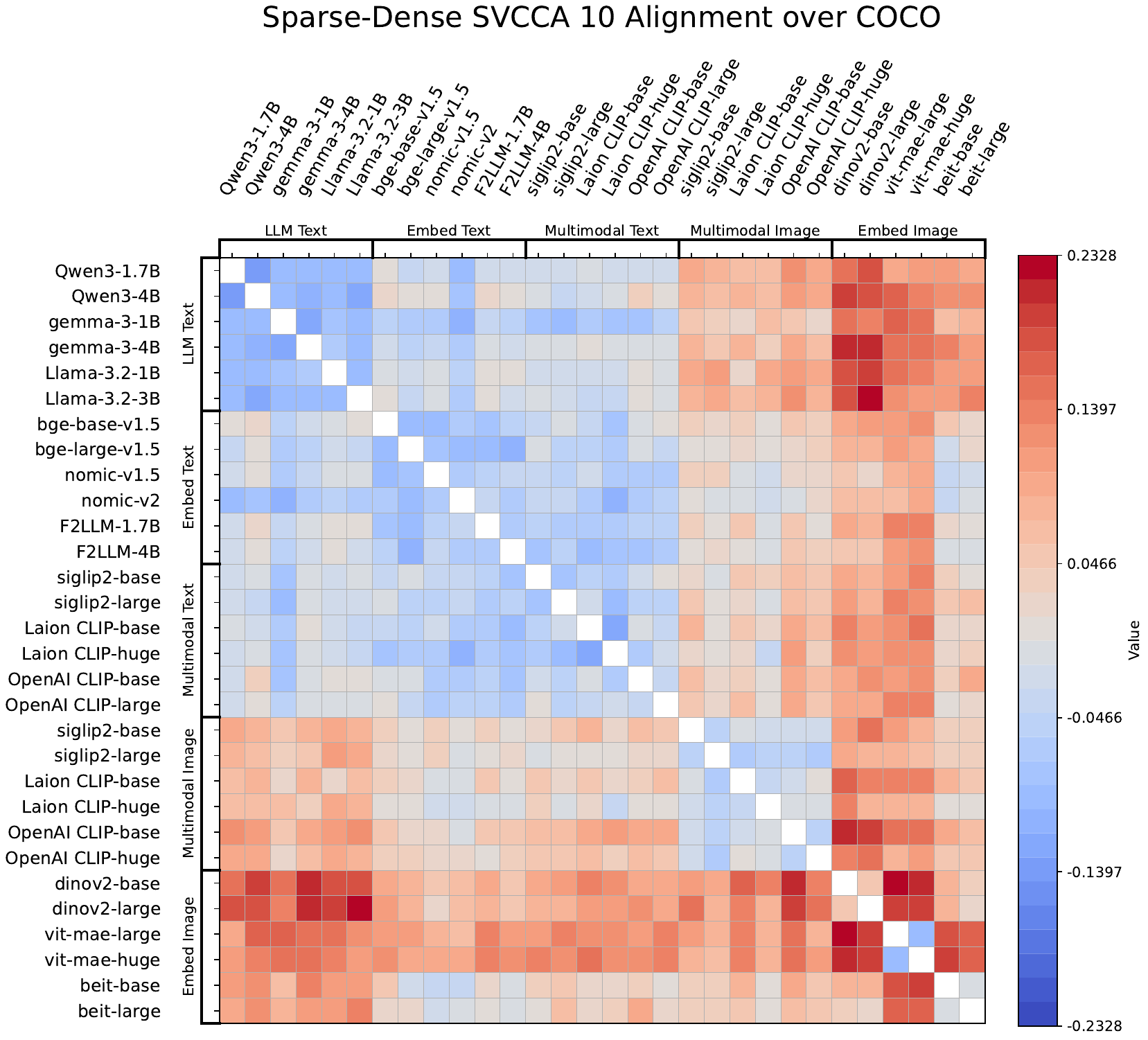}
    \end{minipage}
    \hfill
    \begin{minipage}[t]{0.3\textwidth}
        \centering
        \includegraphics[width = \linewidth]{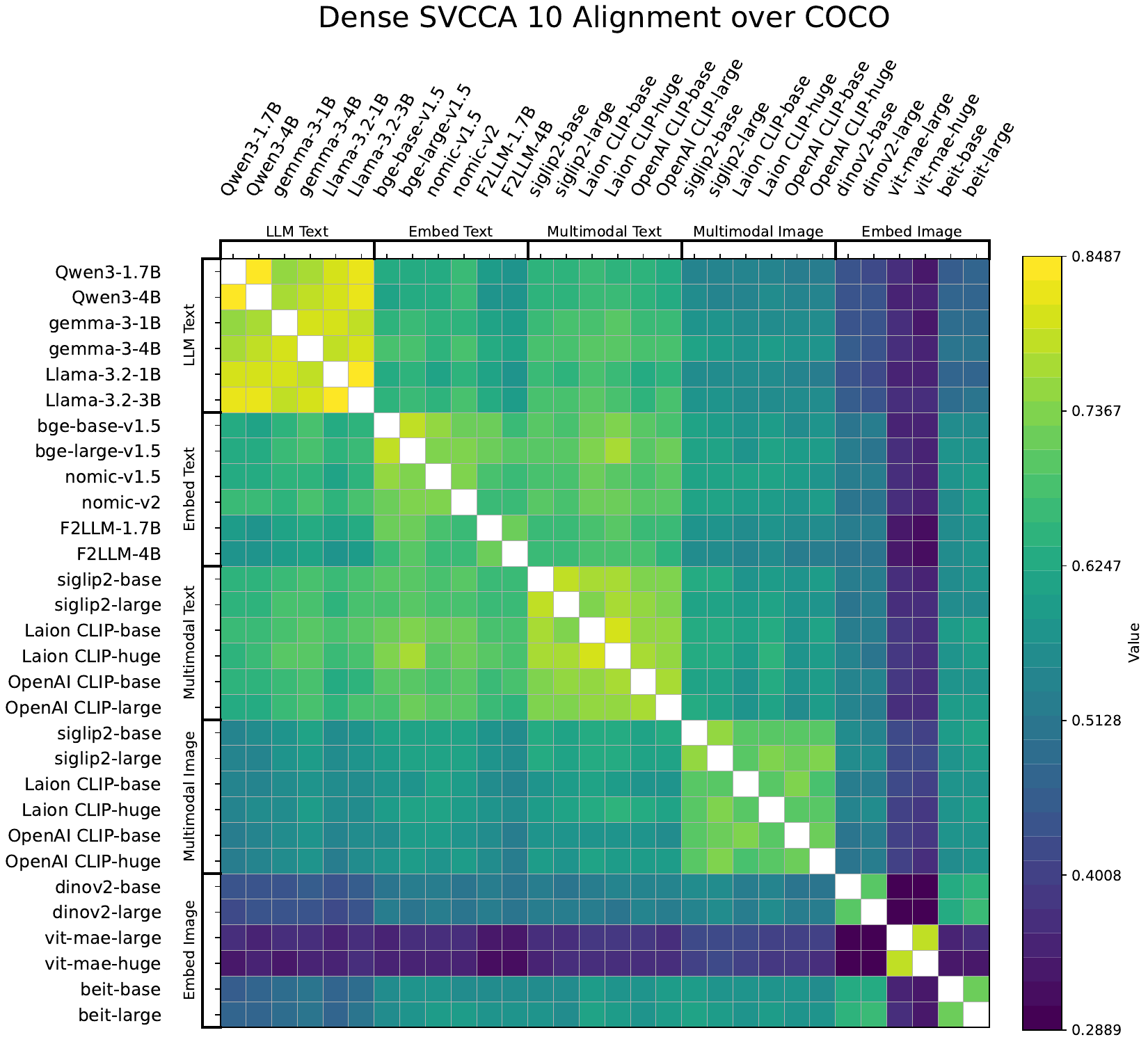}
    \end{minipage}
    \hfill
    \begin{minipage}[t]{0.3\textwidth}
        \centering
        \includegraphics[width = \linewidth]{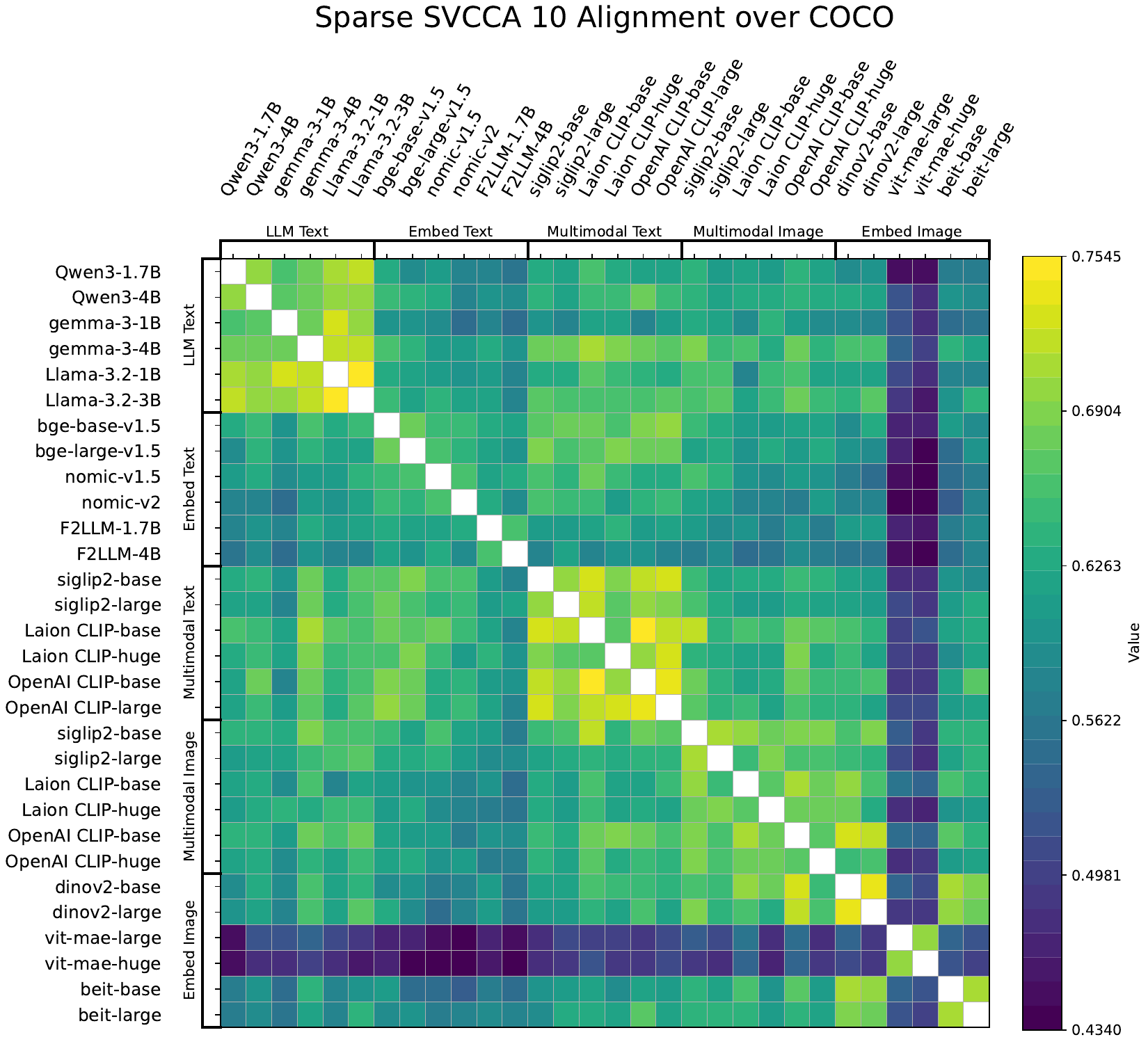}
    \end{minipage}

            \vspace{0.4cm}

    \begin{minipage}[t]{0.3\textwidth}
        \centering
        \includegraphics[width = \linewidth]{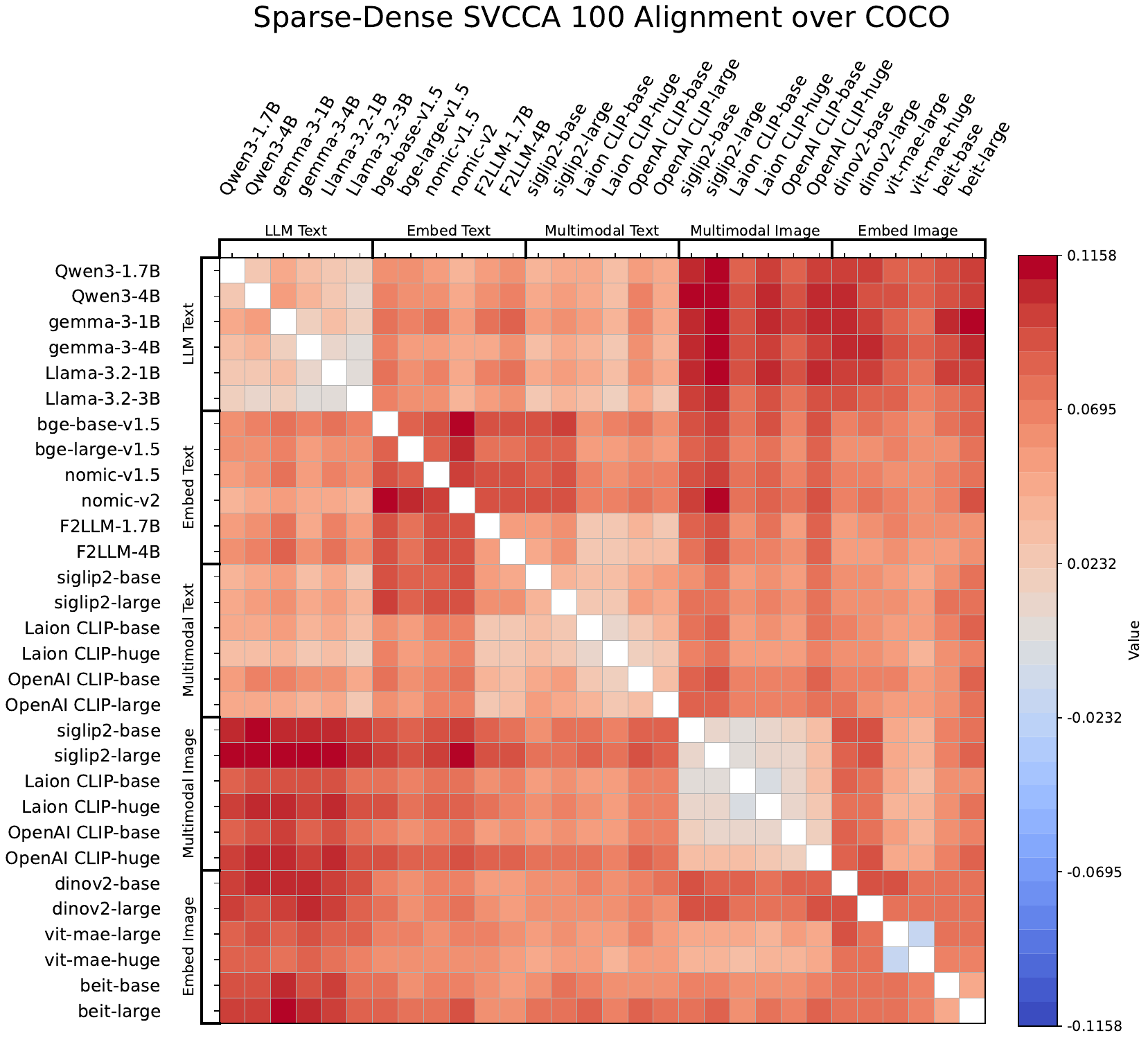}
    \end{minipage}
    \hfill
    \begin{minipage}[t]{0.3\textwidth}
        \centering
        \includegraphics[width = \linewidth]{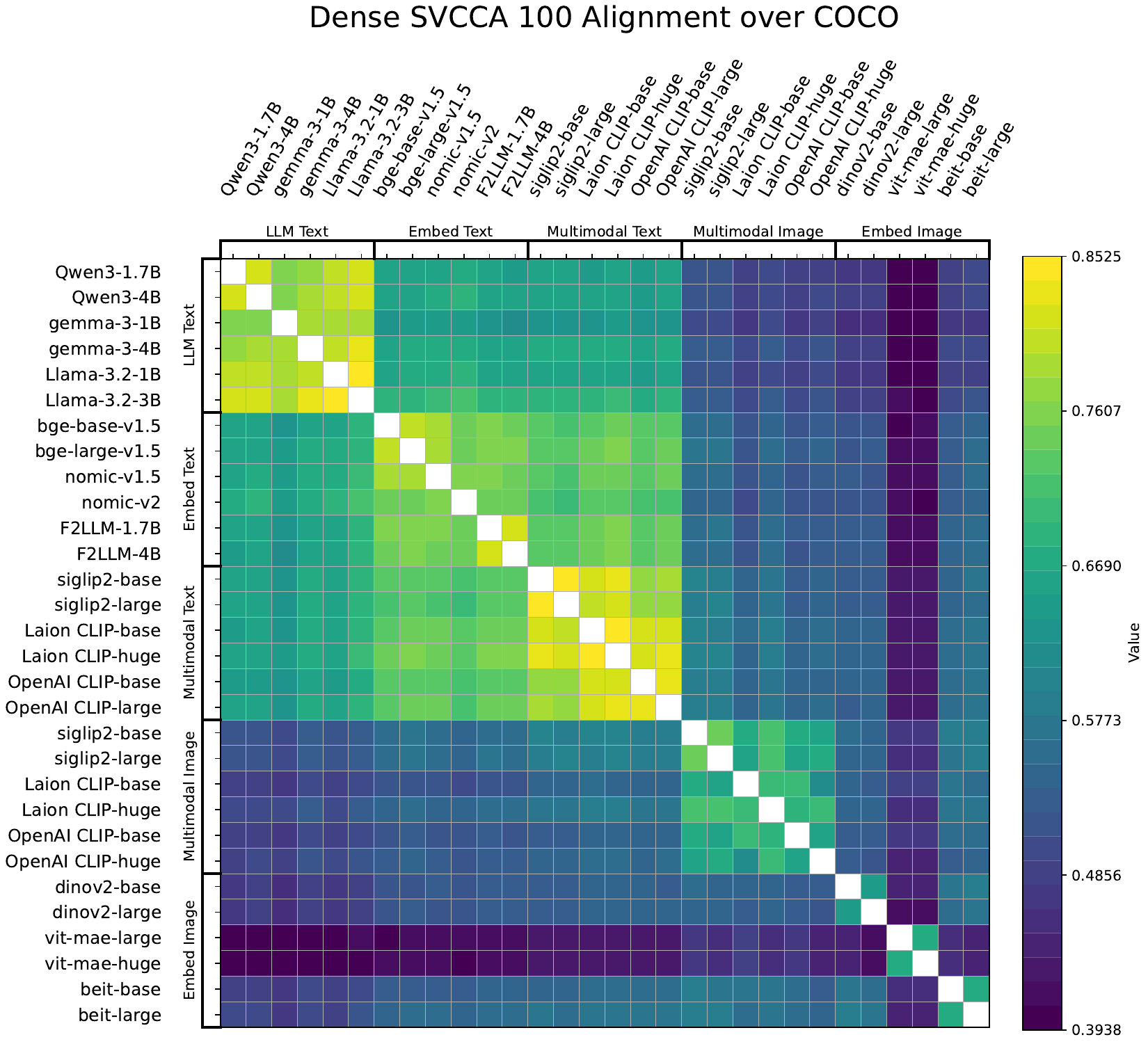}
    \end{minipage}
    \hfill
    \begin{minipage}[t]{0.3\textwidth}
        \centering
        \includegraphics[width = \linewidth]{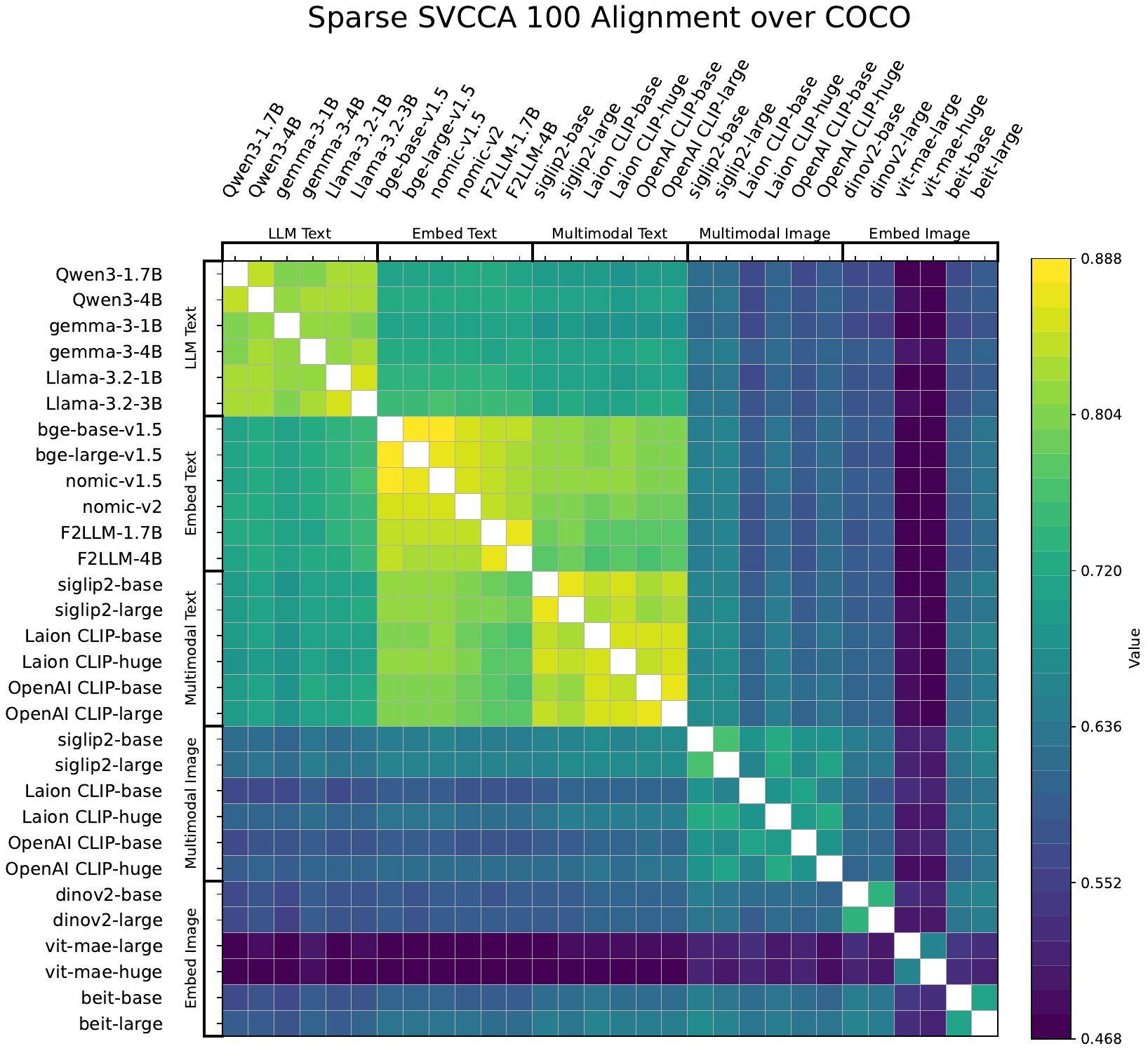}
    \end{minipage}
    \caption{Same plot as in Figure~\ref{fig:signalmain} but for the CKA, Unbiased CKA, SVCCA 10, and SVCCA 100 metrics. In addition to differences, we plot the alignment values for raw dense features and for the sparse features. The SAE dimension is 16384. We do an ablation study with dimension 8192.}
    \label{fig:signalcoco1}
\end{figure}

\clearpage

\begin{figure}[htbp]
    \centering

    \begin{minipage}[t]{0.3\textwidth}
        \centering
        \includegraphics[width = \linewidth]{signal_diagrams/topk_10_coco_diff_16384.pdf}
    \end{minipage}
    \hfill
    \begin{minipage}[t]{0.3\textwidth}
        \centering
        \includegraphics[width = \linewidth]{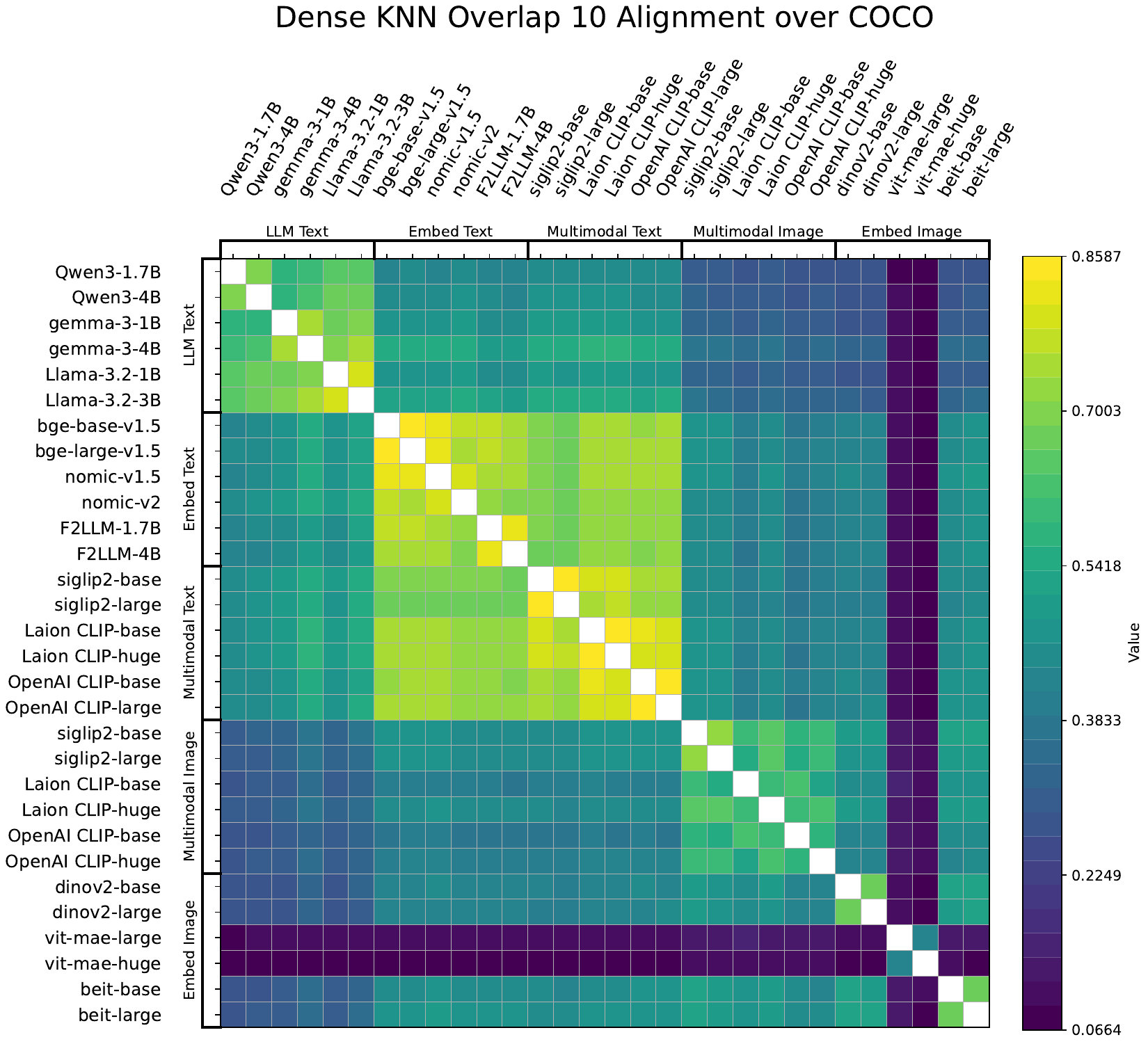}
    \end{minipage}
    \hfill
    \begin{minipage}[t]{0.3\textwidth}
        \centering
        \includegraphics[width = \linewidth]{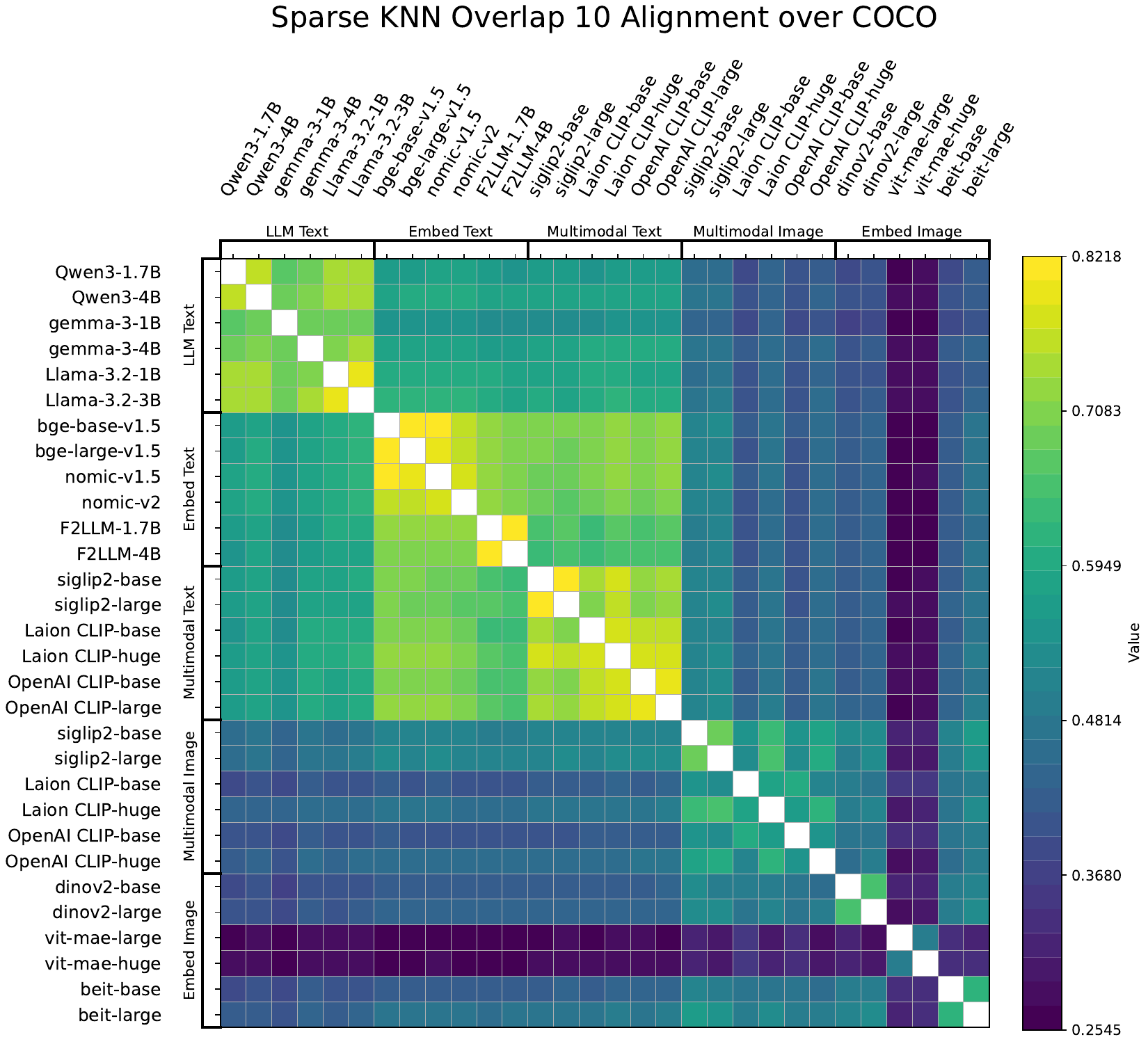}
    \end{minipage}
    
    \vspace{0.4cm}

        \begin{minipage}[t]{0.3\textwidth}
        \centering
        \includegraphics[width = \linewidth]{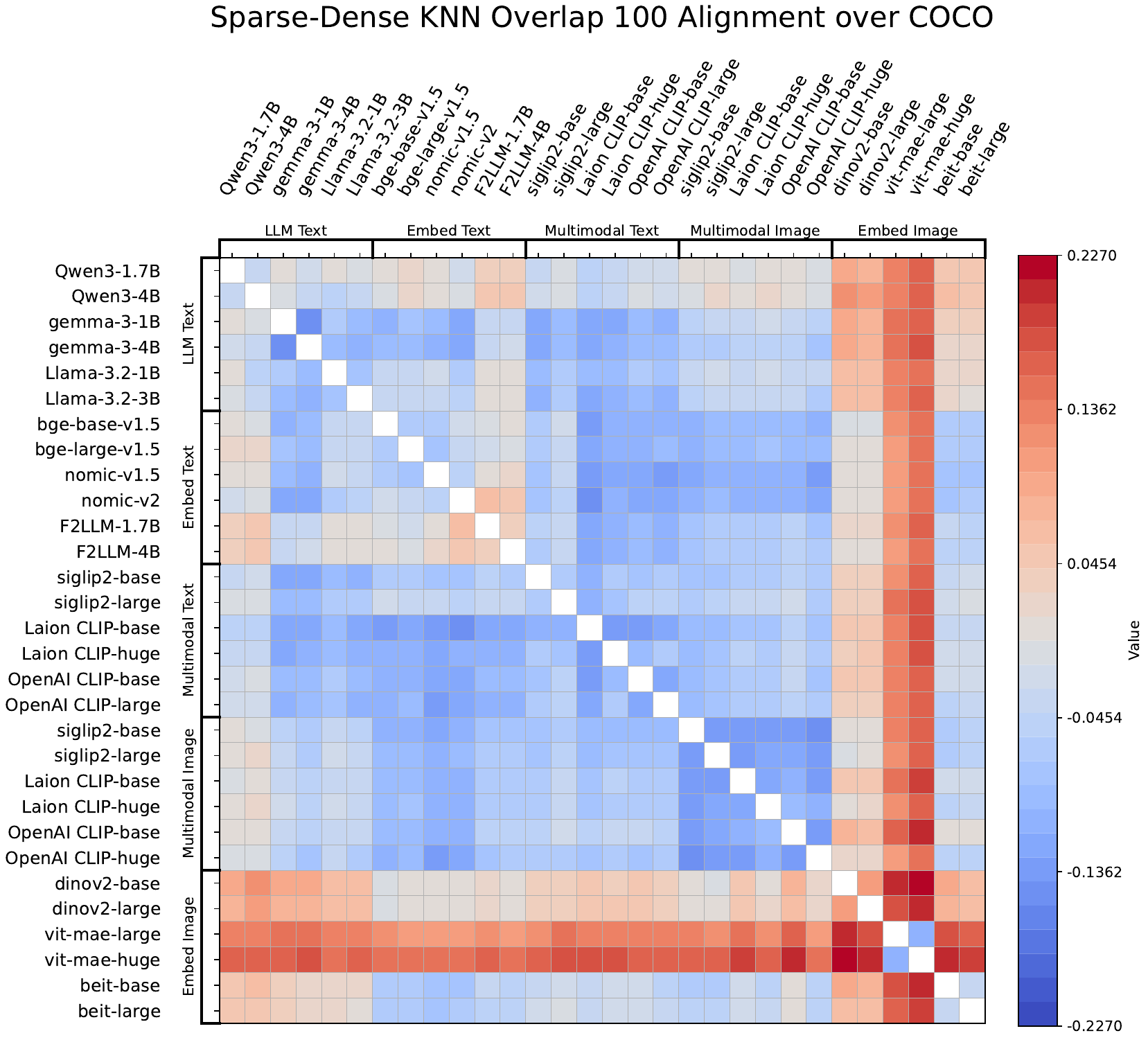}
    \end{minipage}
    \hfill
    \begin{minipage}[t]{0.3\textwidth}
        \centering
        \includegraphics[width = \linewidth]{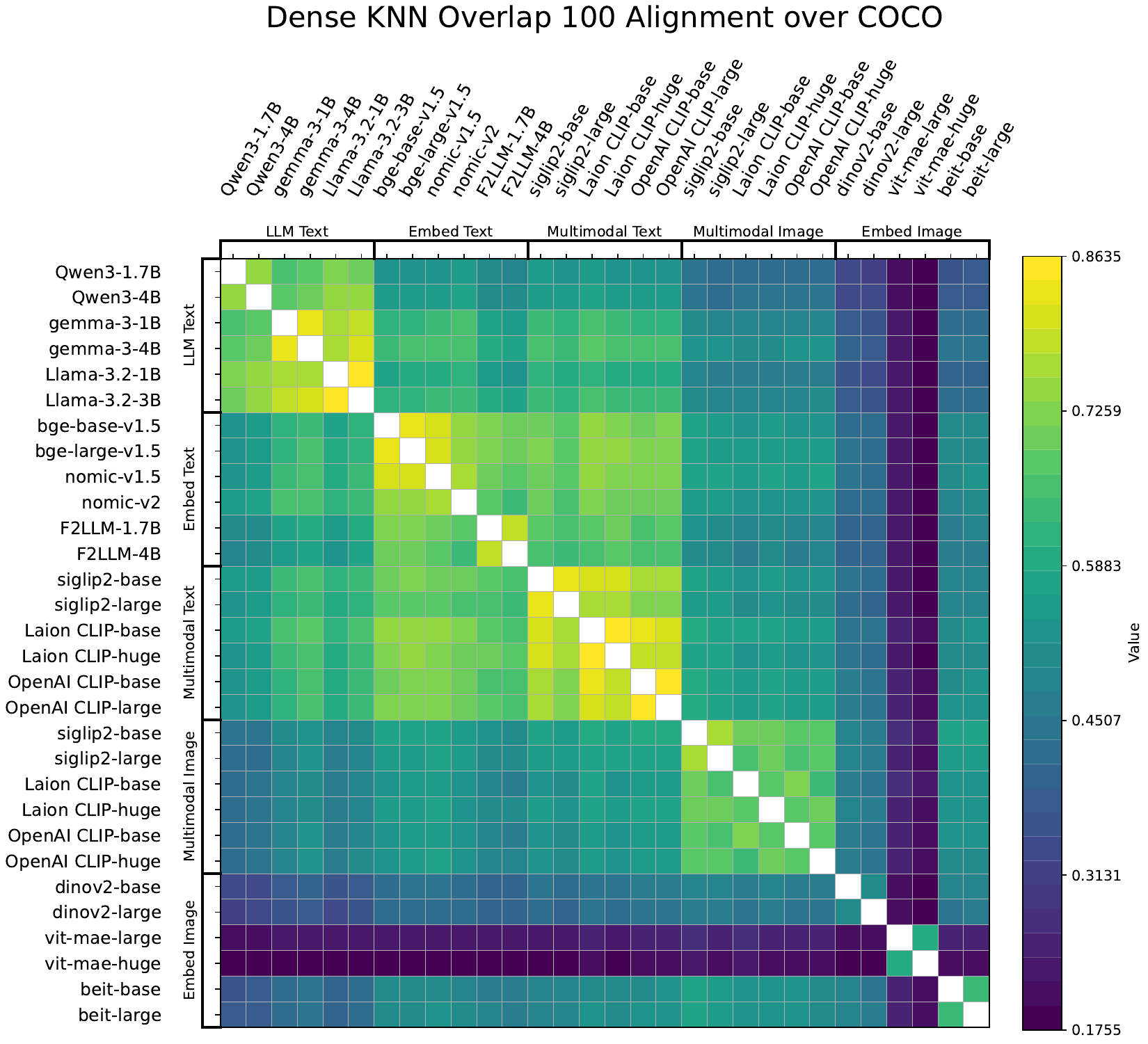}
    \end{minipage}
    \hfill
    \begin{minipage}[t]{0.3\textwidth}
        \centering
        \includegraphics[width = \linewidth]{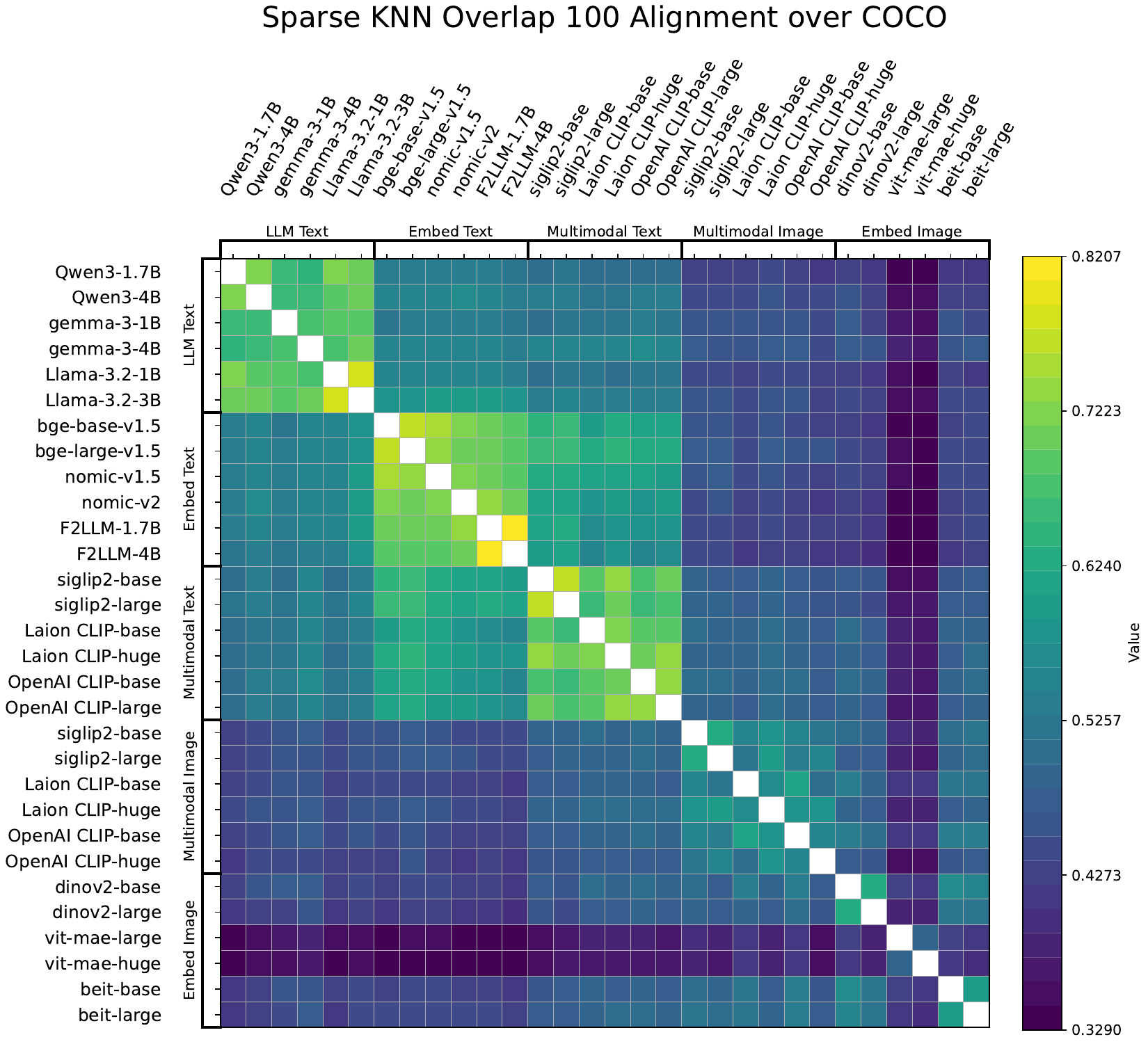}
    \end{minipage}

        \vspace{0.4cm}

        \begin{minipage}[t]{0.3\textwidth}
        \centering
        \includegraphics[width = \linewidth]{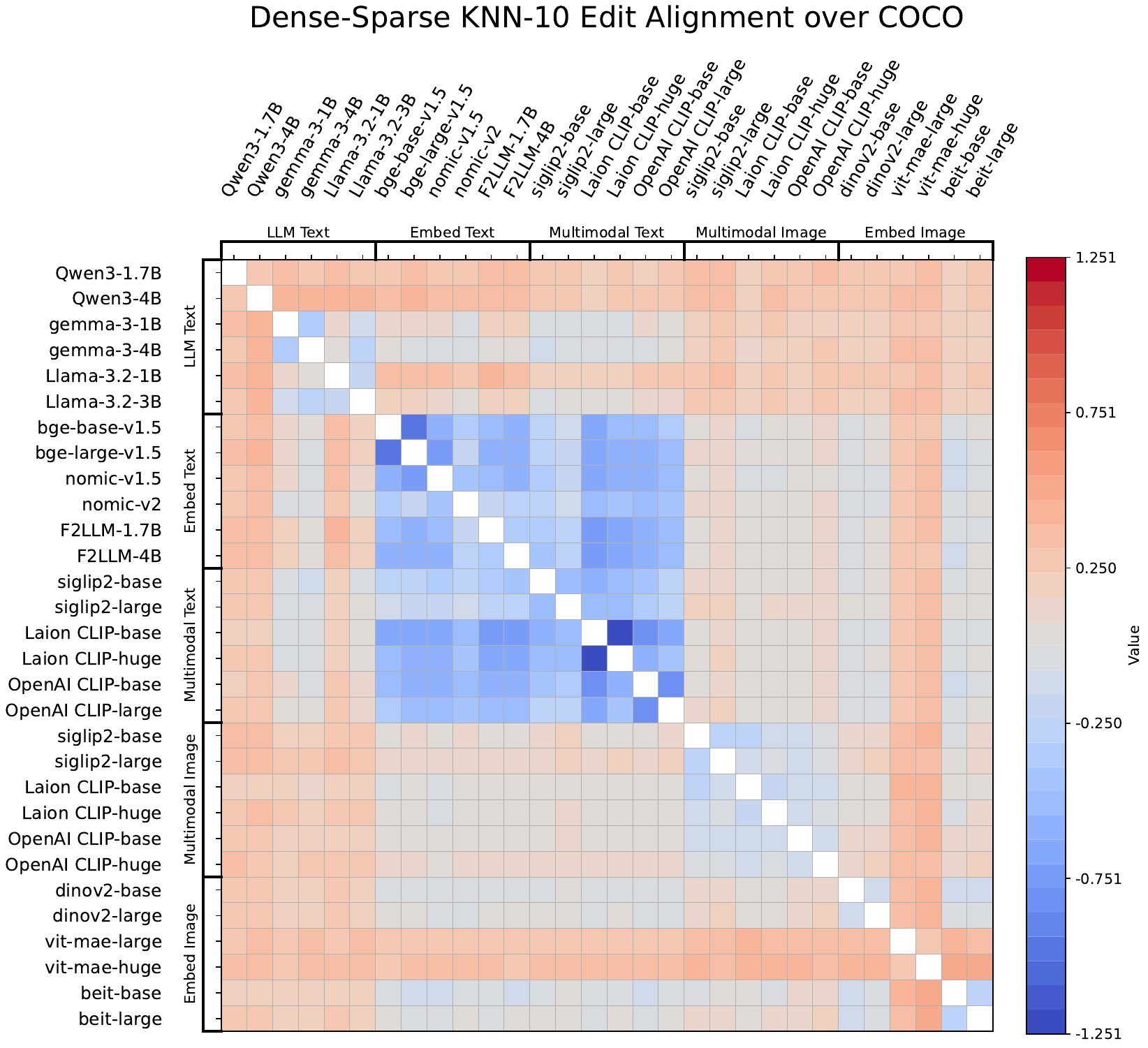}
    \end{minipage}
    \hfill
    \begin{minipage}[t]{0.3\textwidth}
        \centering
        \includegraphics[width = \linewidth]{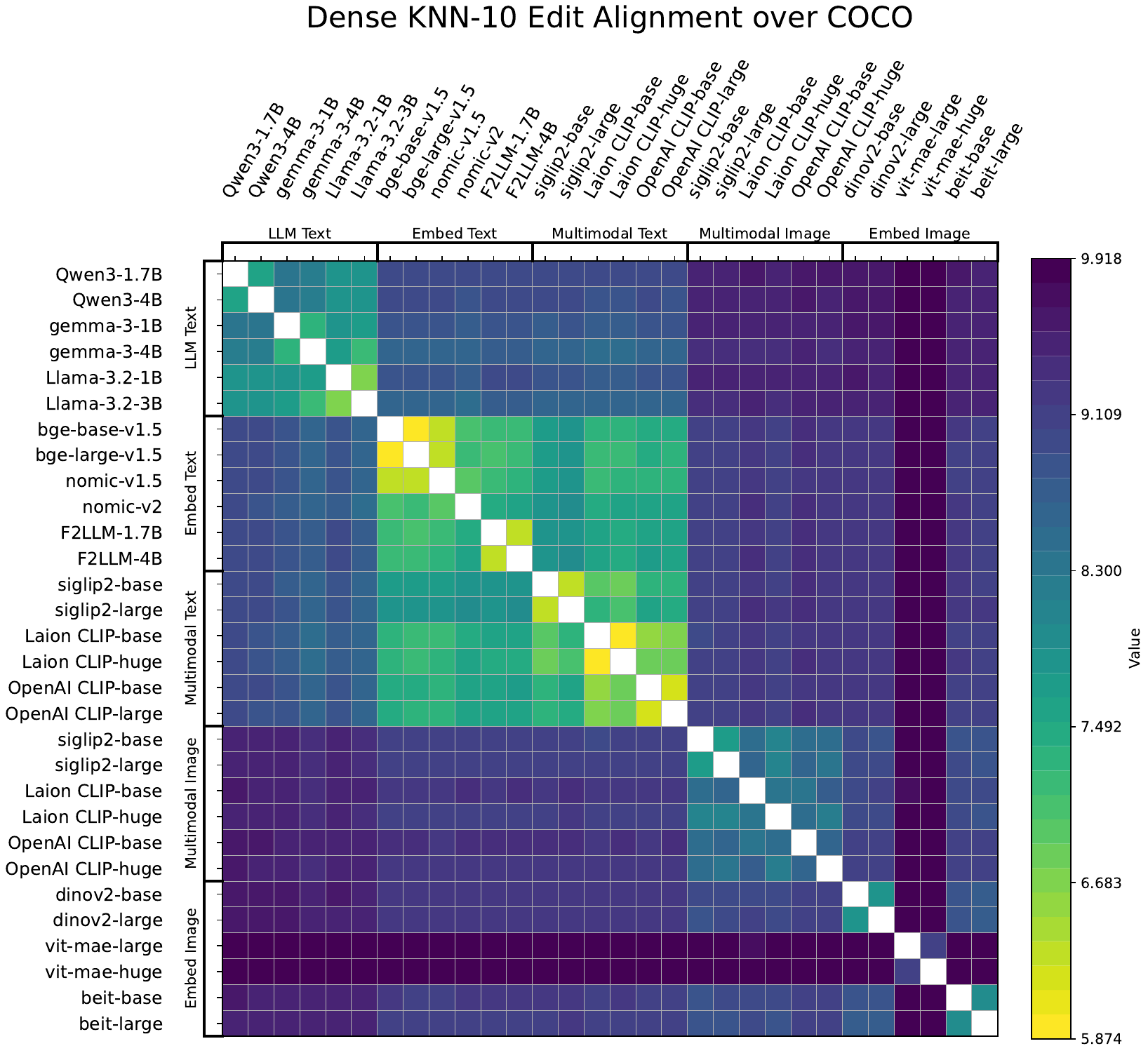}
    \end{minipage}
    \hfill
    \begin{minipage}[t]{0.3\textwidth}
        \centering
        \includegraphics[width = \linewidth]{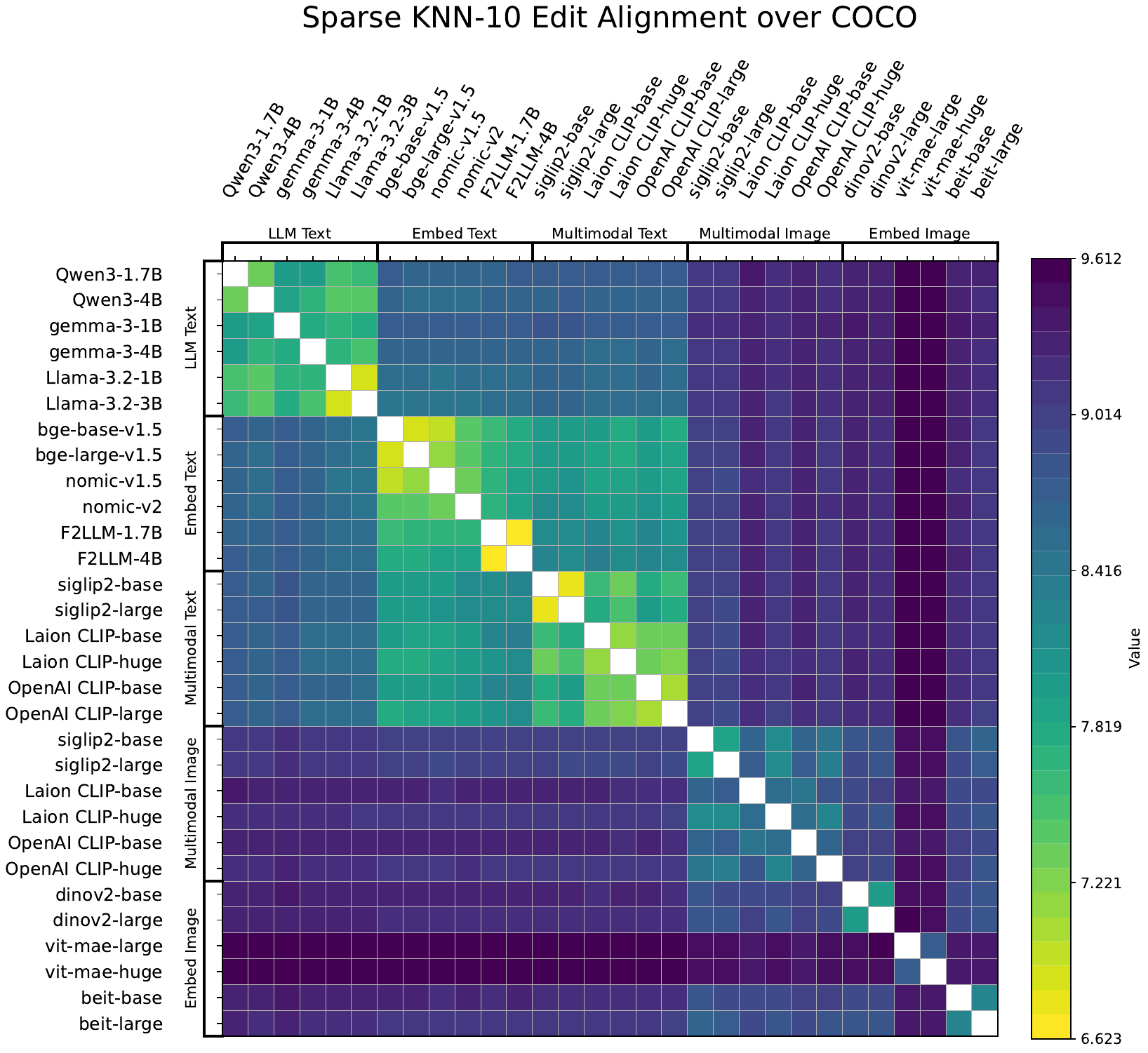}
    \end{minipage}

            \vspace{0.4cm}

    \begin{minipage}[t]{0.3\textwidth}
        \centering
        \includegraphics[width = \linewidth]{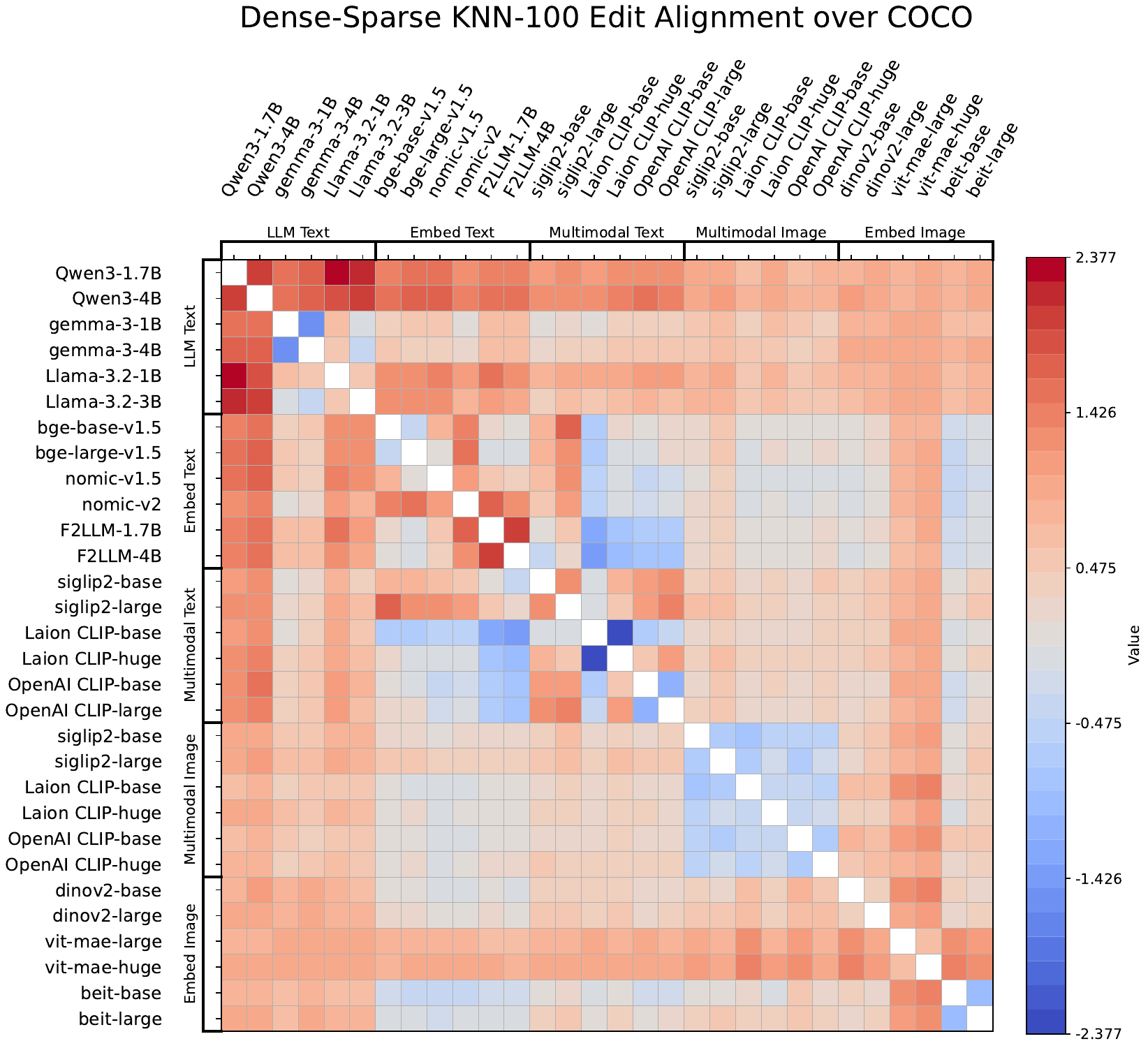}
    \end{minipage}
    \hfill
    \begin{minipage}[t]{0.3\textwidth}
        \centering
        \includegraphics[width = \linewidth]{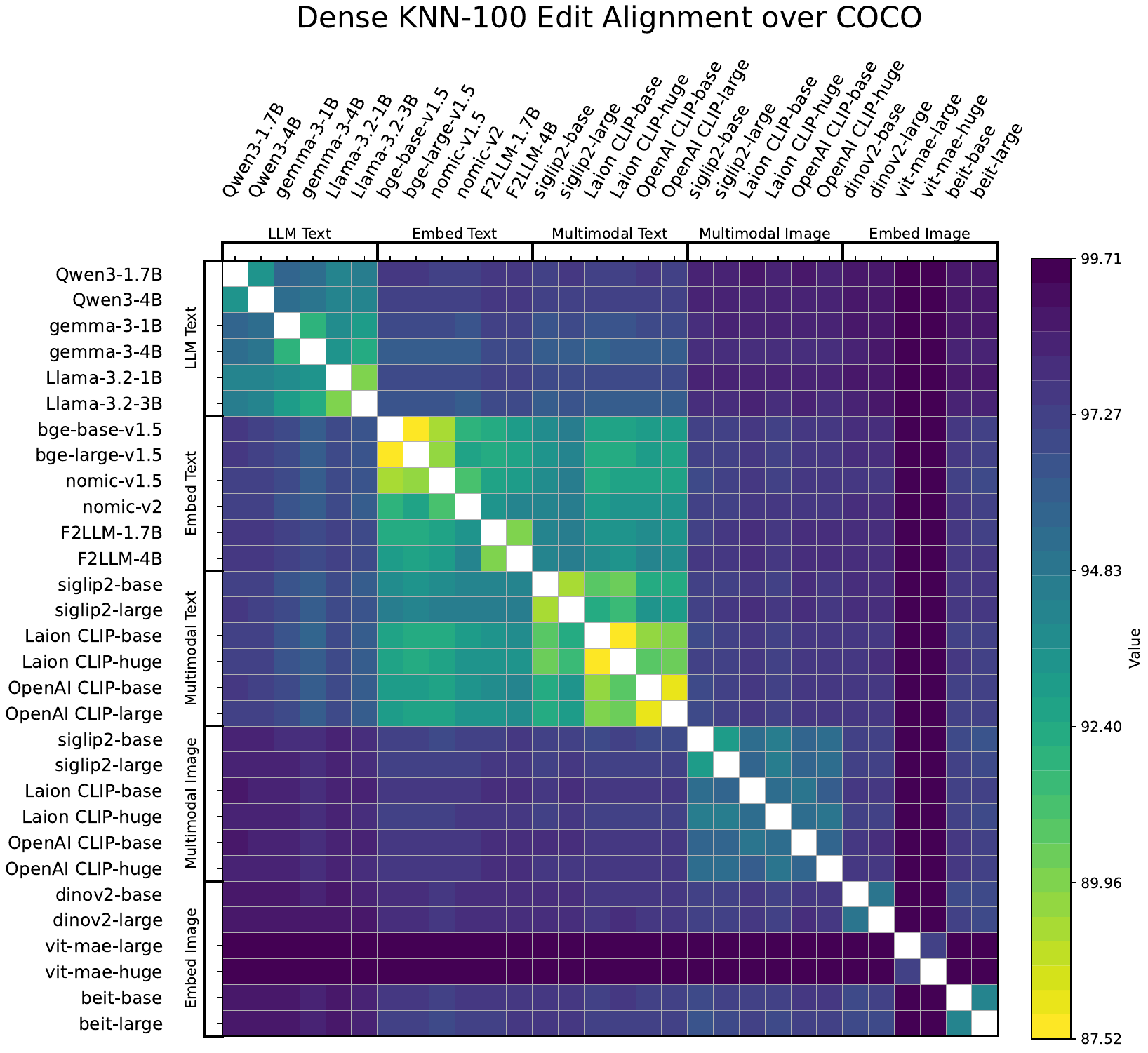}
    \end{minipage}
    \hfill
    \begin{minipage}[t]{0.3\textwidth}
        \centering
        \includegraphics[width = \linewidth]{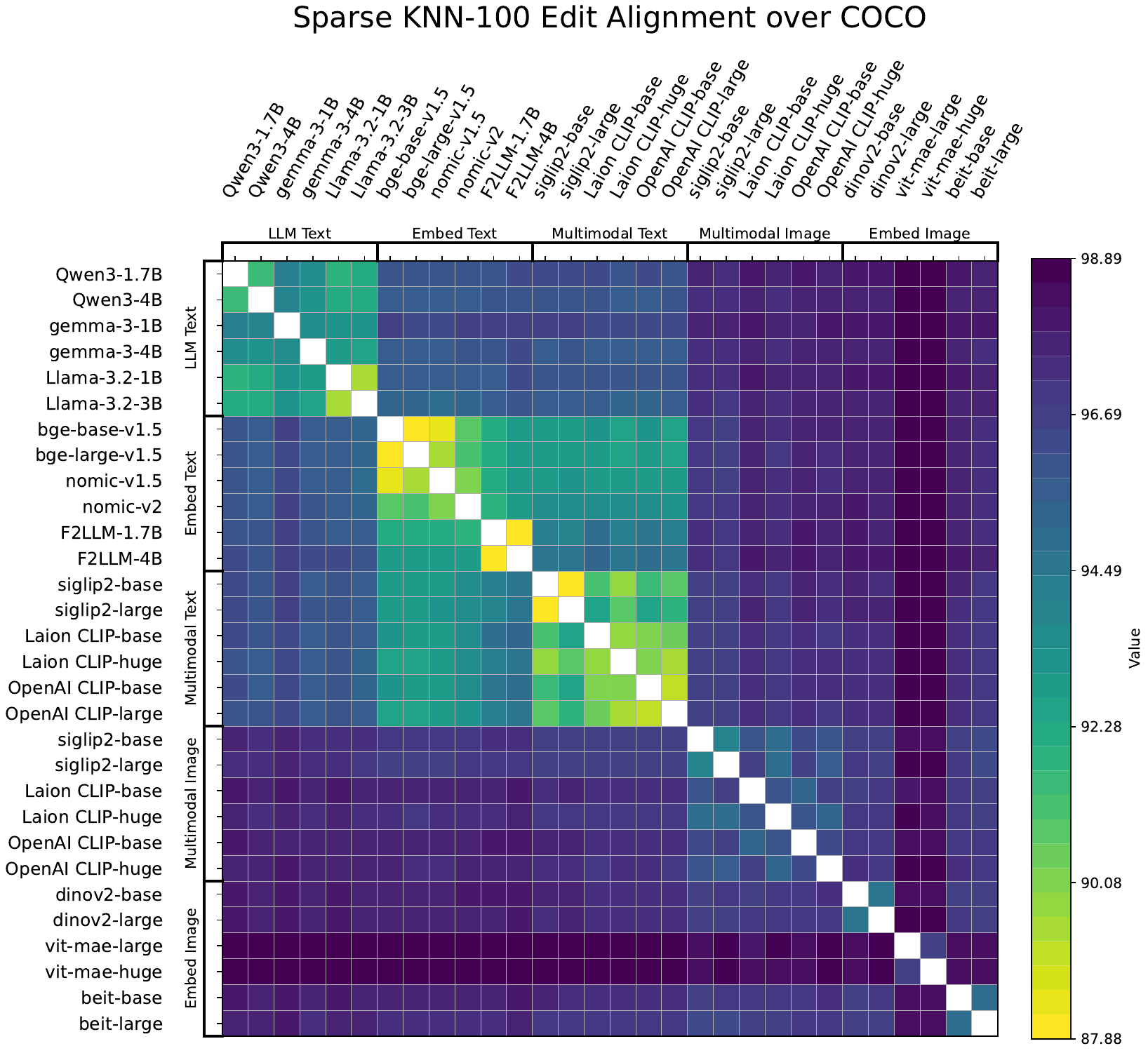}
    \end{minipage}
    \caption{Same plot as in Figure~\ref{fig:signalcoco1} but for the KNN-10 Overlap, KNN-100 Overlap, KNN-10 Edit Distance, and KNN-100 Edit distance metrics. In addition to differences, we plot the alignment values for raw dense features and for the sparse features.}
    \label{fig:signalcoco2}
\end{figure}

We choose not to interpret spectral and geometric metrics for sparse features such as CKA, unbiased CKA, SVCCA. The reason is that sparse features are trained to be non-negative and sparse, both of which induce a strong geometric bias. Instead, ordinal metrics such as KNN overlap and edit distance do not suffer from this issue.

We point out the trend that the KNN-100 overlap similarity consistently decreased with the introduction of sparse features. We attribute this to the local nature of PRH, as discovered in \cite{groger2026prharistotle} and elaborated in our Proposition~\ref{prop:aristotlethm}. When $k = 100,$ for subsets of only $n = 1000$ samples, the KNN graph captures a rather global geometry. We also comment on the KNN-10 overlap similarity in Figure~\ref{fig:signalcoco2} that regions where the improvement of alignment by the introduction of sparse features is low mostly correspond to regions where the alignment is already extremely high -- between LLM models (rows 1-6) and between (multimodal) text embeddings (rows 7 - 18).   

\subsubsection{Experiments on CC3M}

\begin{figure}[htbp]
    \centering

    \begin{minipage}[t]{0.3\textwidth}
        \centering
        \includegraphics[width = \linewidth]{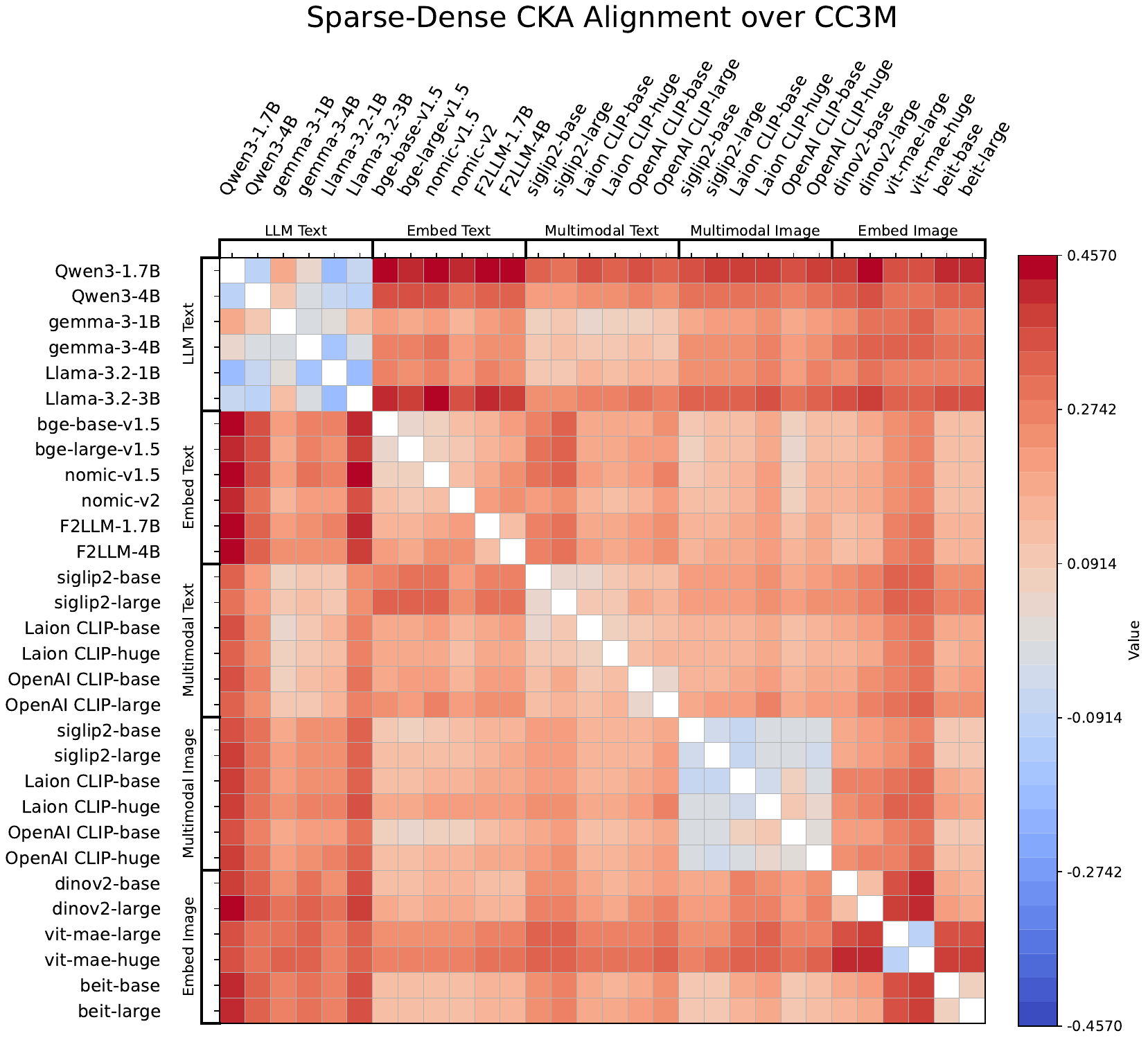}
    \end{minipage}
    \hfill
    \begin{minipage}[t]{0.3\textwidth}
        \centering
        \includegraphics[width = \linewidth]{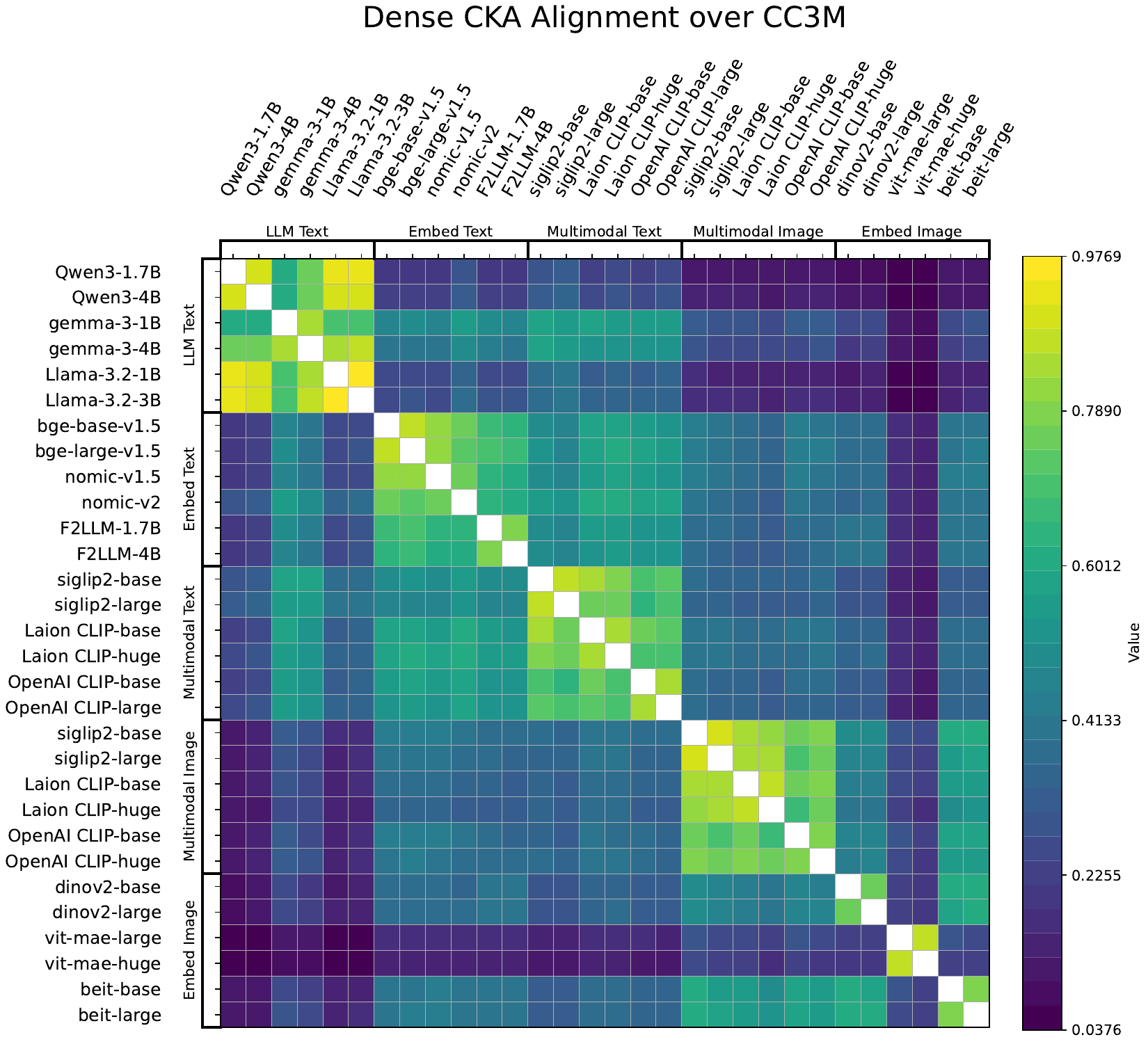}
    \end{minipage}
    \hfill
    \begin{minipage}[t]{0.3\textwidth}
        \centering
        \includegraphics[width = \linewidth]{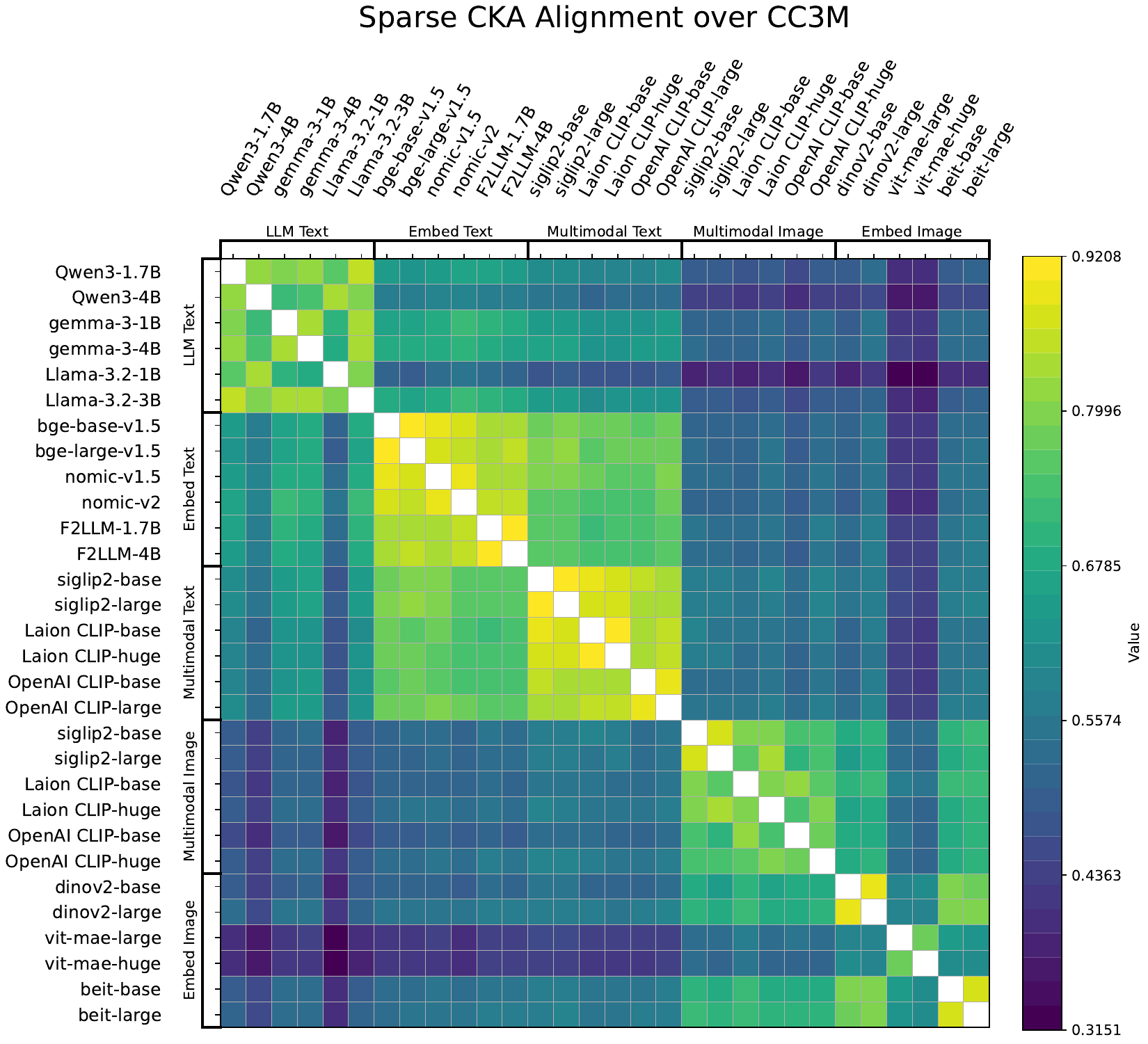}
    \end{minipage}
    
    \vspace{0.4cm}

        \begin{minipage}[t]{0.3\textwidth}
        \centering
        \includegraphics[width = \linewidth]{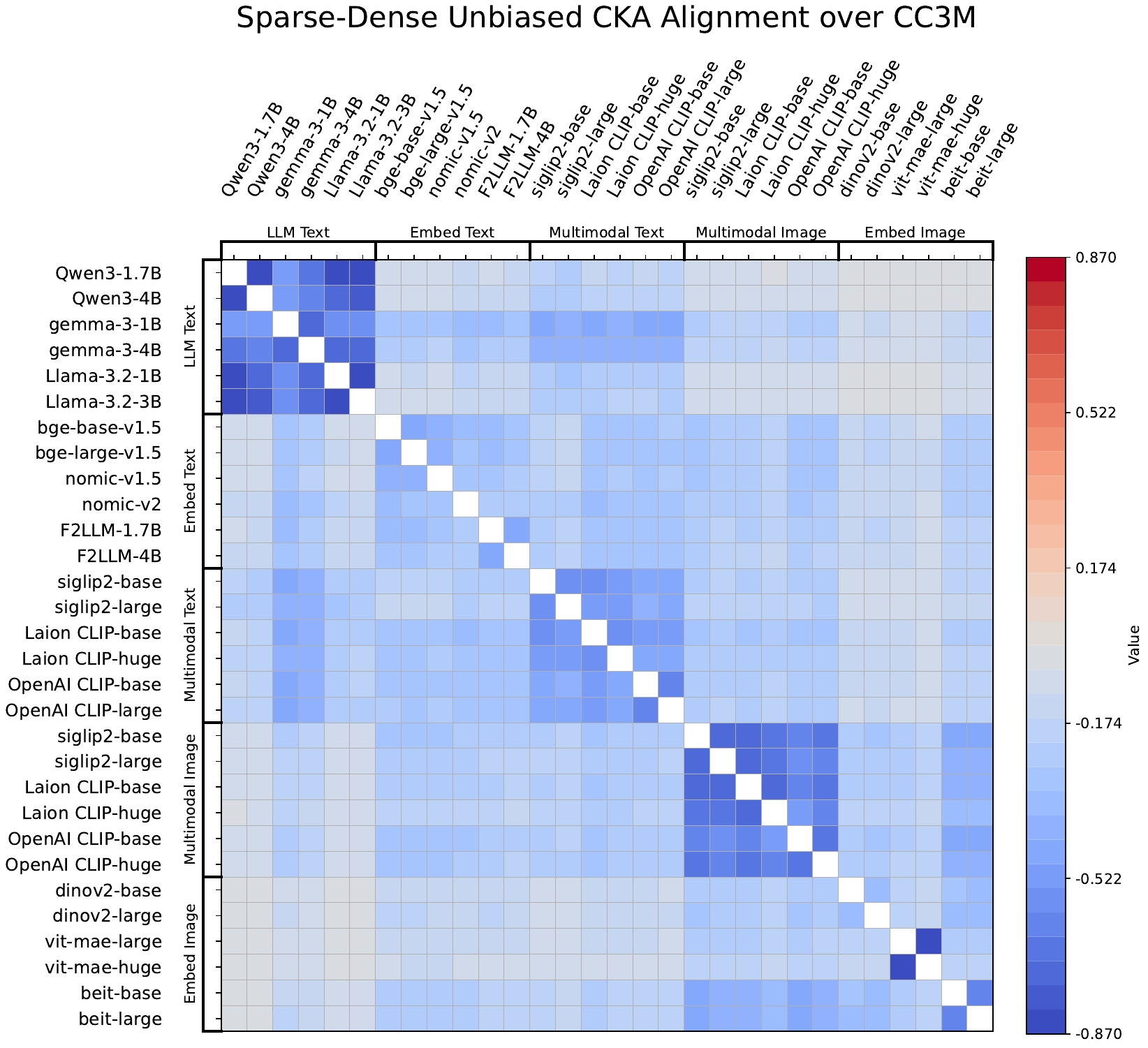}
    \end{minipage}
    \hfill
    \begin{minipage}[t]{0.3\textwidth}
        \centering
        \includegraphics[width = \linewidth]{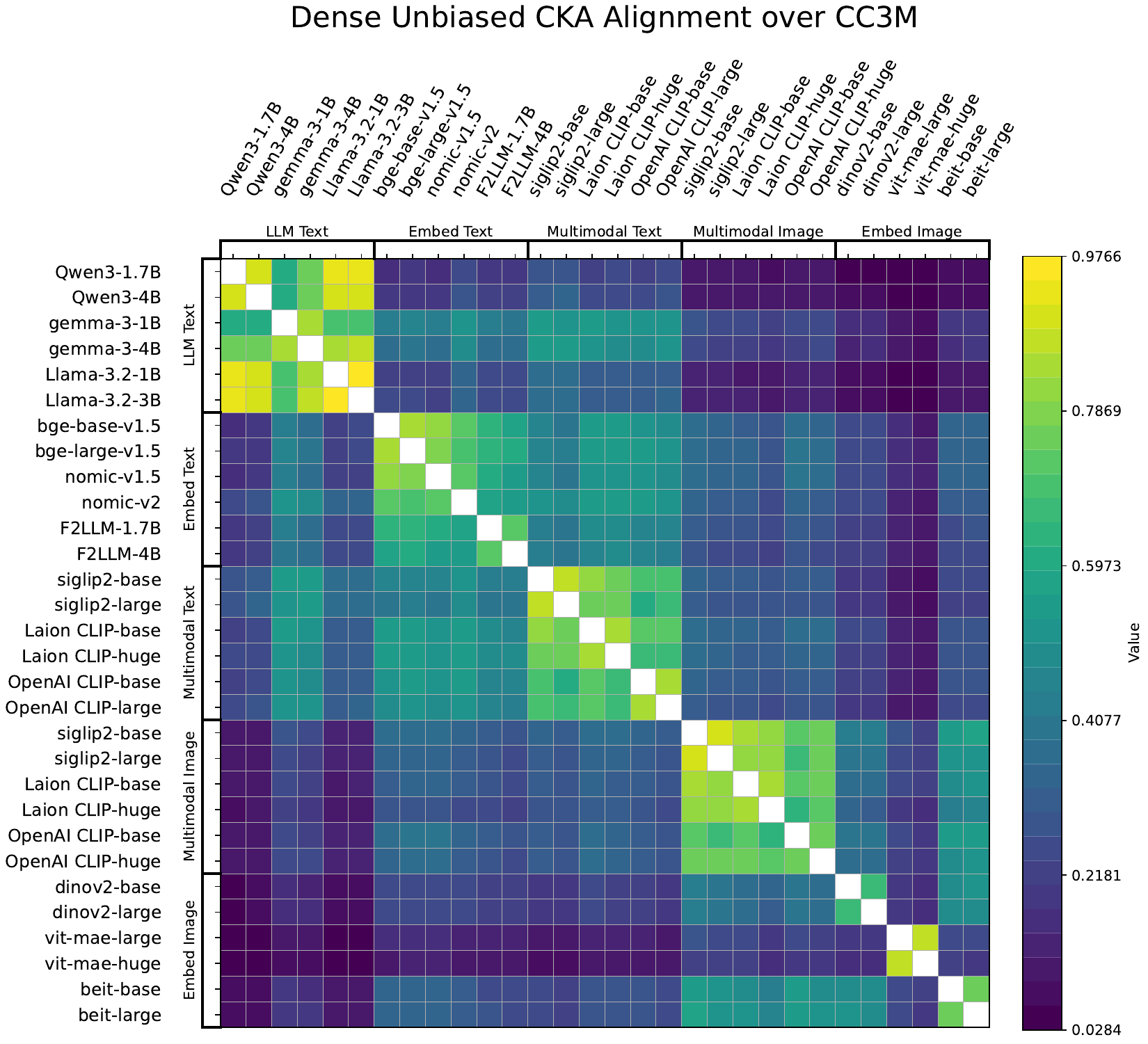}
    \end{minipage}
    \hfill
    \begin{minipage}[t]{0.3\textwidth}
        \centering
        \includegraphics[width = \linewidth]{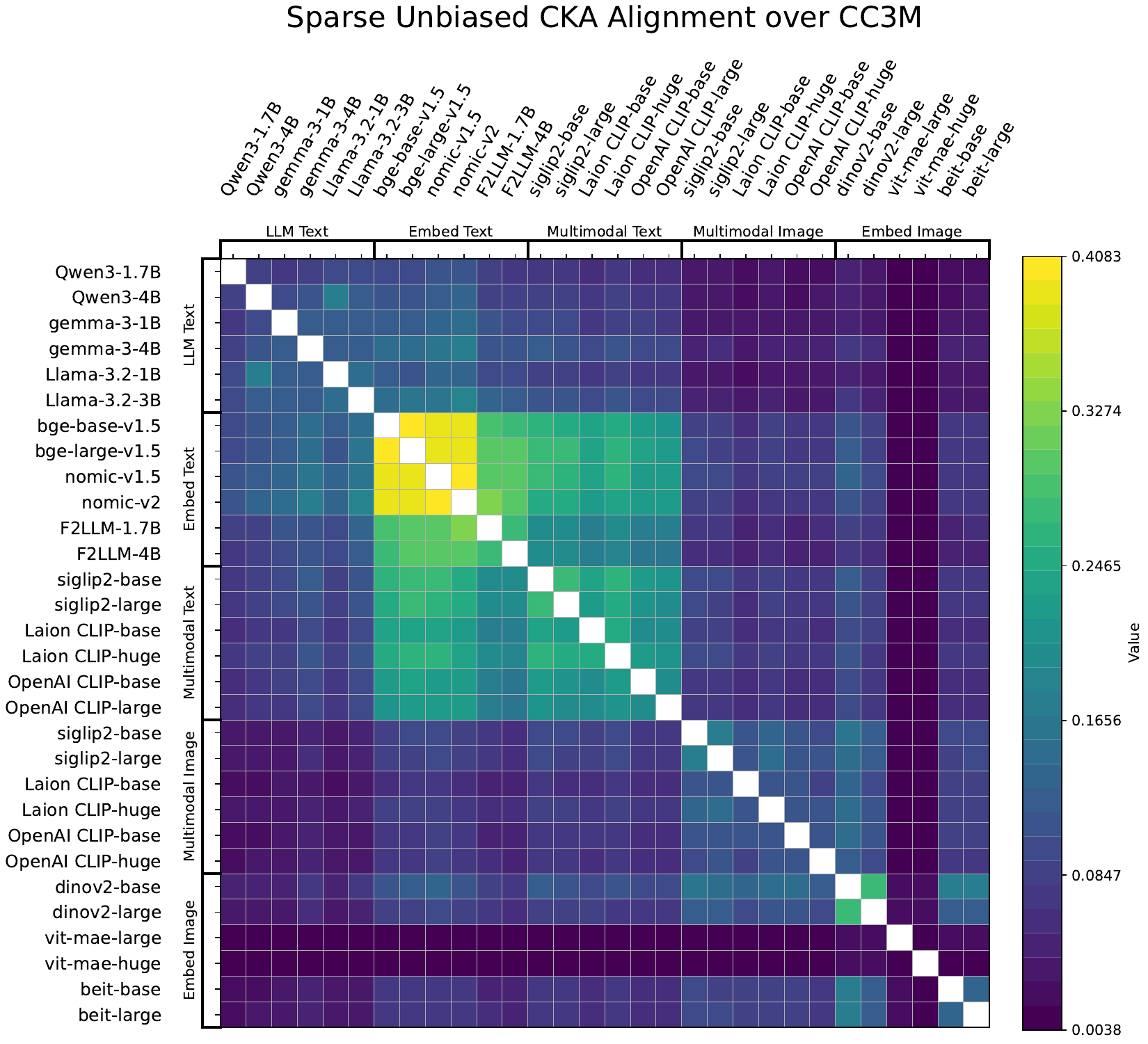}
    \end{minipage}

        \vspace{0.4cm}

        \begin{minipage}[t]{0.3\textwidth}
        \centering
        \includegraphics[width = \linewidth]{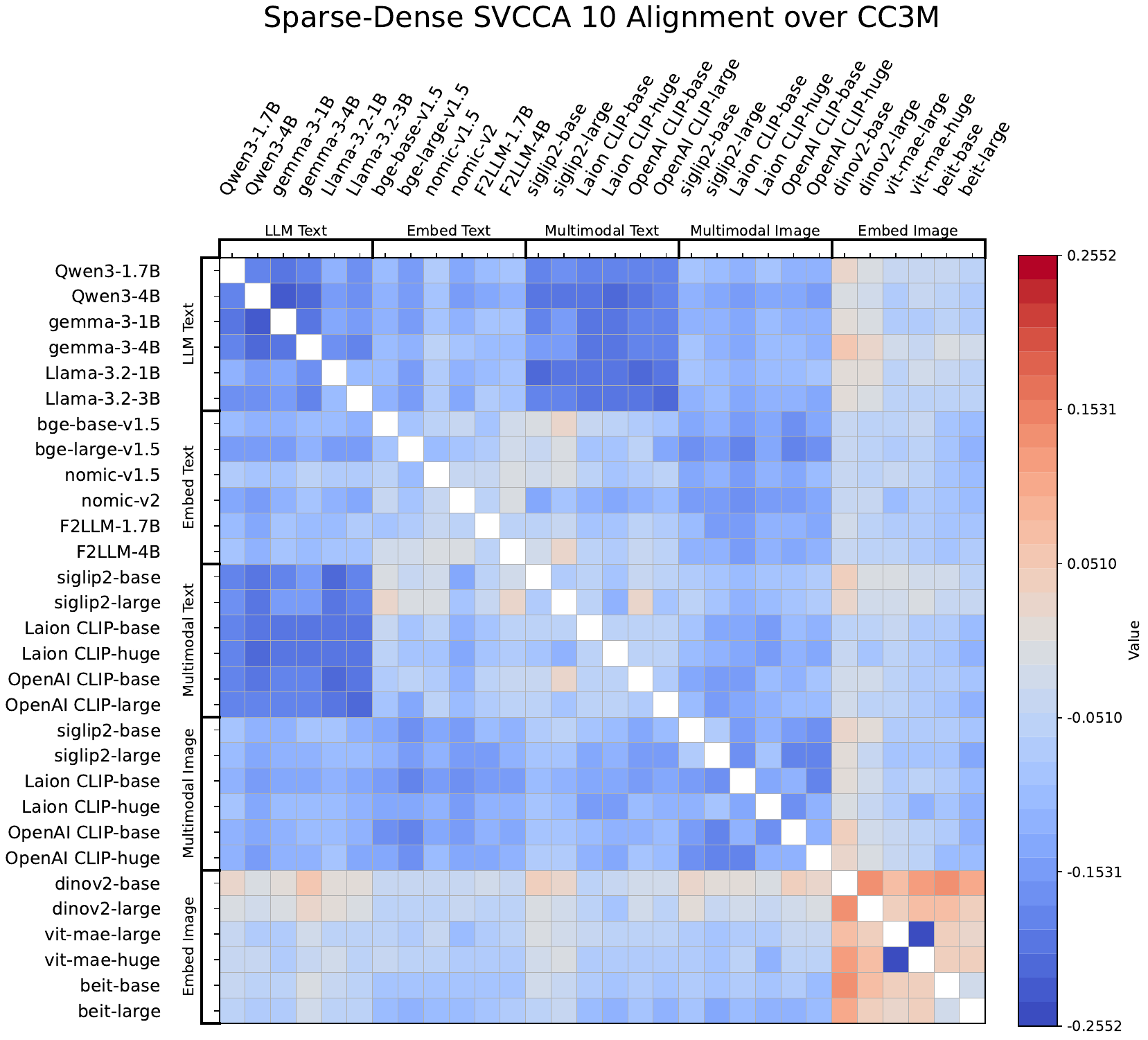}
    \end{minipage}
    \hfill
    \begin{minipage}[t]{0.3\textwidth}
        \centering
        \includegraphics[width = \linewidth]{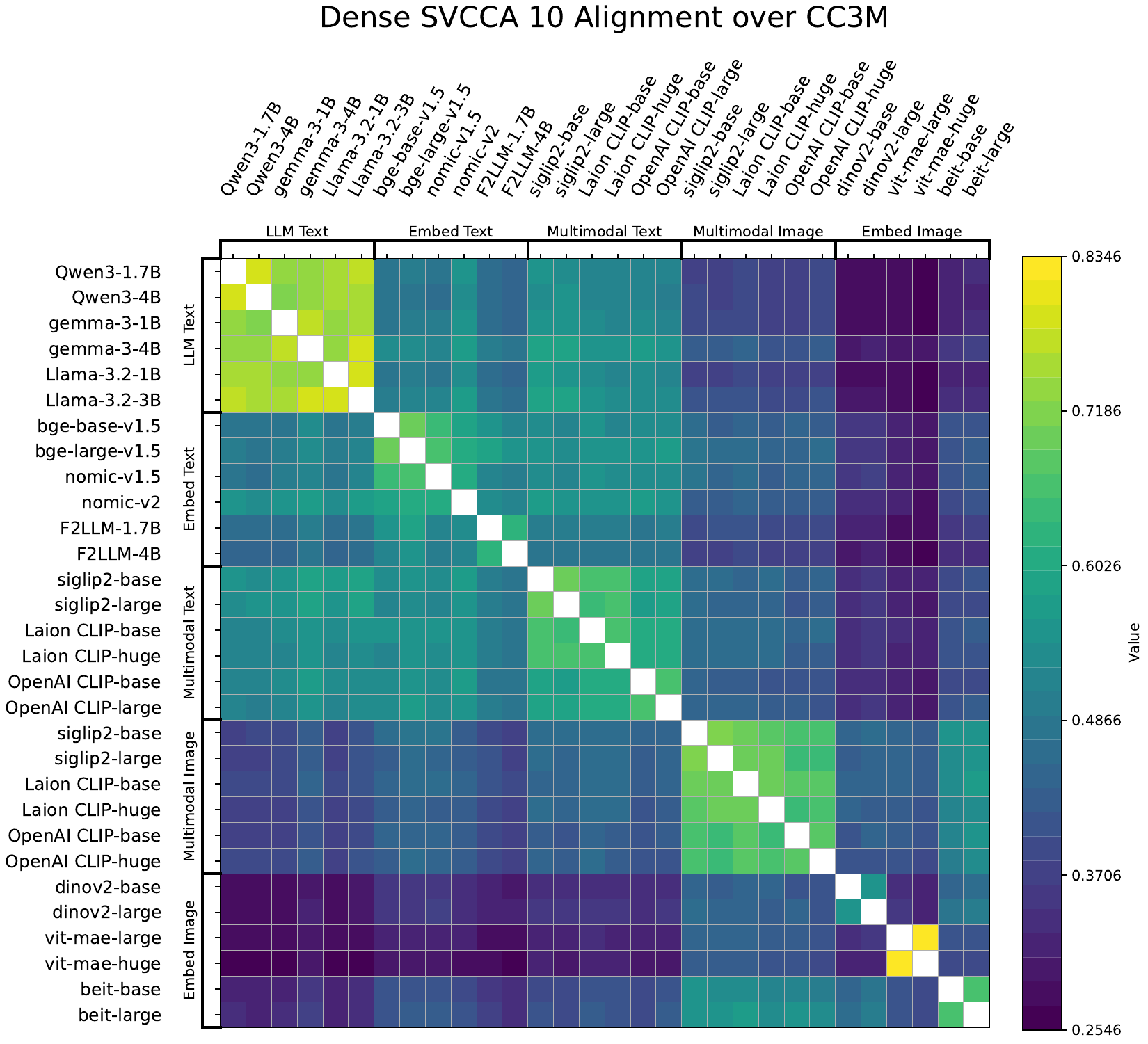}
    \end{minipage}
    \hfill
    \begin{minipage}[t]{0.3\textwidth}
        \centering
        \includegraphics[width = \linewidth]{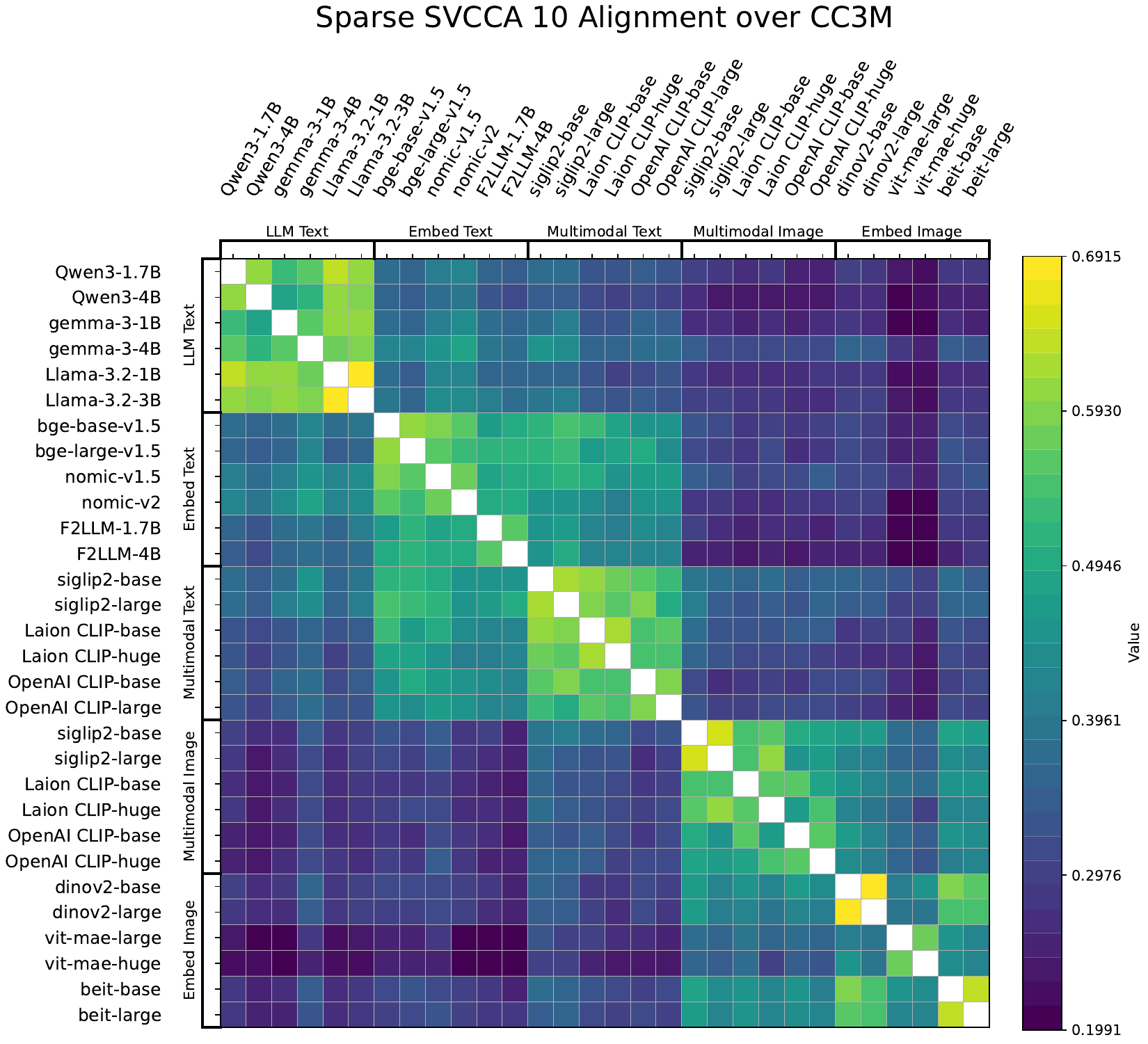}
    \end{minipage}

            \vspace{0.4cm}

    \begin{minipage}[t]{0.3\textwidth}
        \centering
        \includegraphics[width = \linewidth]{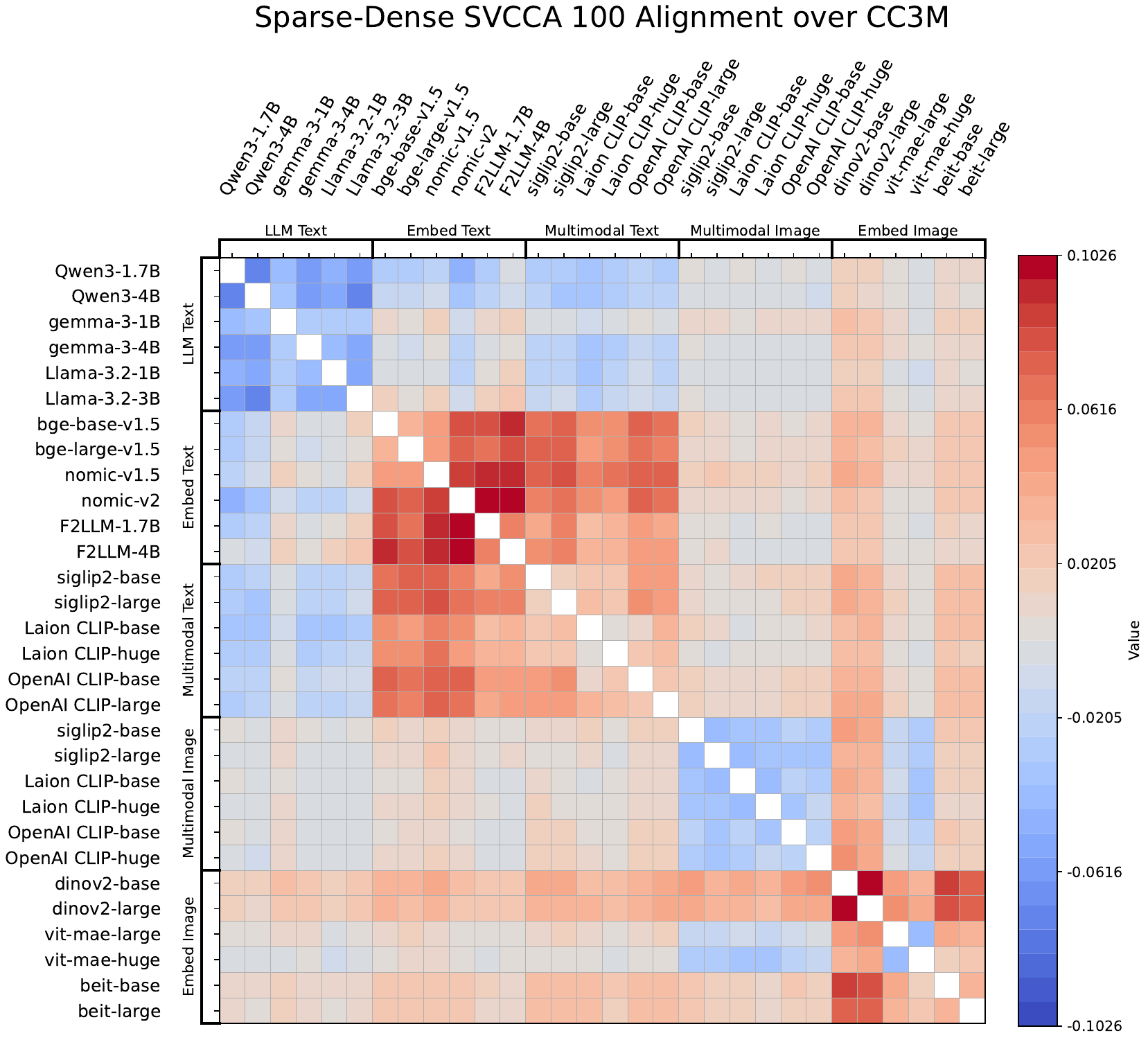}
    \end{minipage}
    \hfill
    \begin{minipage}[t]{0.3\textwidth}
        \centering
        \includegraphics[width = \linewidth]{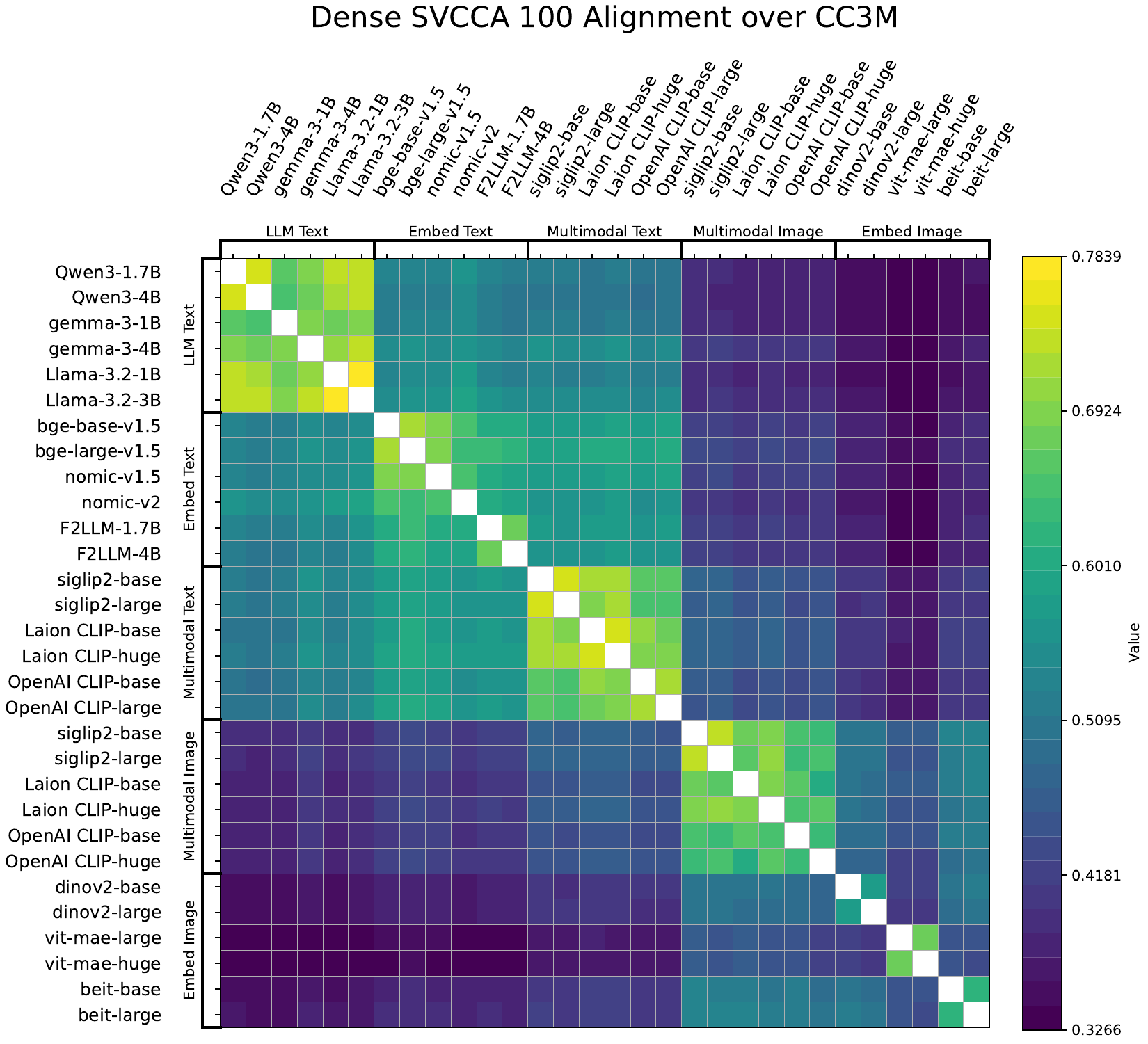}
    \end{minipage}
    \hfill
    \begin{minipage}[t]{0.3\textwidth}
        \centering
        \includegraphics[width = \linewidth]{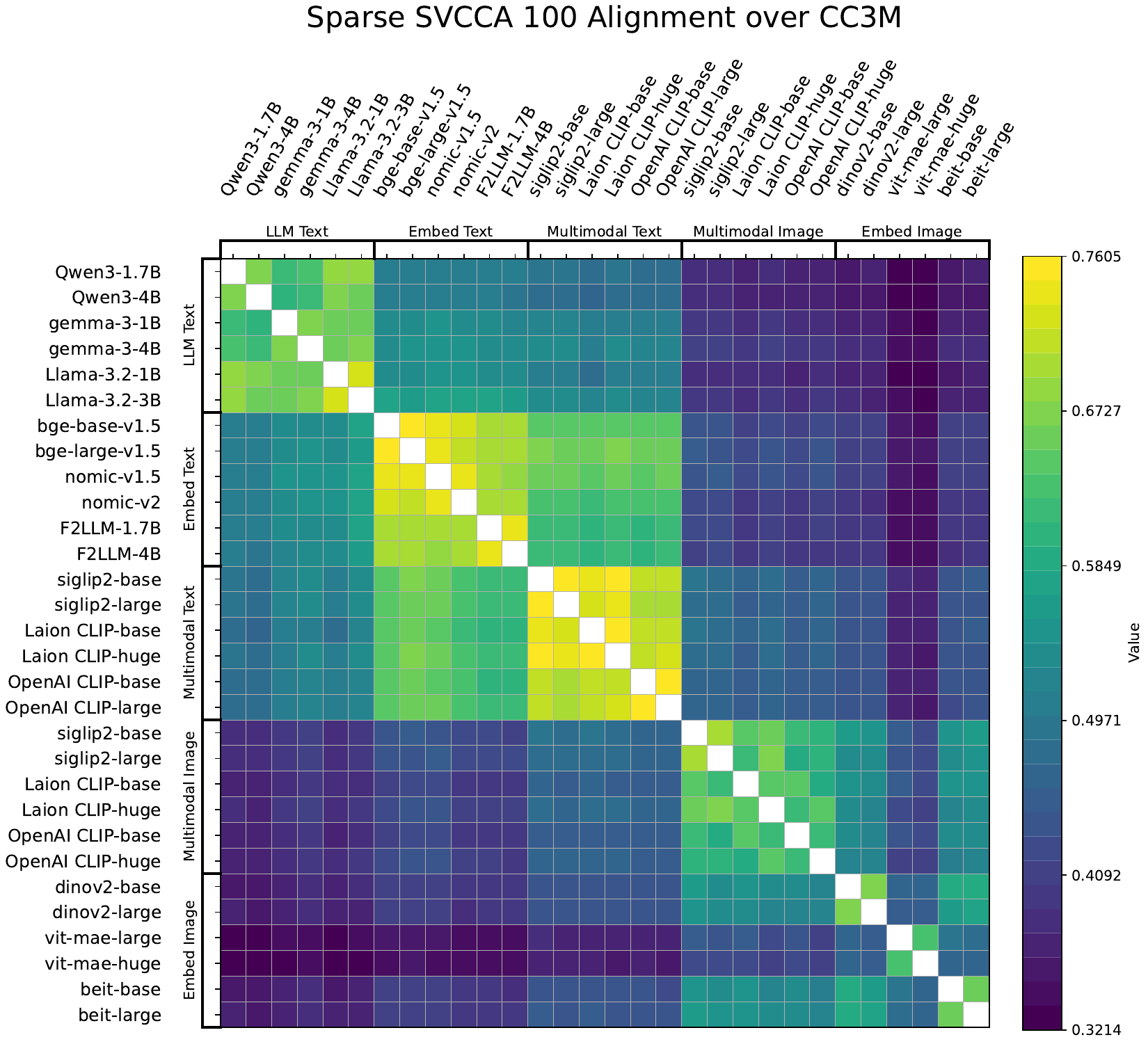}
    \end{minipage}
    \caption{Same plot as in Figure~\ref{fig:signalcoco1} but over CC3M.}
    \label{appendix:signalcc3m1}
\end{figure}

\clearpage

\begin{figure}[htbp]
    \centering

    \begin{minipage}[t]{0.3\textwidth}
        \centering
        \includegraphics[width = \linewidth]{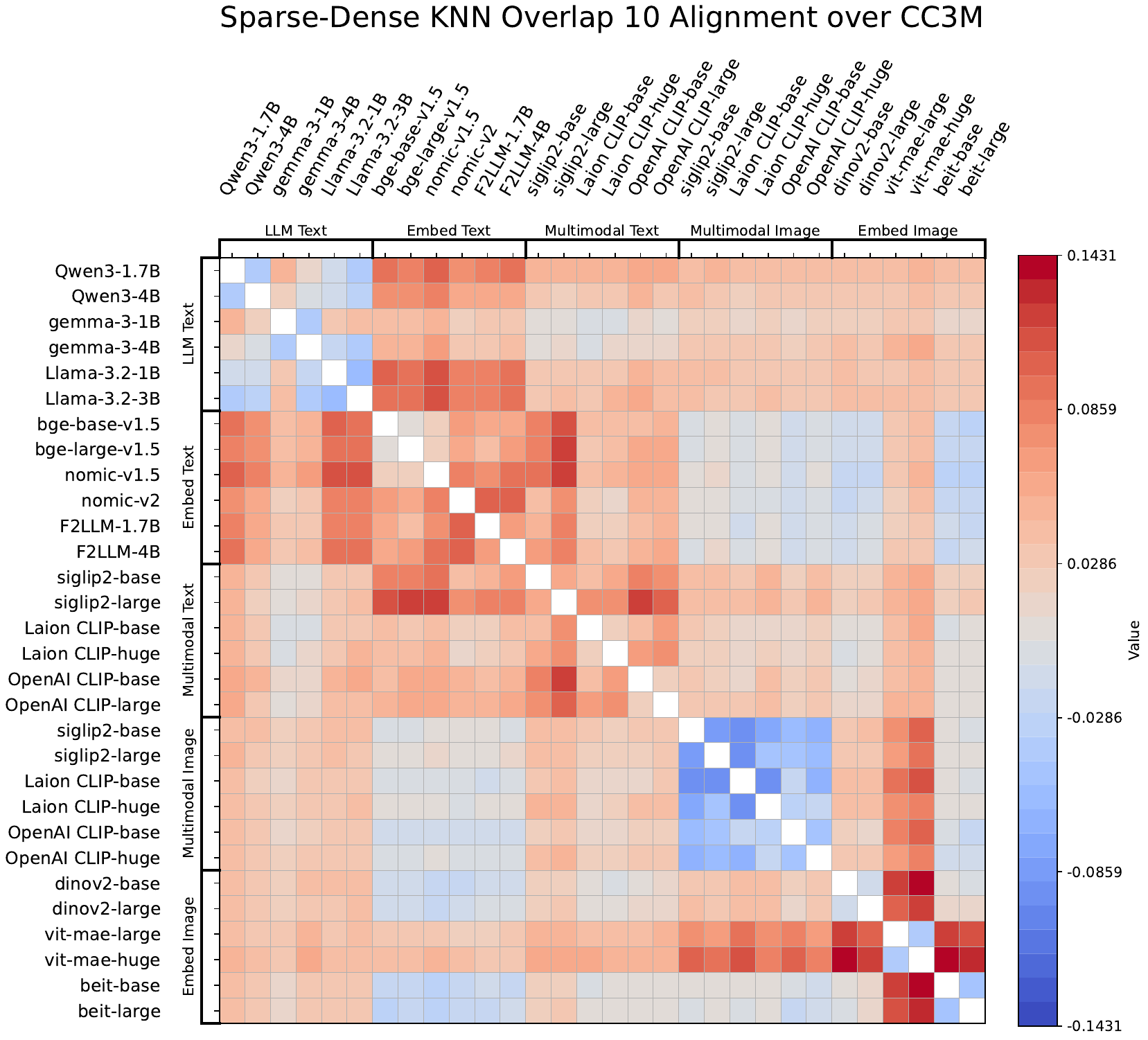}
    \end{minipage}
    \hfill
    \begin{minipage}[t]{0.3\textwidth}
        \centering
        \includegraphics[width = \linewidth]{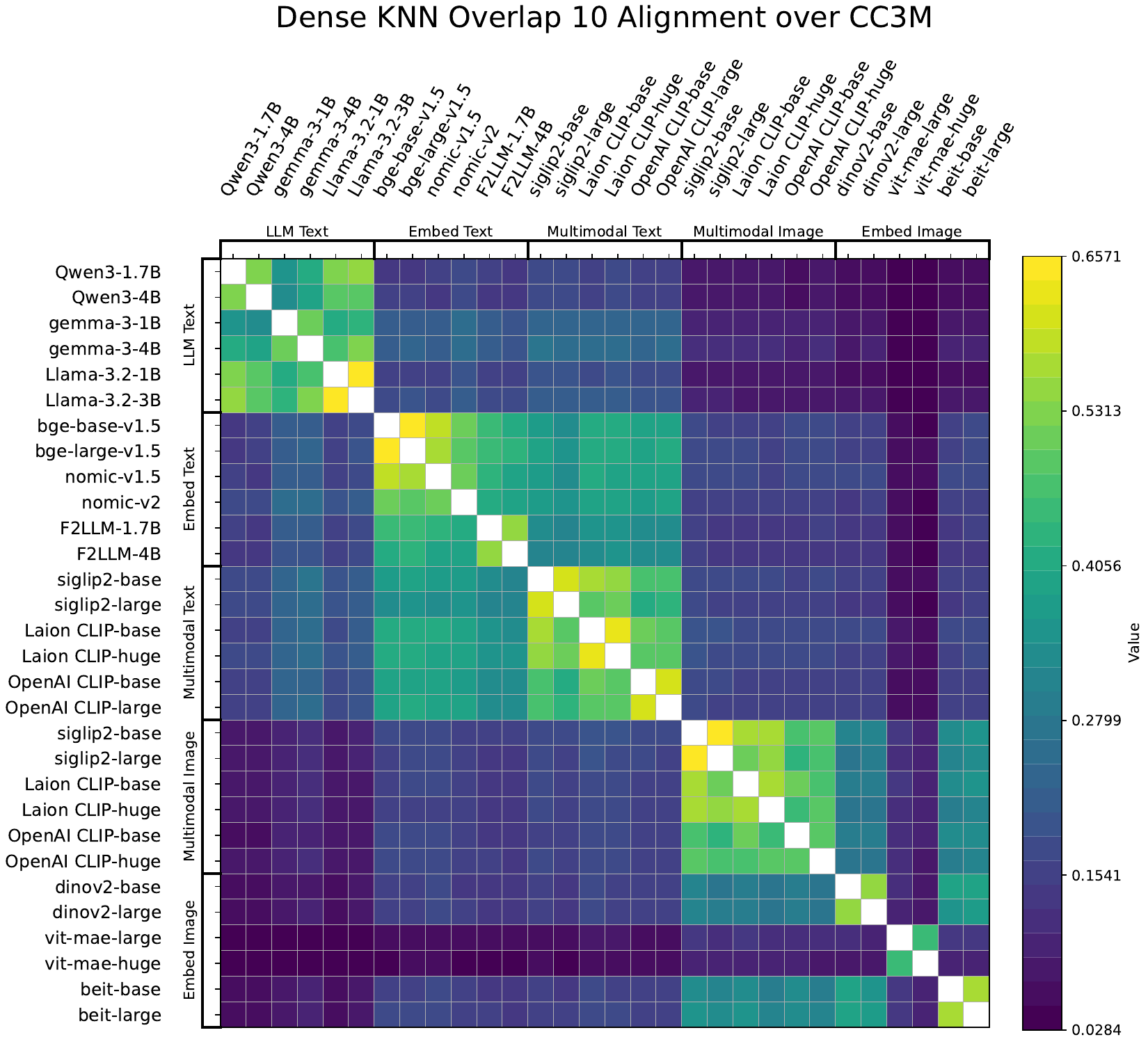}
    \end{minipage}
    \hfill
    \begin{minipage}[t]{0.3\textwidth}
        \centering
        \includegraphics[width = \linewidth]{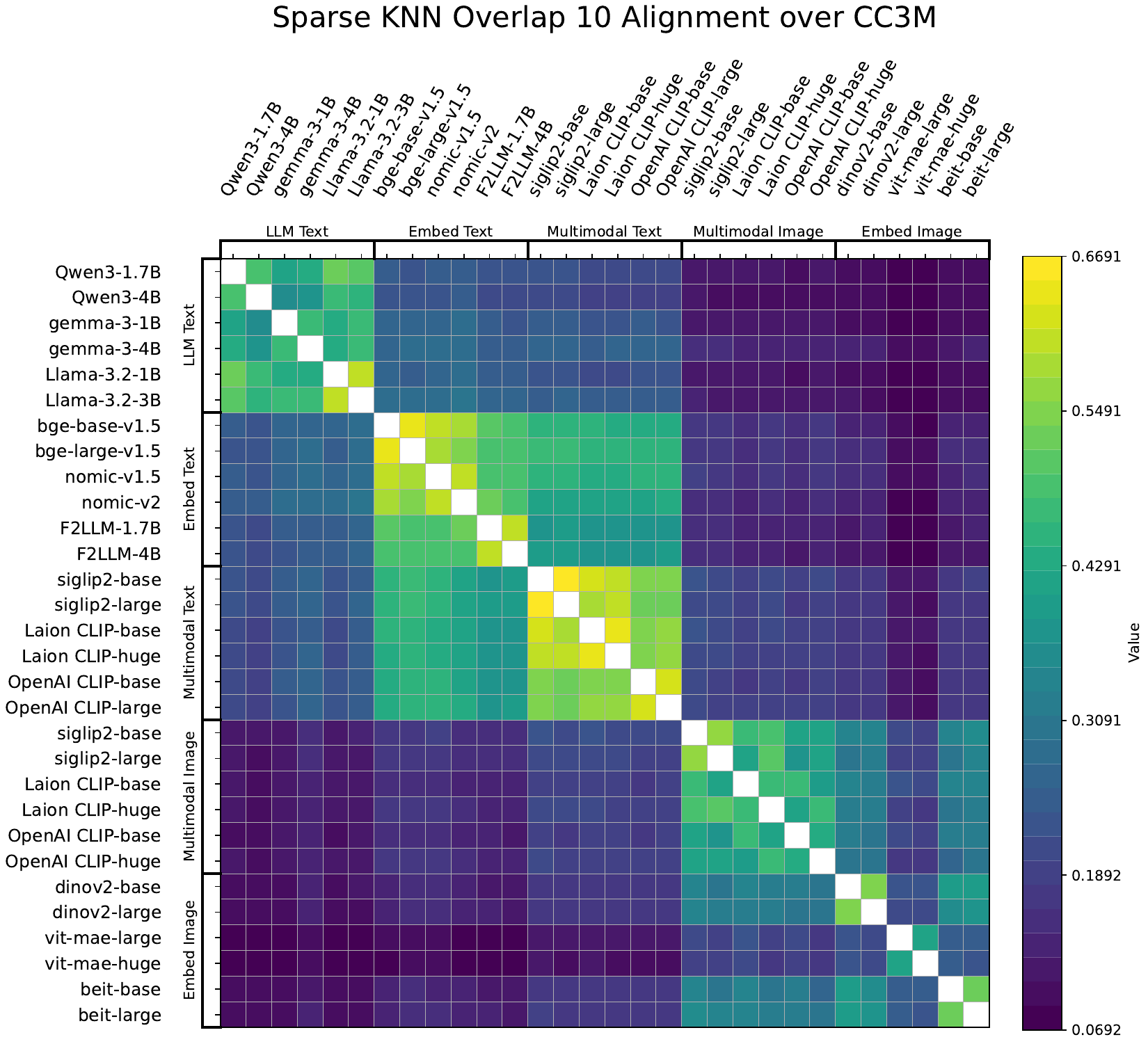}
    \end{minipage}
    
    \vspace{0.4cm}

        \begin{minipage}[t]{0.3\textwidth}
        \centering
        \includegraphics[width = \linewidth]{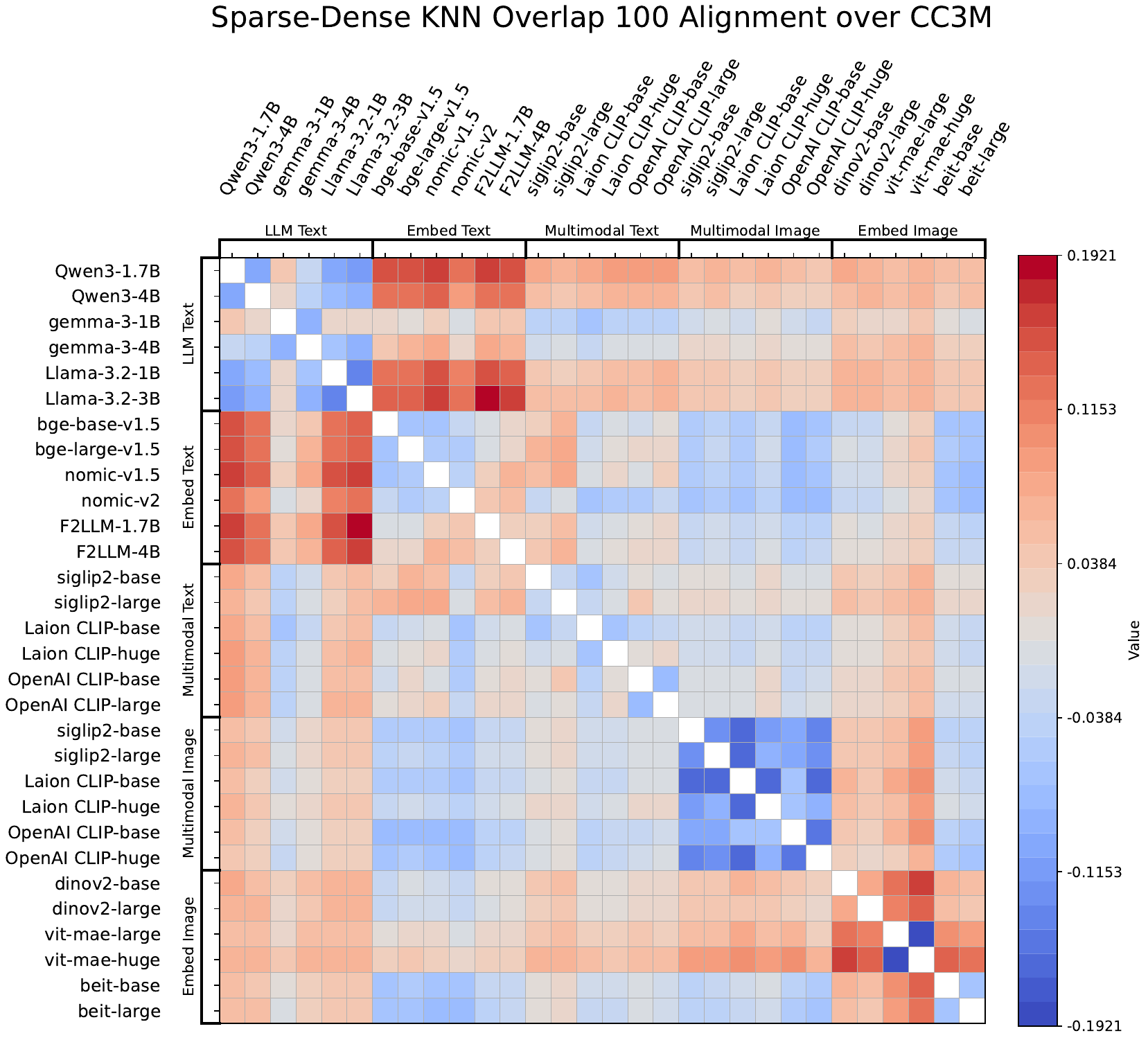}
    \end{minipage}
    \hfill
    \begin{minipage}[t]{0.3\textwidth}
        \centering
        \includegraphics[width = \linewidth]{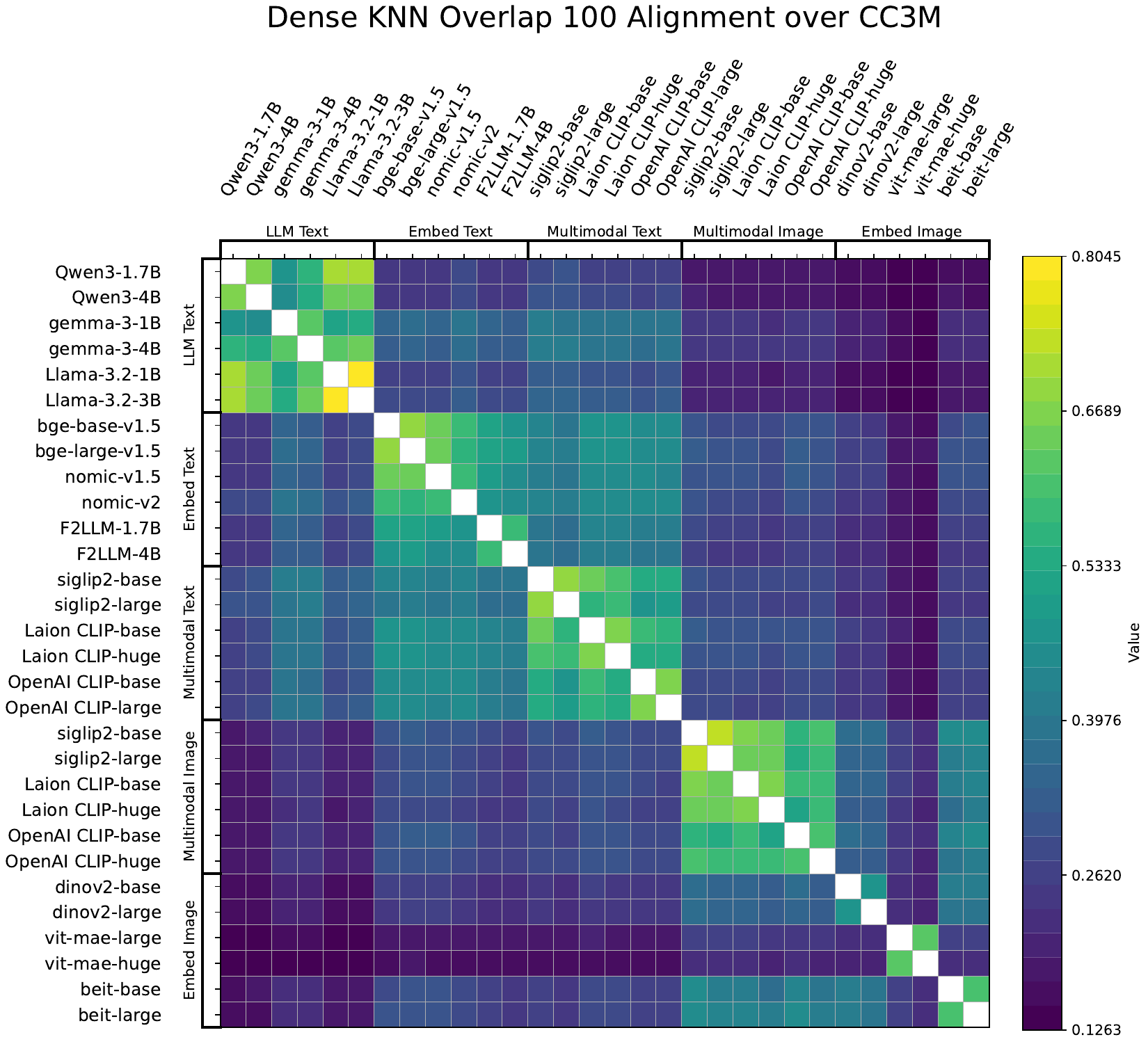}
    \end{minipage}
    \hfill
    \begin{minipage}[t]{0.3\textwidth}
        \centering
        \includegraphics[width = \linewidth]{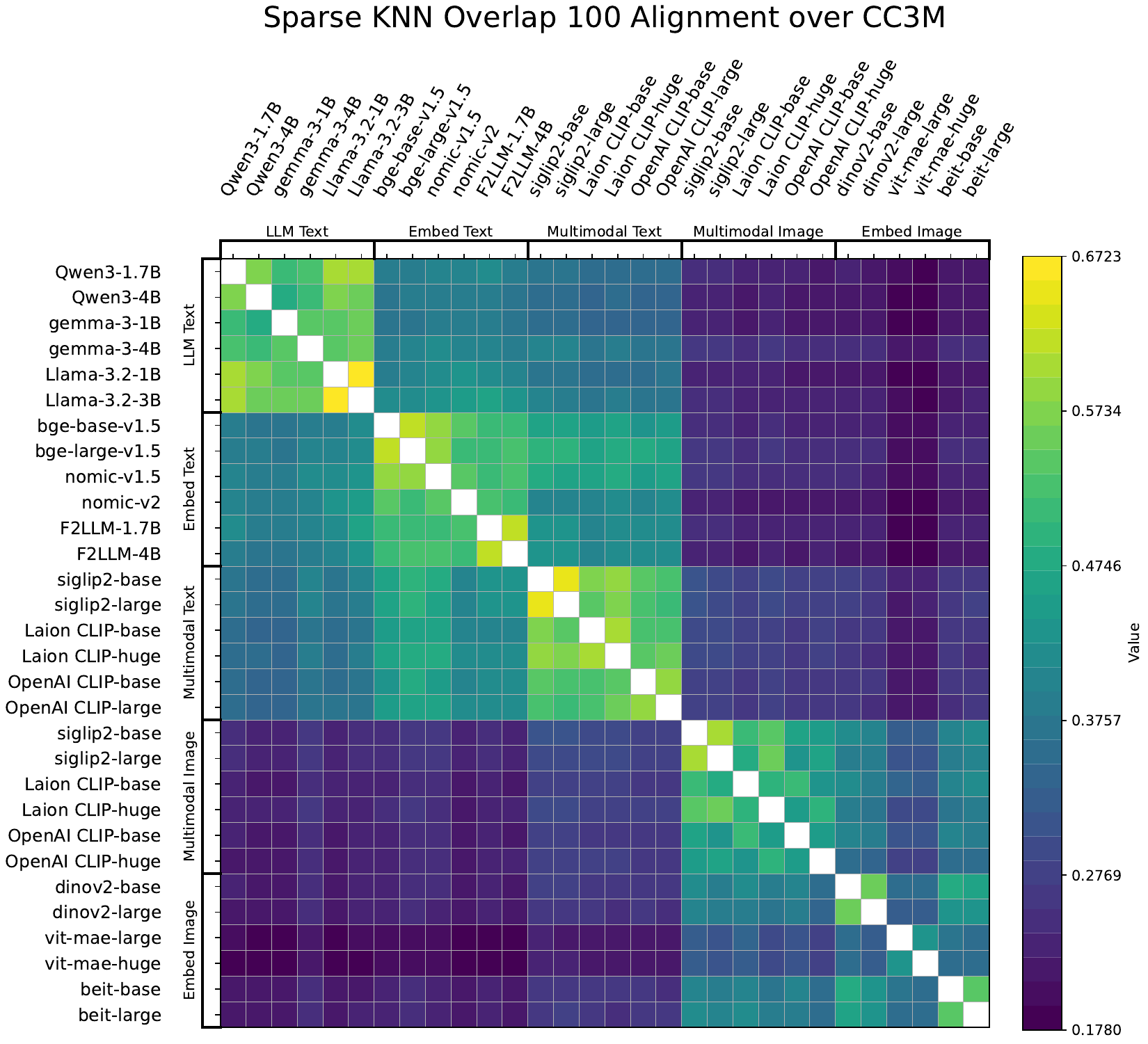}
    \end{minipage}

        \vspace{0.4cm}

        \begin{minipage}[t]{0.3\textwidth}
        \centering
        \includegraphics[width = \linewidth]{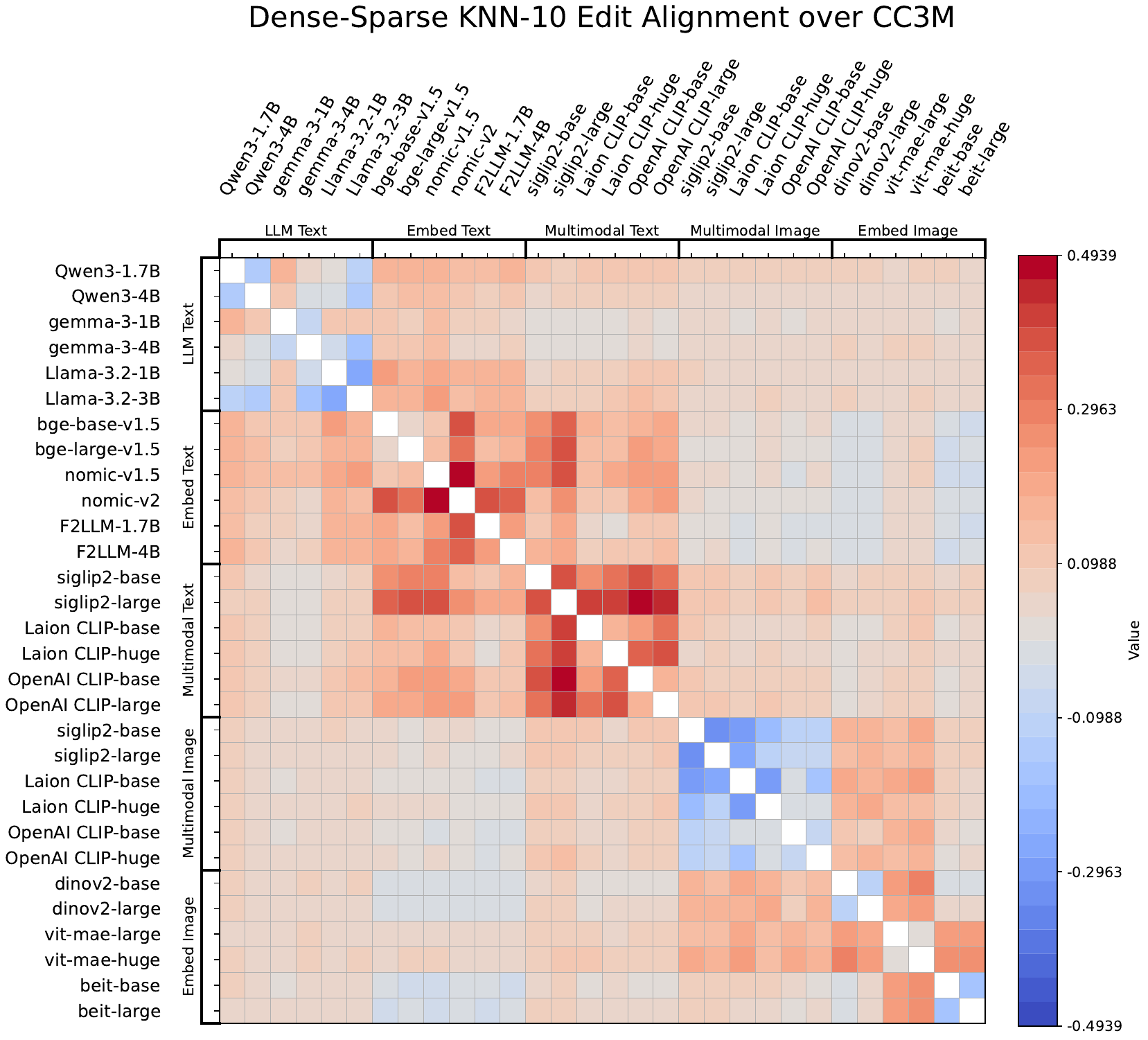}
    \end{minipage}
    \hfill
    \begin{minipage}[t]{0.3\textwidth}
        \centering
        \includegraphics[width = \linewidth]{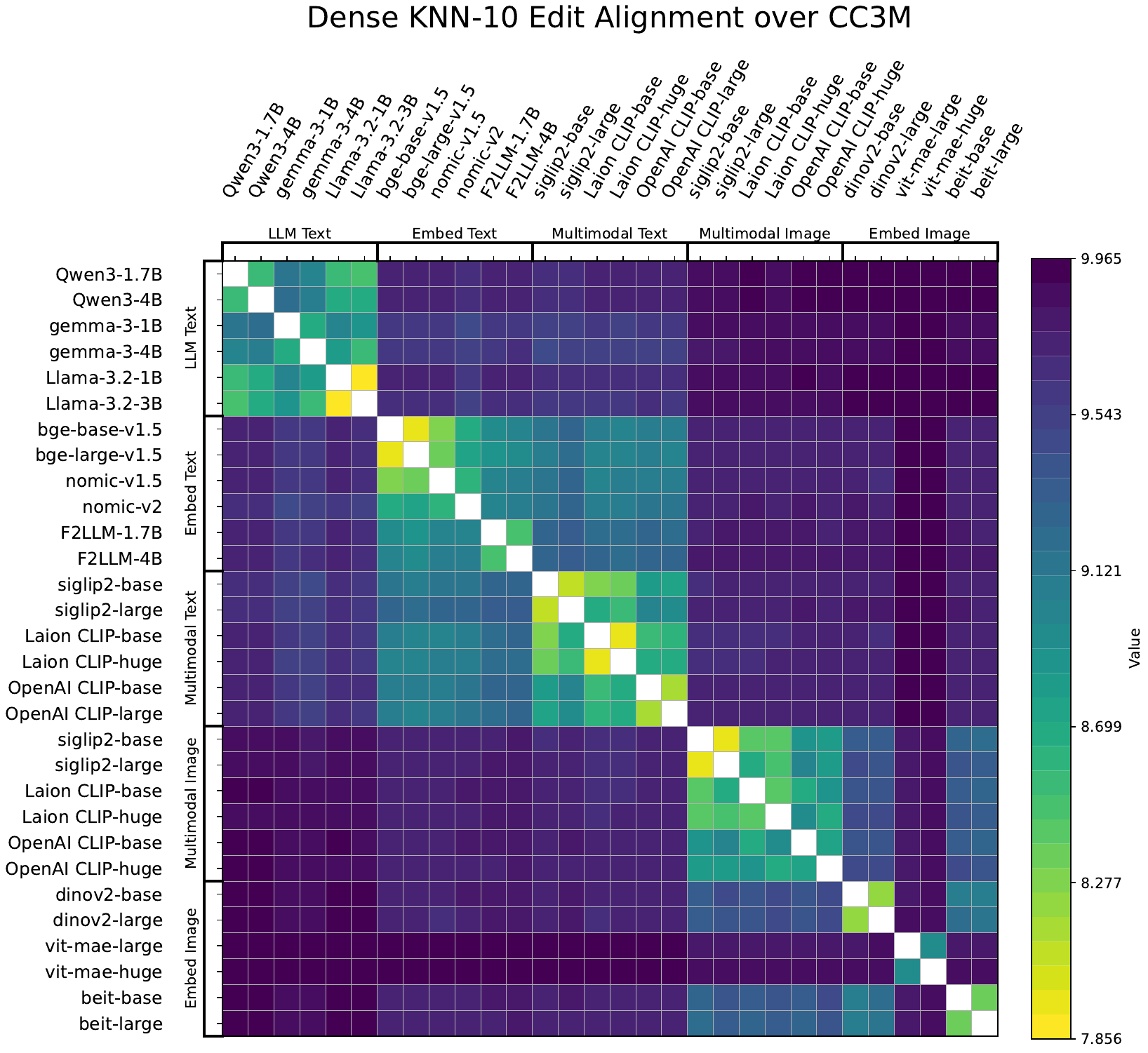}
    \end{minipage}
    \hfill
    \begin{minipage}[t]{0.3\textwidth}
        \centering
        \includegraphics[width = \linewidth]{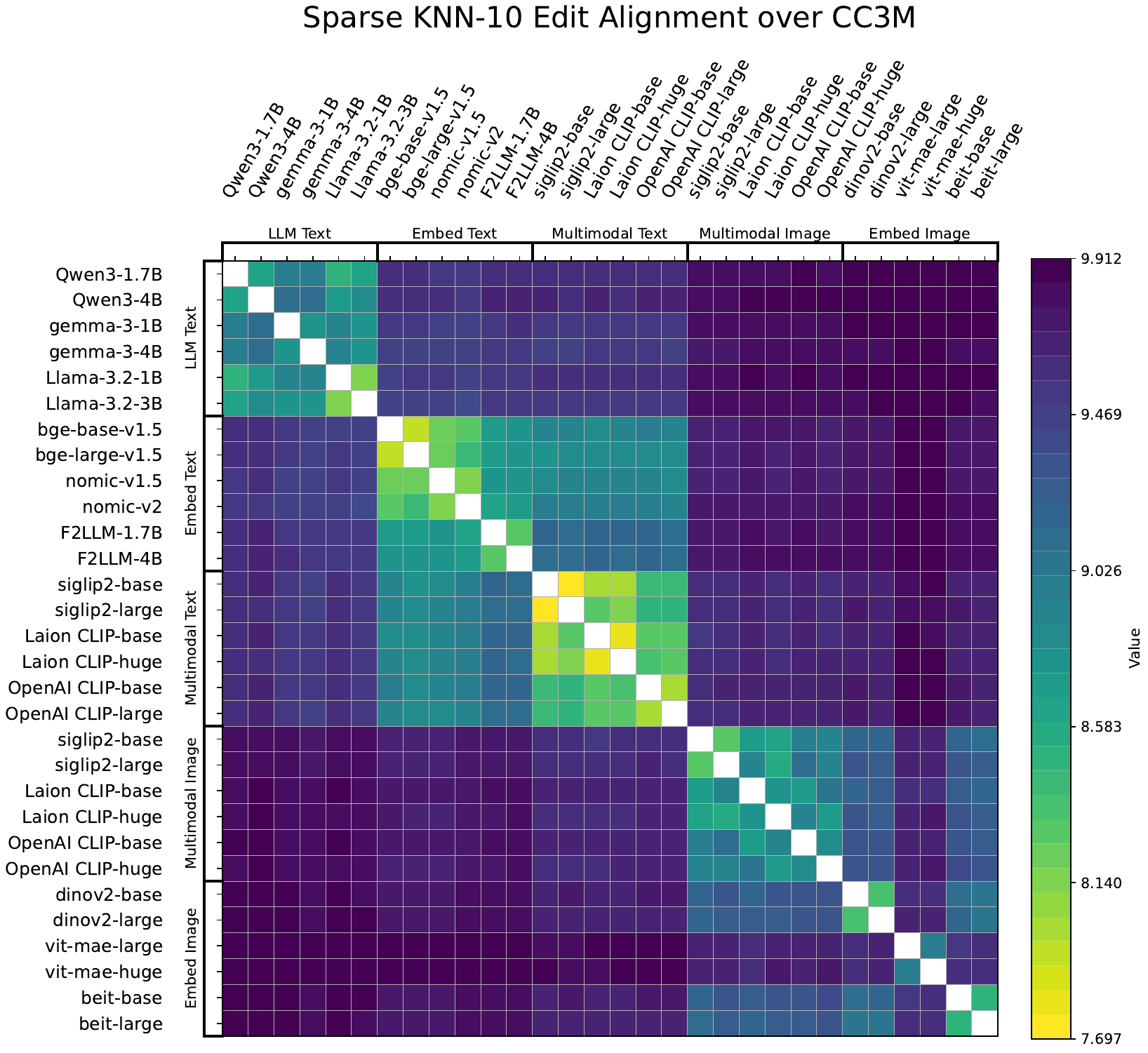}
    \end{minipage}

            \vspace{0.4cm}

    \begin{minipage}[t]{0.3\textwidth}
        \centering
        \includegraphics[width = \linewidth]{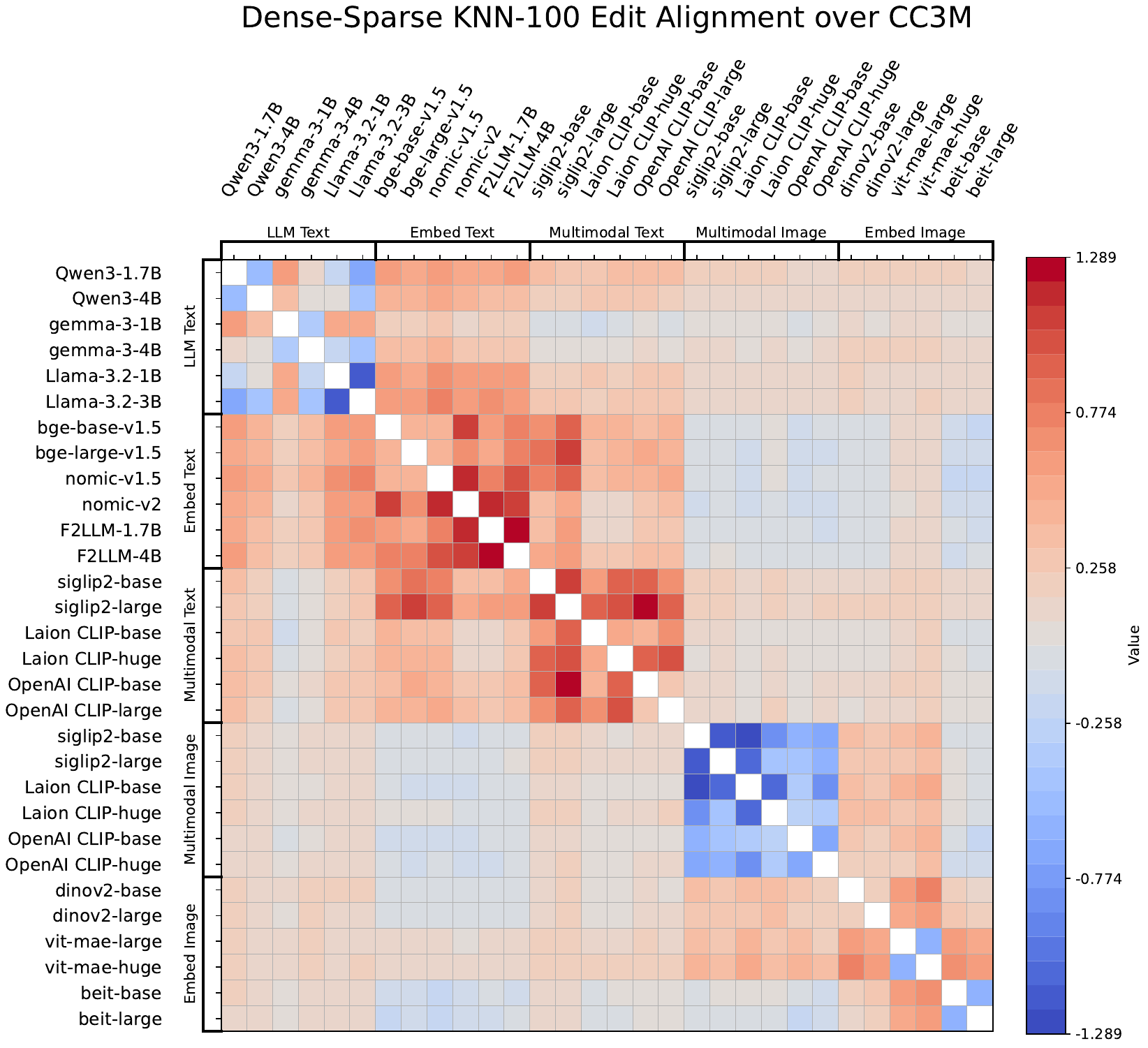}
    \end{minipage}
    \hfill
    \begin{minipage}[t]{0.3\textwidth}
        \centering
        \includegraphics[width = \linewidth]{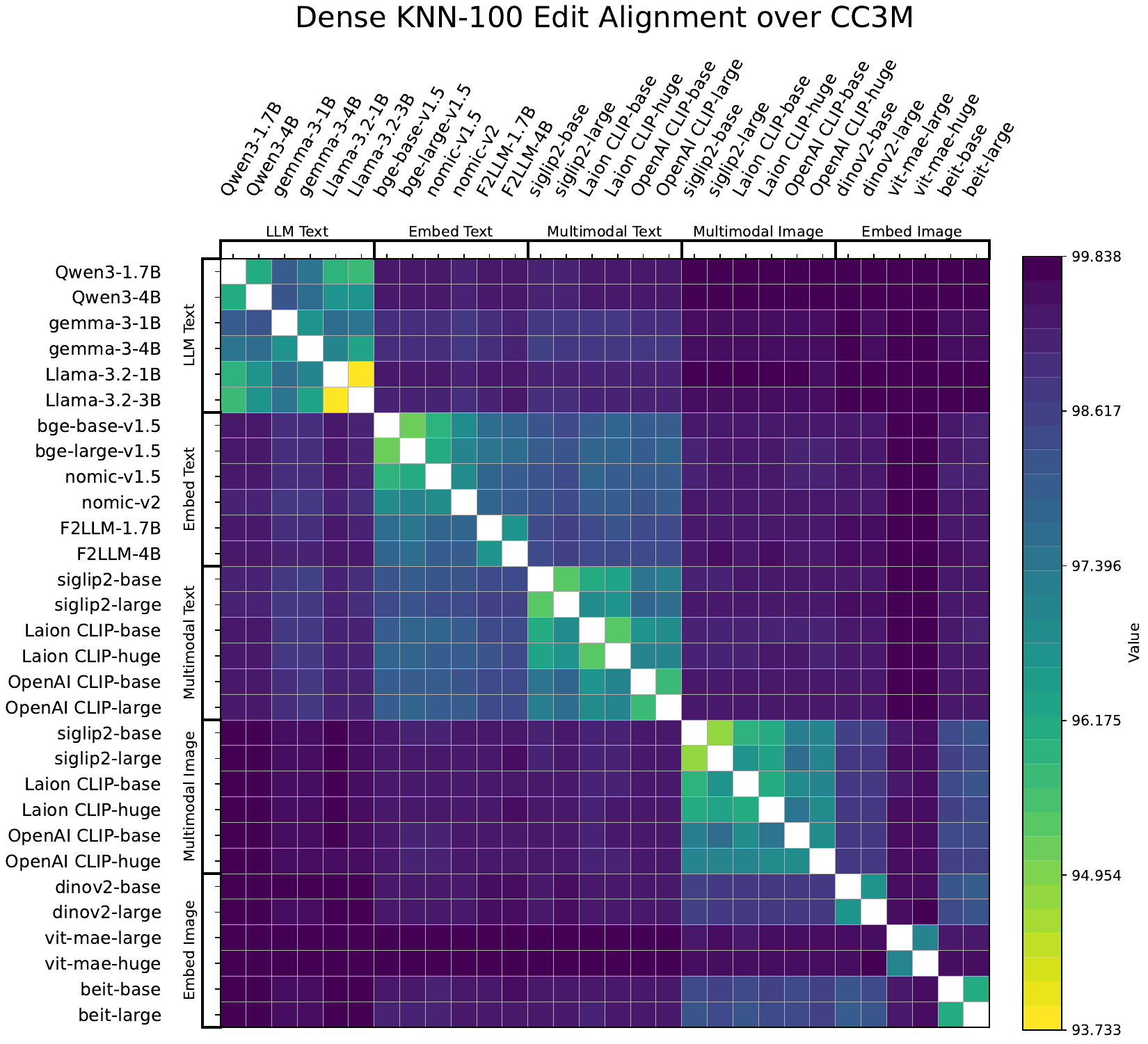}
    \end{minipage}
    \hfill
    \begin{minipage}[t]{0.3\textwidth}
        \centering
        \includegraphics[width = \linewidth]{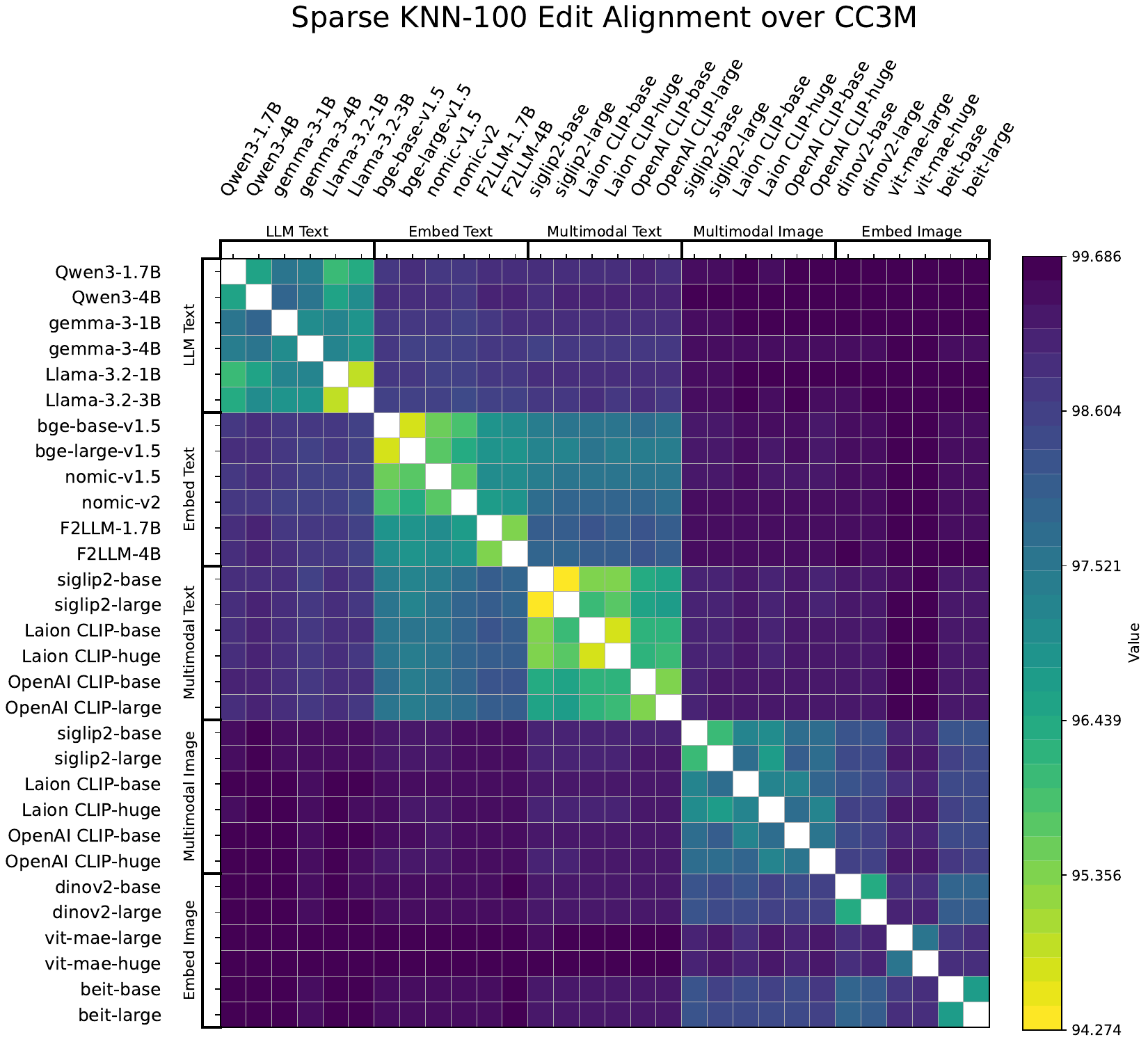}
    \end{minipage}
    \caption{Same plot as in Figure~\ref{fig:signalcoco2} but over CC3M.}
    \label{fig:signalcc3m2}
\end{figure}

\clearpage

\subsubsection{Experiments on Visual Genome}

\begin{figure}[htbp]
    \centering

    \begin{minipage}[t]{0.3\textwidth}
        \centering
        \includegraphics[width = \linewidth]{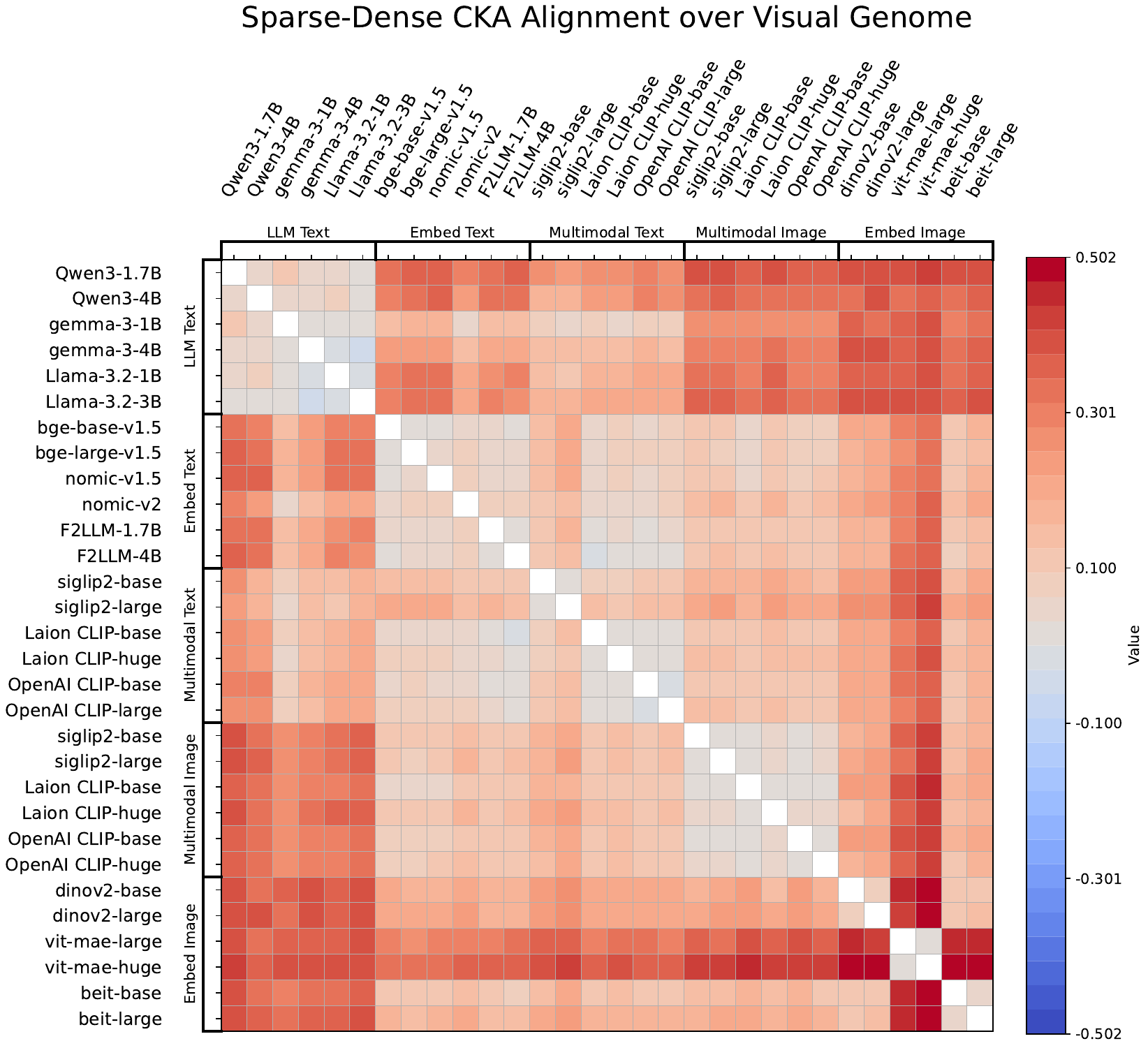}
    \end{minipage}
    \hfill
    \begin{minipage}[t]{0.3\textwidth}
        \centering
        \includegraphics[width = \linewidth]{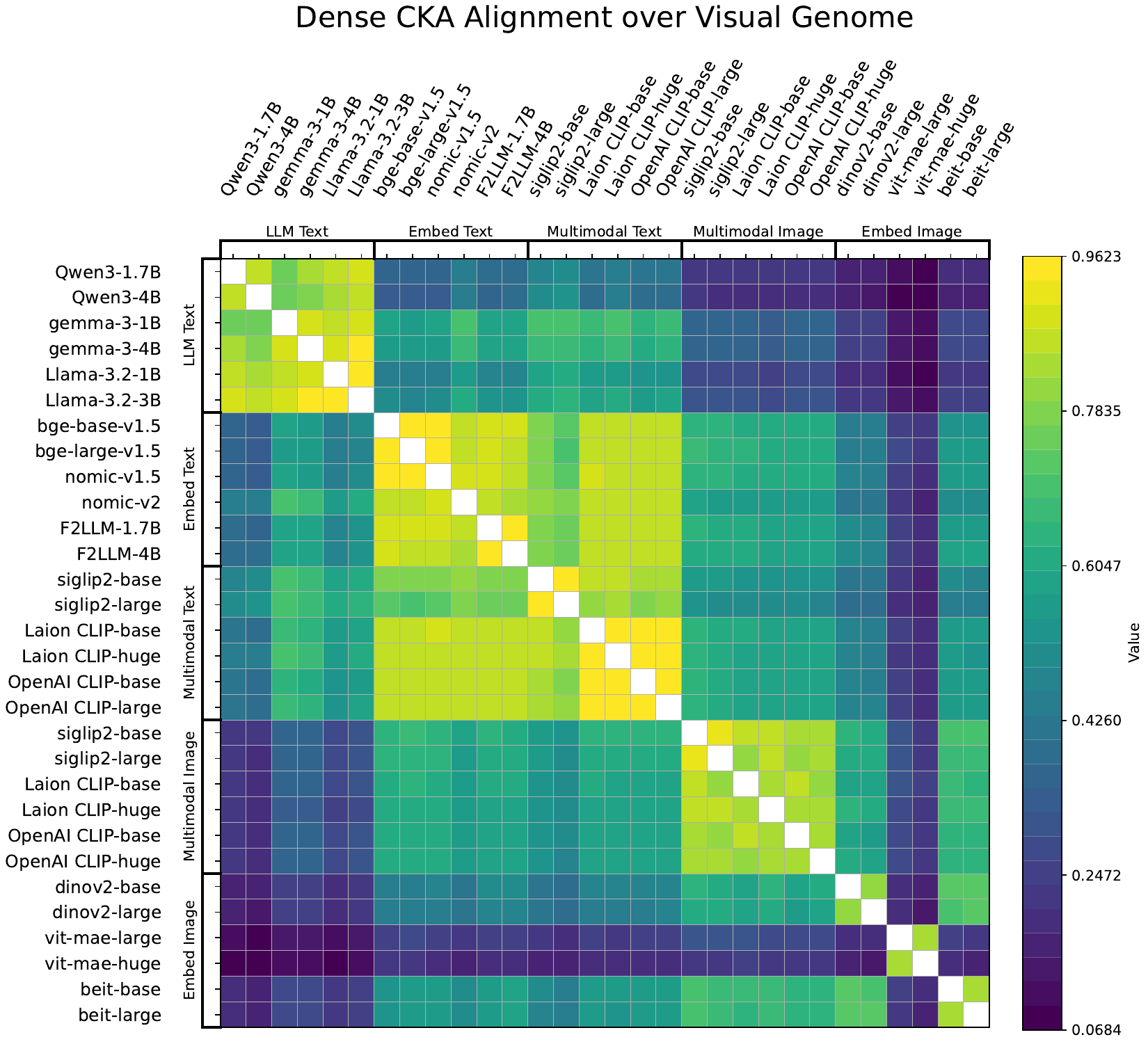}
    \end{minipage}
    \hfill
    \begin{minipage}[t]{0.3\textwidth}
        \centering
        \includegraphics[width = \linewidth]{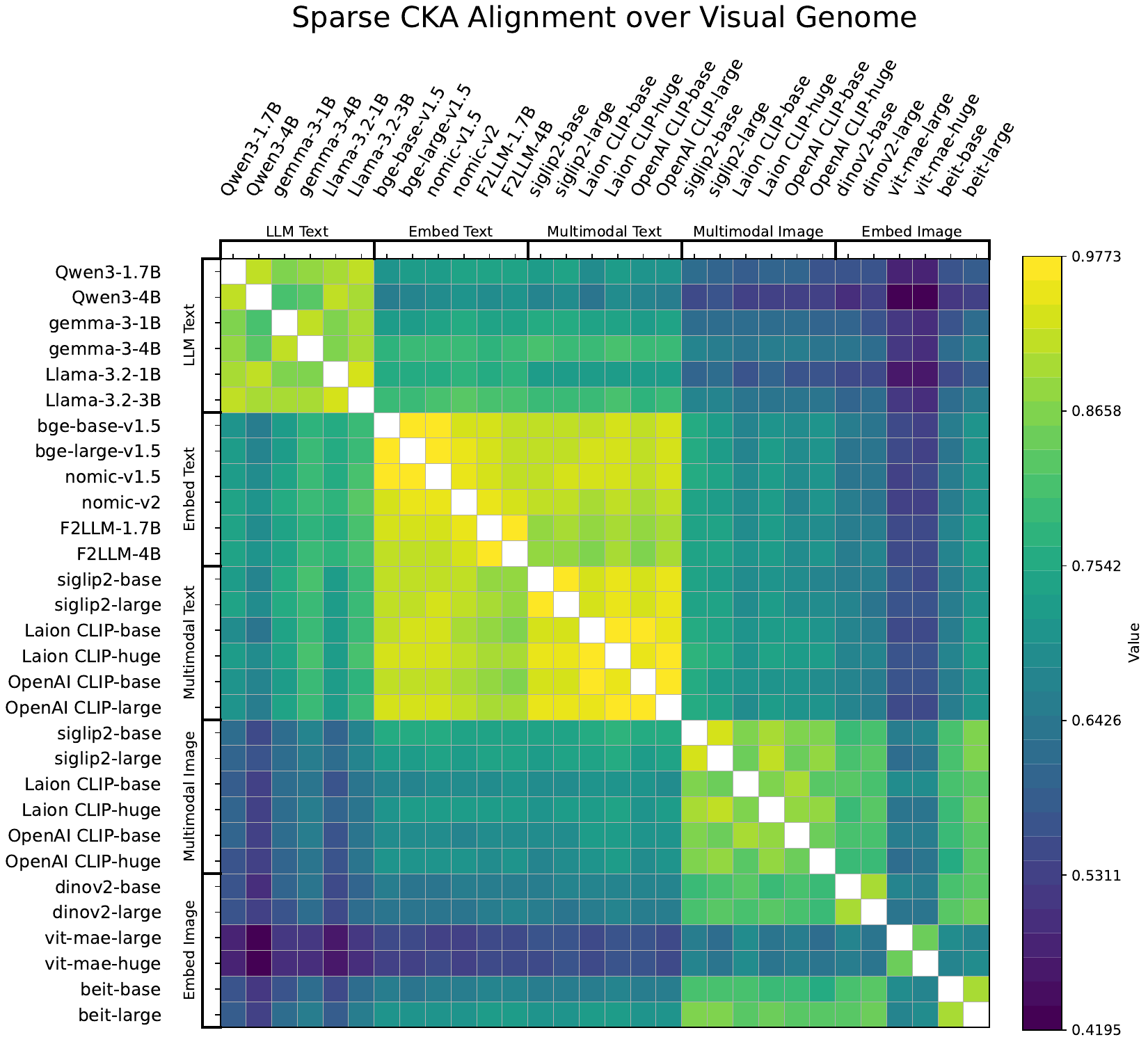}
    \end{minipage}
    
    \vspace{0.4cm}

        \begin{minipage}[t]{0.3\textwidth}
        \centering
        \includegraphics[width = \linewidth]{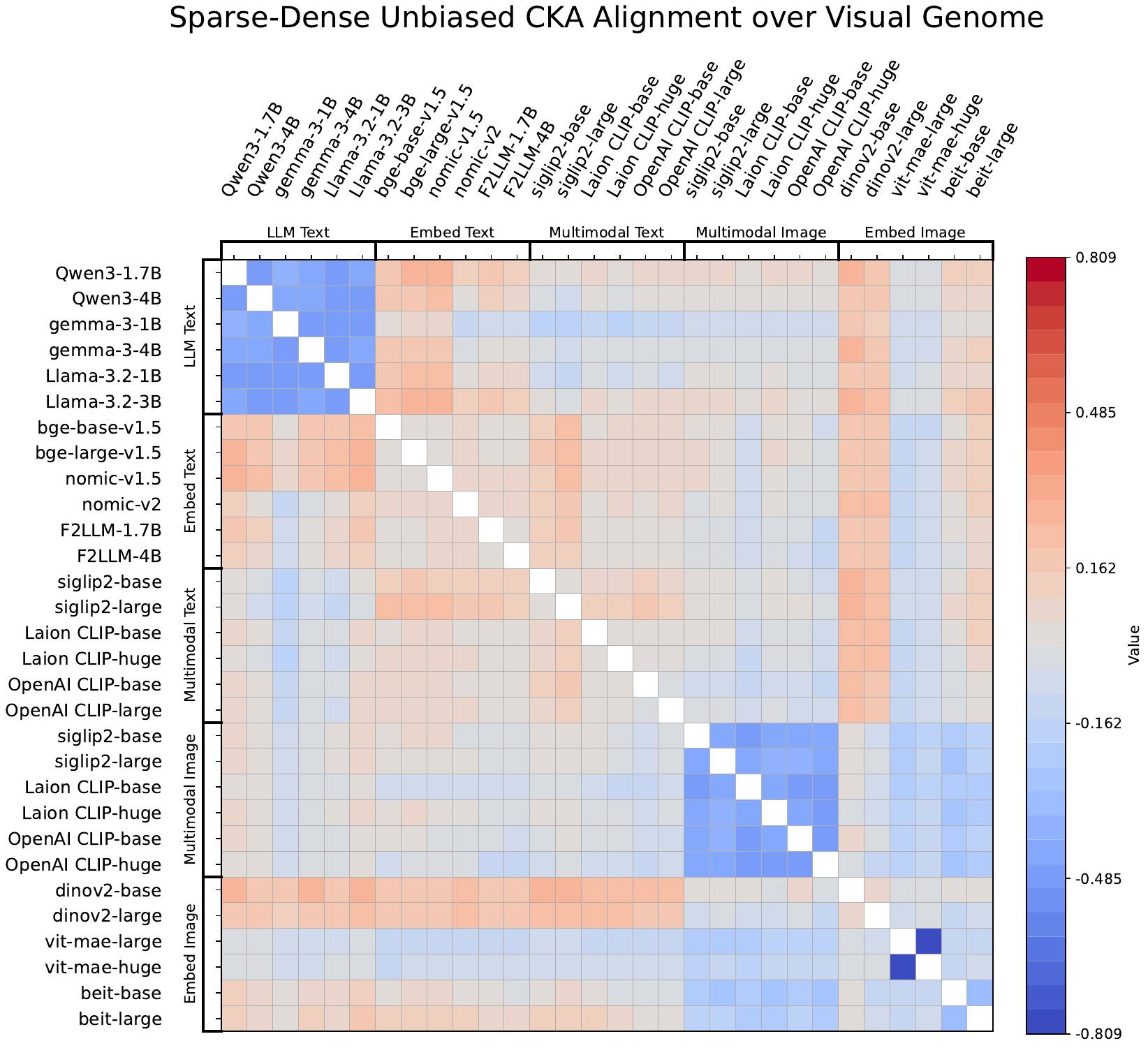}
    \end{minipage}
    \hfill
    \begin{minipage}[t]{0.3\textwidth}
        \centering
        \includegraphics[width = \linewidth]{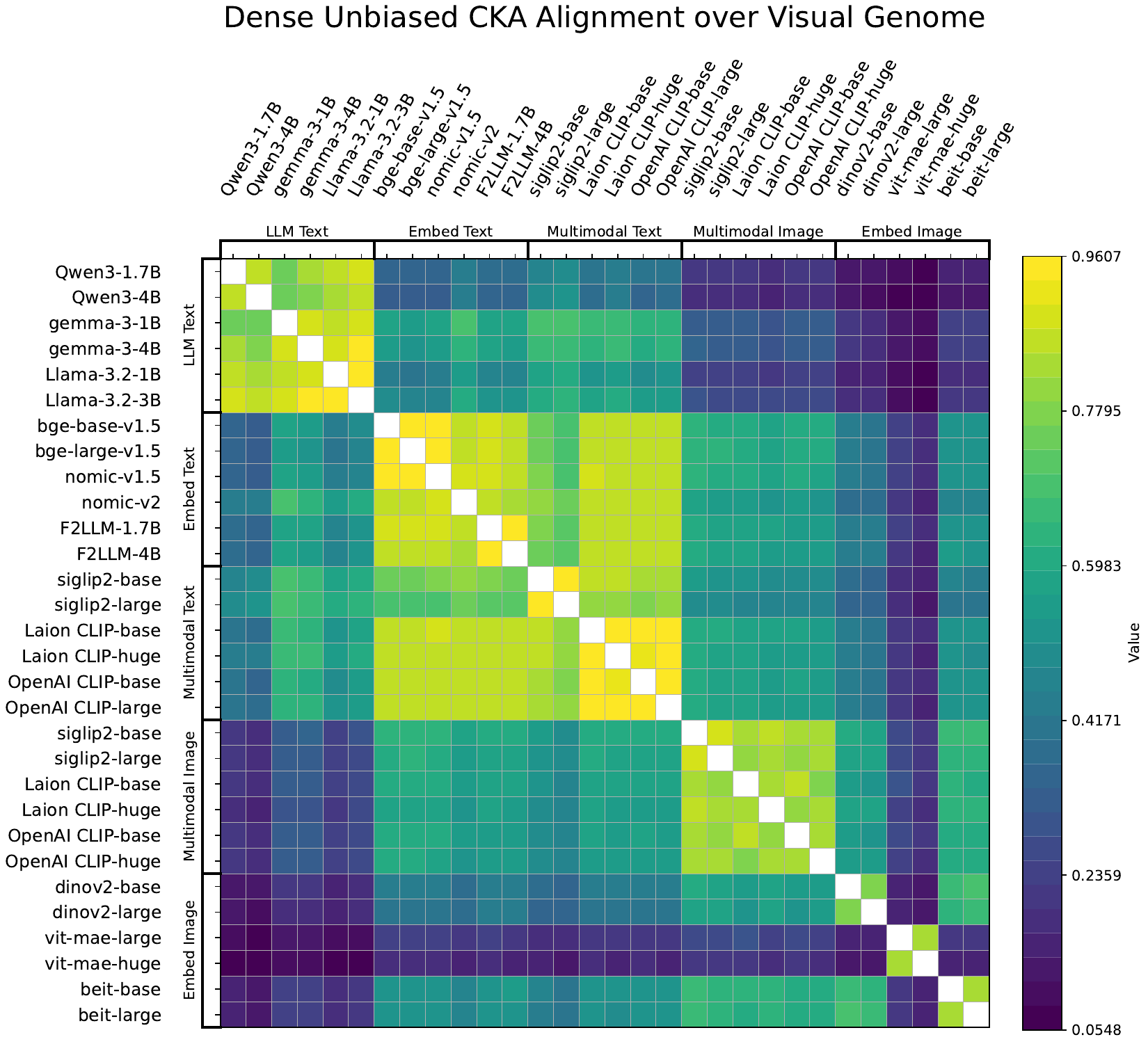}
    \end{minipage}
    \hfill
    \begin{minipage}[t]{0.3\textwidth}
        \centering
        \includegraphics[width = \linewidth]{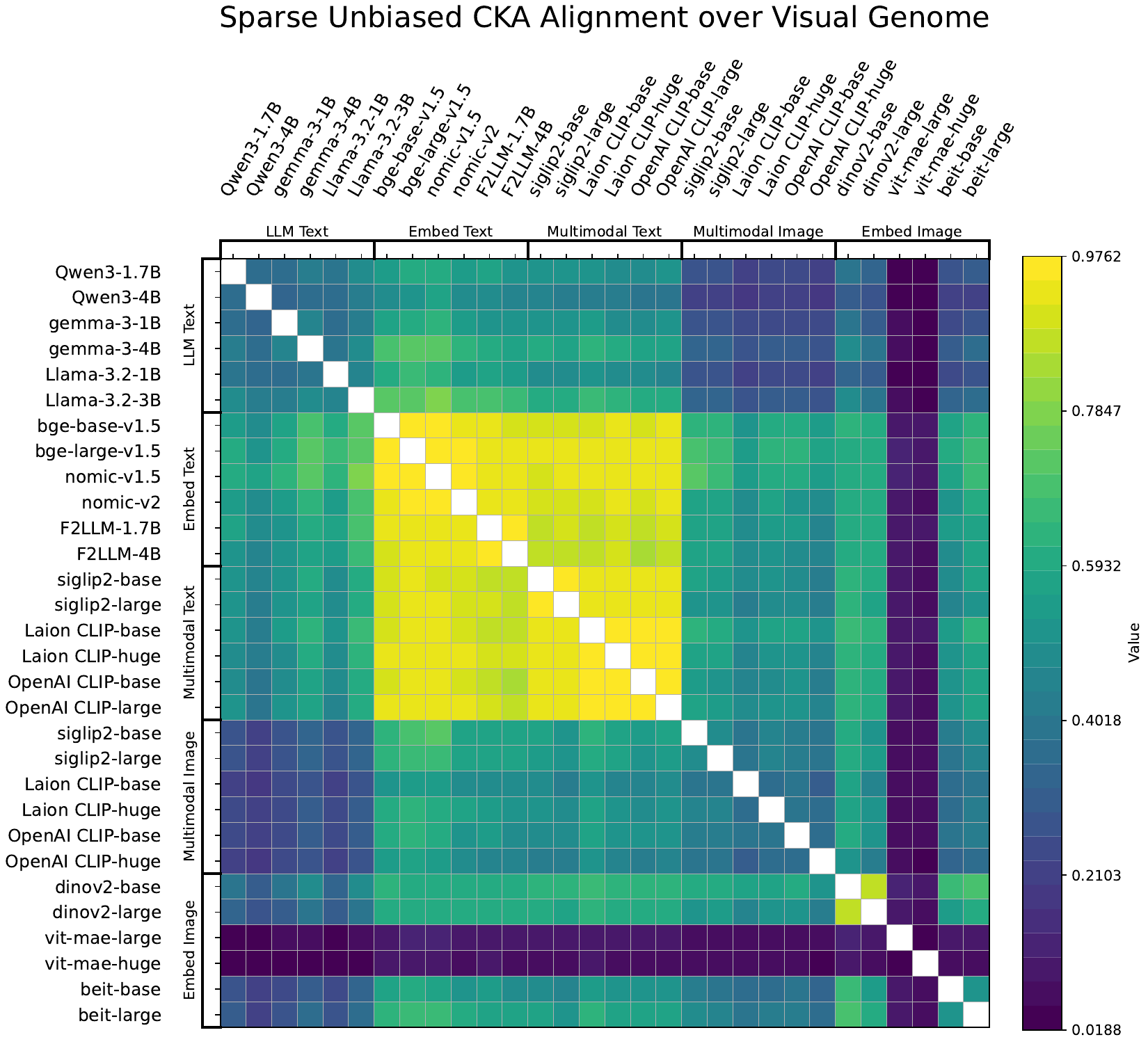}
    \end{minipage}

        \vspace{0.4cm}

        \begin{minipage}[t]{0.3\textwidth}
        \centering
        \includegraphics[width = \linewidth]{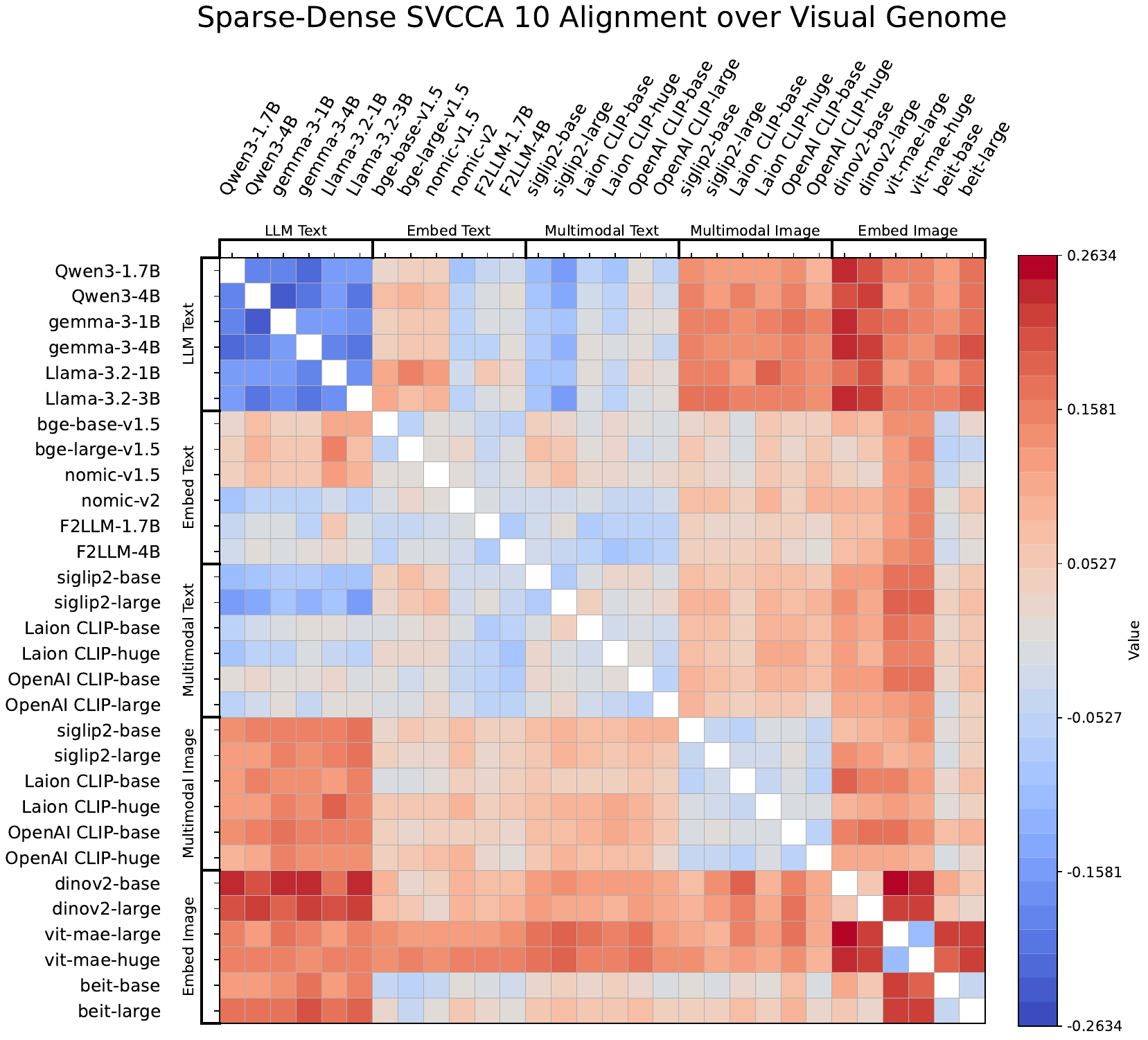}
    \end{minipage}
    \hfill
    \begin{minipage}[t]{0.3\textwidth}
        \centering
        \includegraphics[width = \linewidth]{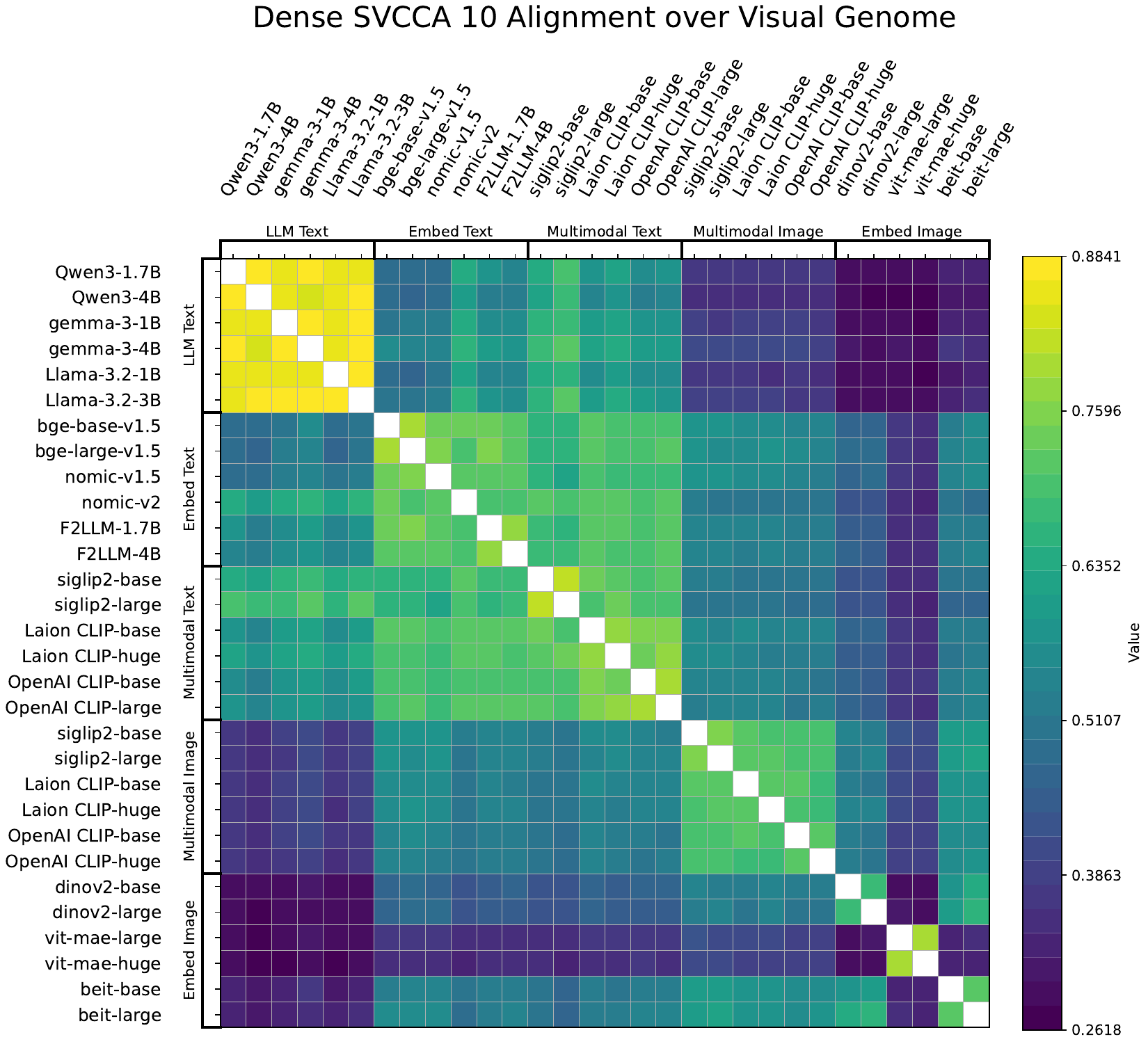}
    \end{minipage}
    \hfill
    \begin{minipage}[t]{0.3\textwidth}
        \centering
        \includegraphics[width = \linewidth]{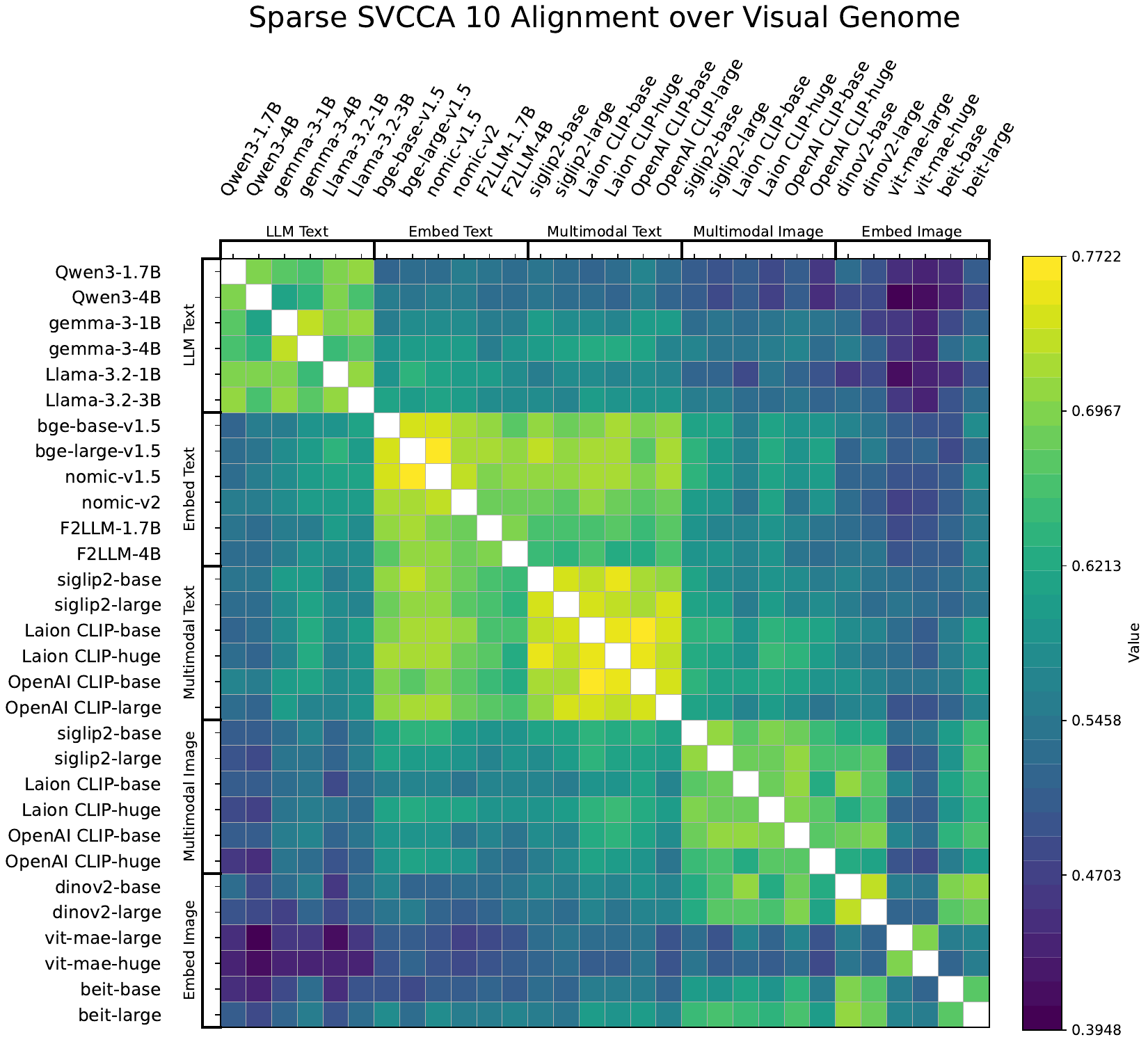}
    \end{minipage}

            \vspace{0.4cm}

    \begin{minipage}[t]{0.3\textwidth}
        \centering
        \includegraphics[width = \linewidth]{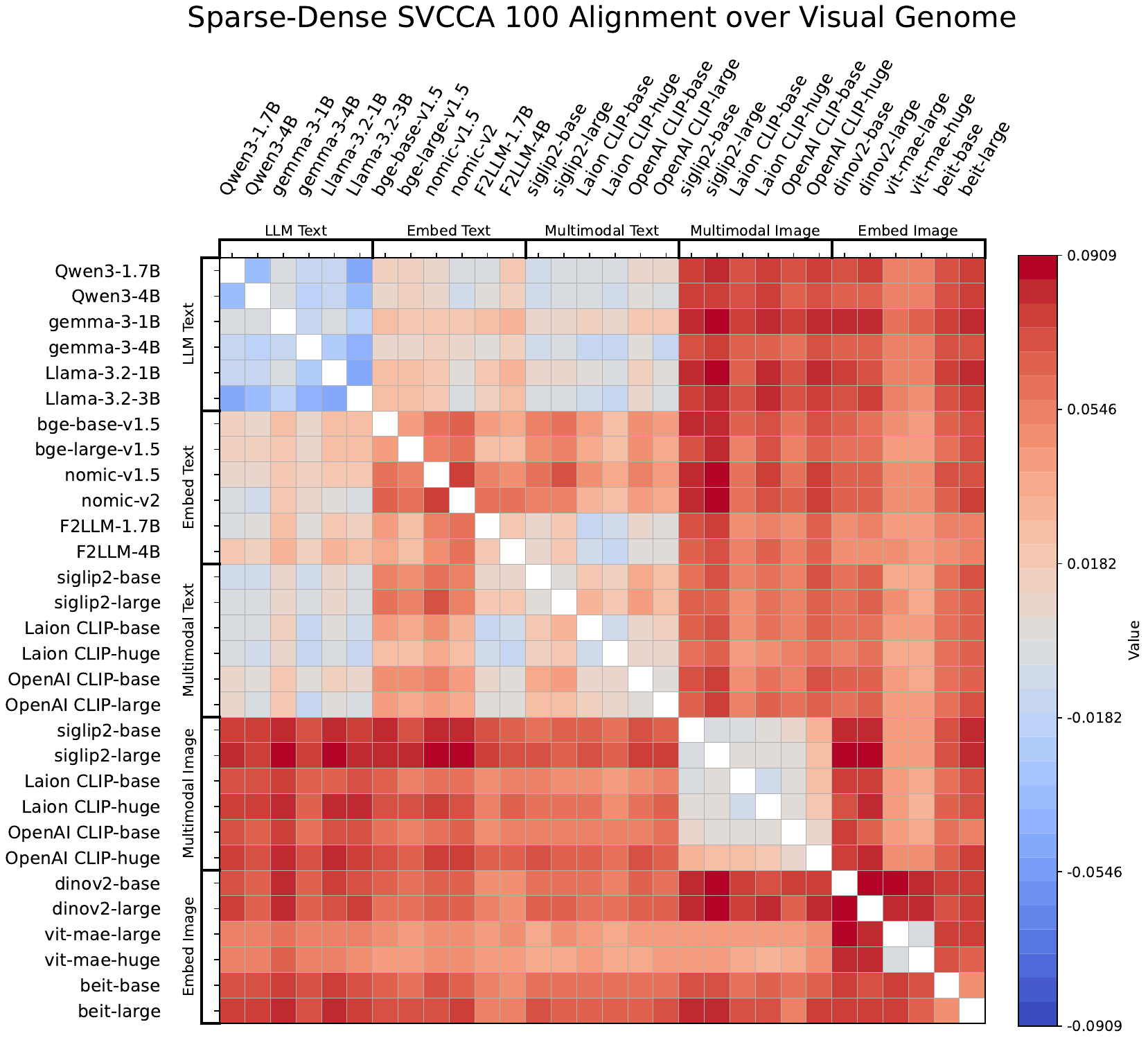}
    \end{minipage}
    \hfill
    \begin{minipage}[t]{0.3\textwidth}
        \centering
        \includegraphics[width = \linewidth]{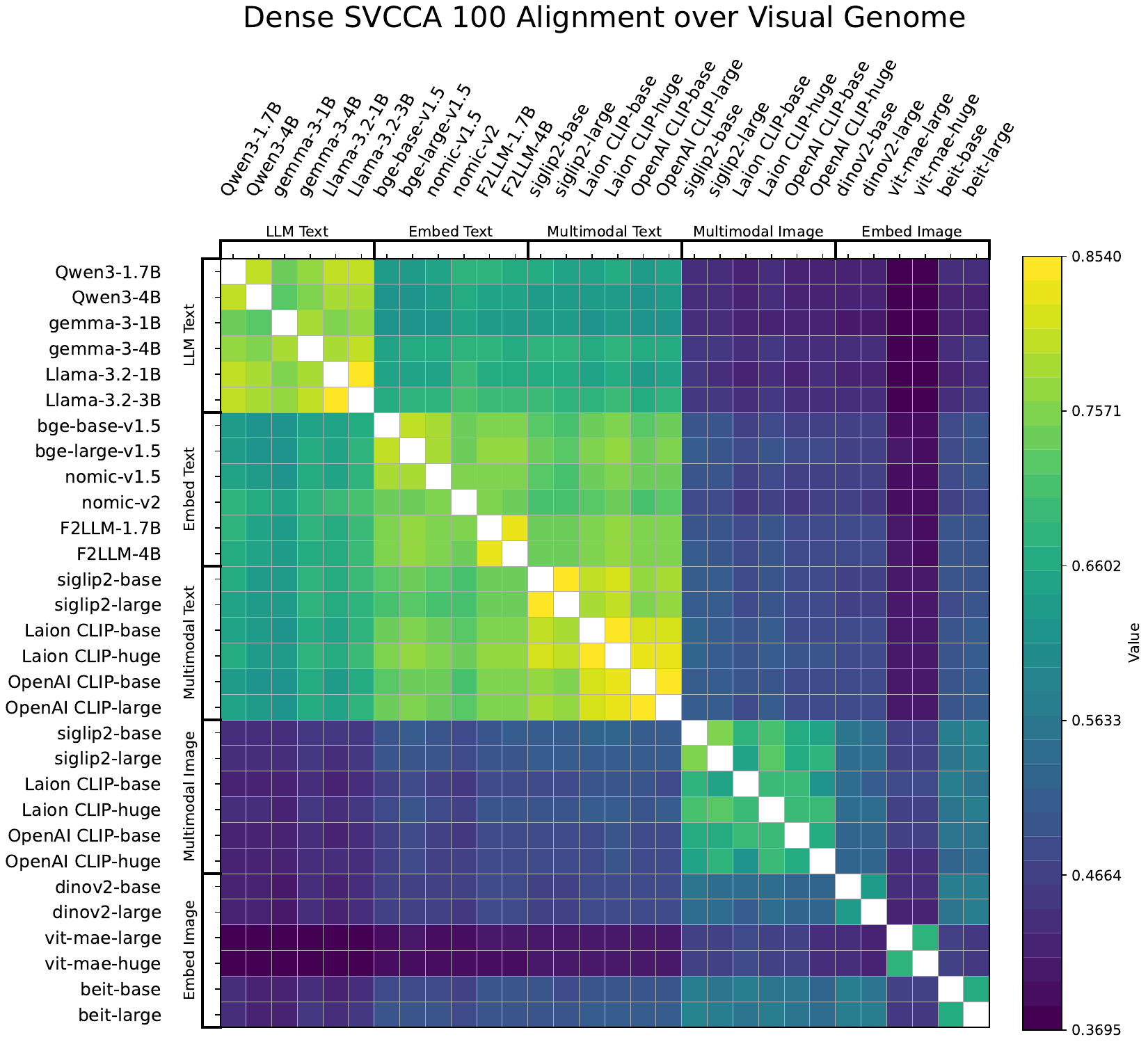}
    \end{minipage}
    \hfill
    \begin{minipage}[t]{0.3\textwidth}
        \centering
        \includegraphics[width = \linewidth]{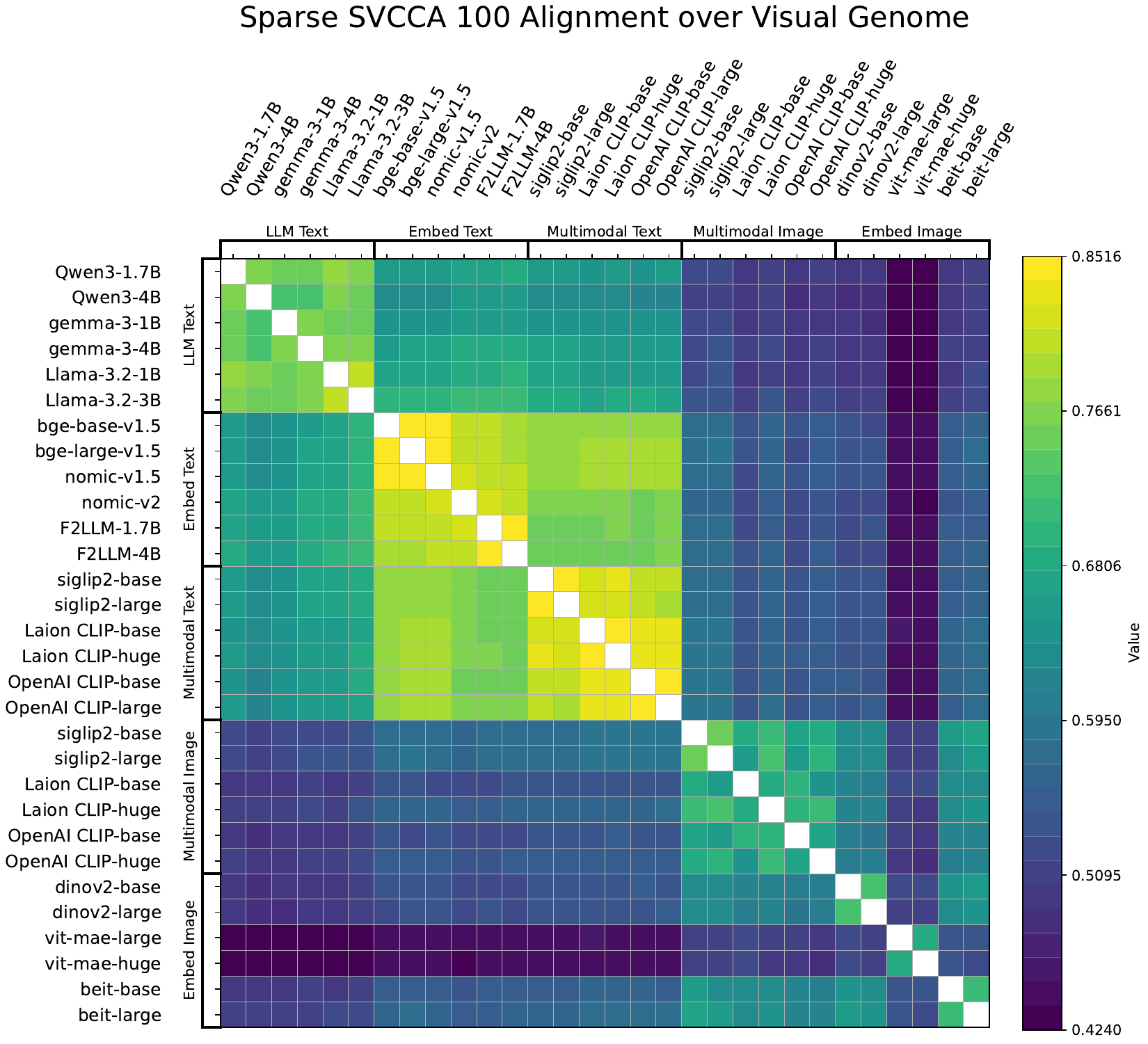}
    \end{minipage}
    \caption{Same plot as in Figure~\ref{fig:signalcoco1} but over Visual Genome.}
    \label{appendix:signalvisual_genome1}
\end{figure}

\clearpage

\begin{figure}[htbp]
    \centering

    \begin{minipage}[t]{0.3\textwidth}
        \centering
        \includegraphics[width = \linewidth]{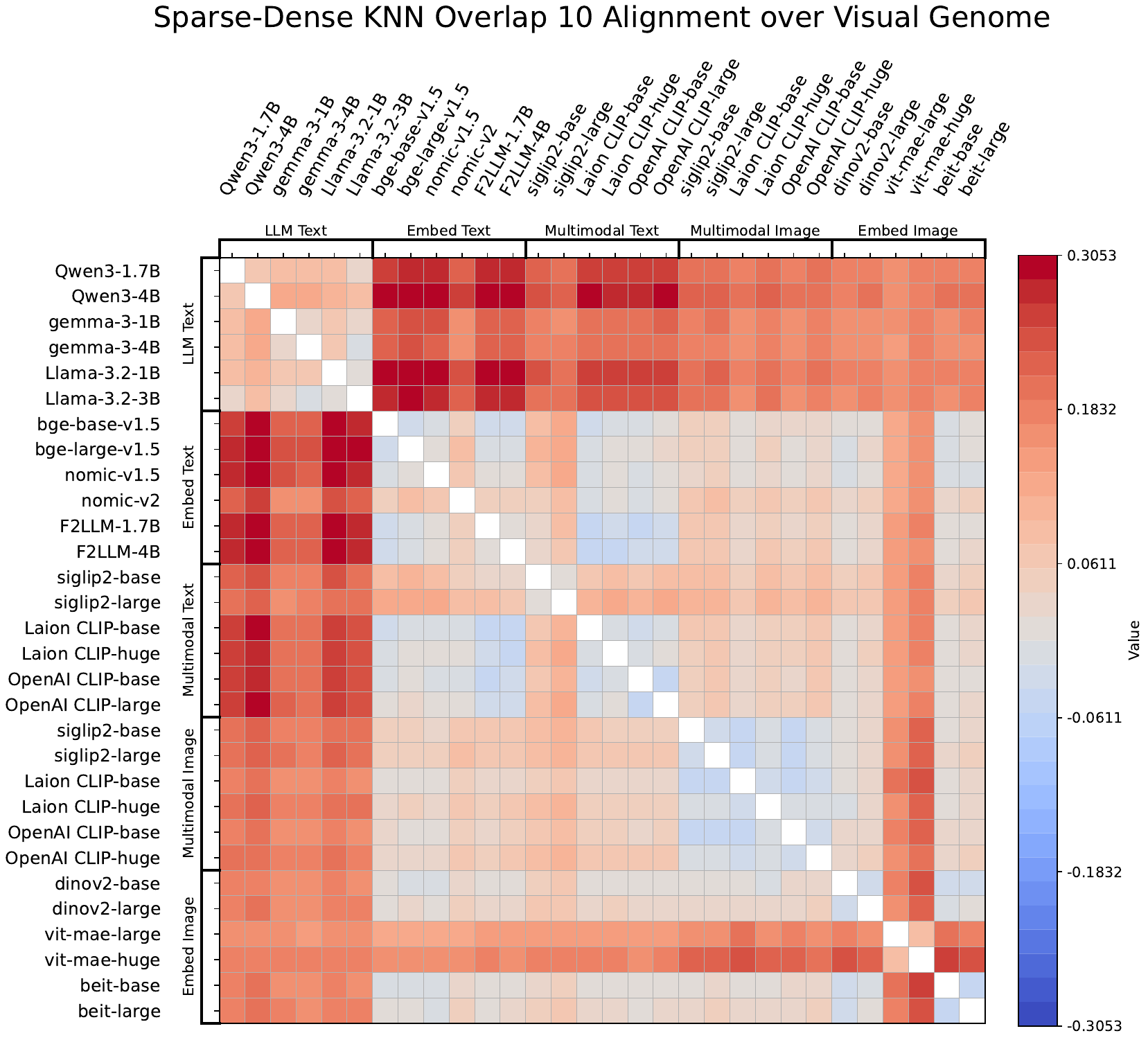}
    \end{minipage}
    \hfill
    \begin{minipage}[t]{0.3\textwidth}
        \centering
        \includegraphics[width = \linewidth]{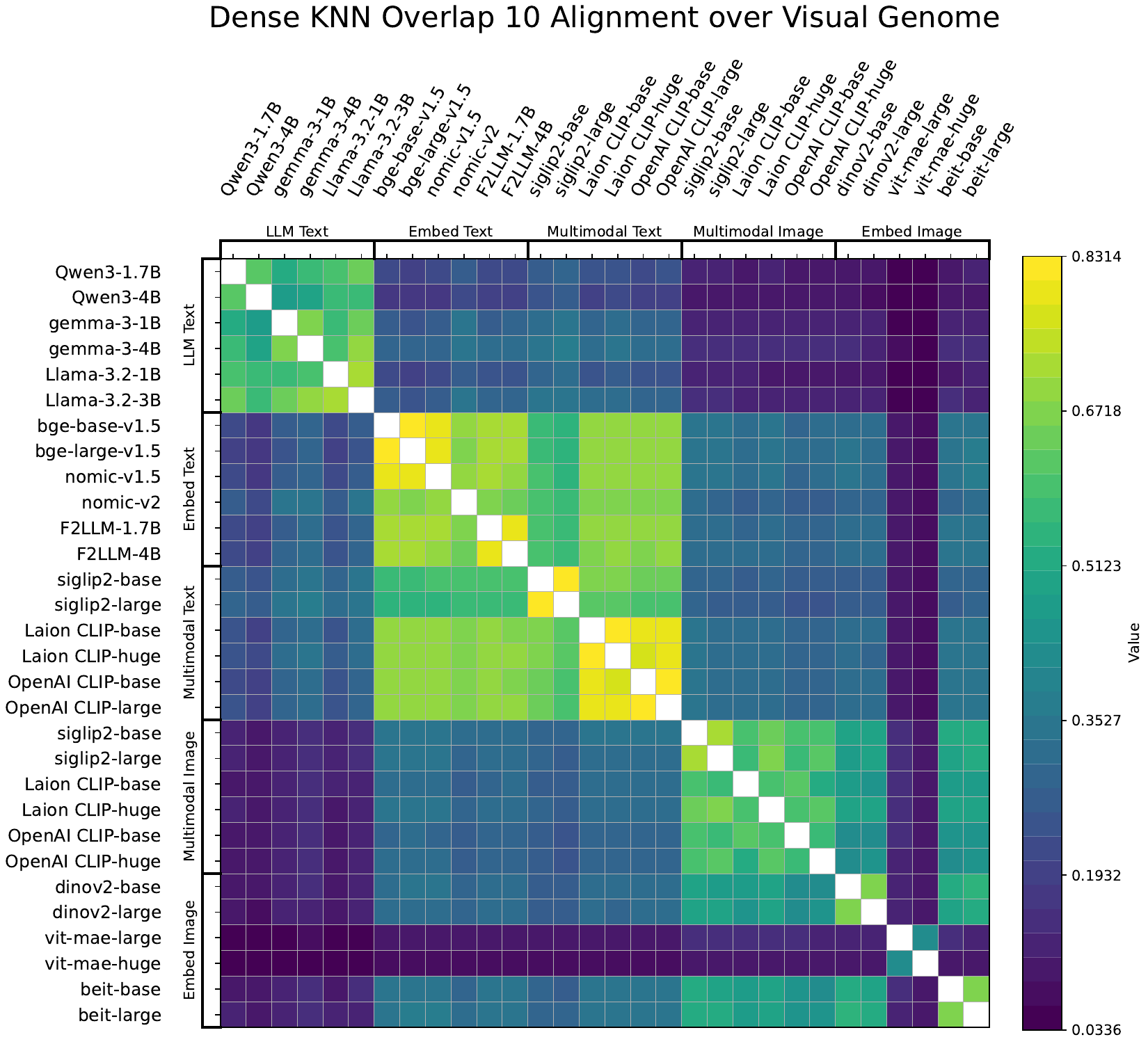}
    \end{minipage}
    \hfill
    \begin{minipage}[t]{0.3\textwidth}
        \centering
        \includegraphics[width = \linewidth]{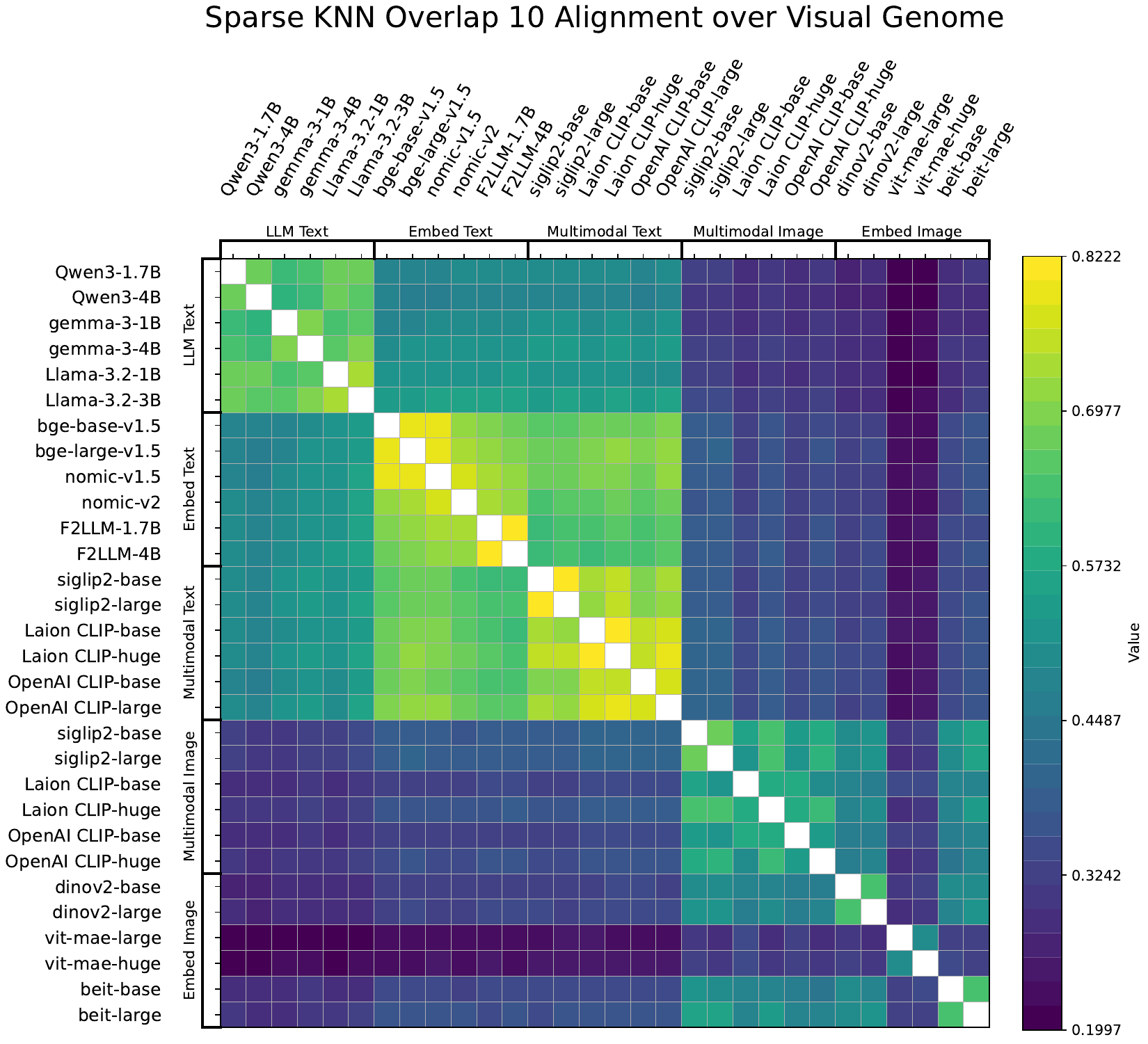}
    \end{minipage}
    
    \vspace{0.4cm}

        \begin{minipage}[t]{0.3\textwidth}
        \centering
        \includegraphics[width = \linewidth]{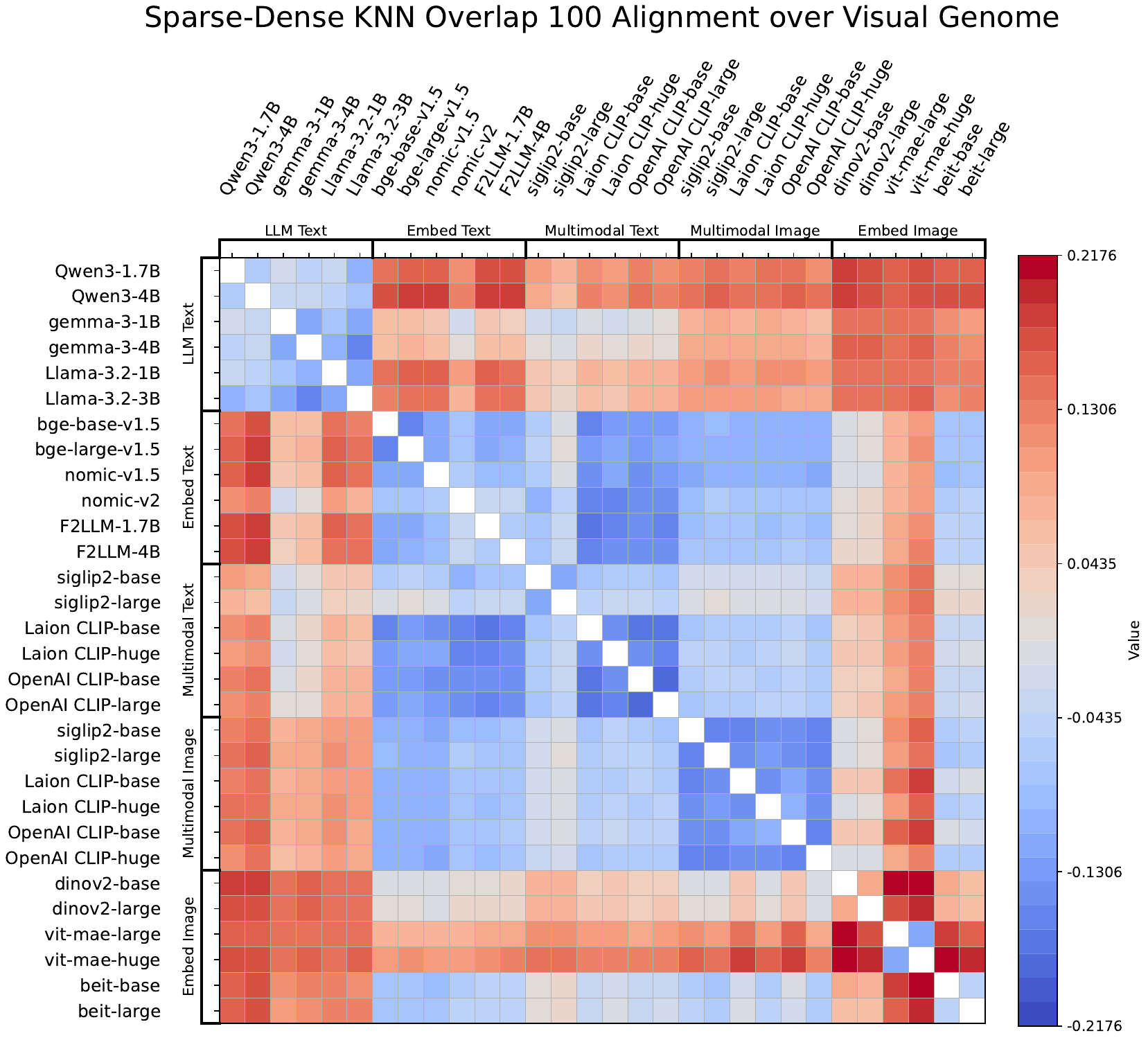}
    \end{minipage}
    \hfill
    \begin{minipage}[t]{0.3\textwidth}
        \centering
        \includegraphics[width = \linewidth]{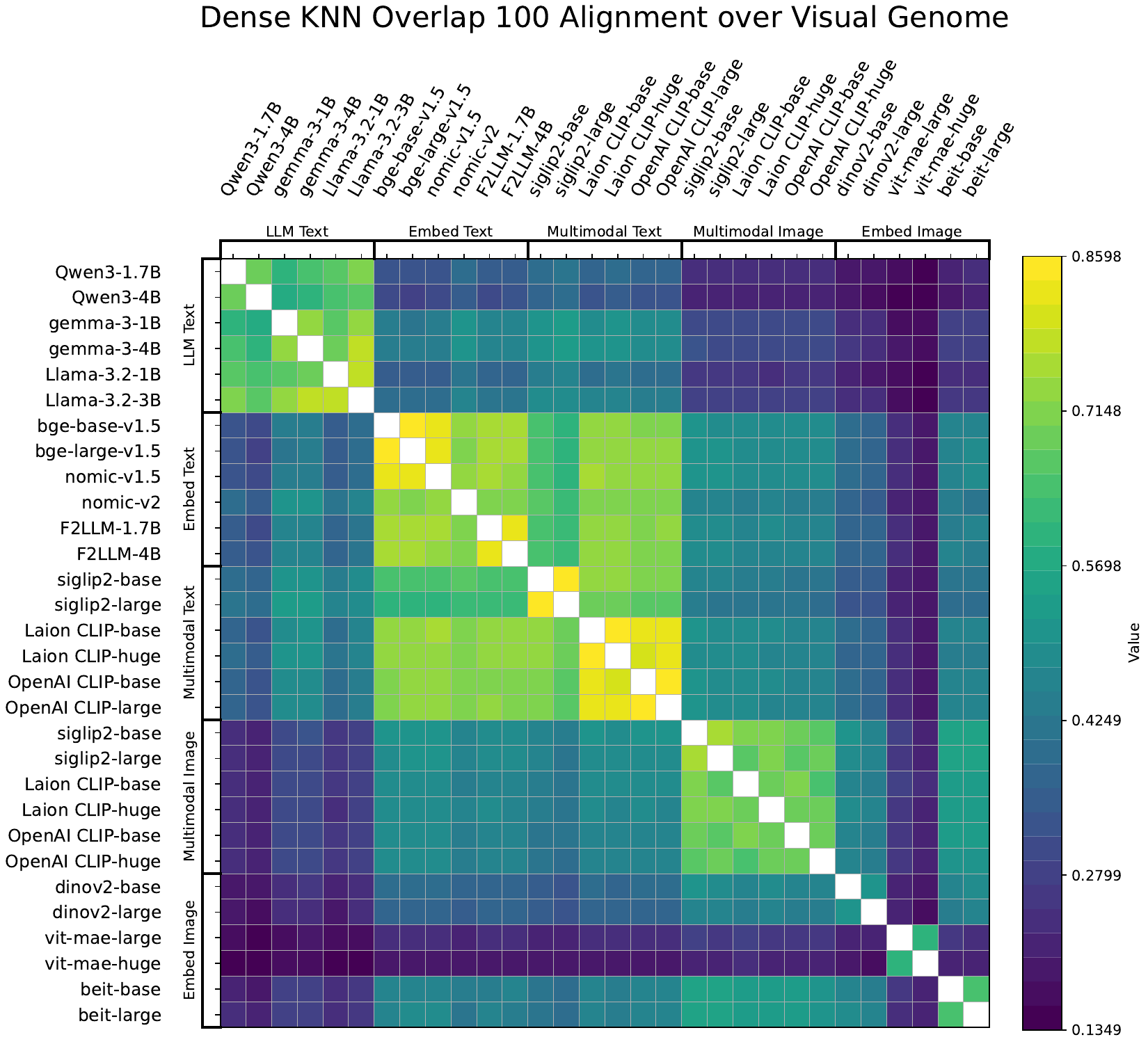}
    \end{minipage}
    \hfill
    \begin{minipage}[t]{0.3\textwidth}
        \centering
        \includegraphics[width = \linewidth]{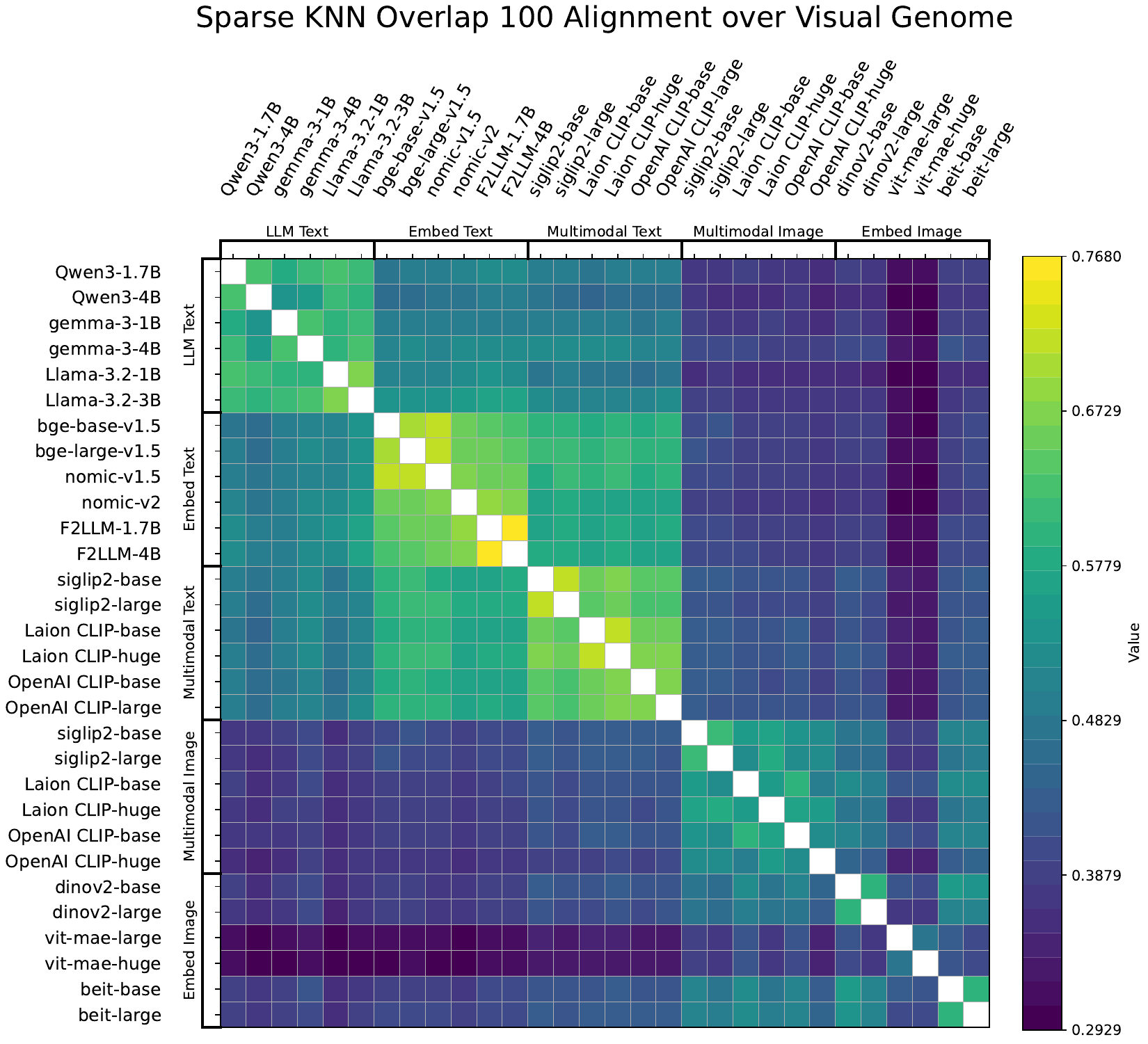}
    \end{minipage}

        \vspace{0.4cm}

        \begin{minipage}[t]{0.3\textwidth}
        \centering
        \includegraphics[width = \linewidth]{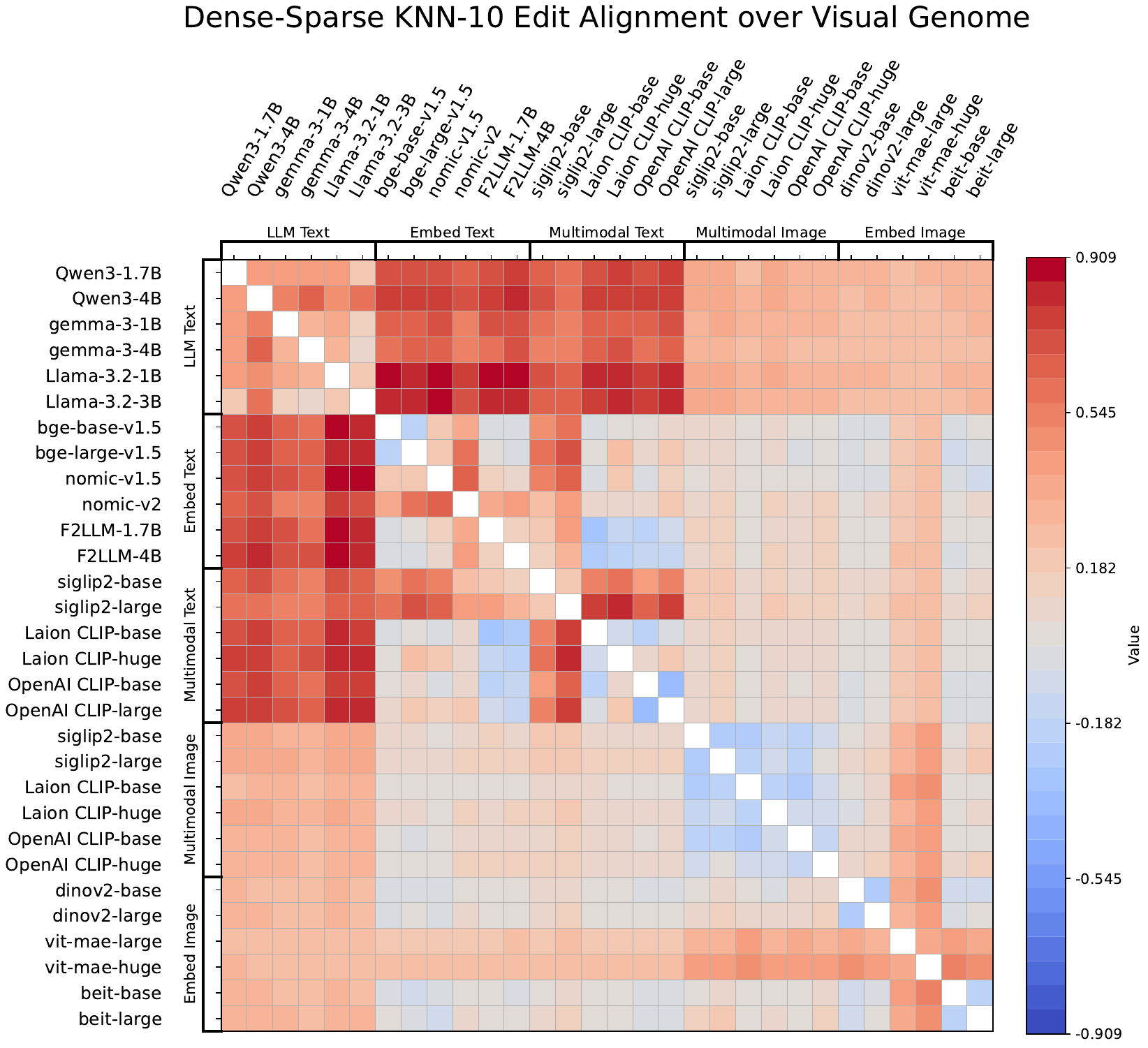}
    \end{minipage}
    \hfill
    \begin{minipage}[t]{0.3\textwidth}
        \centering
        \includegraphics[width = \linewidth]{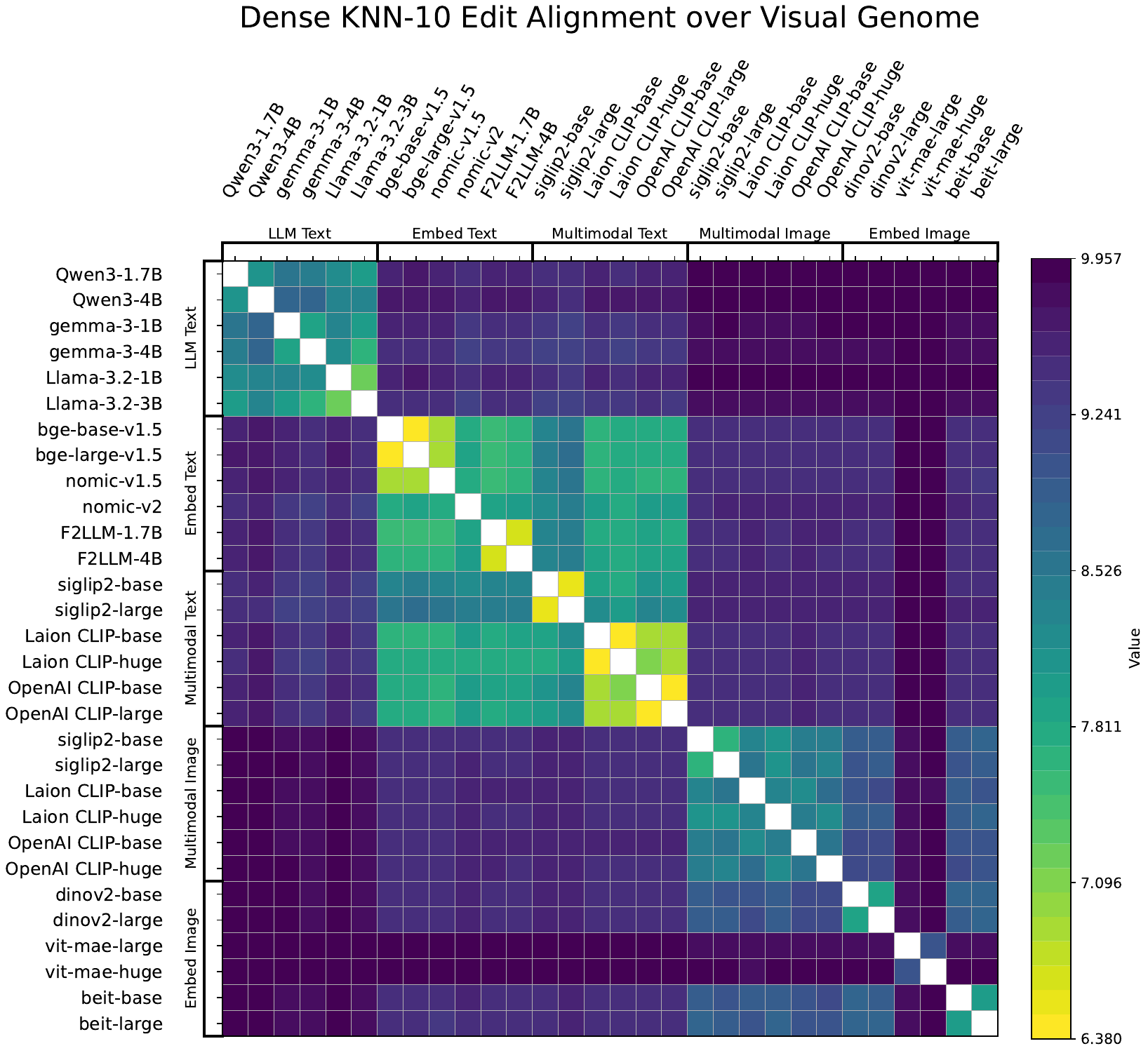}
    \end{minipage}
    \hfill
    \begin{minipage}[t]{0.3\textwidth}
        \centering
        \includegraphics[width = \linewidth]{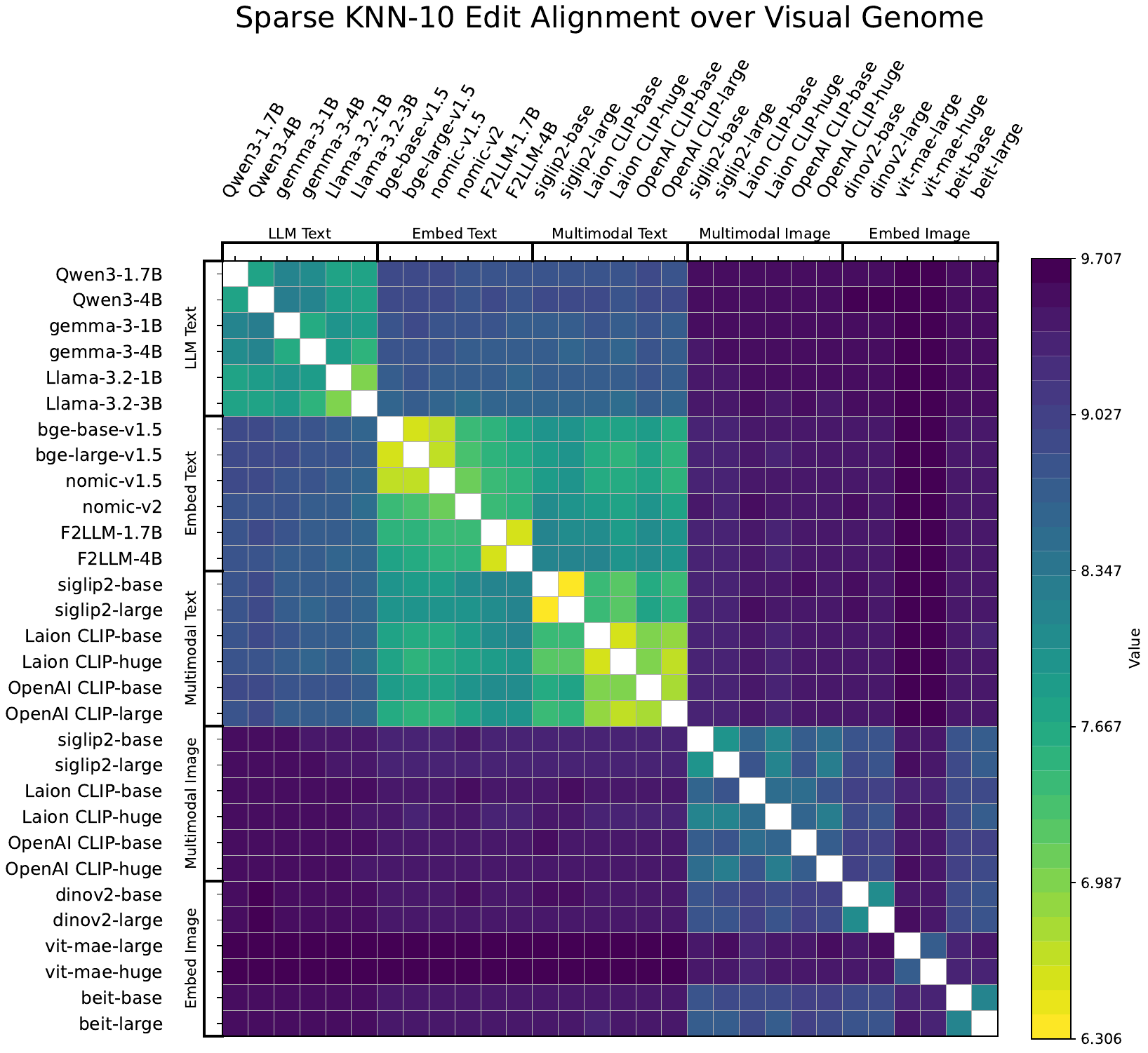}
    \end{minipage}

            \vspace{0.4cm}

    \begin{minipage}[t]{0.3\textwidth}
        \centering
        \includegraphics[width = \linewidth]{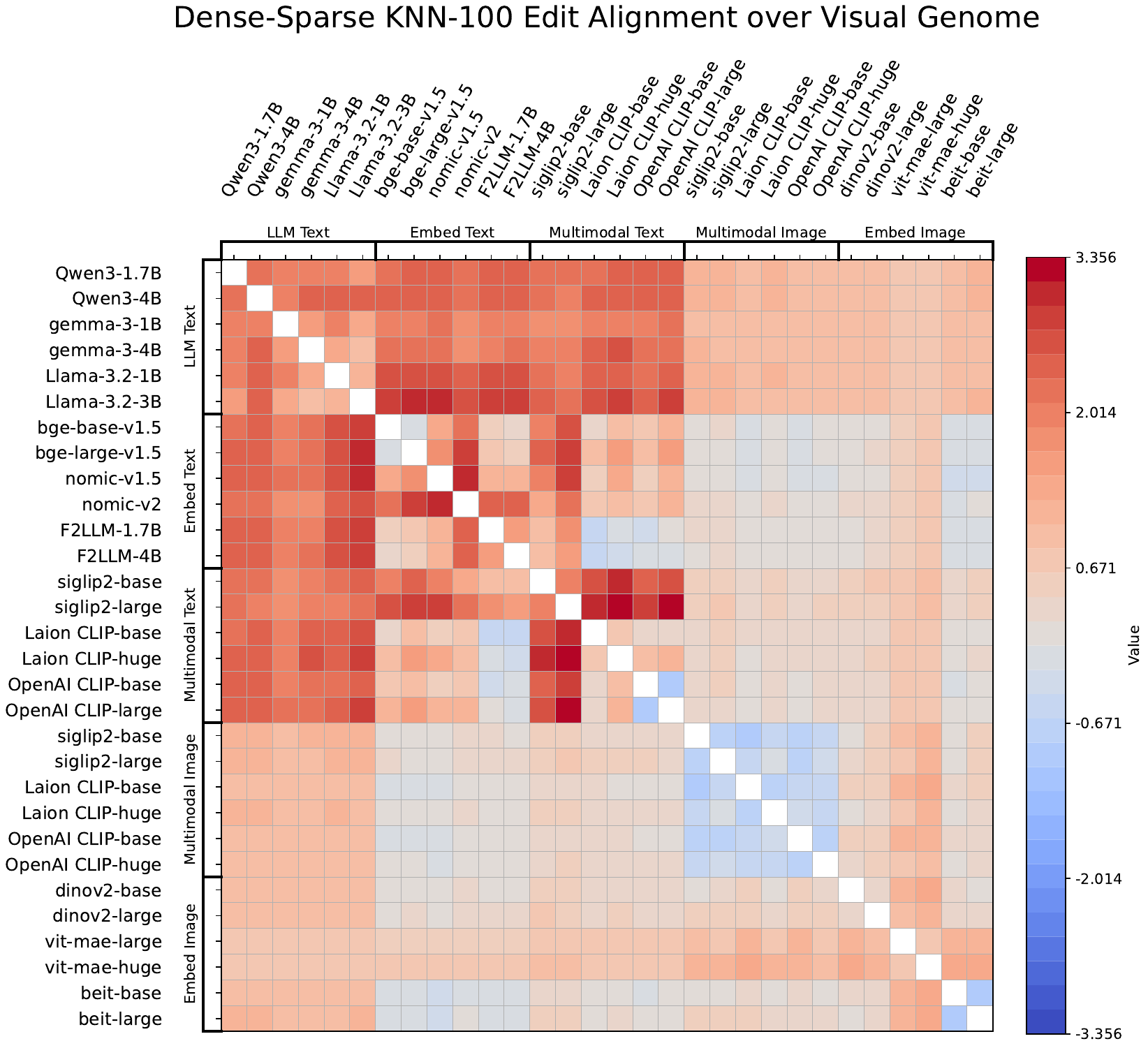}
    \end{minipage}
    \hfill
    \begin{minipage}[t]{0.3\textwidth}
        \centering
        \includegraphics[width = \linewidth]{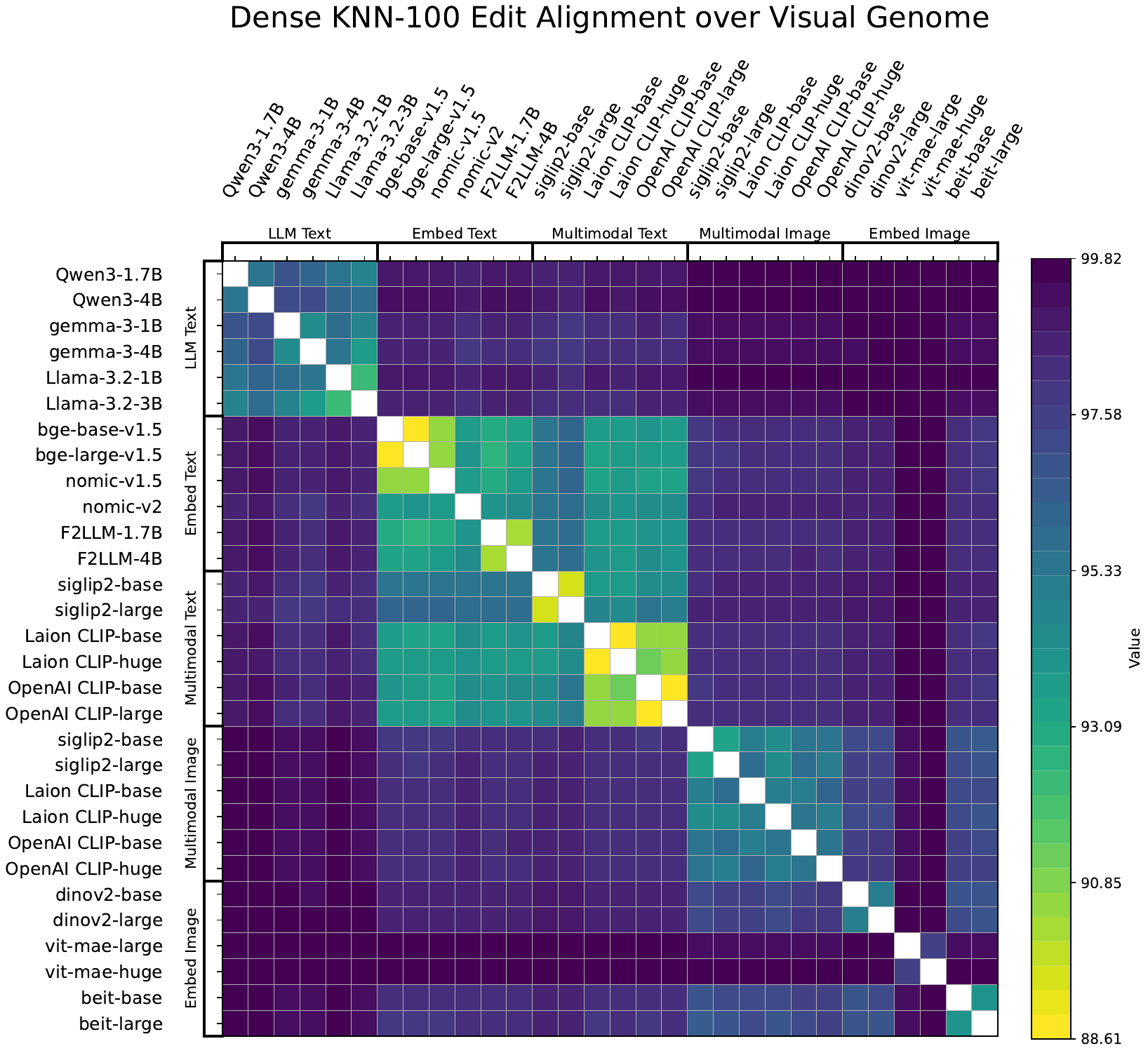}
    \end{minipage}
    \hfill
    \begin{minipage}[t]{0.3\textwidth}
        \centering
        \includegraphics[width = \linewidth]{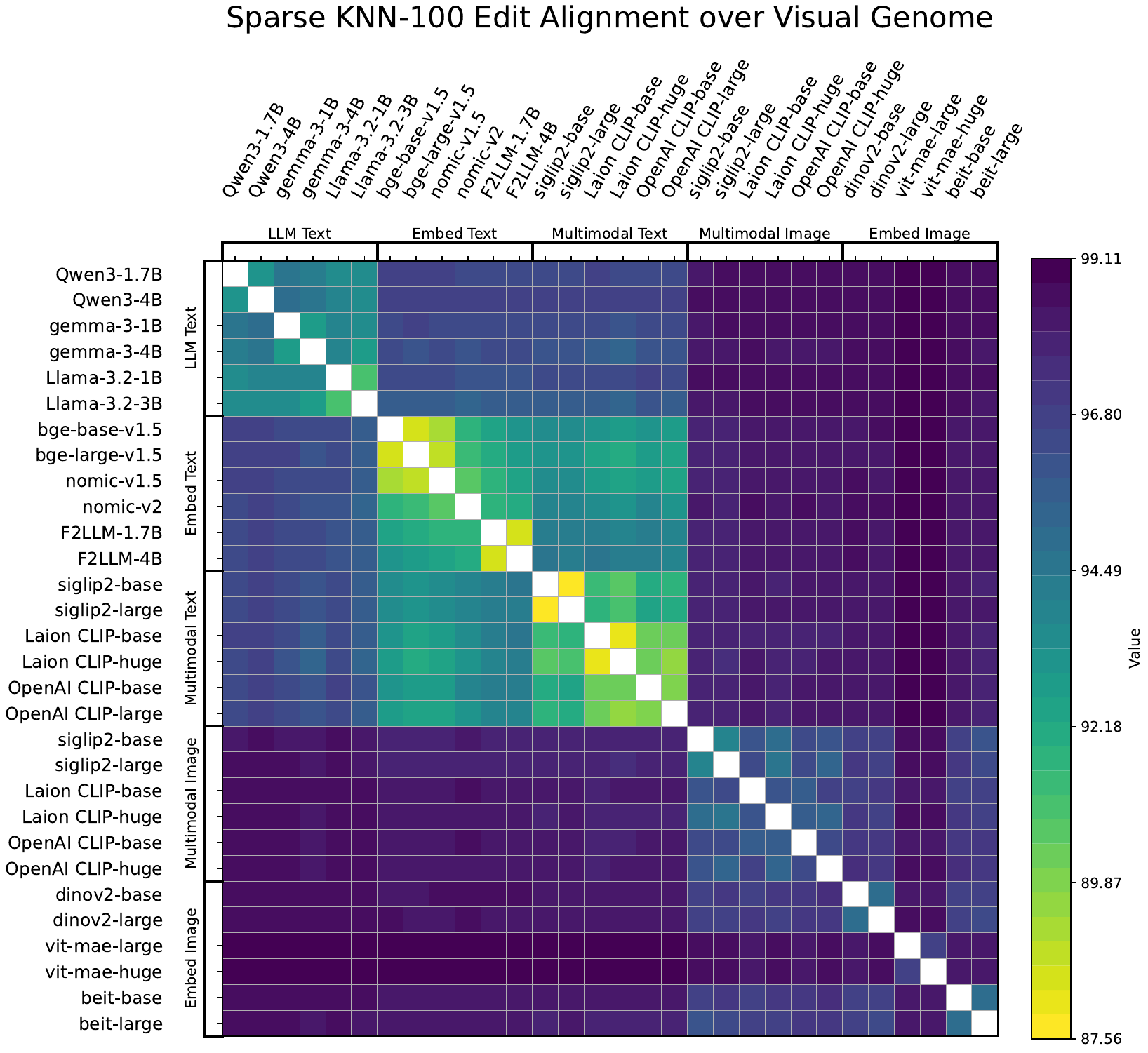}
    \end{minipage}
    \caption{Same plot as in Figure~\ref{fig:signalcoco2} but over Visual Genome.}
    \label{fig:signalvisual_genome2}
\end{figure}

\clearpage

\subsubsection{SAE Dimension Ablation: Experiments on COCO}
\label{sec:dimensionablation}
We provide an ablation study where we change the SAE dimension to $d = 8192.$ We present the experiments only for COCO since there is no qualitative or quantitative difference from $d = 16384.$ 

\begin{figure}[htbp]
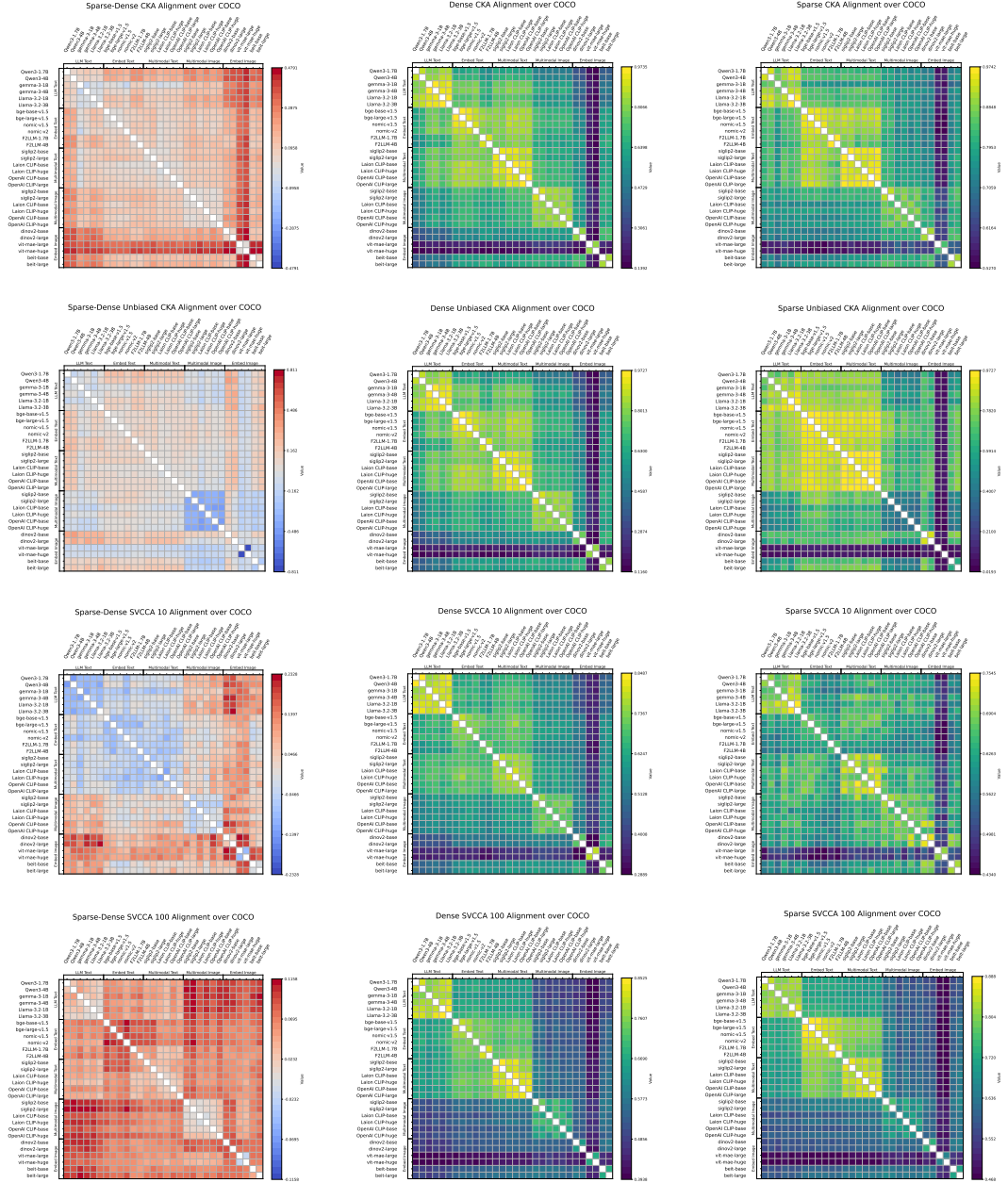

    \centering

    \begin{minipage}[t]{0.3\textwidth}
        \centering
        \includegraphics[width = \linewidth]{signal_diagrams/cka_coco_diff_16384.pdf}
    \end{minipage}
    \hfill
    \begin{minipage}[t]{0.3\textwidth}
        \centering
        \includegraphics[width = \linewidth]{signal_diagrams/cka_coco_raw_16384.pdf}
    \end{minipage}
    \hfill
    \begin{minipage}[t]{0.3\textwidth}
        \centering
        \includegraphics[width = \linewidth]{signal_diagrams/cka_coco_filtered_16384.pdf}
    \end{minipage}
    
    \vspace{0.4cm}

        \begin{minipage}[t]{0.3\textwidth}
        \centering
        \includegraphics[width = \linewidth]{signal_diagrams/cka_unbiased_coco_diff_16384.pdf}
    \end{minipage}
    \hfill
    \begin{minipage}[t]{0.3\textwidth}
        \centering
        \includegraphics[width = \linewidth]{signal_diagrams/cka_unbiased_coco_raw_16384.pdf}
    \end{minipage}
    \hfill
    \begin{minipage}[t]{0.3\textwidth}
        \centering
        \includegraphics[width = \linewidth]{signal_diagrams/cka_unbiased_coco_filtered_16384.pdf}
    \end{minipage}

        \vspace{0.4cm}

        \begin{minipage}[t]{0.3\textwidth}
        \centering
        \includegraphics[width = \linewidth]{signal_diagrams/svcca_10_coco_diff_16384.pdf}
    \end{minipage}
    \hfill
    \begin{minipage}[t]{0.3\textwidth}
        \centering
        \includegraphics[width = \linewidth]{signal_diagrams/svcca_10_coco_raw_16384.pdf}
    \end{minipage}
    \hfill
    \begin{minipage}[t]{0.3\textwidth}
        \centering
        \includegraphics[width = \linewidth]{signal_diagrams/svcca_10_coco_filtered_16384.pdf}
    \end{minipage}

            \vspace{0.4cm}

    \begin{minipage}[t]{0.3\textwidth}
        \centering
        \includegraphics[width = \linewidth]{signal_diagrams/svcca_100_coco_diff_16384.pdf}
    \end{minipage}
    \hfill
    \begin{minipage}[t]{0.3\textwidth}
        \centering
        \includegraphics[width = \linewidth]{signal_diagrams/svcca_100_coco_raw_16384.pdf}
    \end{minipage}
    \hfill
    \begin{minipage}[t]{0.3\textwidth}
        \centering
        \includegraphics[width = \linewidth]{signal_diagrams/svcca_100_coco_filtered_16384.pdf}
    \end{minipage}
    \caption{Same plot as in Figure~\ref{fig:signalmain} but for the CKA, Unbiased CKA, SVCCA 10 and SVCCA 100 metrics. In addition to differences, we plot the alignment values for raw dense features and for the sparse features. We do an ablation study with dimension 8192.}
    \label{fig:signalcocoablate1}
\end{figure}

\clearpage

\begin{figure}[htbp]
    \centering

    \begin{minipage}[t]{0.3\textwidth}
        \centering
        \includegraphics[width = \linewidth]{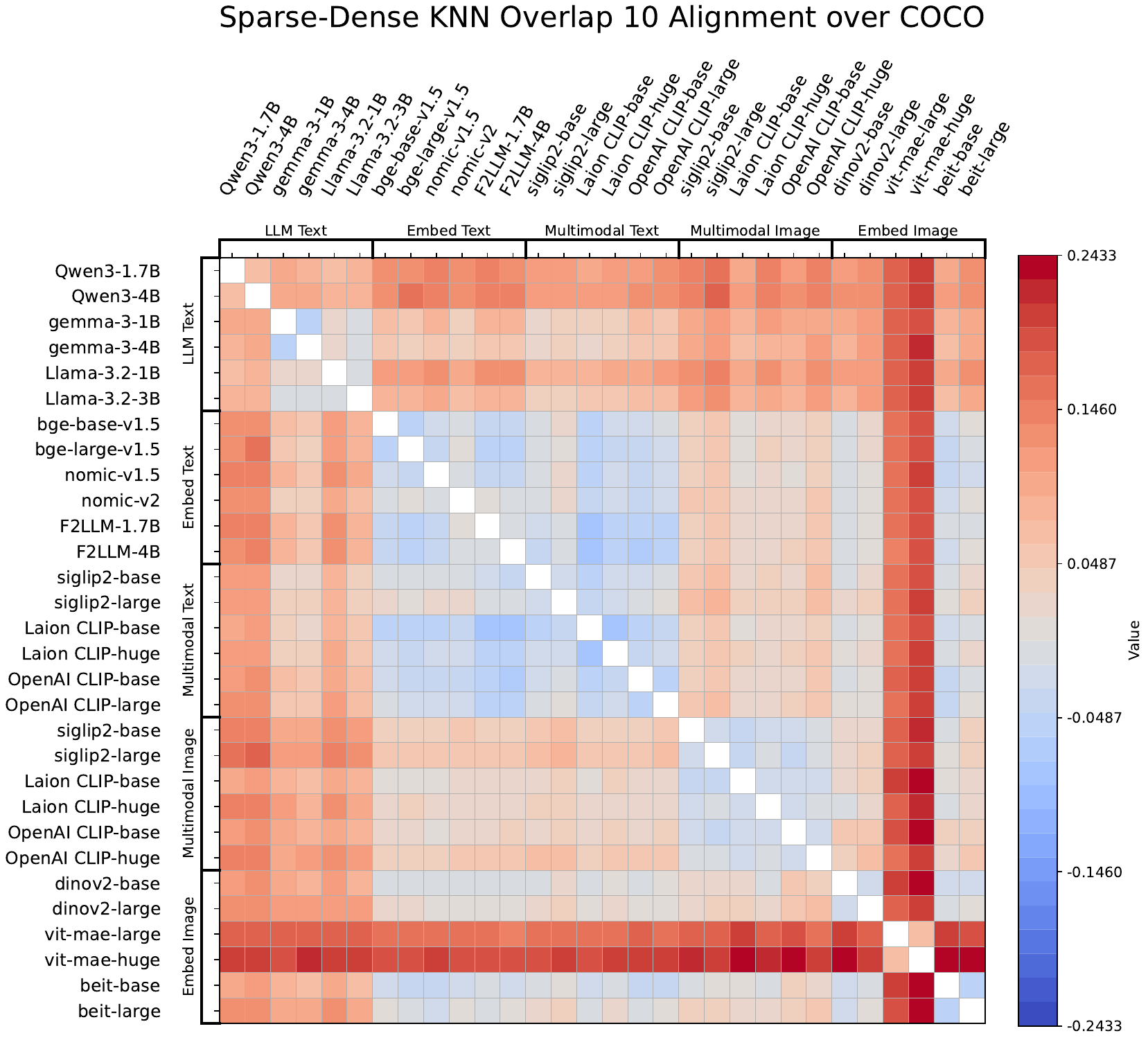}
    \end{minipage}
    \hfill
    \begin{minipage}[t]{0.3\textwidth}
        \centering
        \includegraphics[width = \linewidth]{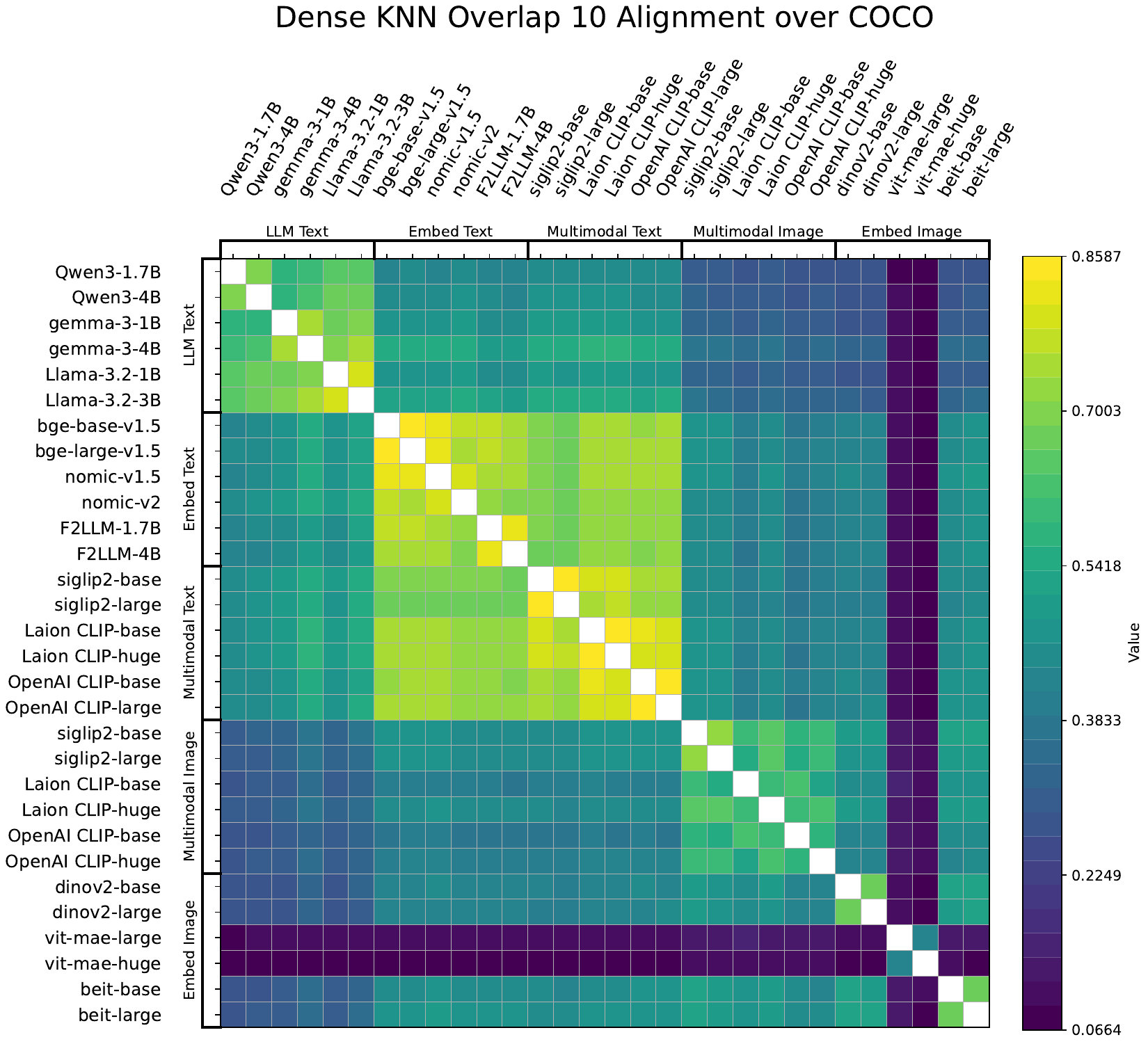}
    \end{minipage}
    \hfill
    \begin{minipage}[t]{0.3\textwidth}
        \centering
        \includegraphics[width = \linewidth]{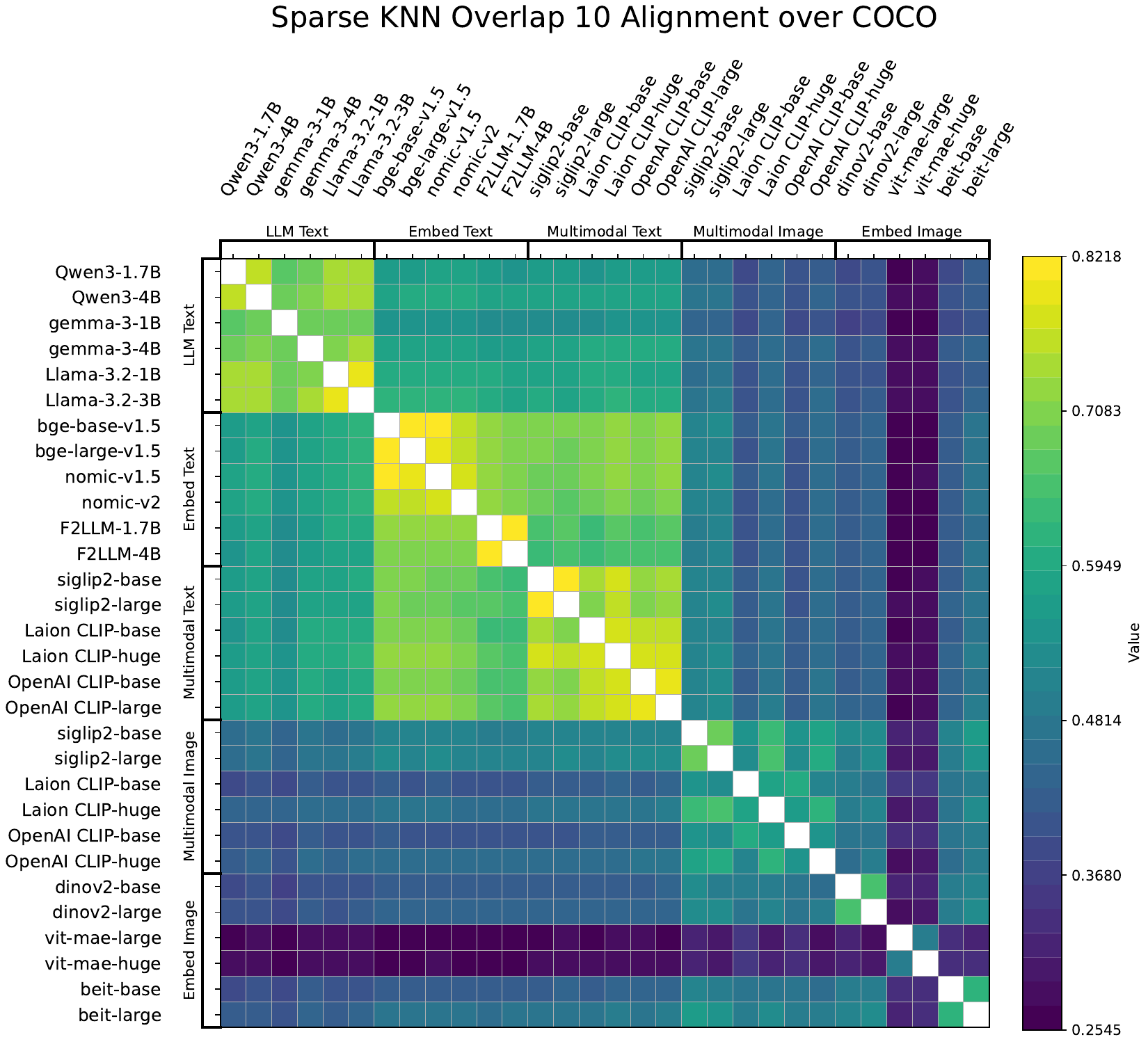}
    \end{minipage}
    
    \vspace{0.4cm}

        \begin{minipage}[t]{0.3\textwidth}
        \centering
        \includegraphics[width = \linewidth]{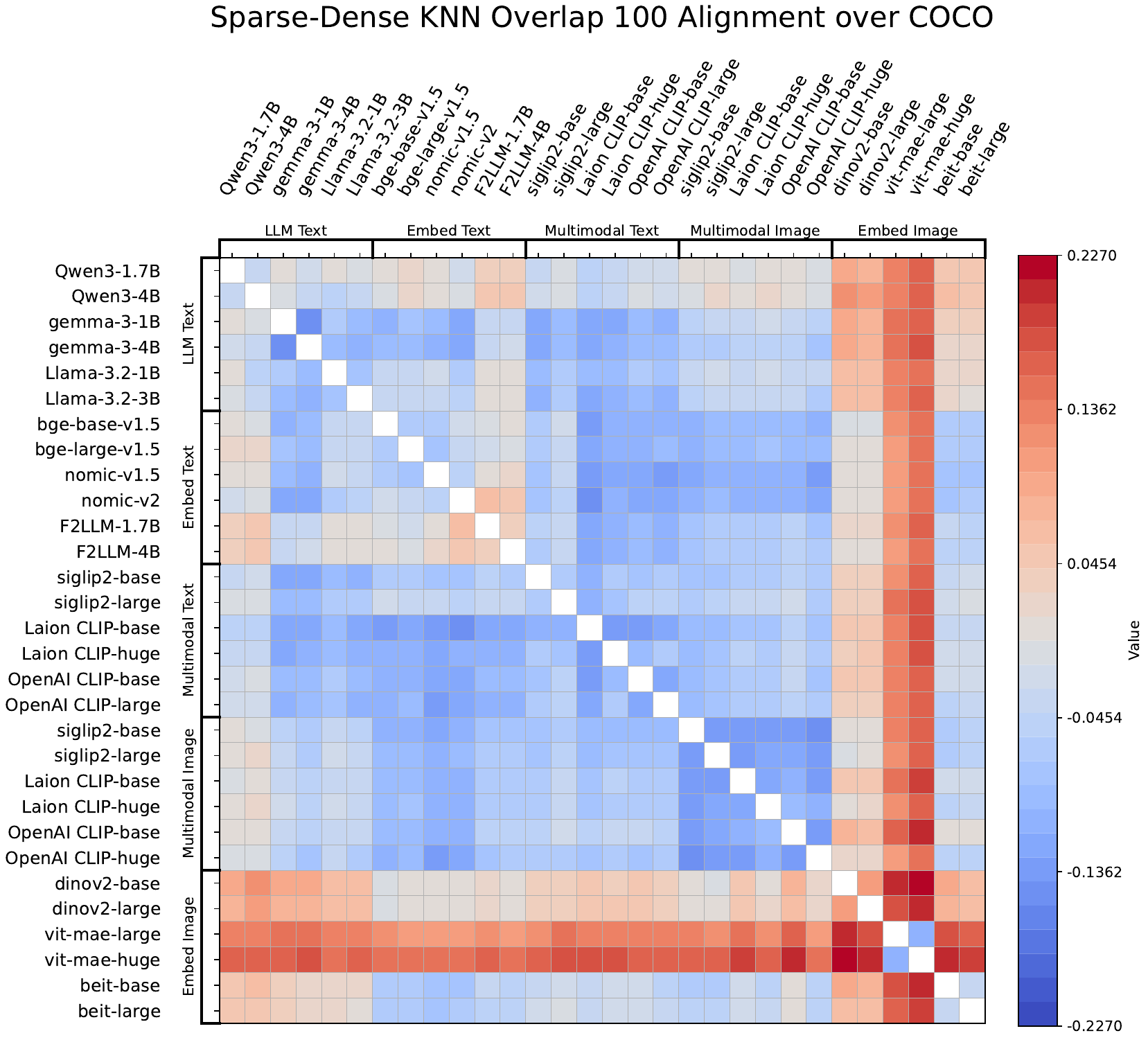}
    \end{minipage}
    \hfill
    \begin{minipage}[t]{0.3\textwidth}
        \centering
        \includegraphics[width = \linewidth]{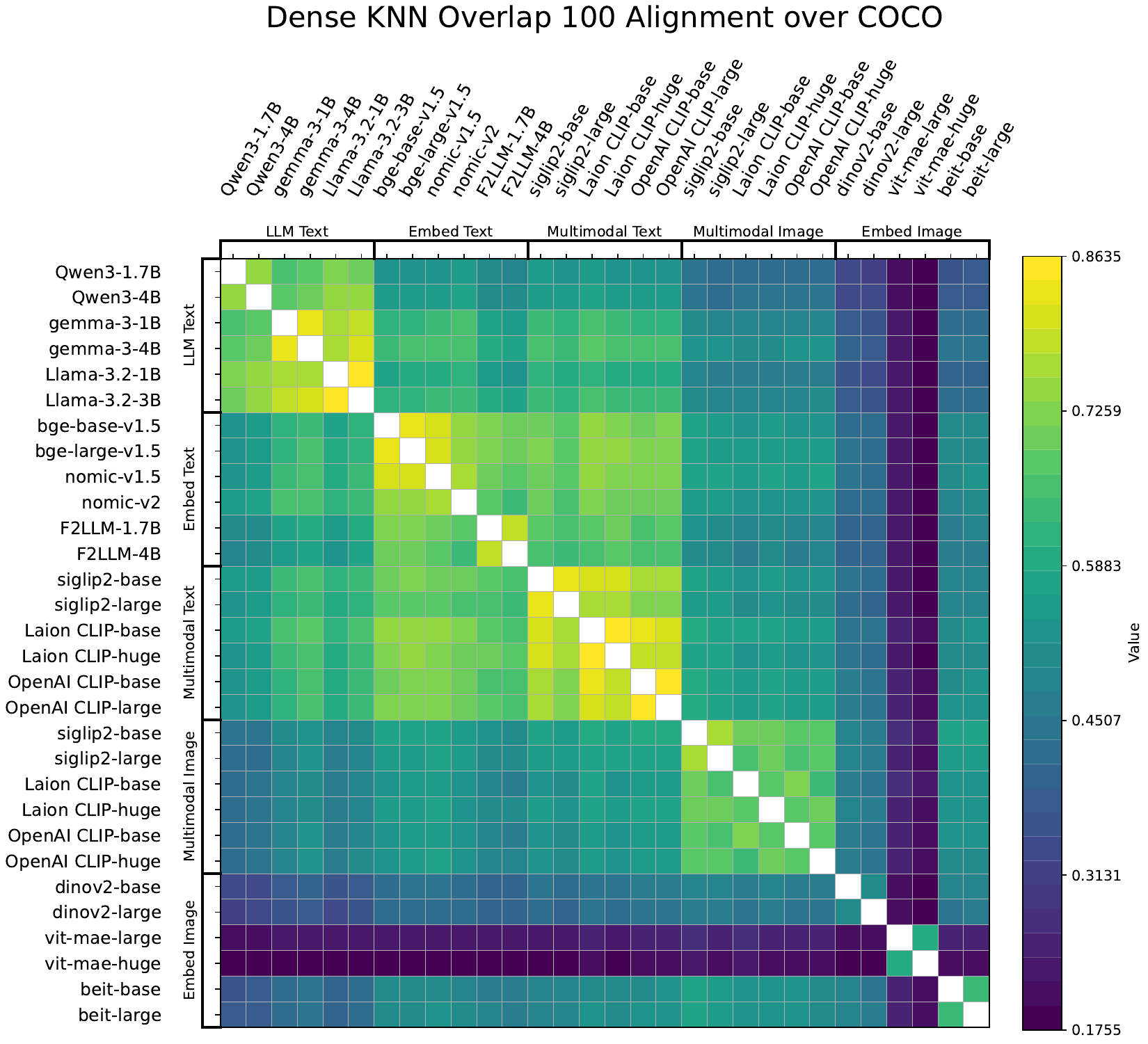}
    \end{minipage}
    \hfill
    \begin{minipage}[t]{0.3\textwidth}
        \centering
        \includegraphics[width = \linewidth]{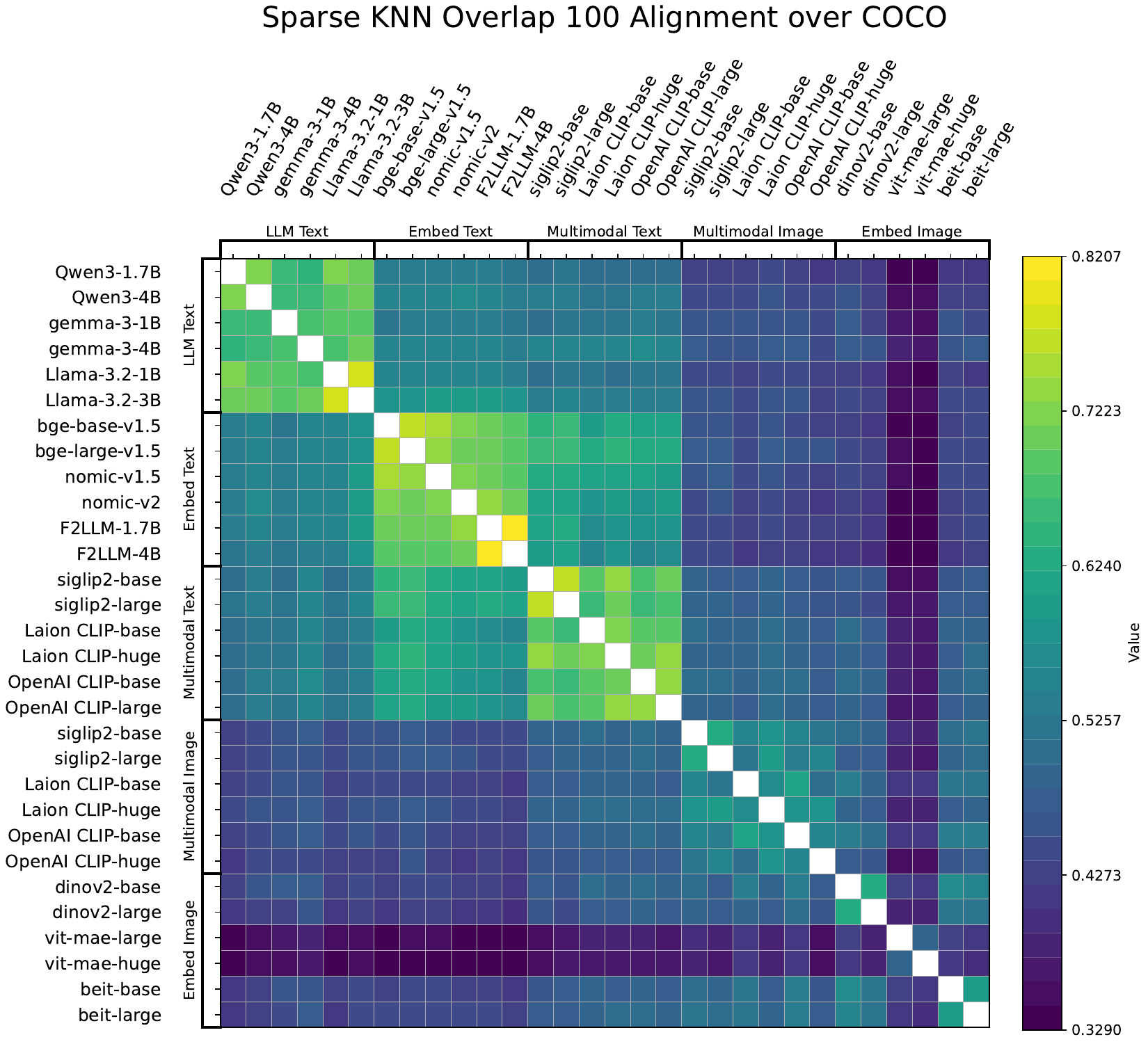}
    \end{minipage}

        \vspace{0.4cm}

        \begin{minipage}[t]{0.3\textwidth}
        \centering
        \includegraphics[width = \linewidth]{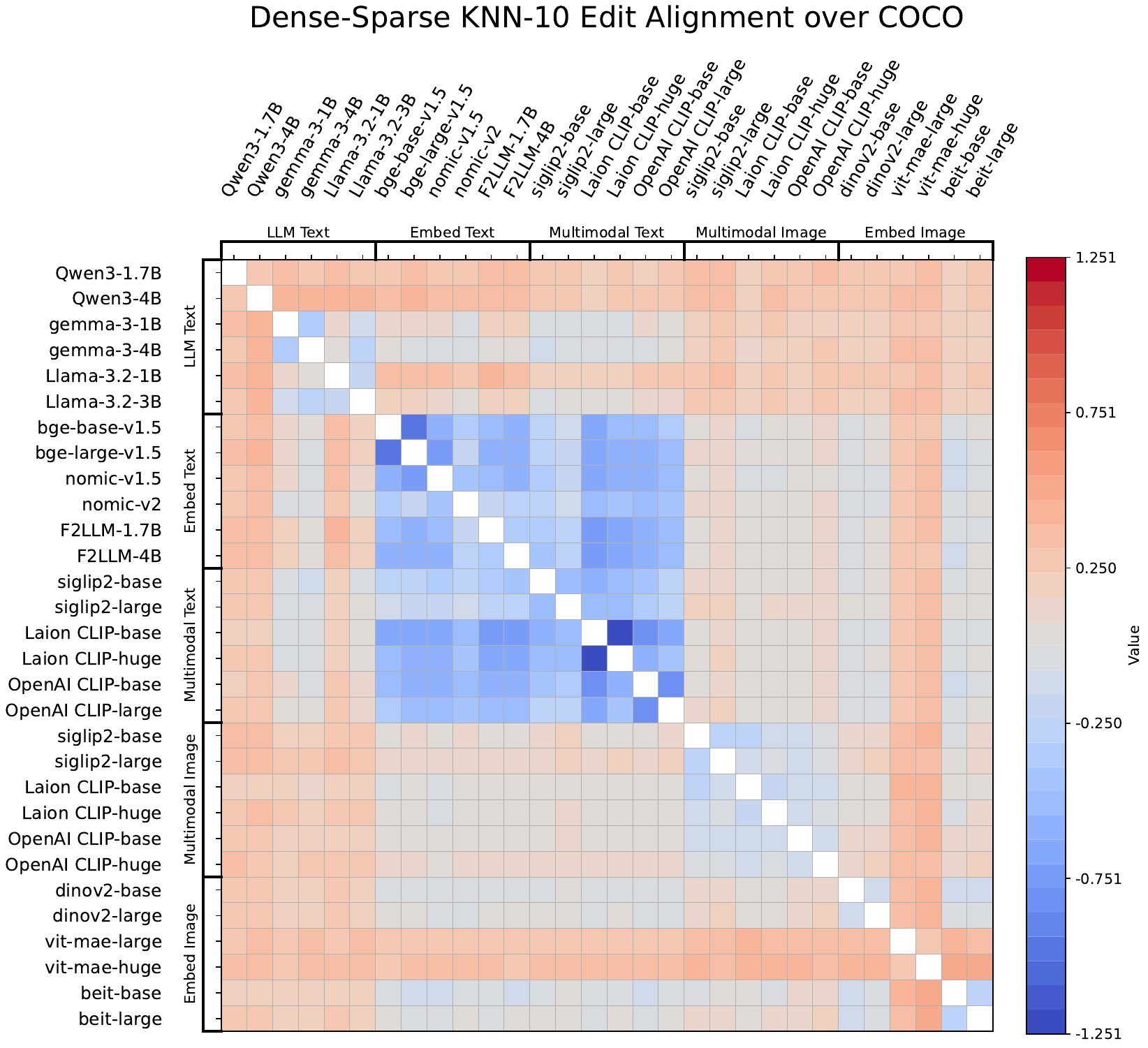}
    \end{minipage}
    \hfill
    \begin{minipage}[t]{0.3\textwidth}
        \centering
        \includegraphics[width = \linewidth]{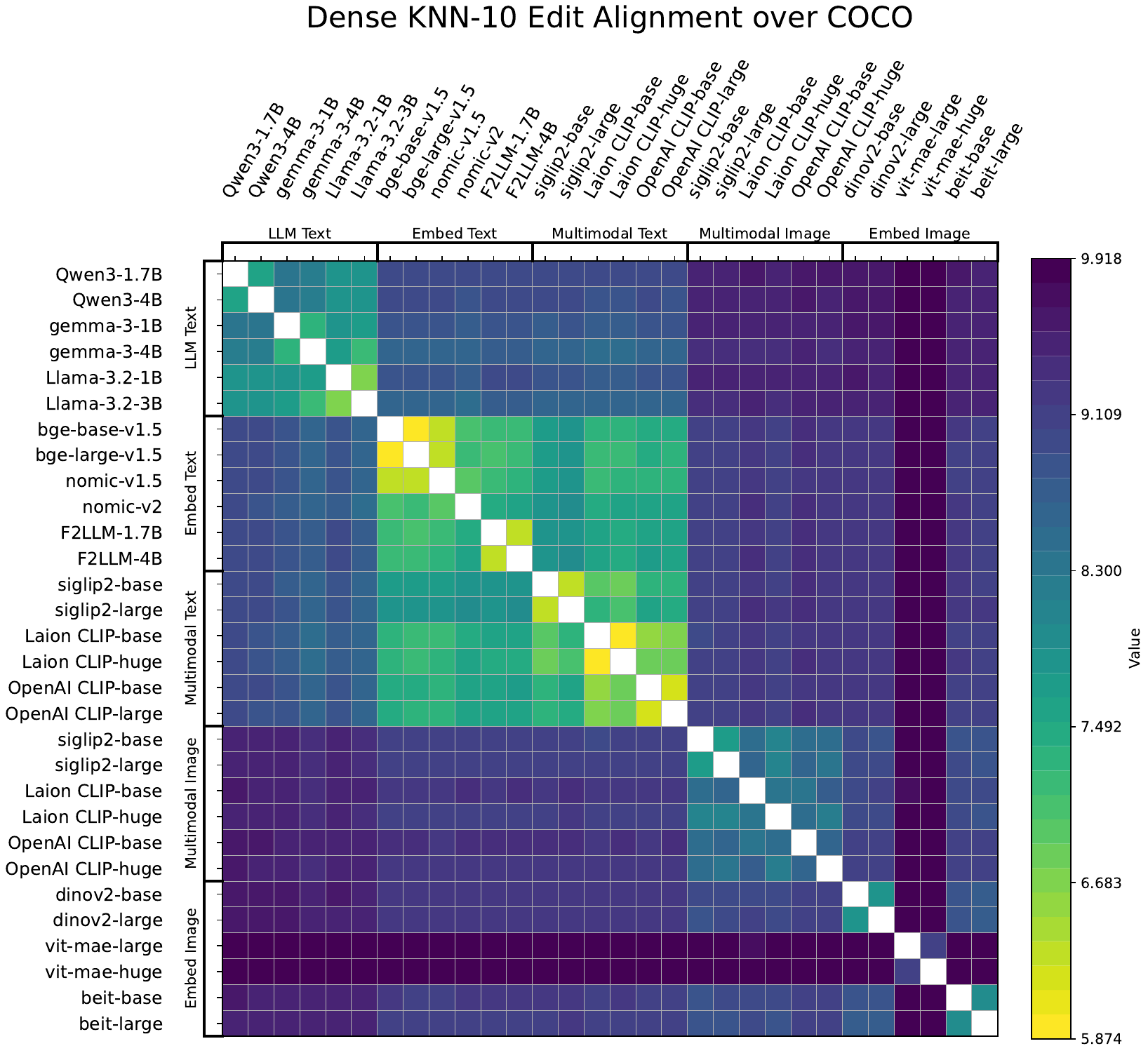}
    \end{minipage}
    \hfill
    \begin{minipage}[t]{0.3\textwidth}
        \centering
        \includegraphics[width = \linewidth]{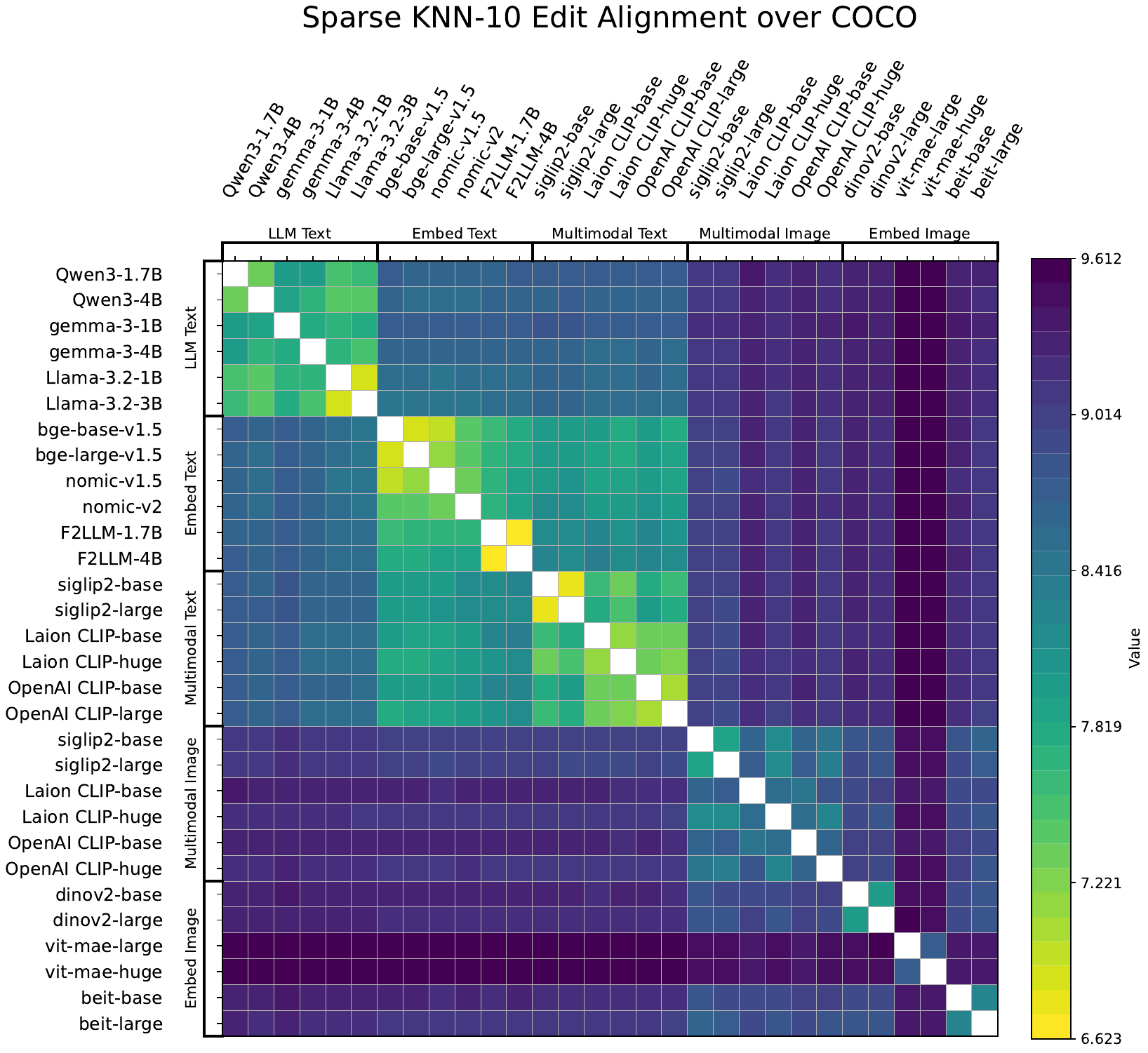}
    \end{minipage}

            \vspace{0.4cm}

    \begin{minipage}[t]{0.3\textwidth}
        \centering
        \includegraphics[width = \linewidth]{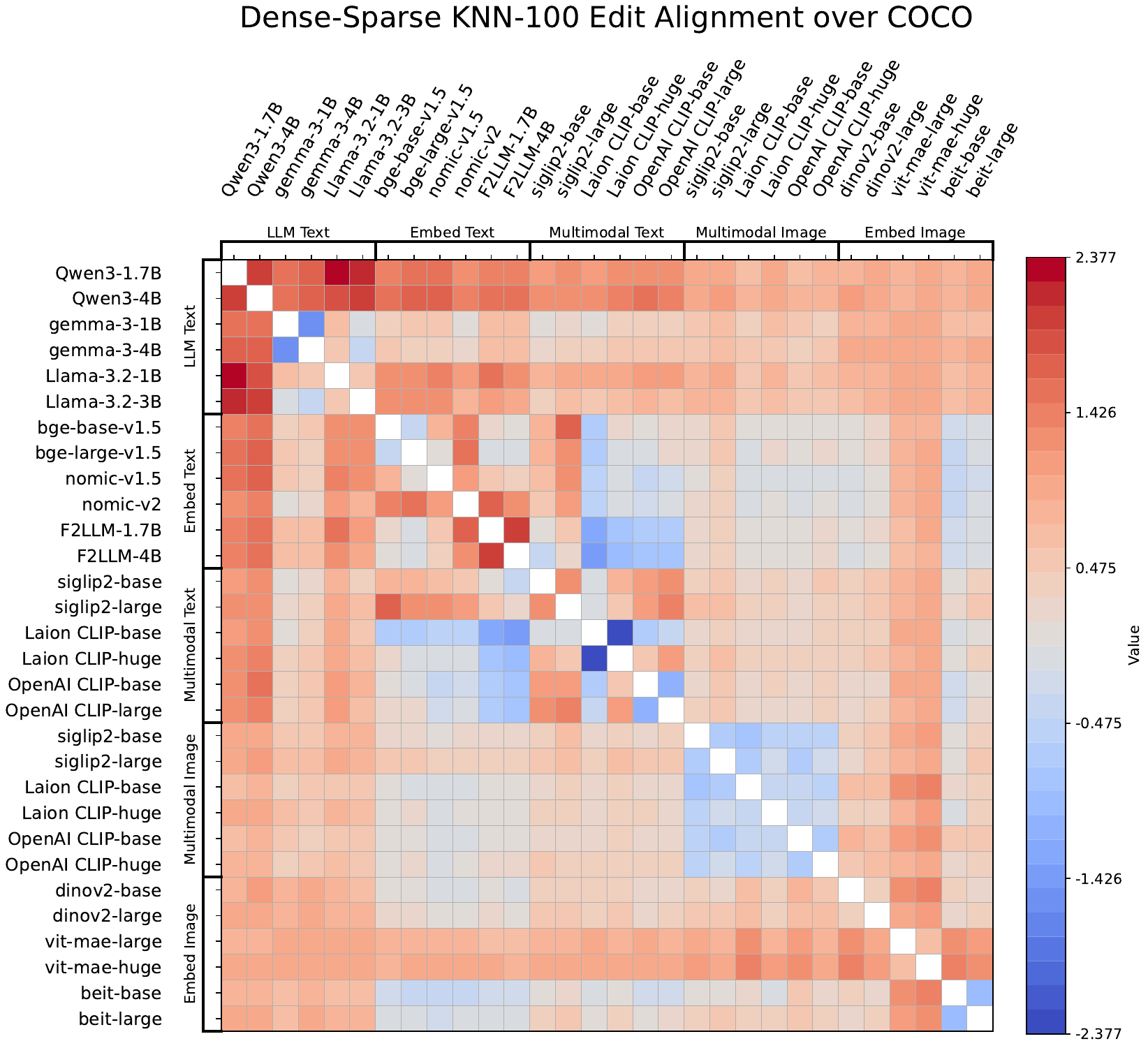}
    \end{minipage}
    \hfill
    \begin{minipage}[t]{0.3\textwidth}
        \centering
        \includegraphics[width = \linewidth]{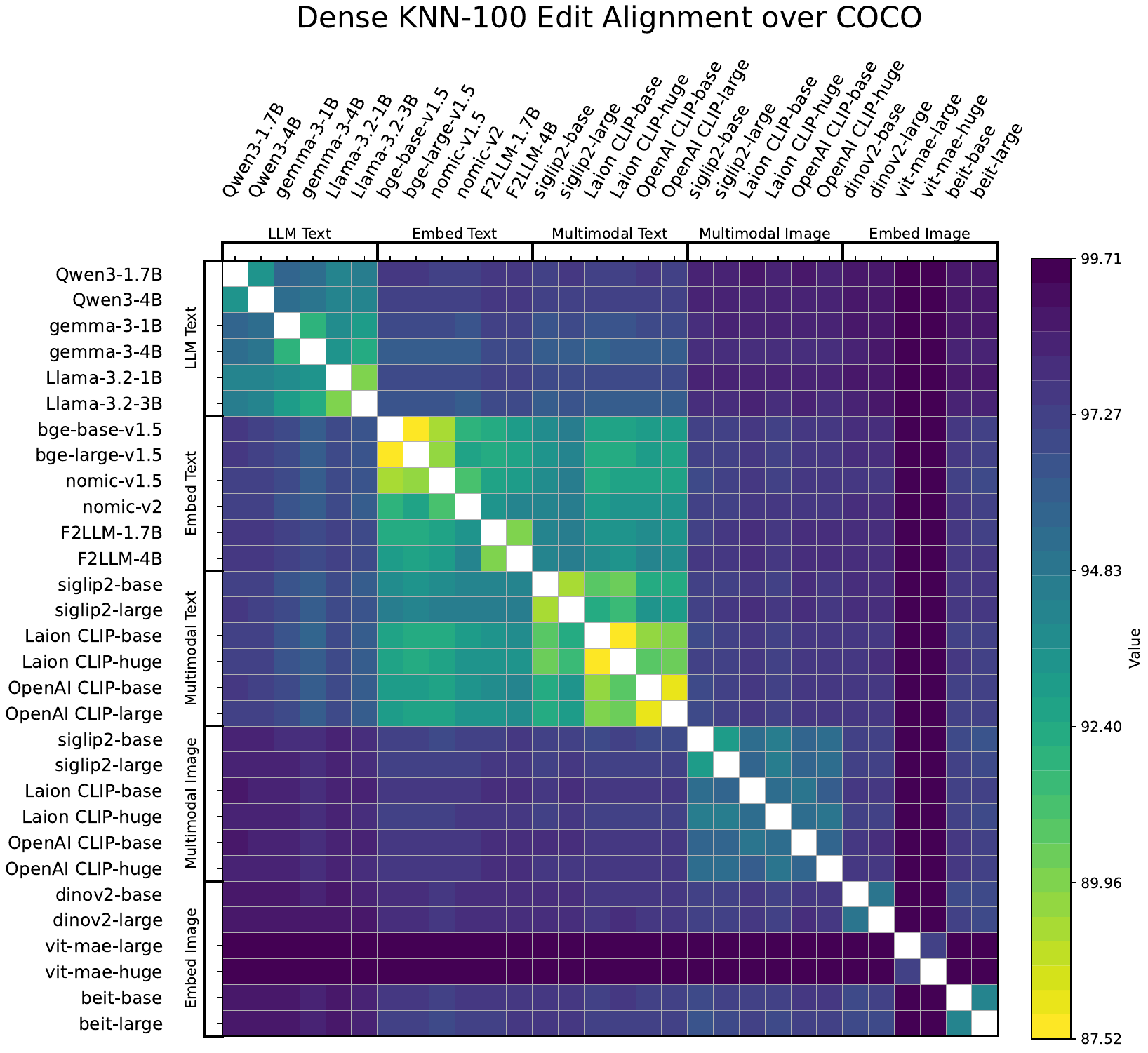}
    \end{minipage}
    \hfill
    \begin{minipage}[t]{0.3\textwidth}
        \centering
        \includegraphics[width = \linewidth]{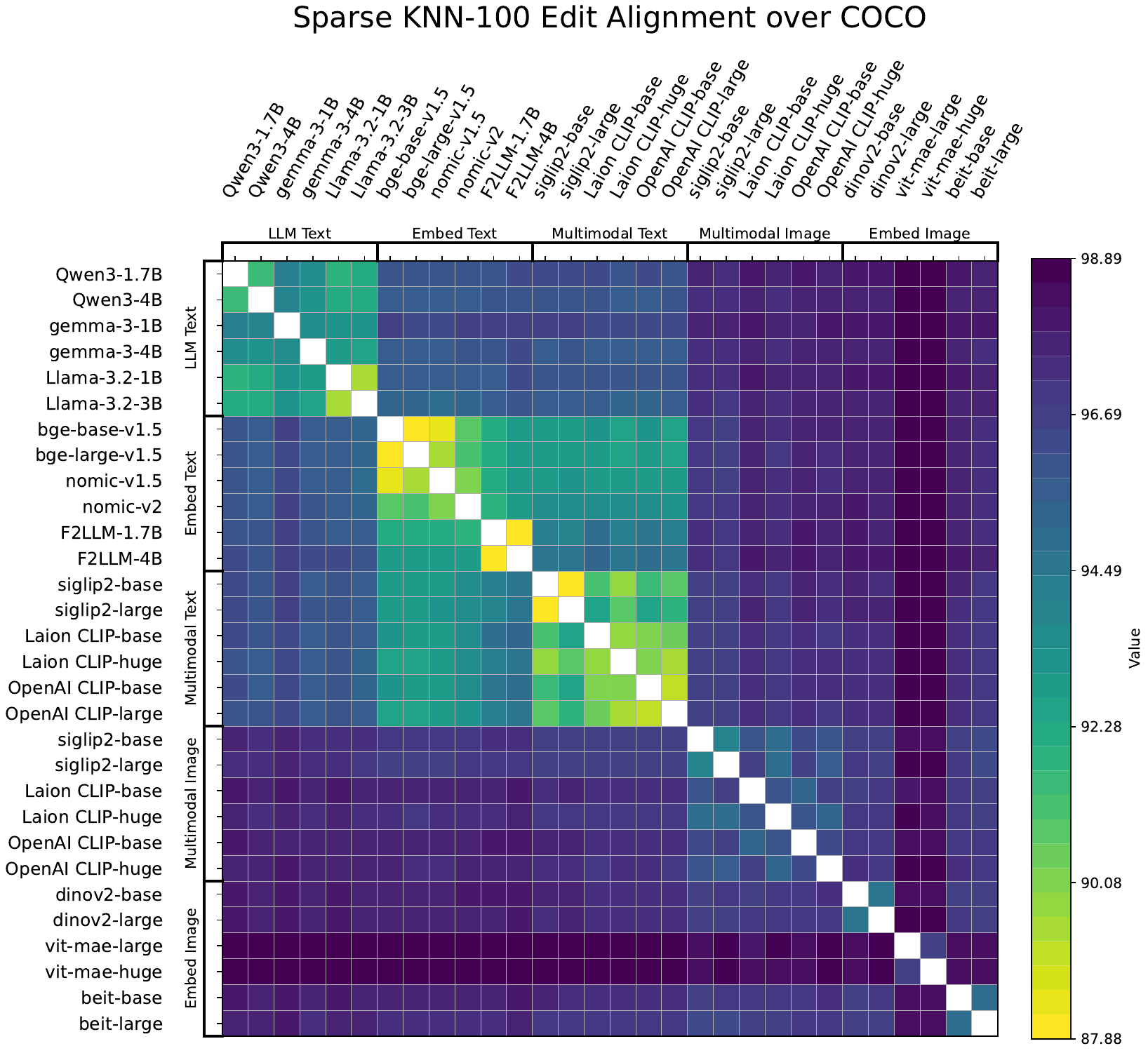}
    \end{minipage}
    \caption{Same plot as in Figure~\ref{fig:signalcoco2} but  with $d = 8192.$}
    \label{fig:signalcocoablate2}
\end{figure}

 \clearpage

\subsection{Signal: Correlation of Sparse Features}
\label{appendix:signalcorrelation}

\subsubsection{Experiments on COCO}

\textbf{With varying $k.$} 32 for text models, 64 for multimodal models, 128 for image-only models.
\vspace*{.3cm}

\begin{figure}[htbp]
    \centering

    \begin{minipage}[t]{0.3\textwidth}
        \centering
        \includegraphics[width = \linewidth]{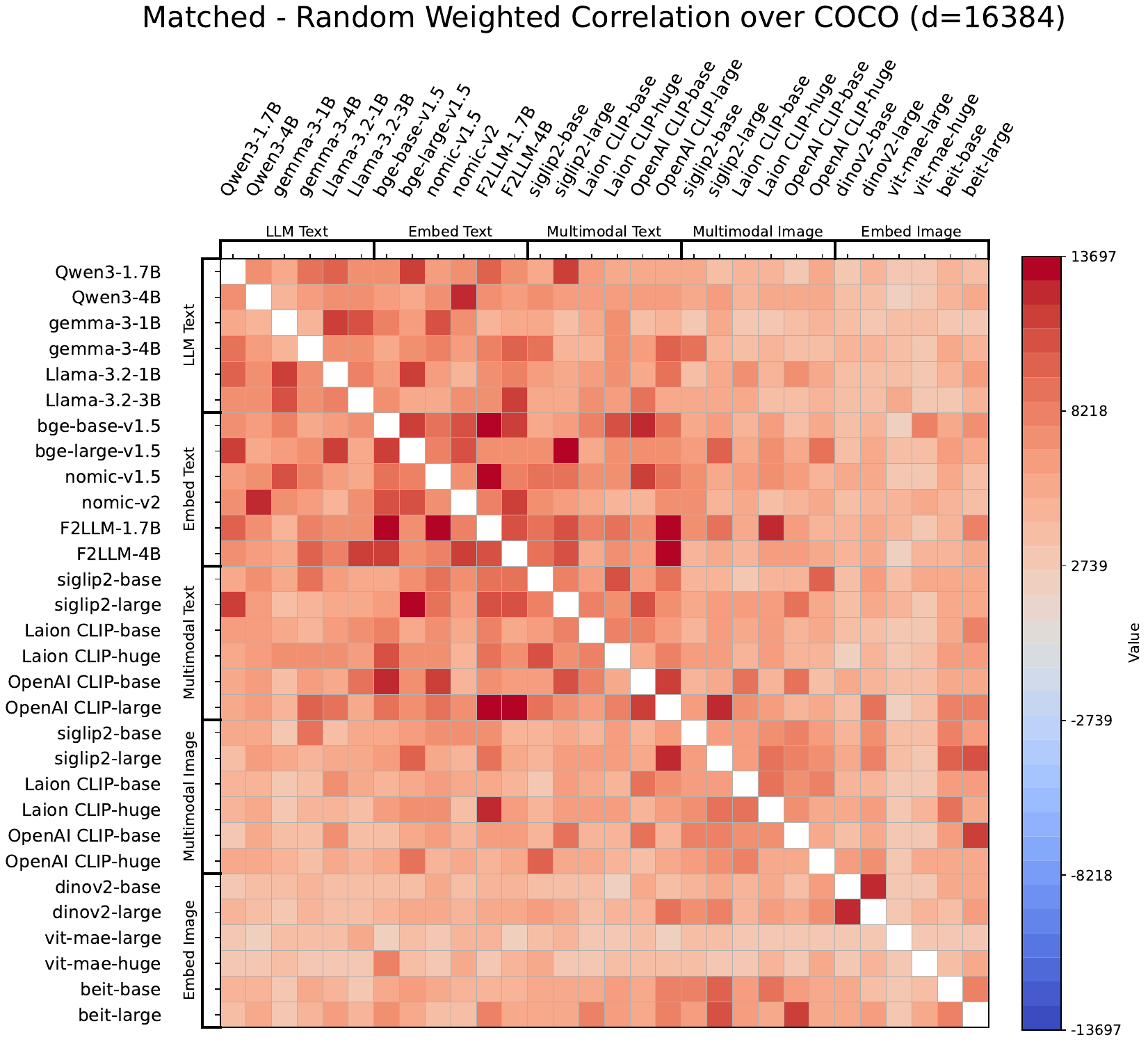}
    \end{minipage}
    \hfill
    \begin{minipage}[t]{0.3\textwidth}
        \centering
        \includegraphics[width = \linewidth]{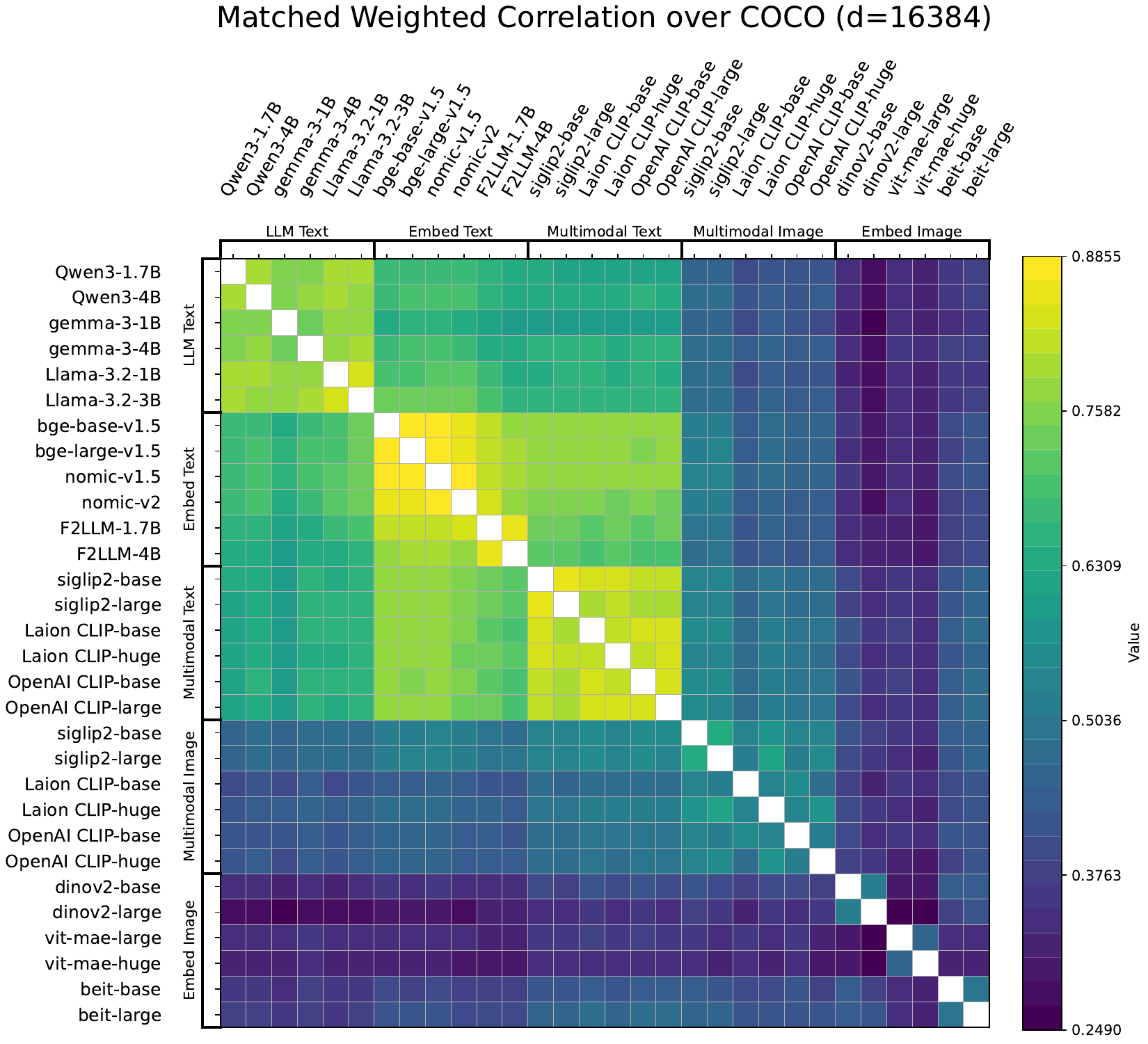}
    \end{minipage}
    \hfill
    \begin{minipage}[t]{0.3\textwidth}
        \centering
        \includegraphics[width = \linewidth]{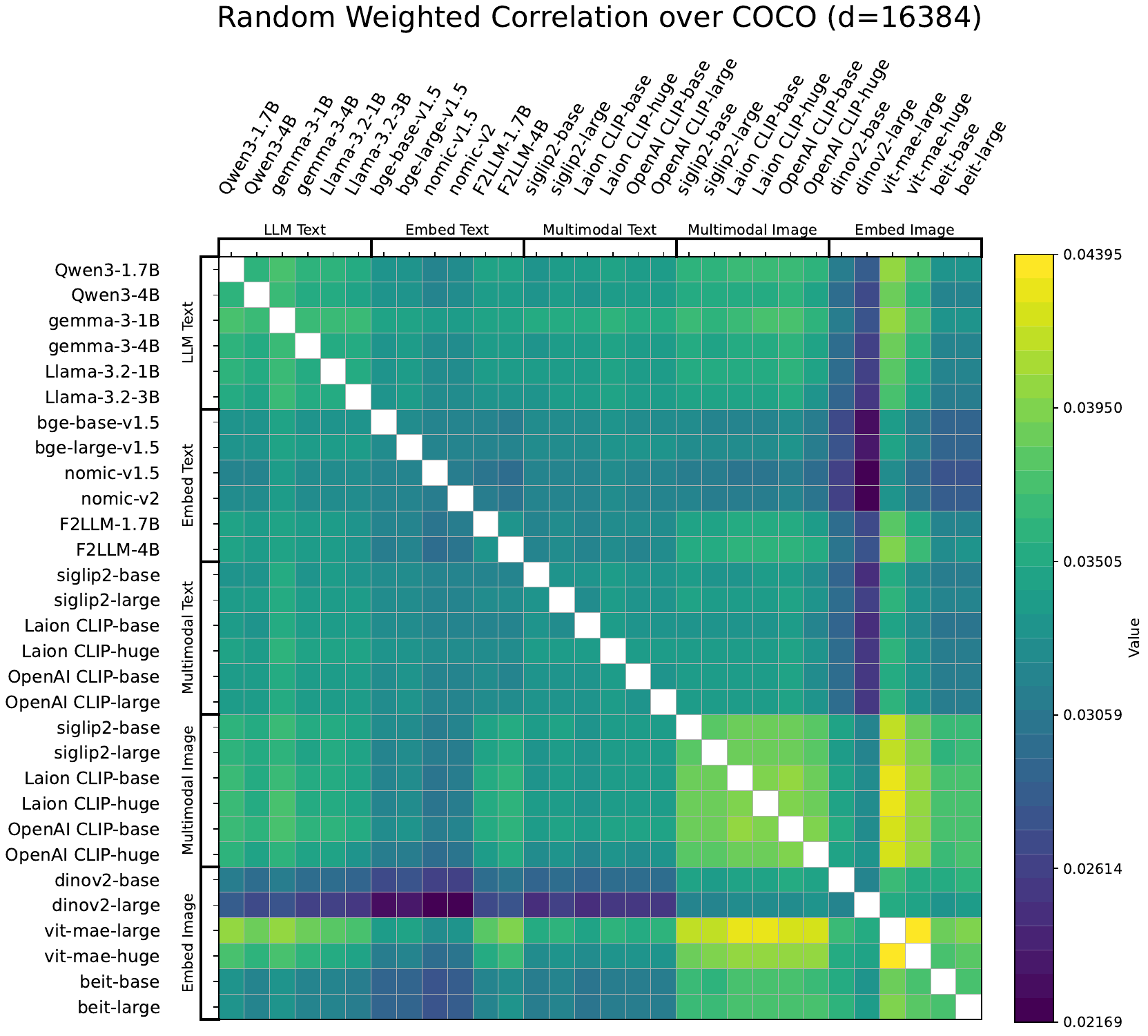}
    \end{minipage}
    
    \vspace{0.4cm}

        \begin{minipage}[t]{0.3\textwidth}
        \centering
        \includegraphics[width = \linewidth]{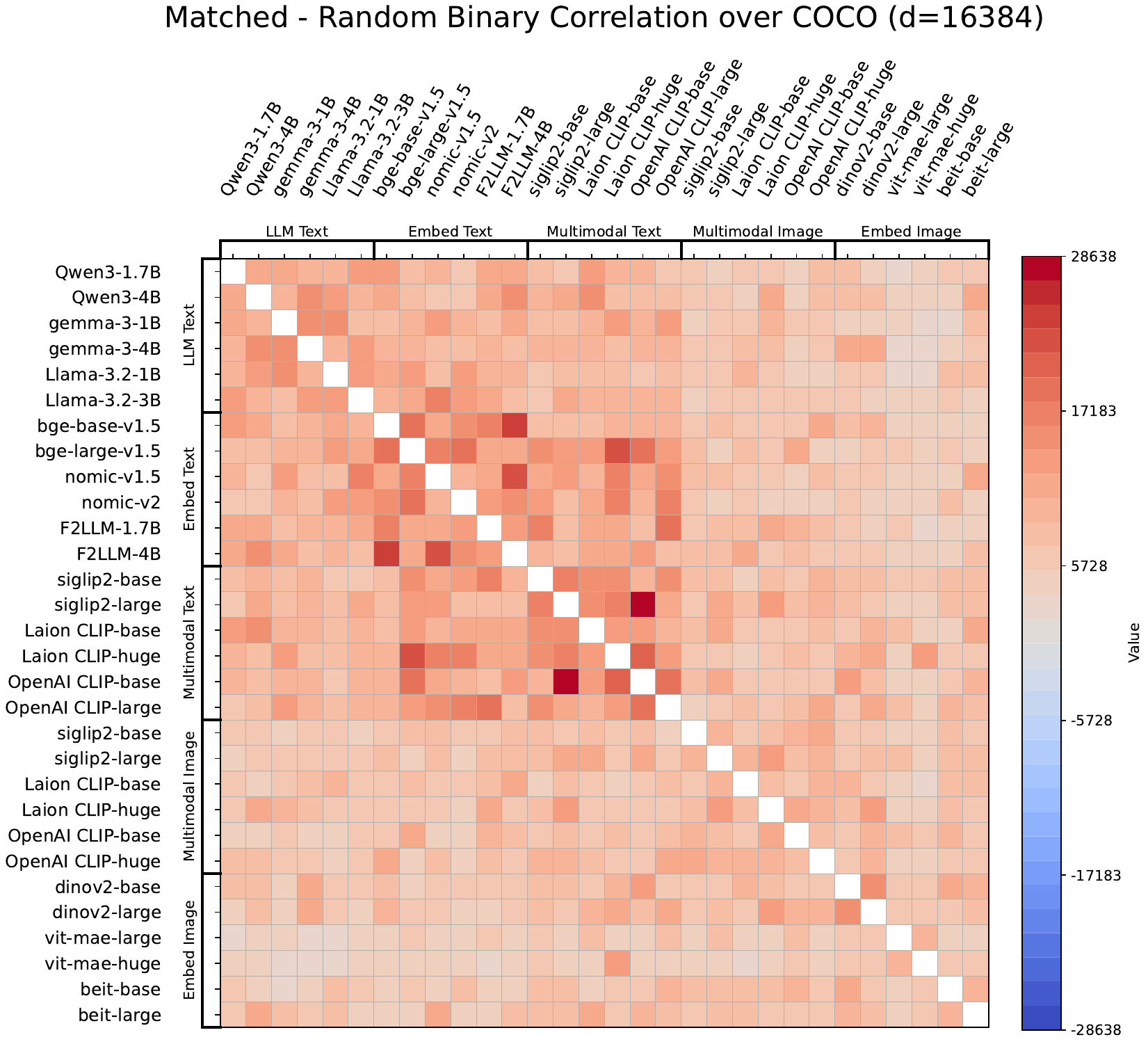}
    \end{minipage}
    \hfill
    \begin{minipage}[t]{0.3\textwidth}
        \centering
        \includegraphics[width = \linewidth]{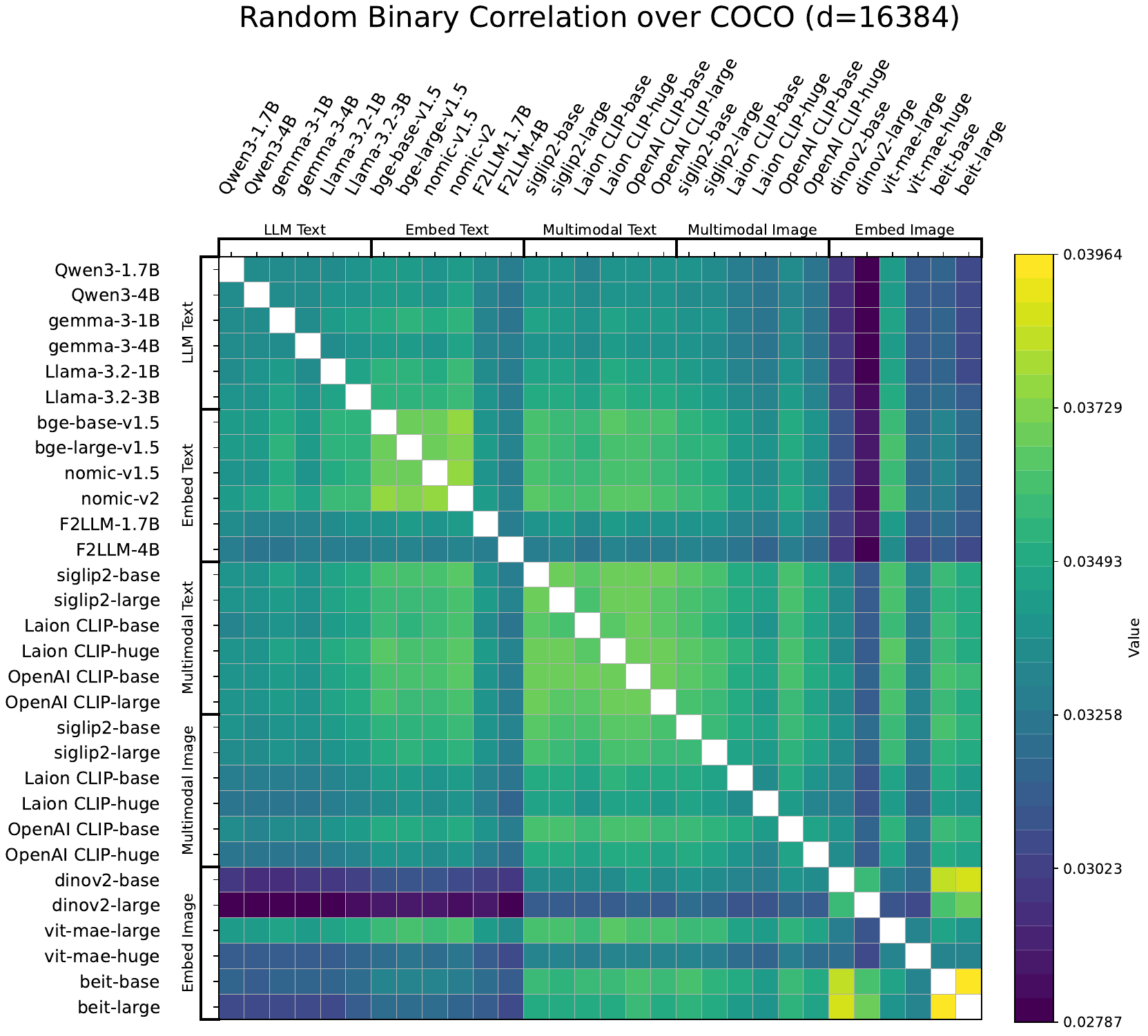}
    \end{minipage}
    \hfill
    \begin{minipage}[t]{0.3\textwidth}
        \centering
        \includegraphics[width = \linewidth]{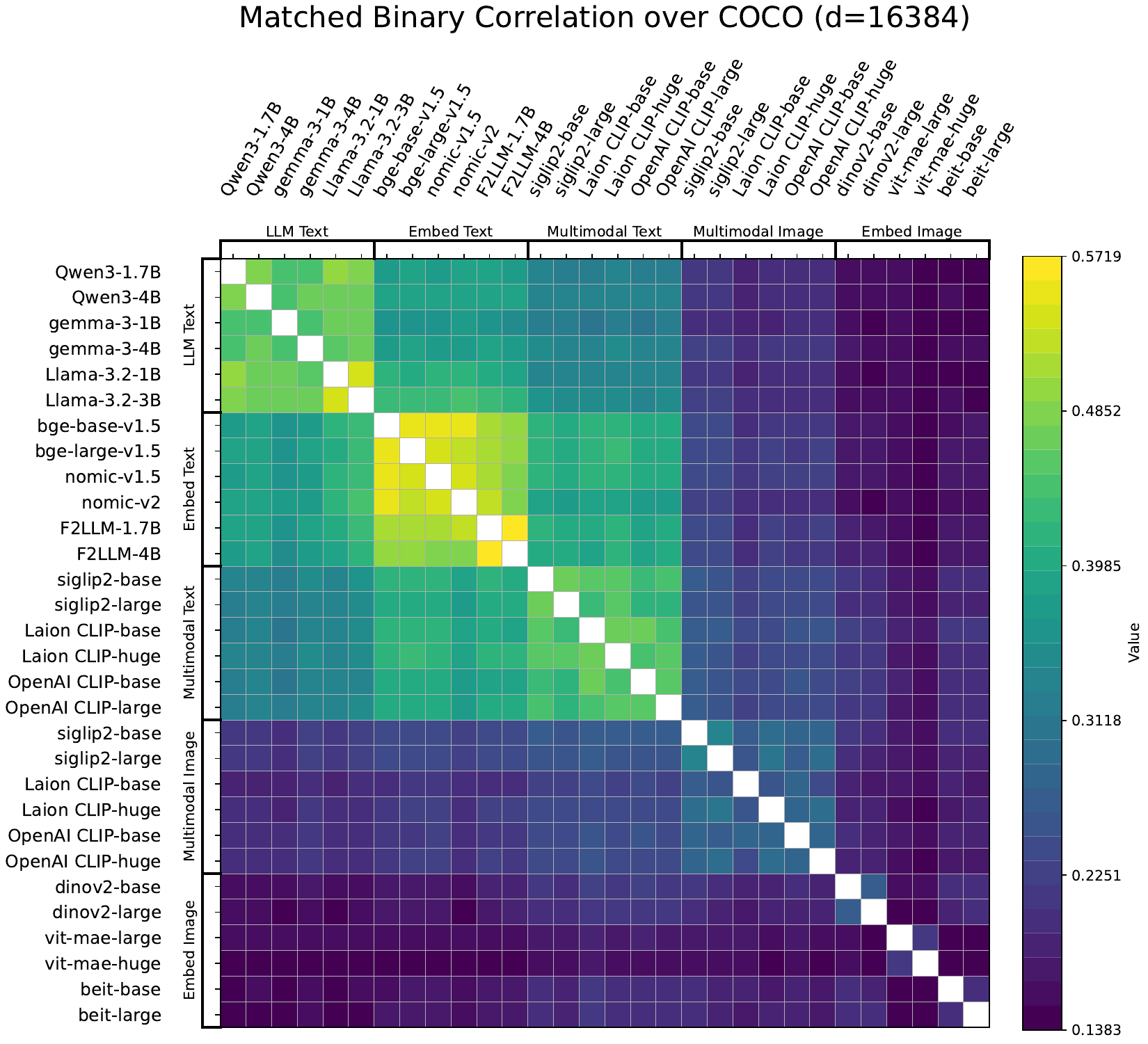}
    \end{minipage}

        \vspace{0.4cm}
    \caption{Plotting Experiment 2 from Section~\ref{sec:signalexperiment}. Concretely, from left to right, we have $[\corr(Z_1, Z_2) - \mathbb{E}_\Phi\corr_\Phi(Z_1,Z_2)]/\mathbb{V}_\Phi(\corr_\Phi(Z_1,Z_2))^{1/2},$ then in the middle plot $\corr(Z_1, Z_2)$ and in the right-most plot $\mathbb{E}_\Phi\corr_\Phi(Z_1,Z_2).$ Here, $\mathbb{E},\mathbb{V}$ correspond to variance and expectation.}
    \label{fig:signalcorrelationcocovar}
\end{figure}

\clearpage
\subsubsection{Experiments on CC3M}
\textbf{With varying $k.$} 32 for text models, 64 for multimodal models, 128 for image-only models.

\begin{figure}[!htb]
    \centering

    \begin{minipage}[t]{0.3\textwidth}
        \centering
        \includegraphics[width = \linewidth]{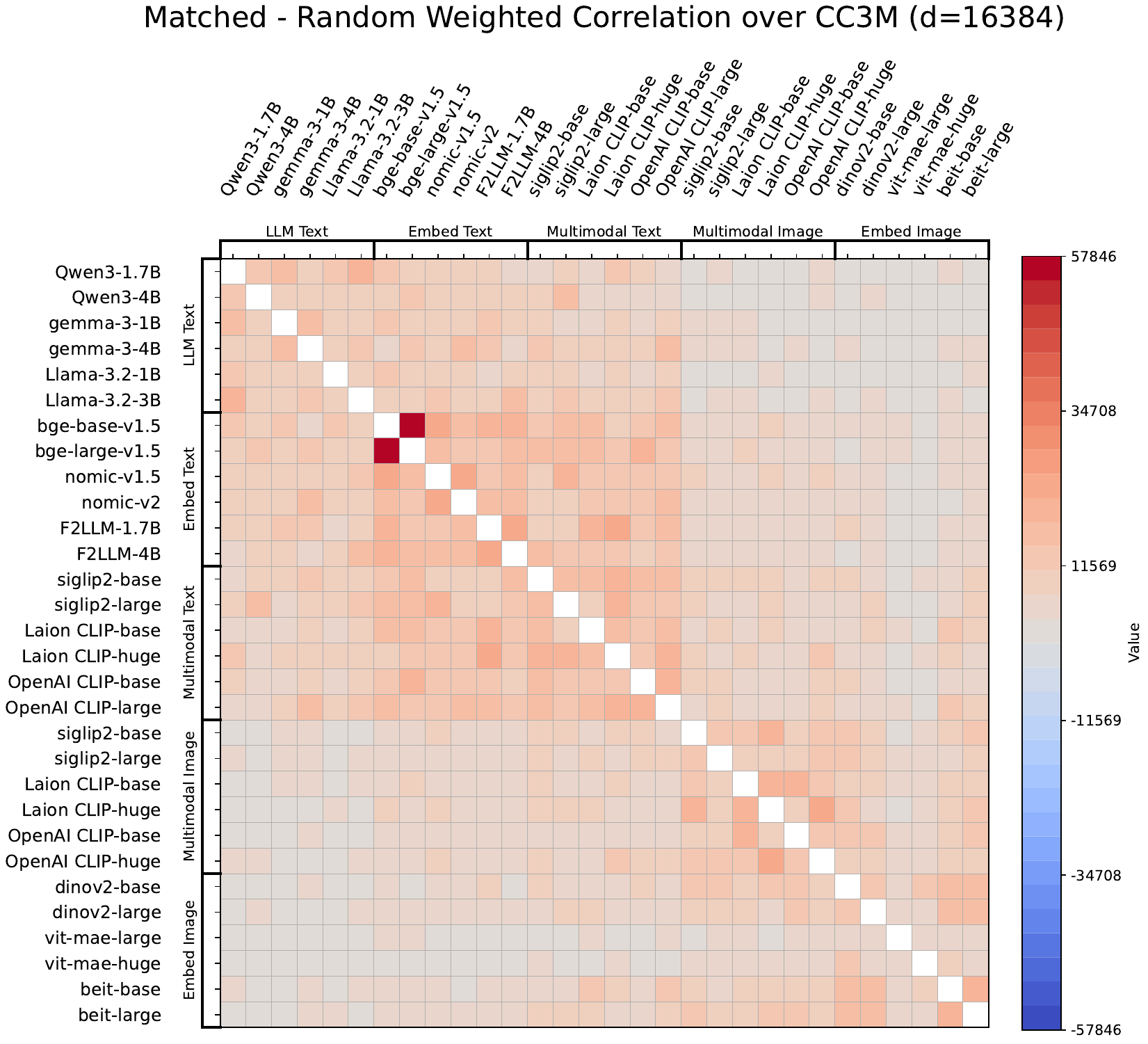}
    \end{minipage}
    \hfill
    \begin{minipage}[t]{0.3\textwidth}
        \centering
        \includegraphics[width = \linewidth]{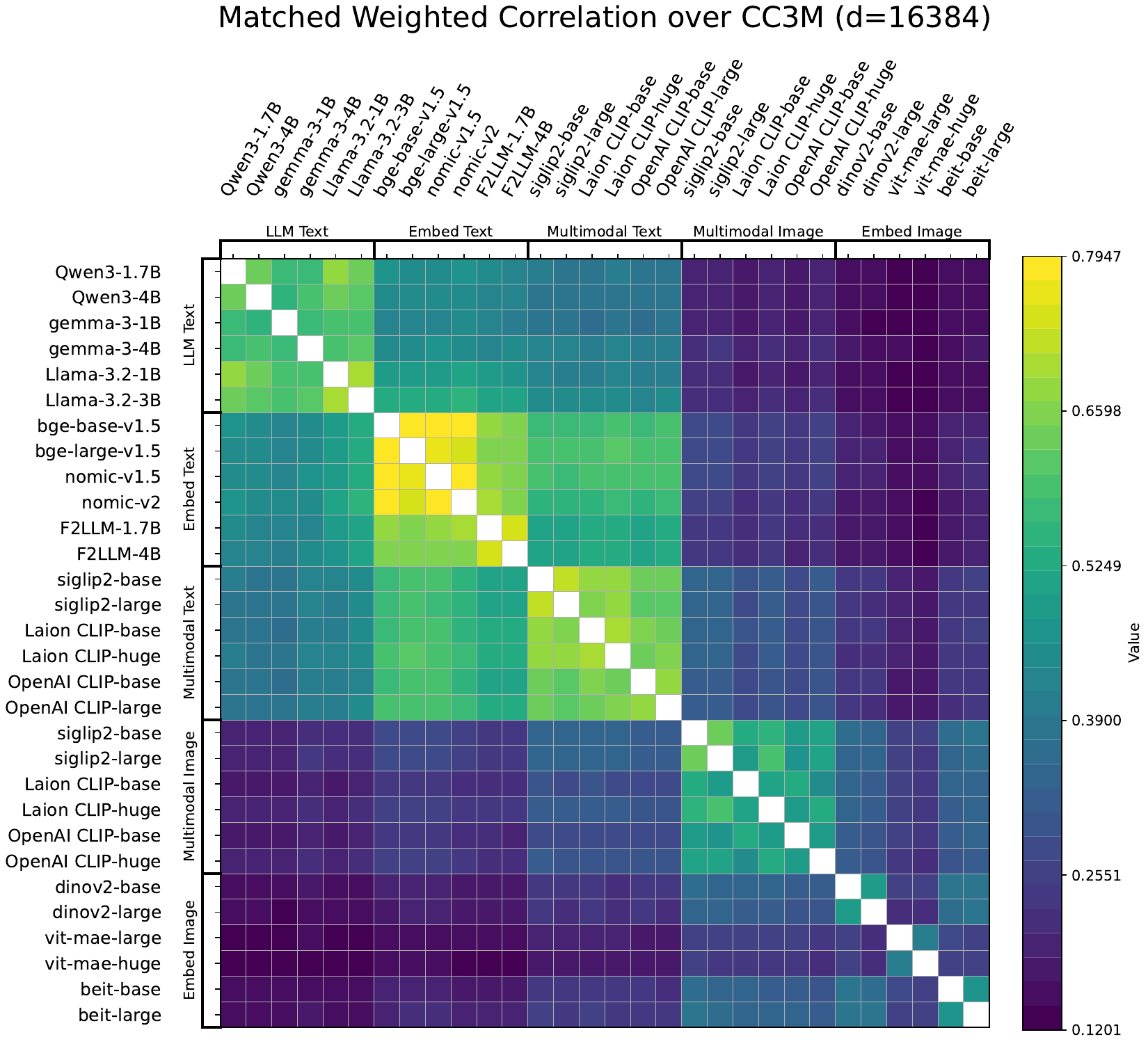}
    \end{minipage}
    \hfill
    \begin{minipage}[t]{0.3\textwidth}
        \centering
        \includegraphics[width = \linewidth]{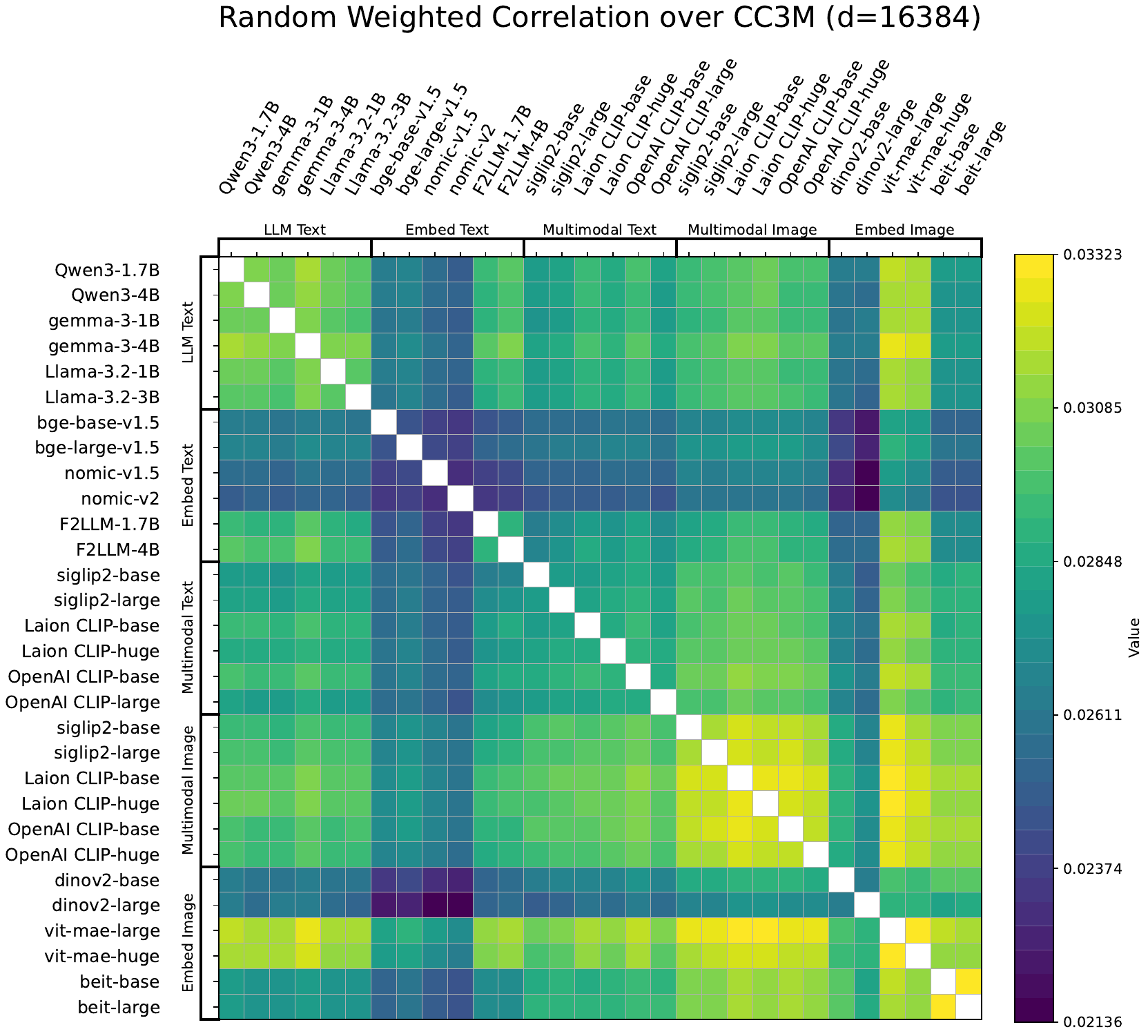}
    \end{minipage}
    
    \vspace{0.4cm}

        \begin{minipage}[t]{0.3\textwidth}
        \centering
        \includegraphics[width = \linewidth]{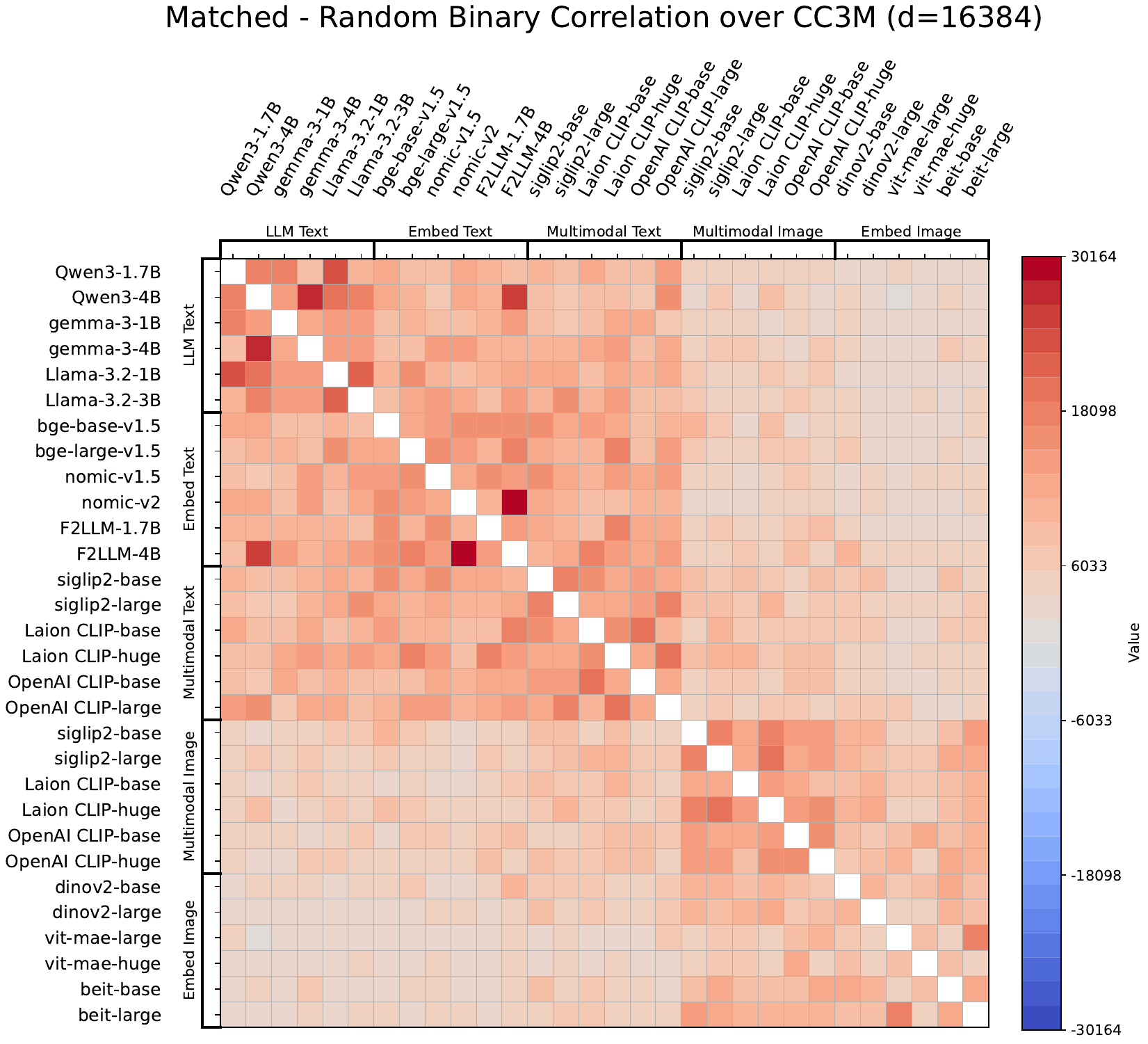}
    \end{minipage}
    \hfill
    \begin{minipage}[t]{0.3\textwidth}
        \centering
        \includegraphics[width = \linewidth]{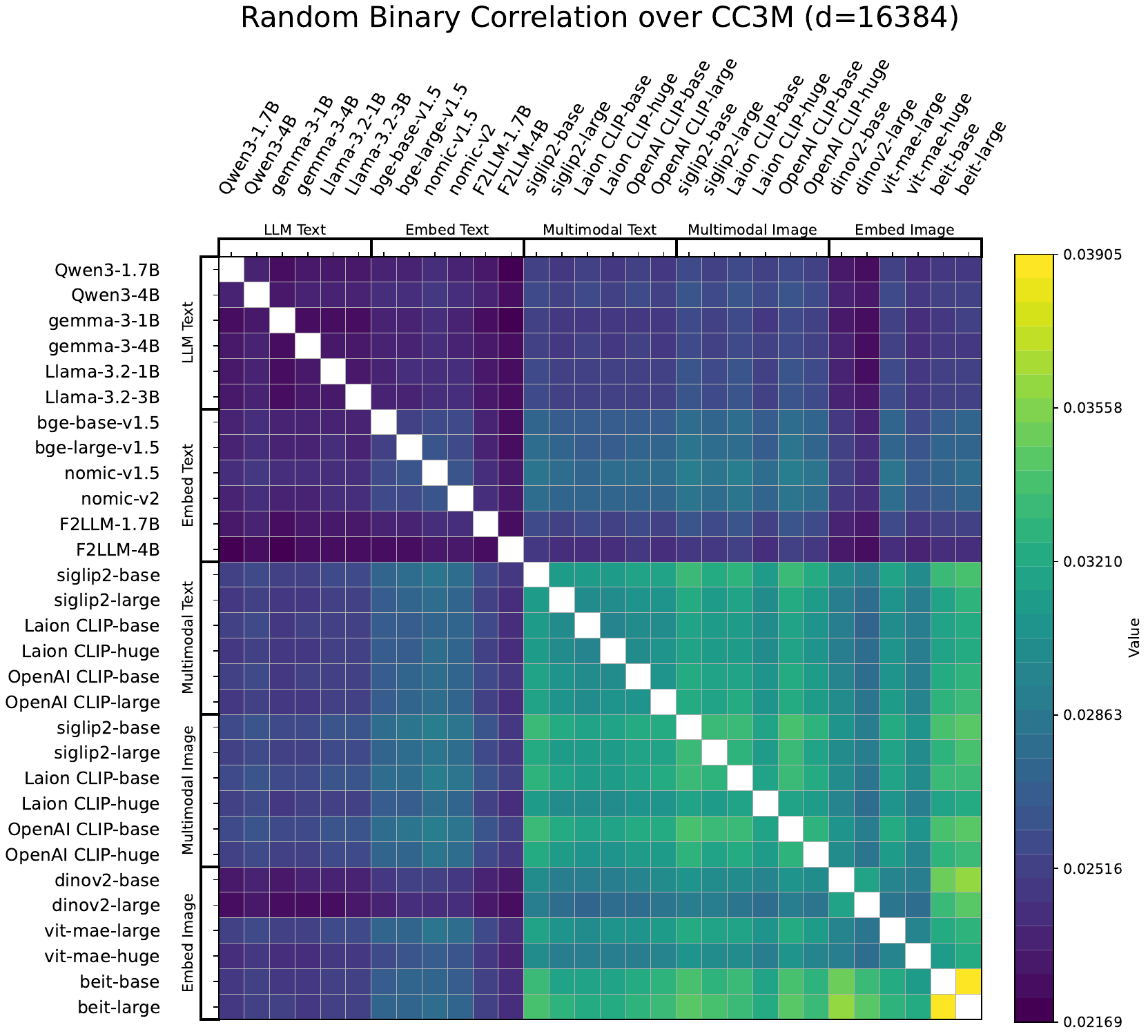}
    \end{minipage}
    \hfill
    \begin{minipage}[t]{0.3\textwidth}
        \centering
        \includegraphics[width = \linewidth]{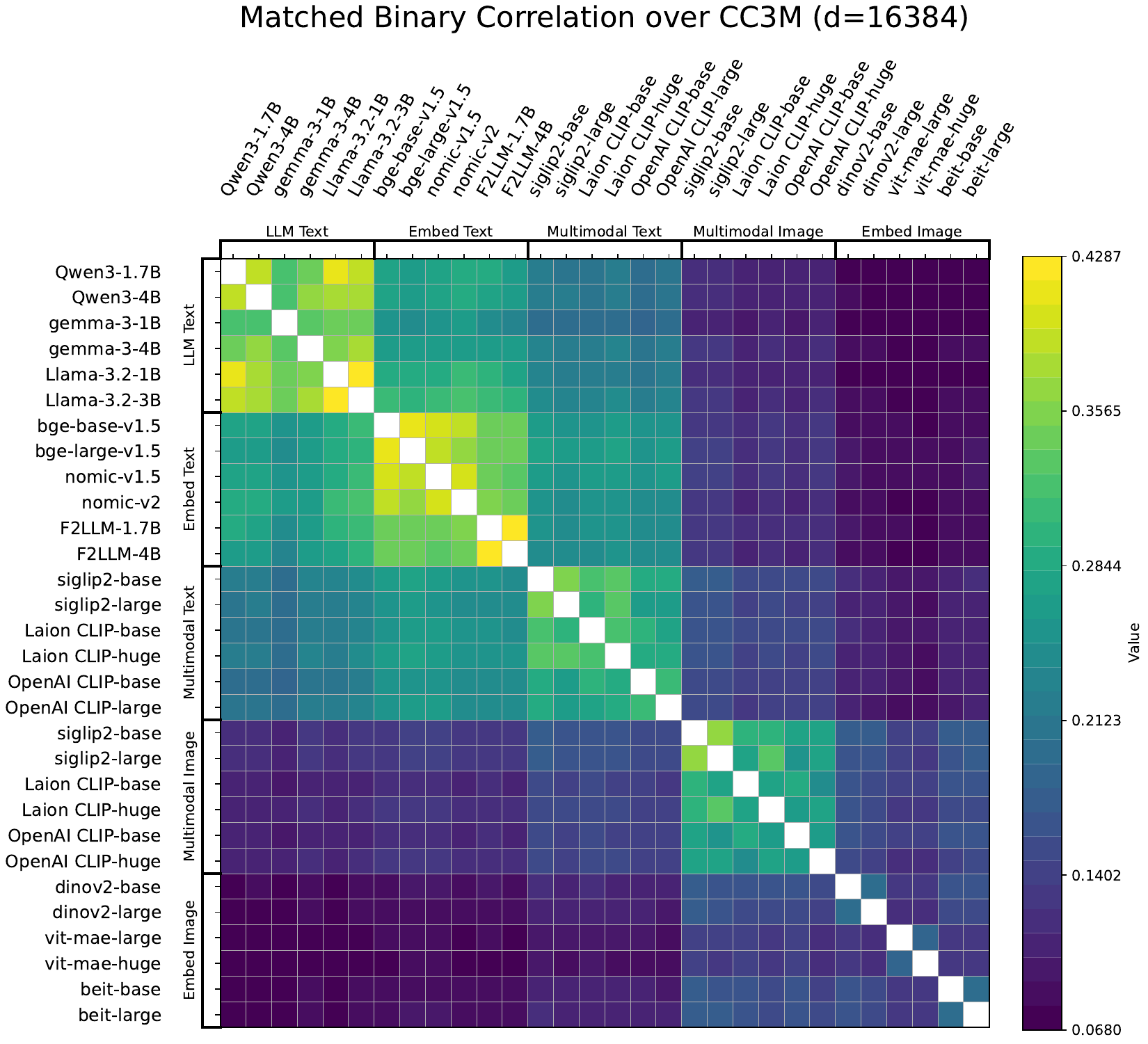}
    \end{minipage}

        \vspace{0.4cm}
    \caption{Same plot as in Figure~\ref{fig:signalcorrelationcocovar} but over CC3M.}
    \label{fig:signalcorrelationcc3mvar}
\end{figure}

\clearpage

\subsubsection{Experiments on Visual Genome}

\textbf{With varying $k.$} 32 for text models, 64 for multimodal models, 128 for image-only models.
\vspace*{.3cm}
\begin{figure}[htbp]
    \centering

    \begin{minipage}[t]{0.3\textwidth}
        \centering
        \includegraphics[width = \linewidth]{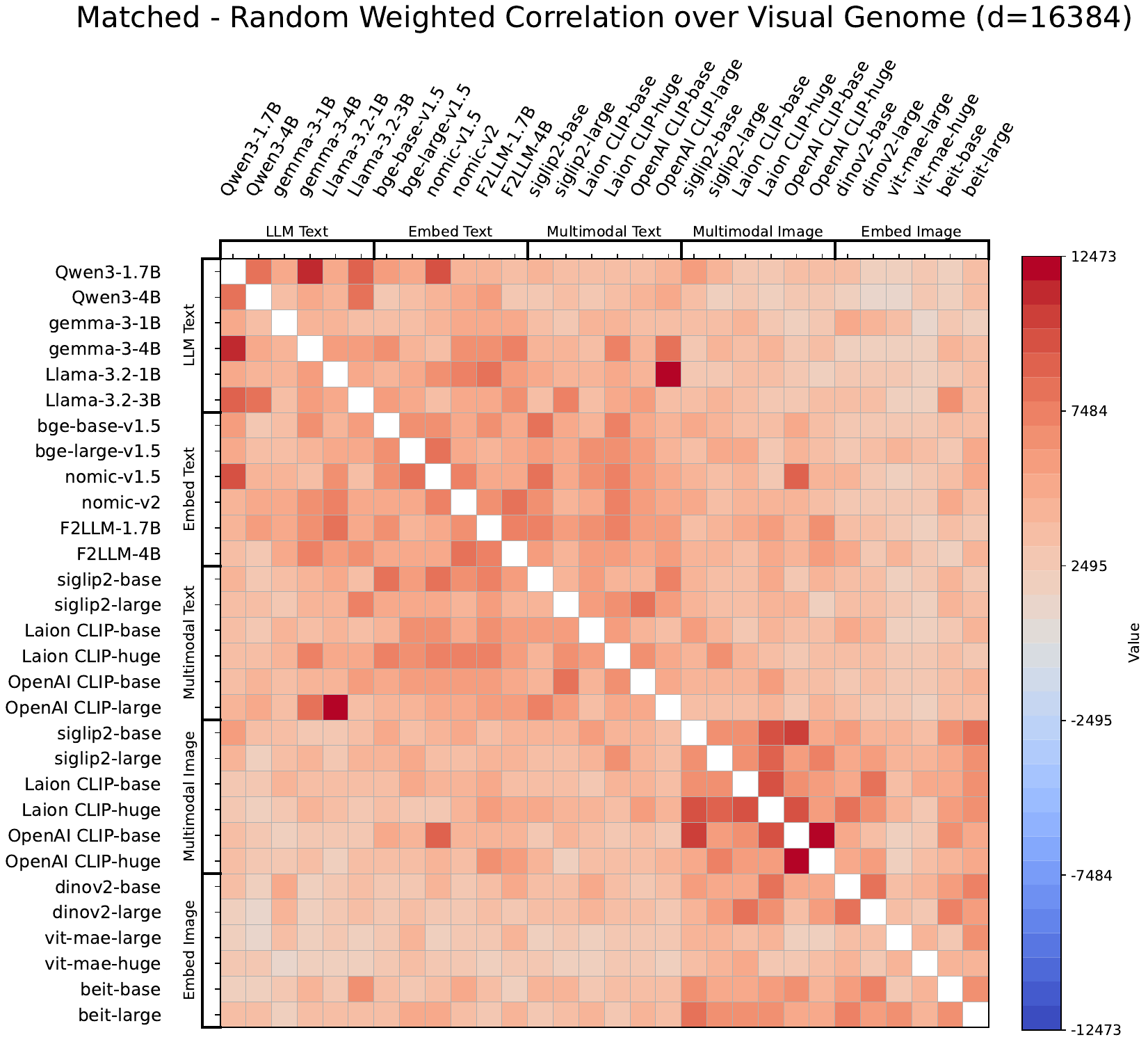}
    \end{minipage}
    \hfill
    \begin{minipage}[t]{0.3\textwidth}
        \centering
        \includegraphics[width = \linewidth]{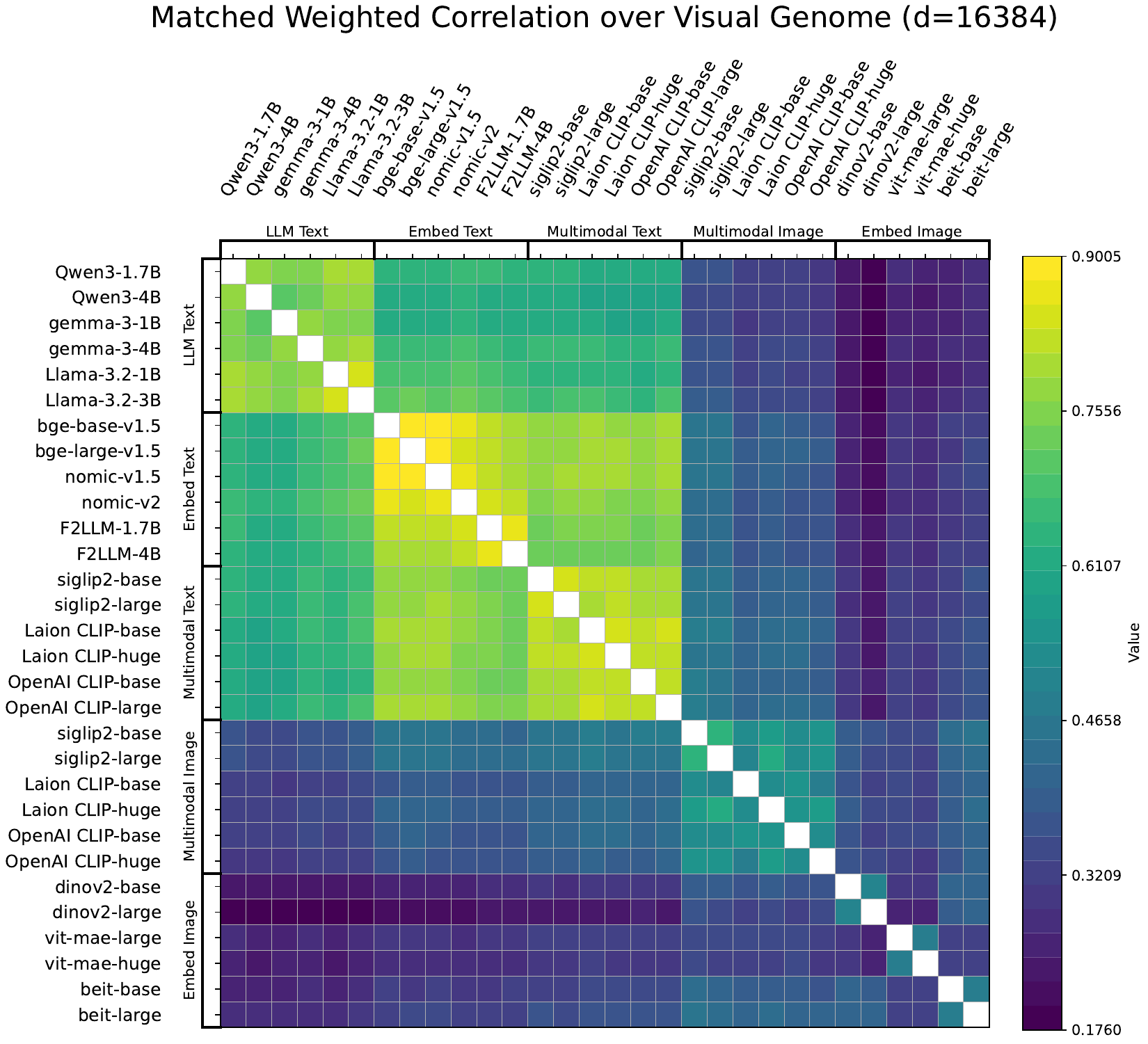}
    \end{minipage}
    \hfill
    \begin{minipage}[t]{0.3\textwidth}
        \centering
        \includegraphics[width = \linewidth]{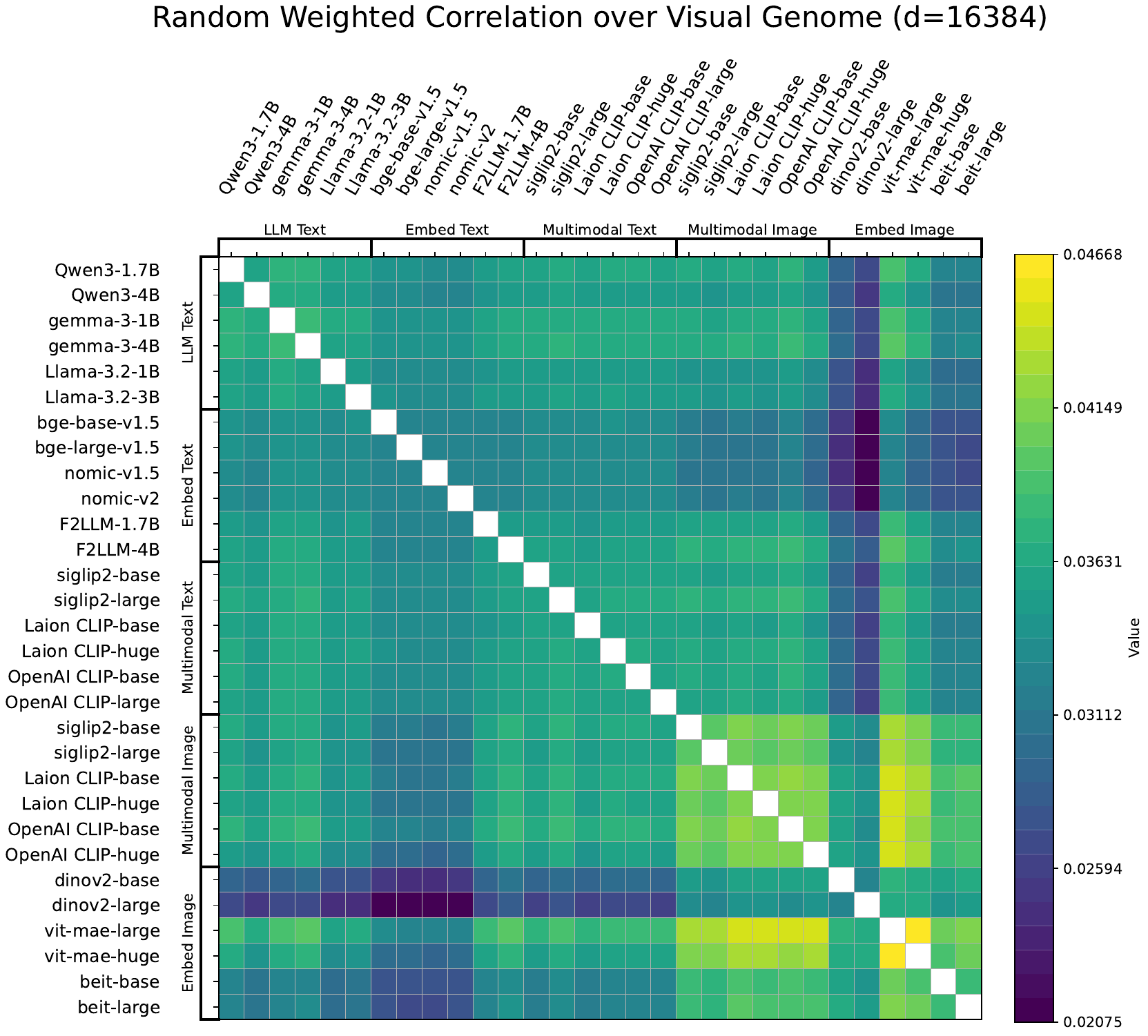}
    \end{minipage}
    
    \vspace{0.4cm}

        \begin{minipage}[t]{0.3\textwidth}
        \centering
        \includegraphics[width = \linewidth]{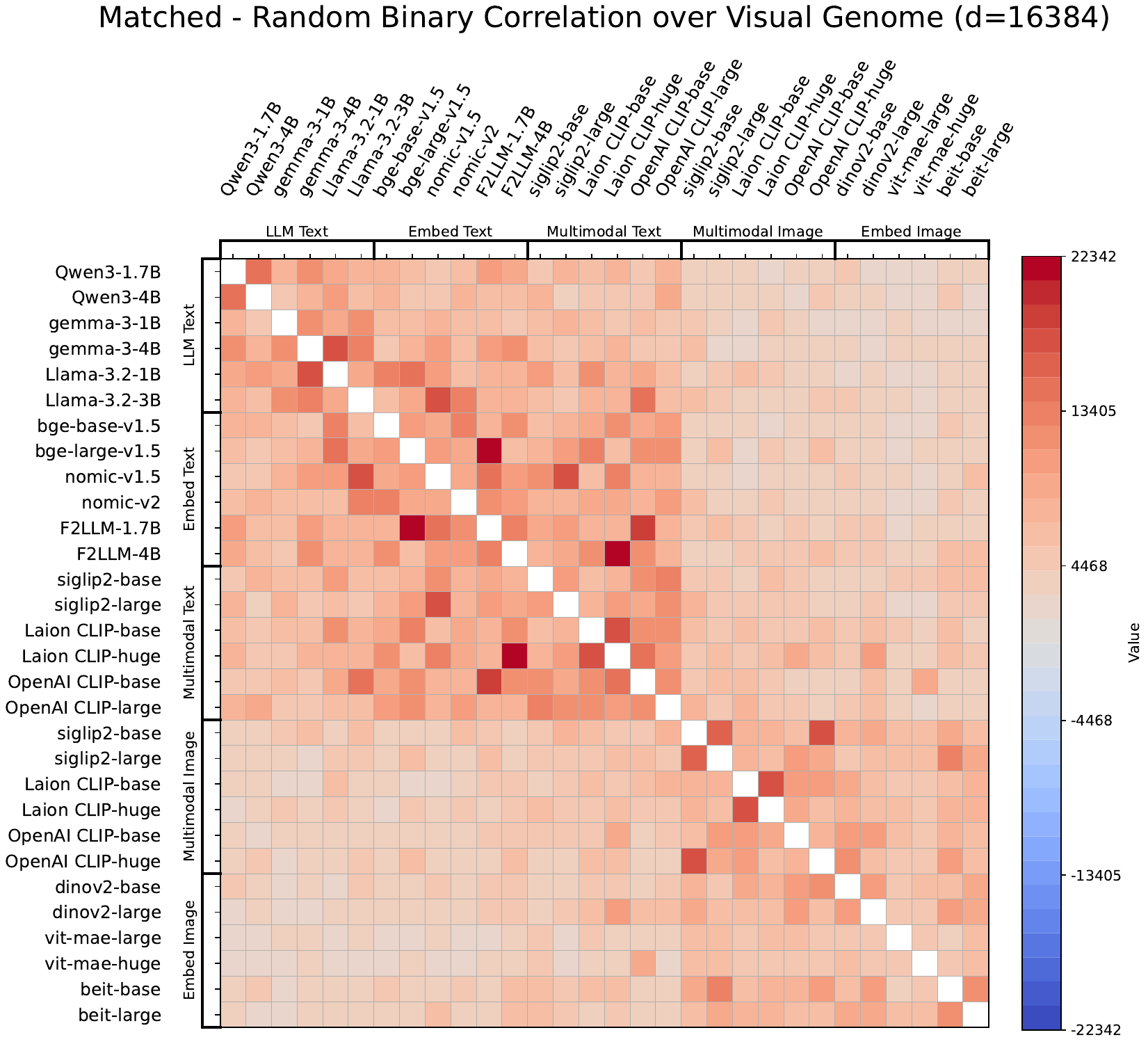}
    \end{minipage}
    \hfill
    \begin{minipage}[t]{0.3\textwidth}
        \centering
        \includegraphics[width = \linewidth]{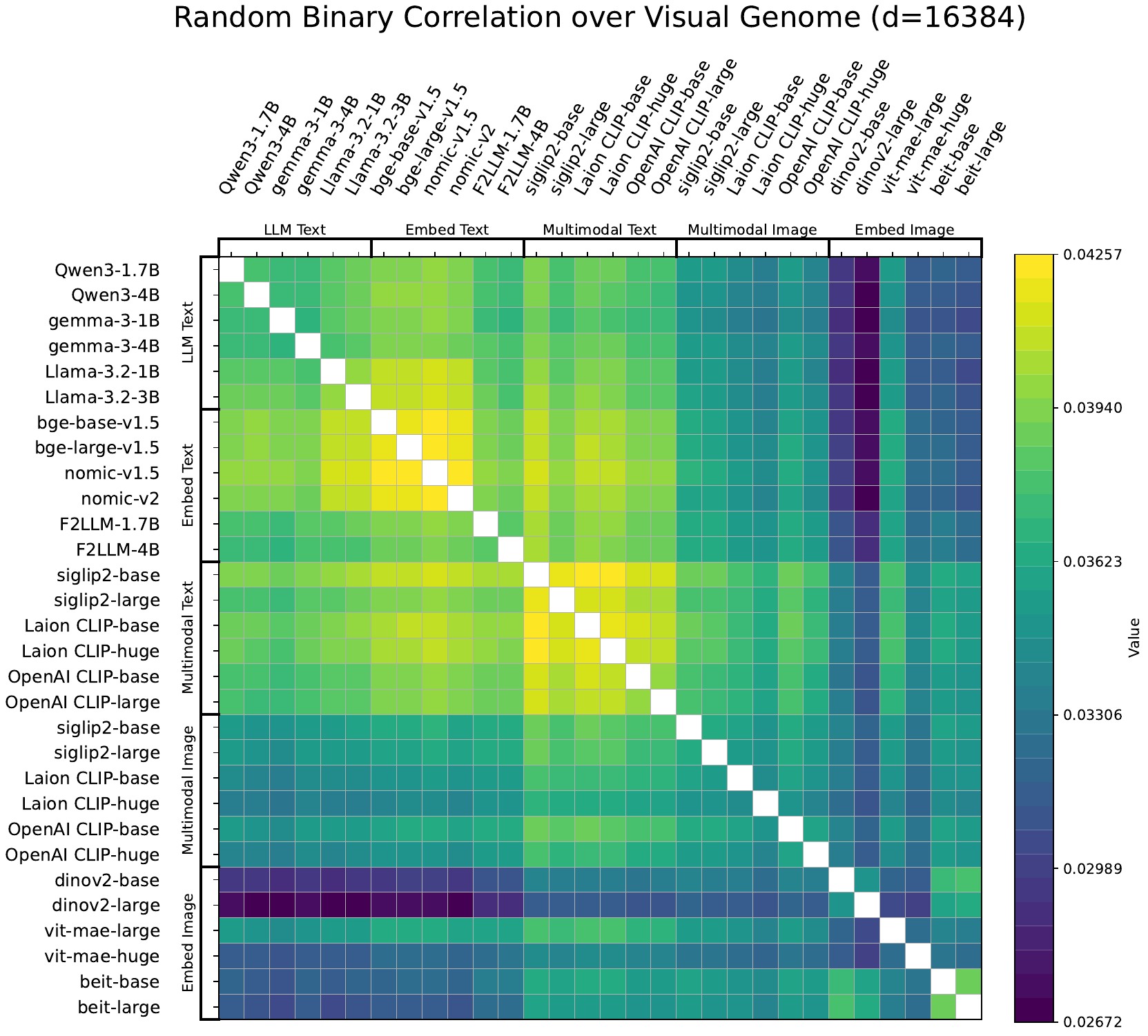}
    \end{minipage}
    \hfill
    \begin{minipage}[t]{0.3\textwidth}
        \centering
        \includegraphics[width = \linewidth]{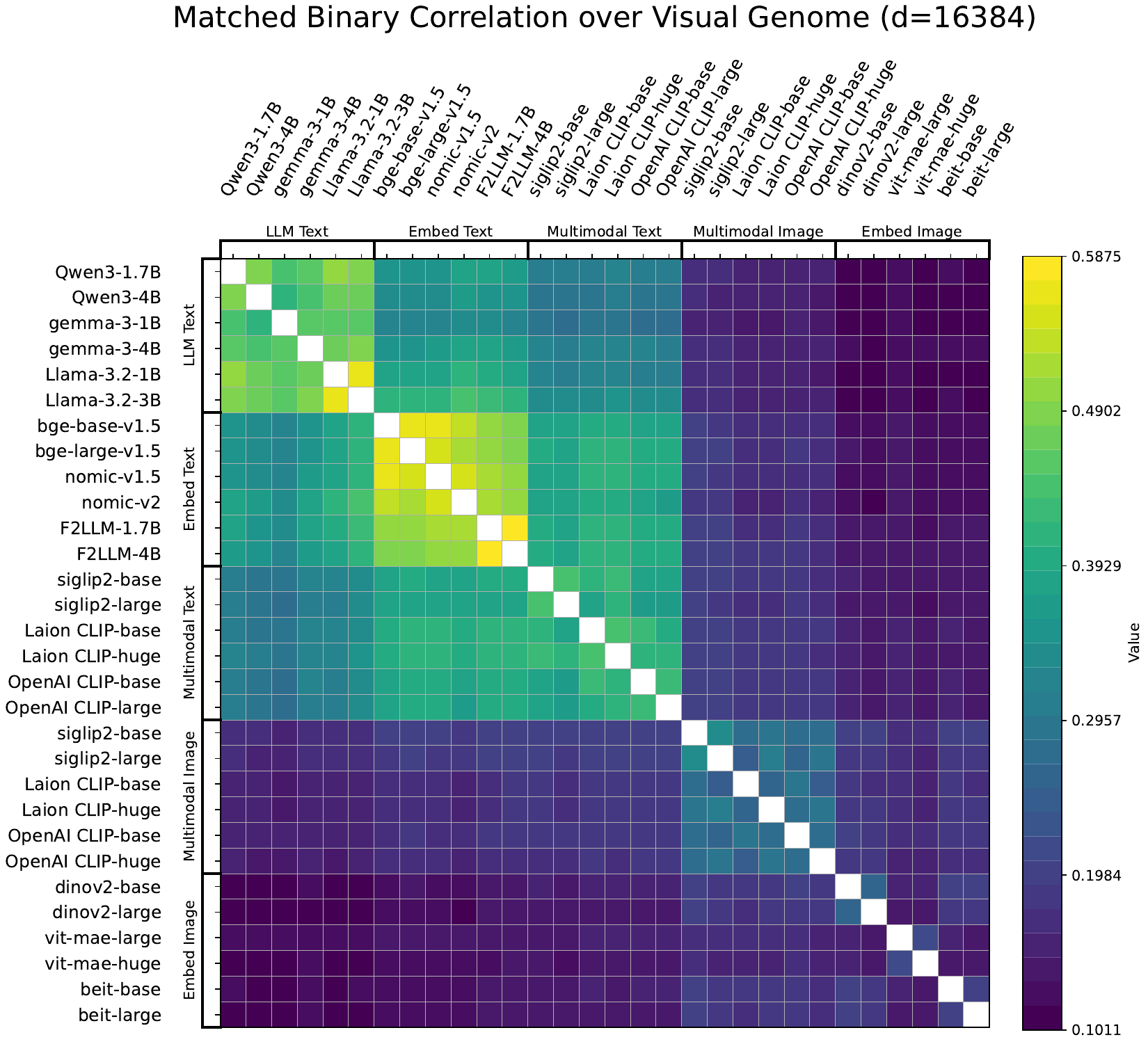}
    \end{minipage}

        \vspace{0.4cm}
    \caption{Same plot as in Figure~\ref{fig:signalcorrelationcocovar} but over Visual Genome.}
    \label{fig:signalcorrelationvisual_genomevar}
\end{figure}

\clearpage

\subsection{Bias: Centering Improves Alignment}
\label{appendix:biasexperiments}
\subsubsection{Experiments on COCO}

\begin{figure}[htbp]
    \centering

    \begin{minipage}[t]{0.3\textwidth}
        \centering
        \includegraphics[width = \linewidth]{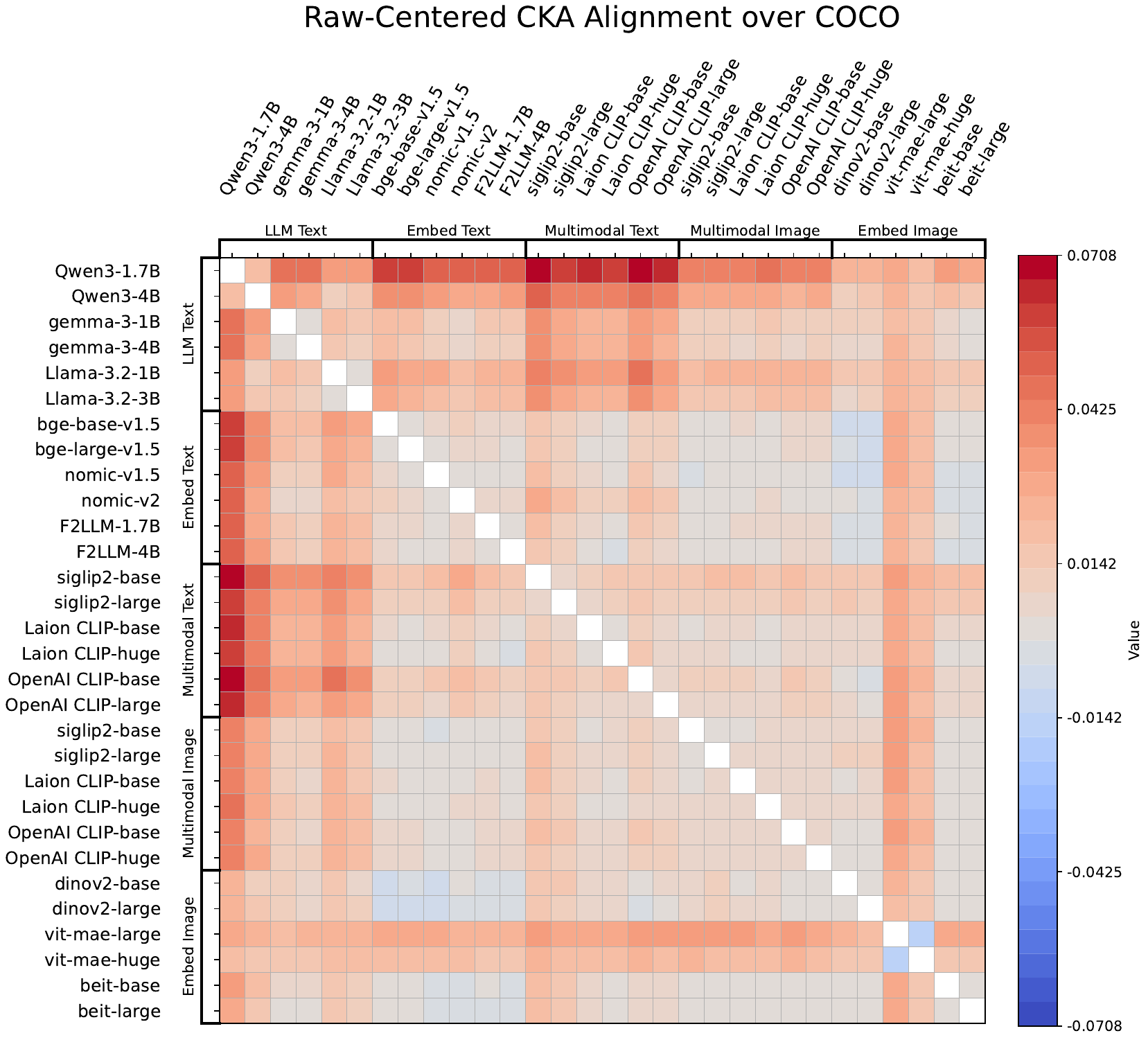}
    \end{minipage}
    \hfill
    \begin{minipage}[t]{0.3\textwidth}
        \centering
        \includegraphics[width = \linewidth]{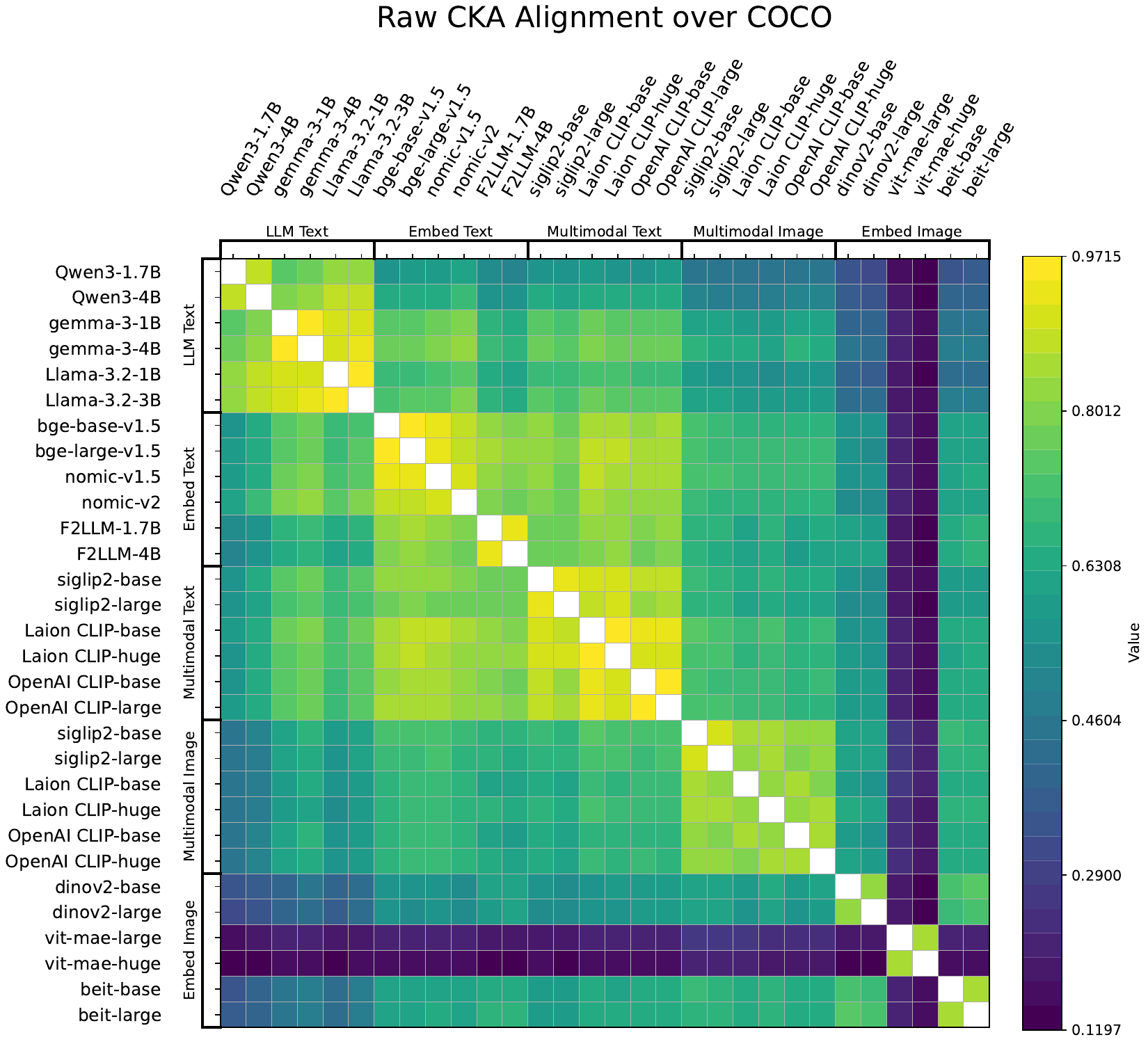}
    \end{minipage}
    \hfill
    \begin{minipage}[t]{0.3\textwidth}
        \centering
        \includegraphics[width = \linewidth]{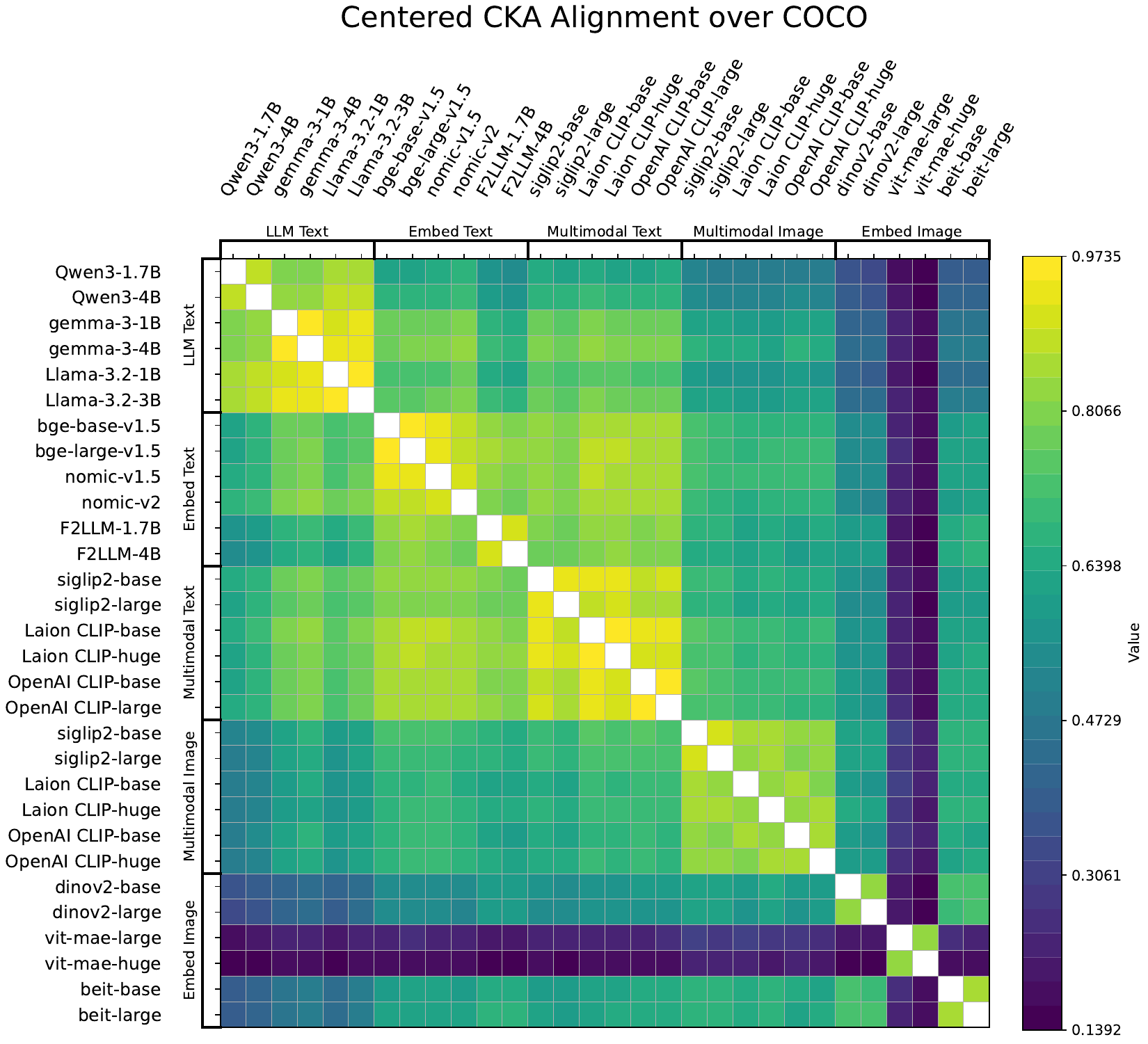}
    \end{minipage}
    
    \vspace{0.4cm}

        \begin{minipage}[t]{0.3\textwidth}
        \centering
        \includegraphics[width = \linewidth]{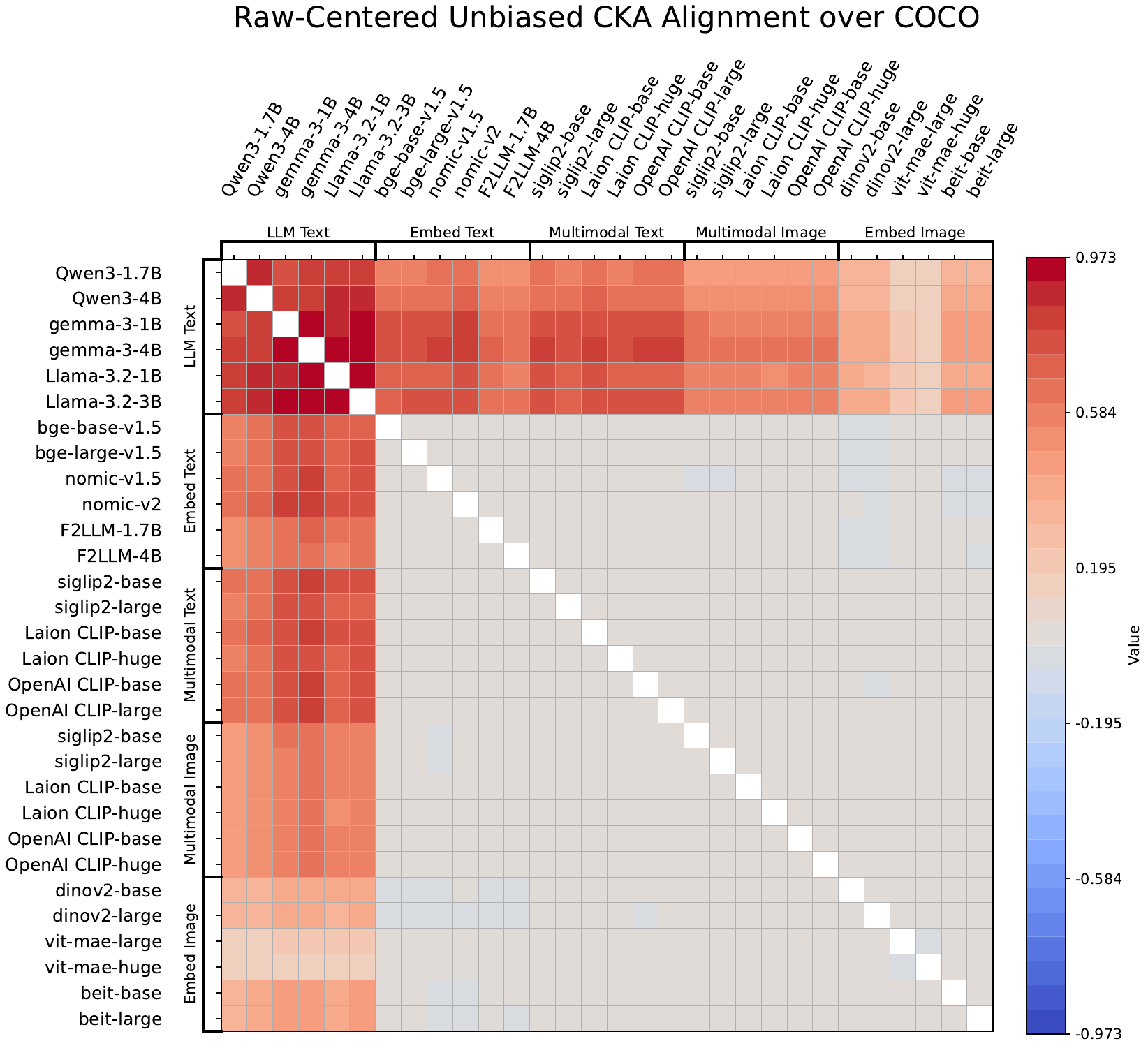}
    \end{minipage}
    \hfill
    \begin{minipage}[t]{0.3\textwidth}
        \centering
        \includegraphics[width = \linewidth]{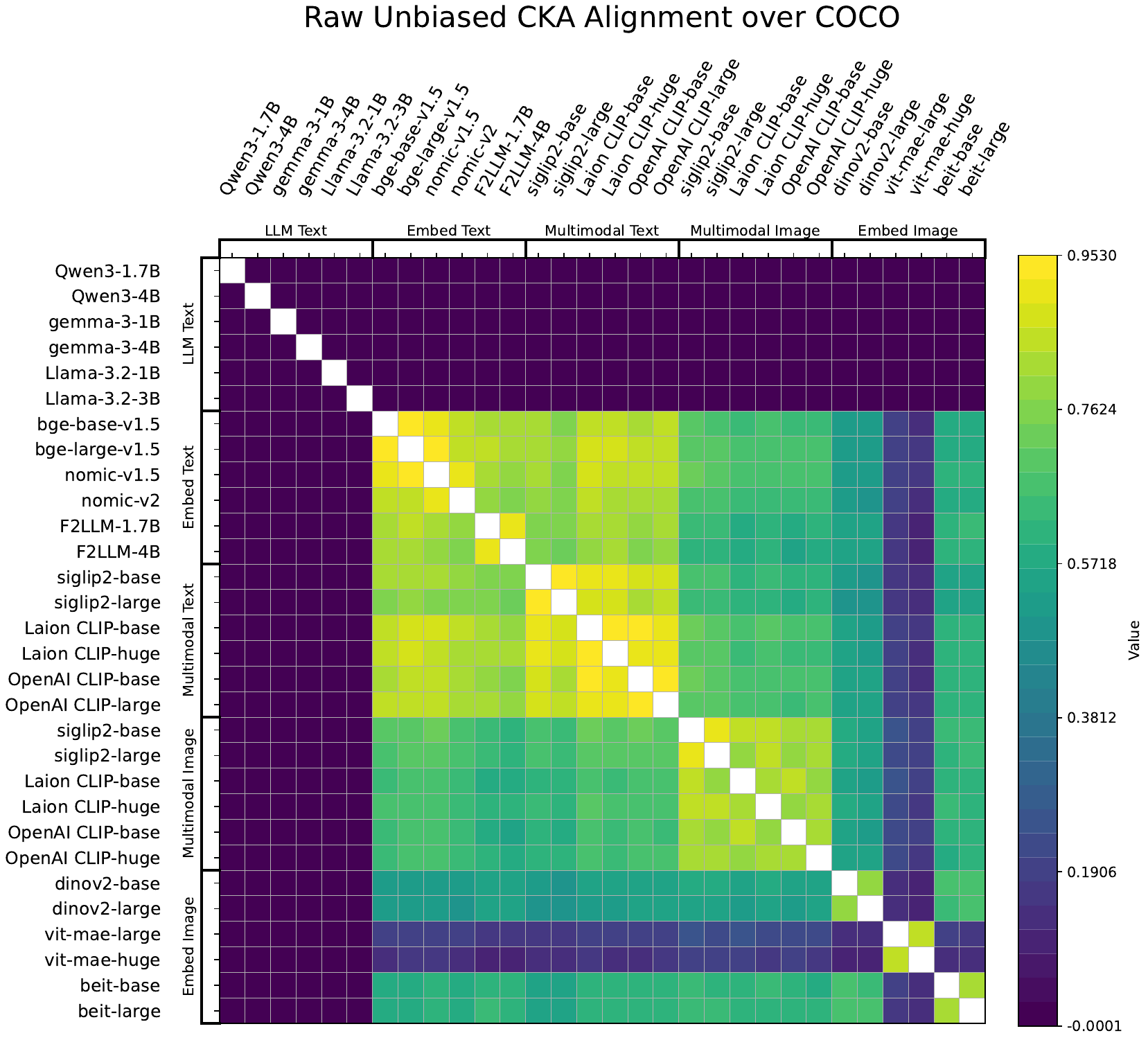}
    \end{minipage}
    \hfill
    \begin{minipage}[t]{0.3\textwidth}
        \centering
        \includegraphics[width = \linewidth]{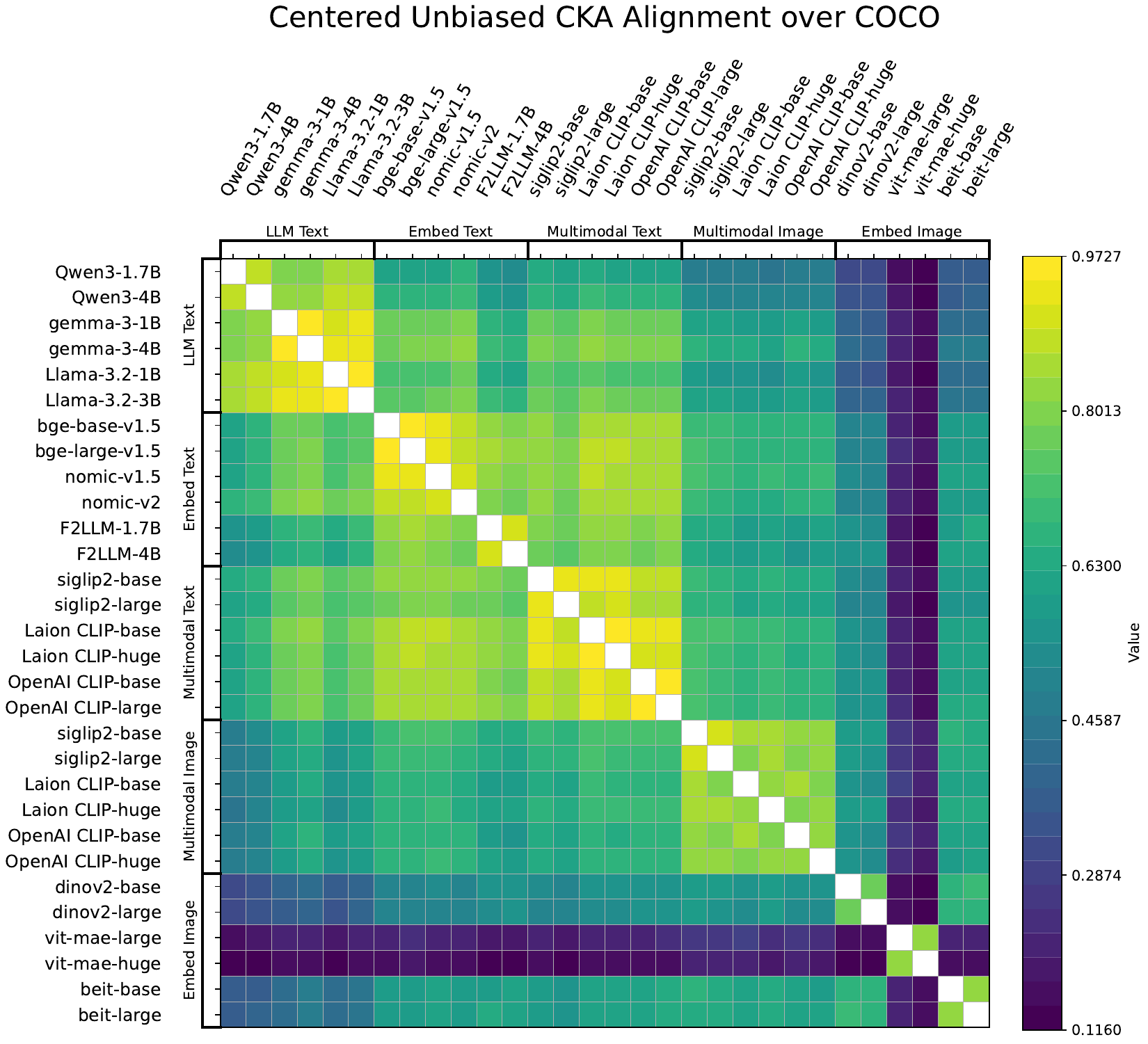}
    \end{minipage}

        \vspace{0.4cm}

        \begin{minipage}[t]{0.3\textwidth}
        \centering
        \includegraphics[width = \linewidth]{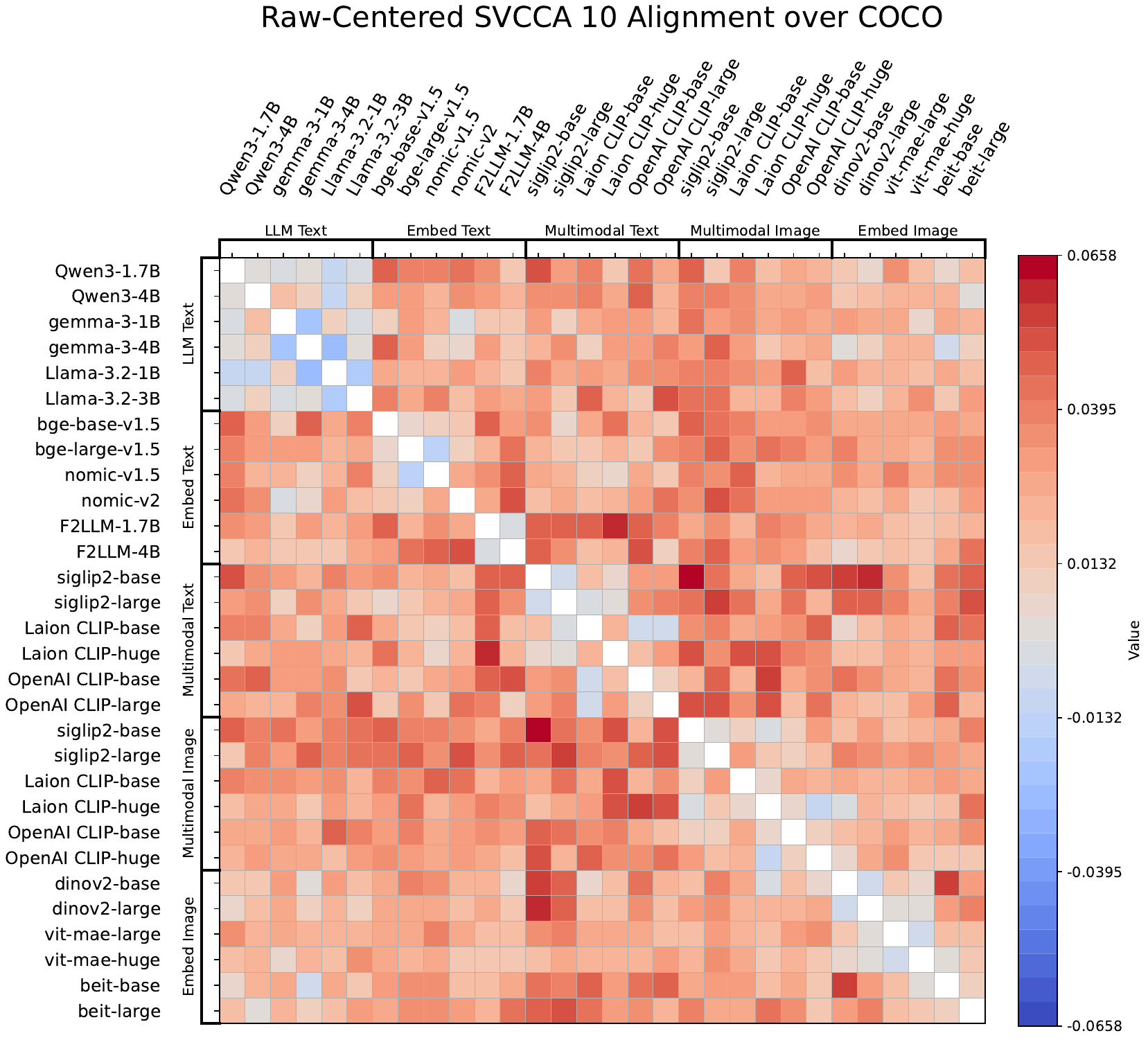}
    \end{minipage}
    \hfill
    \begin{minipage}[t]{0.3\textwidth}
        \centering
        \includegraphics[width = \linewidth]{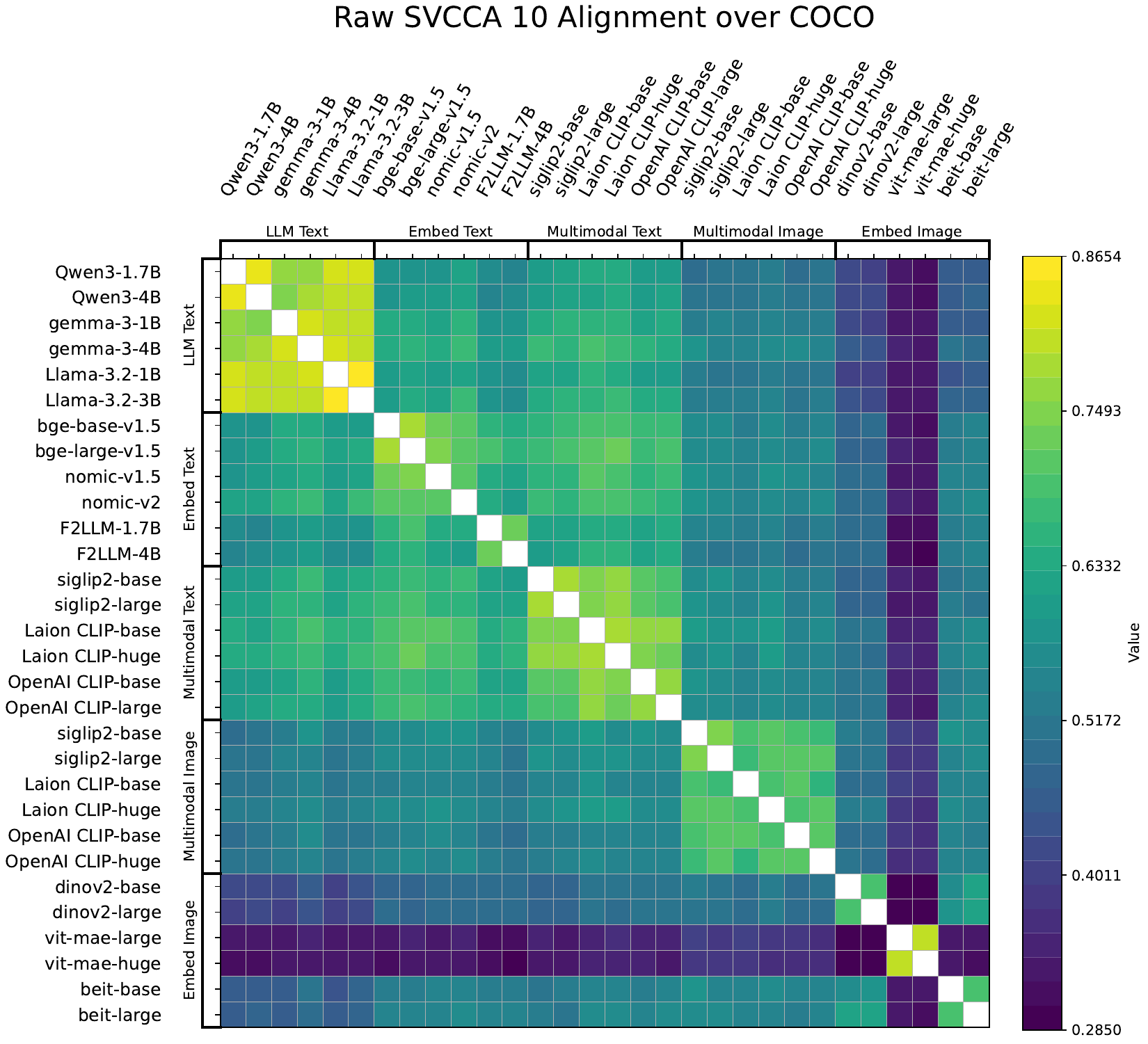}
    \end{minipage}
    \hfill
    \begin{minipage}[t]{0.3\textwidth}
        \centering
        \includegraphics[width = \linewidth]{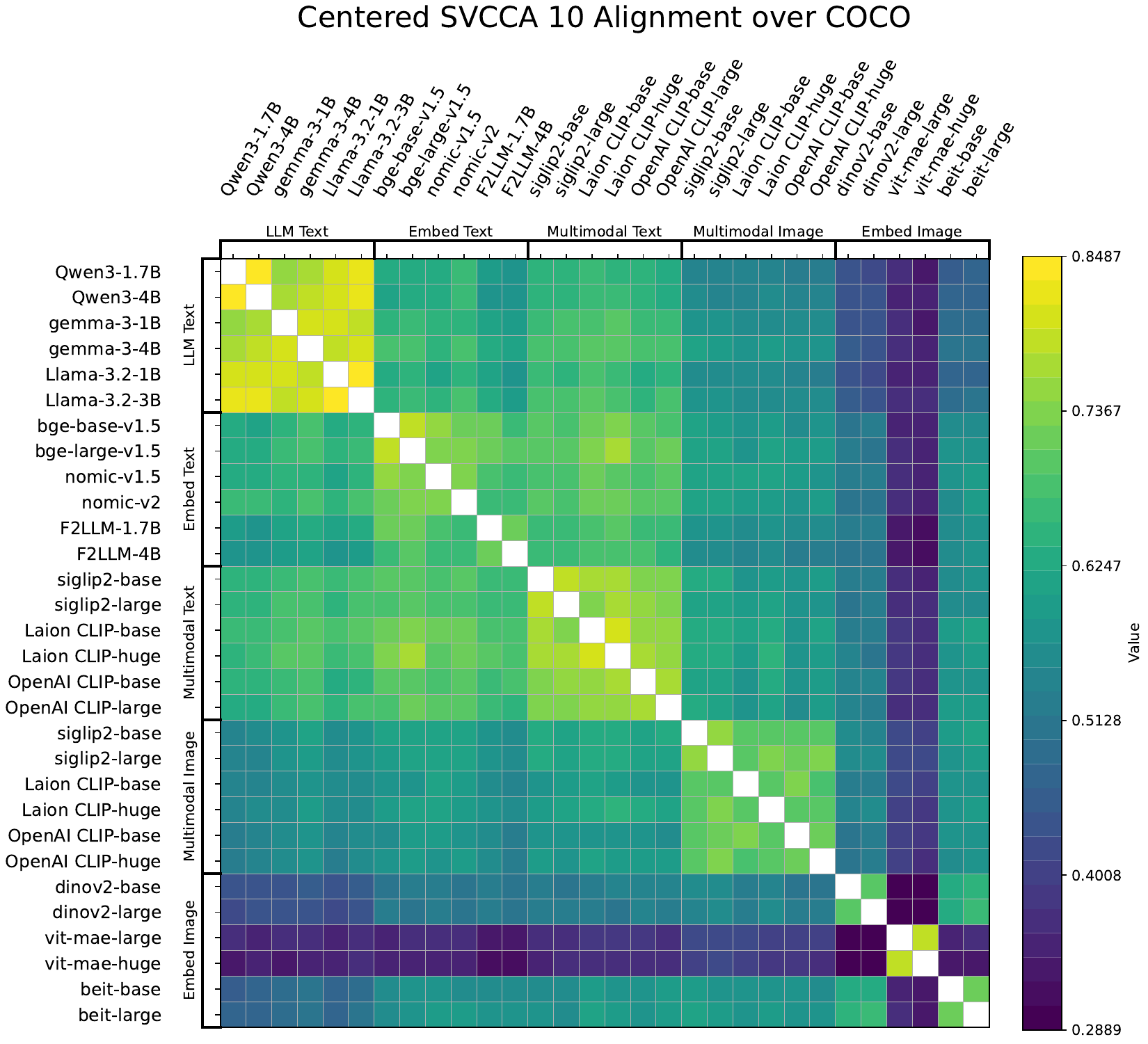}
    \end{minipage}

            \vspace{0.4cm}

    \begin{minipage}[t]{0.3\textwidth}
        \centering
        \includegraphics[width = \linewidth]{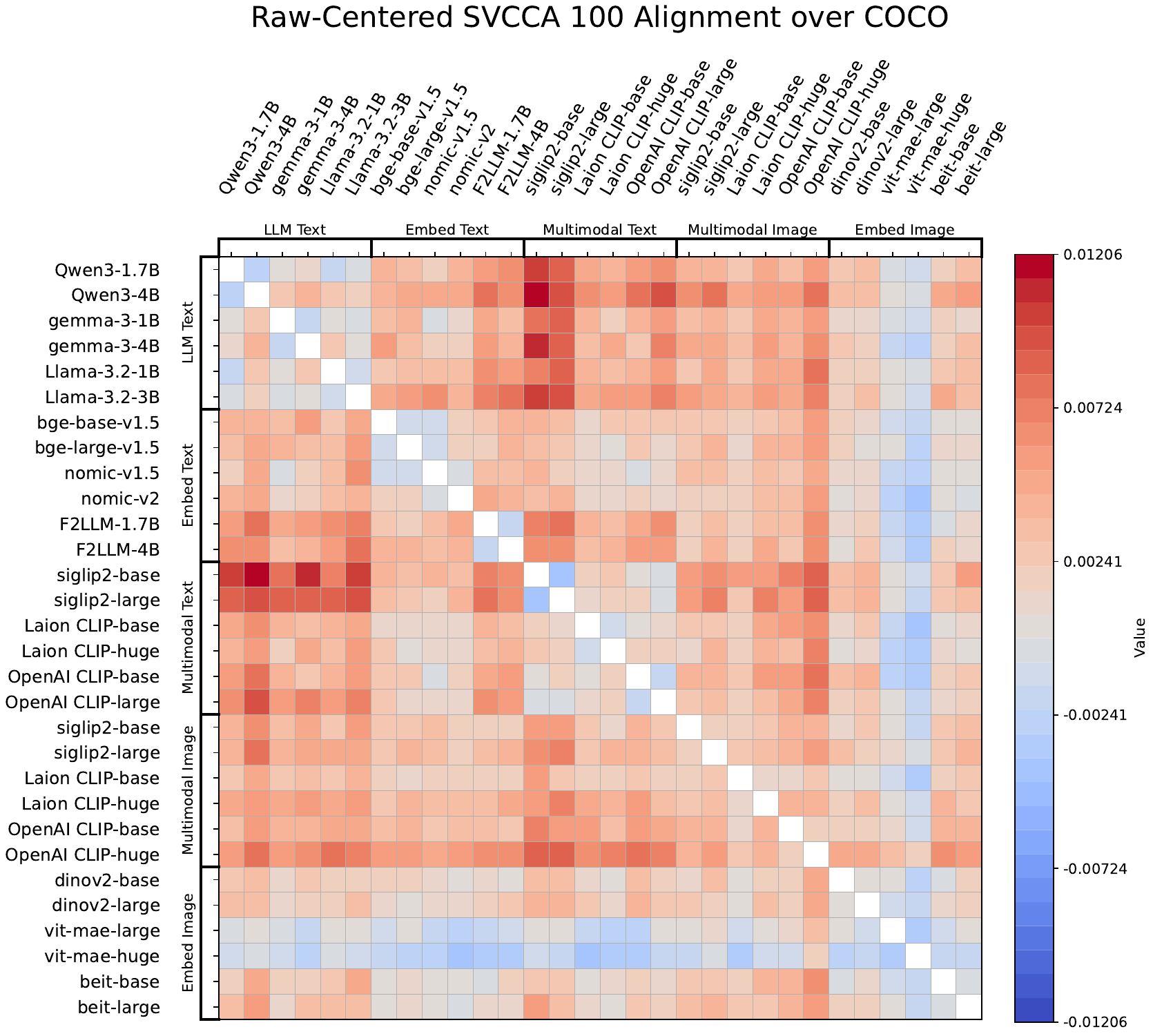}
    \end{minipage}
    \hfill
    \begin{minipage}[t]{0.3\textwidth}
        \centering
        \includegraphics[width = \linewidth]{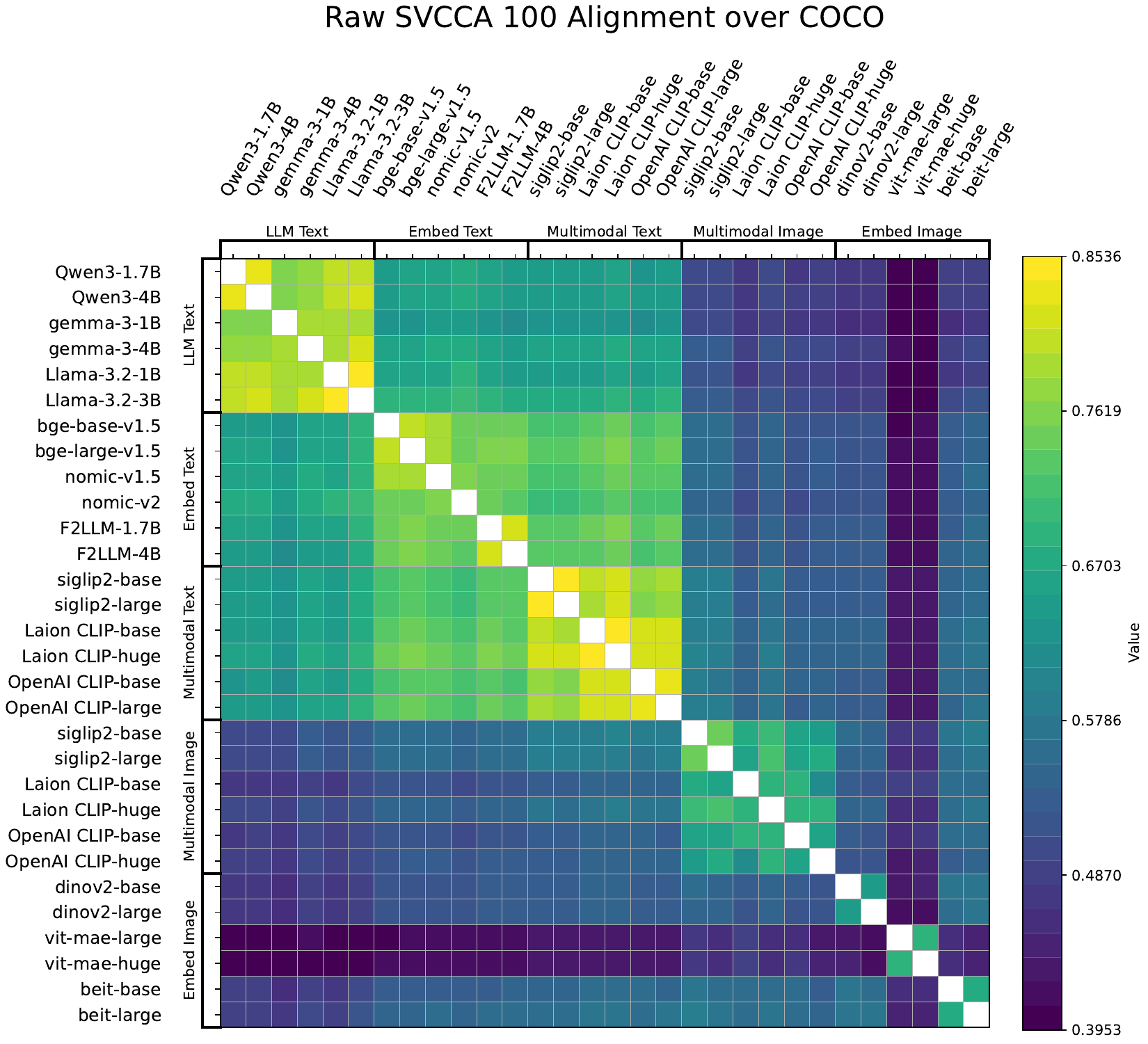}
    \end{minipage}
    \hfill
    \begin{minipage}[t]{0.3\textwidth}
        \centering
        \includegraphics[width = \linewidth]{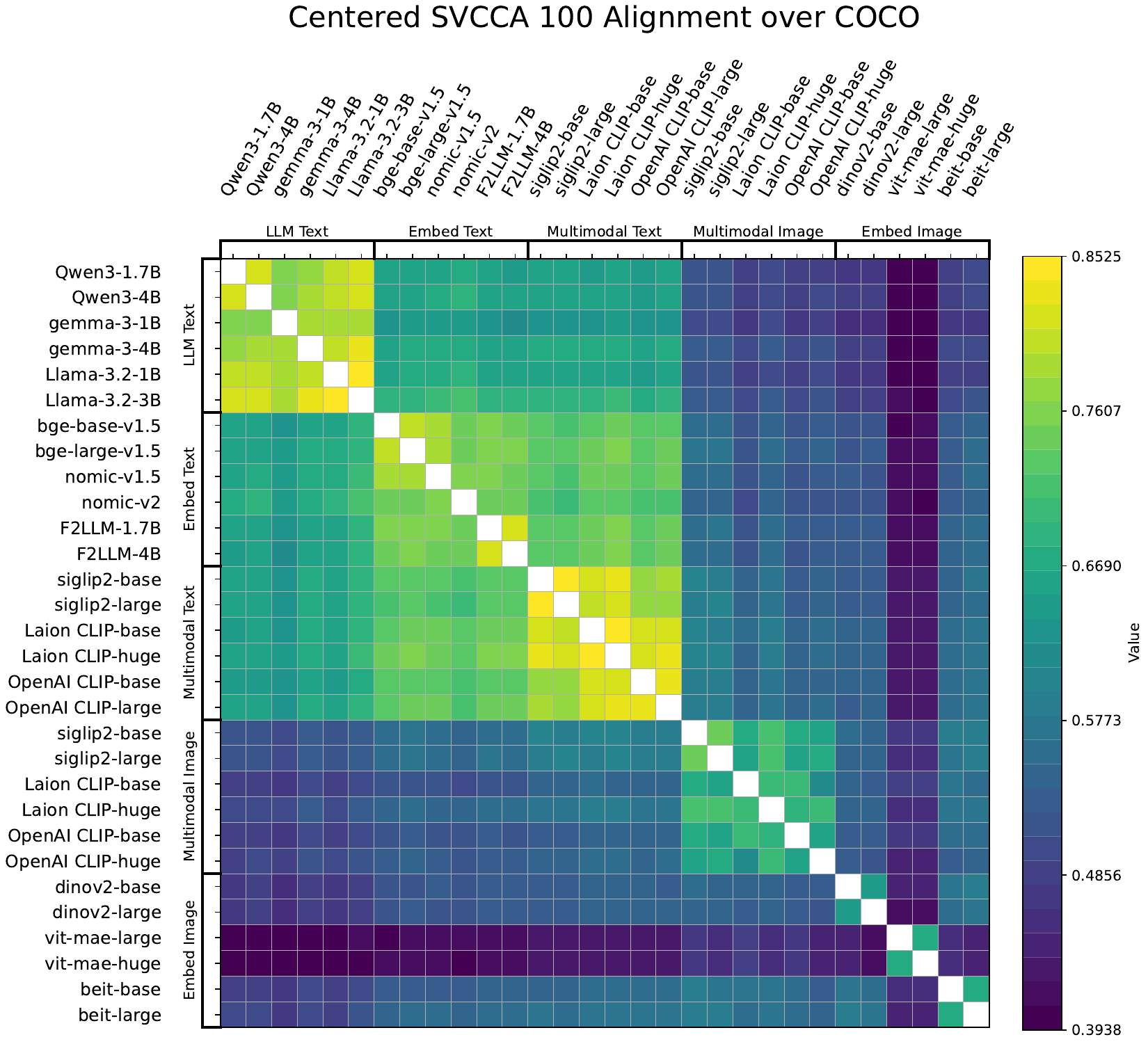}
    \end{minipage}
    \caption{Same plot as in Figure~\ref{fig:biasmain} but for the CKA, Unbiased CKA, SVCCA 10 and SVCCA 100 metrics. In addition to differences, we plot the alignment values for raw dense features and for the sparse features.}
    \label{fig:biascoco1}
\end{figure}

\clearpage

\begin{figure}[htbp]
    \centering

    \begin{minipage}[t]{0.3\textwidth}
        \centering
        \includegraphics[width = \linewidth]{bias_diagrams/topk_10_coco_diff.pdf}
    \end{minipage}
    \hfill
    \begin{minipage}[t]{0.3\textwidth}
        \centering
        \includegraphics[width = \linewidth]{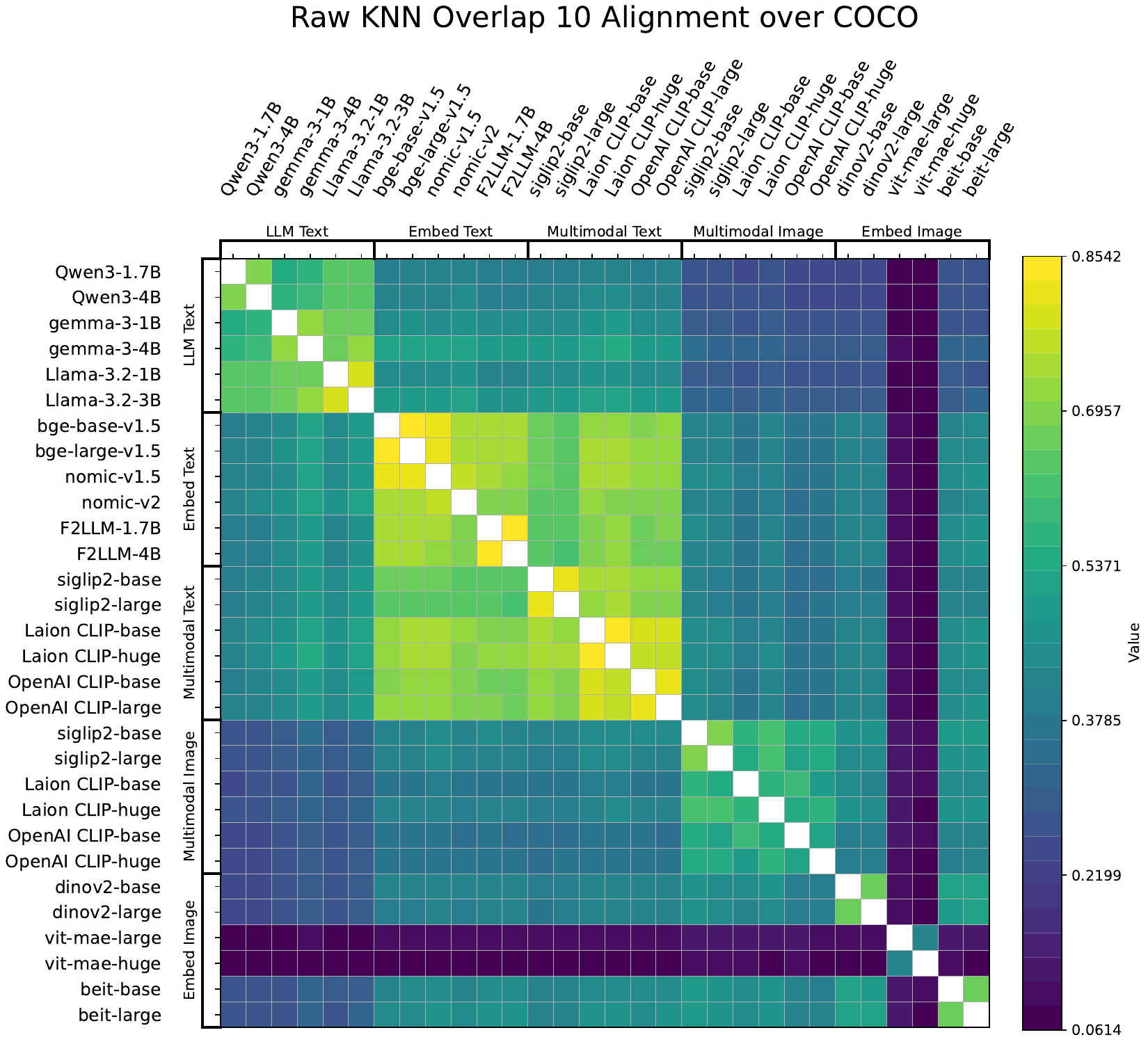}
    \end{minipage}
    \hfill
    \begin{minipage}[t]{0.3\textwidth}
        \centering
        \includegraphics[width = \linewidth]{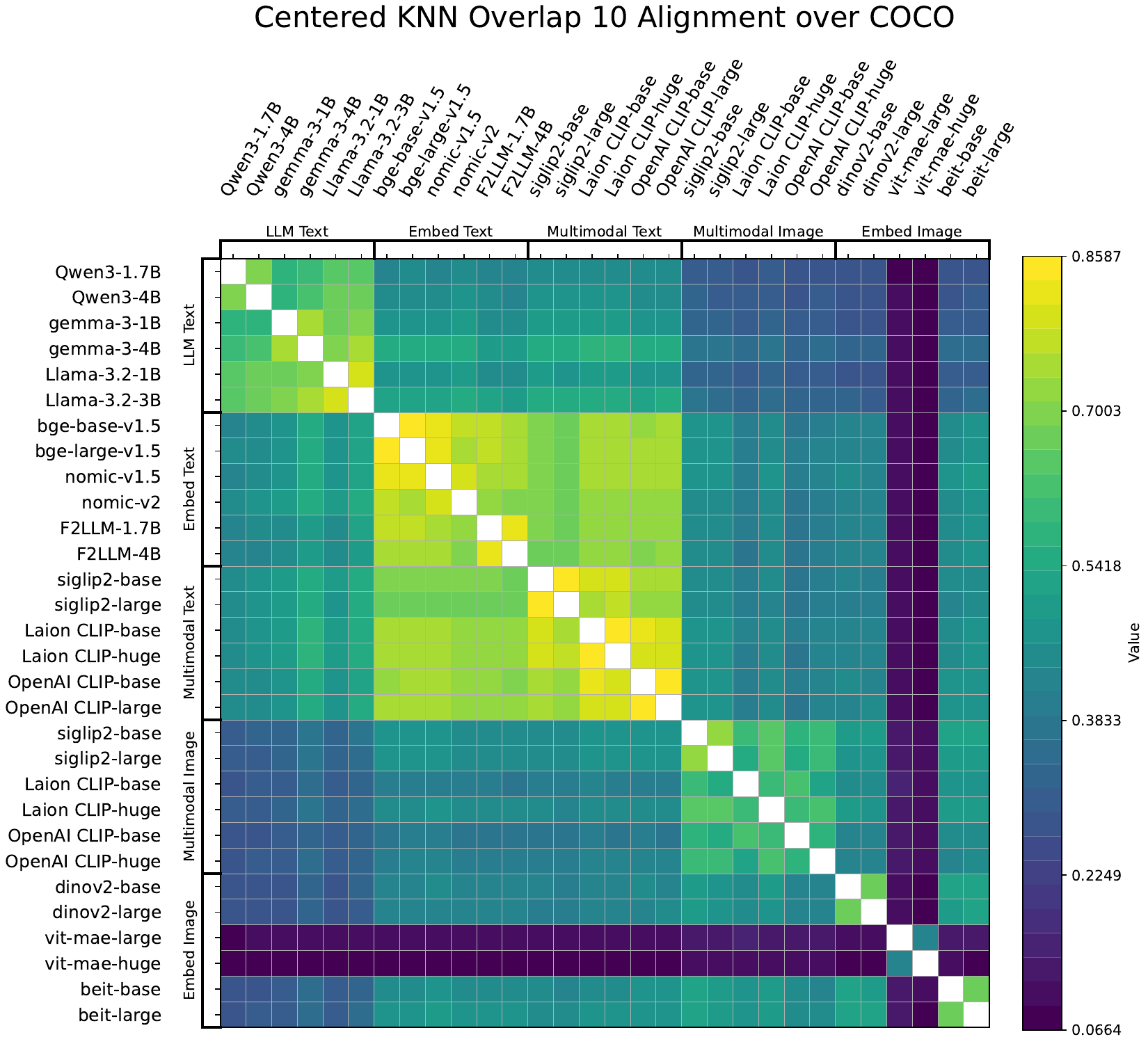}
    \end{minipage}
    
    \vspace{0.4cm}

        \begin{minipage}[t]{0.3\textwidth}
        \centering
        \includegraphics[width = \linewidth]{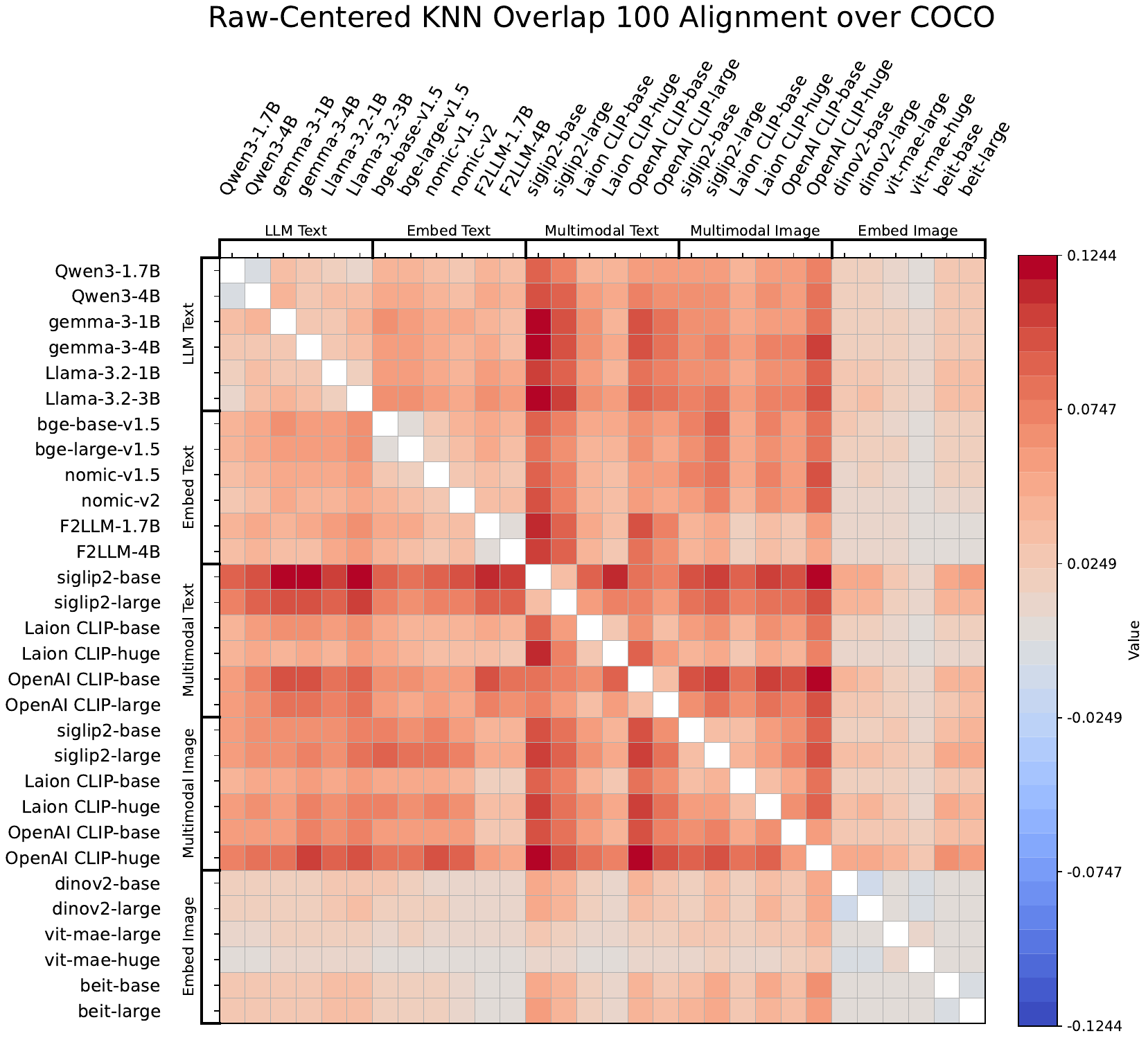}
    \end{minipage}
    \hfill
    \begin{minipage}[t]{0.3\textwidth}
        \centering
        \includegraphics[width = \linewidth]{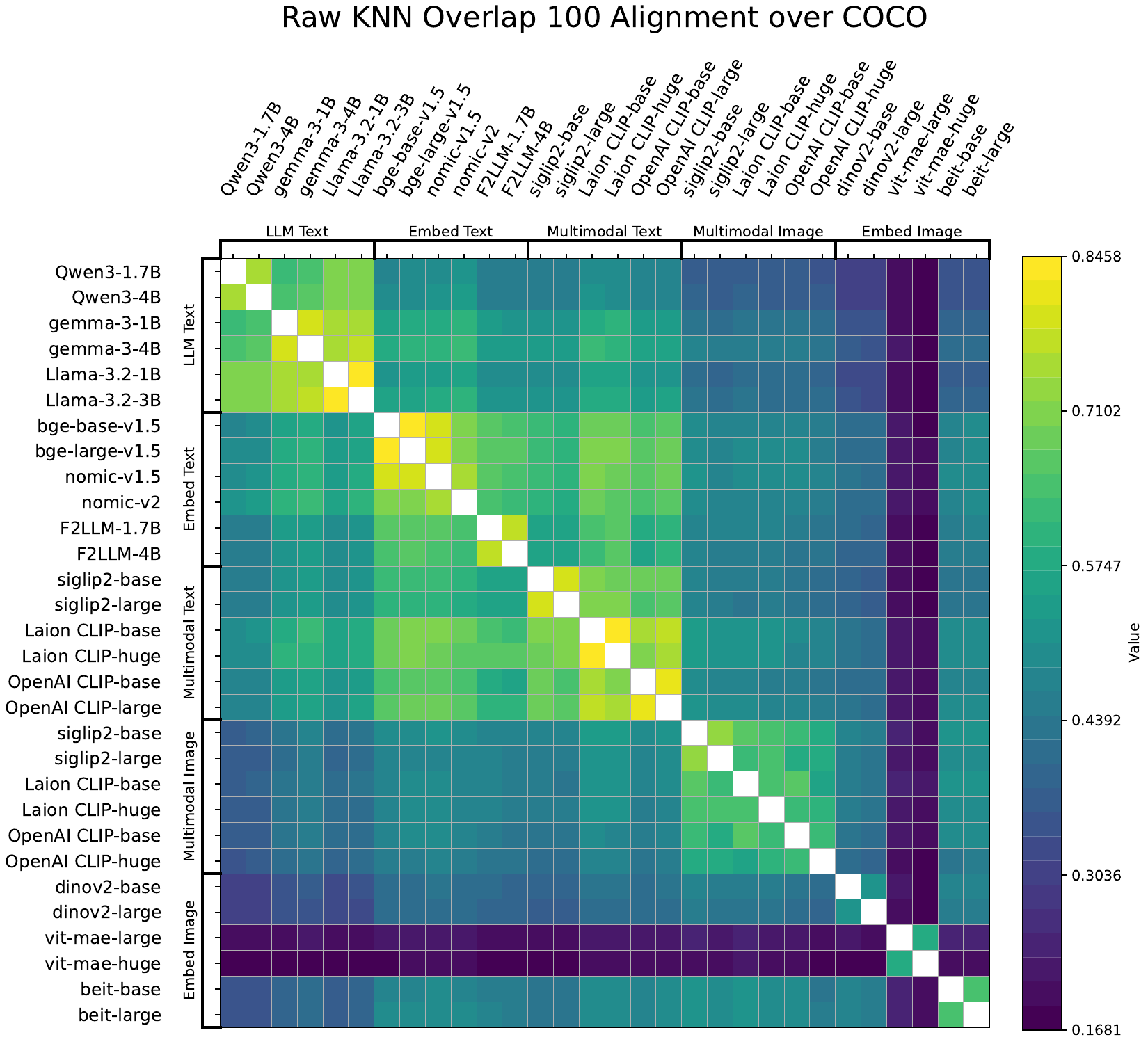}
    \end{minipage}
    \hfill
    \begin{minipage}[t]{0.3\textwidth}
        \centering
        \includegraphics[width = \linewidth]{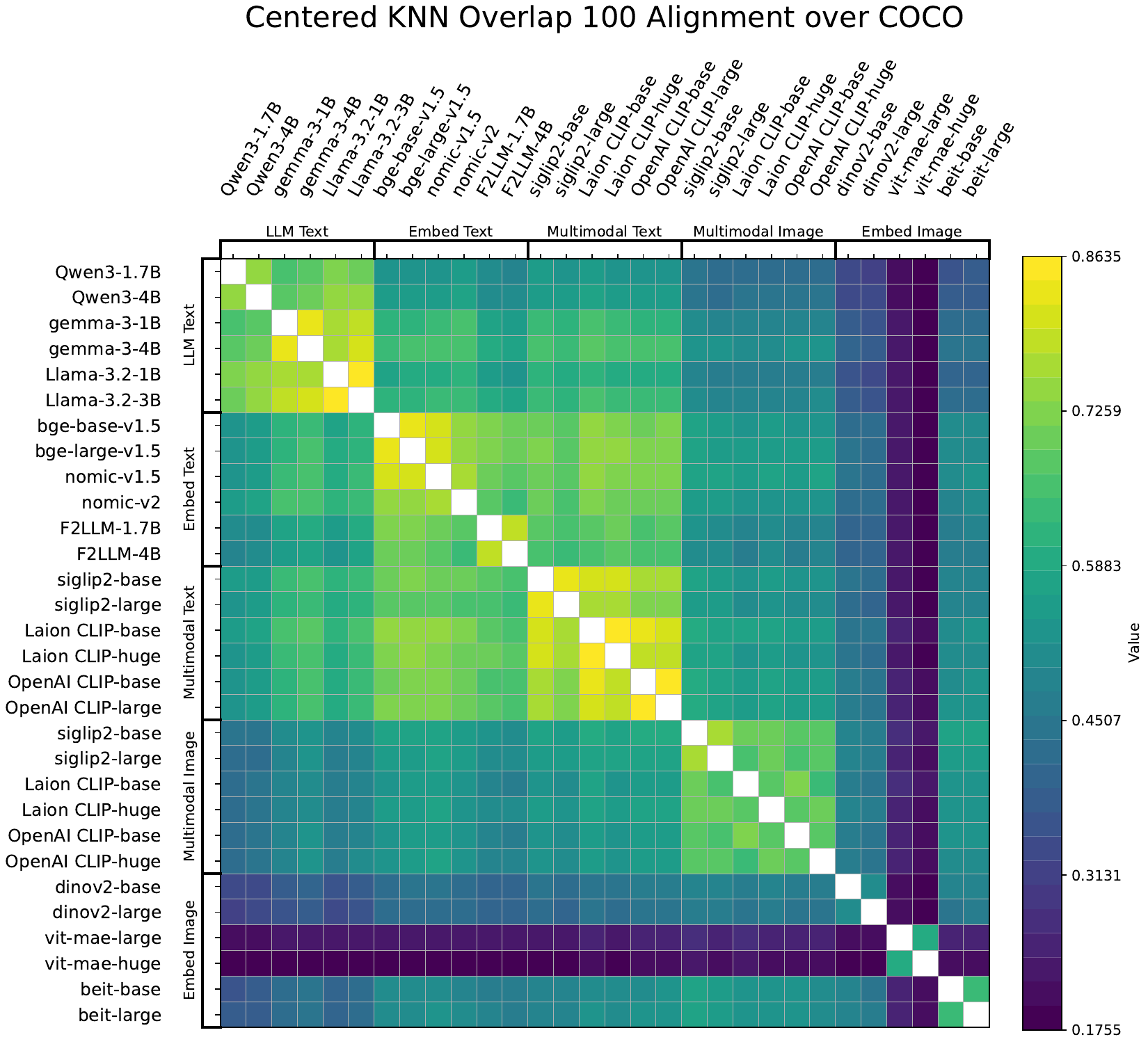}
    \end{minipage}

        \vspace{0.4cm}

        \begin{minipage}[t]{0.3\textwidth}
        \centering
        \includegraphics[width = \linewidth]{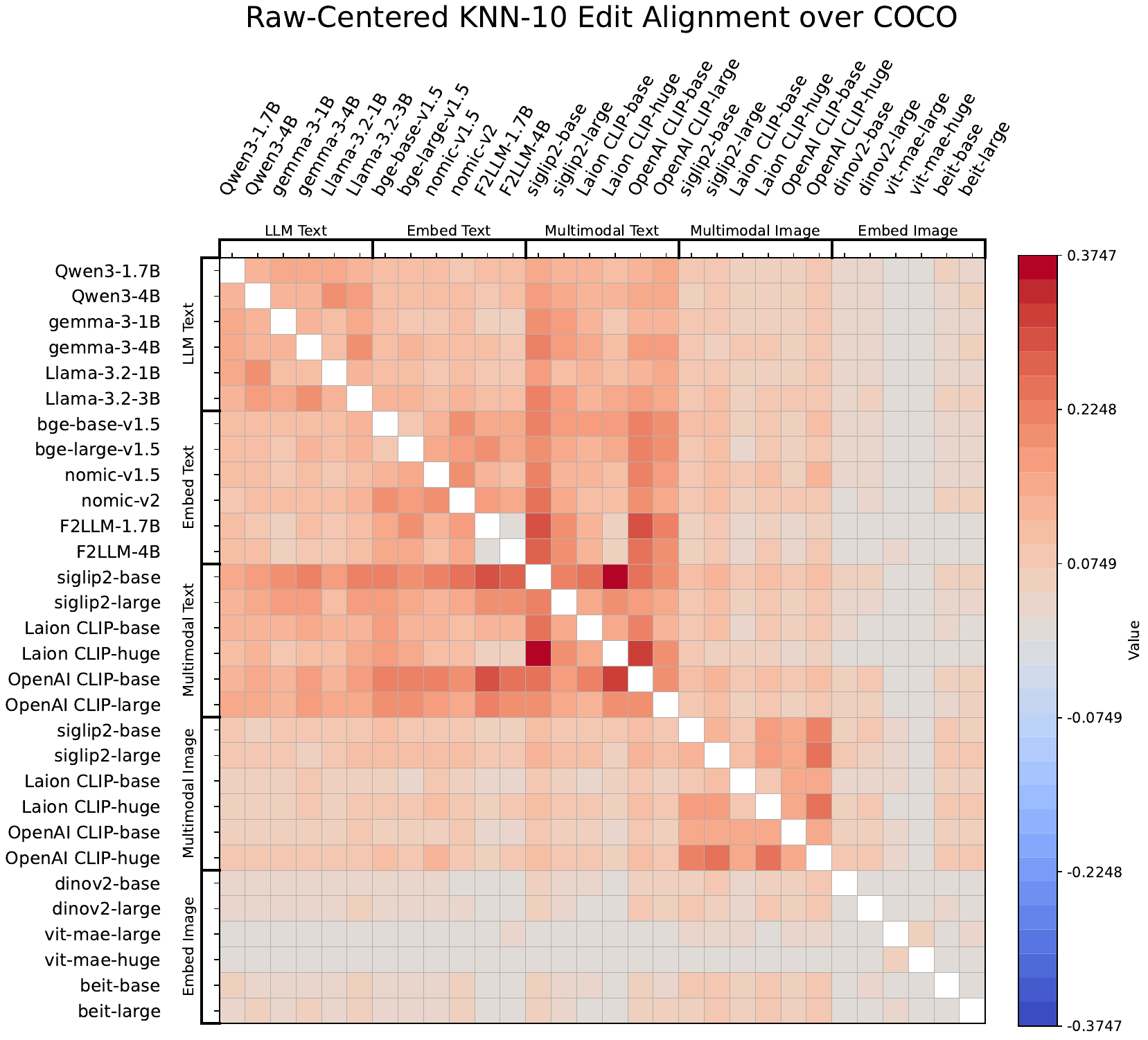}
    \end{minipage}
    \hfill
    \begin{minipage}[t]{0.3\textwidth}
        \centering
        \includegraphics[width = \linewidth]{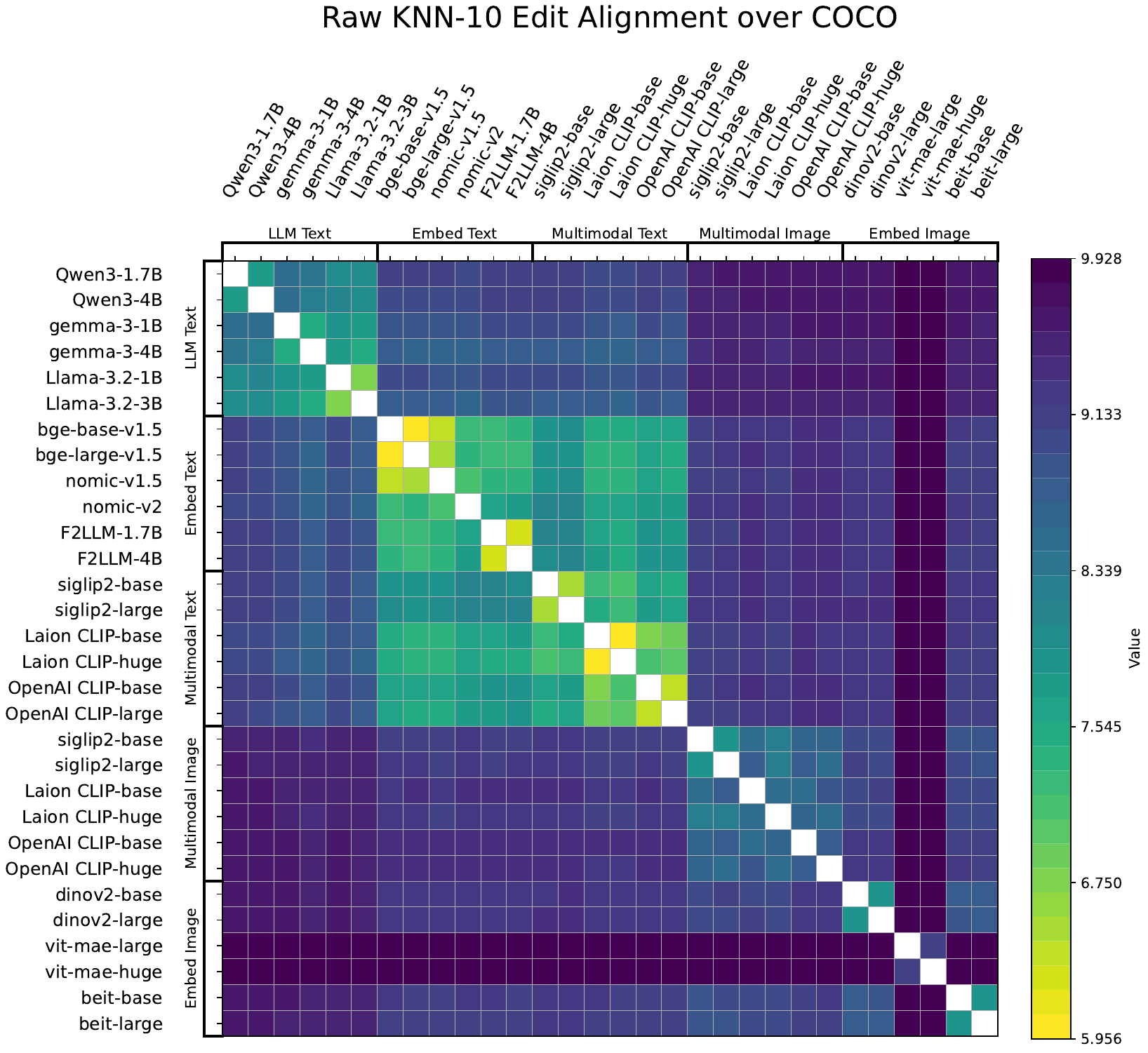}
    \end{minipage}
    \hfill
    \begin{minipage}[t]{0.3\textwidth}
        \centering
        \includegraphics[width = \linewidth]{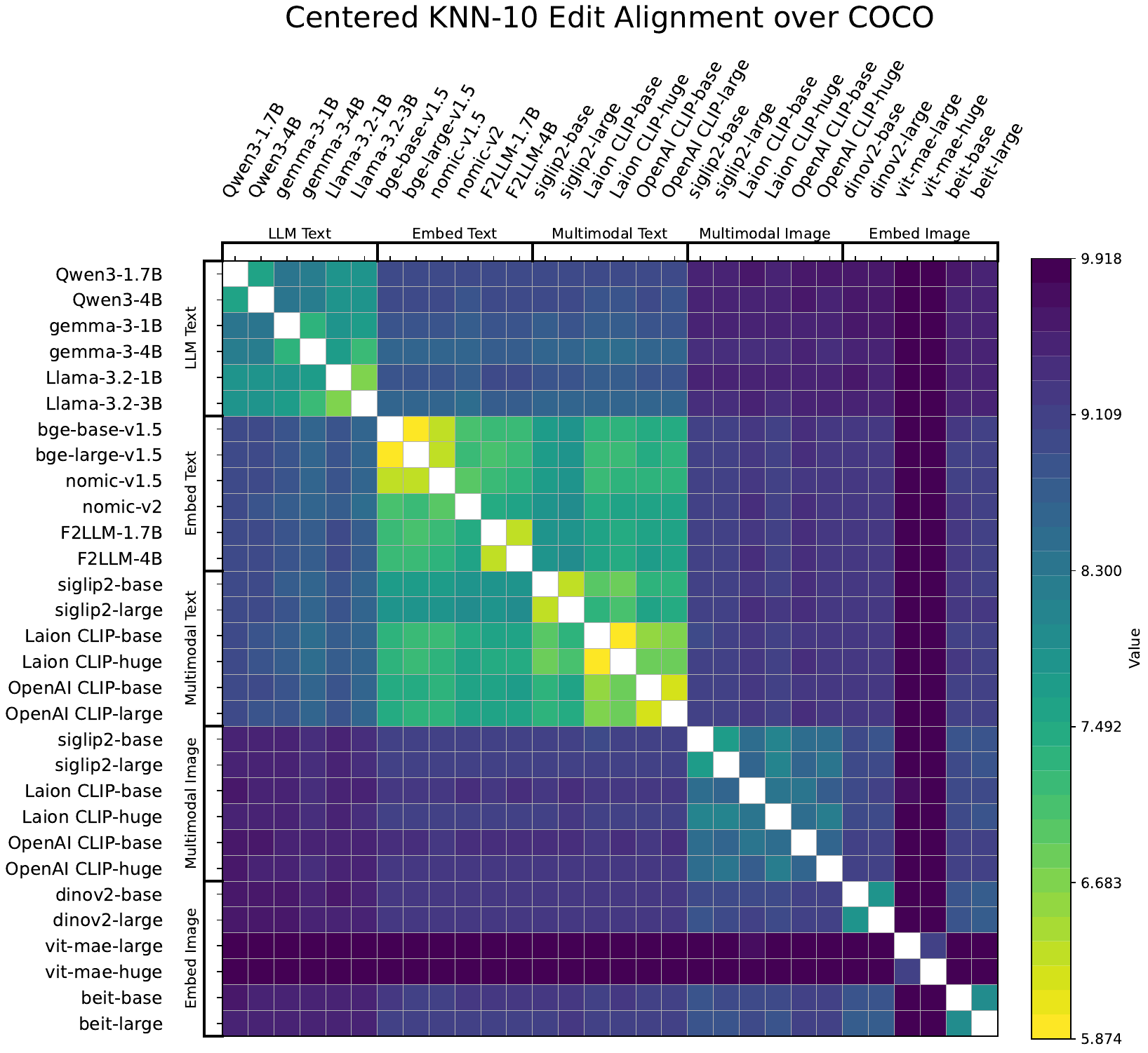}
    \end{minipage}

            \vspace{0.4cm}

    \begin{minipage}[t]{0.3\textwidth}
        \centering
        \includegraphics[width = \linewidth]{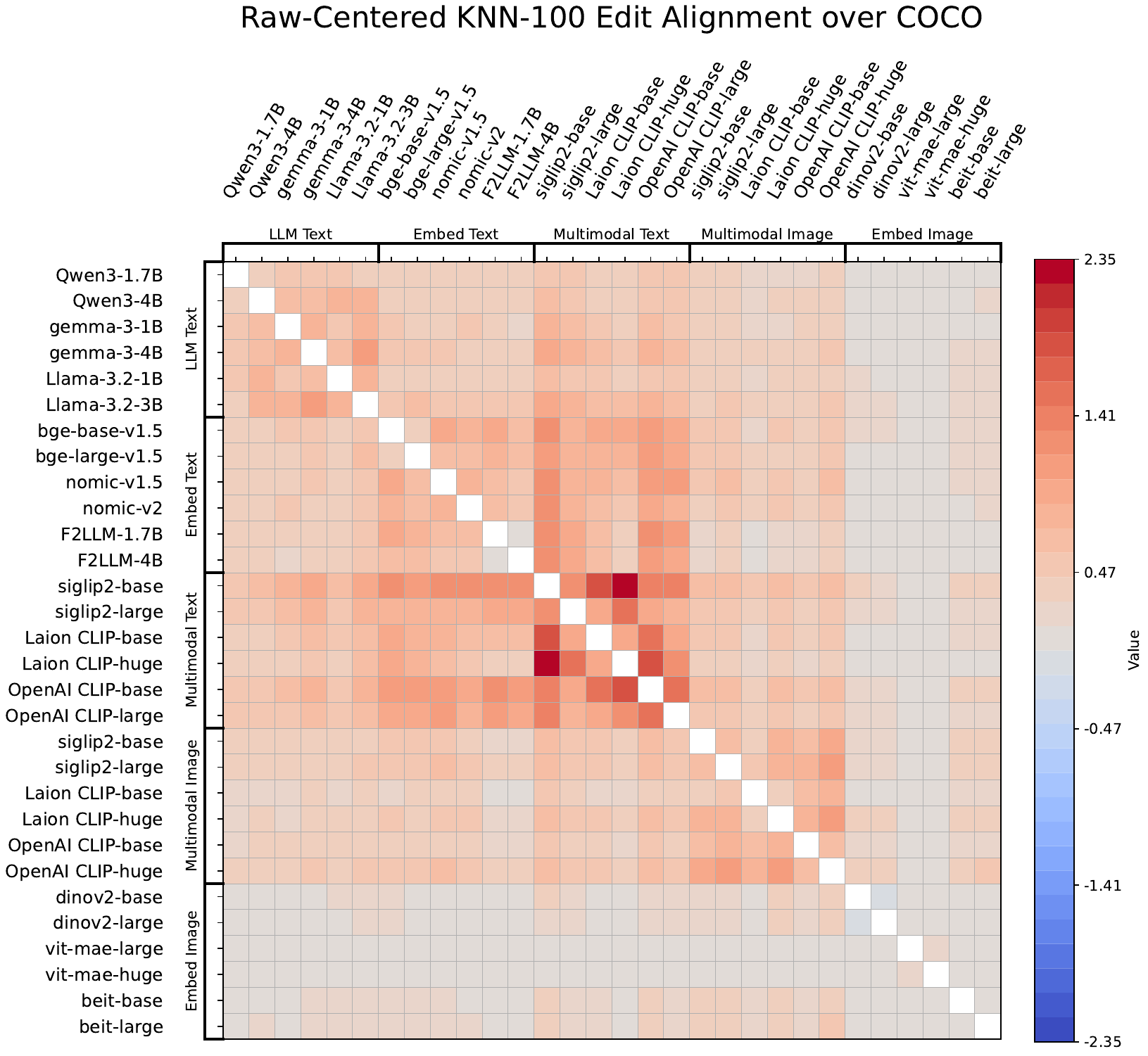}
    \end{minipage}
    \hfill
    \begin{minipage}[t]{0.3\textwidth}
        \centering
        \includegraphics[width = \linewidth]{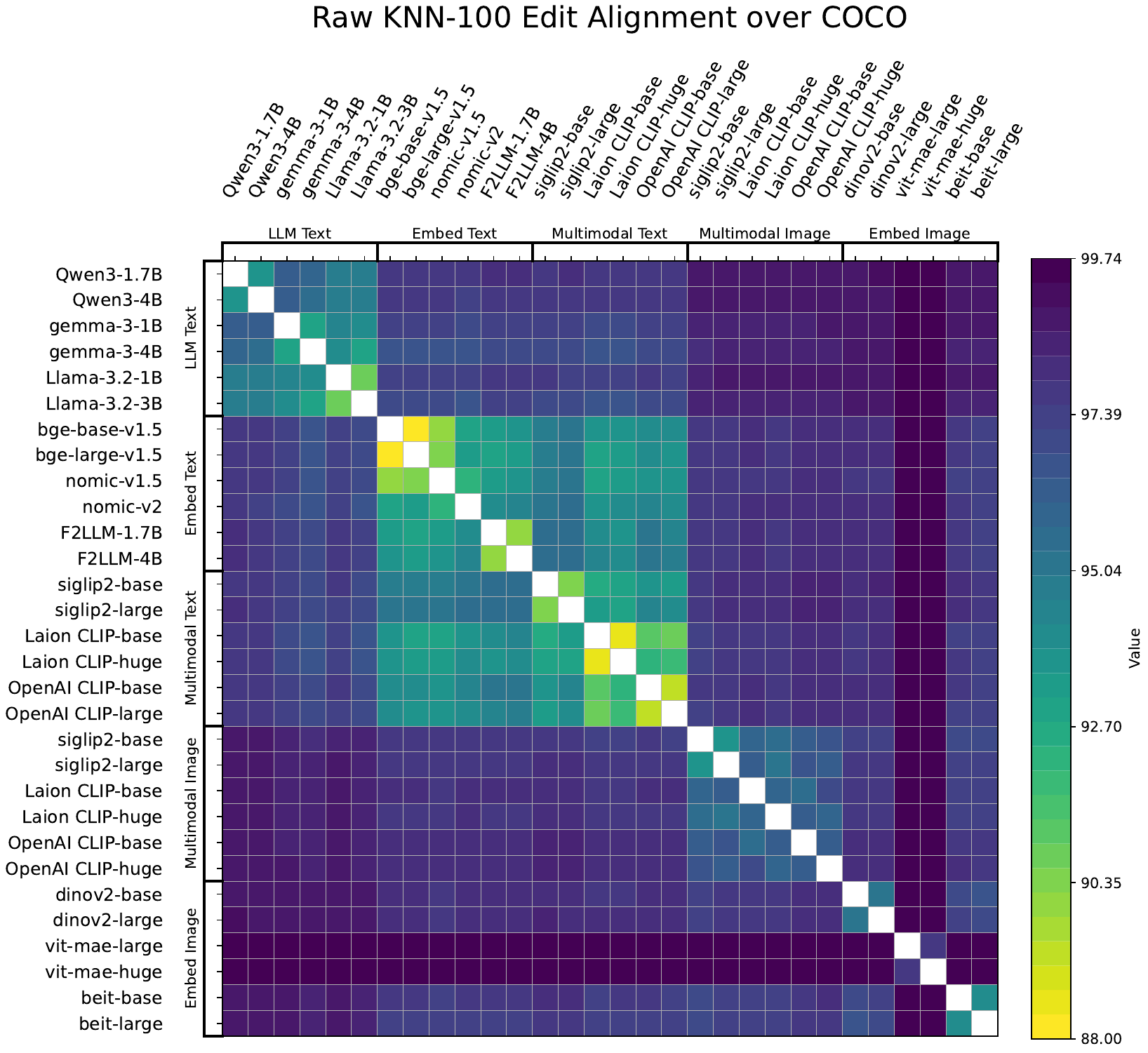}
    \end{minipage}
    \hfill
    \begin{minipage}[t]{0.3\textwidth}
        \centering
        \includegraphics[width = \linewidth]{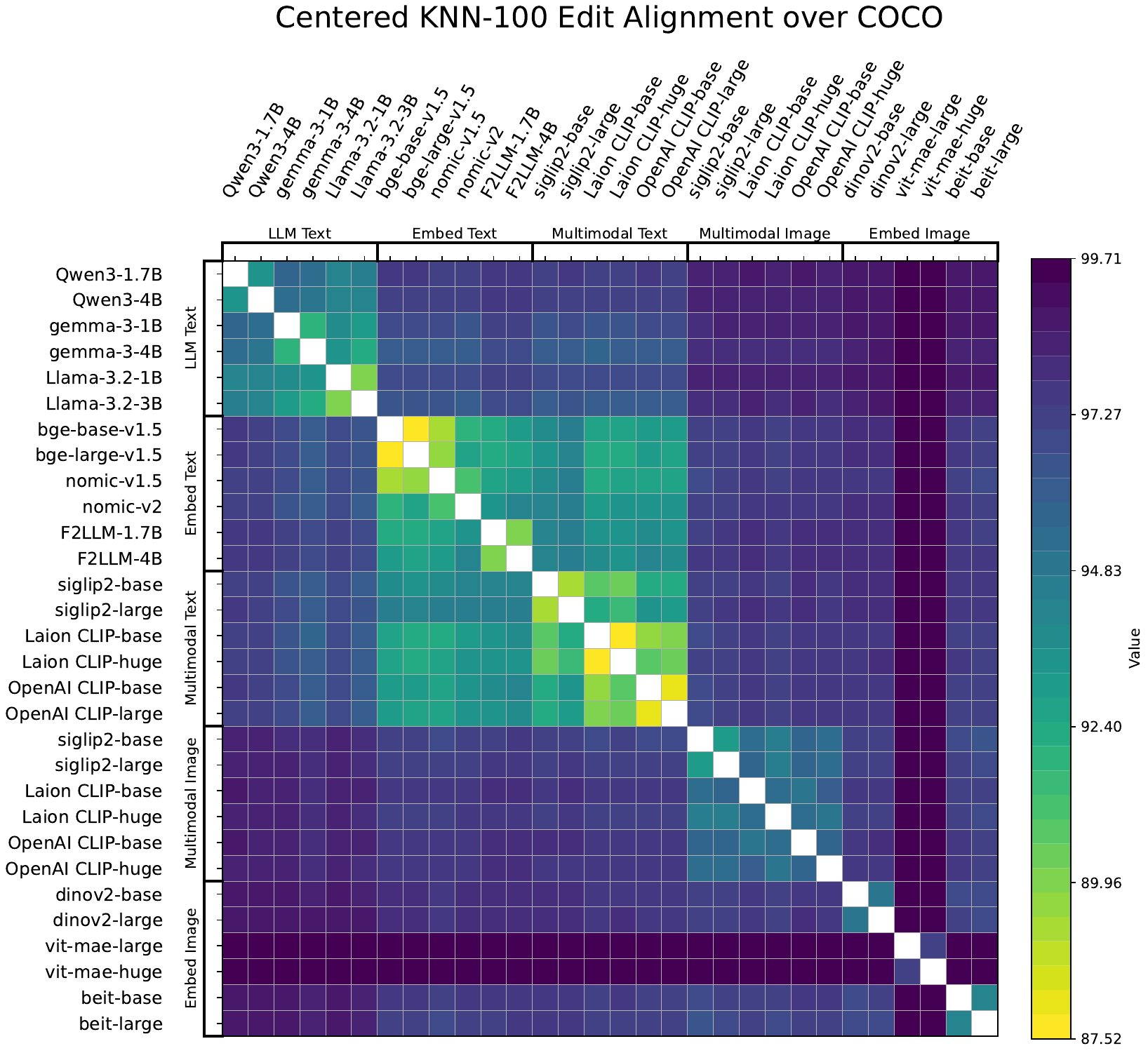}
    \end{minipage}
    \caption{Same plot as in Figure~\ref{fig:biasmain} but for the KNN-10 Overlap, KNN-100 Overlap, KNN-10 Edit Distance, and KNN-100 Edit distance metrics. In addition to differences, we plot the alignment values for raw features and for the centered features.}
    \label{fig:biascoco2}
\end{figure}
 We remark how consistent the improvement is across metrics. For nearly all pairs of models and metrics, there is improvement in the alignment after centering and re-normalizing.

\clearpage

\subsubsection{Experiments on CC3M}

\begin{figure}[htbp]
    \centering

    \begin{minipage}[t]{0.3\textwidth}
        \centering
        \includegraphics[width = \linewidth]{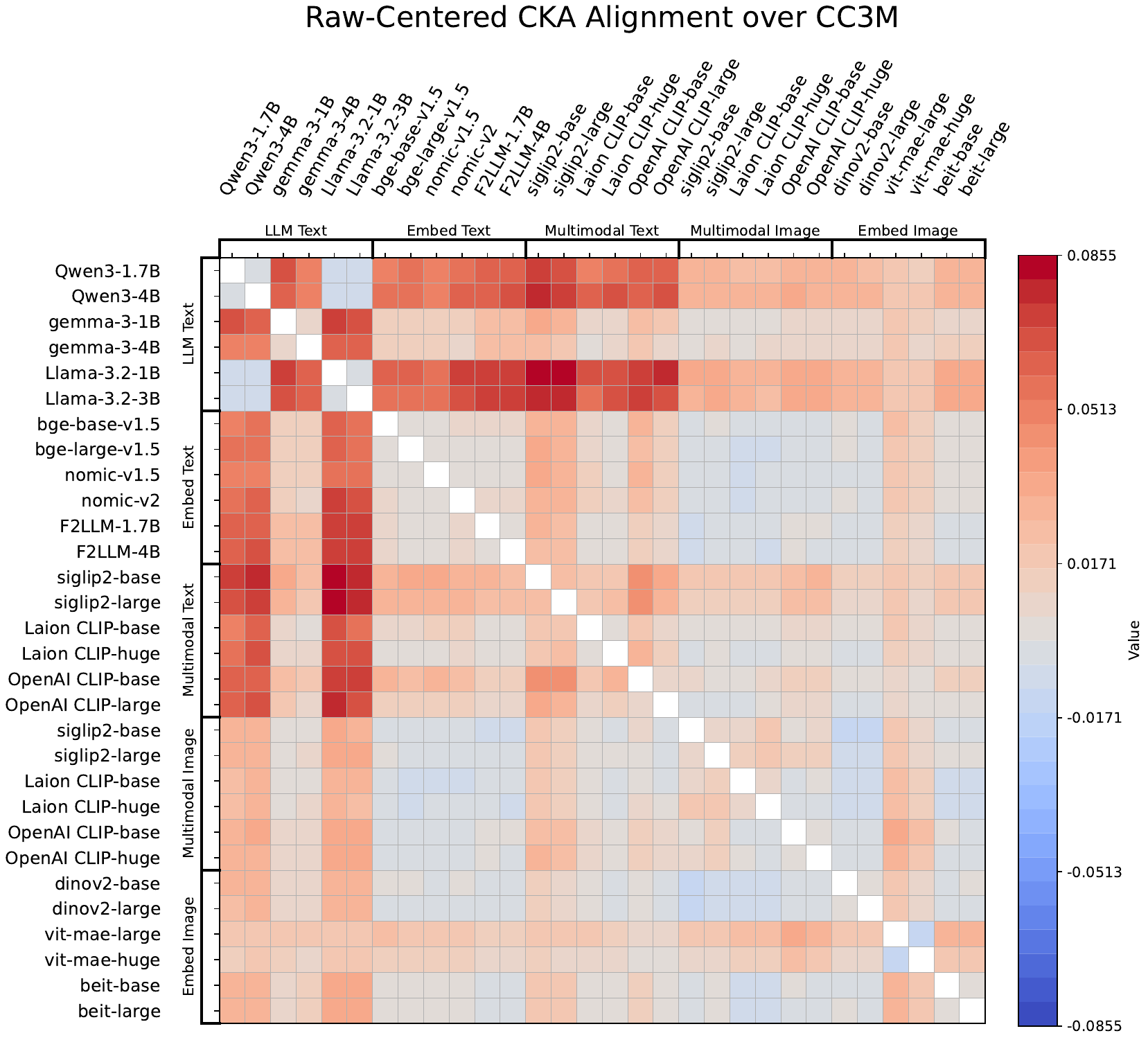}
    \end{minipage}
    \hfill
    \begin{minipage}[t]{0.3\textwidth}
        \centering
        \includegraphics[width = \linewidth]{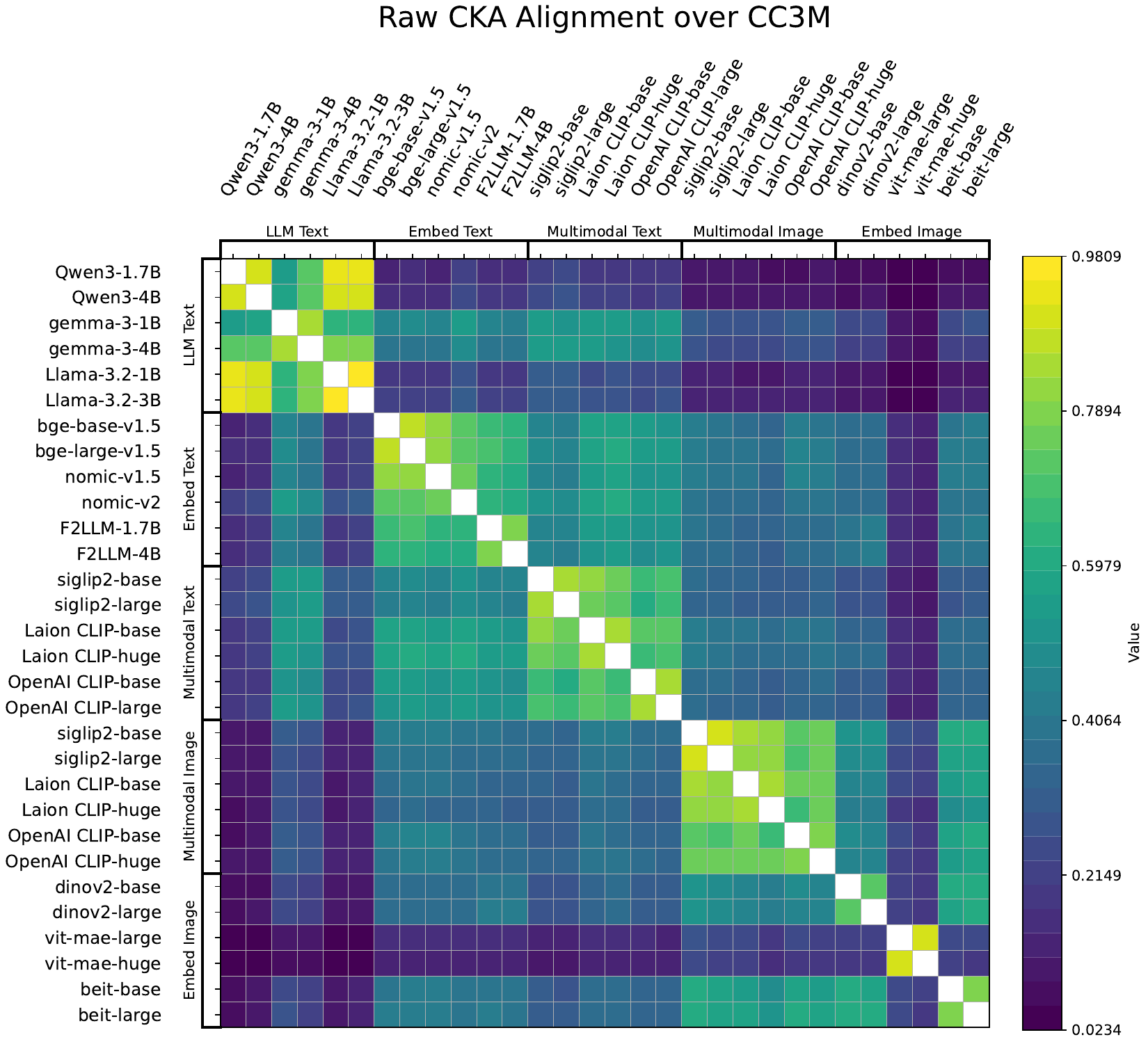}
    \end{minipage}
    \hfill
    \begin{minipage}[t]{0.3\textwidth}
        \centering
        \includegraphics[width = \linewidth]{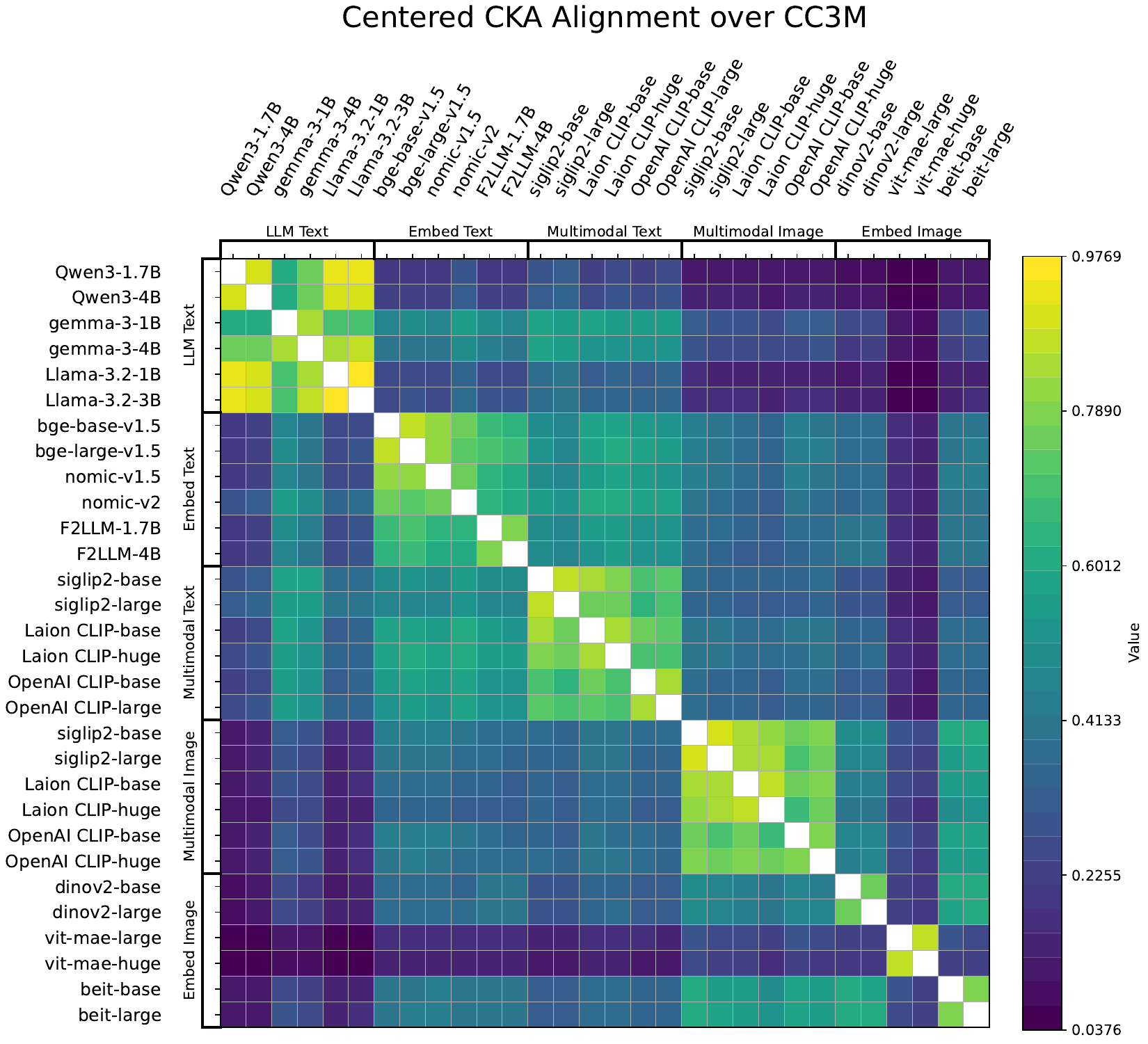}
    \end{minipage}
    
    \vspace{0.4cm}

        \begin{minipage}[t]{0.3\textwidth}
        \centering
        \includegraphics[width = \linewidth]{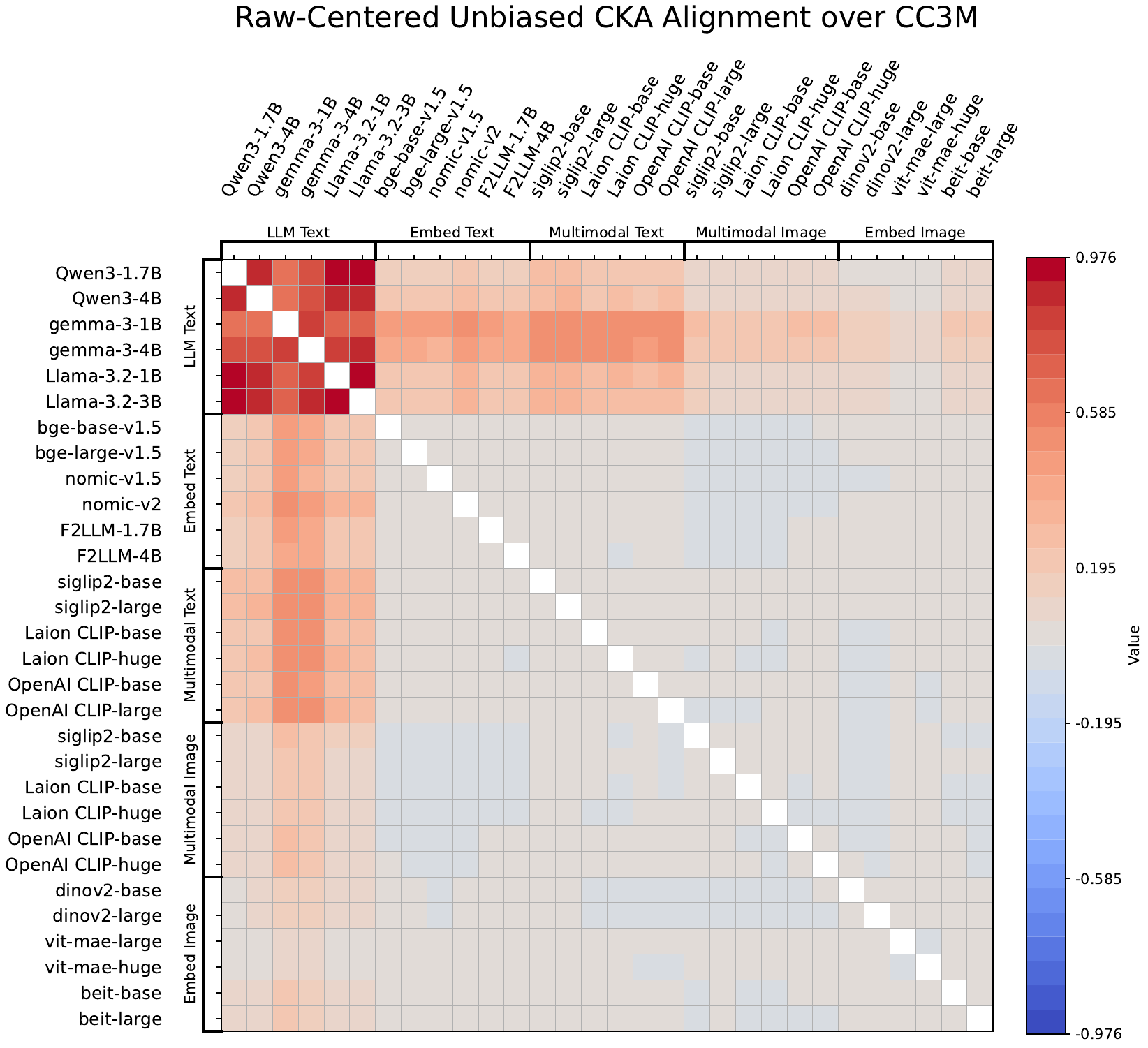}
    \end{minipage}
    \hfill
    \begin{minipage}[t]{0.3\textwidth}
        \centering
        \includegraphics[width = \linewidth]{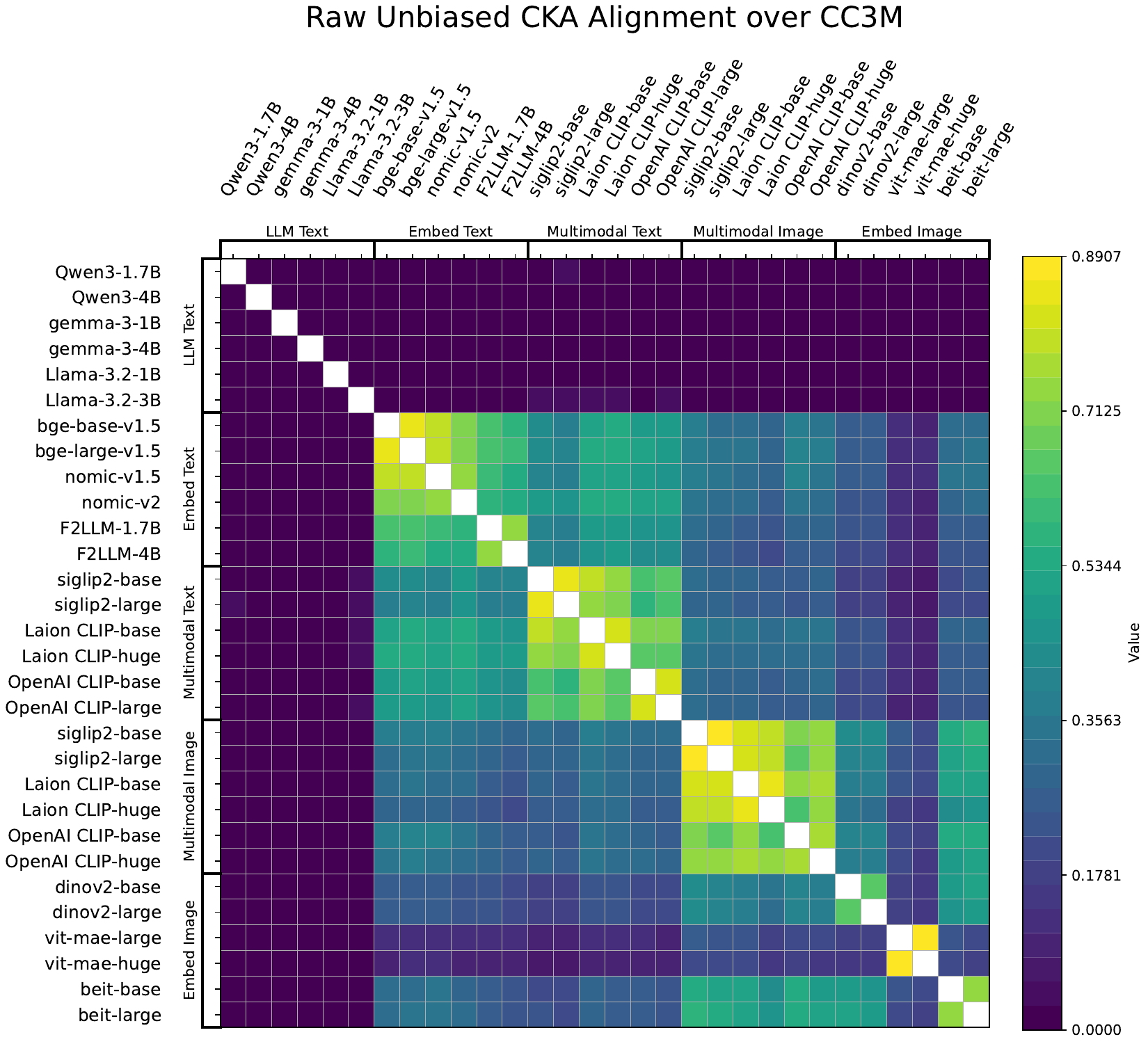}
    \end{minipage}
    \hfill
    \begin{minipage}[t]{0.3\textwidth}
        \centering
        \includegraphics[width = \linewidth]{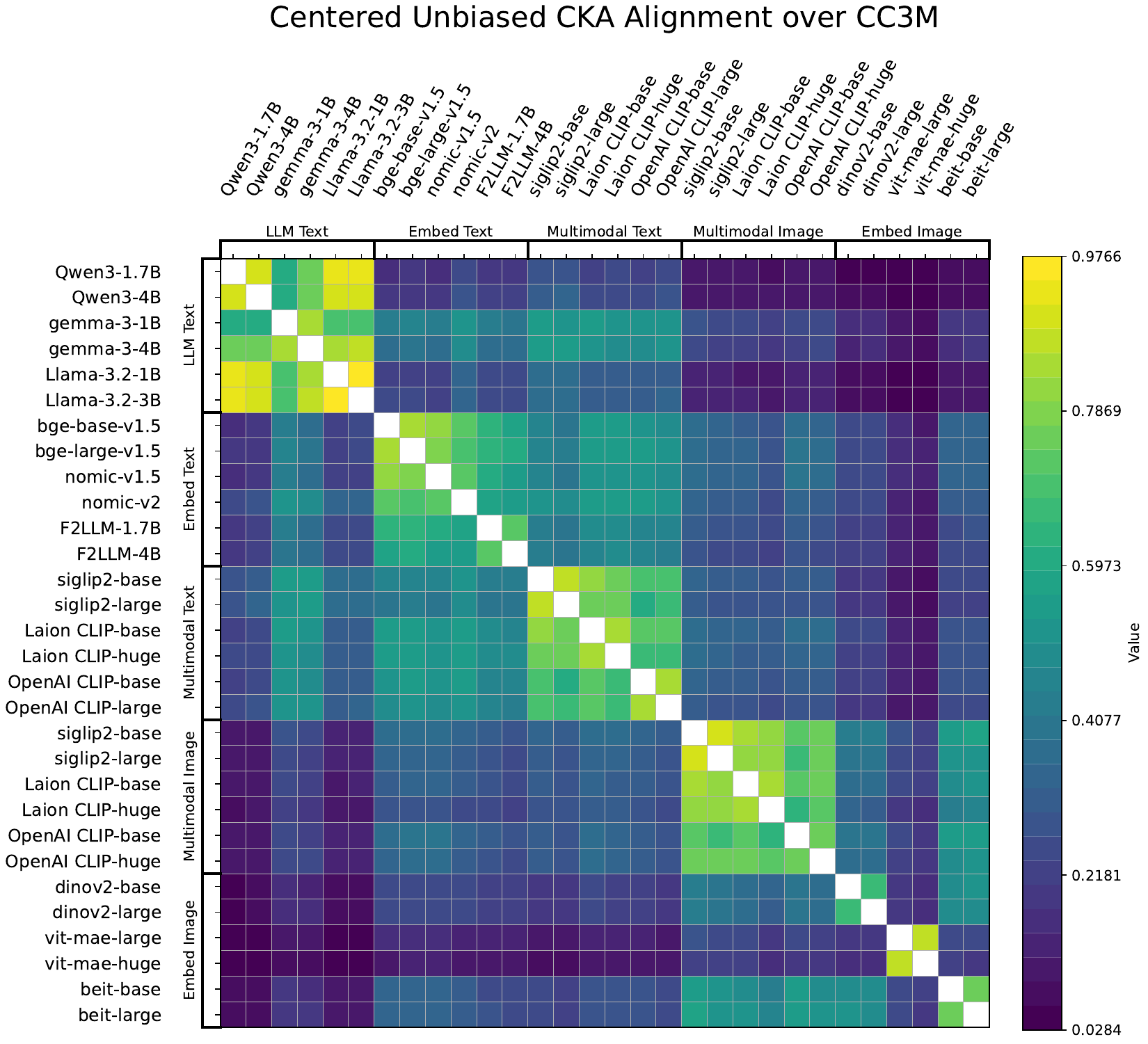}
    \end{minipage}

        \vspace{0.4cm}

        \begin{minipage}[t]{0.3\textwidth}
        \centering
        \includegraphics[width = \linewidth]{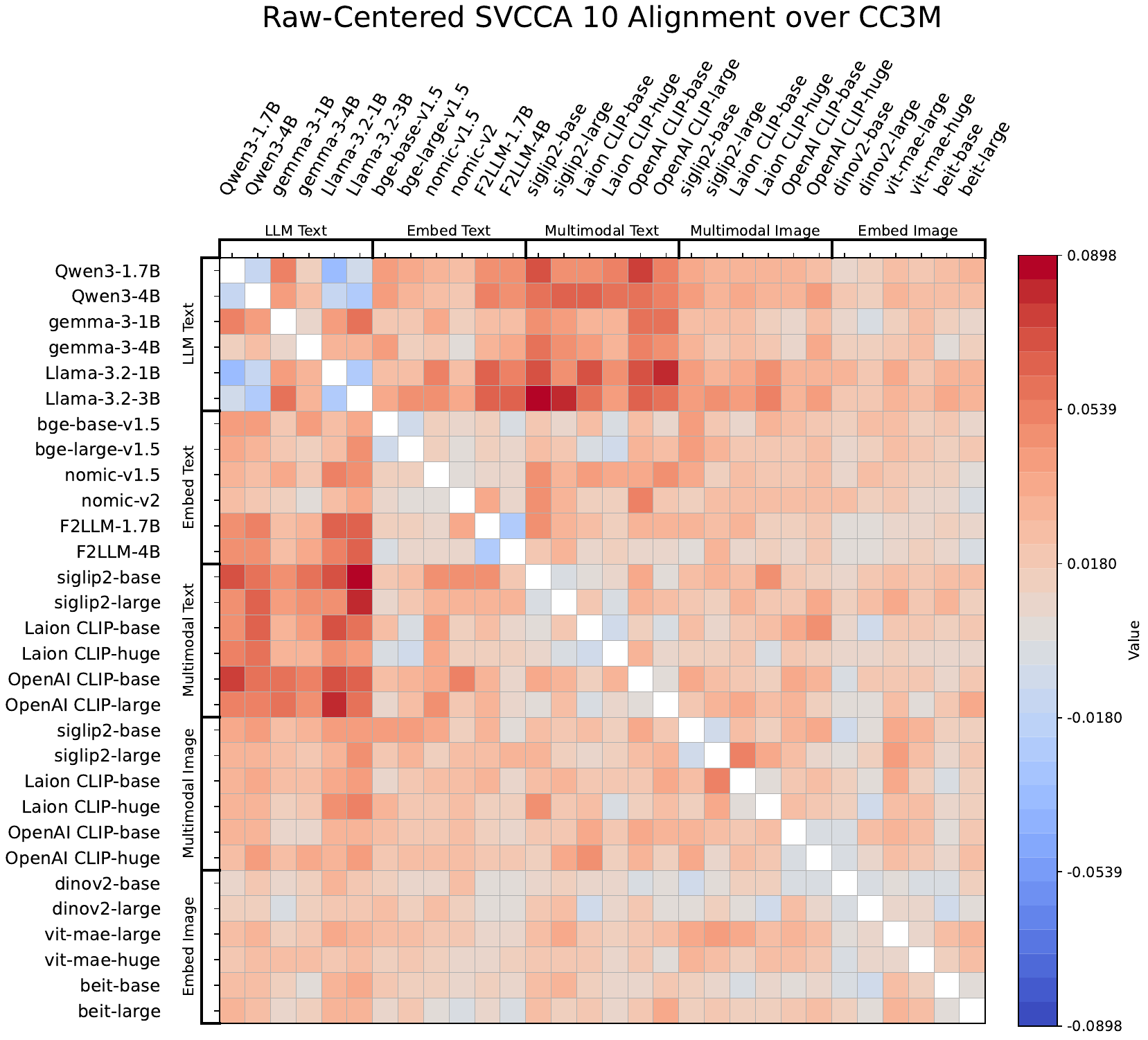}
    \end{minipage}
    \hfill
    \begin{minipage}[t]{0.3\textwidth}
        \centering
        \includegraphics[width = \linewidth]{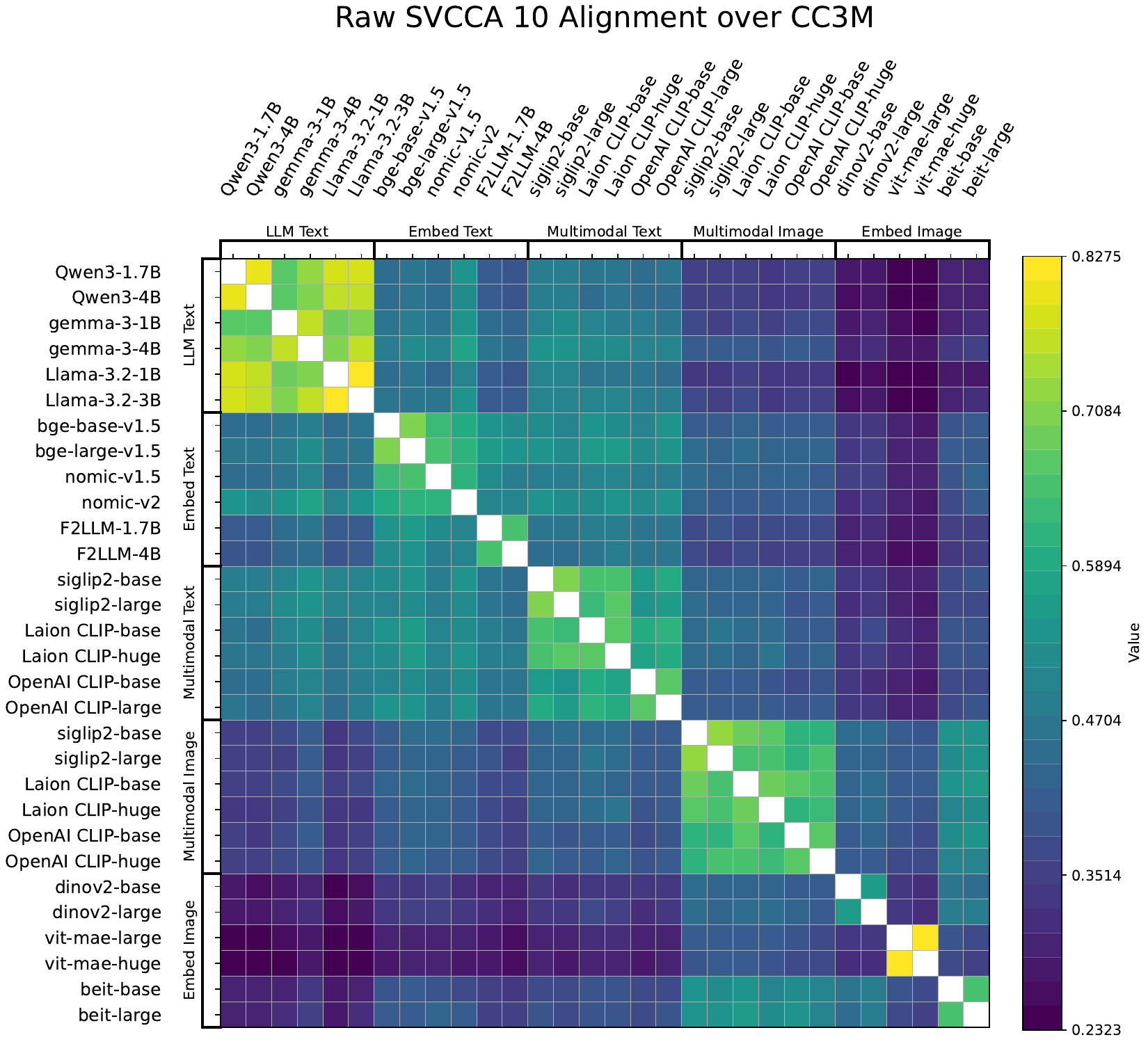}
    \end{minipage}
    \hfill
    \begin{minipage}[t]{0.3\textwidth}
        \centering
        \includegraphics[width = \linewidth]{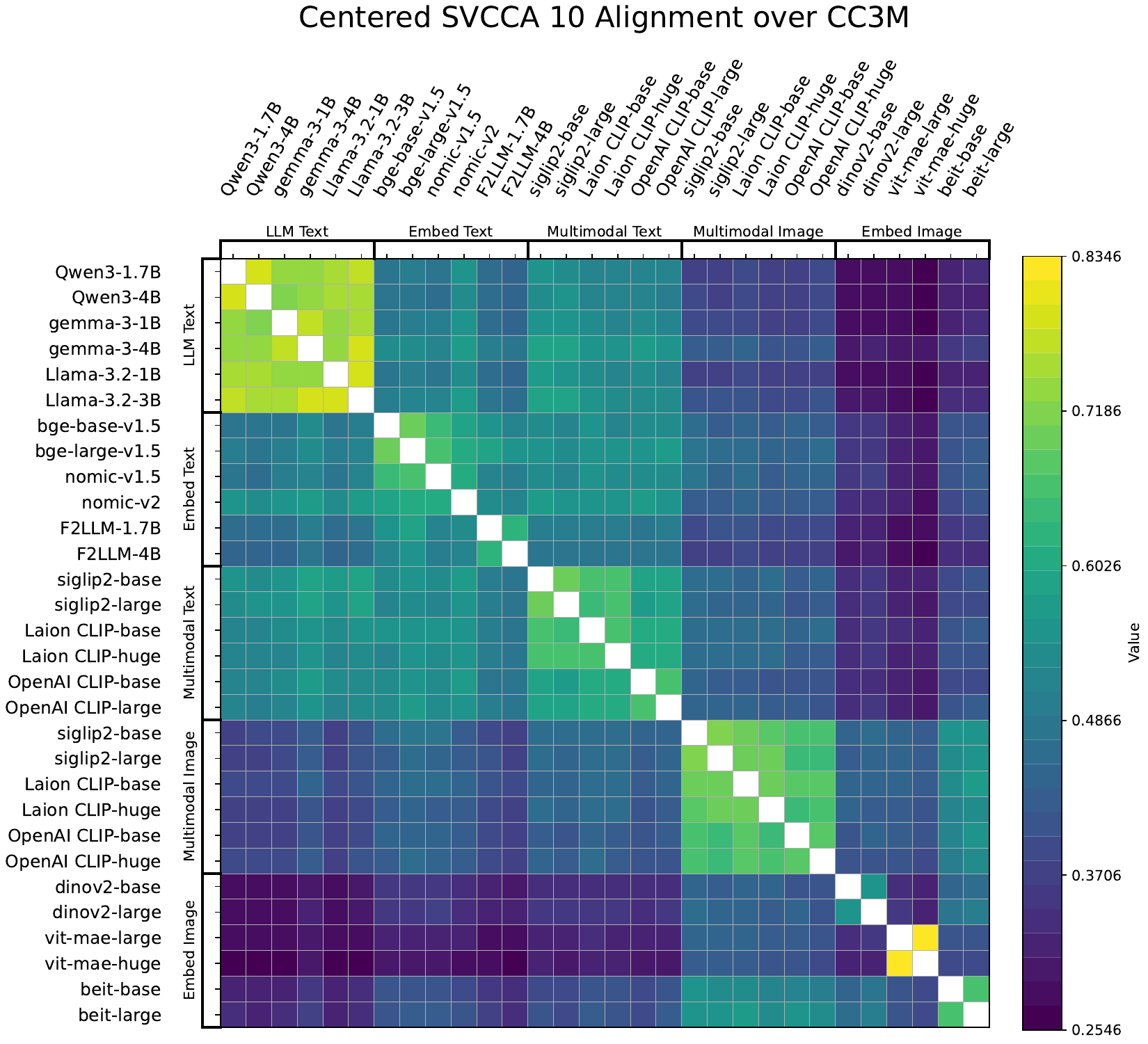}
    \end{minipage}

            \vspace{0.4cm}

    \begin{minipage}[t]{0.3\textwidth}
        \centering
        \includegraphics[width = \linewidth]{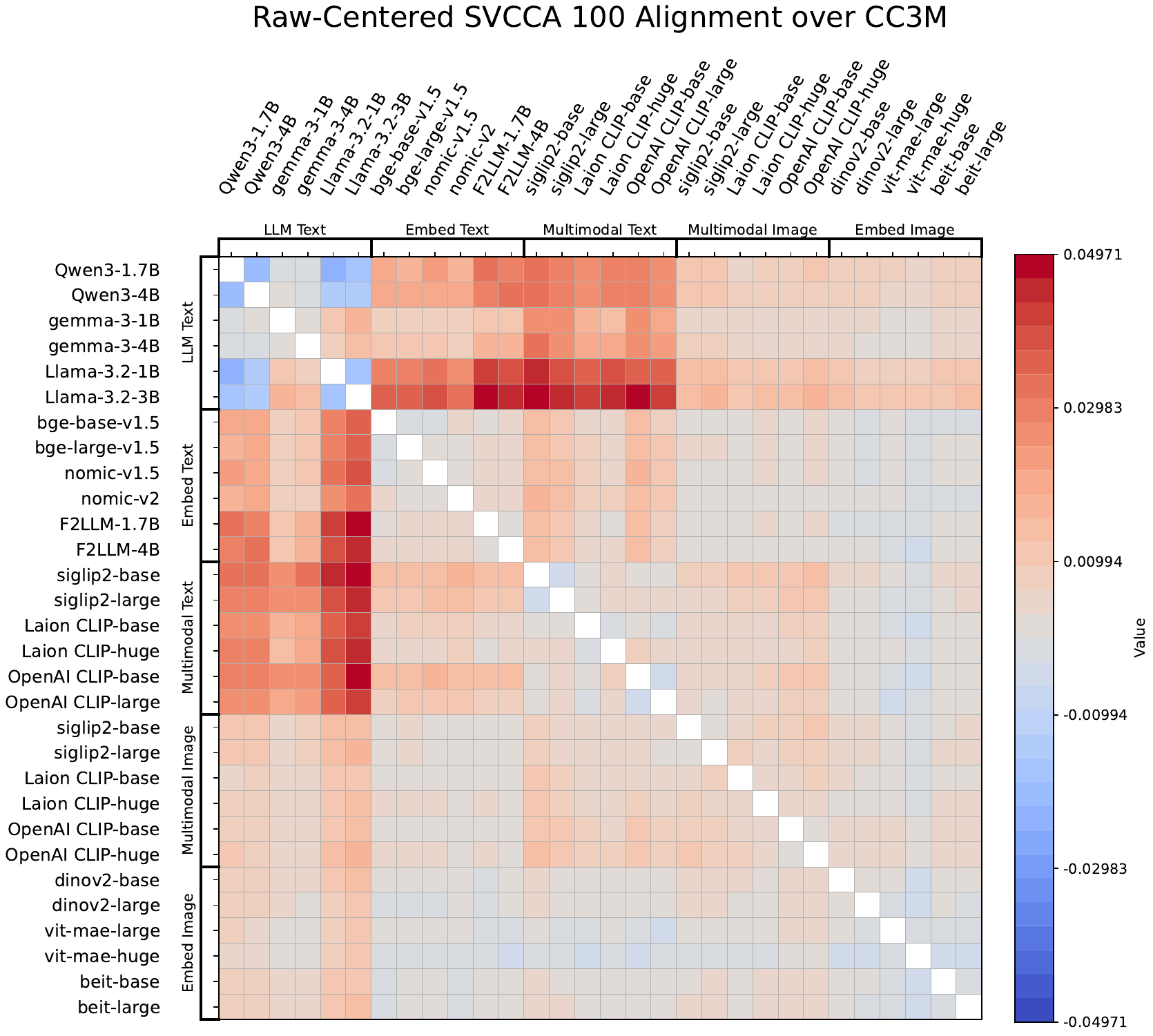}
    \end{minipage}
    \hfill
    \begin{minipage}[t]{0.3\textwidth}
        \centering
        \includegraphics[width = \linewidth]{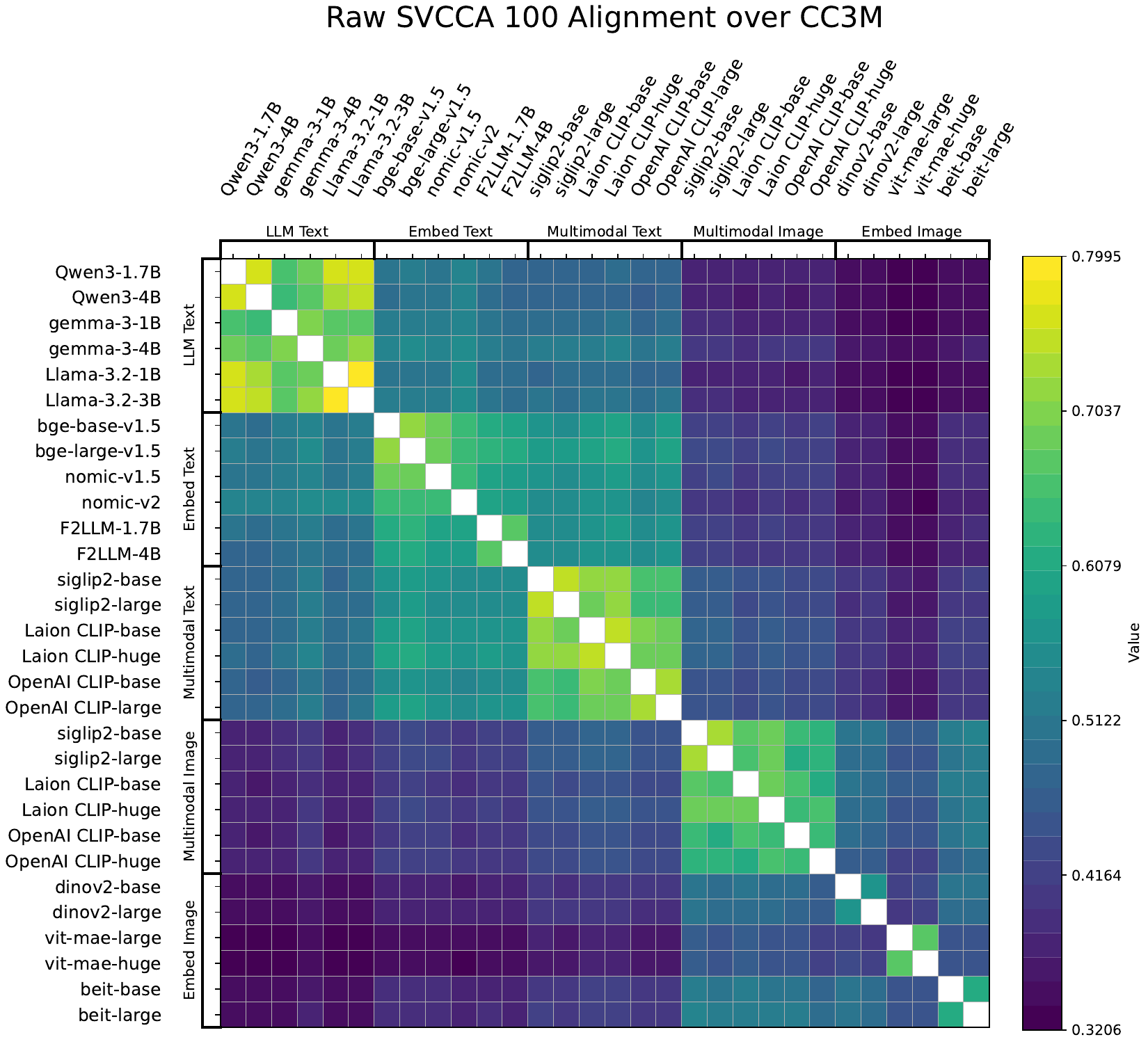}
    \end{minipage}
    \hfill
    \begin{minipage}[t]{0.3\textwidth}
        \centering
        \includegraphics[width = \linewidth]{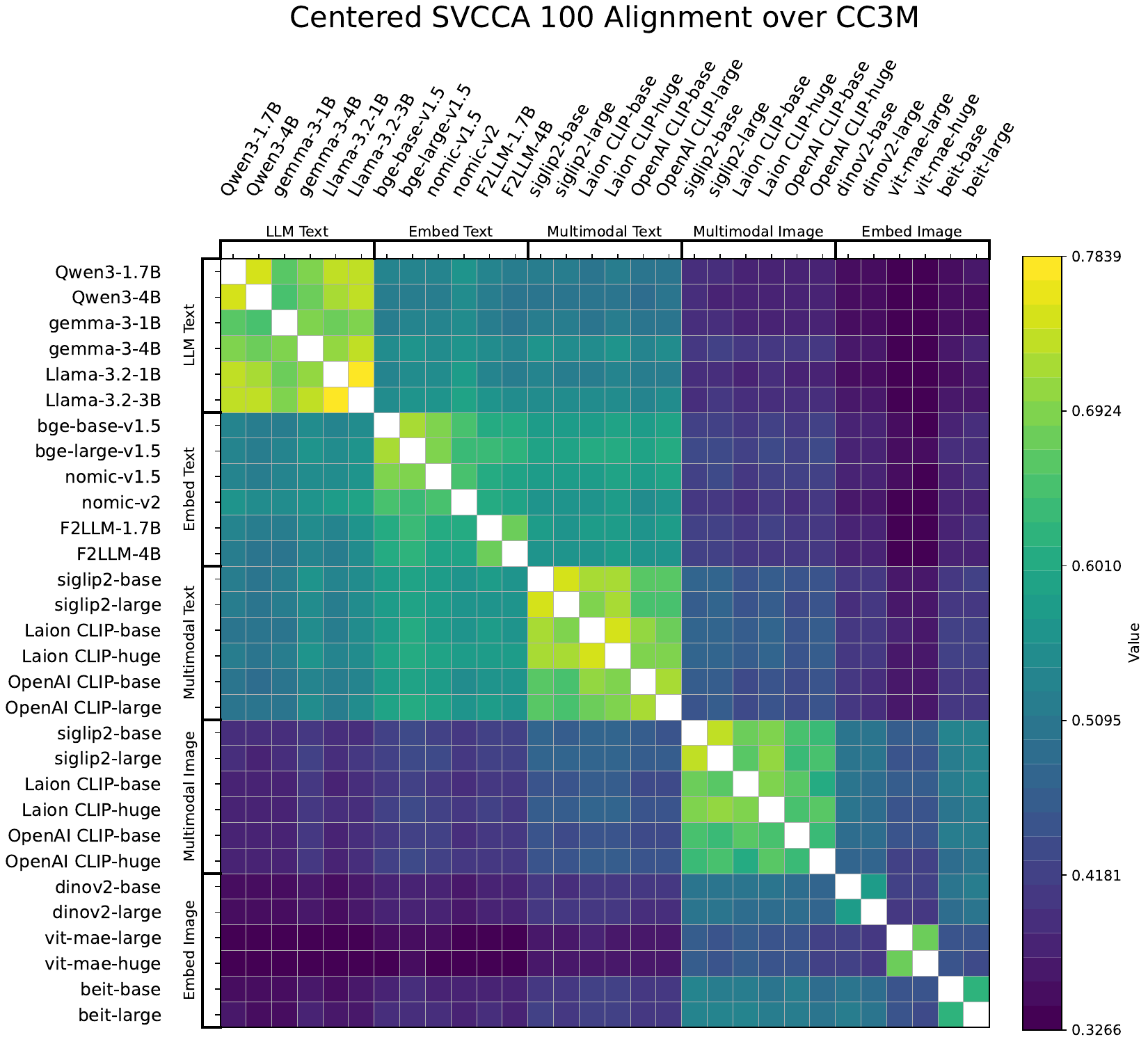}
    \end{minipage}
    \caption{Same plot as in Figure~\ref{fig:biascoco1} but over CC3M.}
    \label{fig:biascc3m1}
\end{figure}

\clearpage

\begin{figure}[htbp]
    \centering

    \begin{minipage}[t]{0.3\textwidth}
        \centering
        \includegraphics[width = \linewidth]{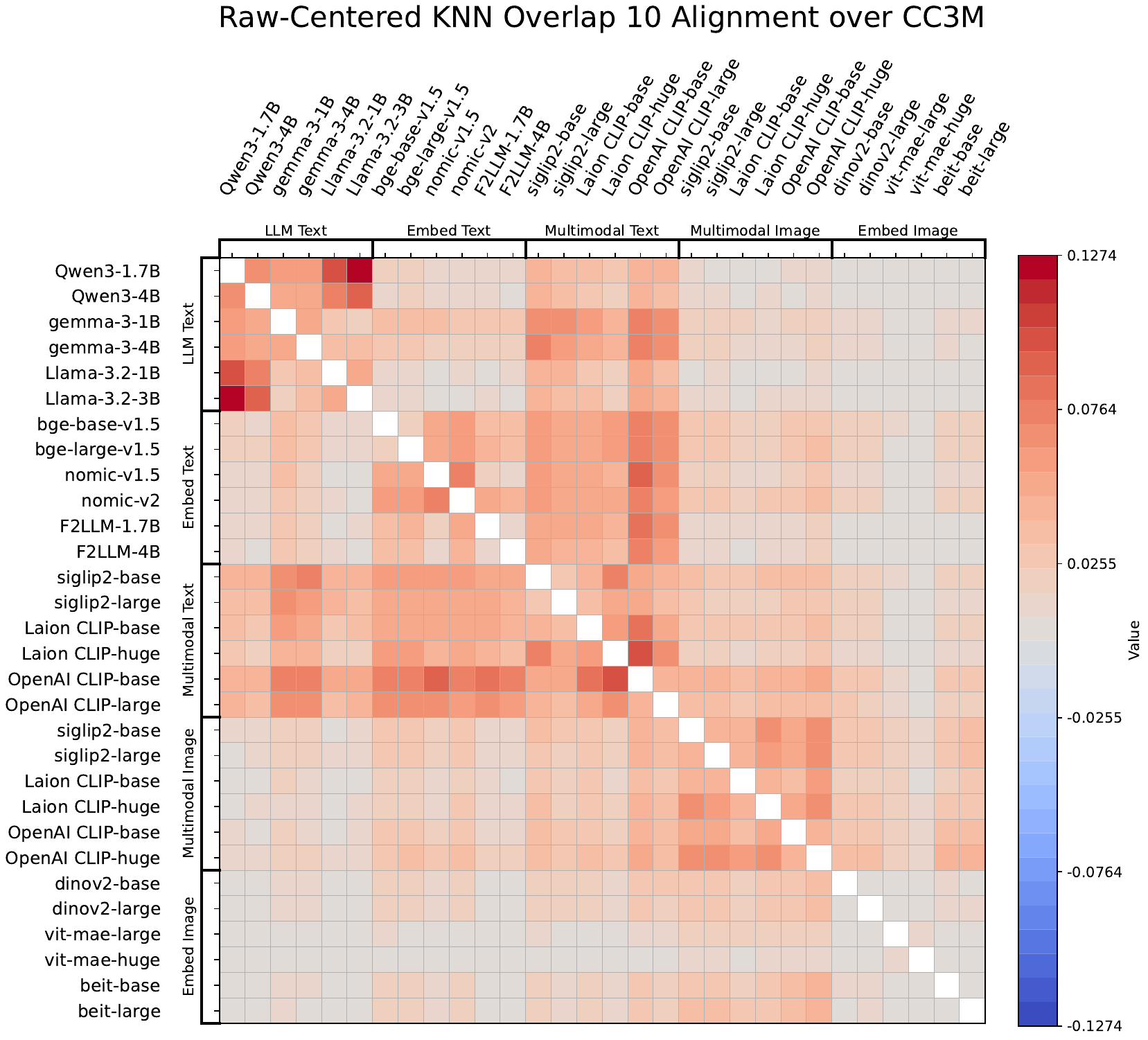}
    \end{minipage}
    \hfill
    \begin{minipage}[t]{0.3\textwidth}
        \centering
        \includegraphics[width = \linewidth]{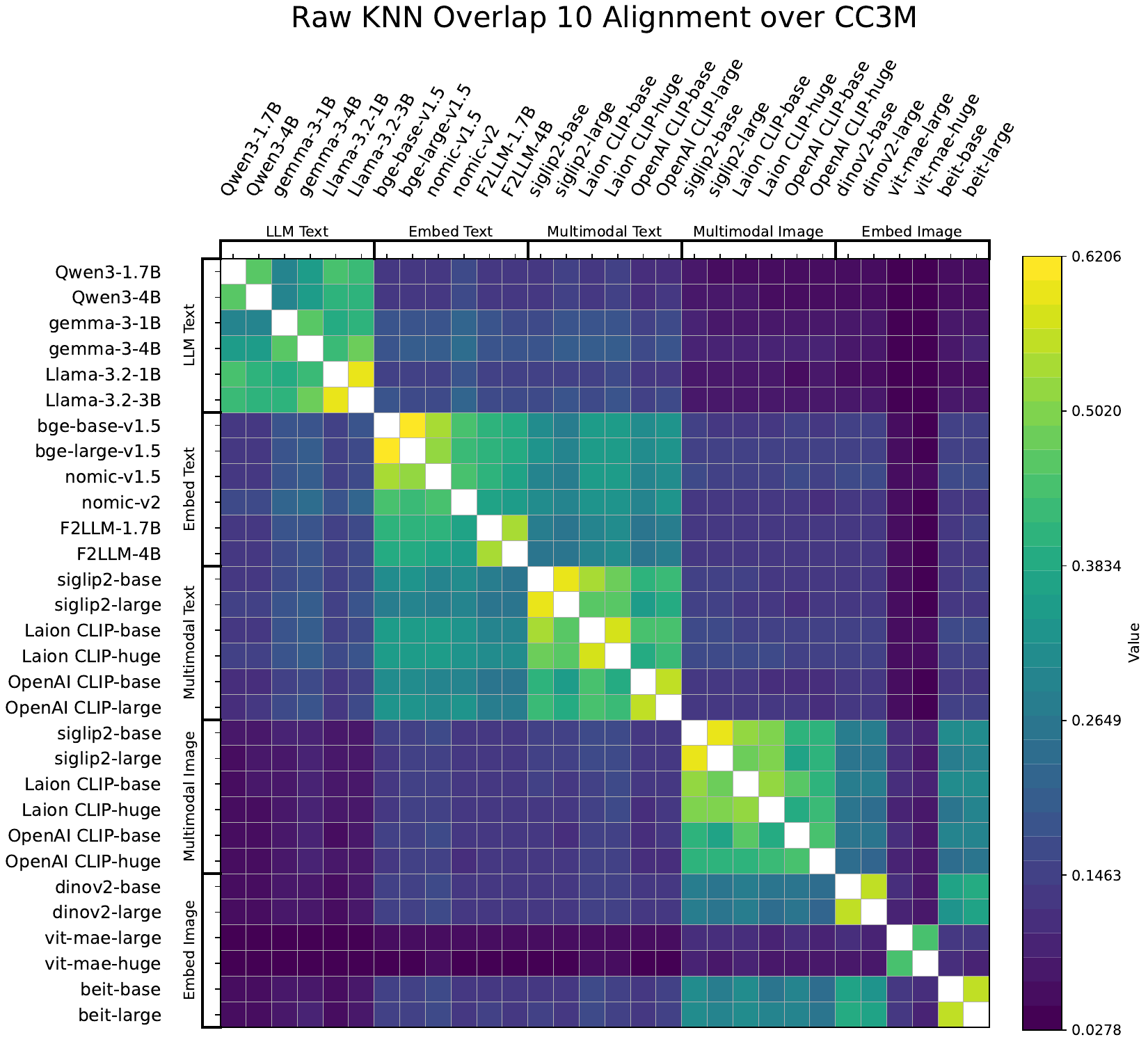}
    \end{minipage}
    \hfill
    \begin{minipage}[t]{0.3\textwidth}
        \centering
        \includegraphics[width = \linewidth]{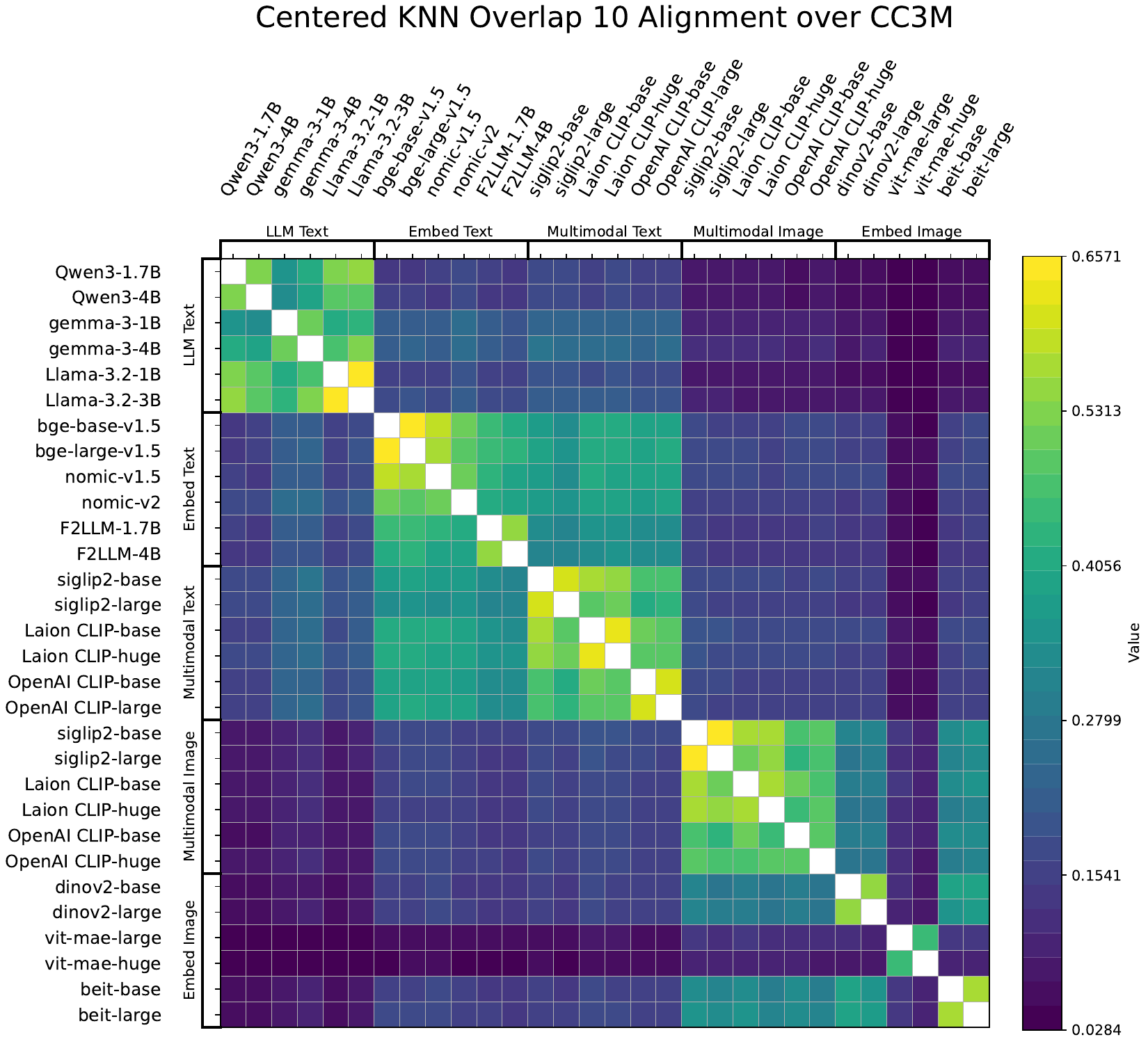}
    \end{minipage}
    
    \vspace{0.4cm}

        \begin{minipage}[t]{0.3\textwidth}
        \centering
        \includegraphics[width = \linewidth]{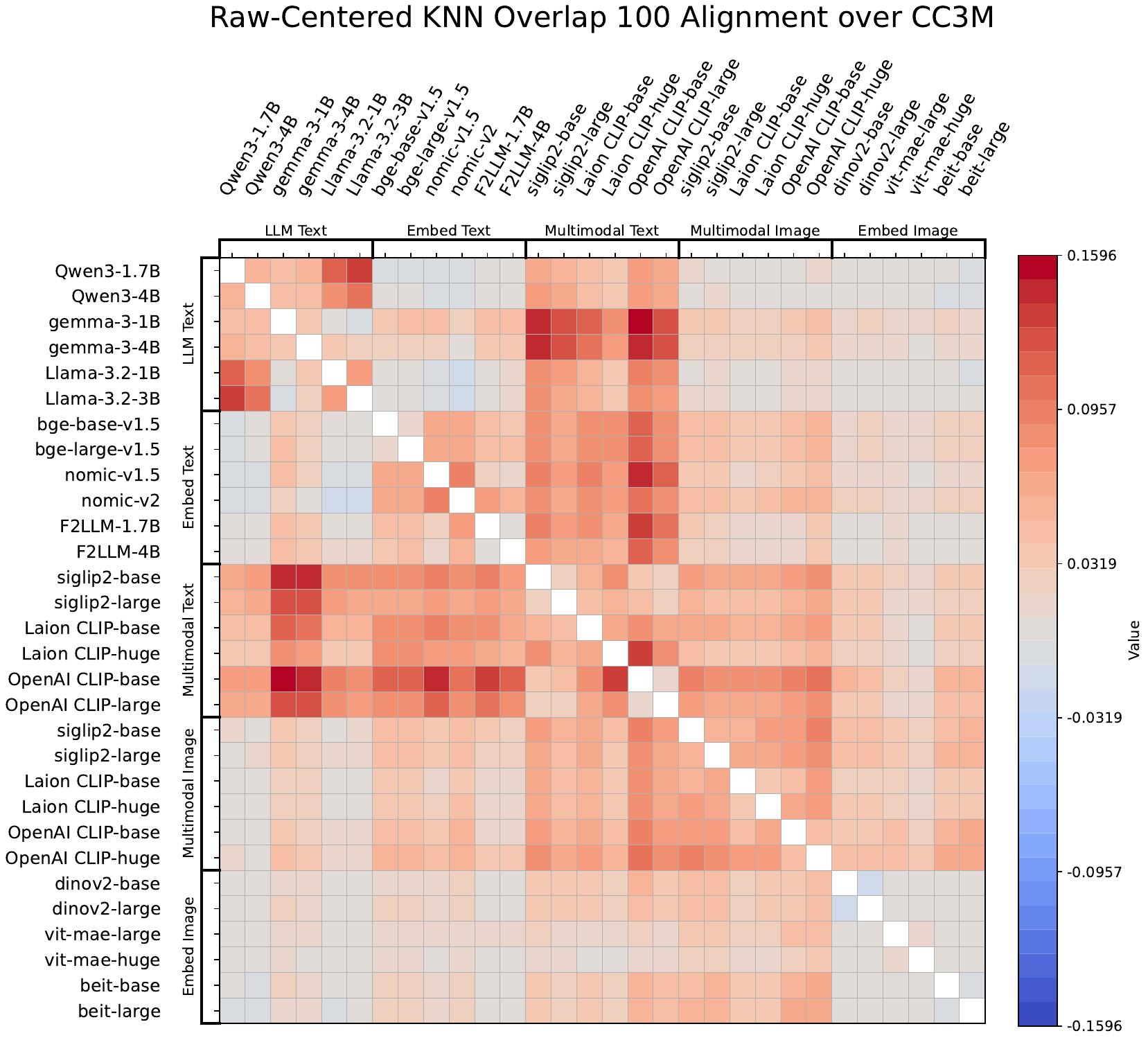}
    \end{minipage}
    \hfill
    \begin{minipage}[t]{0.3\textwidth}
        \centering
        \includegraphics[width = \linewidth]{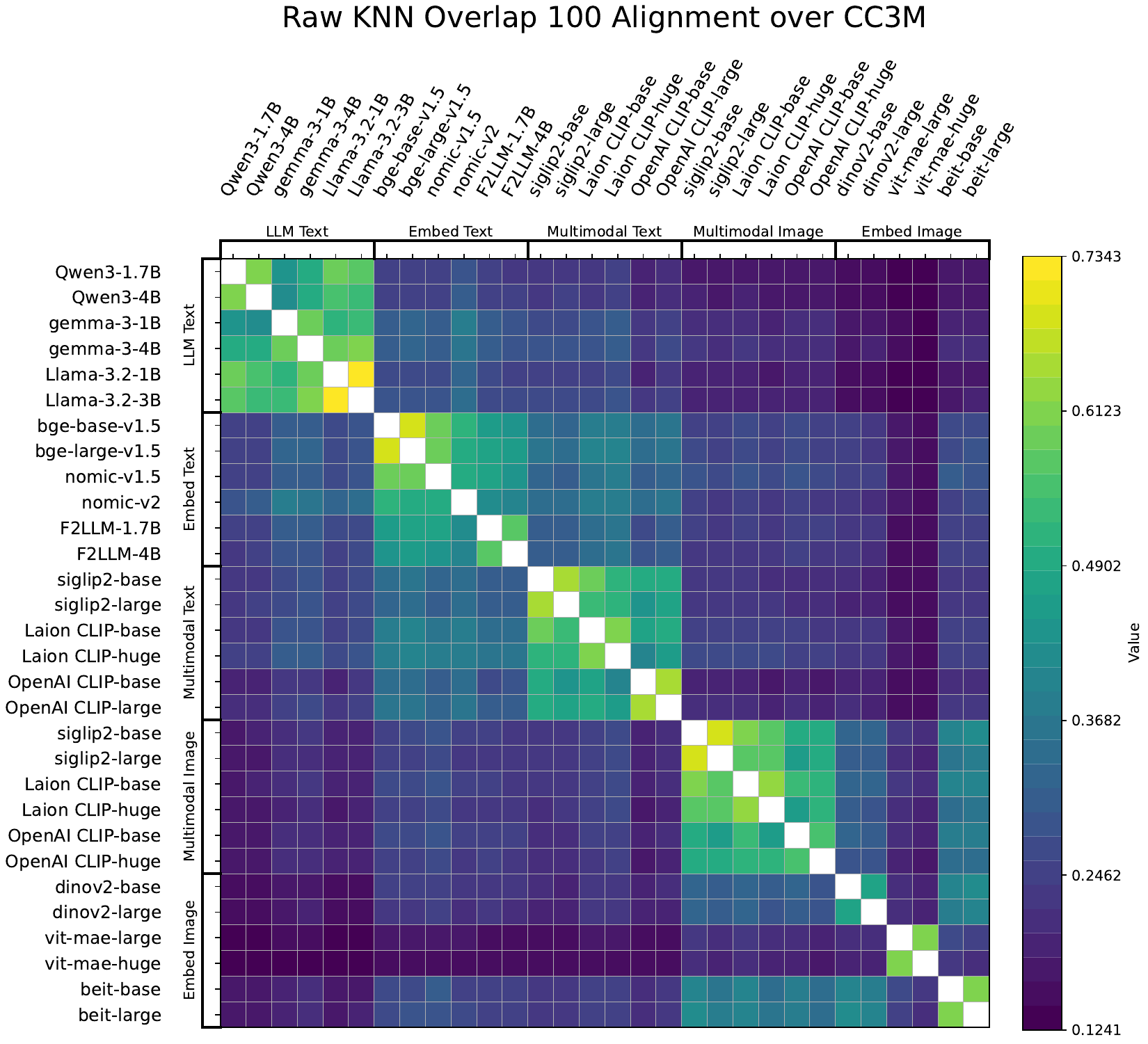}
    \end{minipage}
    \hfill
    \begin{minipage}[t]{0.3\textwidth}
        \centering
        \includegraphics[width = \linewidth]{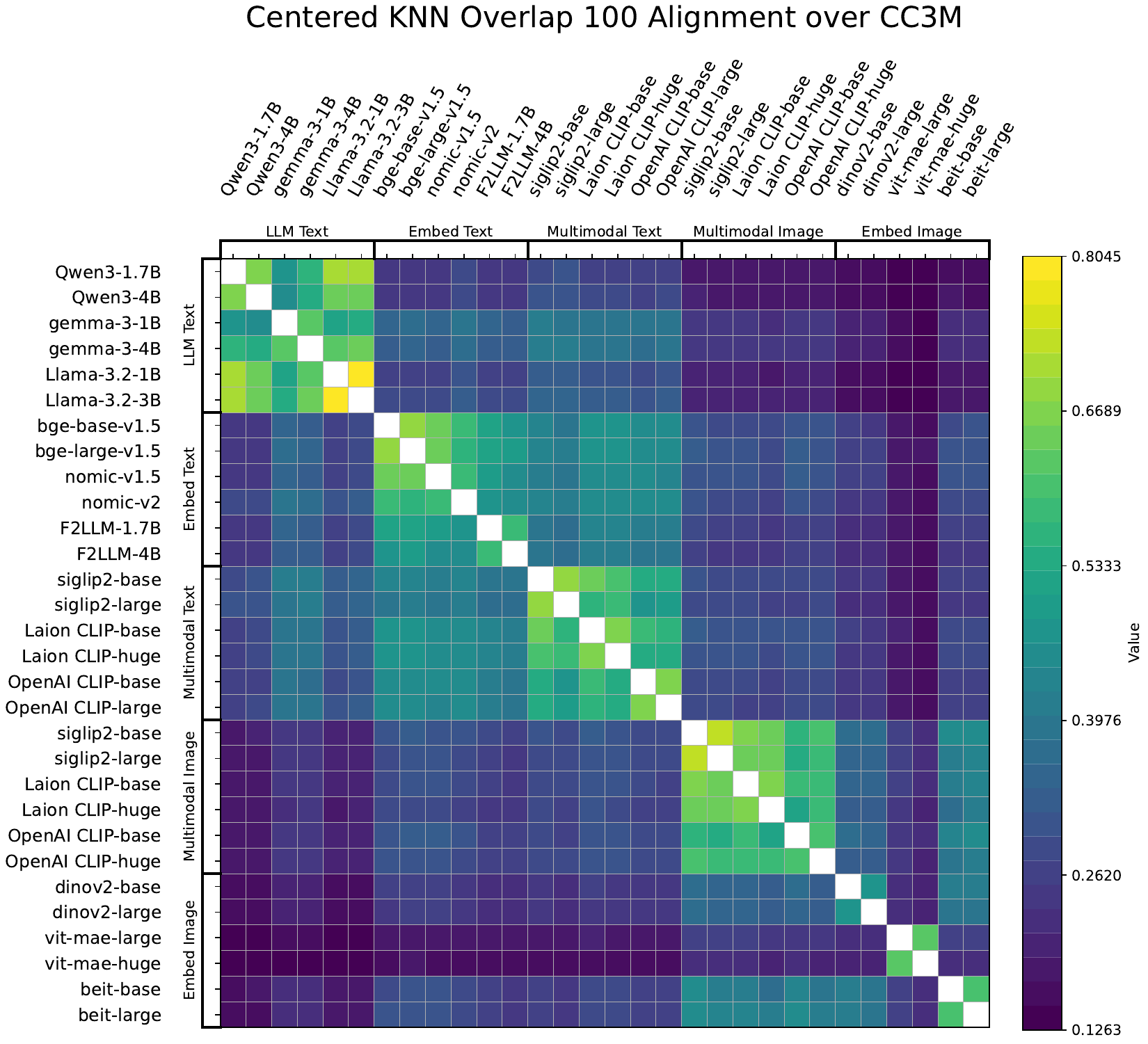}
    \end{minipage}

        \vspace{0.4cm}

        \begin{minipage}[t]{0.3\textwidth}
        \centering
        \includegraphics[width = \linewidth]{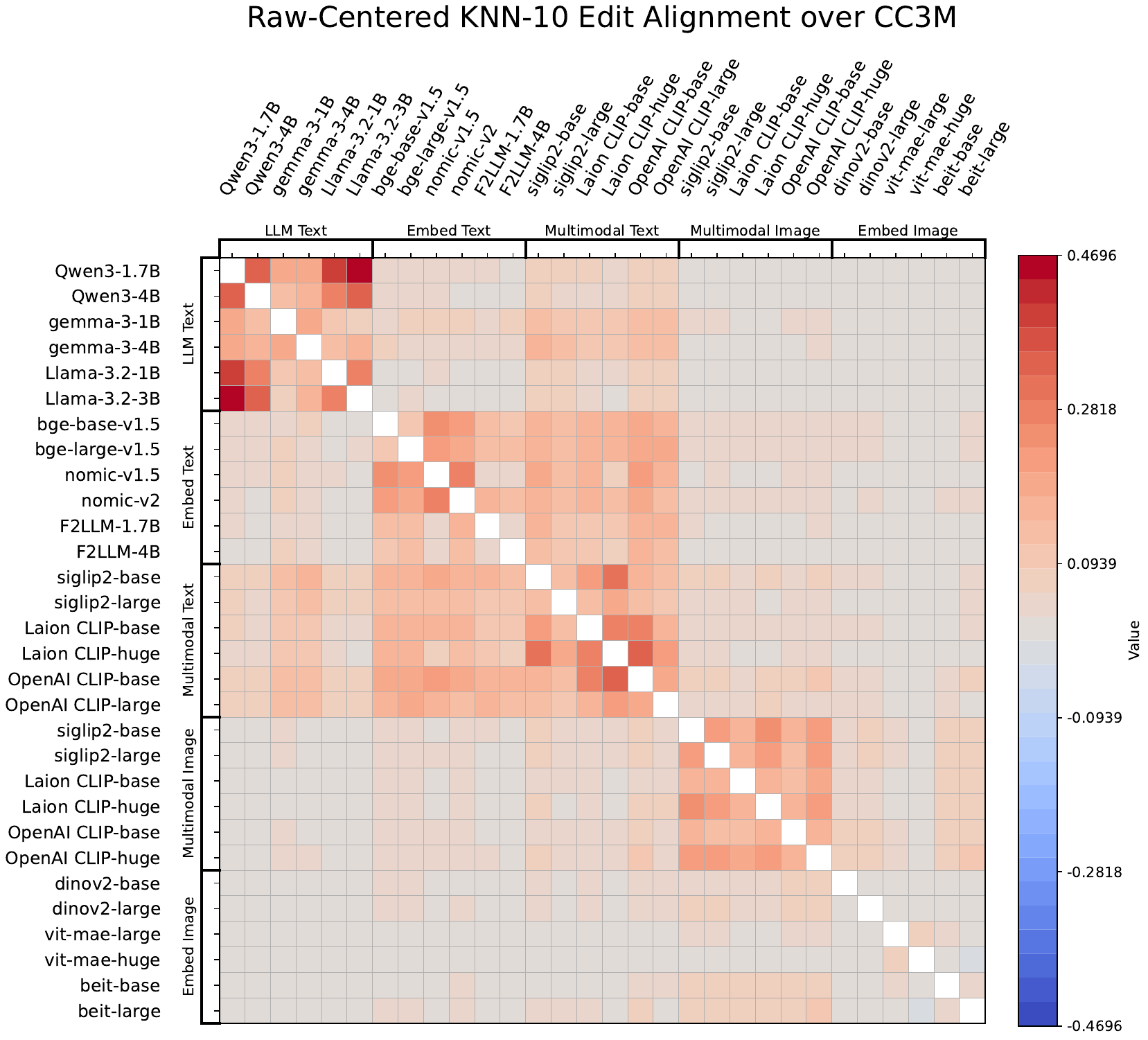}
    \end{minipage}
    \hfill
    \begin{minipage}[t]{0.3\textwidth}
        \centering
        \includegraphics[width = \linewidth]{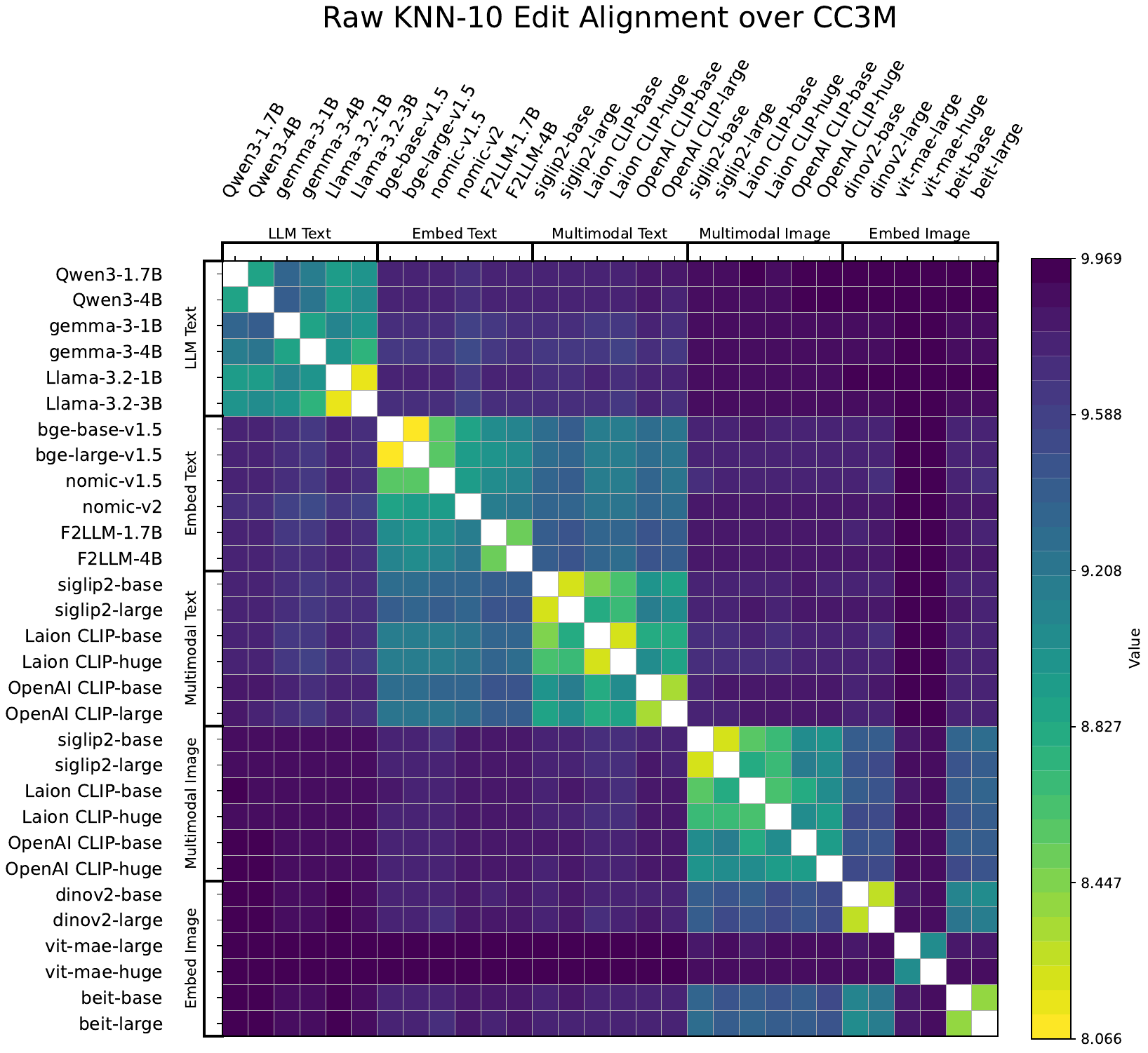}
    \end{minipage}
    \hfill
    \begin{minipage}[t]{0.3\textwidth}
        \centering
        \includegraphics[width = \linewidth]{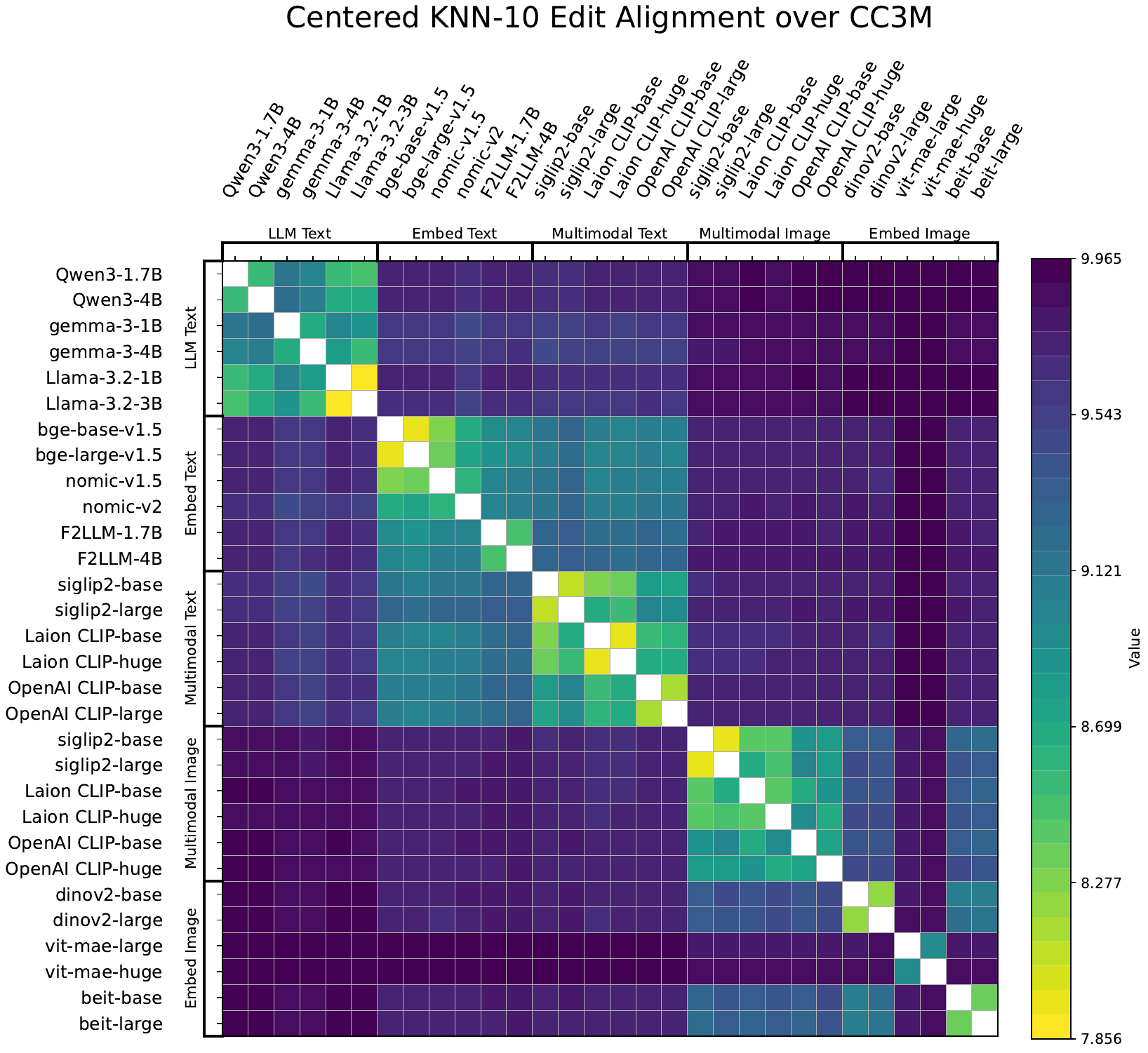}
    \end{minipage}

            \vspace{0.4cm}

    \begin{minipage}[t]{0.3\textwidth}
        \centering
        \includegraphics[width = \linewidth]{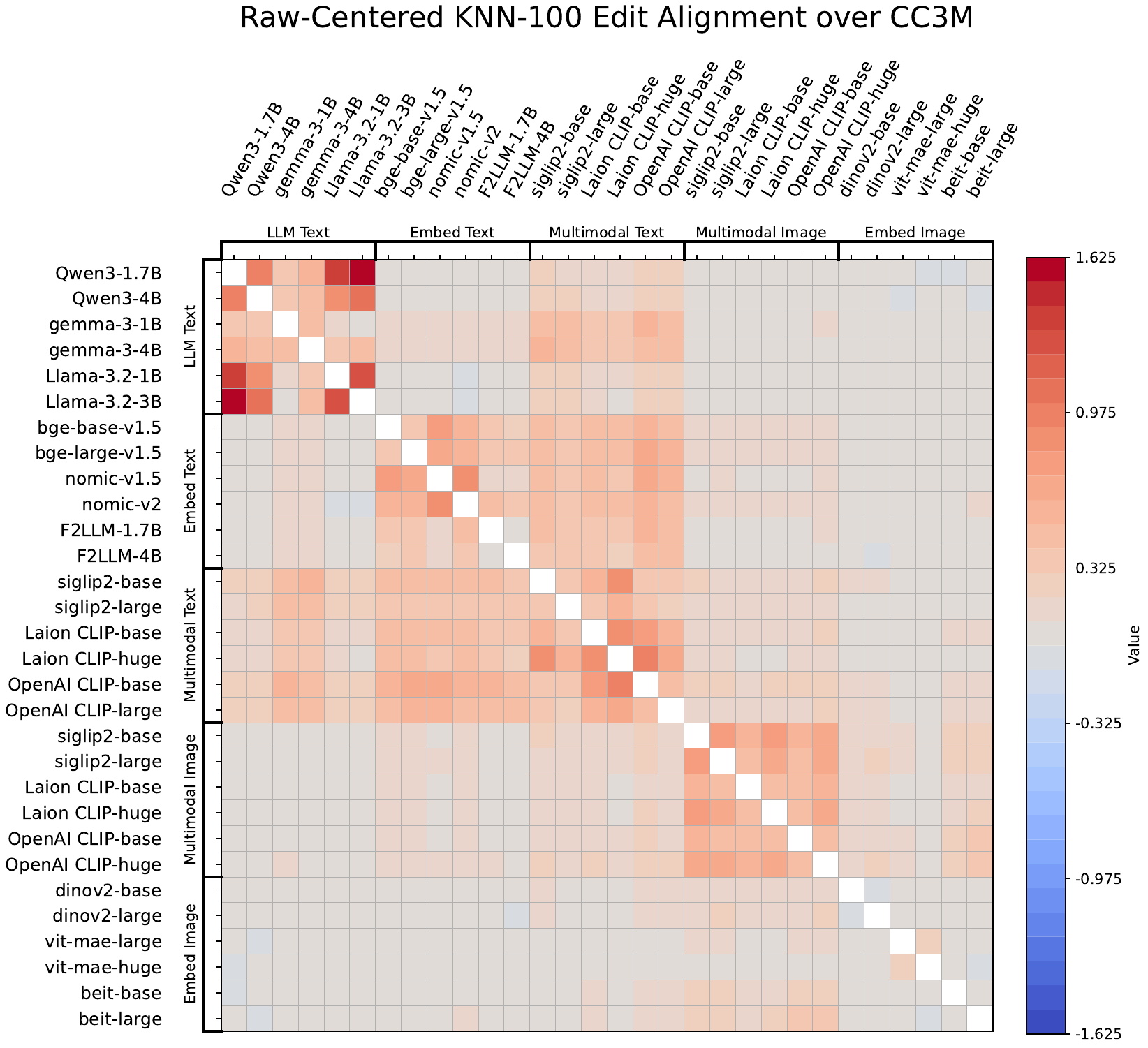}
    \end{minipage}
    \hfill
    \begin{minipage}[t]{0.3\textwidth}
        \centering
        \includegraphics[width = \linewidth]{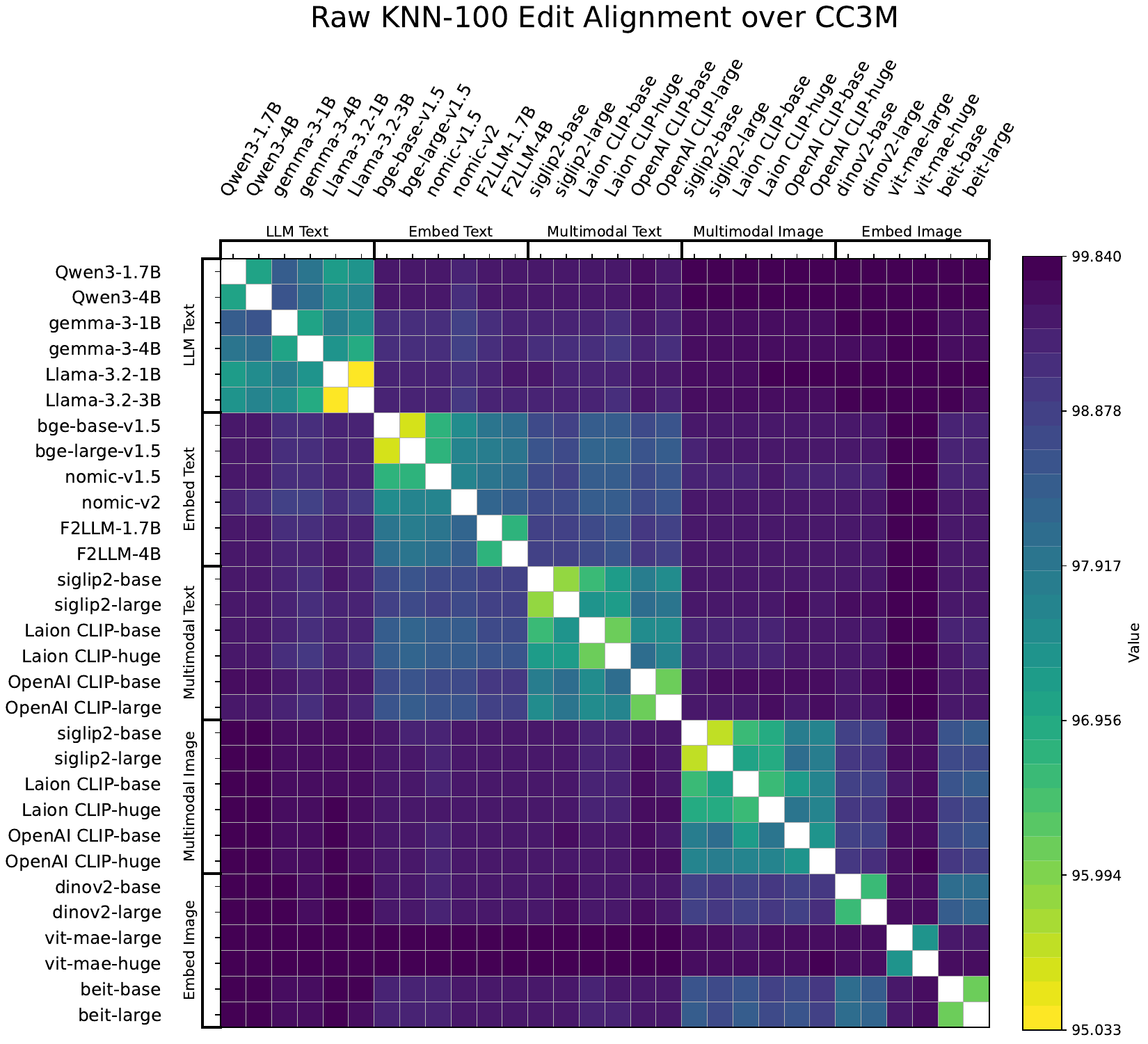}
    \end{minipage}
    \hfill
    \begin{minipage}[t]{0.3\textwidth}
        \centering
        \includegraphics[width = \linewidth]{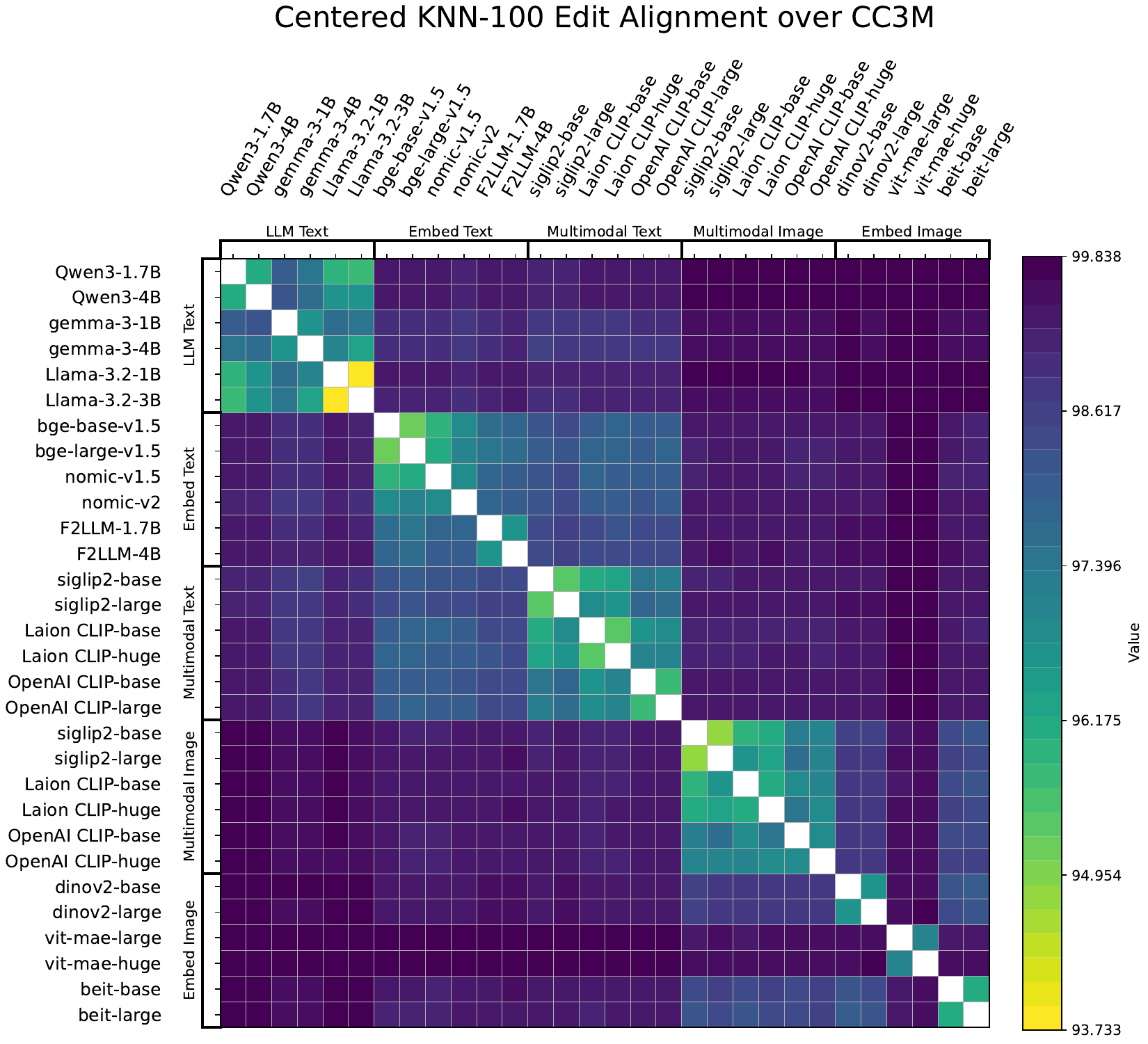}
    \end{minipage}
    \caption{Same plot as in Figure~\ref{fig:biasmain} but over CC3M.}
    \label{fig:biascc3m2}
\end{figure}

\clearpage 

\subsubsection{Experiments on Visual Genome}

\begin{figure}[htbp]
    \centering

    \begin{minipage}[t]{0.3\textwidth}
        \centering
        \includegraphics[width = \linewidth]{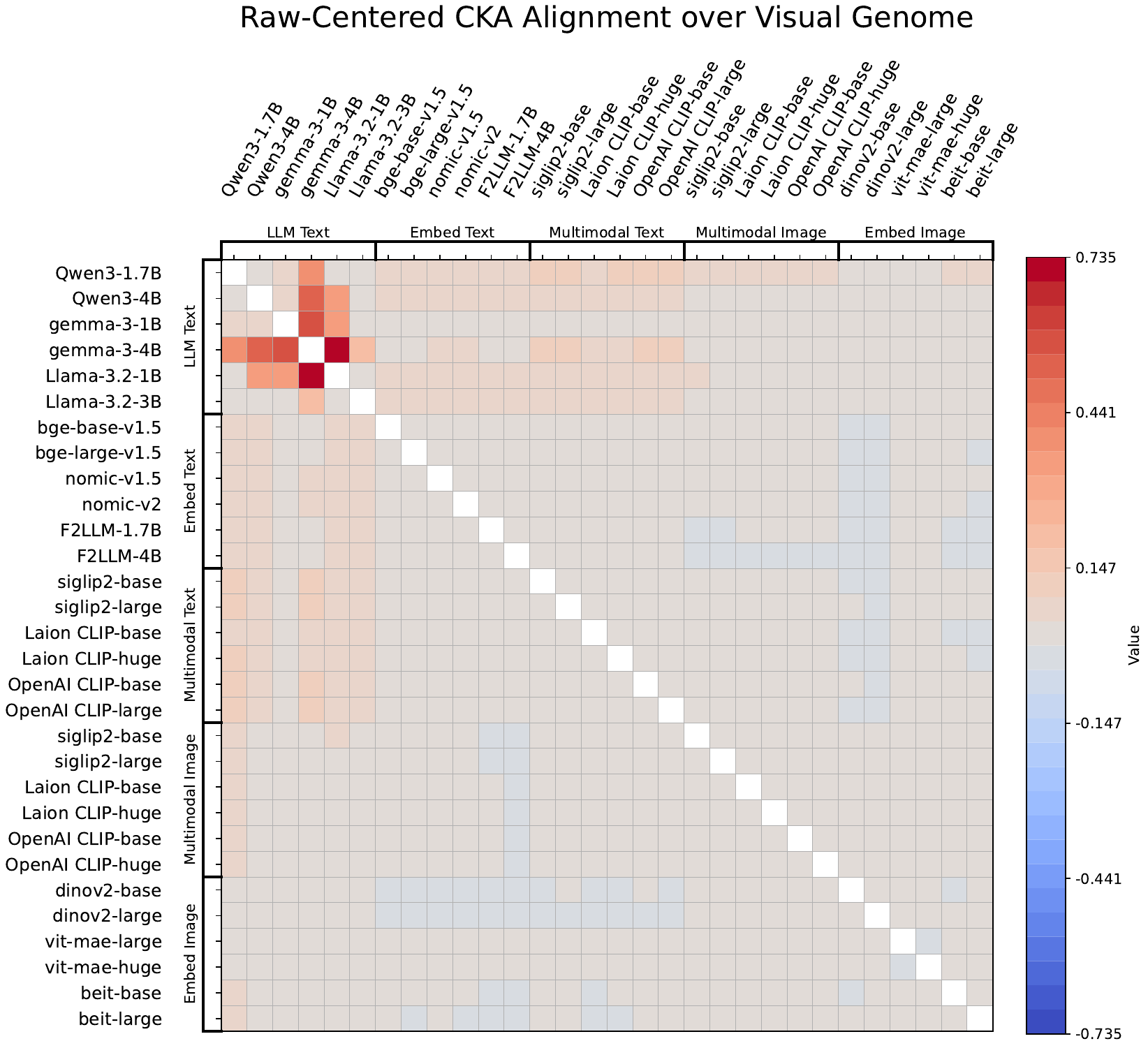}
    \end{minipage}
    \hfill
    \begin{minipage}[t]{0.3\textwidth}
        \centering
        \includegraphics[width = \linewidth]{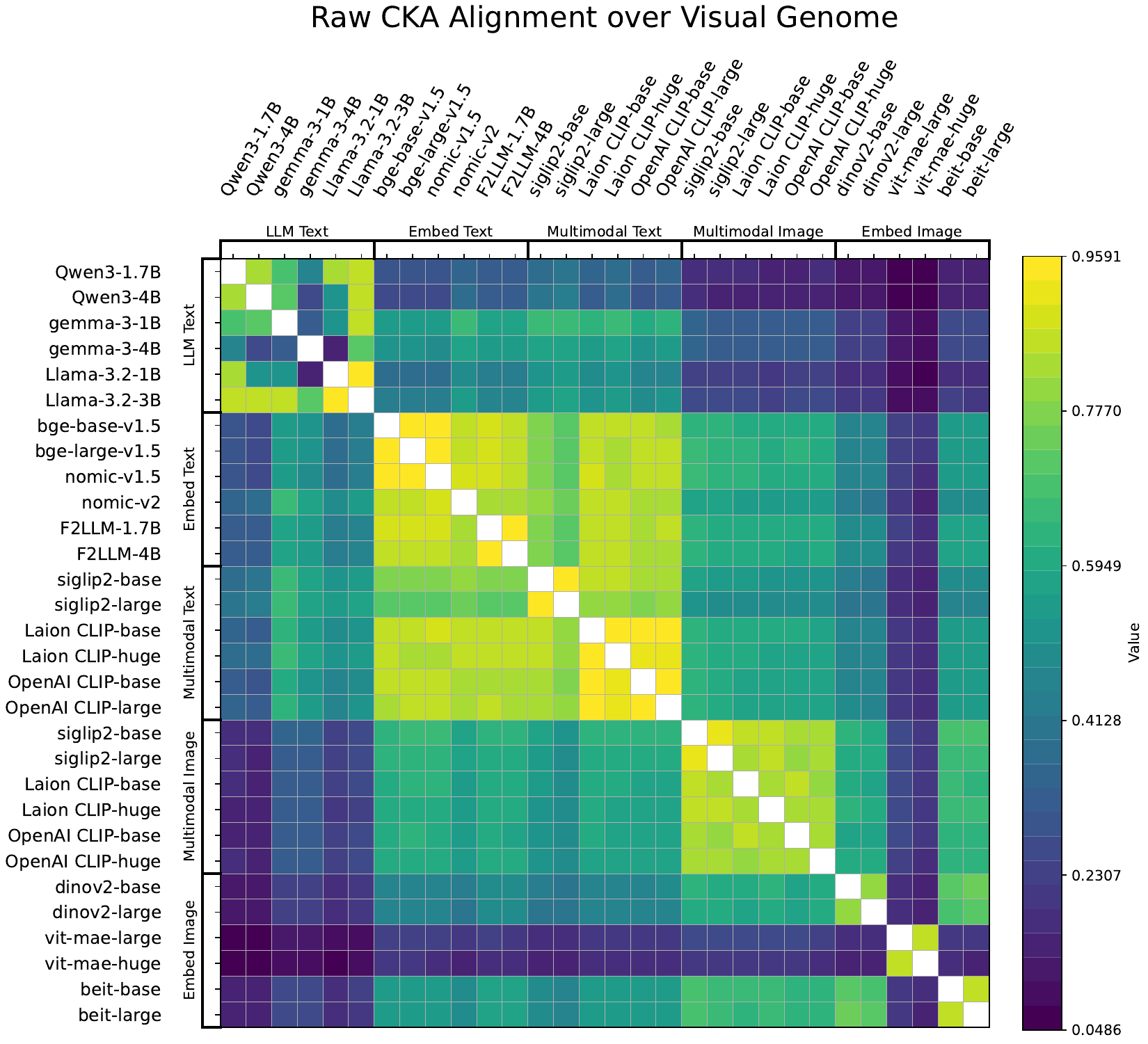}
    \end{minipage}
    \hfill
    \begin{minipage}[t]{0.3\textwidth}
        \centering
        \includegraphics[width = \linewidth]{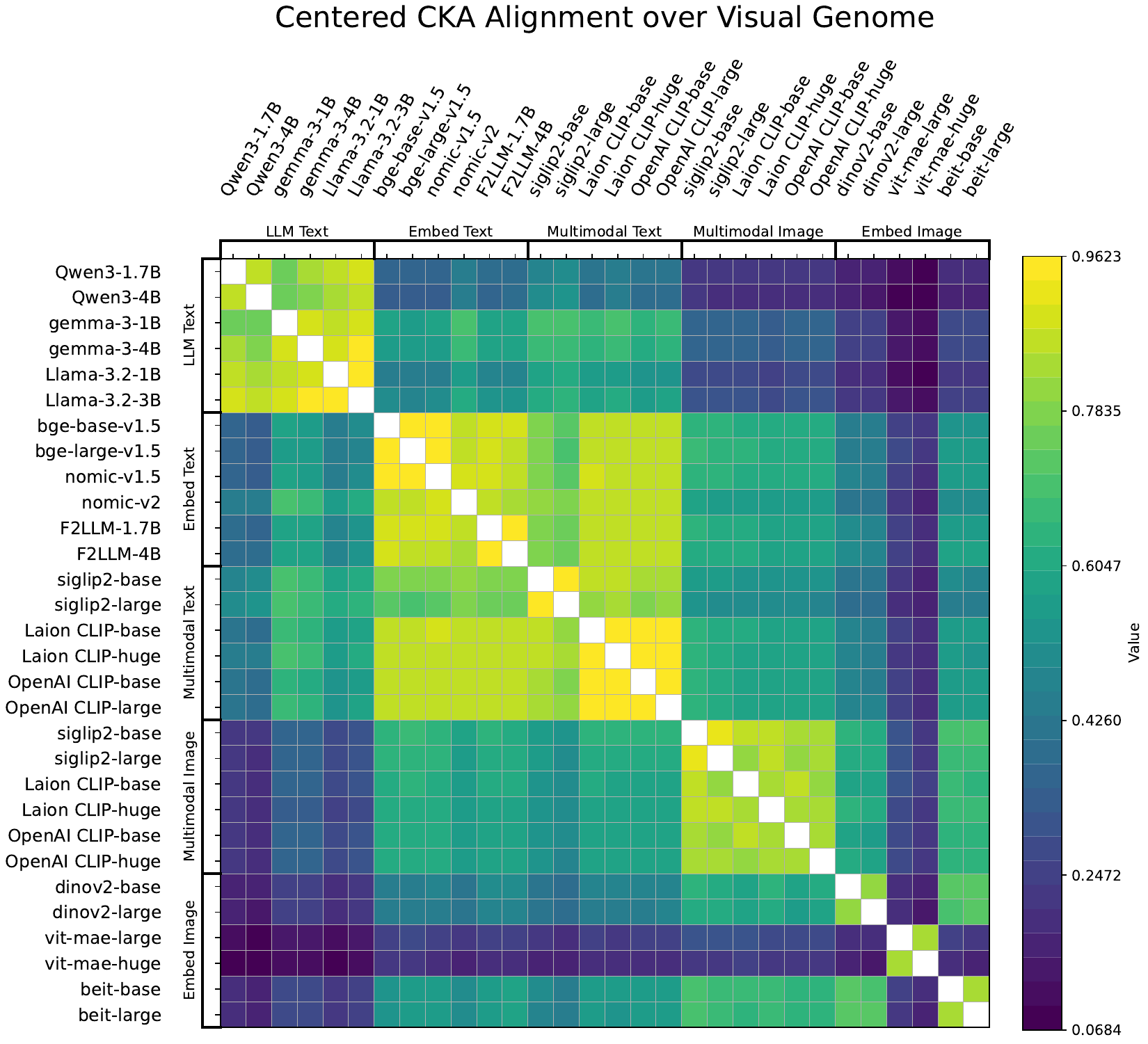}
    \end{minipage}
    
    \vspace{0.4cm}

        \begin{minipage}[t]{0.3\textwidth}
        \centering
        \includegraphics[width = \linewidth]{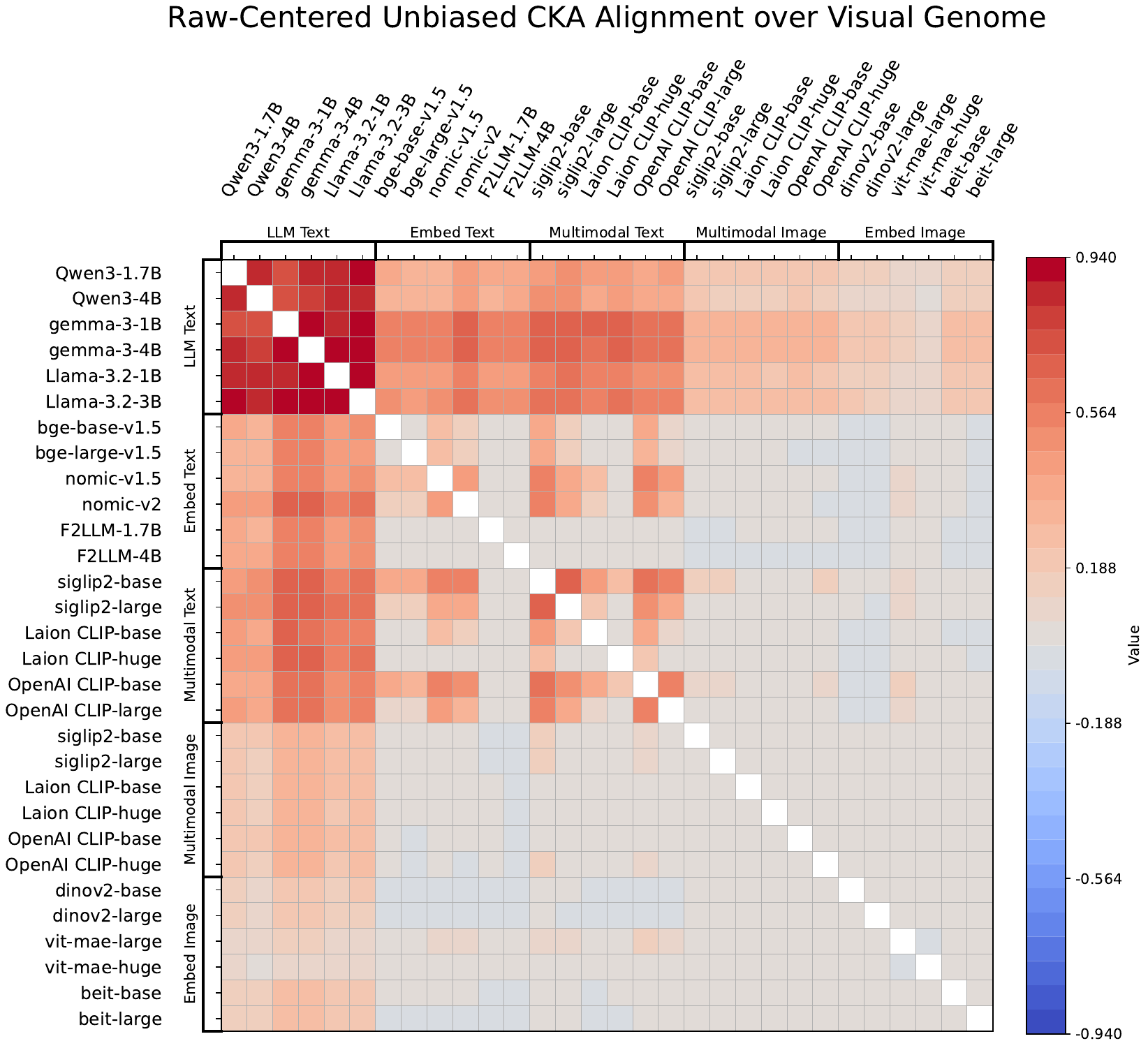}
    \end{minipage}
    \hfill
    \begin{minipage}[t]{0.3\textwidth}
        \centering
        \includegraphics[width = \linewidth]{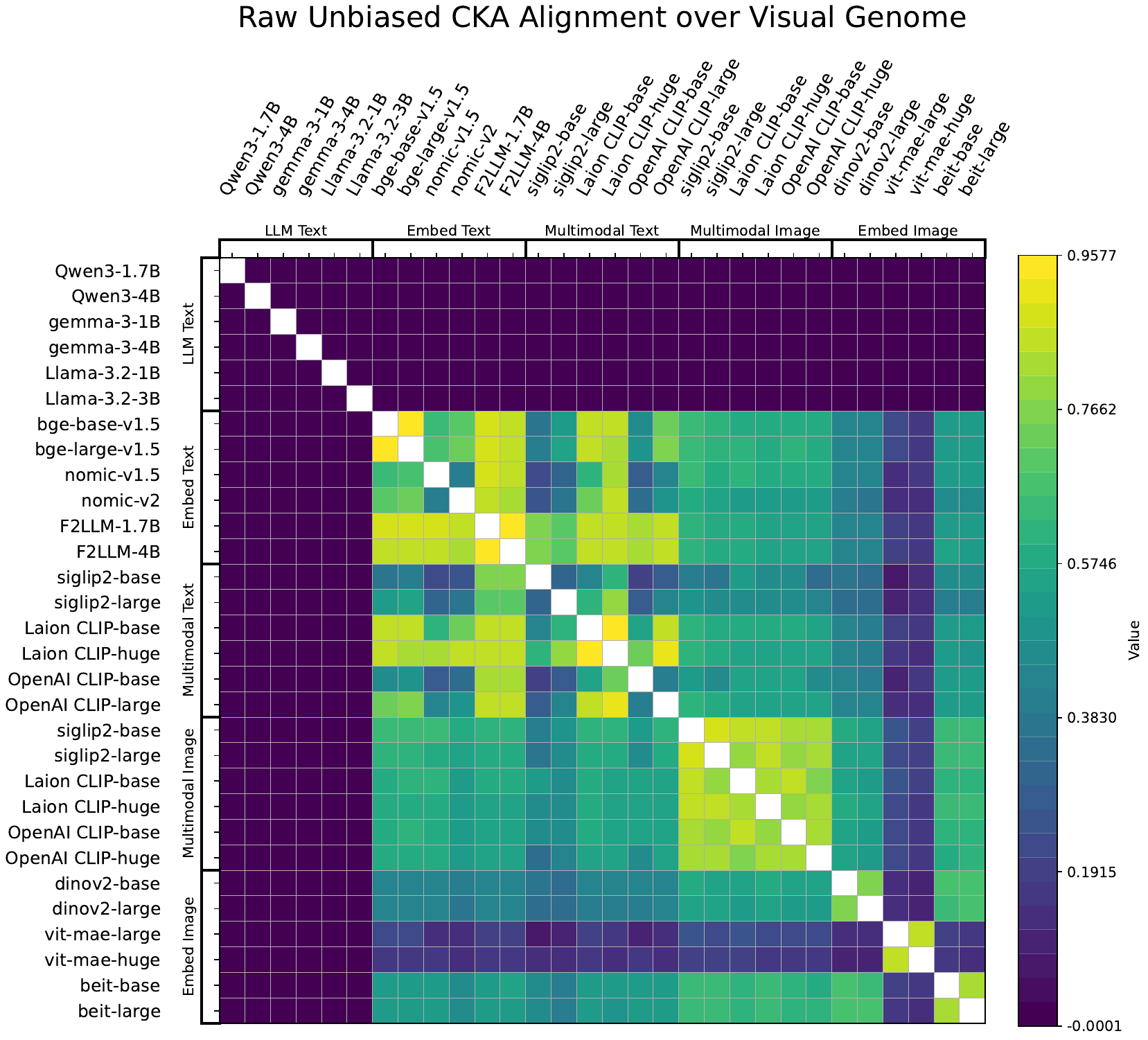}
    \end{minipage}
    \hfill
    \begin{minipage}[t]{0.3\textwidth}
        \centering
        \includegraphics[width = \linewidth]{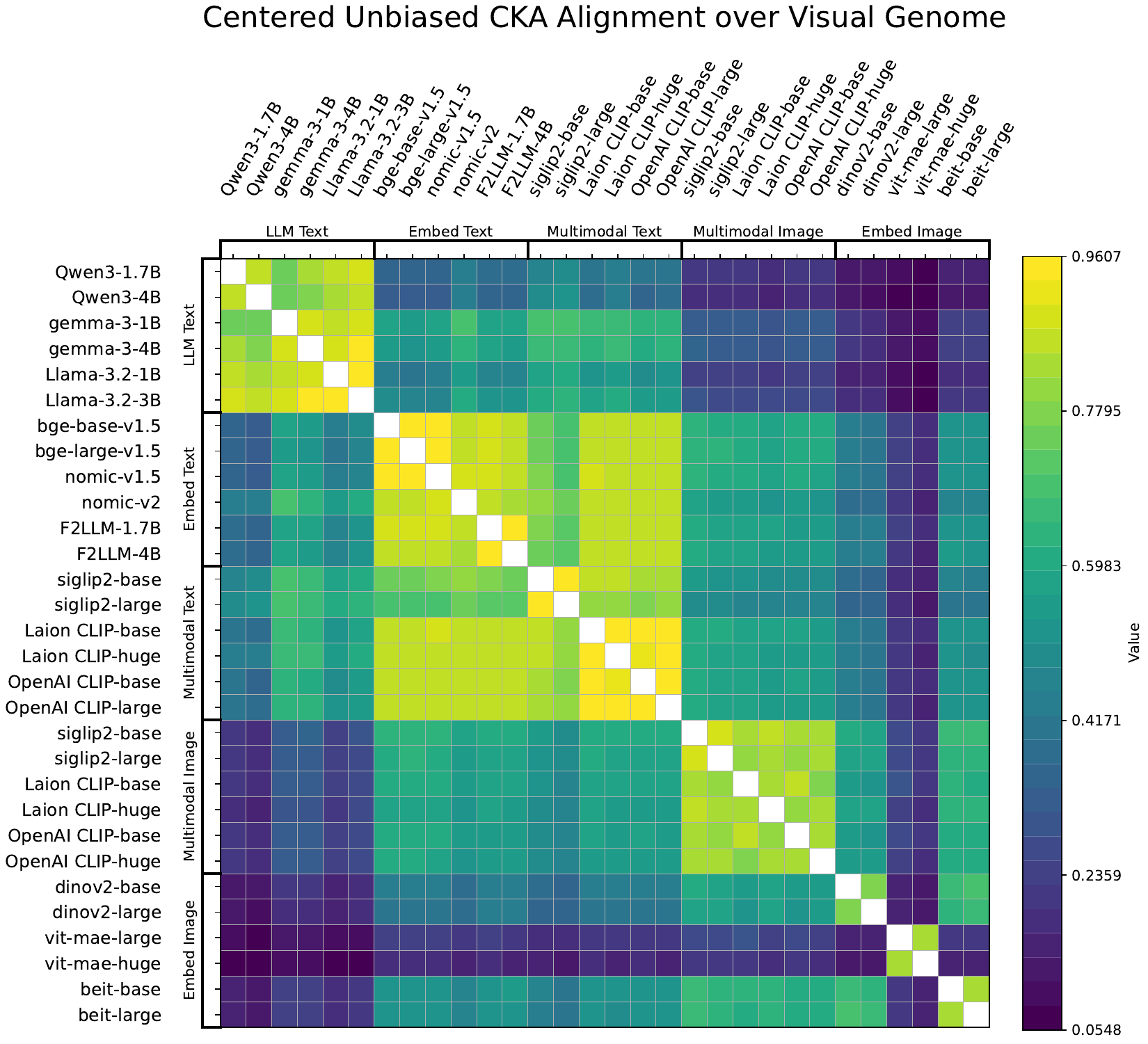}
    \end{minipage}

        \vspace{0.4cm}

        \begin{minipage}[t]{0.3\textwidth}
        \centering
        \includegraphics[width = \linewidth]{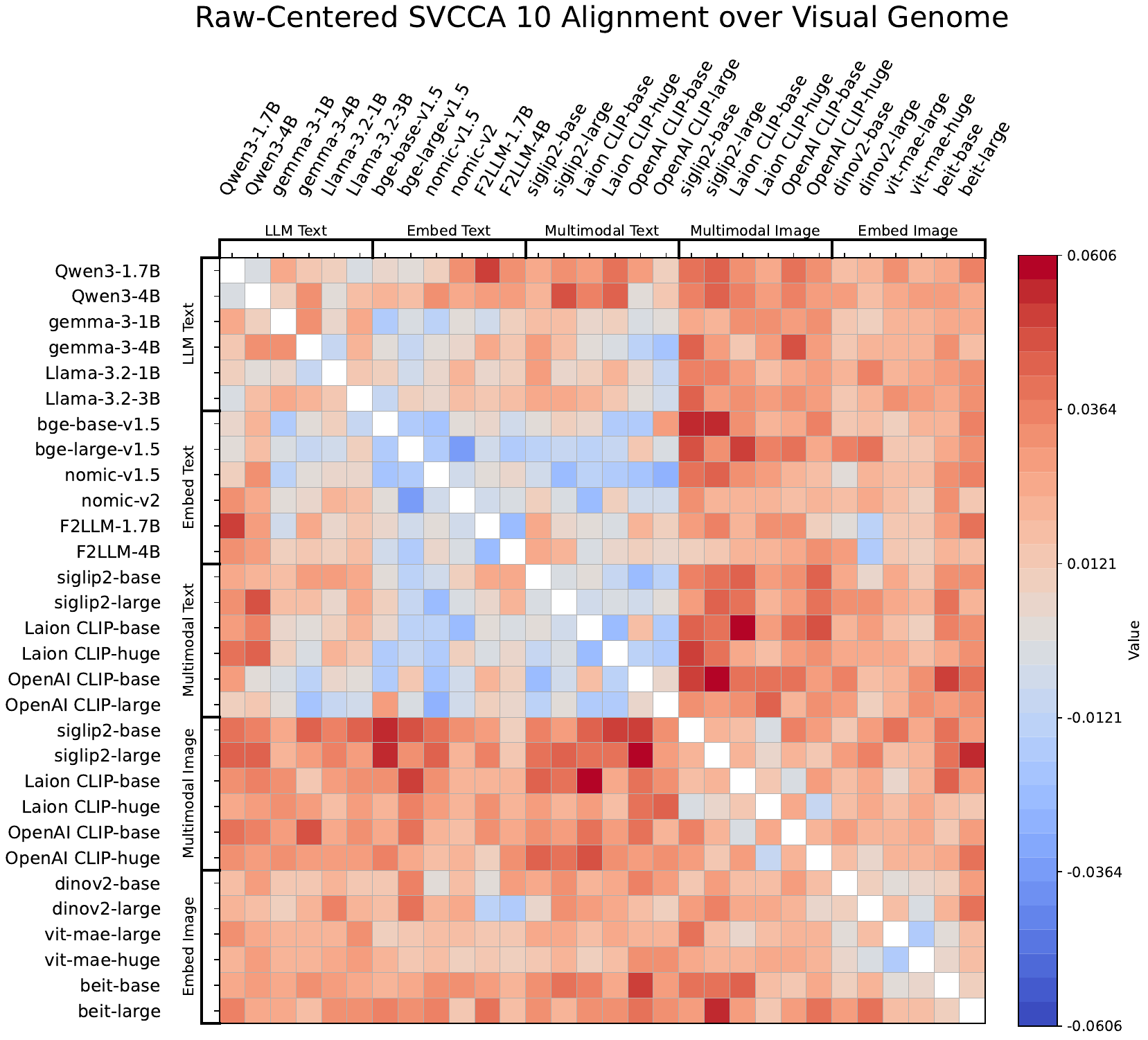}
    \end{minipage}
    \hfill
    \begin{minipage}[t]{0.3\textwidth}
        \centering
        \includegraphics[width = \linewidth]{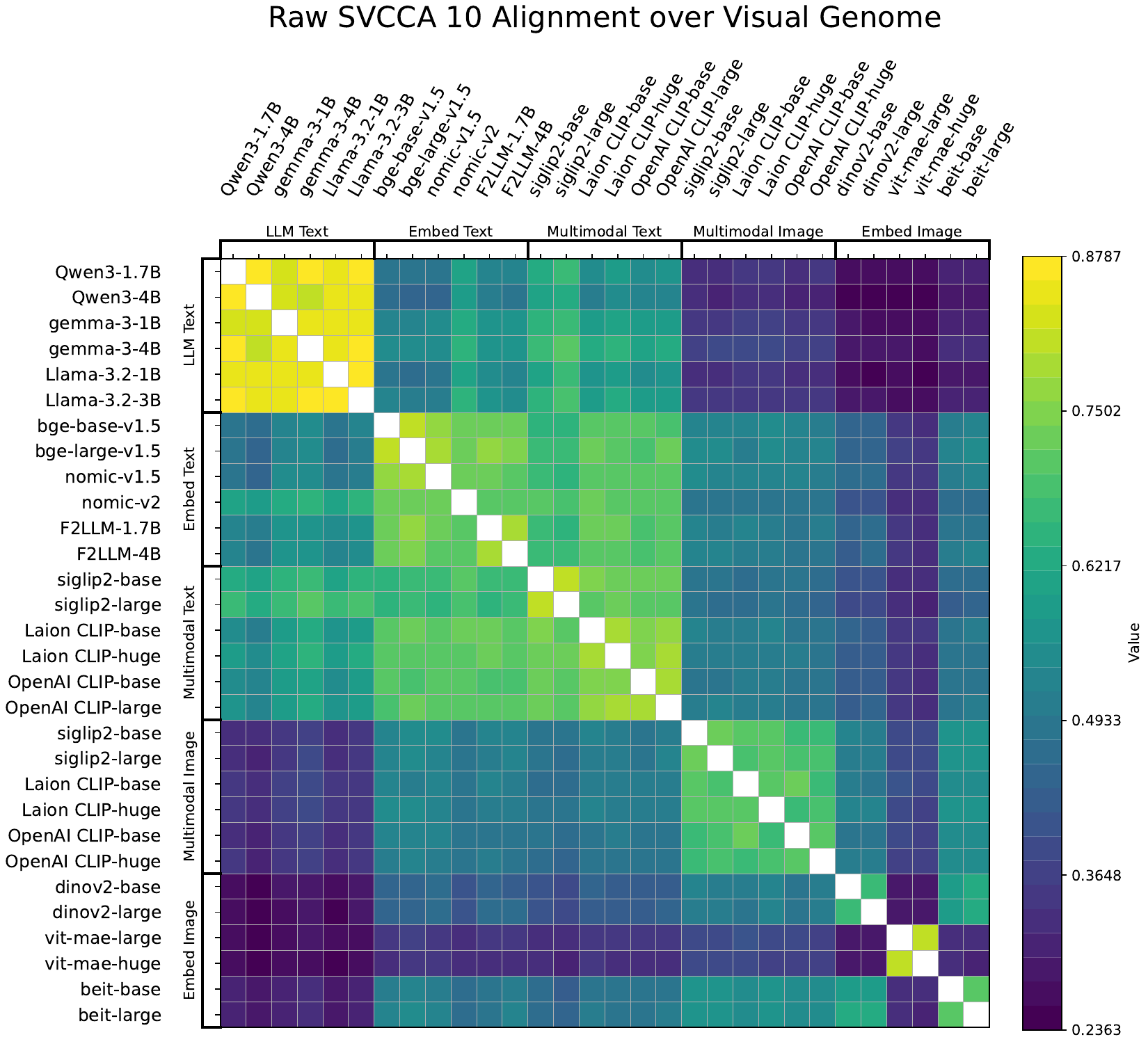}
    \end{minipage}
    \hfill
    \begin{minipage}[t]{0.3\textwidth}
        \centering
        \includegraphics[width = \linewidth]{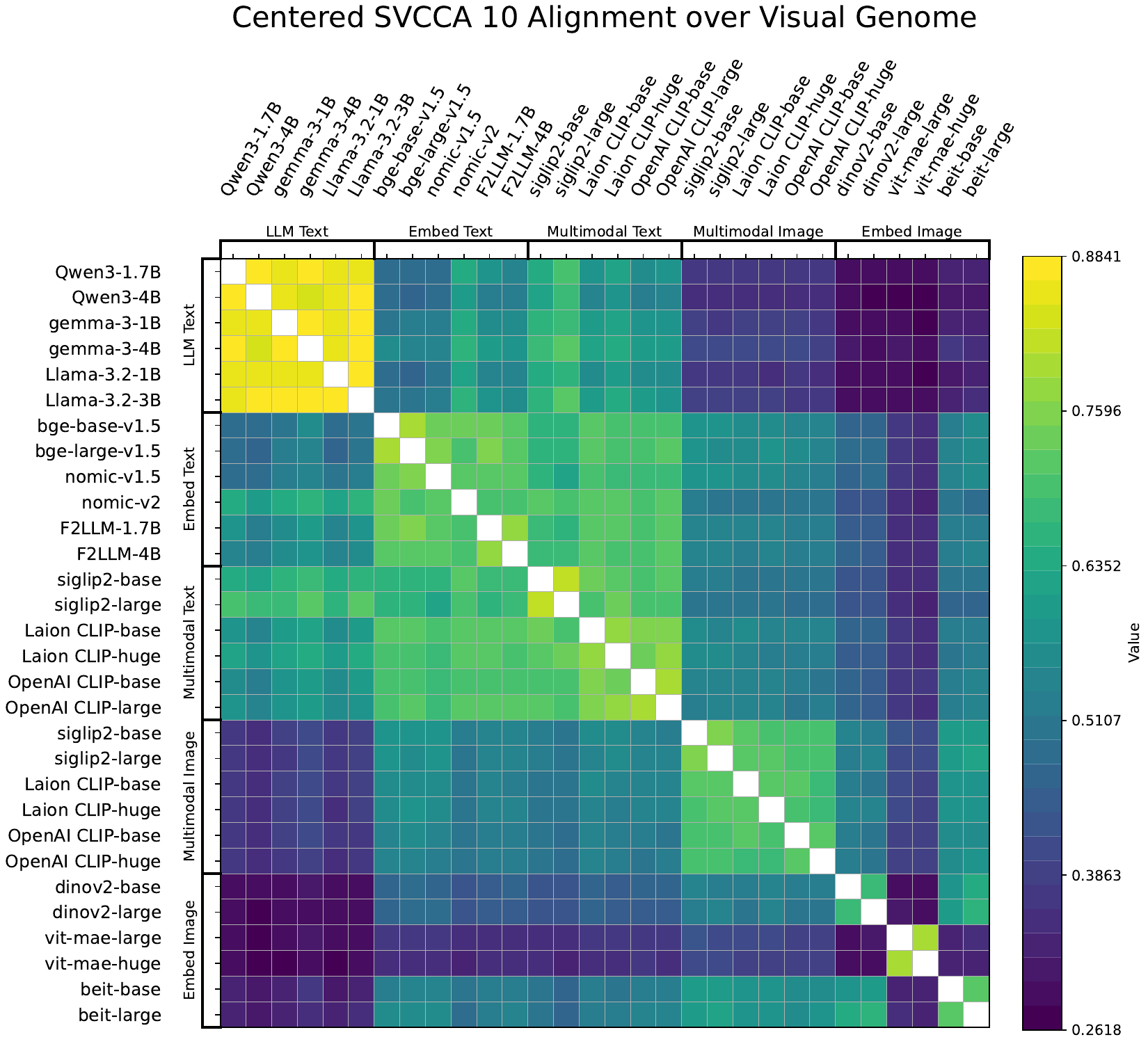}
    \end{minipage}

            \vspace{0.4cm}

    \begin{minipage}[t]{0.3\textwidth}
        \centering
        \includegraphics[width = \linewidth]{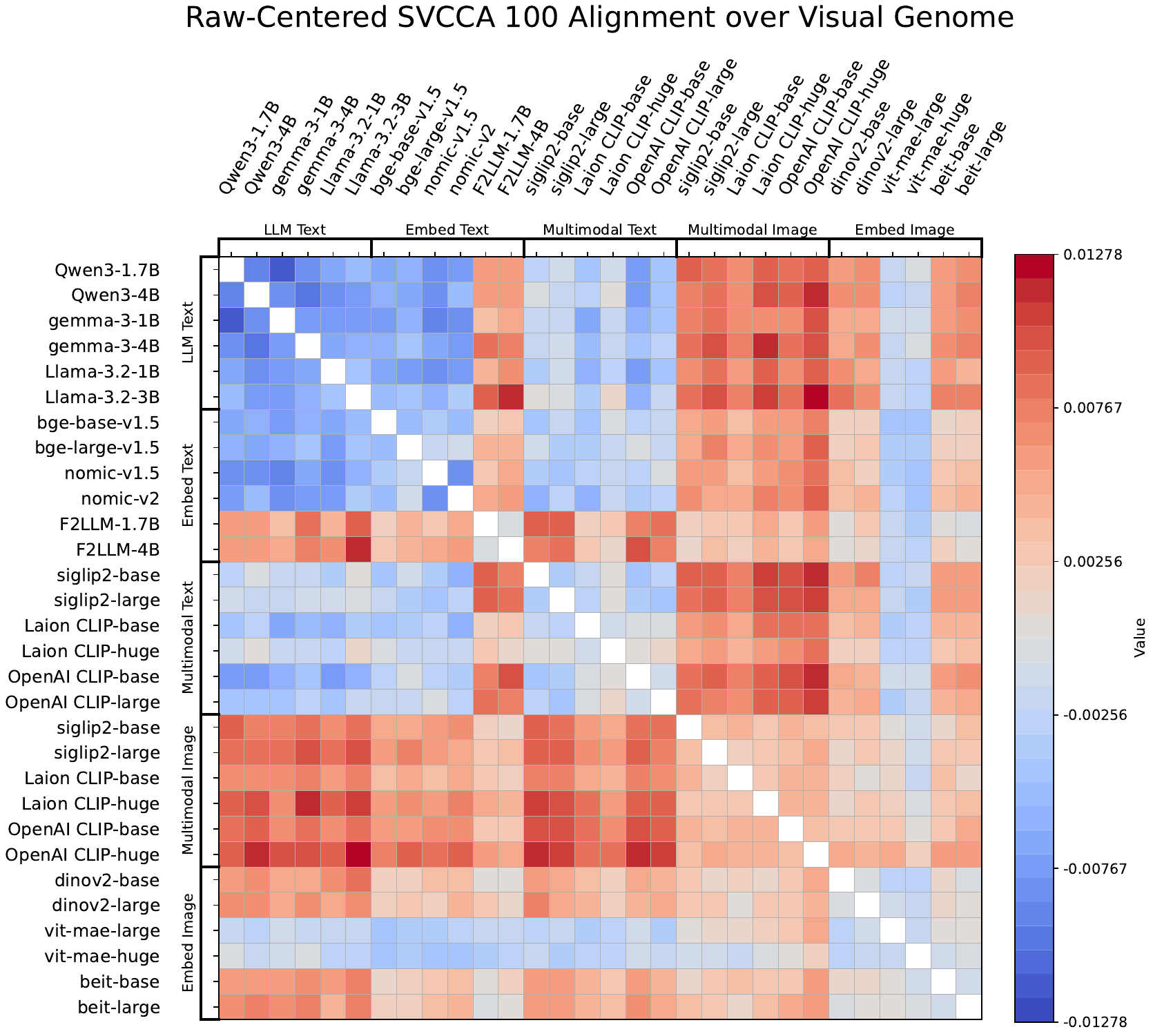}
    \end{minipage}
    \hfill
    \begin{minipage}[t]{0.3\textwidth}
        \centering
        \includegraphics[width = \linewidth]{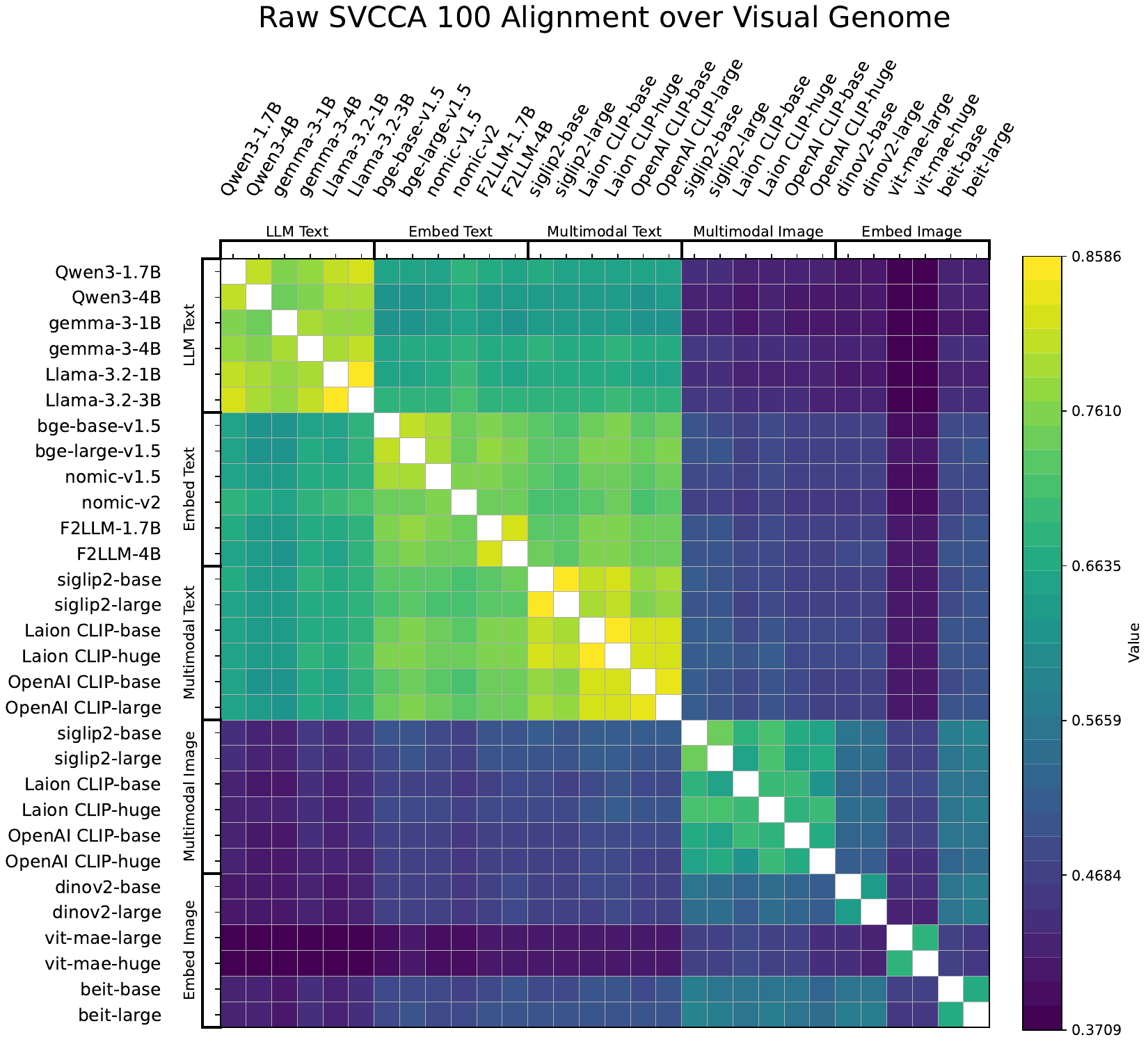}
    \end{minipage}
    \hfill
    \begin{minipage}[t]{0.3\textwidth}
        \centering
        \includegraphics[width = \linewidth]{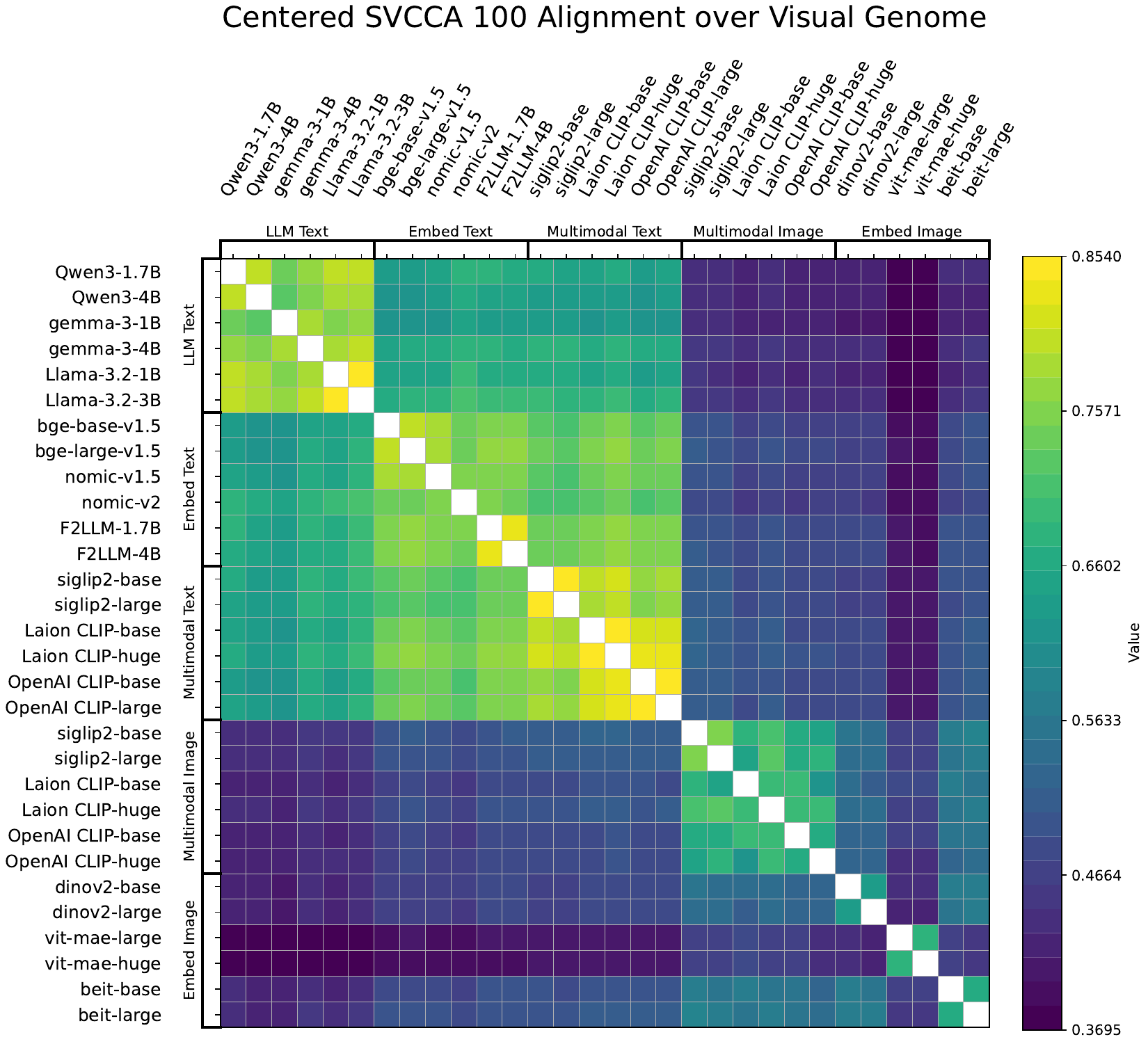}
    \end{minipage}
    \caption{Same plot as in Figure~\ref{fig:biascoco1} but over Visual Genome.}
    \label{fig:biasvisual_genome1}
\end{figure}

\clearpage

\begin{figure}[htbp]
    \centering

    \begin{minipage}[t]{0.3\textwidth}
        \centering
        \includegraphics[width = \linewidth]{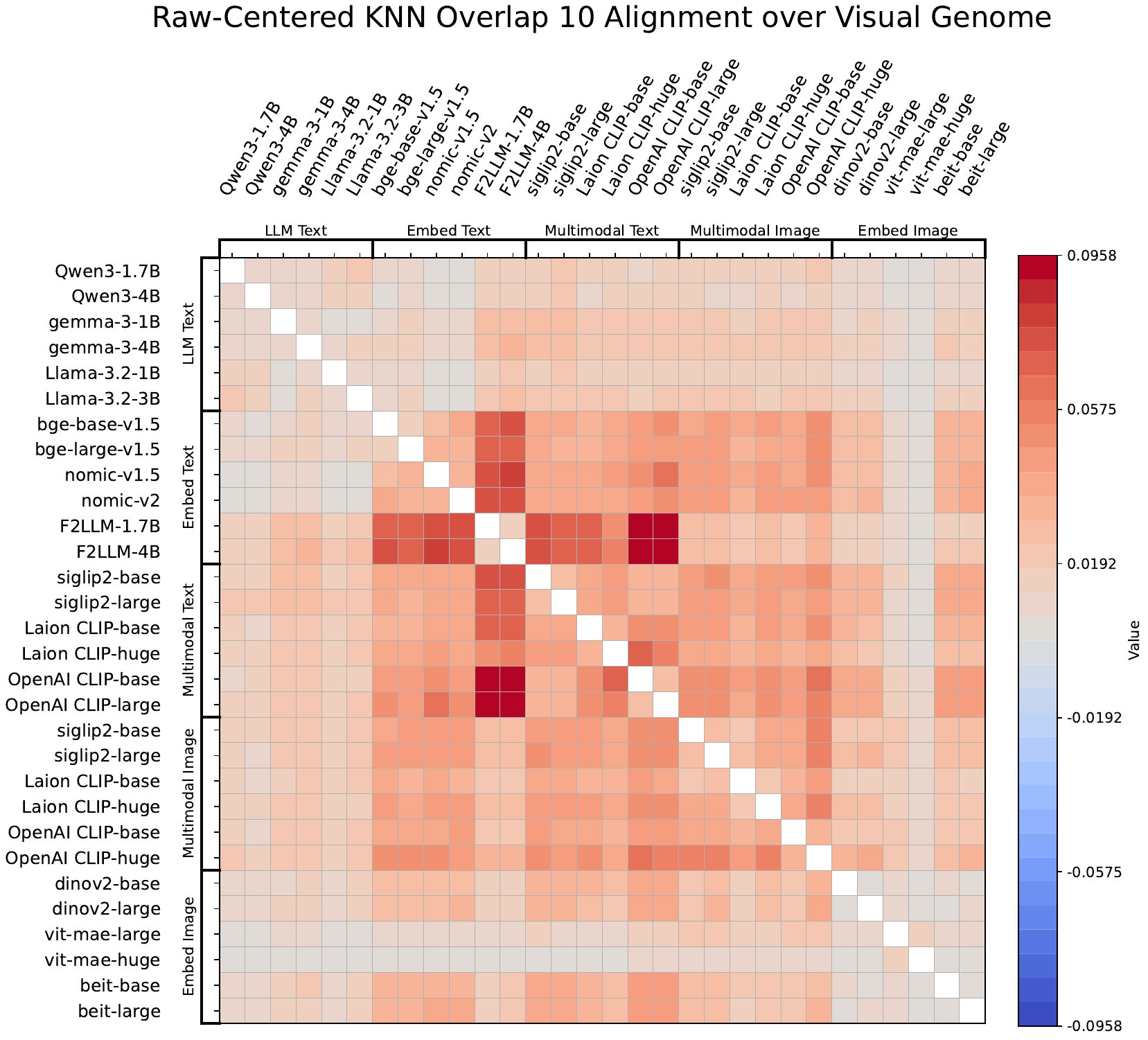}
    \end{minipage}
    \hfill
    \begin{minipage}[t]{0.3\textwidth}
        \centering
        \includegraphics[width = \linewidth]{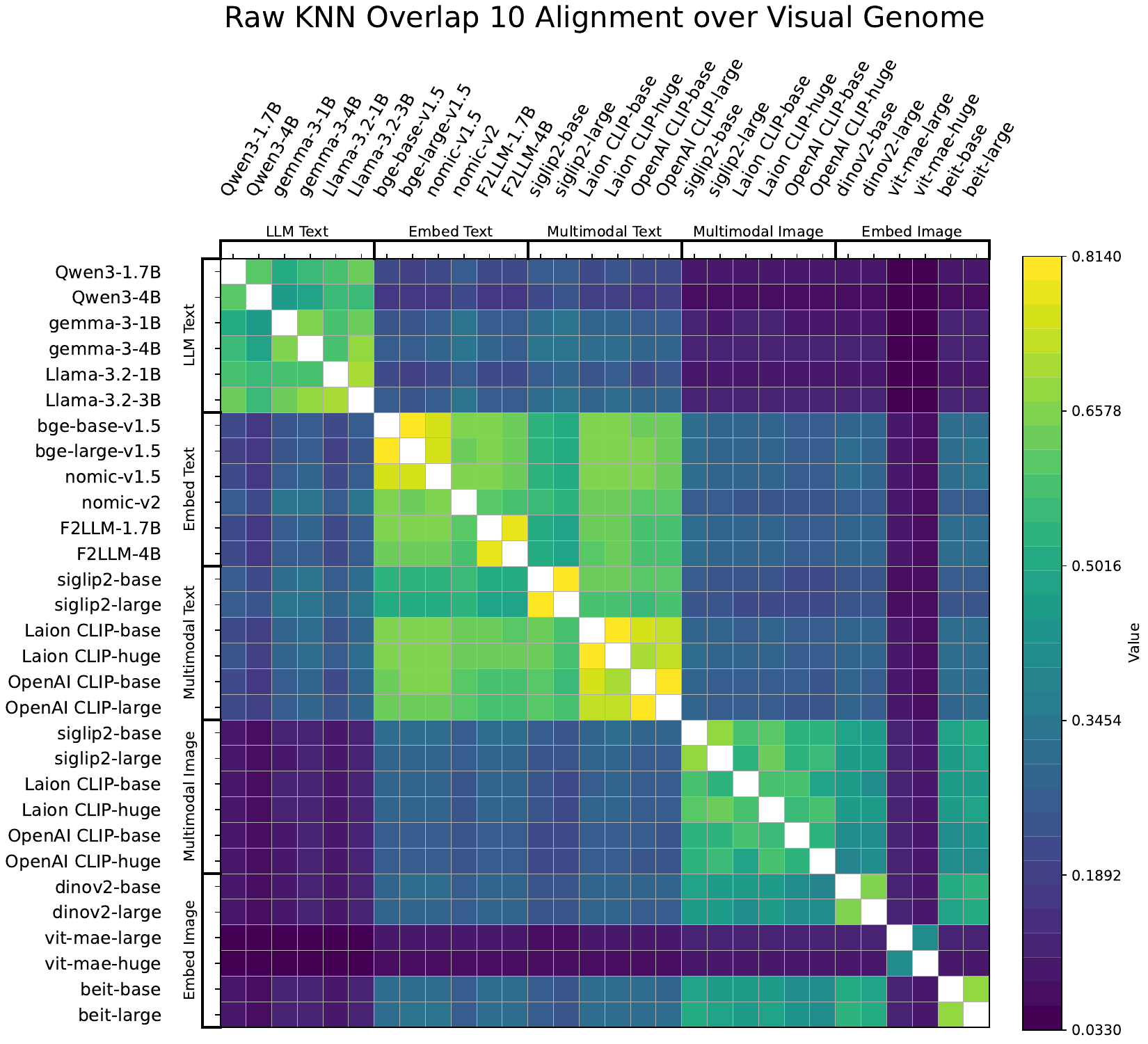}
    \end{minipage}
    \hfill
    \begin{minipage}[t]{0.3\textwidth}
        \centering
        \includegraphics[width = \linewidth]{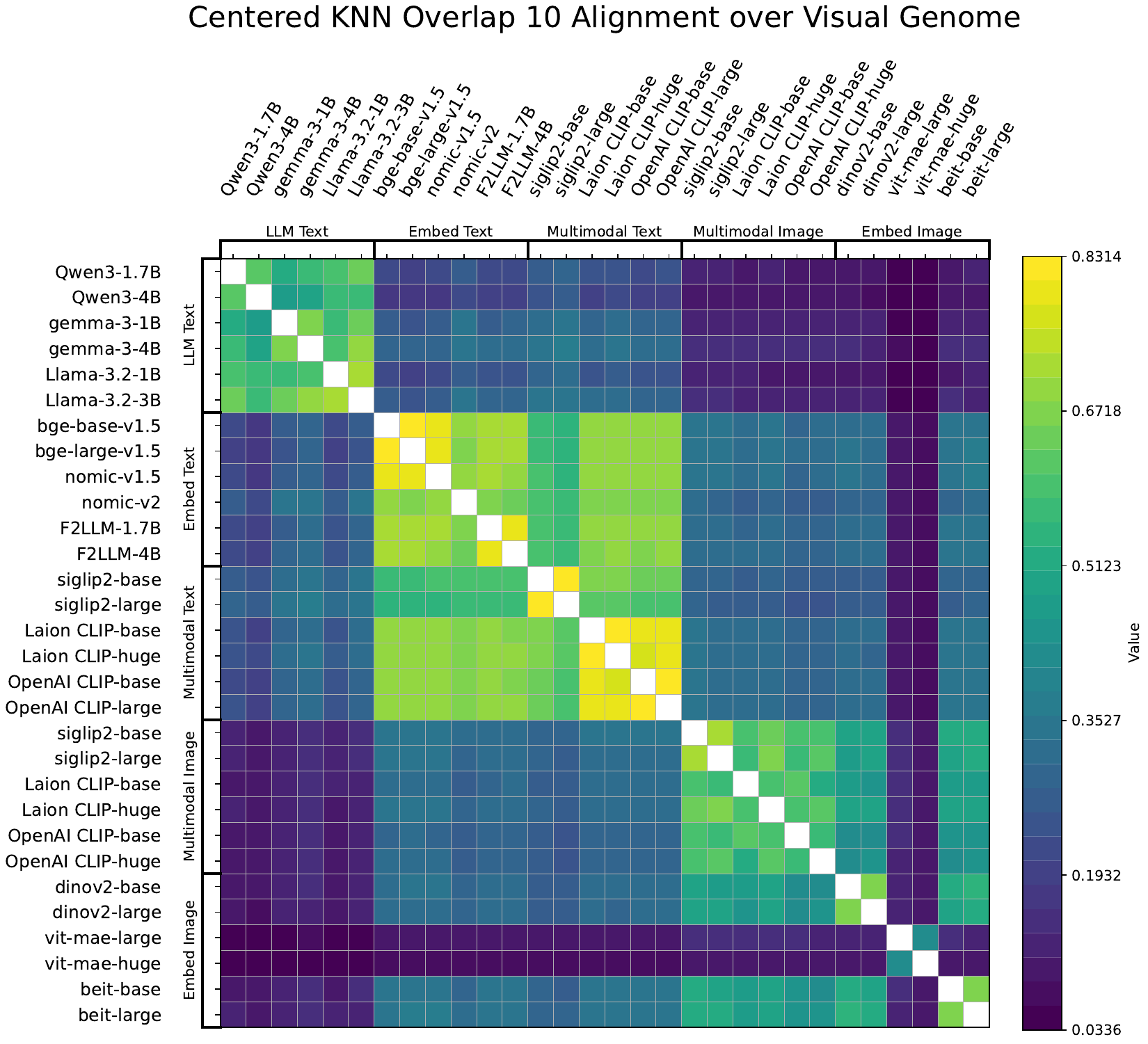}
    \end{minipage}
    
    \vspace{0.4cm}

        \begin{minipage}[t]{0.3\textwidth}
        \centering
        \includegraphics[width = \linewidth]{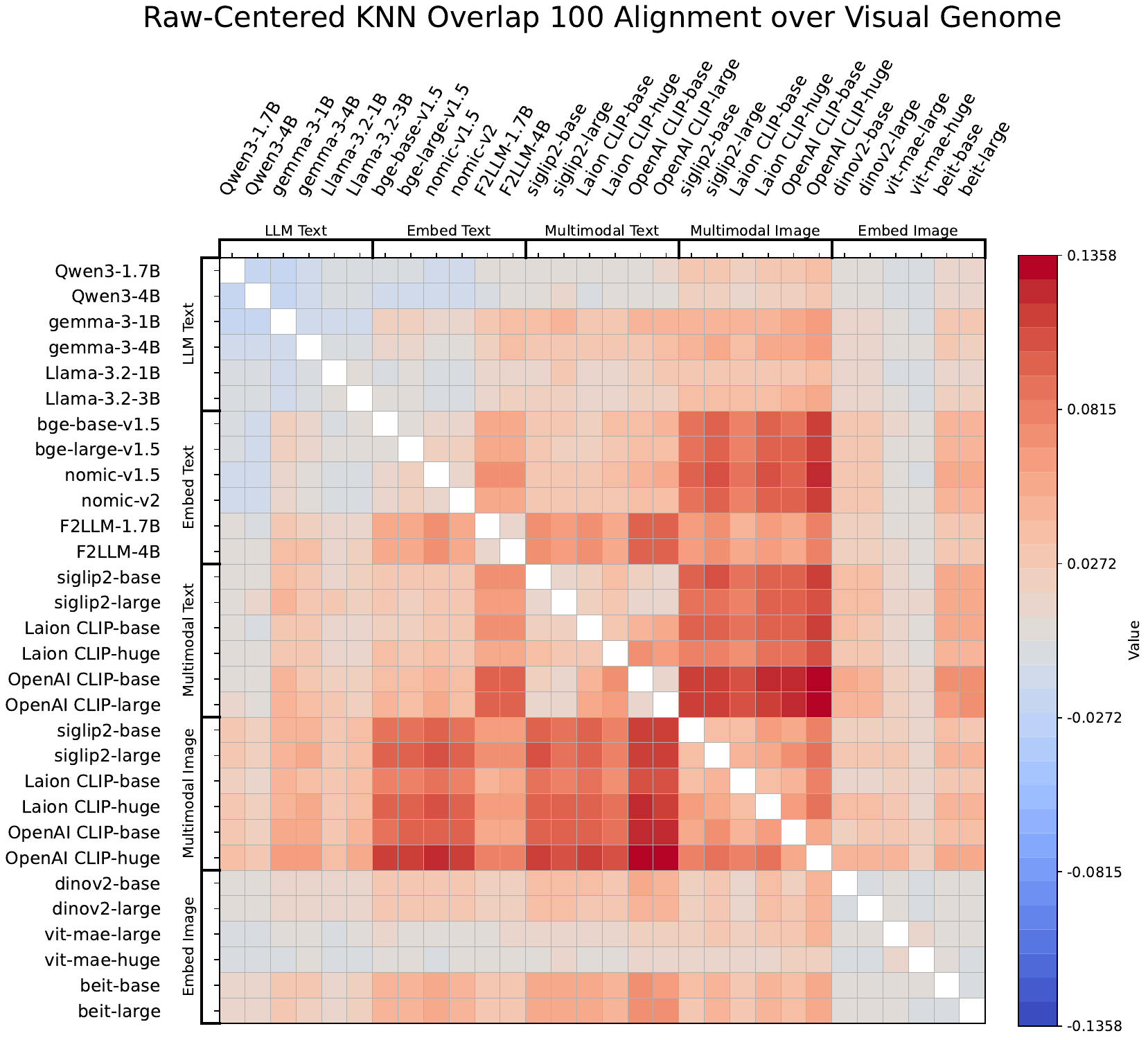}
    \end{minipage}
    \hfill
    \begin{minipage}[t]{0.3\textwidth}
        \centering
        \includegraphics[width = \linewidth]{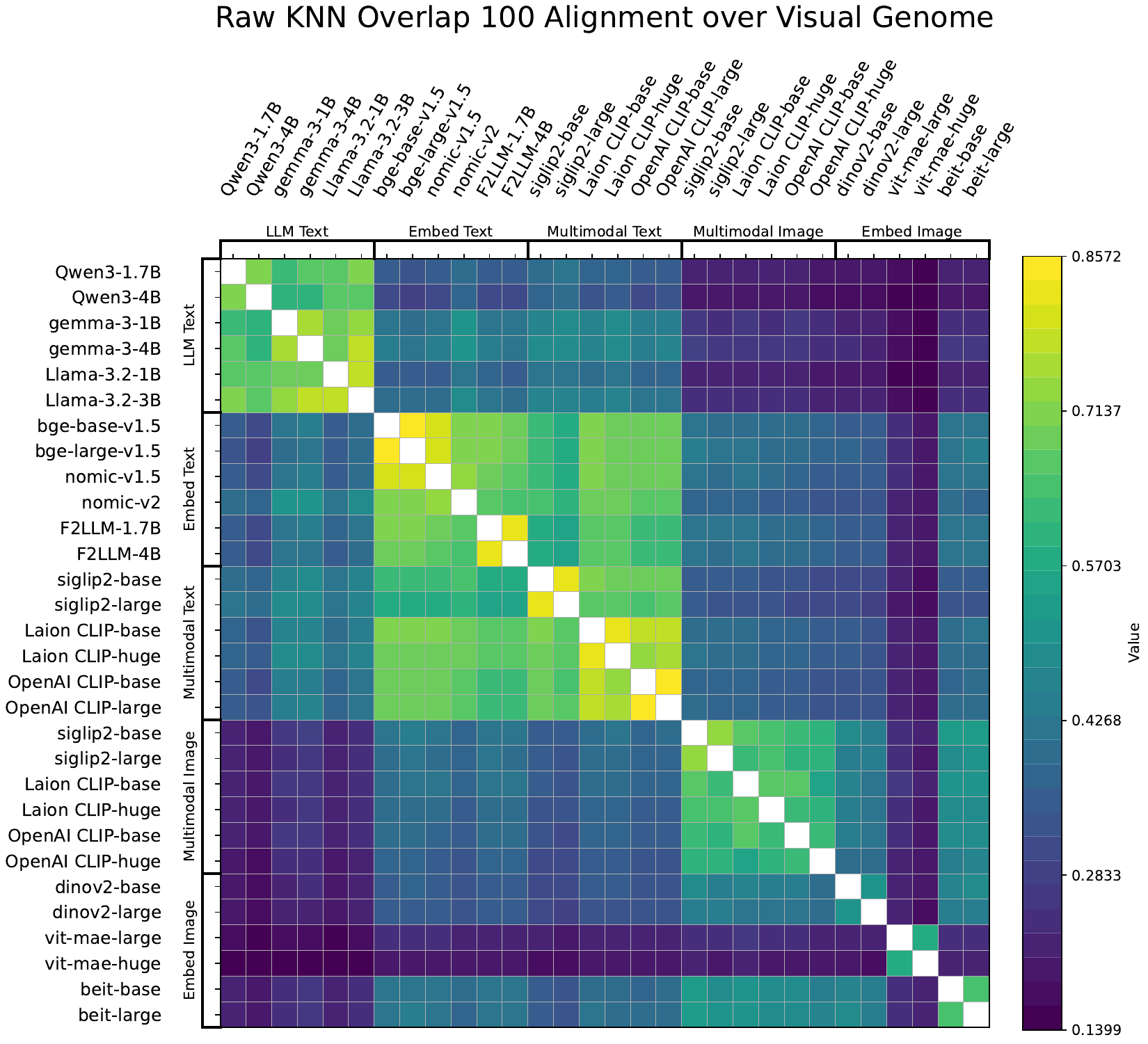}
    \end{minipage}
    \hfill
    \begin{minipage}[t]{0.3\textwidth}
        \centering
        \includegraphics[width = \linewidth]{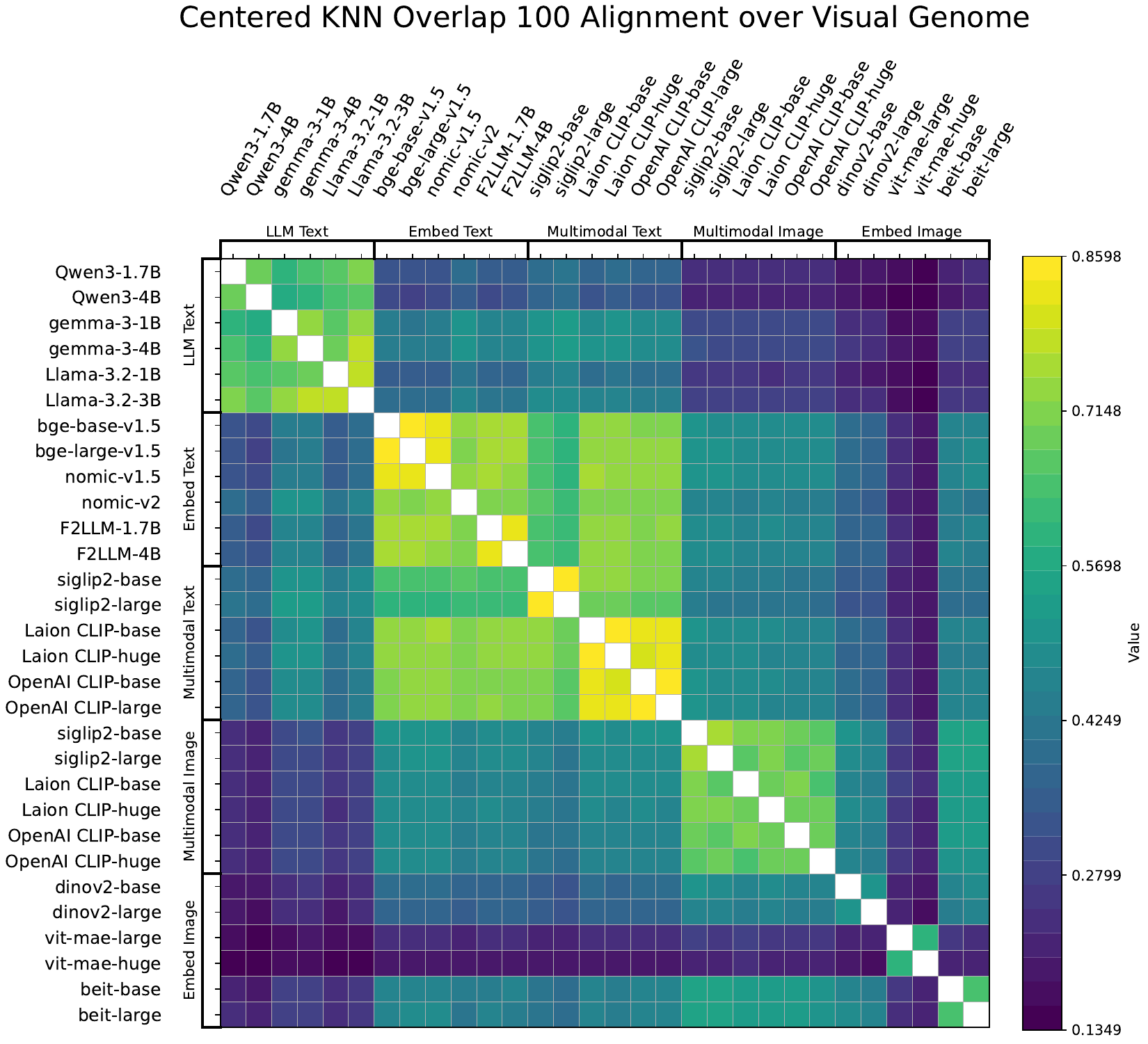}
    \end{minipage}

        \vspace{0.4cm}

        \begin{minipage}[t]{0.3\textwidth}
        \centering
        \includegraphics[width = \linewidth]{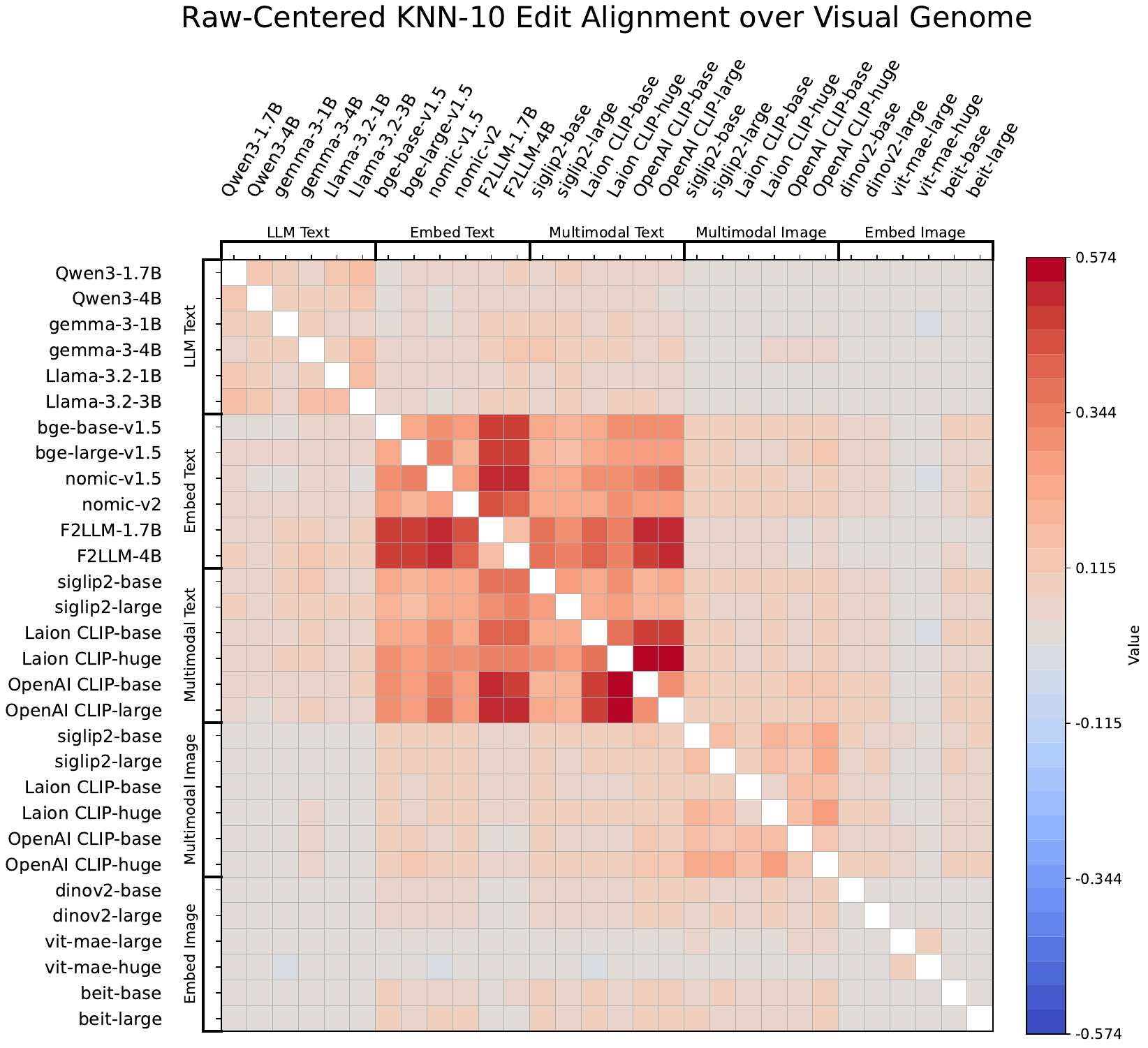}
    \end{minipage}
    \hfill
    \begin{minipage}[t]{0.3\textwidth}
        \centering
        \includegraphics[width = \linewidth]{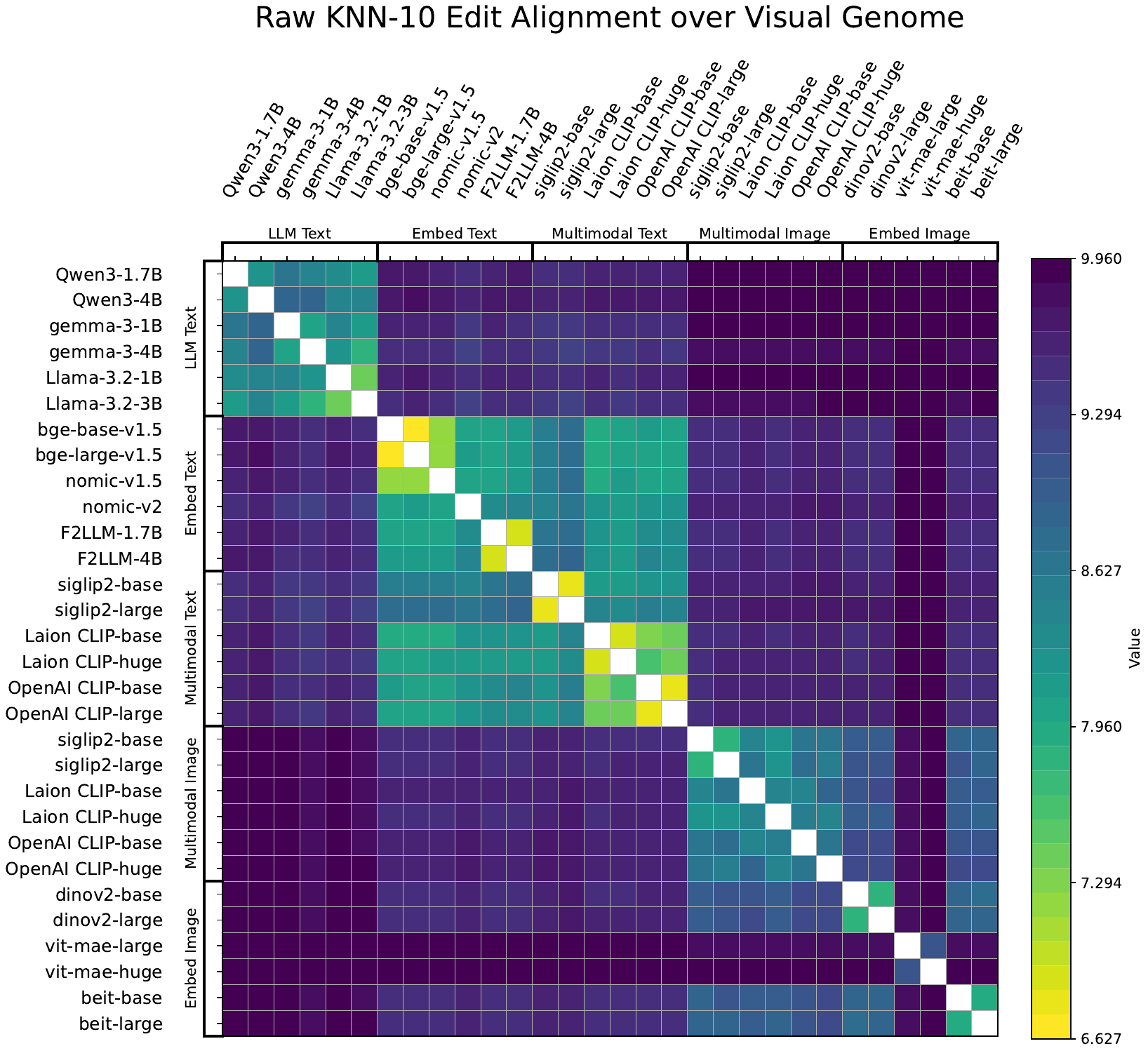}
    \end{minipage}
    \hfill
    \begin{minipage}[t]{0.3\textwidth}
        \centering
        \includegraphics[width = \linewidth]{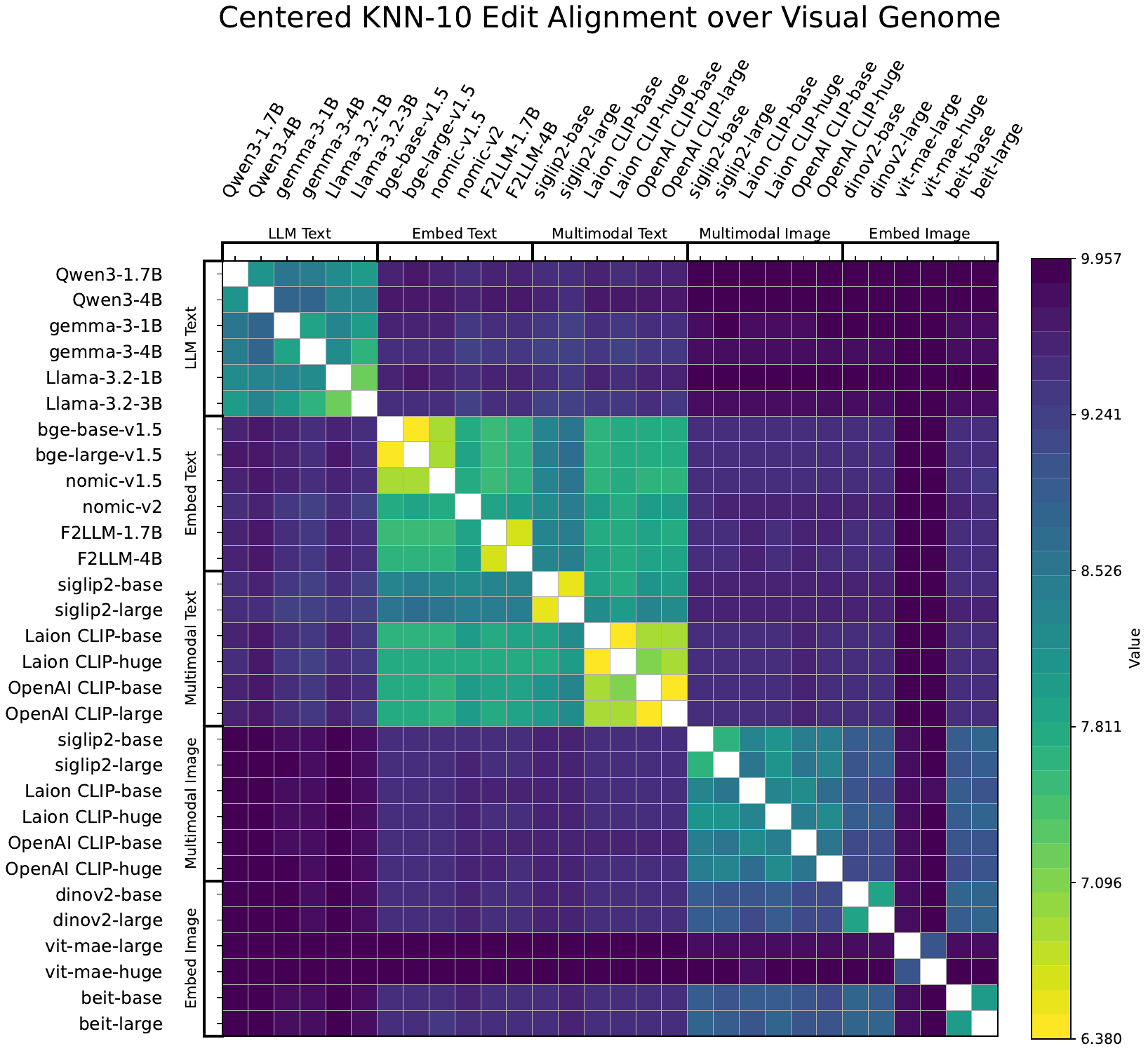}
    \end{minipage}

            \vspace{0.4cm}

    \begin{minipage}[t]{0.3\textwidth}
        \centering
        \includegraphics[width = \linewidth]{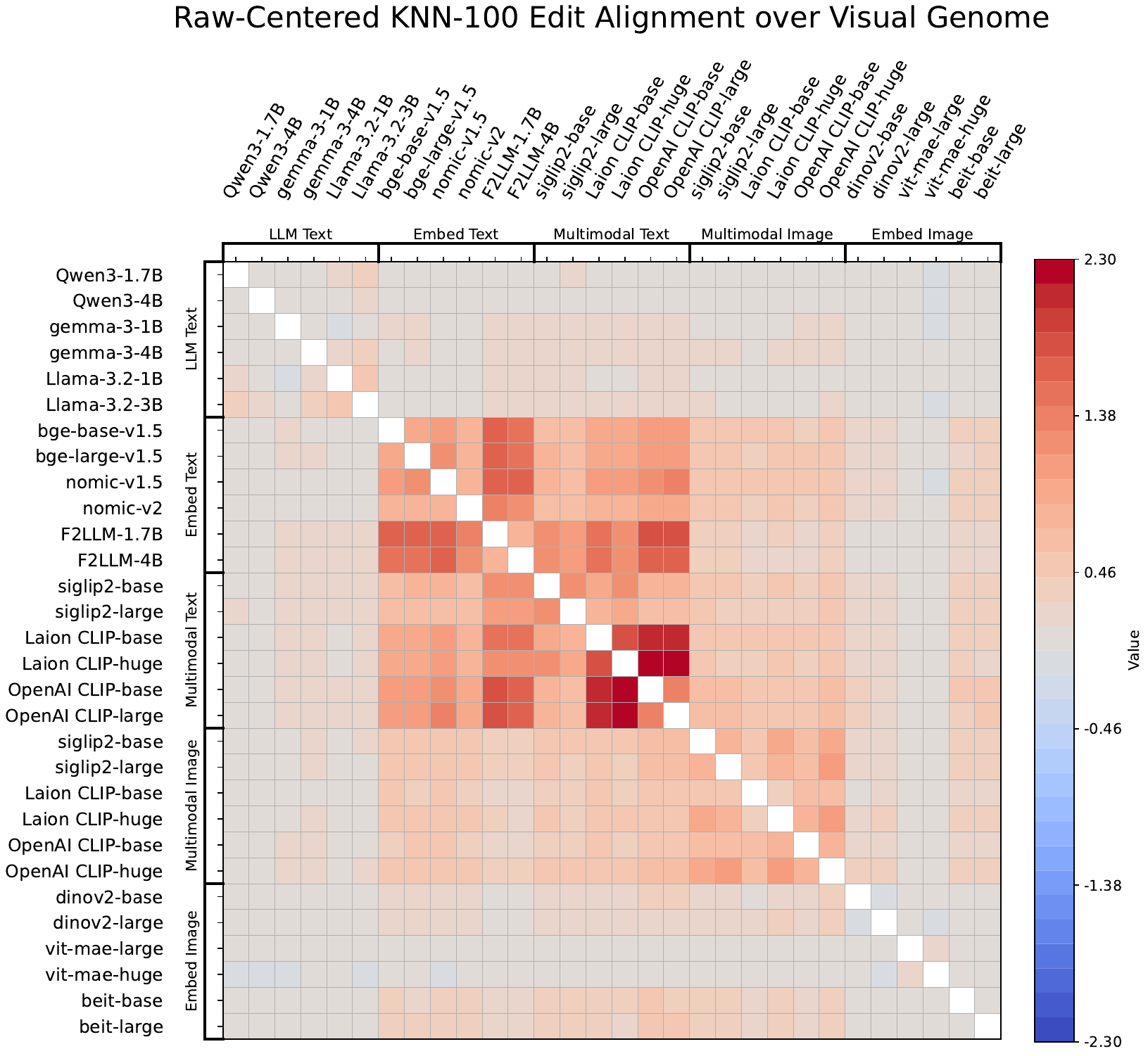}
    \end{minipage}
    \hfill
    \begin{minipage}[t]{0.3\textwidth}
        \centering
        \includegraphics[width = \linewidth]{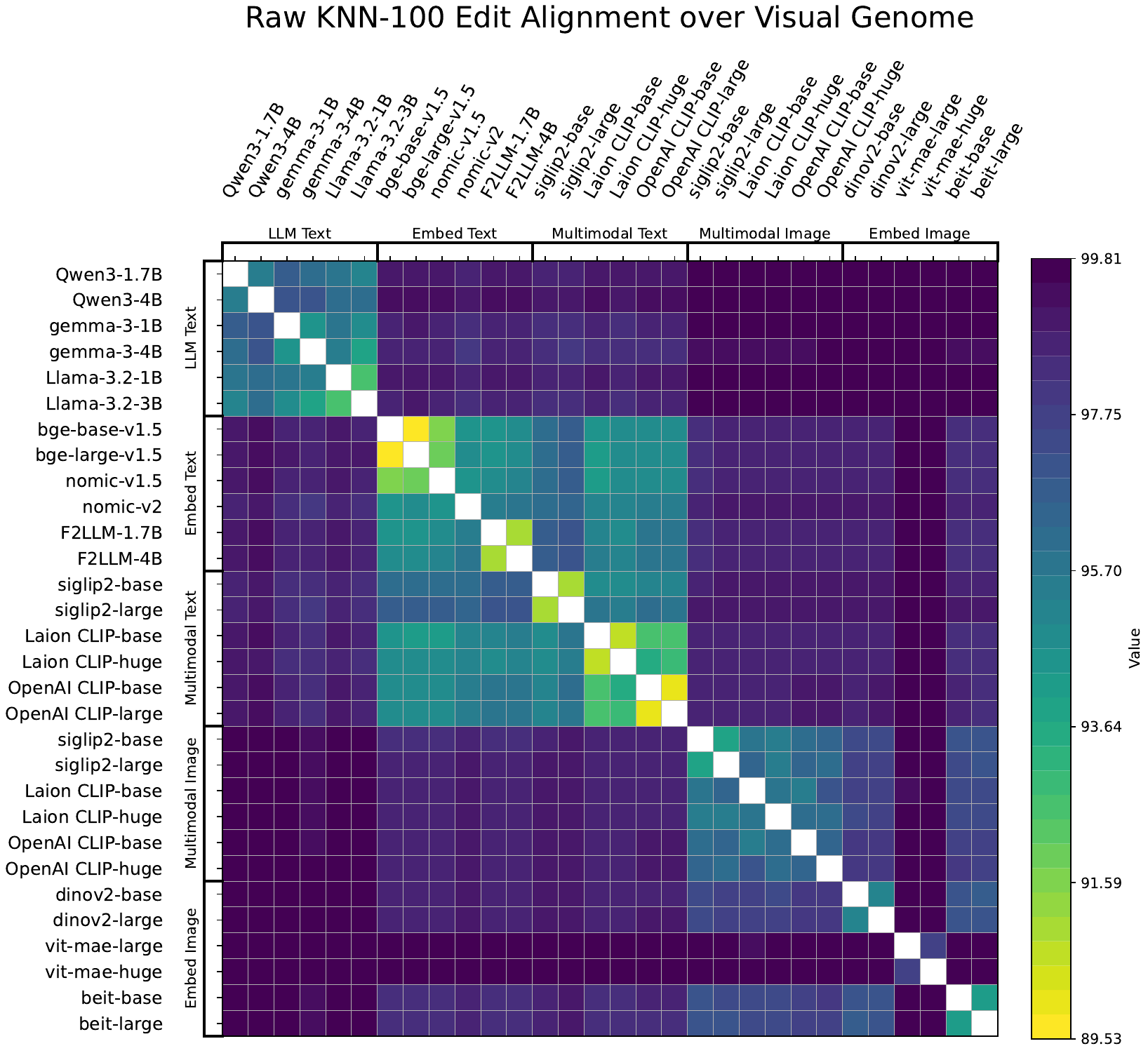}
    \end{minipage}
    \hfill
    \begin{minipage}[t]{0.3\textwidth}
        \centering
        \includegraphics[width = \linewidth]{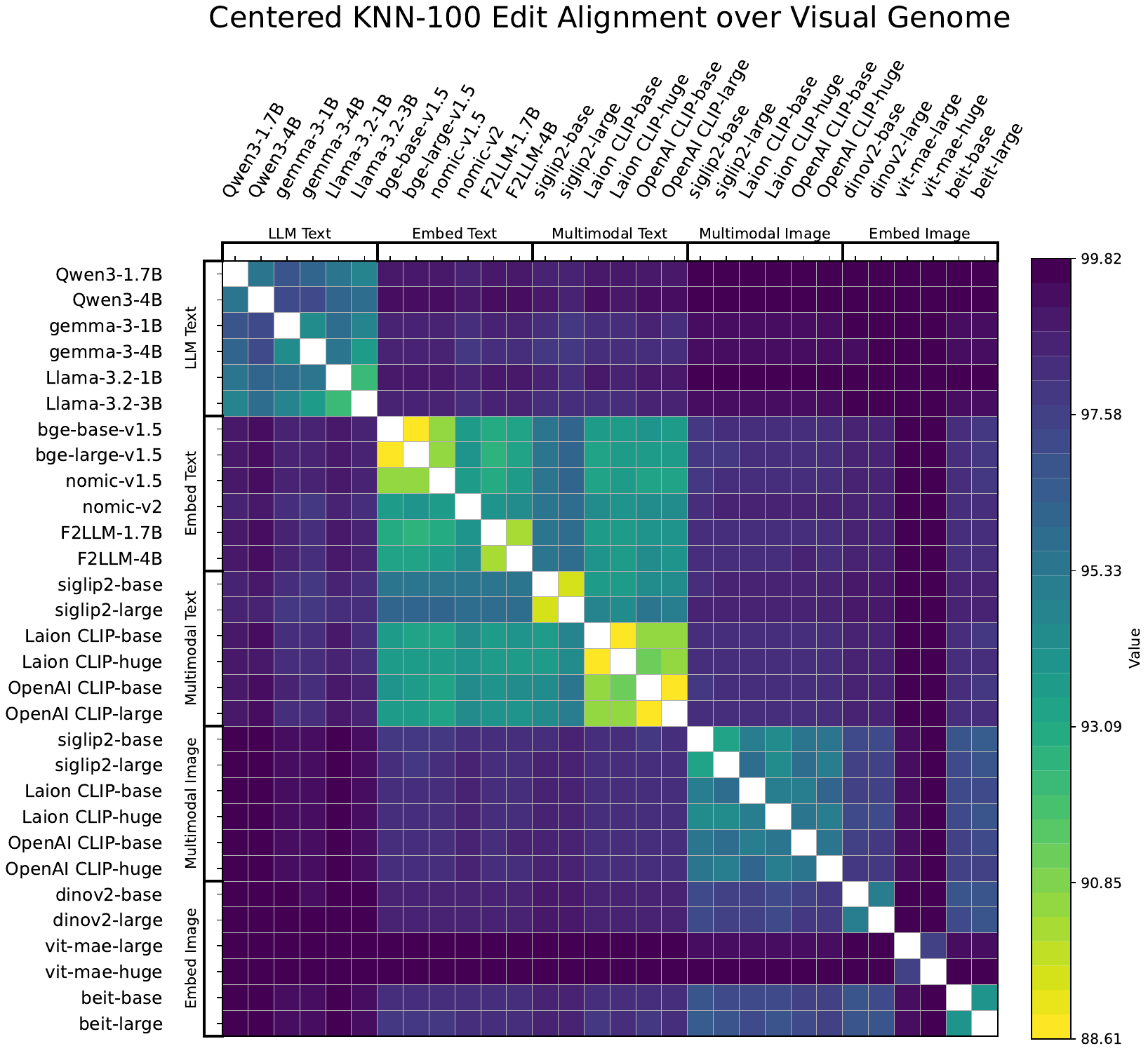}
    \end{minipage}
    \caption{Same plot as in Figure~\ref{fig:biasmain} but over Visual Genome.}
    \label{fig:biasvisual_genome2}
\end{figure}

\clearpage

\subsection{Noise: Higher Alignment for More Frequent Data}
\label{appendix:noise}
We reproduce the experiment from Figure~\ref{fig:noise_experiment_main} for other metrics as well, even though again our focus is on the KNN-10 overlap metric due to the local nature of the Platonic alignment observed in \cite{groger2026prharistotle}.

\subsubsection{Text vs LLM}

\begin{figure}[htbp]
    \centering

    \includegraphics[width = .8\linewidth]{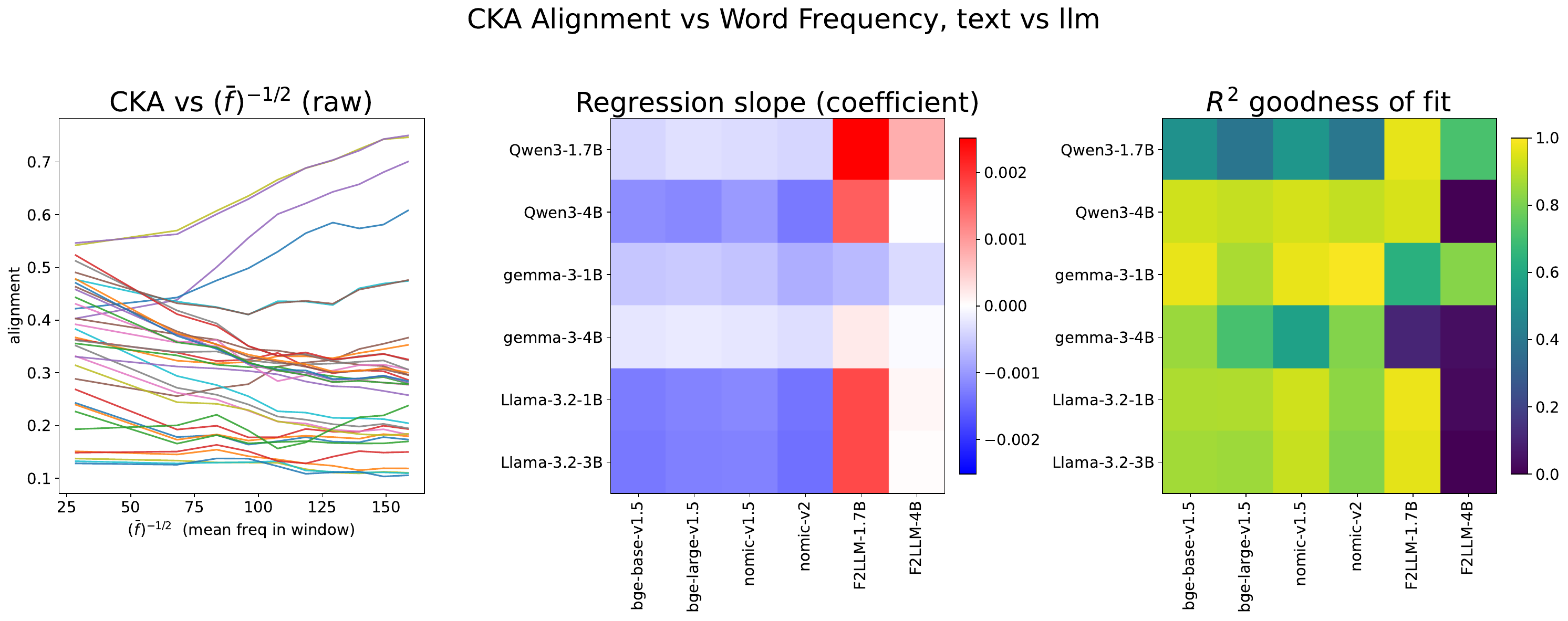}
    
    \vspace{0.4cm}

    \includegraphics[width = .8\linewidth]{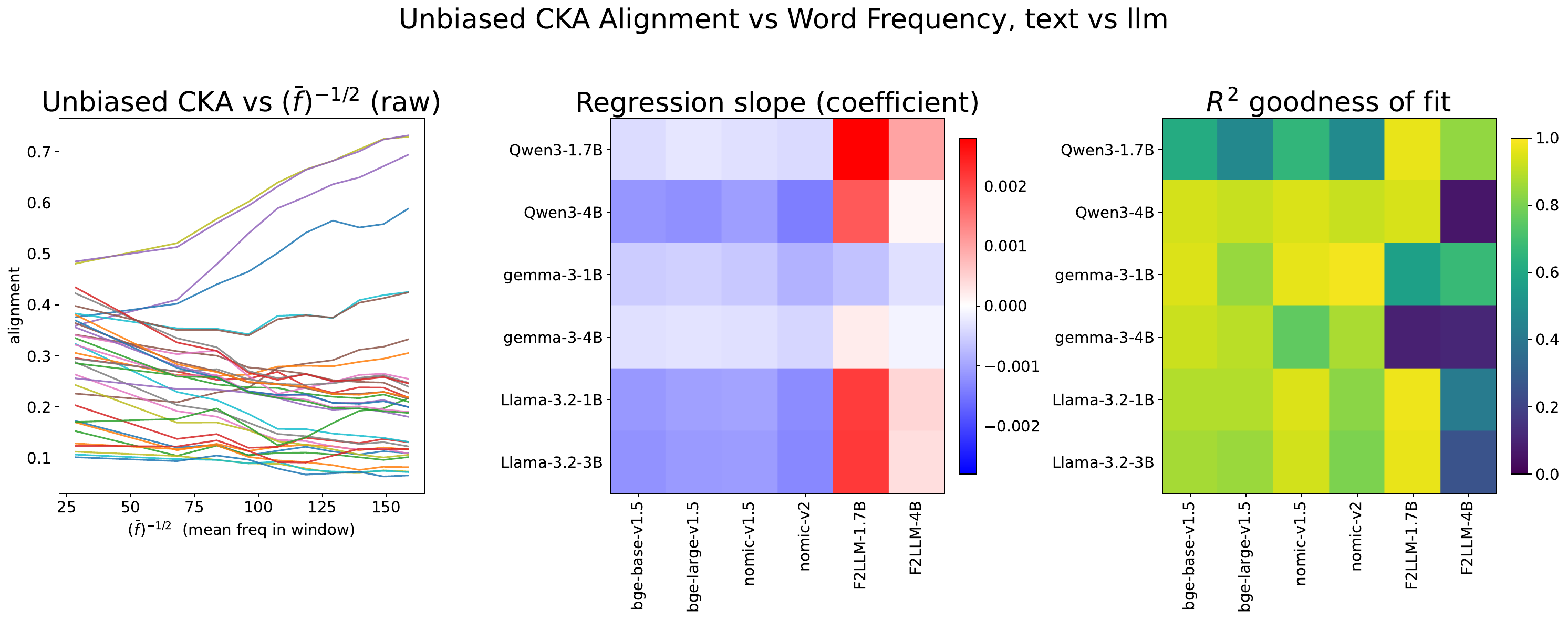}

    \vspace{0.4cm}

    \includegraphics[width = .8\linewidth]{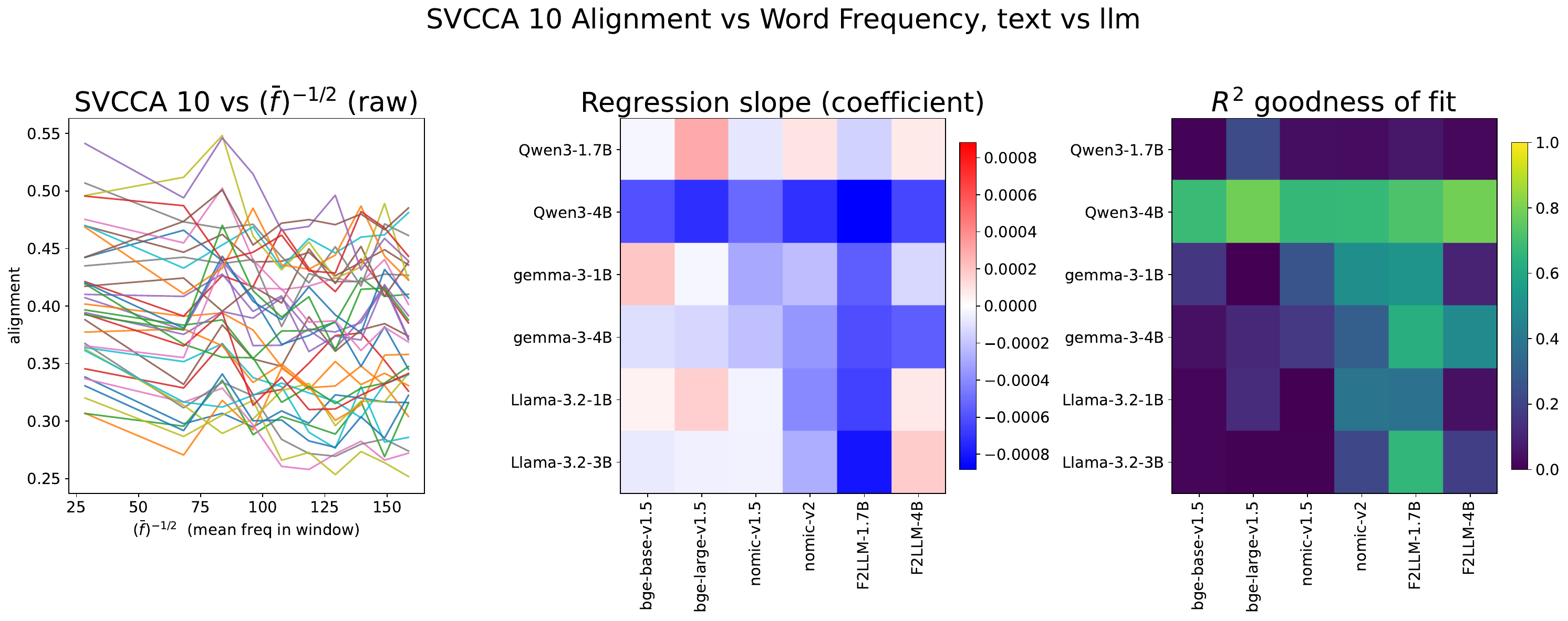}

    \vspace{0.4cm}

    \includegraphics[width = .8\linewidth]{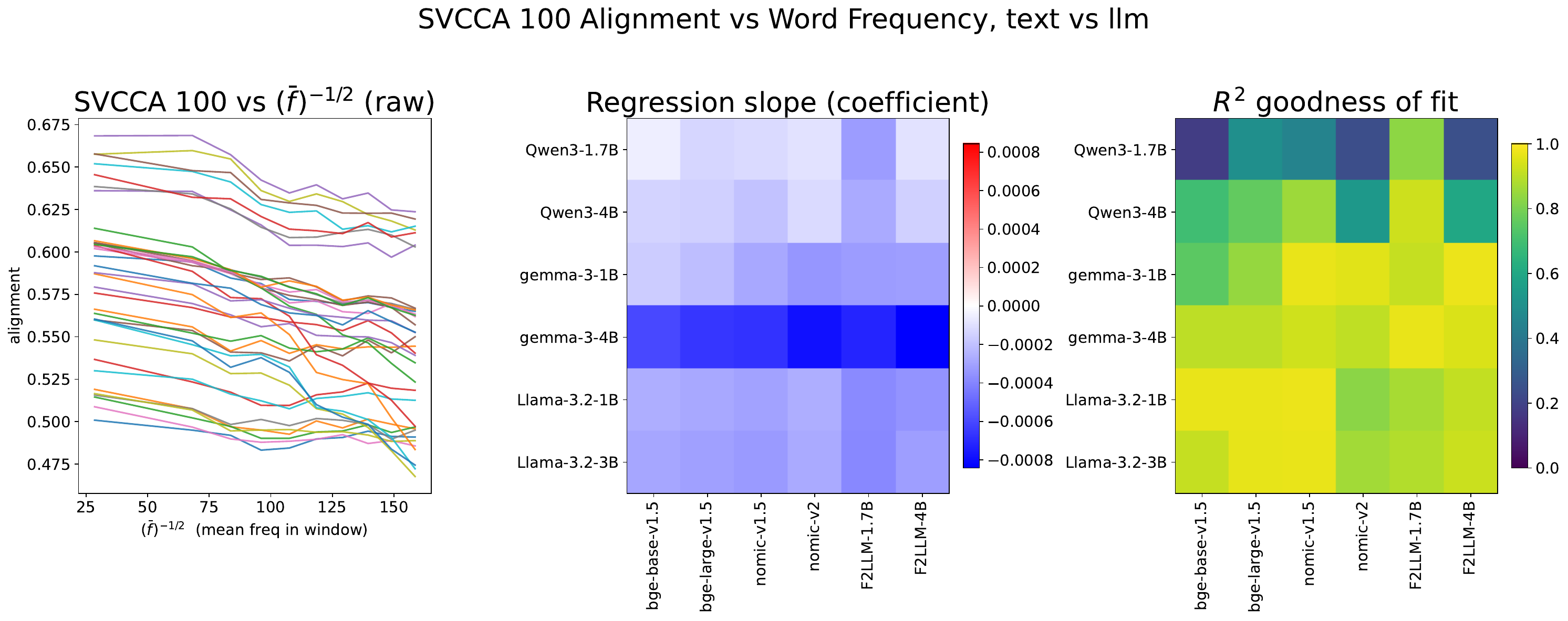}
    
    \caption{Same plot as in Figure~\ref{fig:noise_experiment_main} but for the CKA, Unbiased CKA, SVCCA 10 and SVCCA 100.}
    \label{fig:noisetextllm1}
\end{figure}

\clearpage

\begin{figure}[htb!]
    \centering

    \includegraphics[width = .8\linewidth]{noise_diagrams/KNN_Overlap_10_with_regression_text_llm.pdf}
    
    \vspace{0.4cm}

    \includegraphics[width = .8\linewidth]{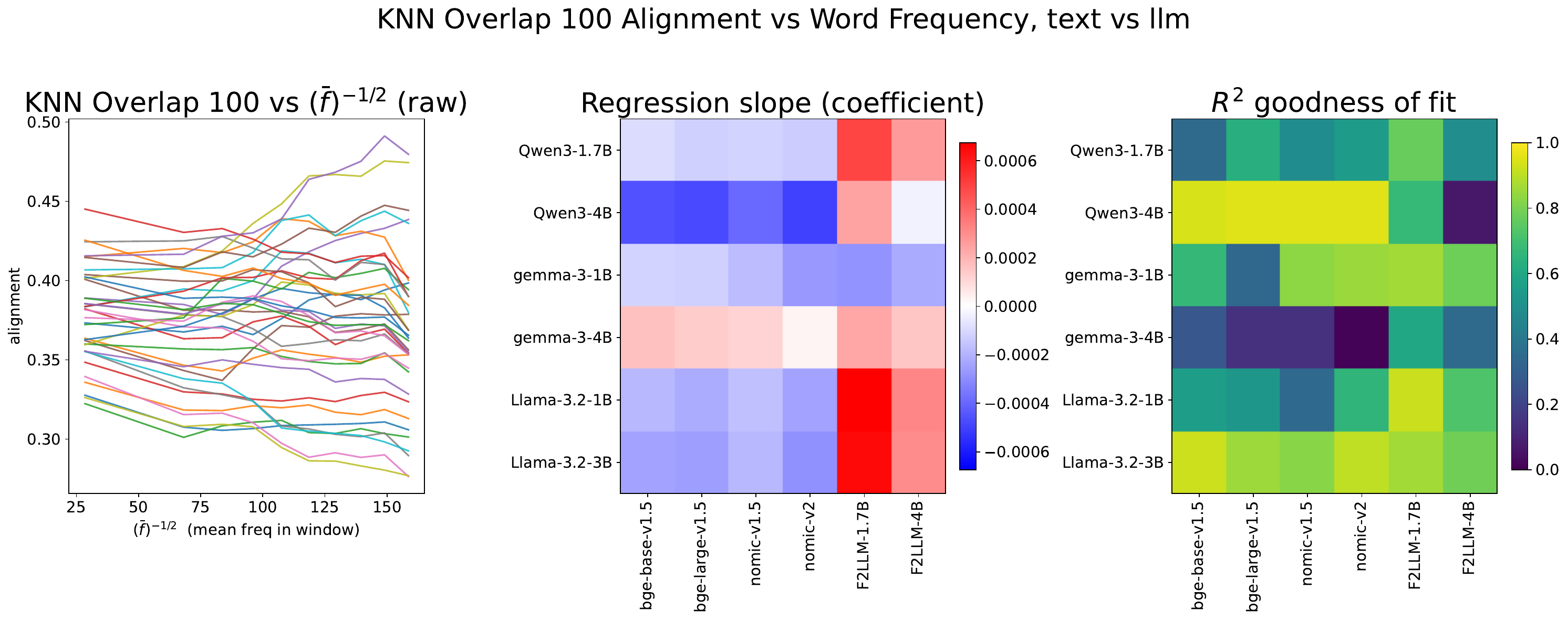}

    \vspace{0.4cm}

    \includegraphics[width = .8\linewidth]{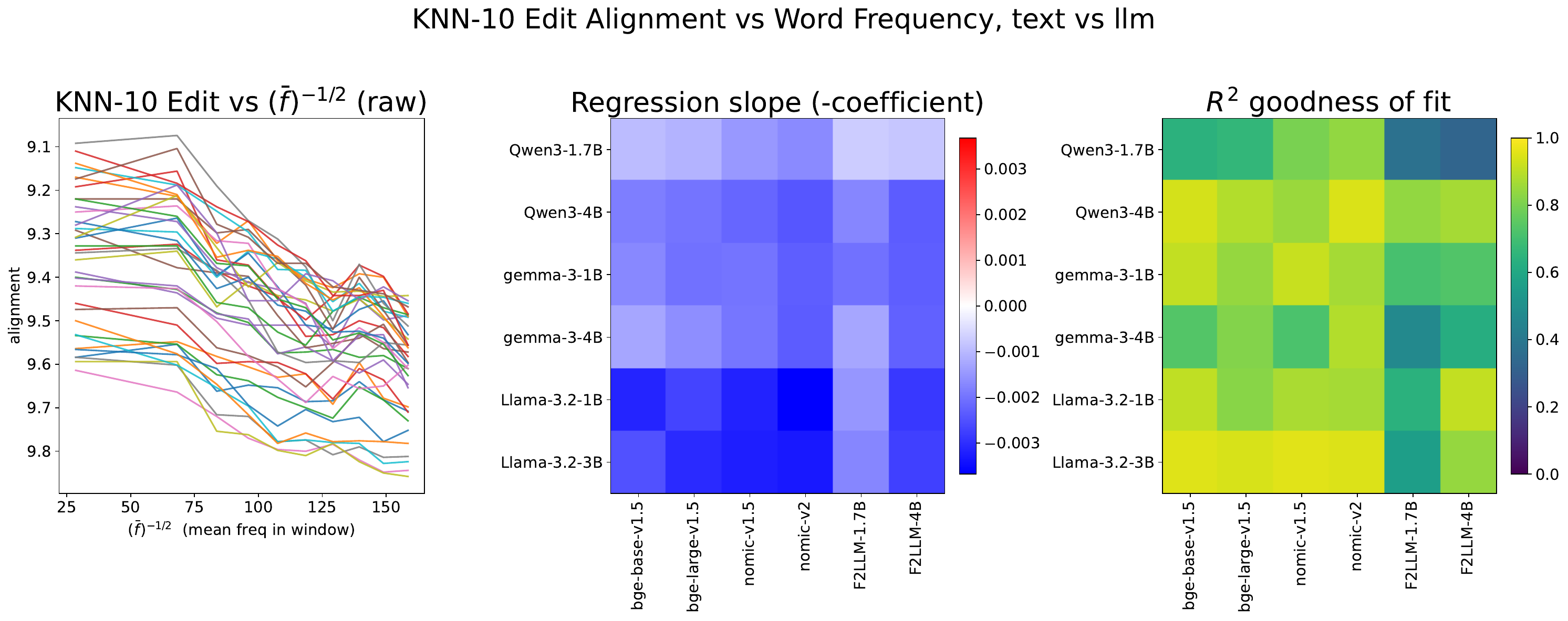}

    \vspace{0.4cm}

    \includegraphics[width = .8\linewidth]{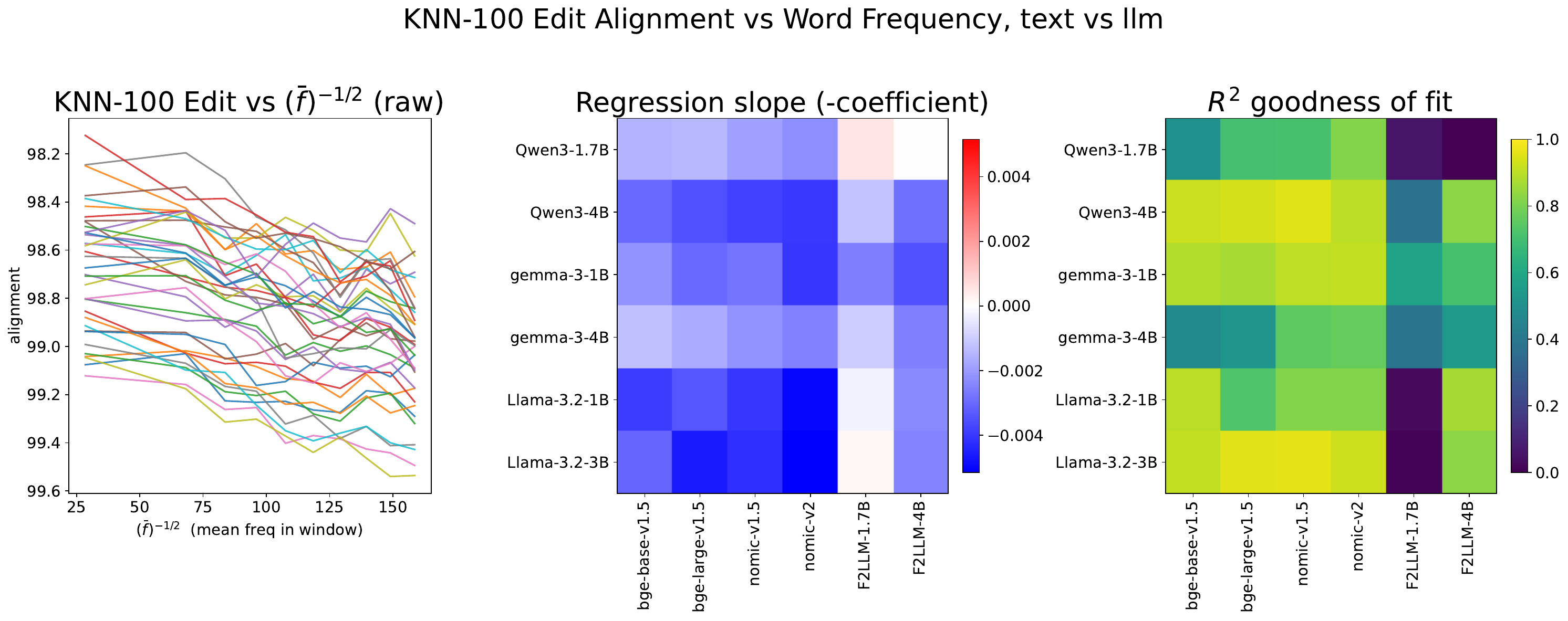}
    
    \caption{Same plot as in Figure~\ref{fig:noise_experiment_main} but for the KNN-10 Overlap, KNN-100 Overlap, KNN-10 Edit Distance, and KNN-100 Edit distance metrics. For the two KNN neighborhood edit distances, we plot the negative regression coefficients as higher values mean lower similarity unlike for other metrics.}
    \label{fig:noisetextllm2}
\end{figure}
We can see that with the exception of the SVCCA-10 metric, the nearly linear negative relationship between alignment and inverse square-root mean frequency is very consistent. 

\clearpage

\subsubsection{Text vs Text}
For the case of text-text and LLM-LLM models, we skip models from the same family as they are trained on similar data, often distilled from bigger models etc., which makes the noises terms highly correlated and the alignment trends uninterpretable.

\begin{figure}[htbp]
    \centering

    \includegraphics[width = .8\linewidth]{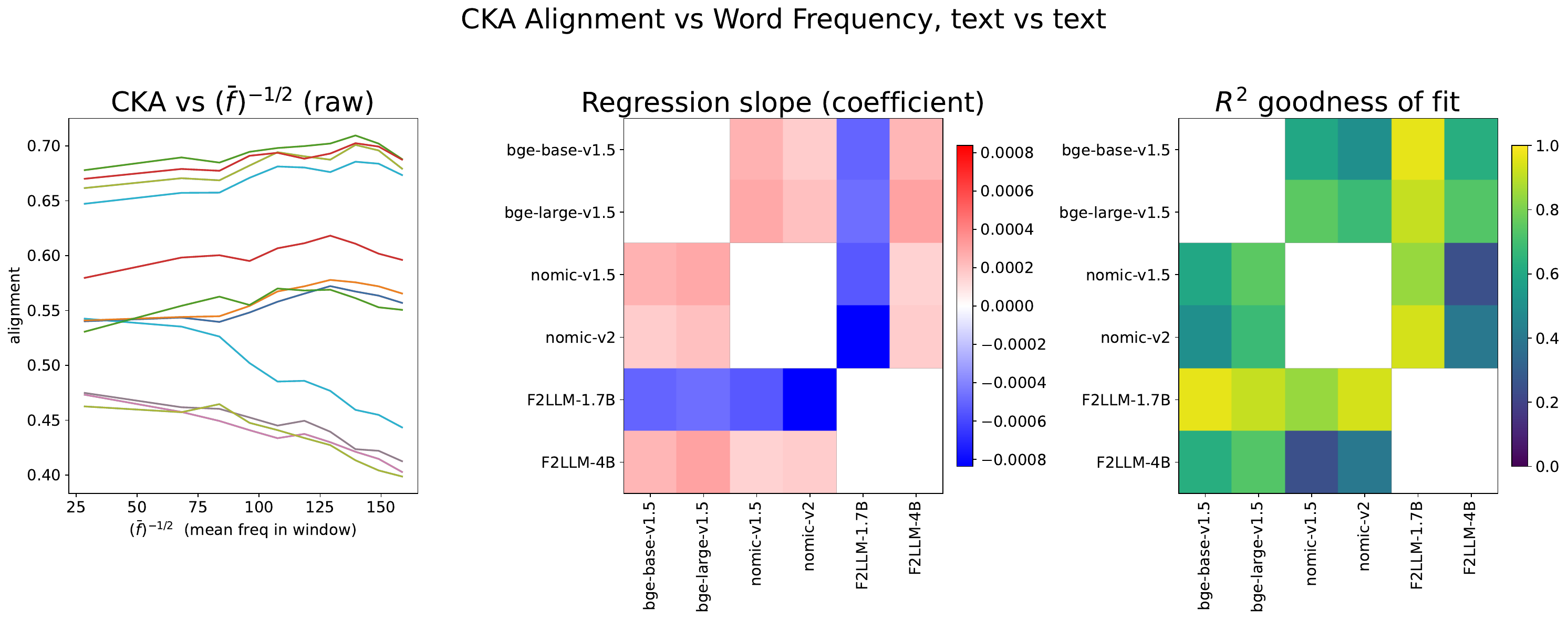}
    
    \vspace{0.4cm}

    \includegraphics[width = .8\linewidth]{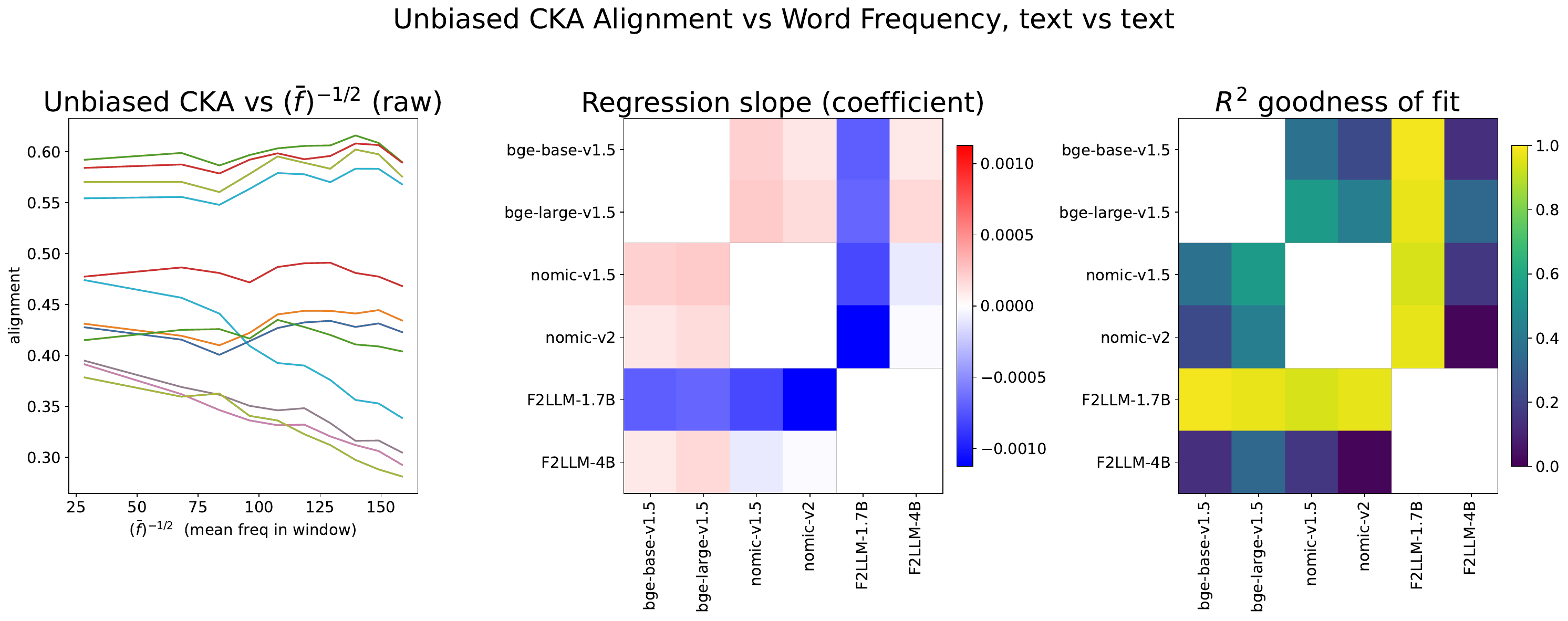}

    \vspace{0.4cm}

    \includegraphics[width = .8\linewidth]{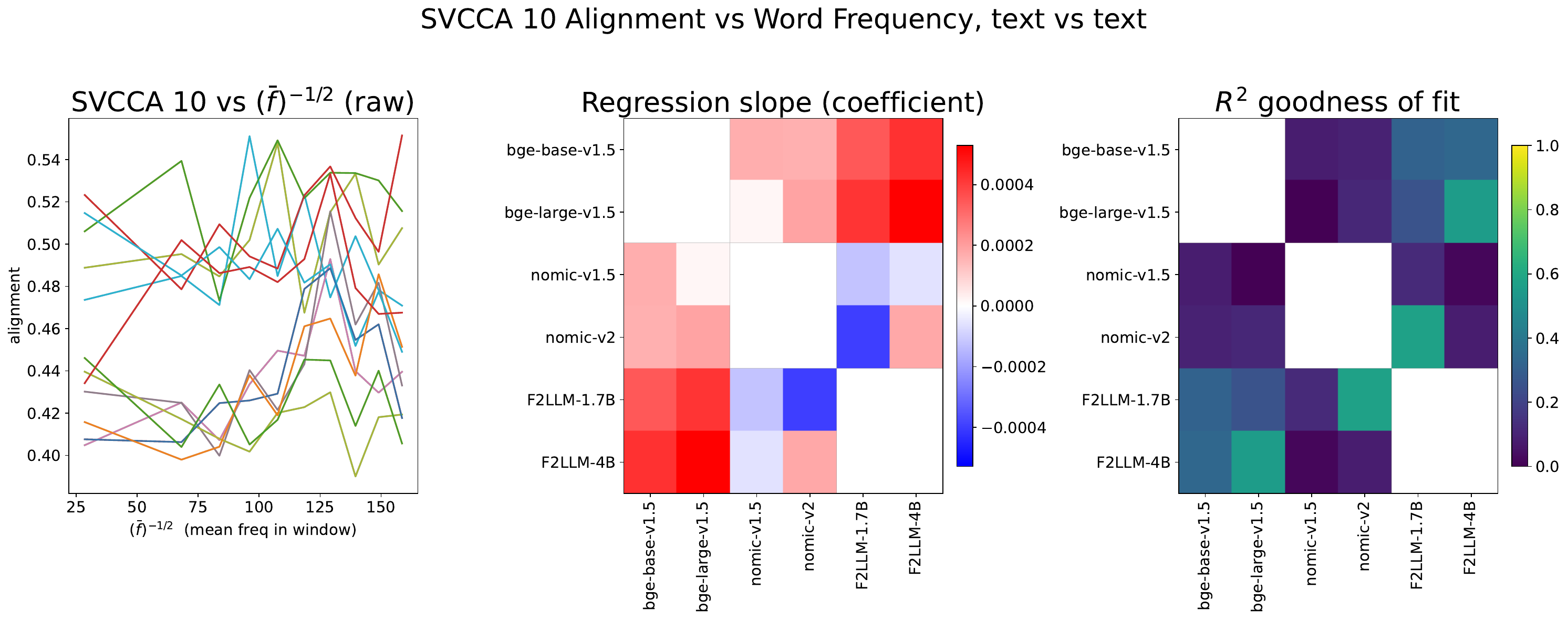}

    \vspace{0.4cm}

    \includegraphics[width = .8\linewidth]{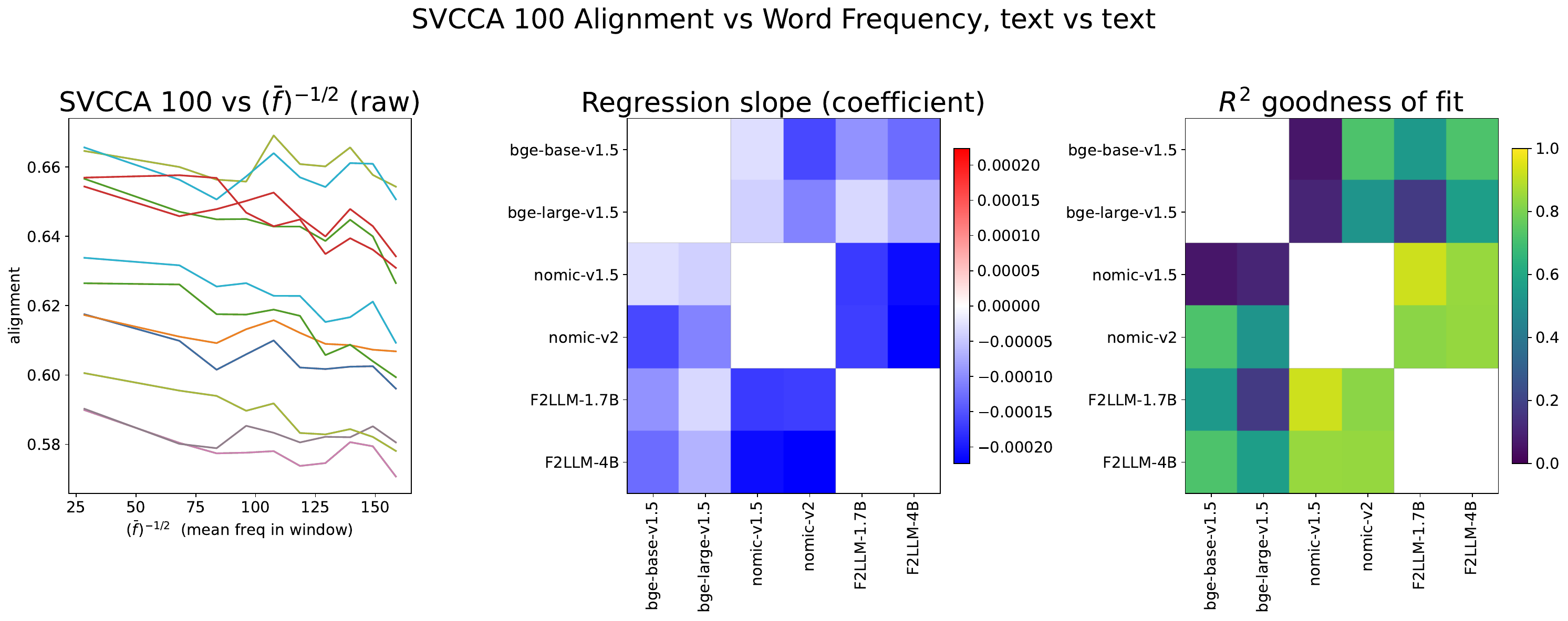}
    
    \caption{Same plot as in Figure~\ref{fig:noisetextllm1} but for text embedding vs text embedding models.}
    \label{fig:noisetexttext1}
\end{figure}

\clearpage

\begin{figure}[htbp]
    \centering

    \includegraphics[width = .8\linewidth]{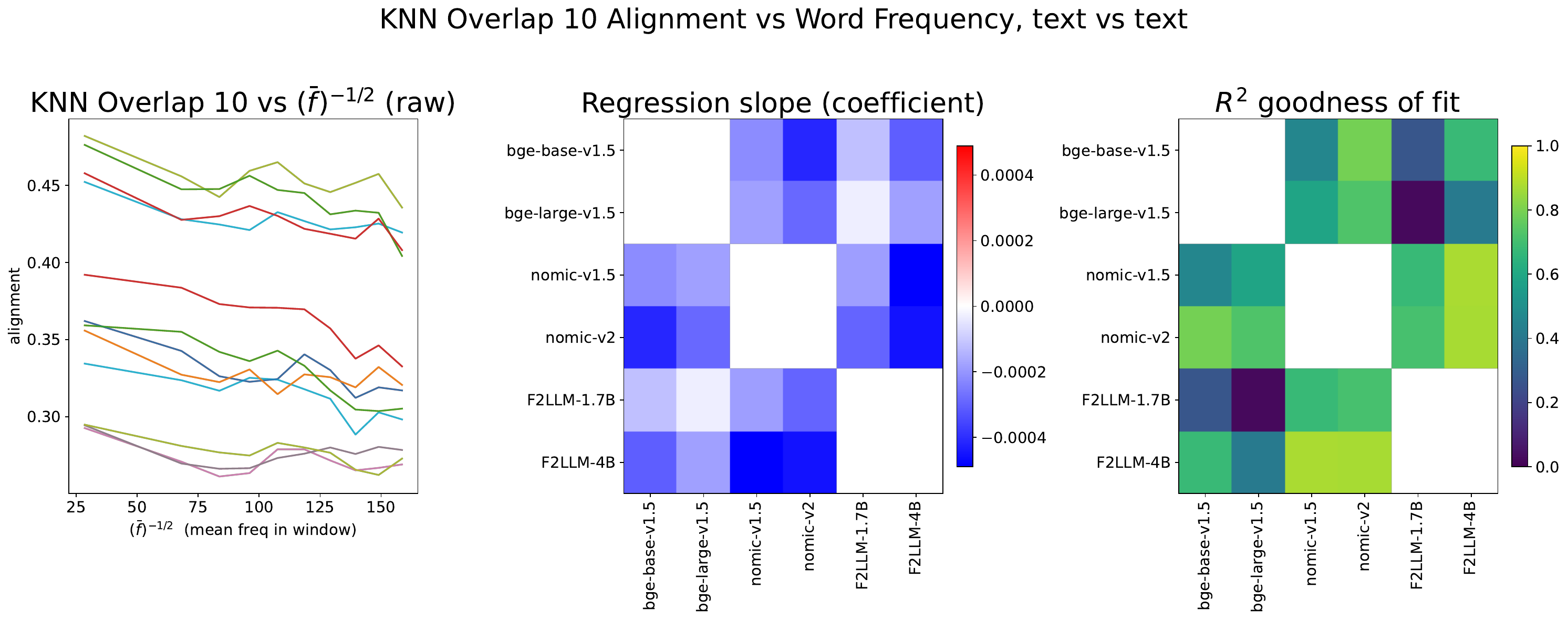}
    
    \vspace{0.4cm}

    \includegraphics[width = .8\linewidth]{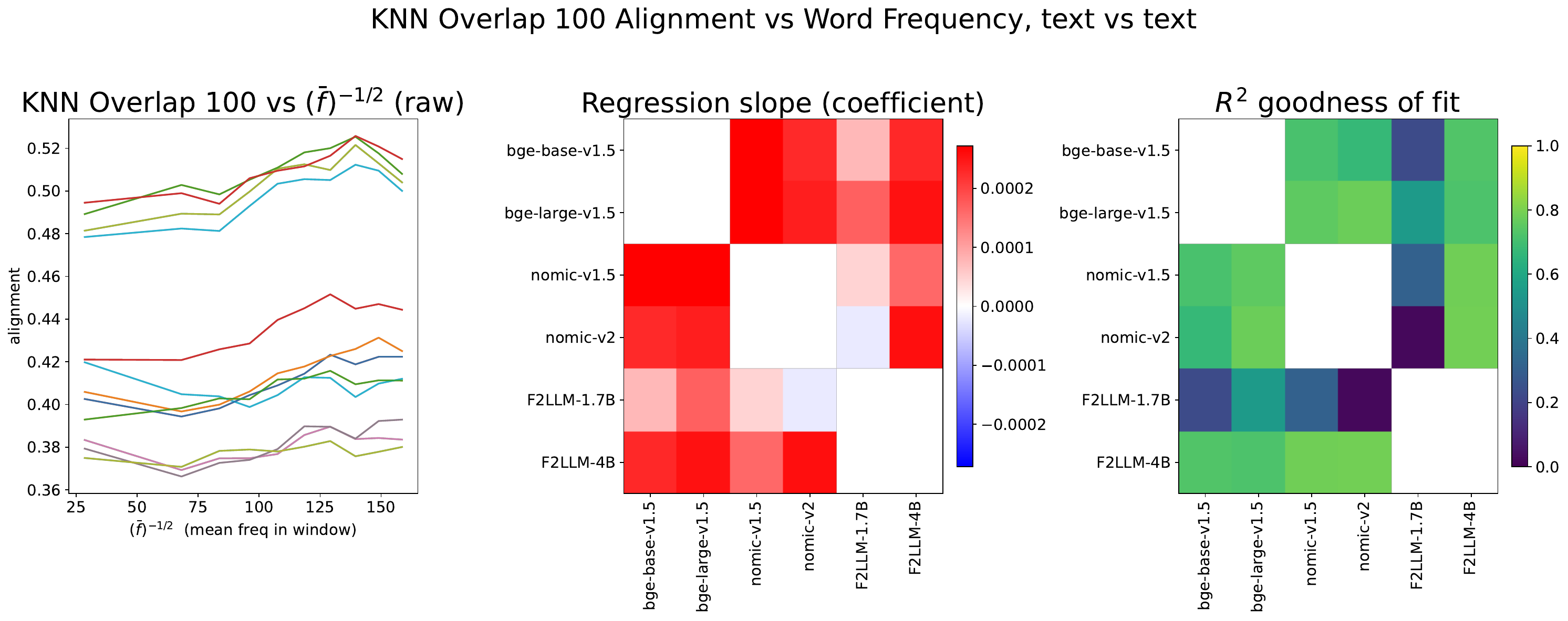}

    \vspace{0.4cm}

    \includegraphics[width = .8\linewidth]{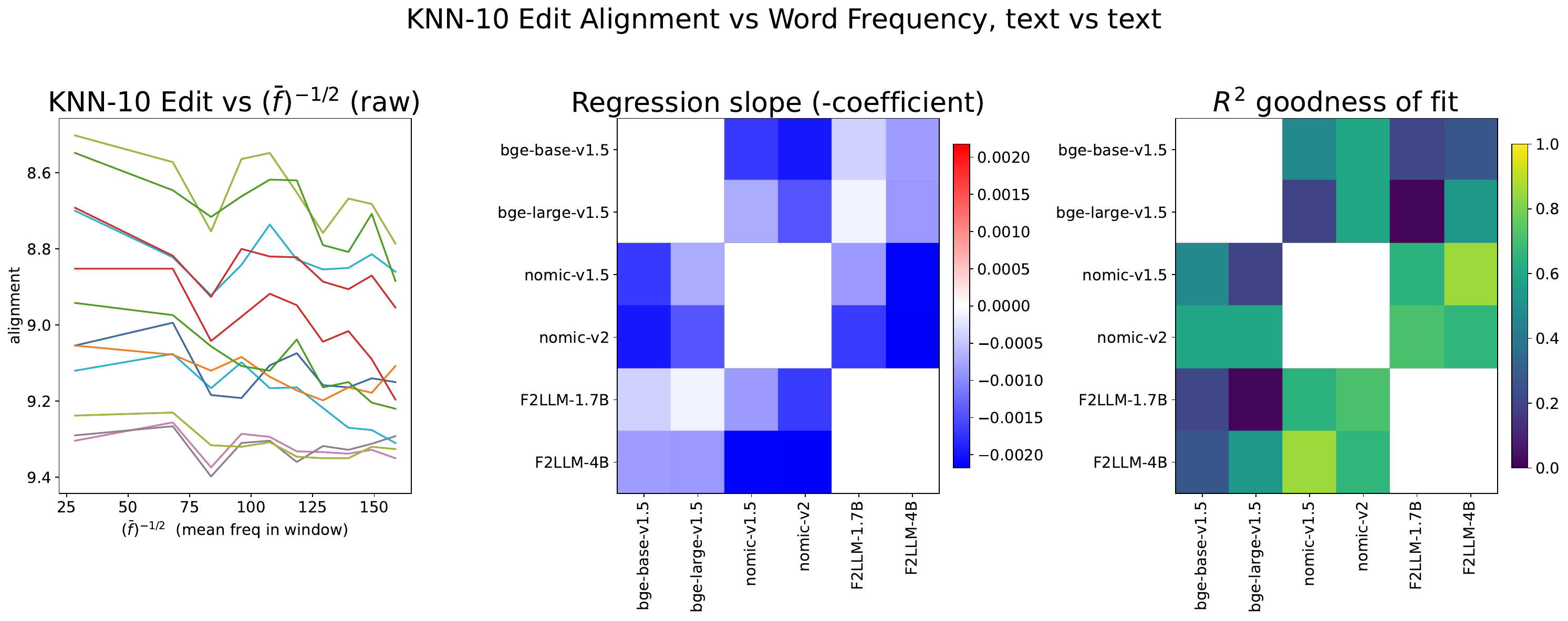}

    \vspace{0.4cm}

    \includegraphics[width = .8\linewidth]{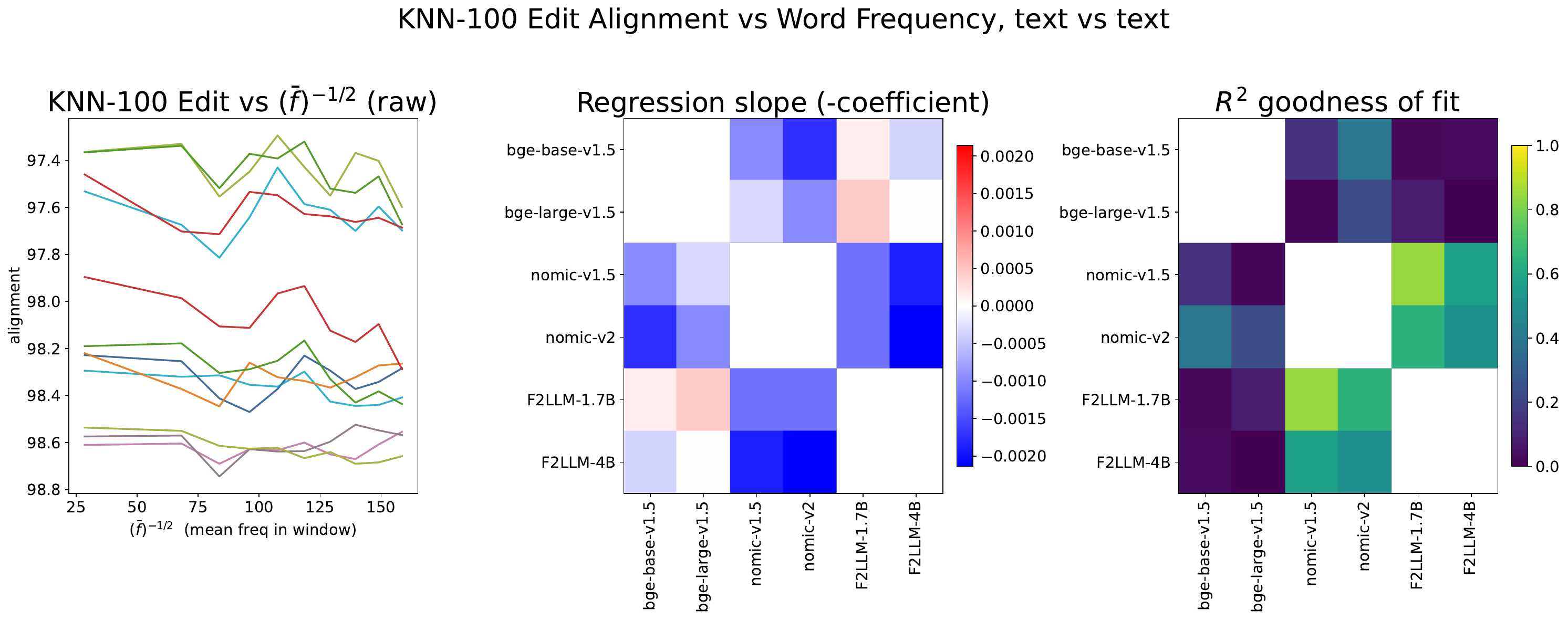}
    
    \caption{Same plot as in Figure~\ref{fig:noisetextllm2} but for text embedding vs text embedding models.}
    \label{fig:noisetexttext2}
\end{figure}

\clearpage

\subsubsection{LLM vs LLM}

\begin{figure}[htbp]
    \centering

    \includegraphics[width = .8\linewidth]{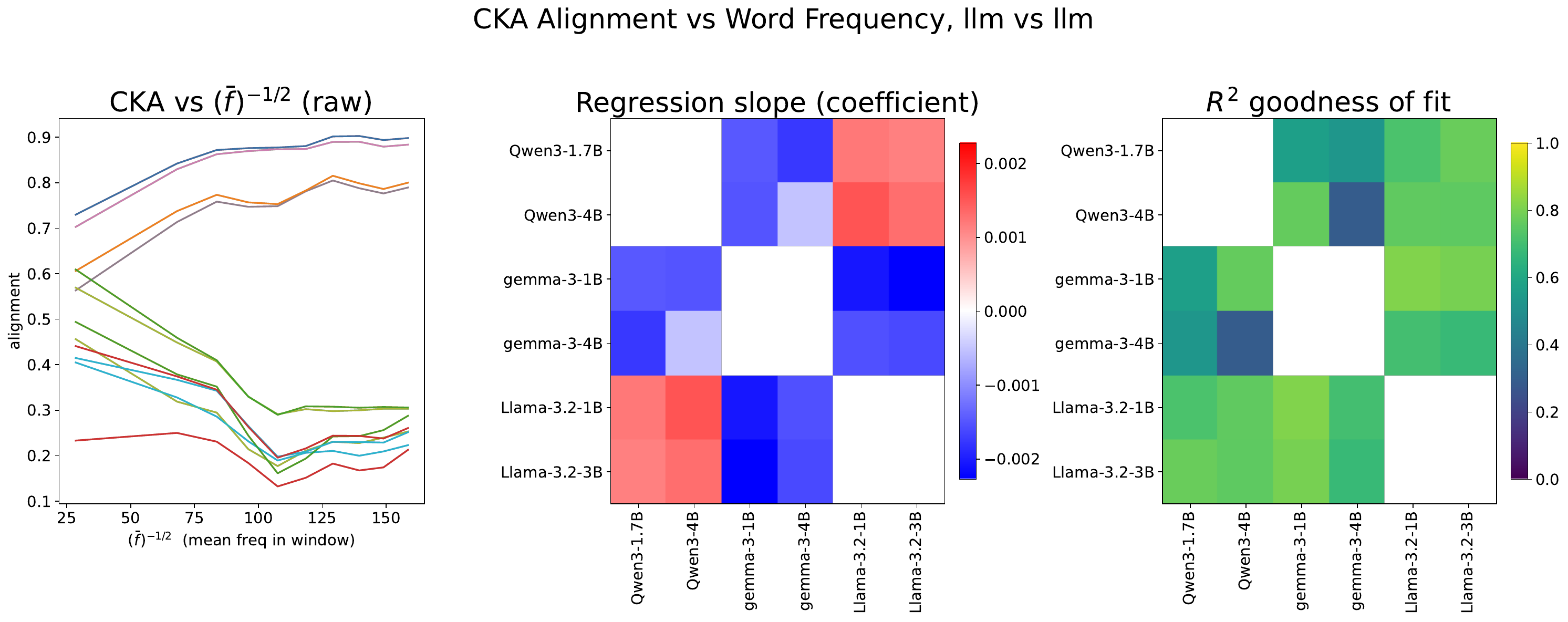}
    
    \vspace{0.4cm}

    \includegraphics[width = .8\linewidth]{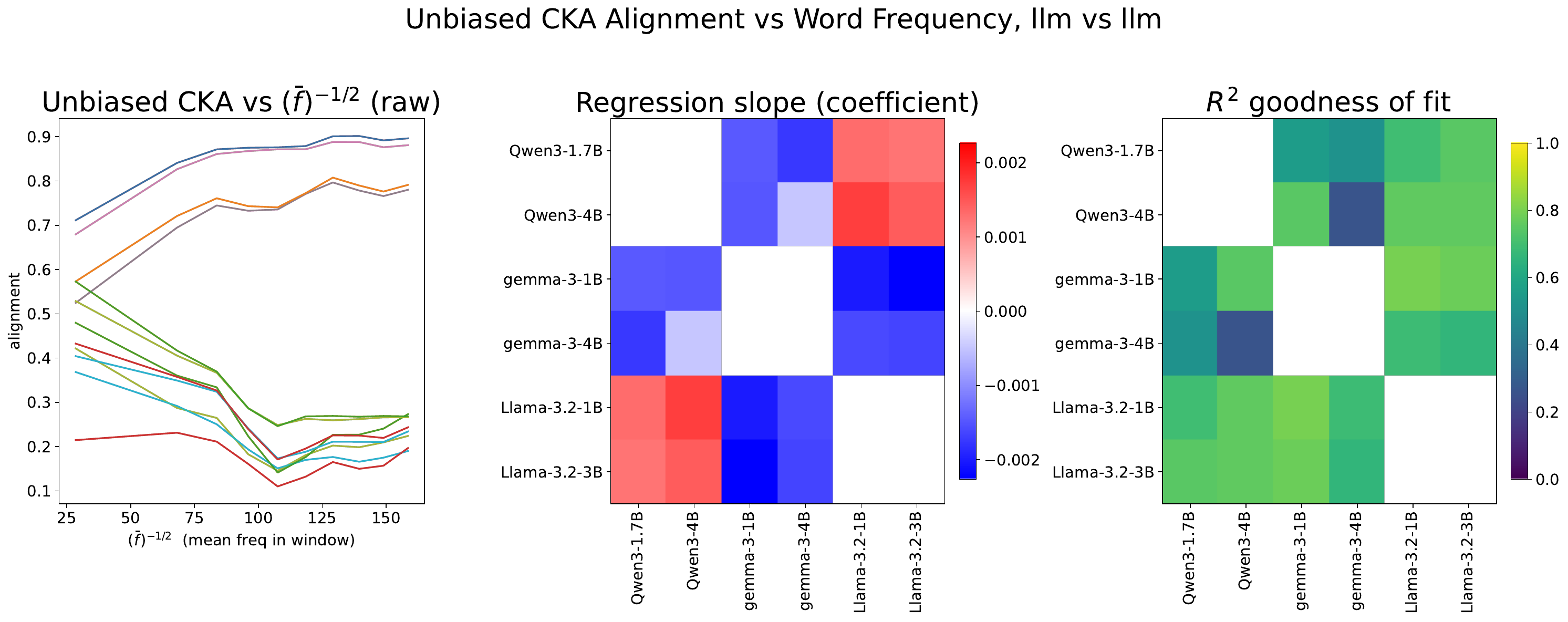}

    \vspace{0.4cm}

    \includegraphics[width = .8\linewidth]{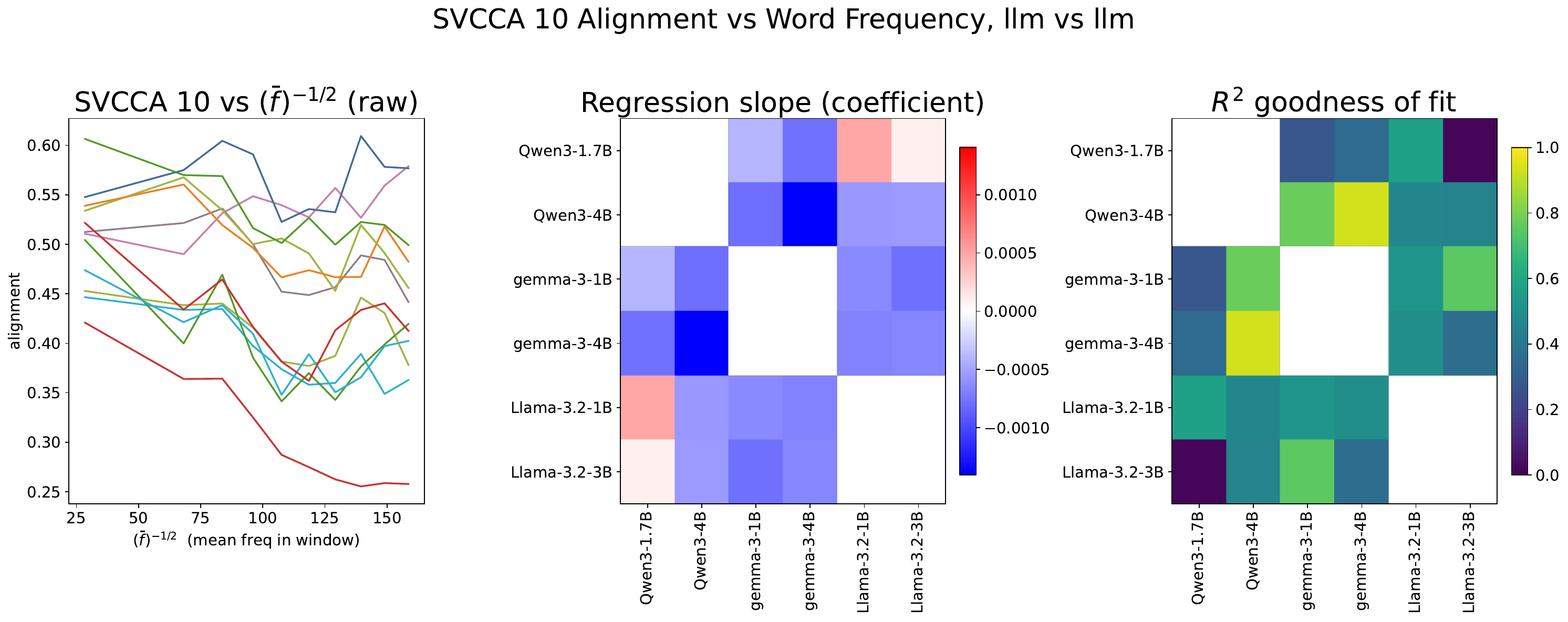}

    \vspace{0.4cm}

    \includegraphics[width = .8\linewidth]{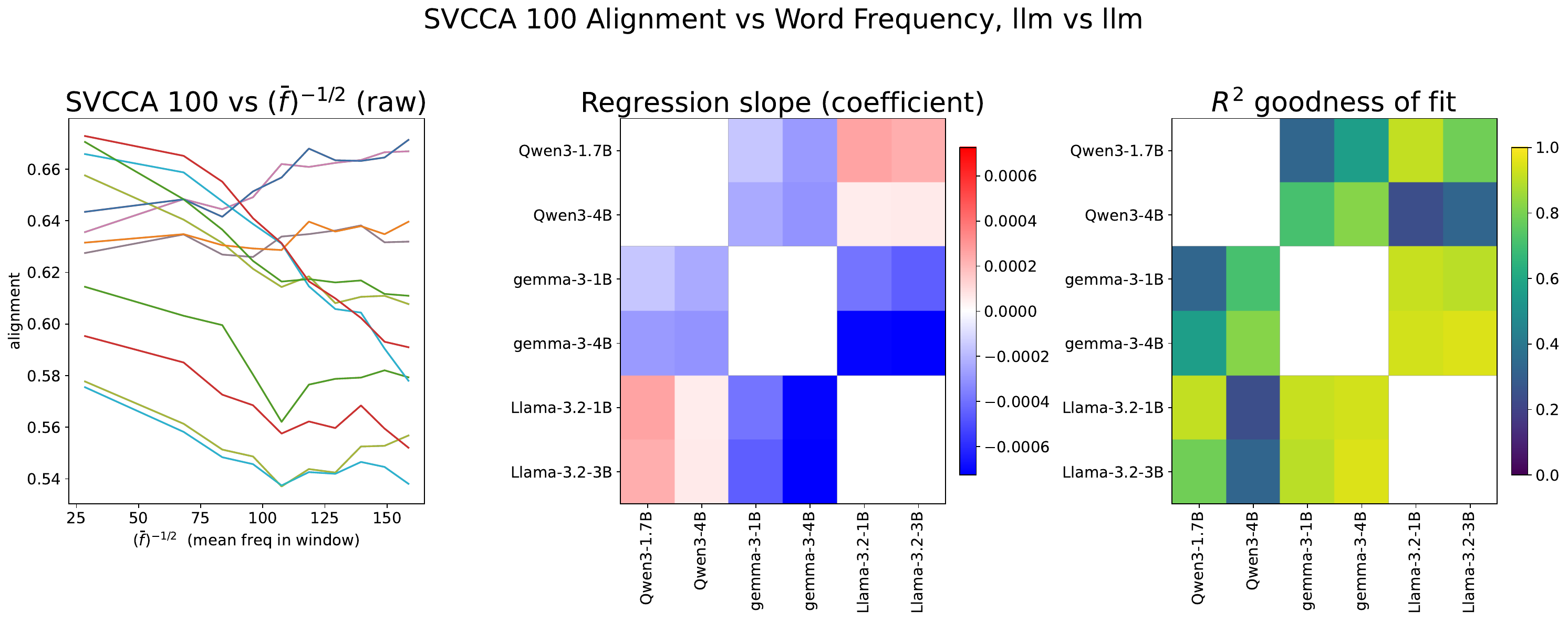}
    
    \caption{Same plot as in Figure~\ref{fig:noisetextllm1} but for LLM vs LLM models.}
    \label{fig:noisellmllm1}
\end{figure}

\clearpage

\begin{figure}[htbp]
    \centering

    \includegraphics[width = .8\linewidth]{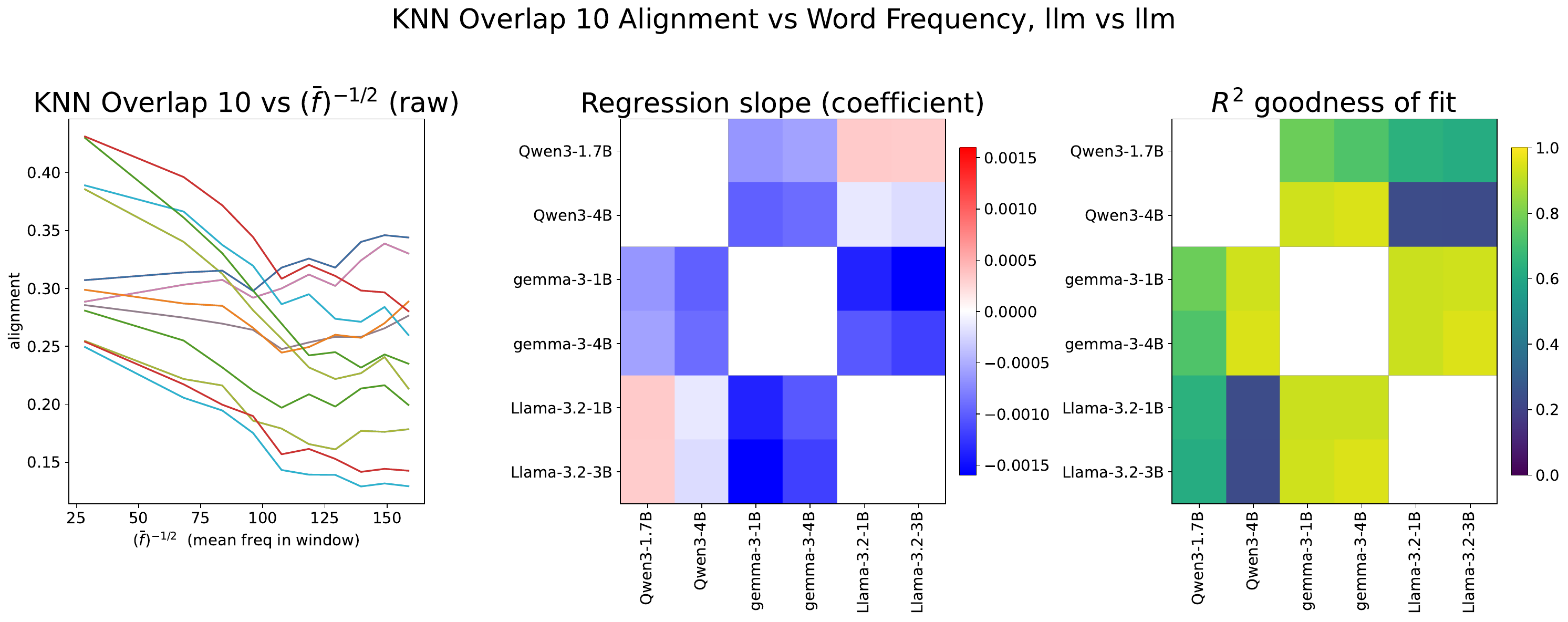}
    
    \vspace{0.4cm}

    \includegraphics[width = .8\linewidth]{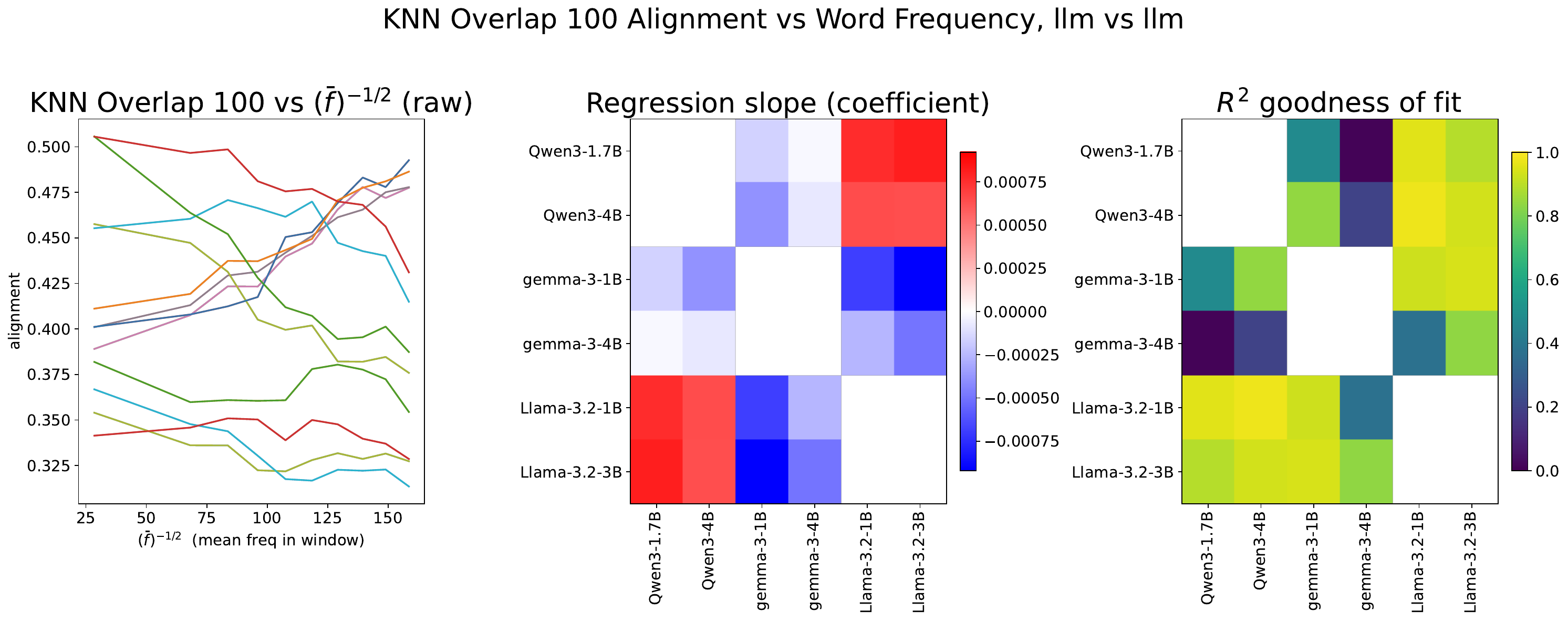}

    \vspace{0.4cm}

    \includegraphics[width = .8\linewidth]{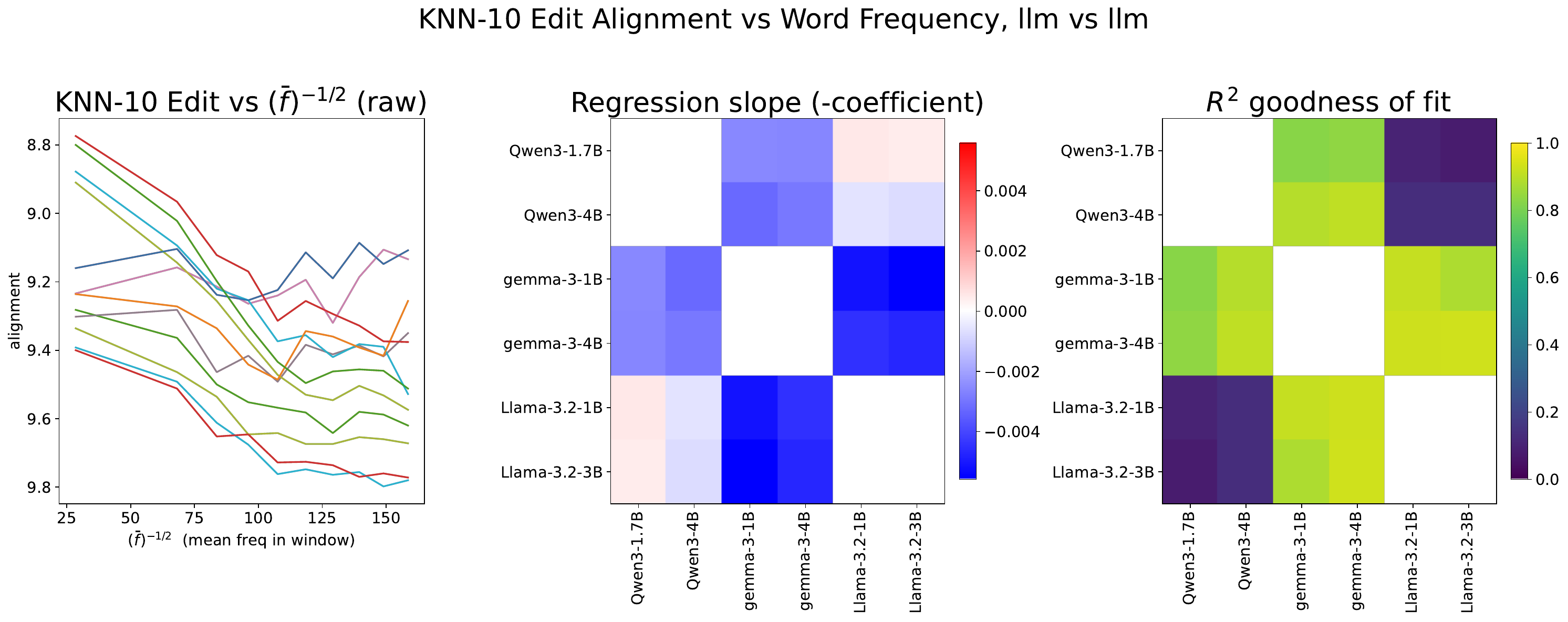}

    \vspace{0.4cm}

    \includegraphics[width = .8\linewidth]{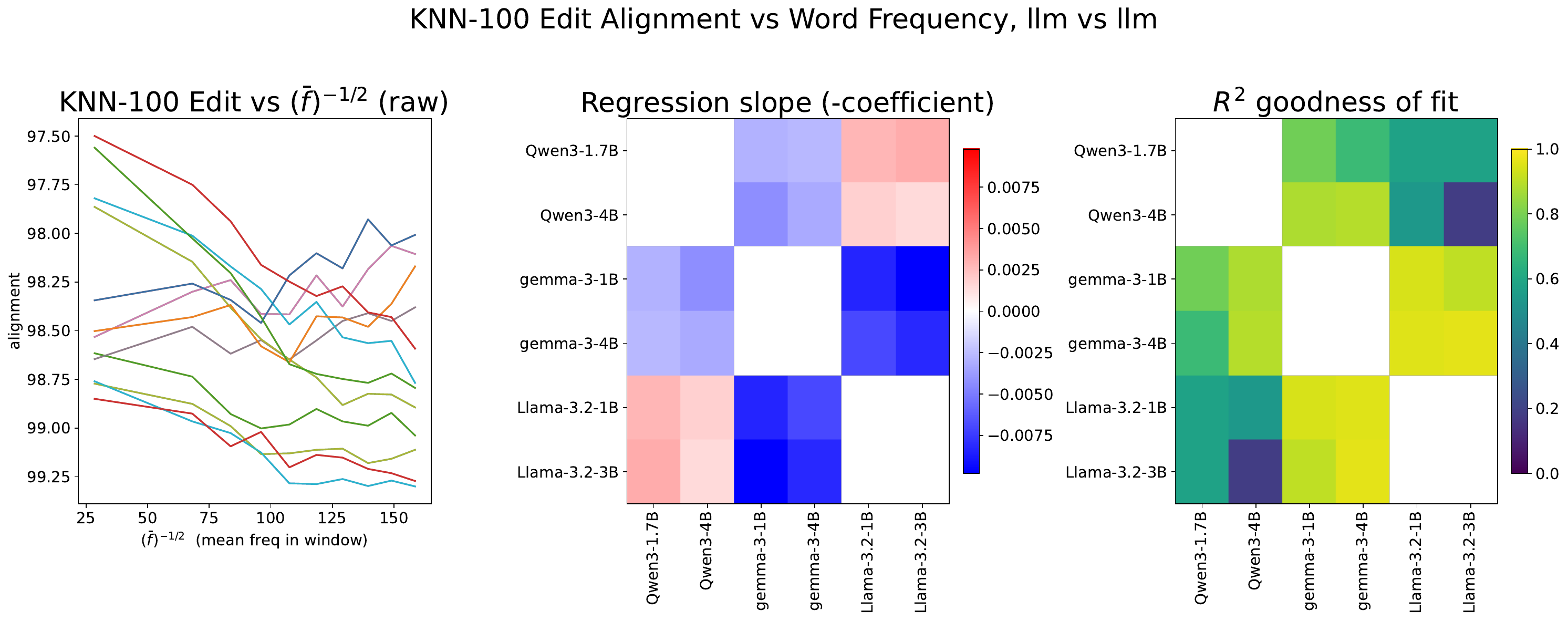}
    
    \caption{Same plot as in Figure~\ref{fig:noisetextllm2} but for LLM vs LLM models.}
    \label{fig:noisellmllm2}
\end{figure}

The trends are somewhat weaker for LLM vs LLM and text embedding vs text embedding models, even though still strong with respect to the KNN-10 Overlap metric (as well as the two KNN neighborhood distance metrics). Again, we notice that typically the positive correlation between data frequency and alignment is stronger for larger models within the same family. This is consistent with our \eqref{eq:statmodel} since one could expect larger models to have a smaller bias due to their greater expressivity, so alignment trends are more dependent on noise. 
\newpage

\subsubsection{Ablating The Number of Tokens}
We repeat the same experiments but restricting only to words that are composed of a single token in each model, thus controlling for the number of tokens. The results are consistent. We only present them for the KNN-10 metric for compactness.

\begin{figure}[htbp]
    \centering

    \includegraphics[width = .8\linewidth]{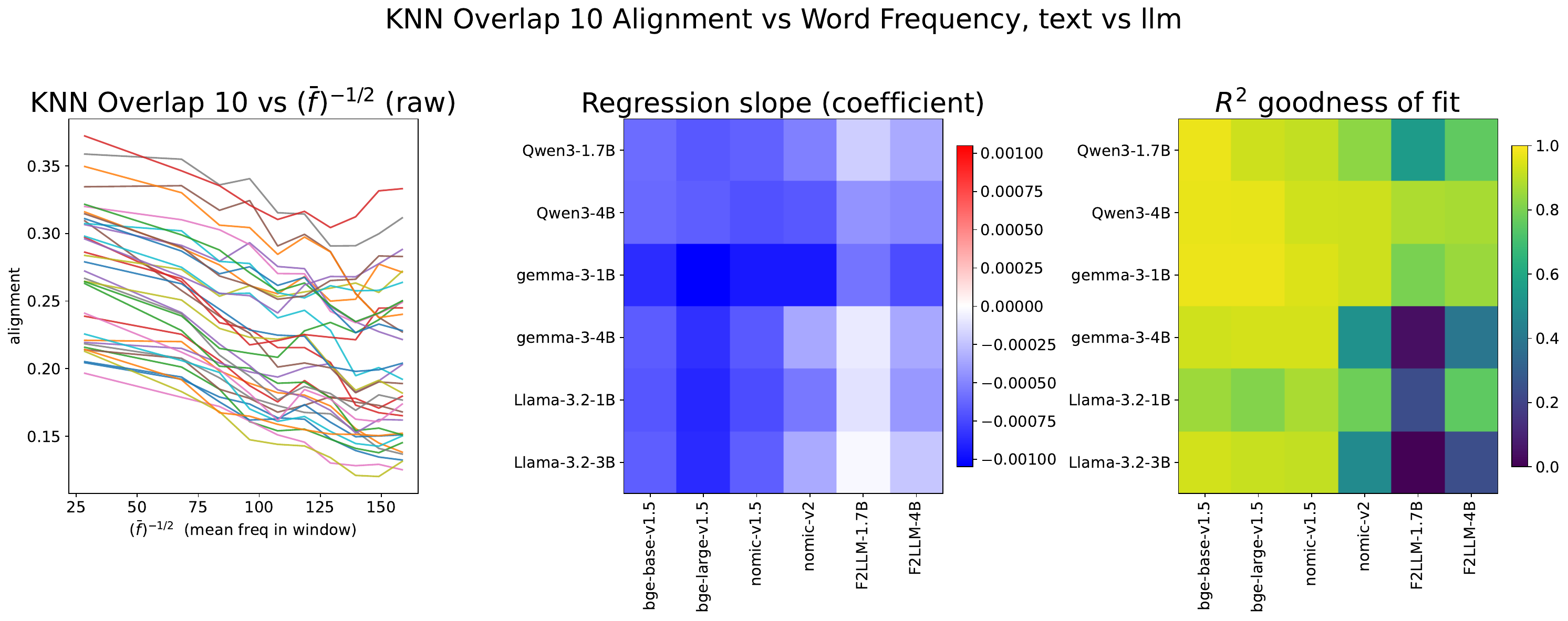}
    
    \vspace{0.4cm}

    \includegraphics[width = .8\linewidth]{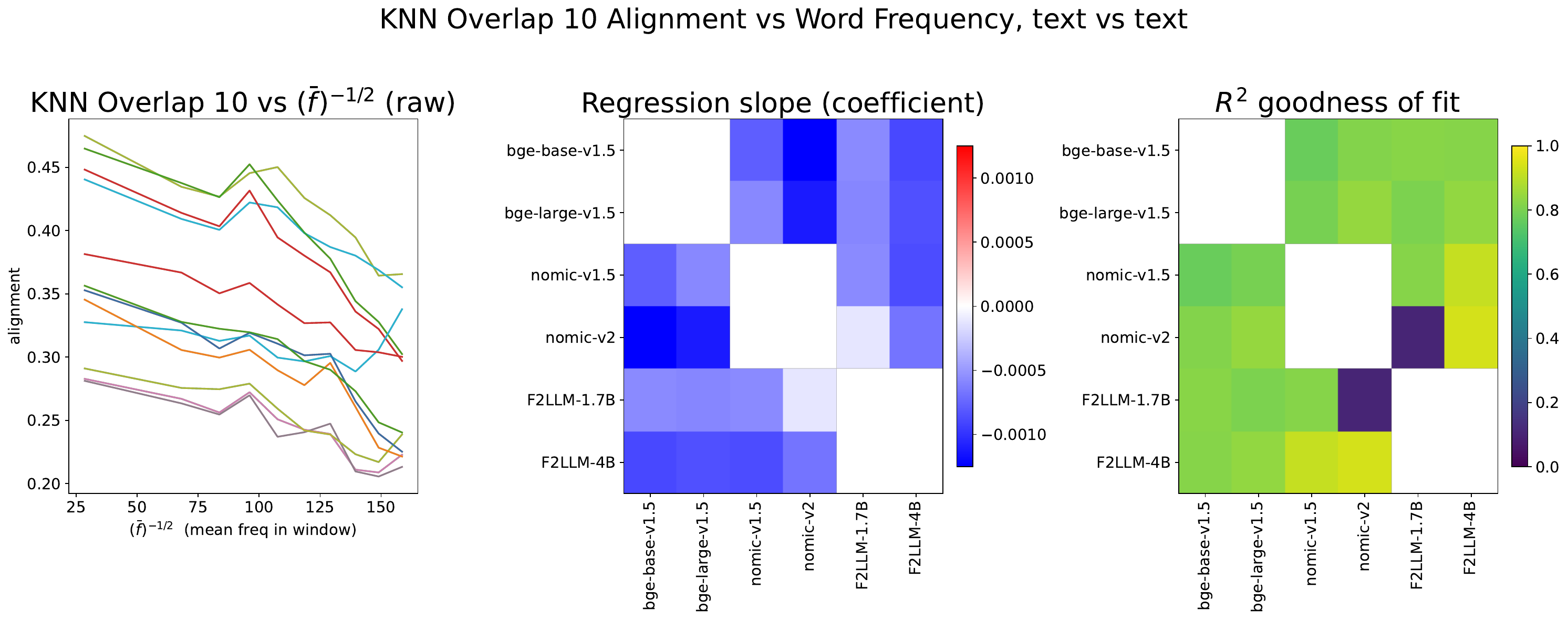}

    \vspace{0.4cm}

    \includegraphics[width = .8\linewidth]{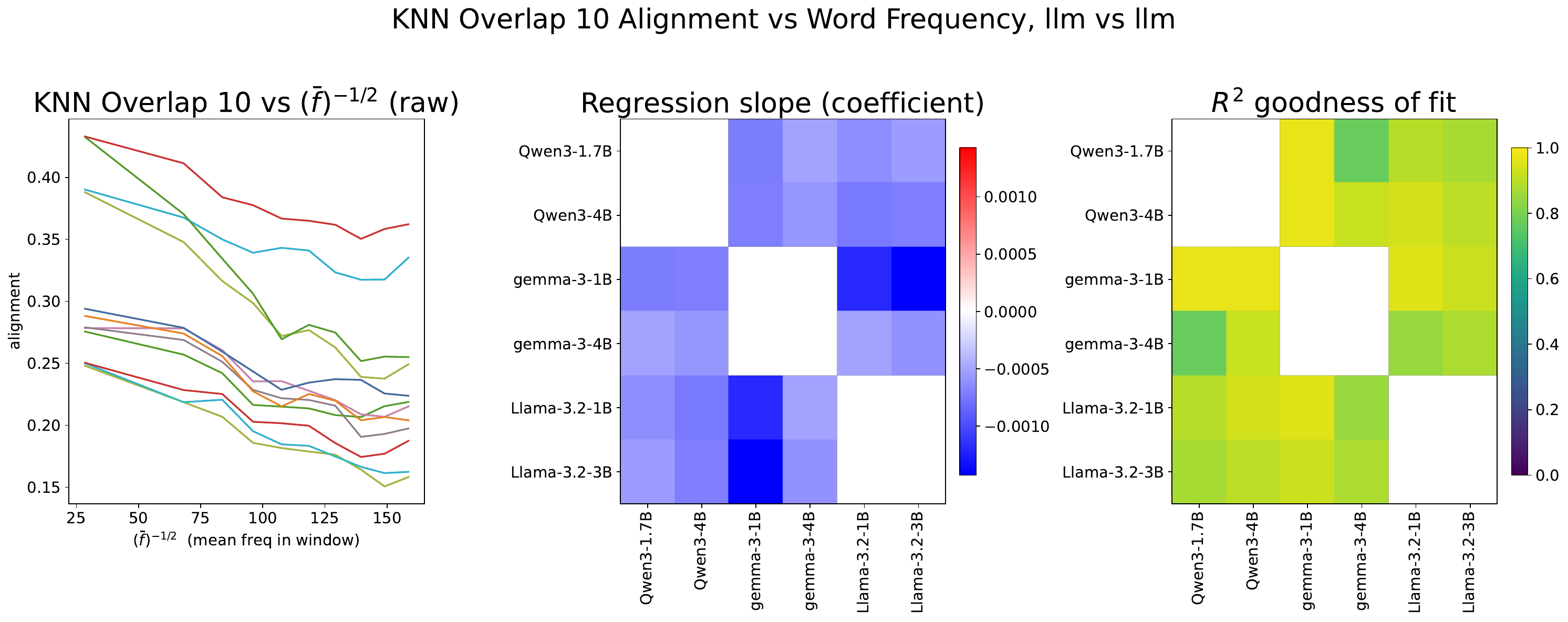}
    
    \caption{Same plot as in Figure~\ref{fig:noisetextllm2} but only for single-token words.}
    \label{fig:noiseablate}
\end{figure}

\clearpage

\subsubsection{Ablating The Word Type}
We repeat the same experiments but restricting only to words that are the same part of speech, thus controlling for the type of word. The results are consistent. We only present them for the KNN-10 metric and nouns for compactness.

\begin{figure}[htbp]
    \centering

    \includegraphics[width = .8\linewidth]{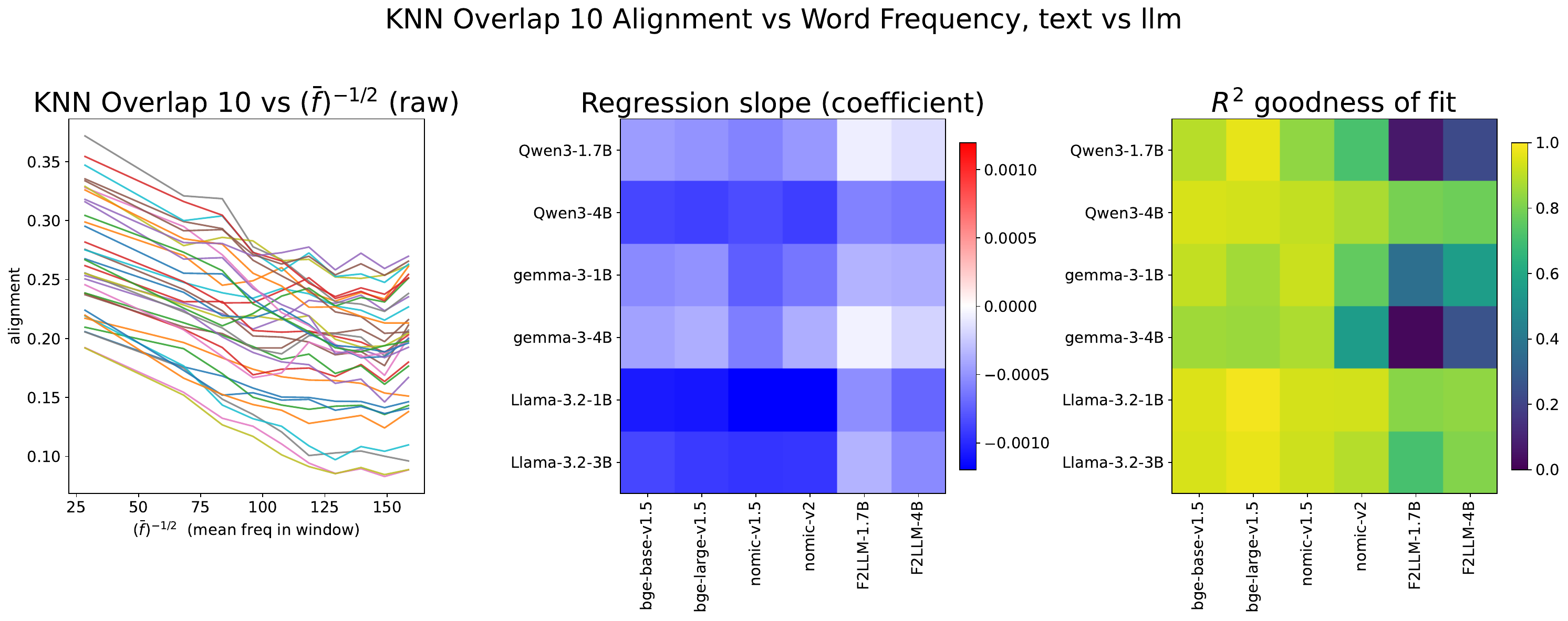}
    
    \vspace{0.4cm}

    \includegraphics[width = .8\linewidth]{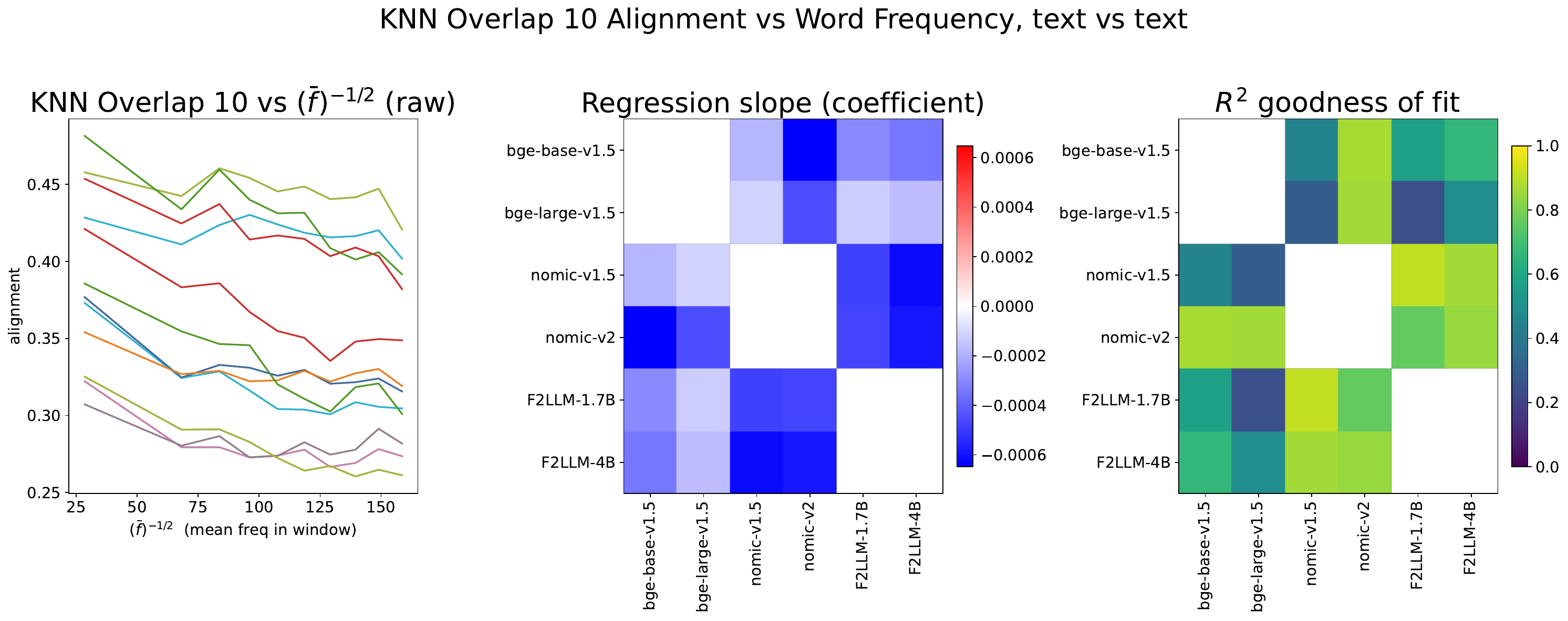}

    \vspace{0.4cm}

    \includegraphics[width = .8\linewidth]{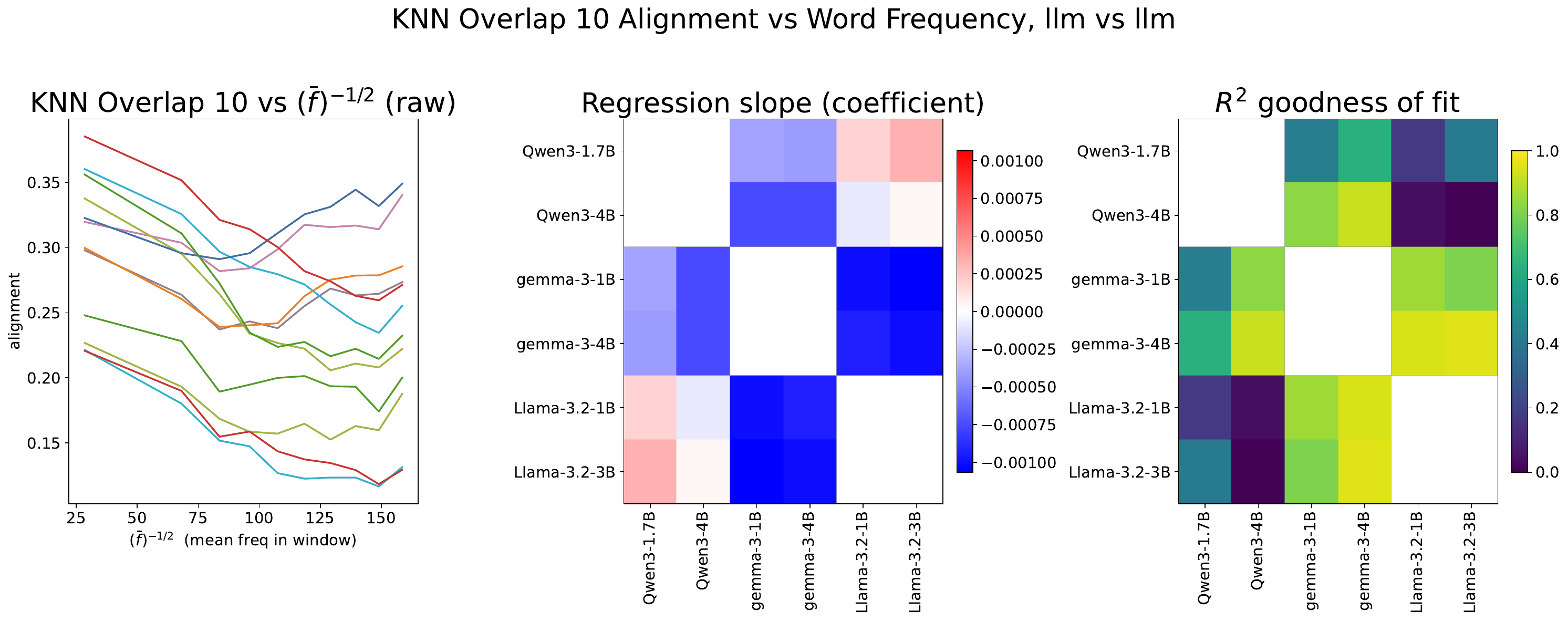}
    
    \caption{Same plot as in Figure~\ref{fig:noisetextllm2} but only for nouns.}
    \label{fig:noisenounablate}
\end{figure}

\clearpage

\subsection{Decomposing the Alignment}
\label{appendix:decomposingthealignment}

\subsubsection{Exact Definition of Features}
\begin{enumerate}
    \item \textsf{min params:} Suppose that the two models $i,j$ have respectively $N_i$ and $N_j$ parameters. Then, first recorded is $\textsf{minparams}'_{ij} = \min(\log N_i,\log N_j).$ Once computed for all models,
    $\textsf{minparams}$ is formed by centering and normalizing to unit norm the vector $(\textsf{minparams}'_{ij})_{i\neq j}.$
    \item \textsf{max params:} Same as above but $\max$ instead of $\min$.
    \item \textsf{min depth:} Suppose that the two models $i,j$ have respectively depths $L_i$ and $L_j.$ Then, first recorded is $\textsf{mindepth}'_{ij} = \min(L_i,L_j).$ Once computed for all models,
    $\textsf{mindepth}$ is formed by centering and normalizing to unit norm the vector $(\textsf{mindepth}'_{ij})_{i\neq j}.$
    \item \textsf{max depth:} Same as above but $\max$ instead of $\min$.
    \item \textsf{min dimension:} Suppose that the two models $i,j$ have respectively dimension $d_i$ and $d_j.$ The dimension is the dimension of the representation, i.e. the output for multimodal and text embedding models. For image models, it is the dimension of the CLS token. In LLMs, it is the width of the respective layer. 
    Then, first recorded is $\textsf{mindimension}'_{ij} = \min(\log d_i,\log d_j).$ Once computed for all models,
    $\textsf{mindimension}$ is formed by centering and normalizing to unit norm the vector $(\textsf{mindimension}'_{ij})_{i\neq j}.$
    \item \textsf{max dimension:} Same as above but $\max$ instead of $\min$.
     \item \textsf{min training images:} Suppose that the two models $i,j$ have been trained respectively on $K_i$ and $K_j$ images. 
    Then, first recorded is $\textsf{minimages}'_{ij} = \min(\log K_i,\log K_j).$ Once computed for all models,
    $\textsf{minimages}$ is formed by centering and normalizing to unit norm the vector $(\textsf{minimages}'_{ij})_{i\neq j}.$
    \item \textsf{max training images:}  Same as above but $\max$ instead of $\min$.
     \item \textsf{min training text tokens:} Suppose that the two models $i,j$ have been trained respectively on $T_i$ and $T_j$ text tokens. 
    Then, first recorded is $\textsf{mintokens}'_{ij} = \min(\log T_i,\log T_j).$ Once computed for all models,
    $\textsf{mintokens}$ is formed by centering and normalizing to unit norm the vector $(\textsf{mintokens}'_{ij})_{i\neq j}.$
    \item \textsf{max training text tokens:}  Same as above but $\max$ instead of $\min$.
    \item \textsf{min year:} Suppose that the two models $i,j$ have been released respectively in years $Y_i$ and $Y_j.$ 
    Then, first recorded is $\textsf{minyear}'_{ij} = \min(Y_i,Y_j).$ Once computed for all models,
    $\textsf{minyear}$ is formed by centering and normalizing to unit norm the vector $(\textsf{minyear}'_{ij})_{i\neq j}.$
    \item \textsf{max year:}  Same as above but $\max$ instead of $\min$.
    \item \textsf{text-text, text-img, img-img:} Those variables one-hot encode the modality of the represented data. Again, they are centered and normalized.  
\end{enumerate}

\clearpage

\subsubsection{Model Specifics}
\label{appendix:featuresofmodels}

\input{tikz/modelspecifics1}

\clearpage

\input{tikz/modelspecifics2}

\clearpage

We repeat the same experiment from Figure~\ref{fig:regressionmain} with varying values of the penalty $\lambda$ and over three different datasets: COCO, CC3M, Visual Genome. The relative importance of features for alignment is very consistent between penalty values and datasets.  
\subsubsection{Experiments on COCO}

\begin{figure}[!htb]
    \centering
    \includegraphics[width=.9\linewidth]{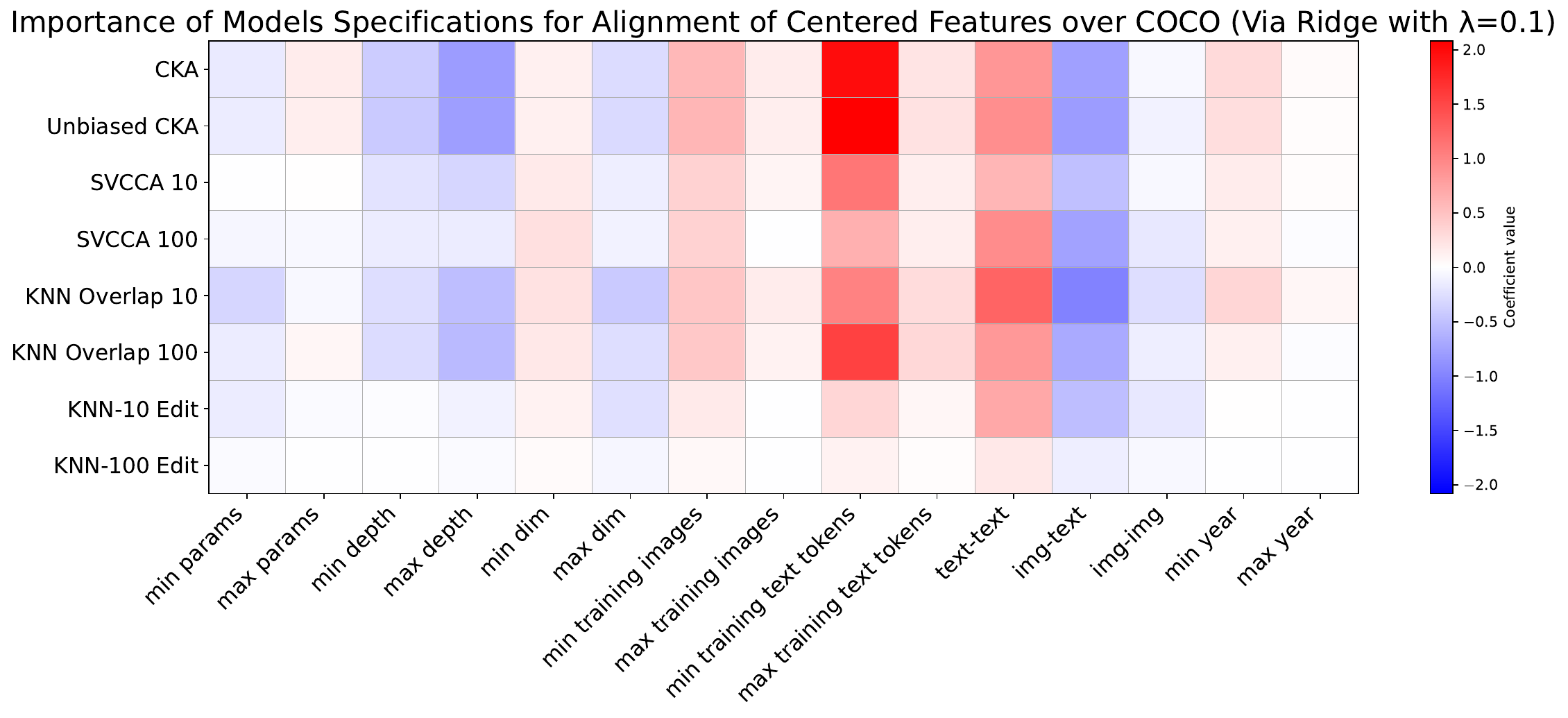}
    \caption*{}
    \label{fig:appendixcococenteredp1}
    \vspace*{-.7cm}
\end{figure}

\begin{figure}[!htb]
    \centering
    \includegraphics[width=.9\linewidth]{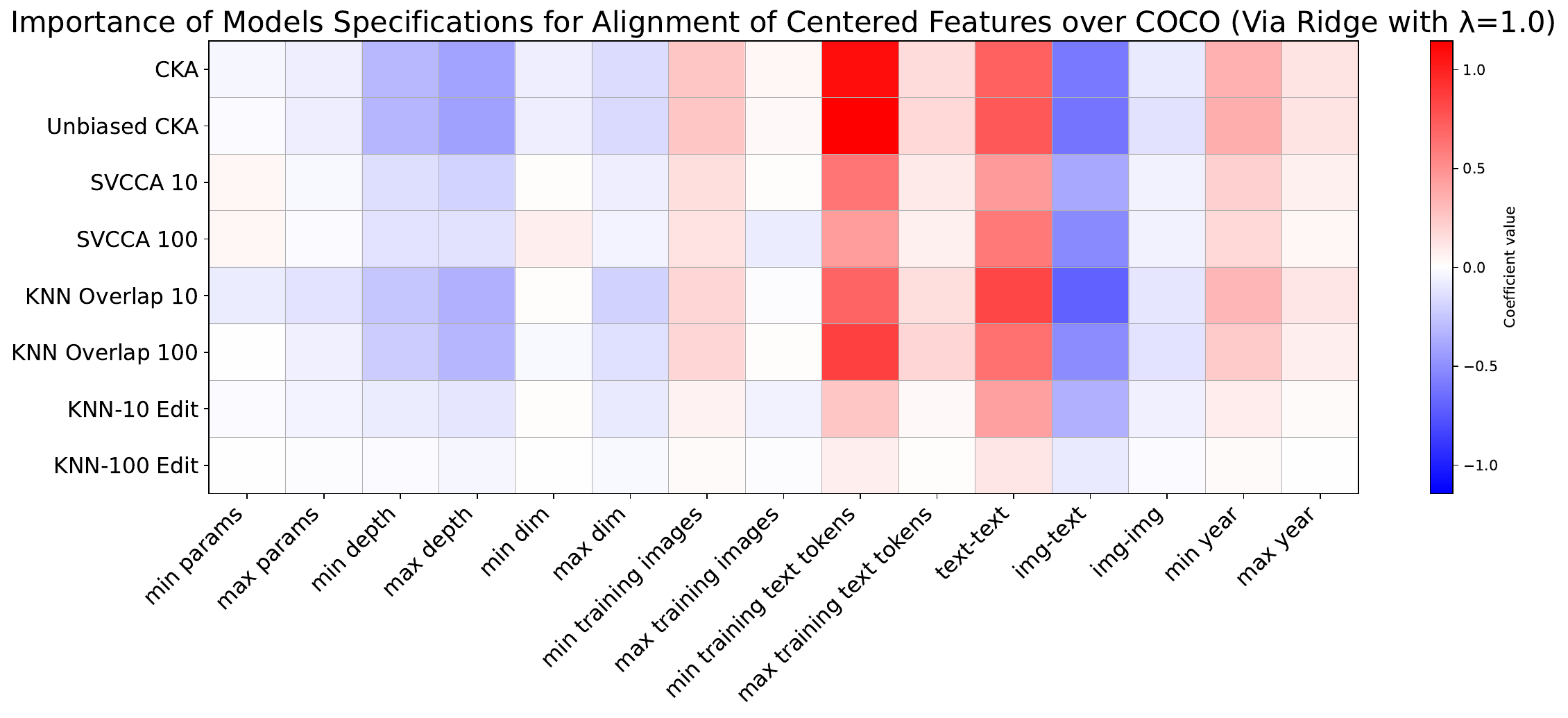}
    \caption*{}
    \label{fig:appendixcococentered1}
    \vspace*{-.7cm}
\end{figure}

\begin{figure}[!htb]
    \centering
    \includegraphics[width=.9\linewidth]{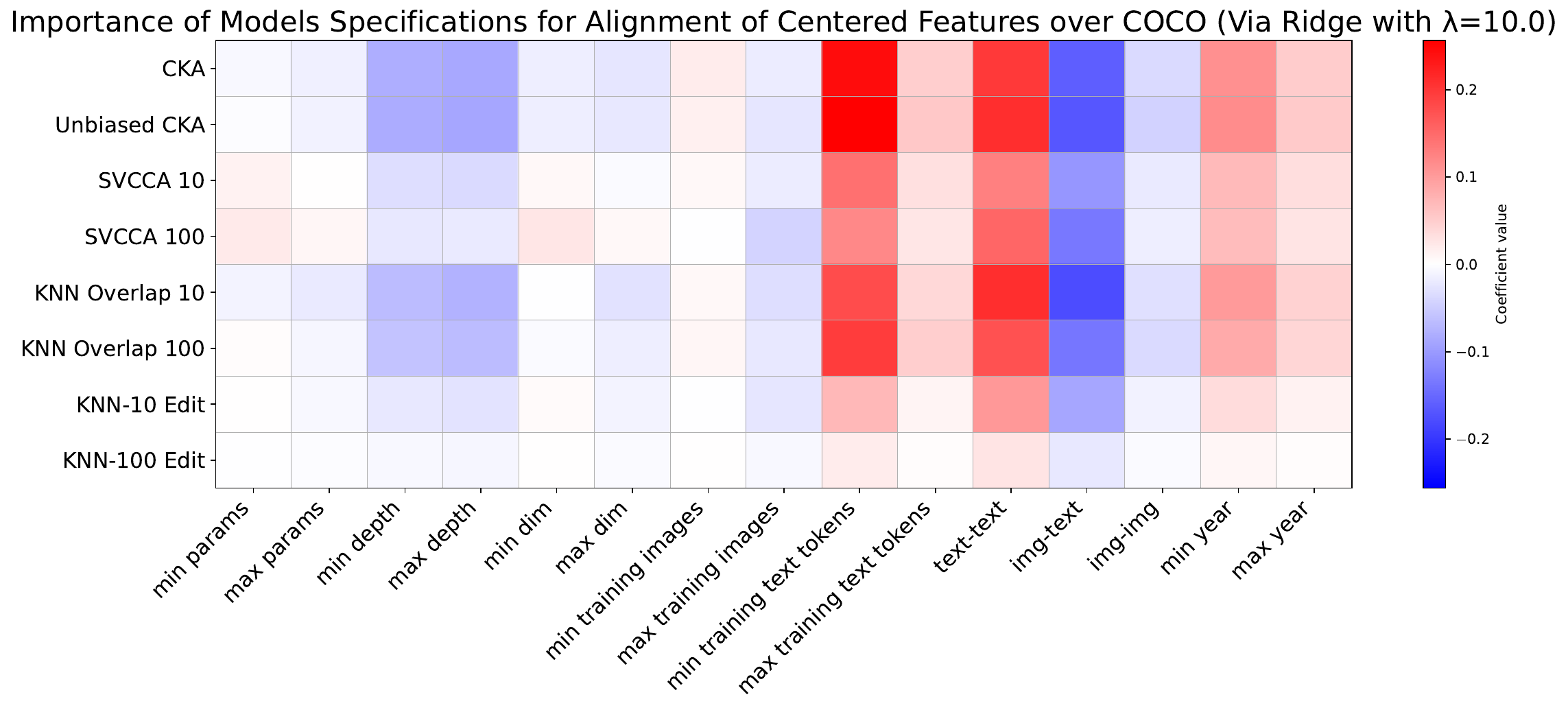}
    \caption*{}
    \label{fig:appendixcococentered10}
\end{figure}

\clearpage

\subsubsection{Experiments on CC3M}

\begin{figure}[!htb]
    \centering
    \includegraphics[width=.9\linewidth]{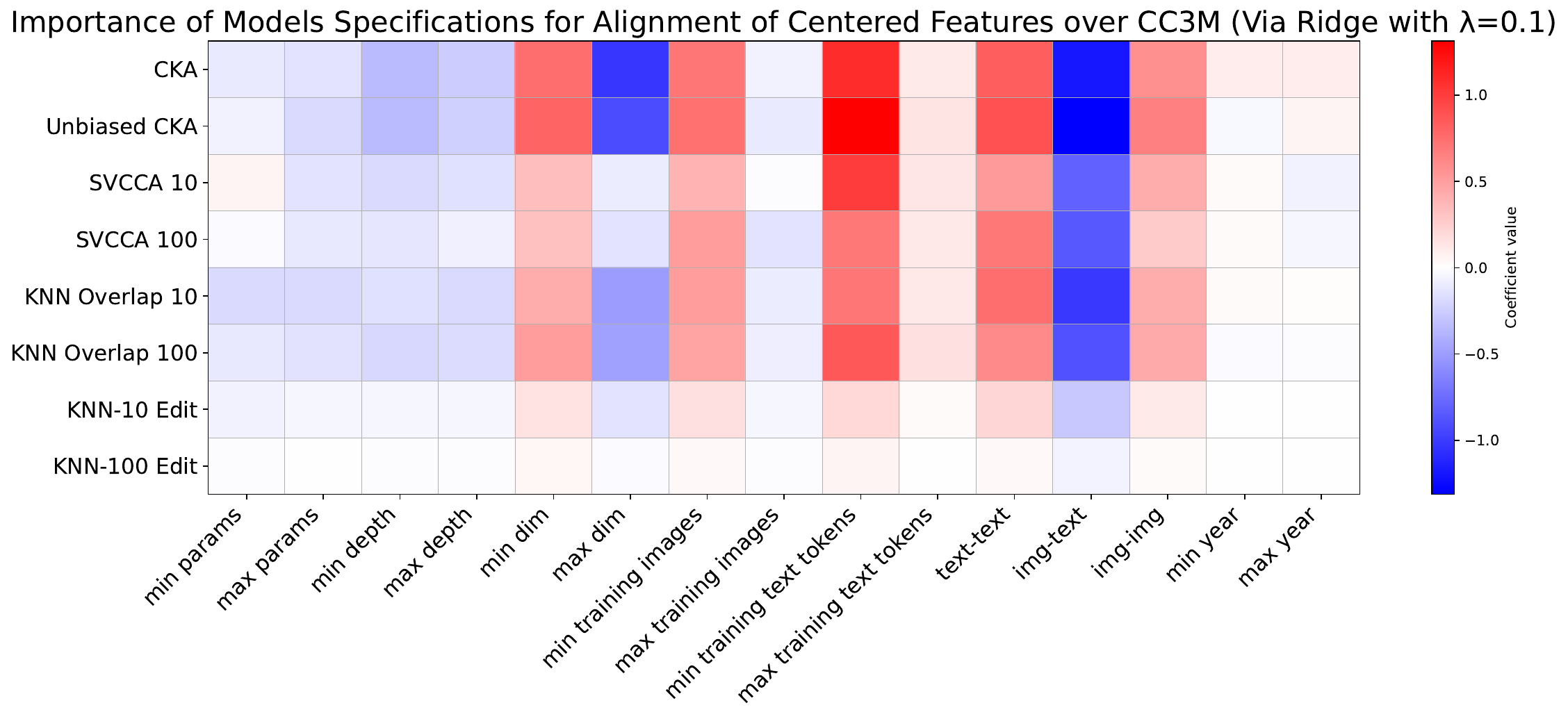}
    \caption*{}
    \label{fig:appendixcc3mcenteredp1}
\end{figure}

\begin{figure}[!htb]
    \centering
    \includegraphics[width=.9\linewidth]{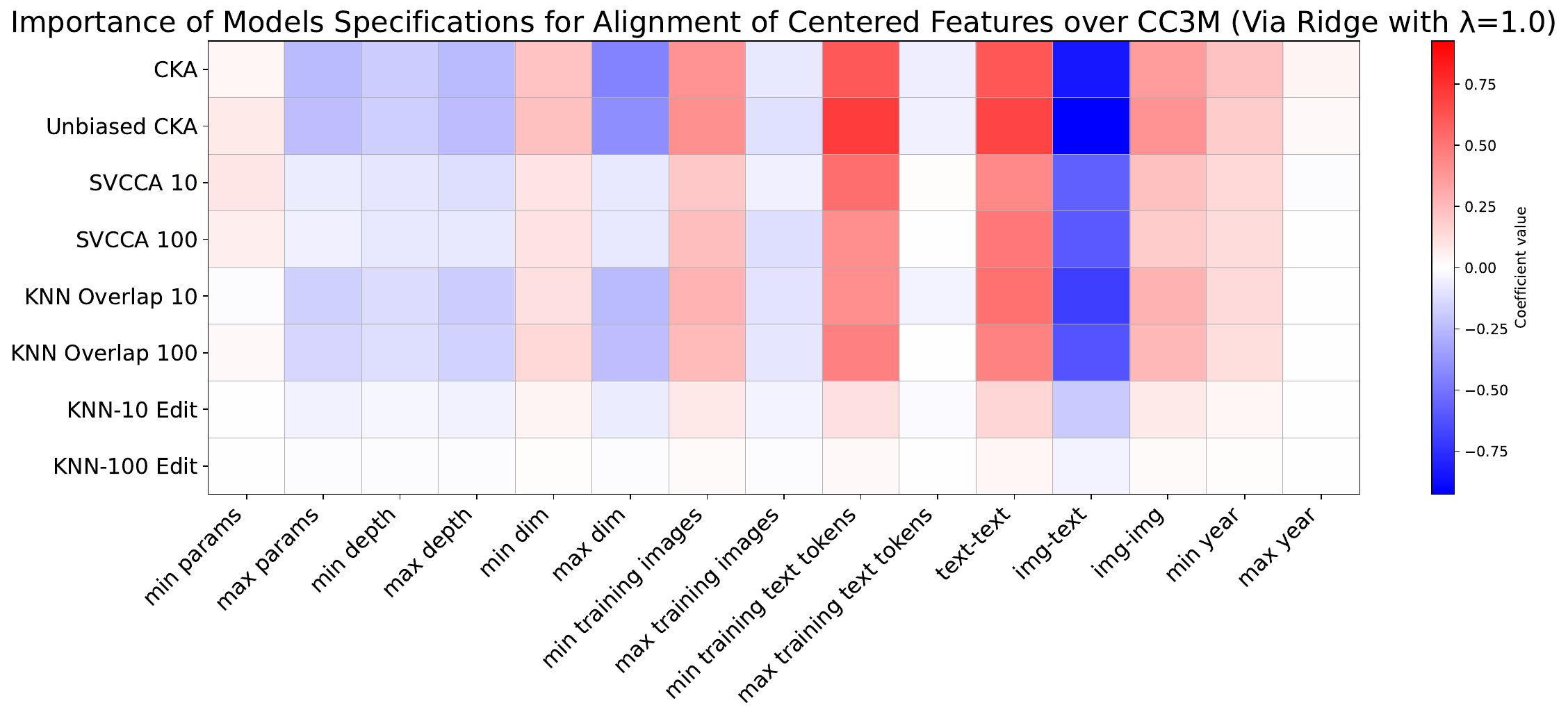}
    \caption*{}
    \label{fig:appendixcc3mcentered1}
\end{figure}

\begin{figure}[!htb]
    \centering
    \includegraphics[width=.9\linewidth]{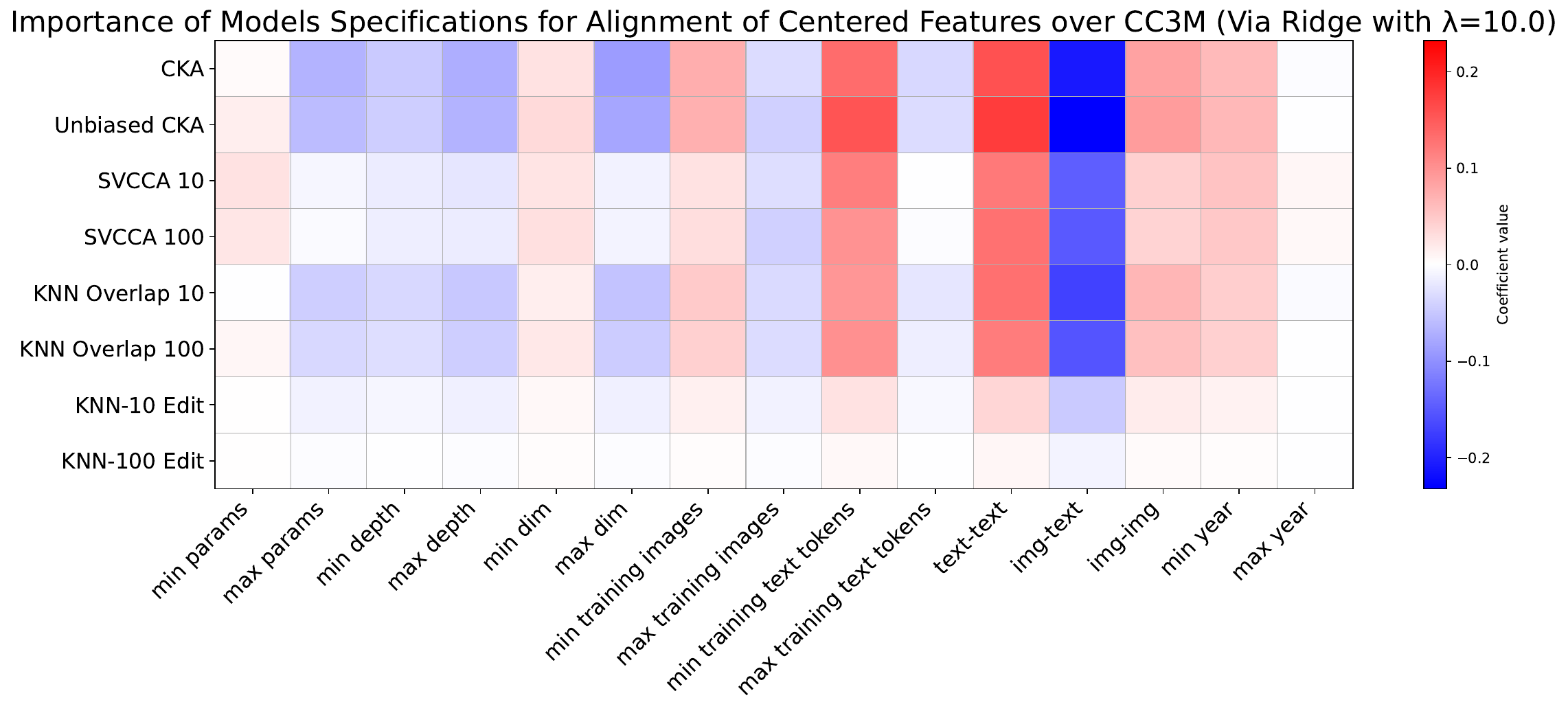}
    \caption*{}
    \label{fig:appendixcc3mcentered10}
\end{figure}

\clearpage

\subsubsection{Experiments on Visual Genome}

\begin{figure}[!htb]
    \centering
    \includegraphics[width=.9\linewidth]{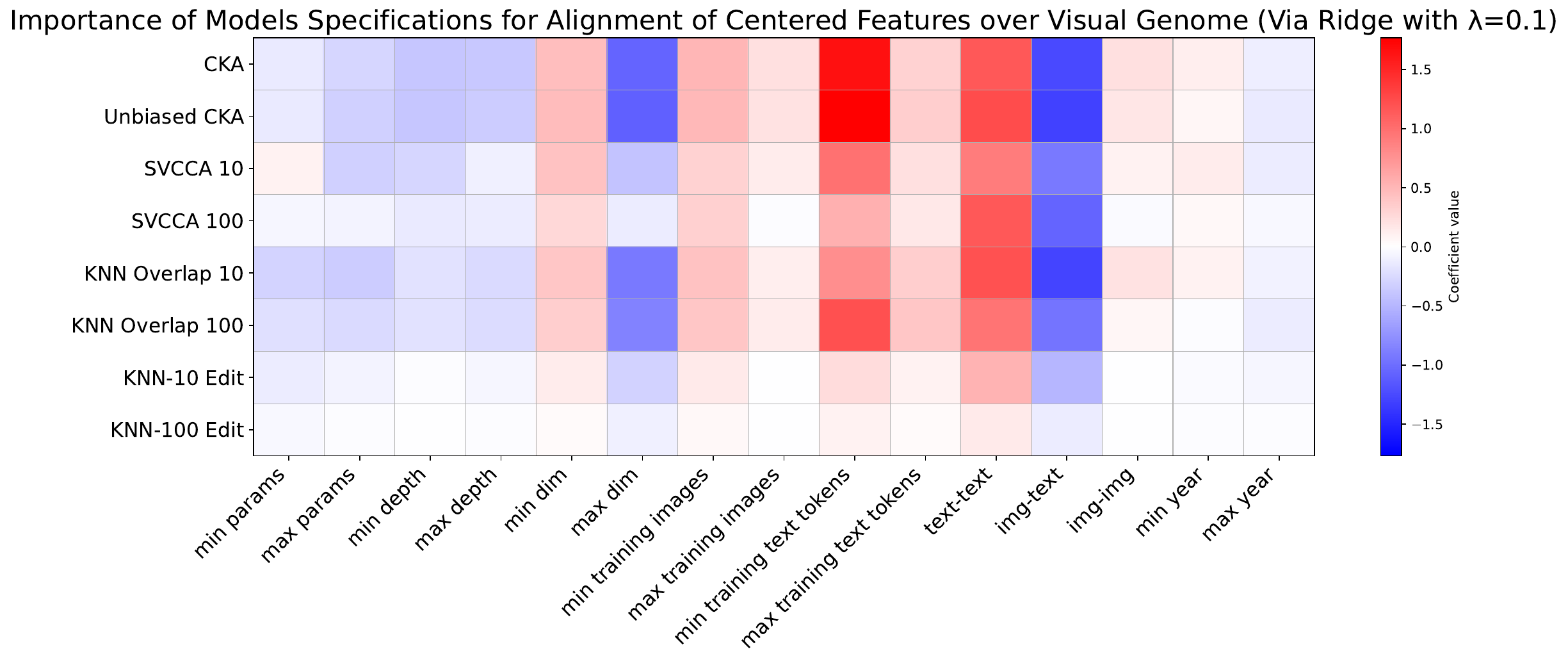}
    \caption*{}
    \label{fig:appendixvgcenteredp1}
\end{figure}

\begin{figure}[!htb]
    \centering
    \includegraphics[width=.9\linewidth]{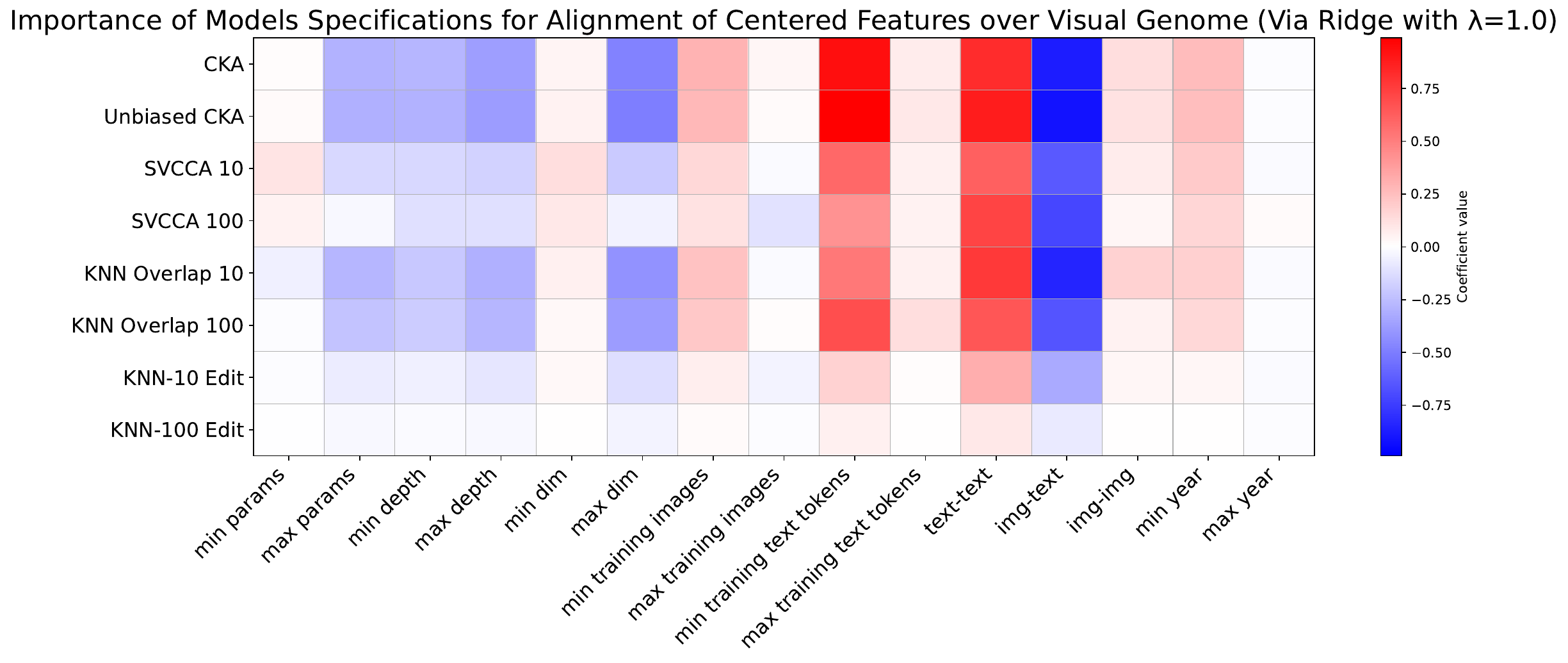}
    \caption*{}
    \label{fig:appendixvgcentered1}
\end{figure}

\begin{figure}[!htb]
    \centering
    \includegraphics[width=.9\linewidth]{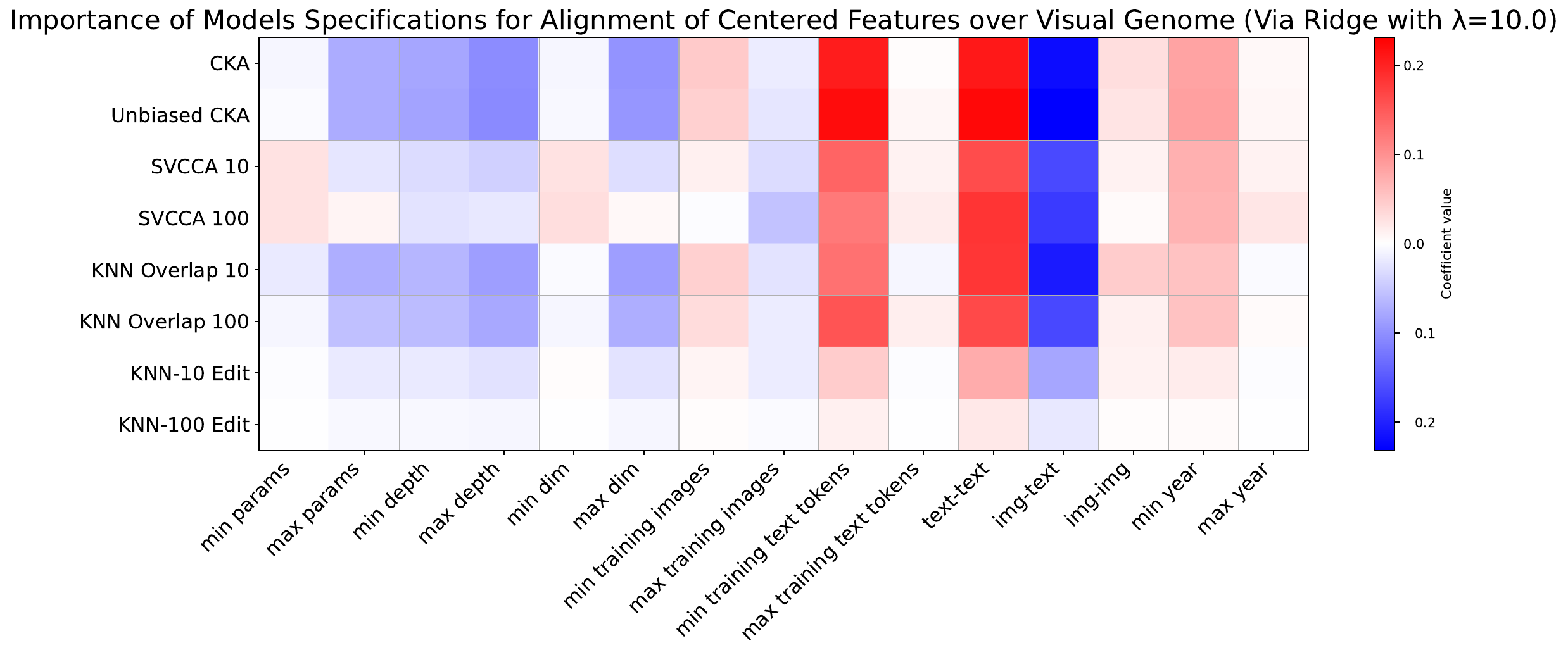}
    \caption*{}
    \label{fig:appendixvgcentered10}
\end{figure}

\clearpage
\subsection{Dimension and Incoherence}
\label{appendix:dimensionandincoherence}

\subsubsection{Experiments on COCO}
\begin{figure}[htbp]
    \centering

    \begin{minipage}[t]{0.48\textwidth}
        \centering
        \includegraphics[width = .8\linewidth]{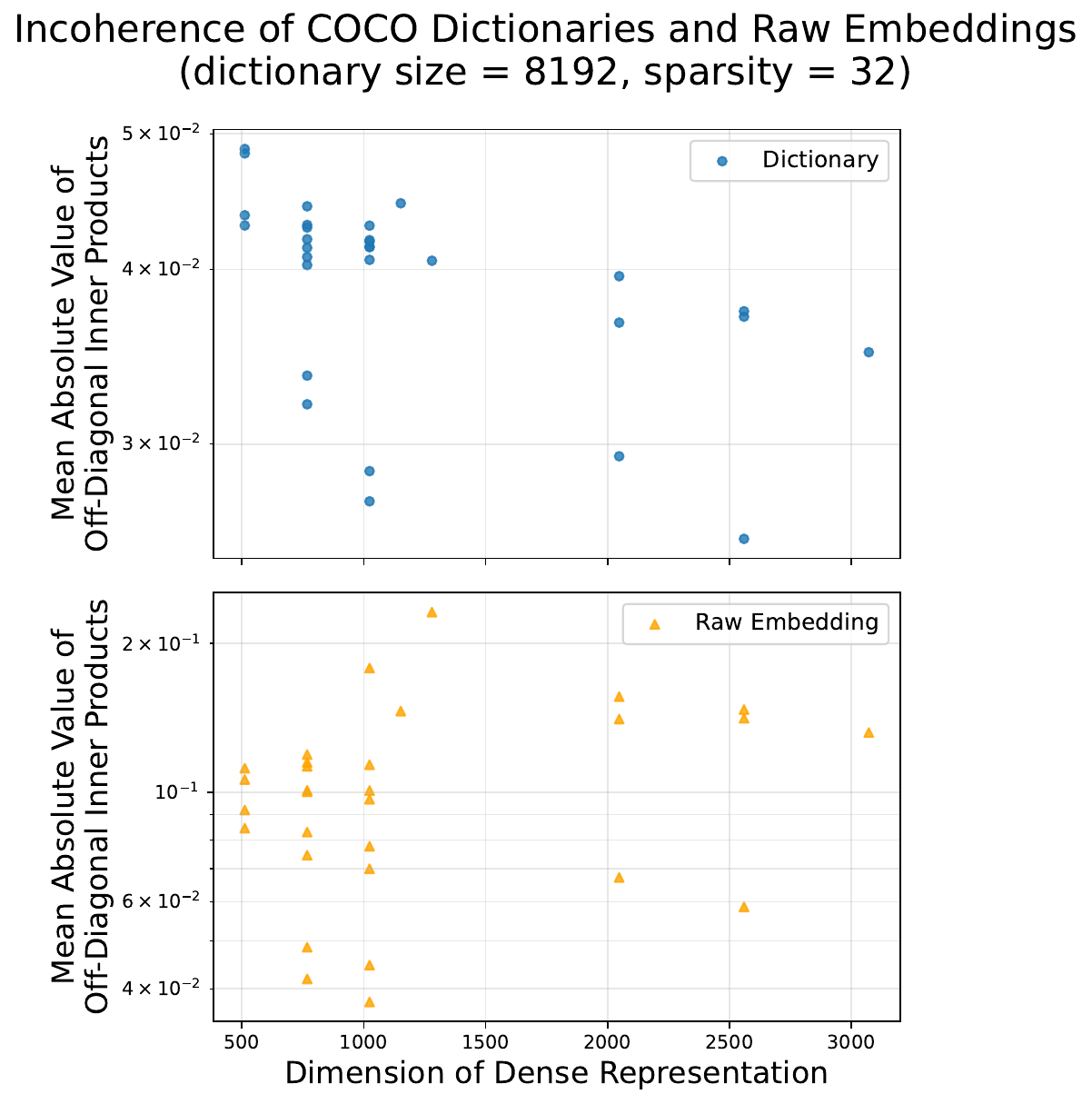}
    \end{minipage}
    \hfill
    \begin{minipage}[t]{0.48\textwidth}
        \centering
        \includegraphics[width = .8\linewidth]{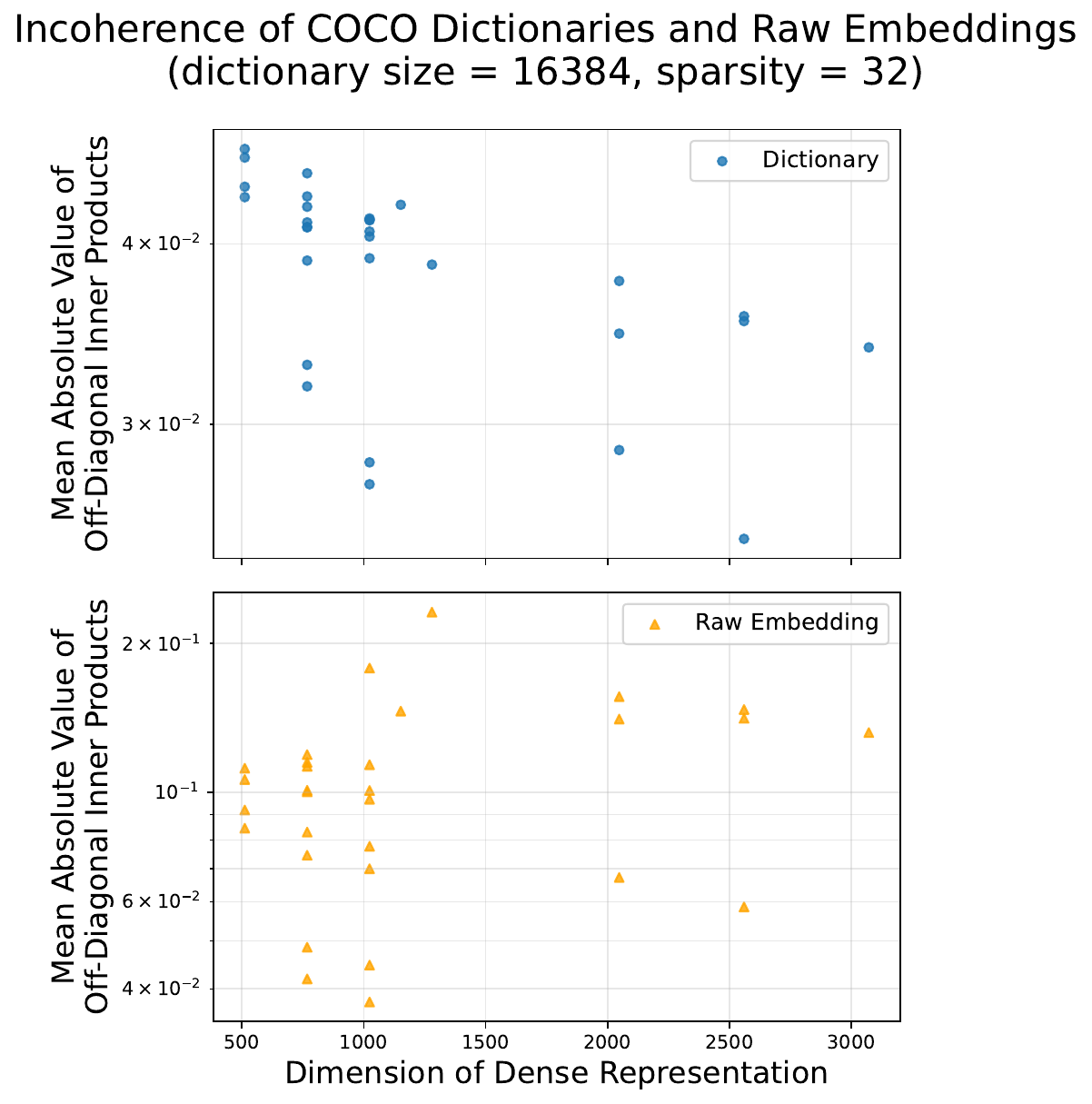}
    \end{minipage}

    \vspace{0.4cm}

    \begin{minipage}[t]{0.48\textwidth}
        \centering
        \includegraphics[width = .8\linewidth]{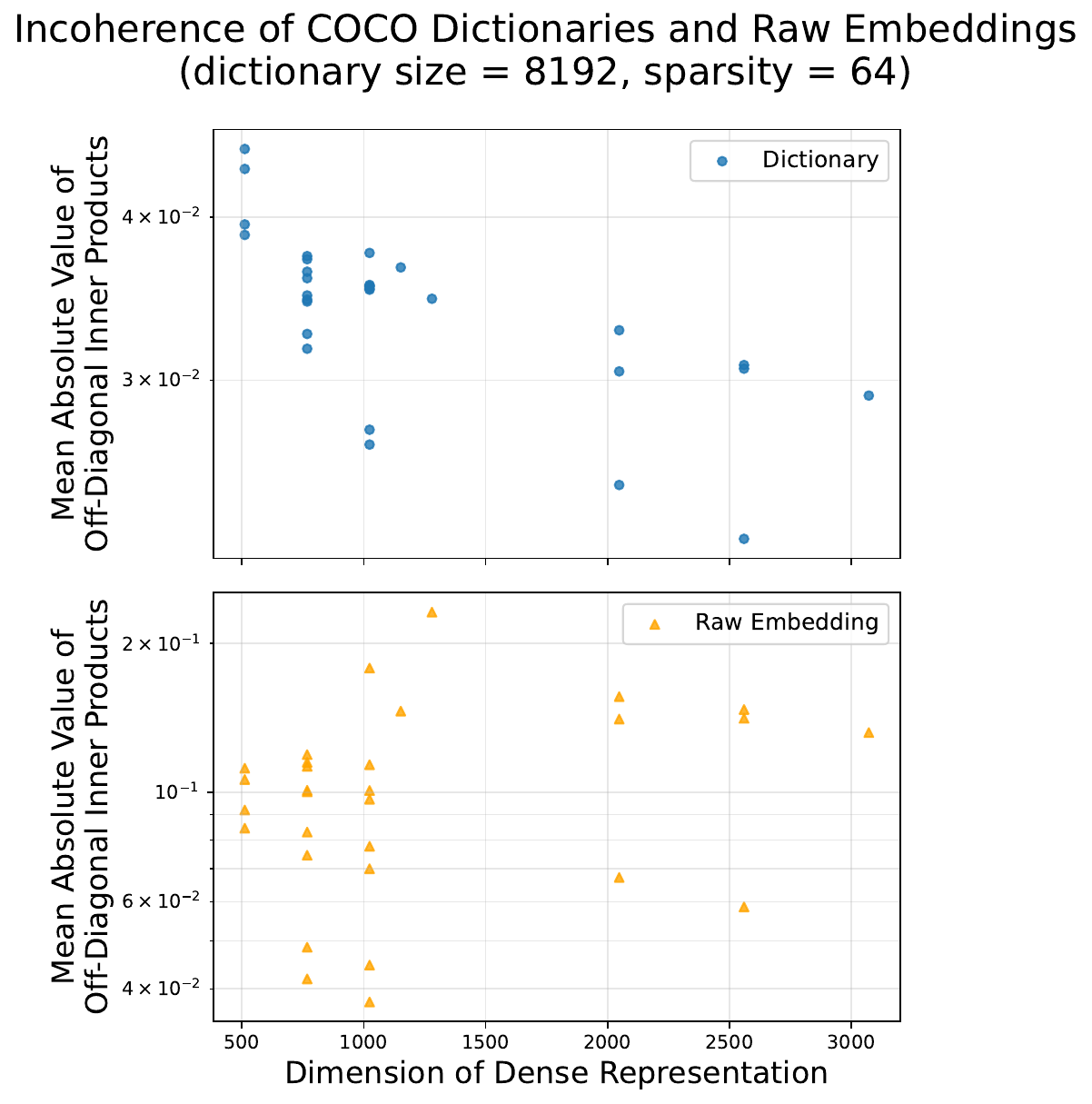}
    \end{minipage}
    \hfill
    \begin{minipage}[t]{0.48\textwidth}
        \centering
        \includegraphics[width = .8\linewidth]{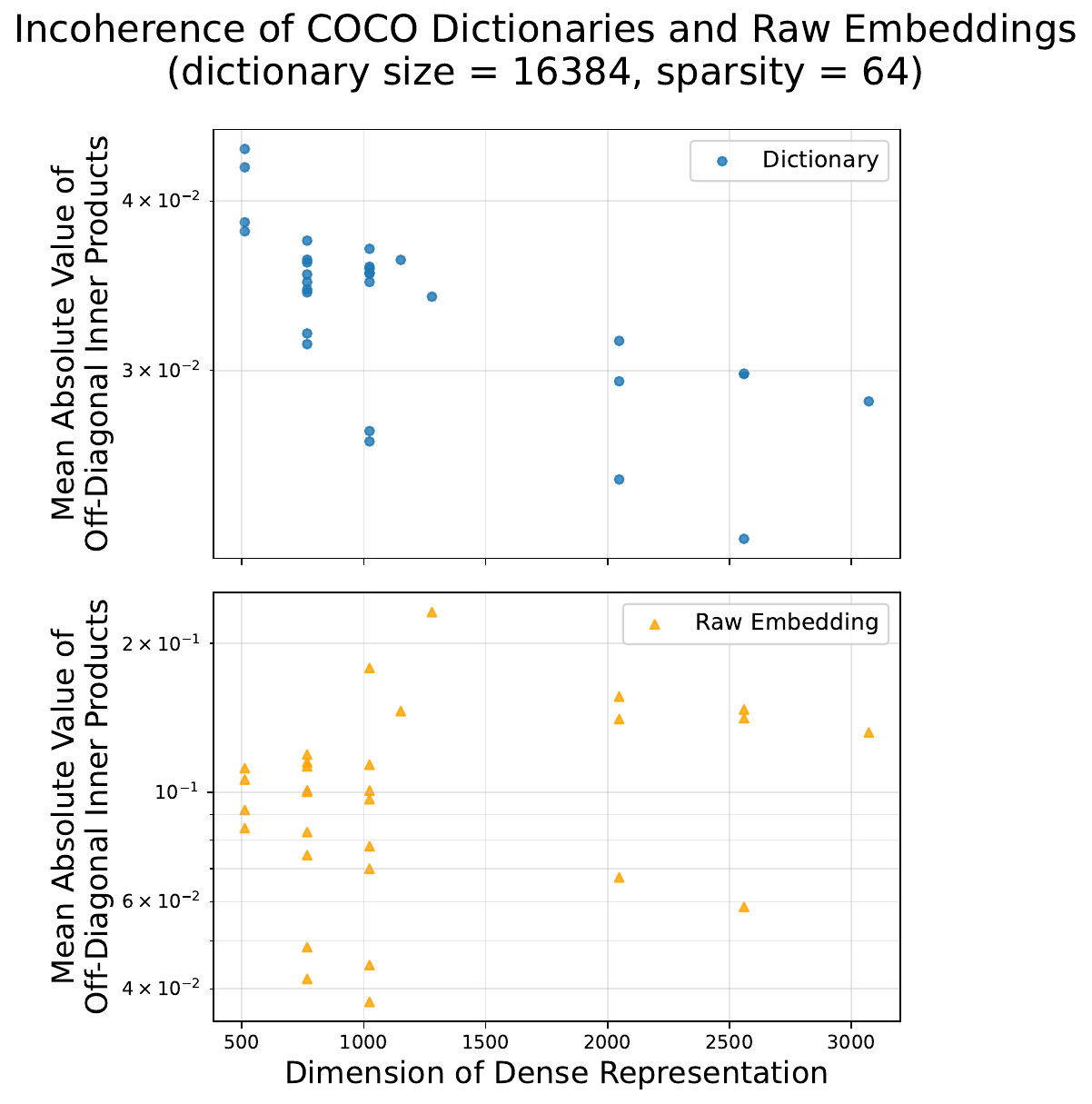}
    \end{minipage}

    \vspace{0.4cm}

    \begin{minipage}[t]{0.48\textwidth}
        \centering
        \includegraphics[width = .8\linewidth]{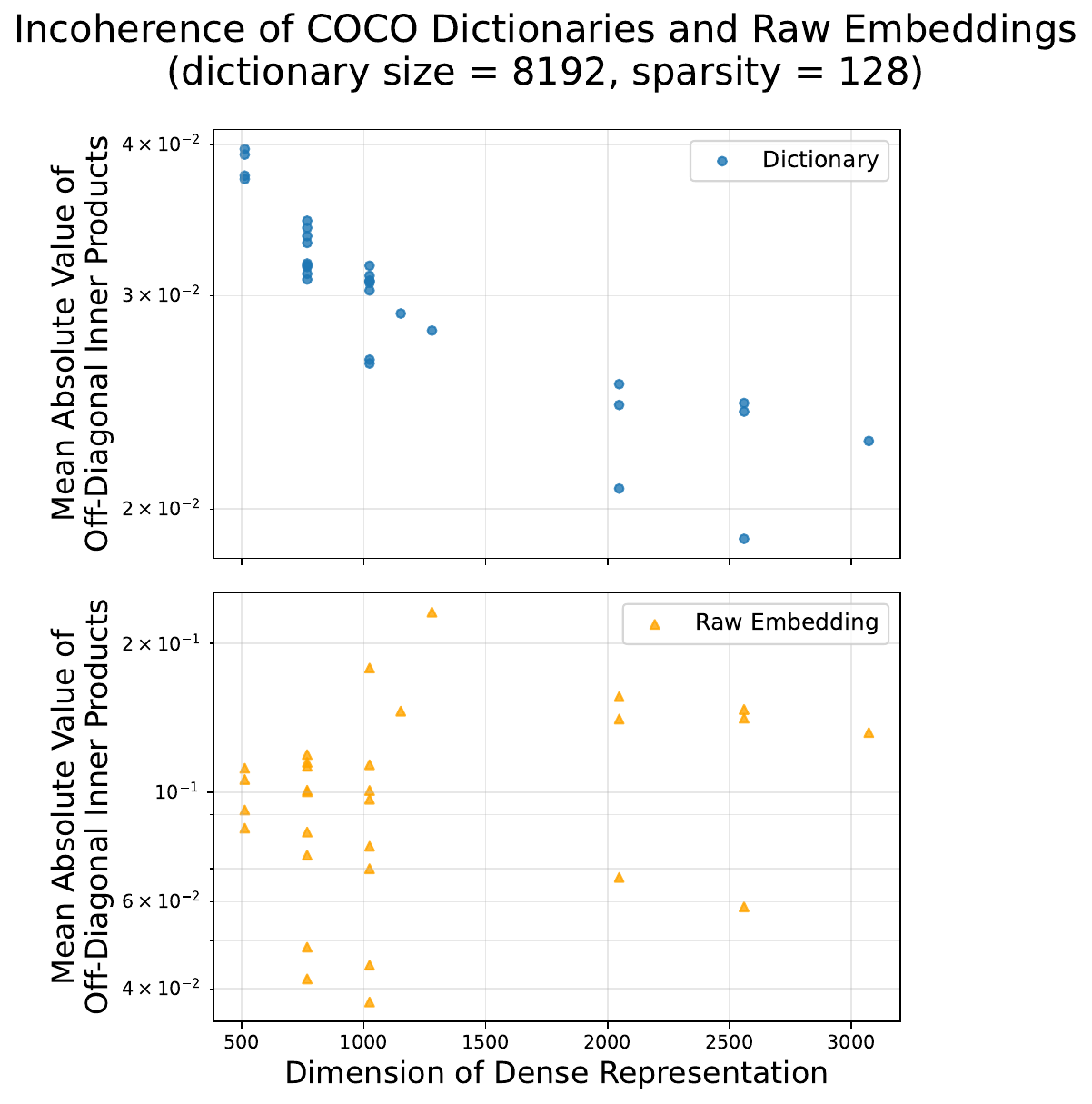}
    \end{minipage}
    \hfill
    \begin{minipage}[t]{0.48\textwidth}
        \centering
        \includegraphics[width = .8\linewidth]{incoherence_diagrams/truncated_incoherence_coco_16384_128.pdf}
    \end{minipage}

    \caption{Plotted is the mean absolute value of inner products of distinct features from SAEs trained for each model in Table~\ref{table:modelspecs_architecture}. The SAEs are trained as in Section~\ref{sec:saepseudo} with truncated features. We can see that the trend that higher dimensionality leads to smaller $\epsilon_{\mathrm{dict}}$ is consistent across SAE parameters, although it is higher for higher sparsities.}
    \label{fig:appendixincoherencecoco}
\end{figure}

\clearpage

\subsubsection{Experiments on CC3M}
\begin{figure}[htbp]
    \centering

    \begin{minipage}[t]{0.48\textwidth}
        \centering
        \includegraphics[width = .8\linewidth]{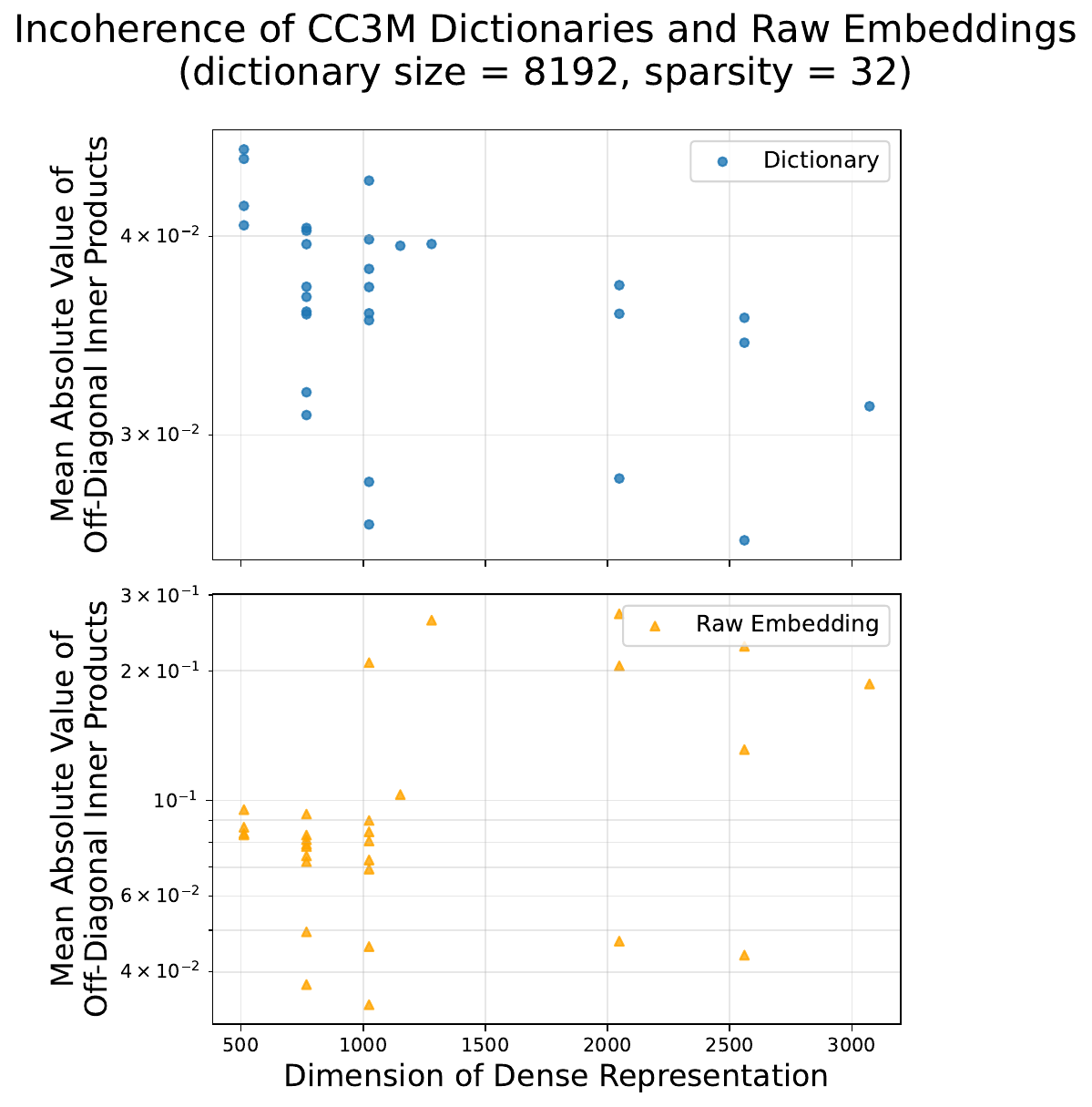}
    \end{minipage}
    \hfill
    \begin{minipage}[t]{0.48\textwidth}
        \centering
        \includegraphics[width = .8\linewidth]{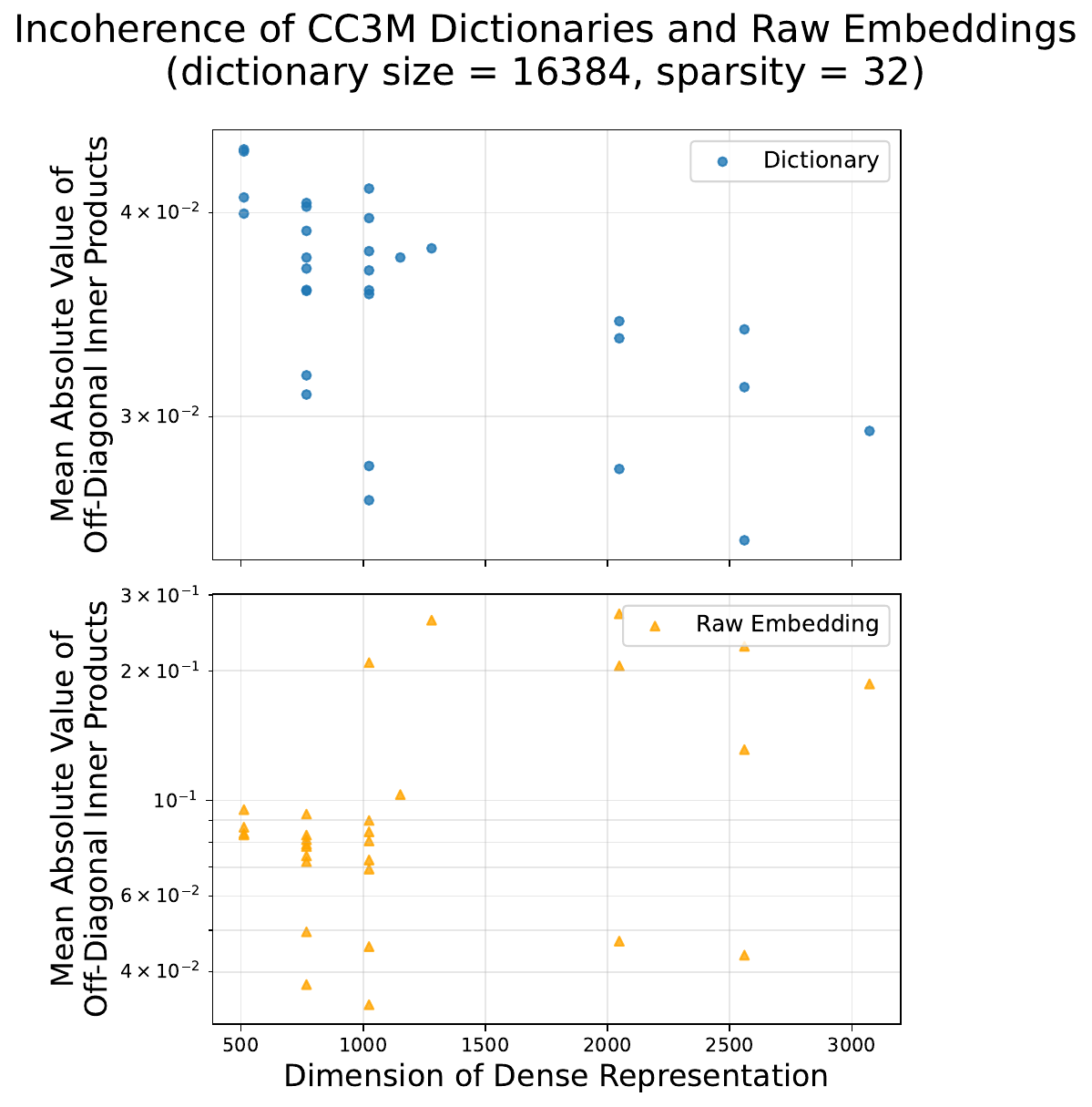}
    \end{minipage}

    \vspace{0.4cm}

    \begin{minipage}[t]{0.48\textwidth}
        \centering
        \includegraphics[width = .8\linewidth]{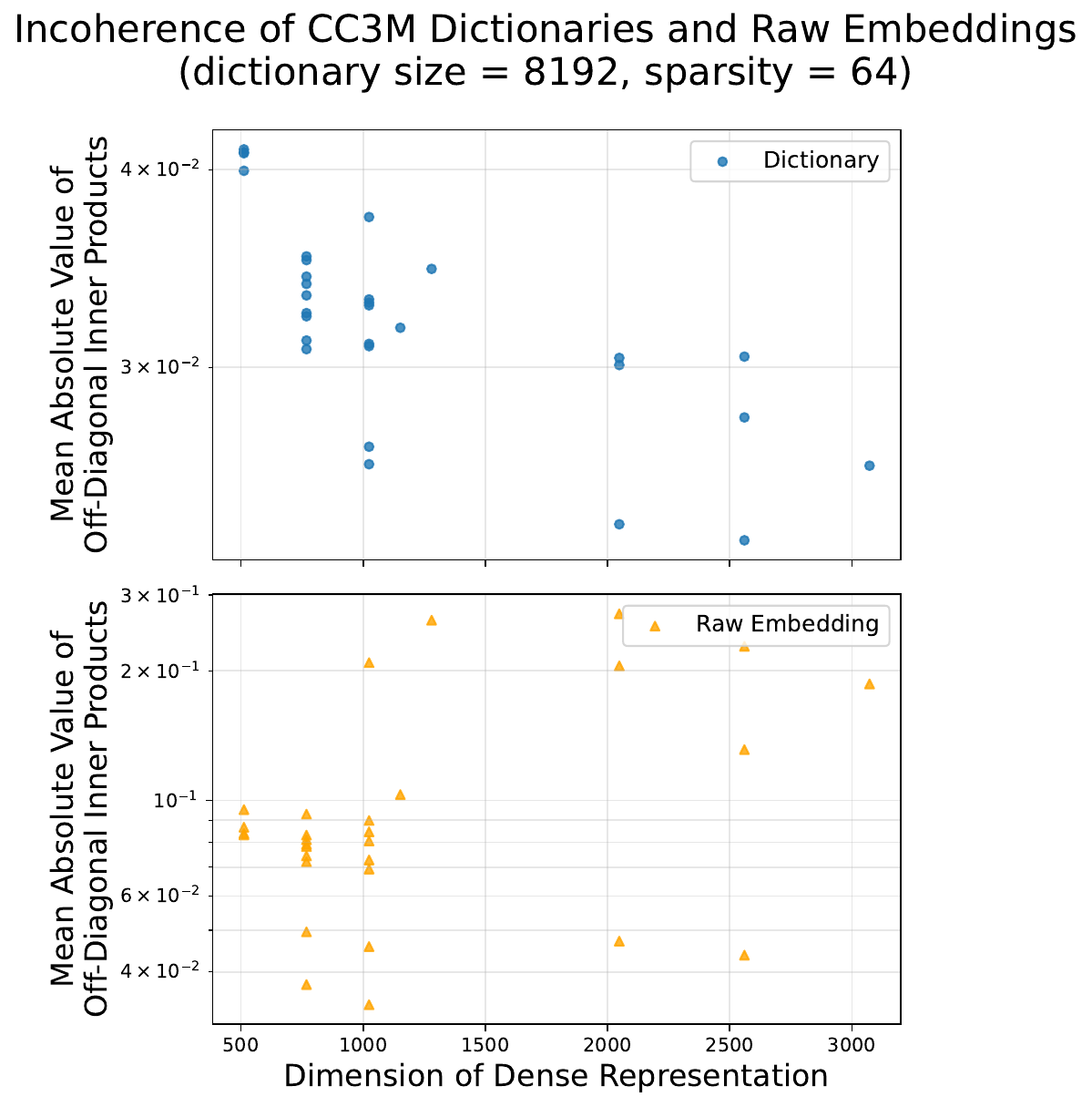}
    \end{minipage}
    \hfill
    \begin{minipage}[t]{0.48\textwidth}
        \centering
        \includegraphics[width = .8\linewidth]{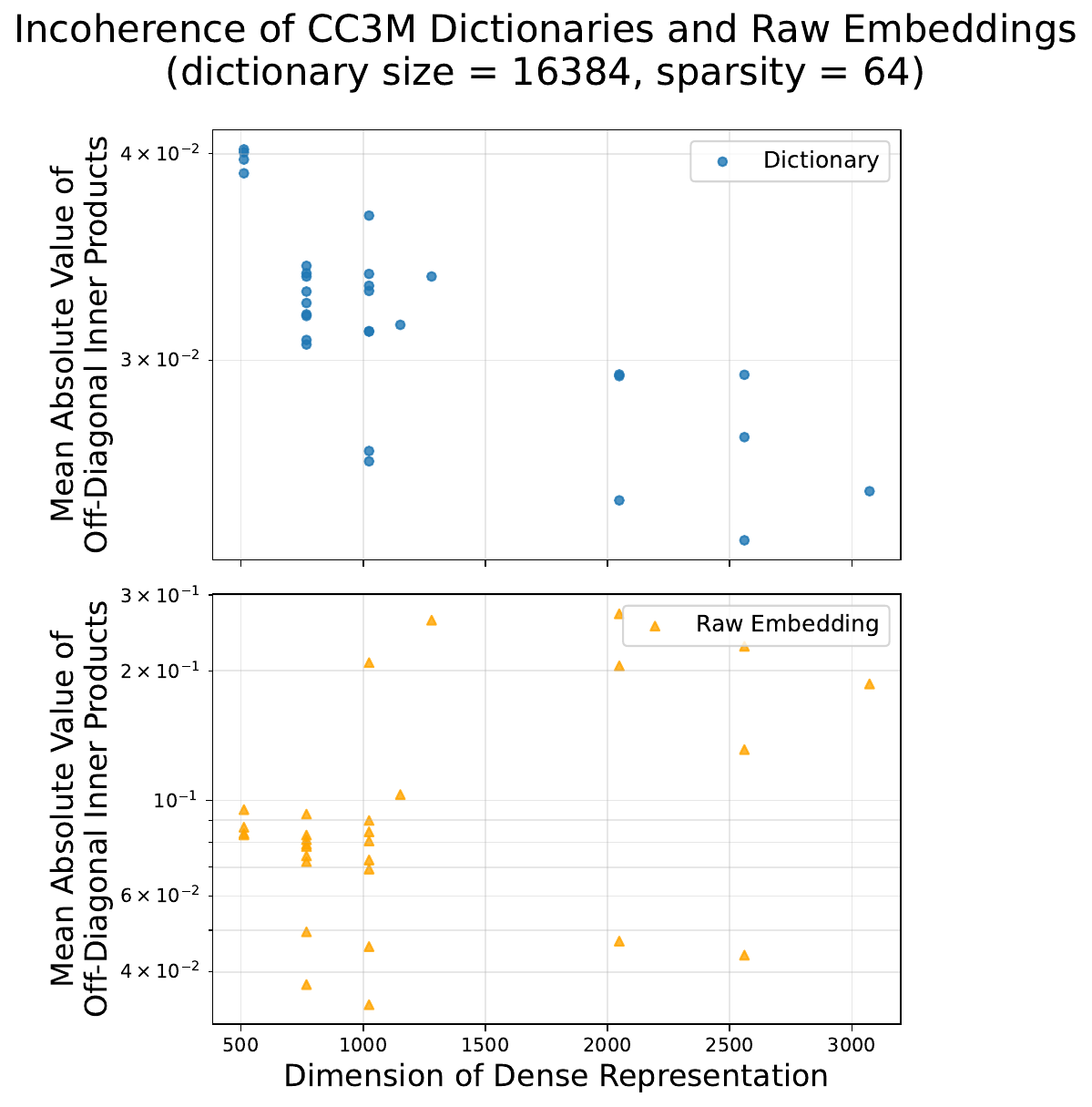}
    \end{minipage}

    \vspace{0.4cm}

    \begin{minipage}[t]{0.48\textwidth}
        \centering
        \includegraphics[width = .8\linewidth]{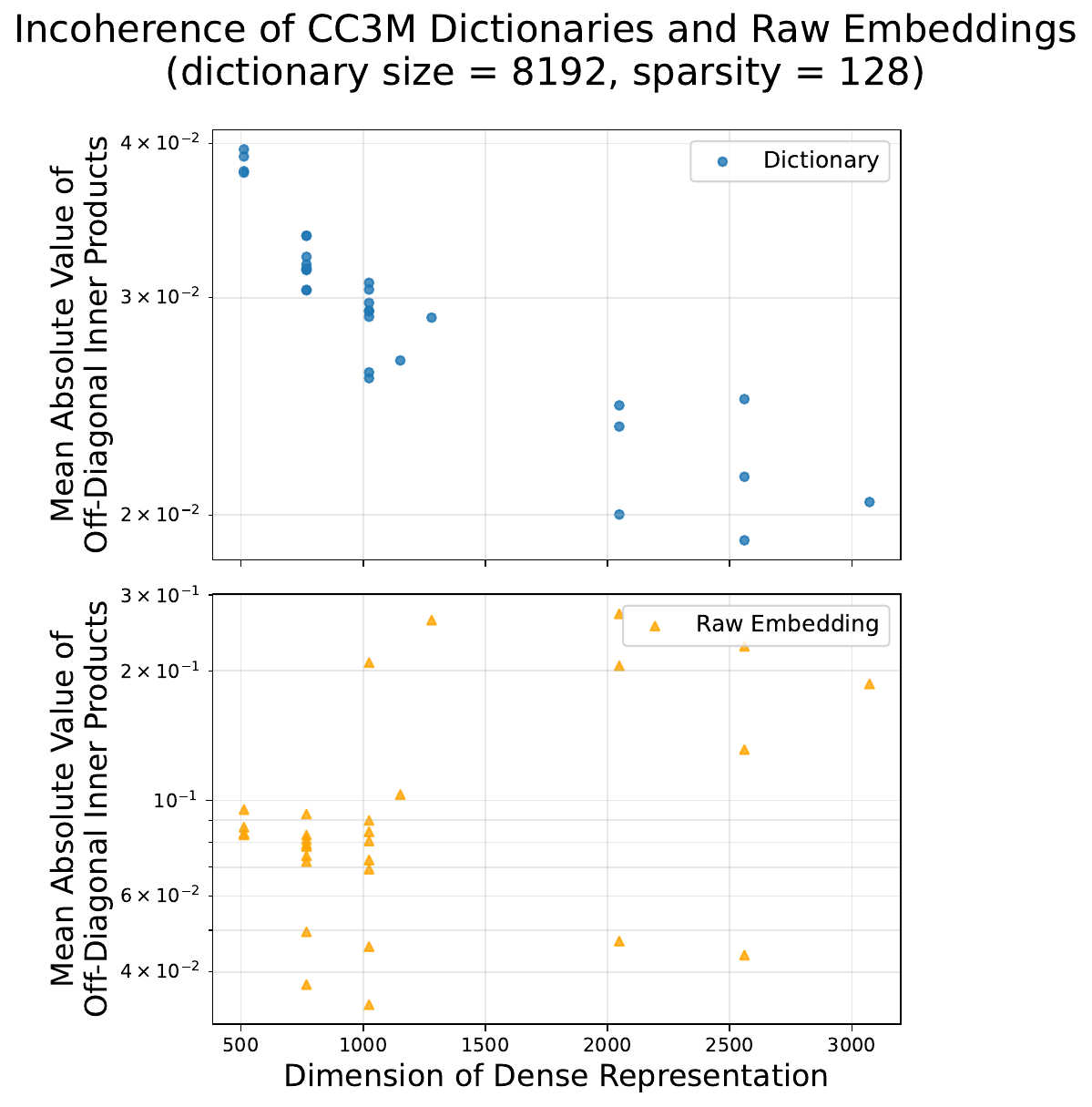}
    \end{minipage}
    \hfill
    \begin{minipage}[t]{0.48\textwidth}
        \centering
        \includegraphics[width = .8\linewidth]{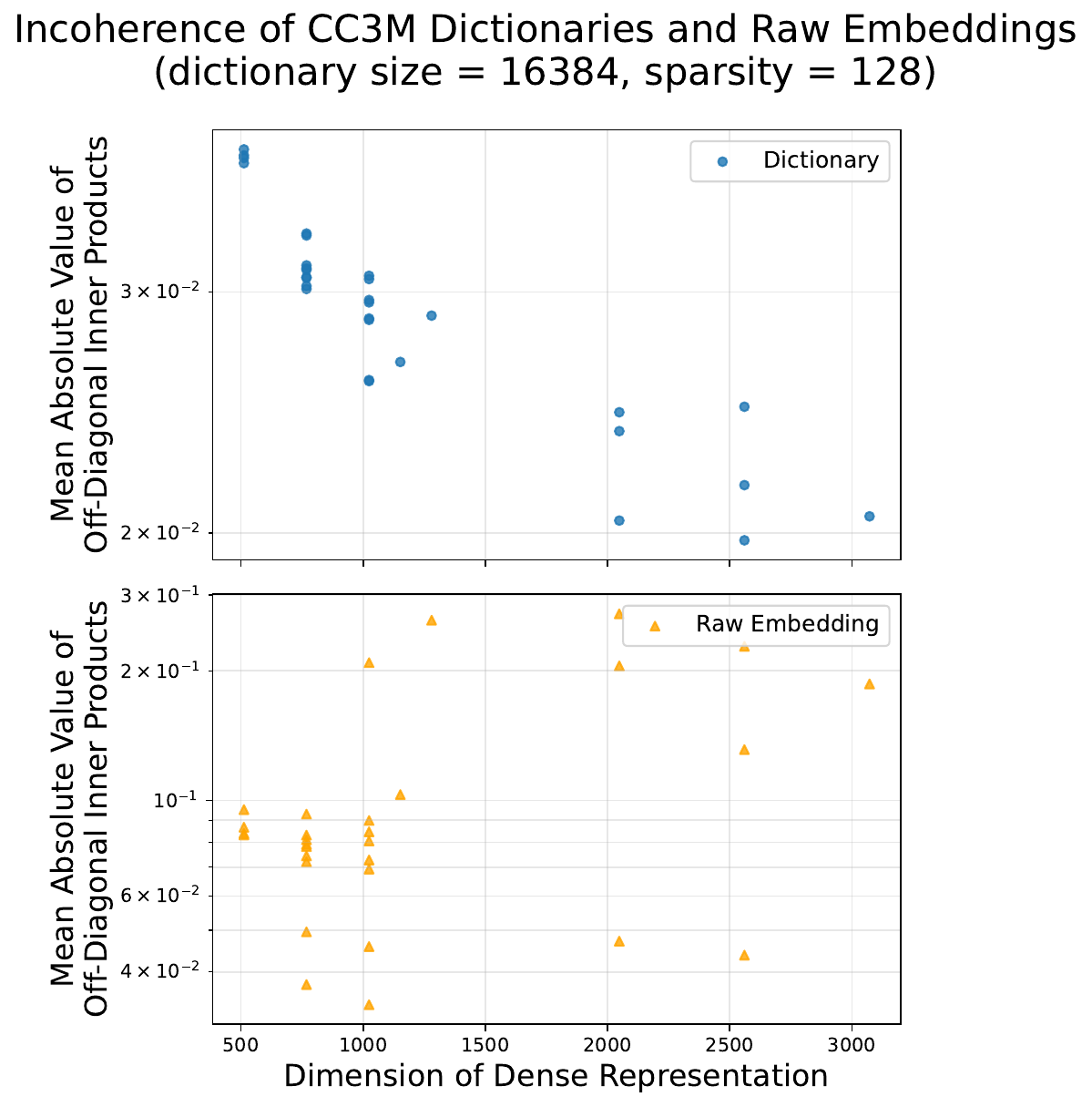}
    \end{minipage}

    \caption{The same figure as Figure~\ref{fig:appendixincoherencecoco} but over the CC3M dataset. Qualitatively, all trends are similar.}
    \label{fig:appendixincoherencecc3m}
\end{figure}

\clearpage
\subsubsection{Experiments on Visual Genome}

\begin{figure}[htbp]
    \centering

    \begin{minipage}[t]{0.48\textwidth}
        \centering
        \includegraphics[width = .8\linewidth]{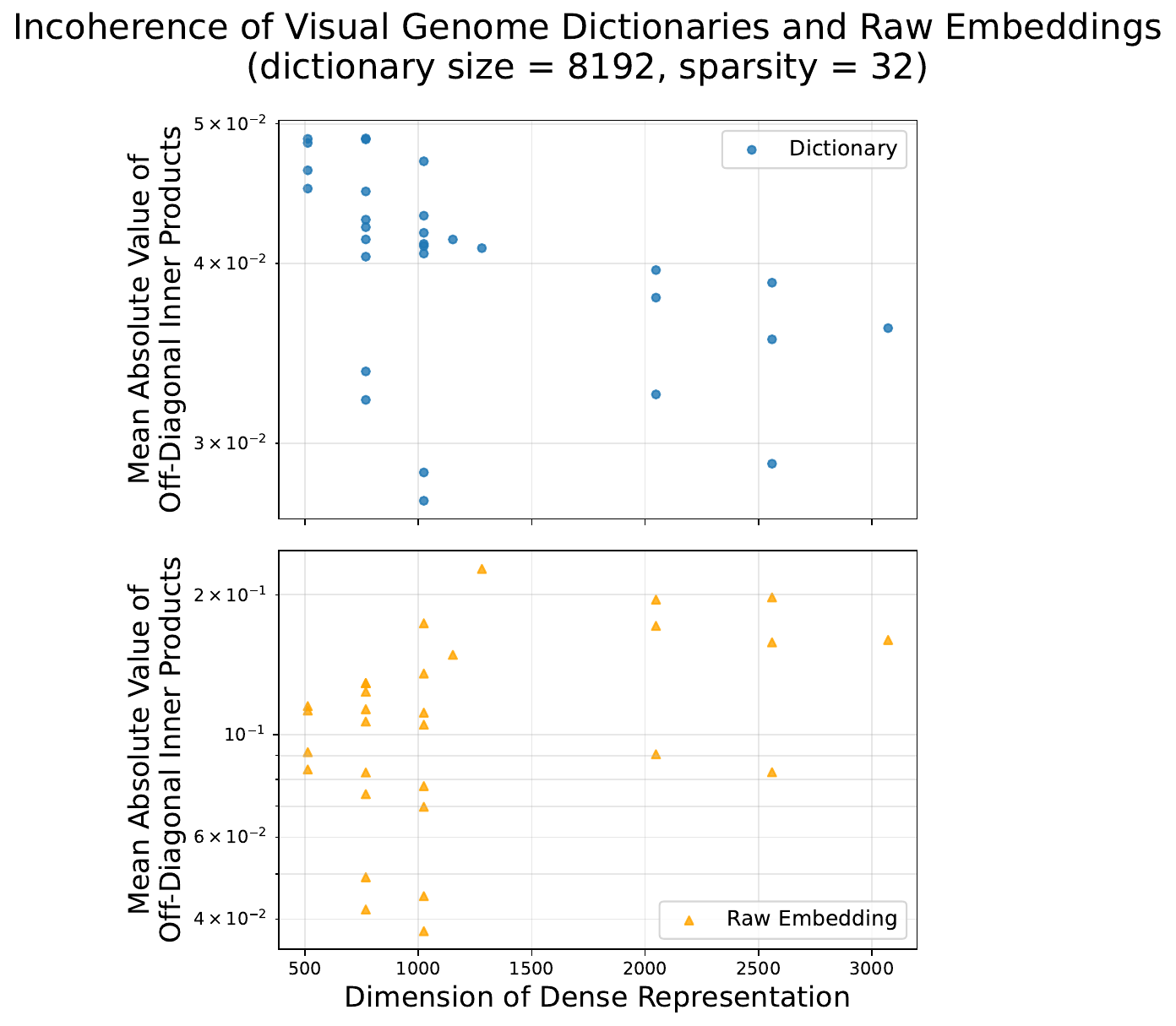}
    \end{minipage}
    \hfill
    \begin{minipage}[t]{0.48\textwidth}
        \centering
        \includegraphics[width = .8\linewidth]{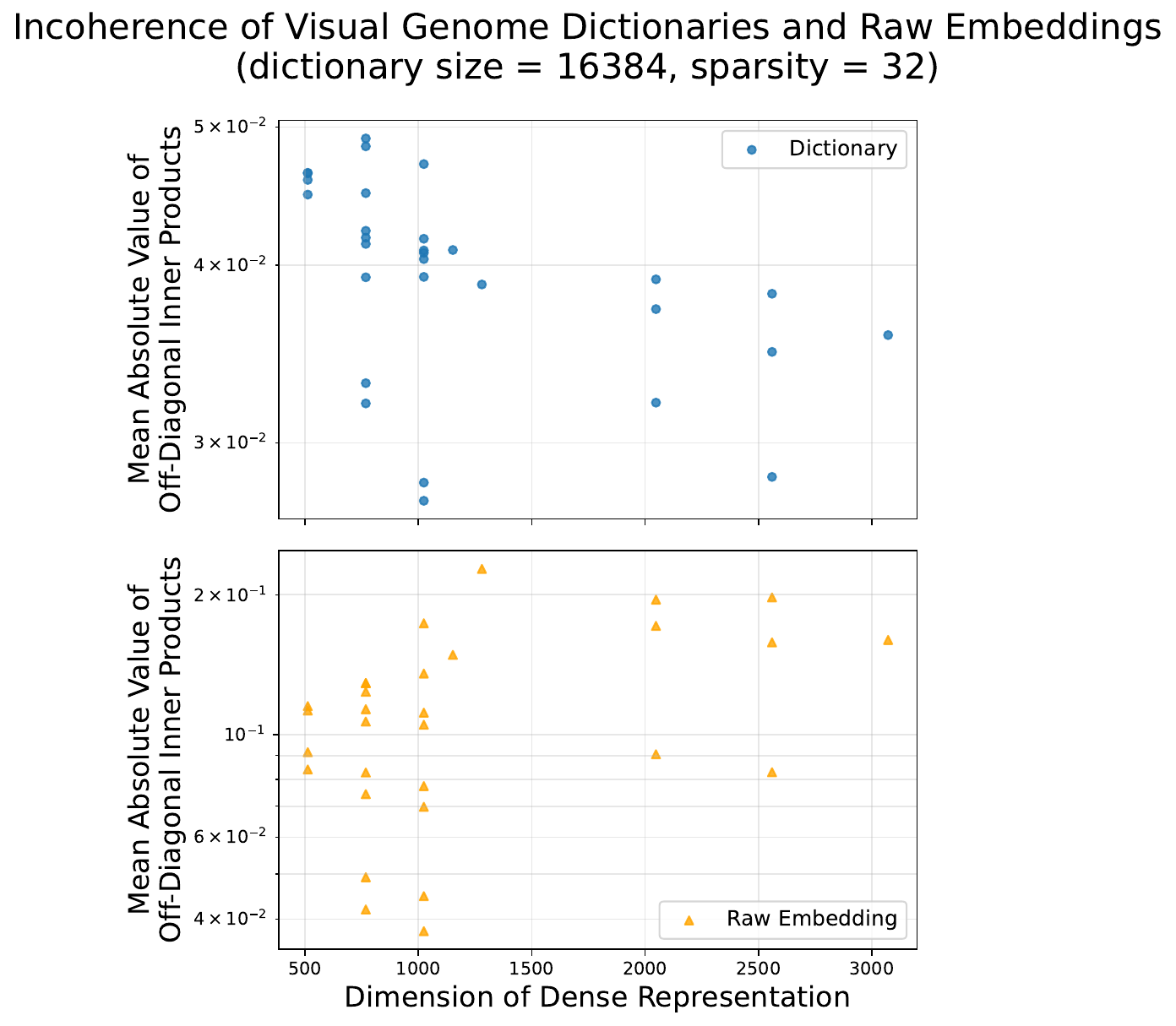}
    \end{minipage}

    \vspace{0.4cm}

    \begin{minipage}[t]{0.48\textwidth}
        \centering
        \includegraphics[width = .8\linewidth]{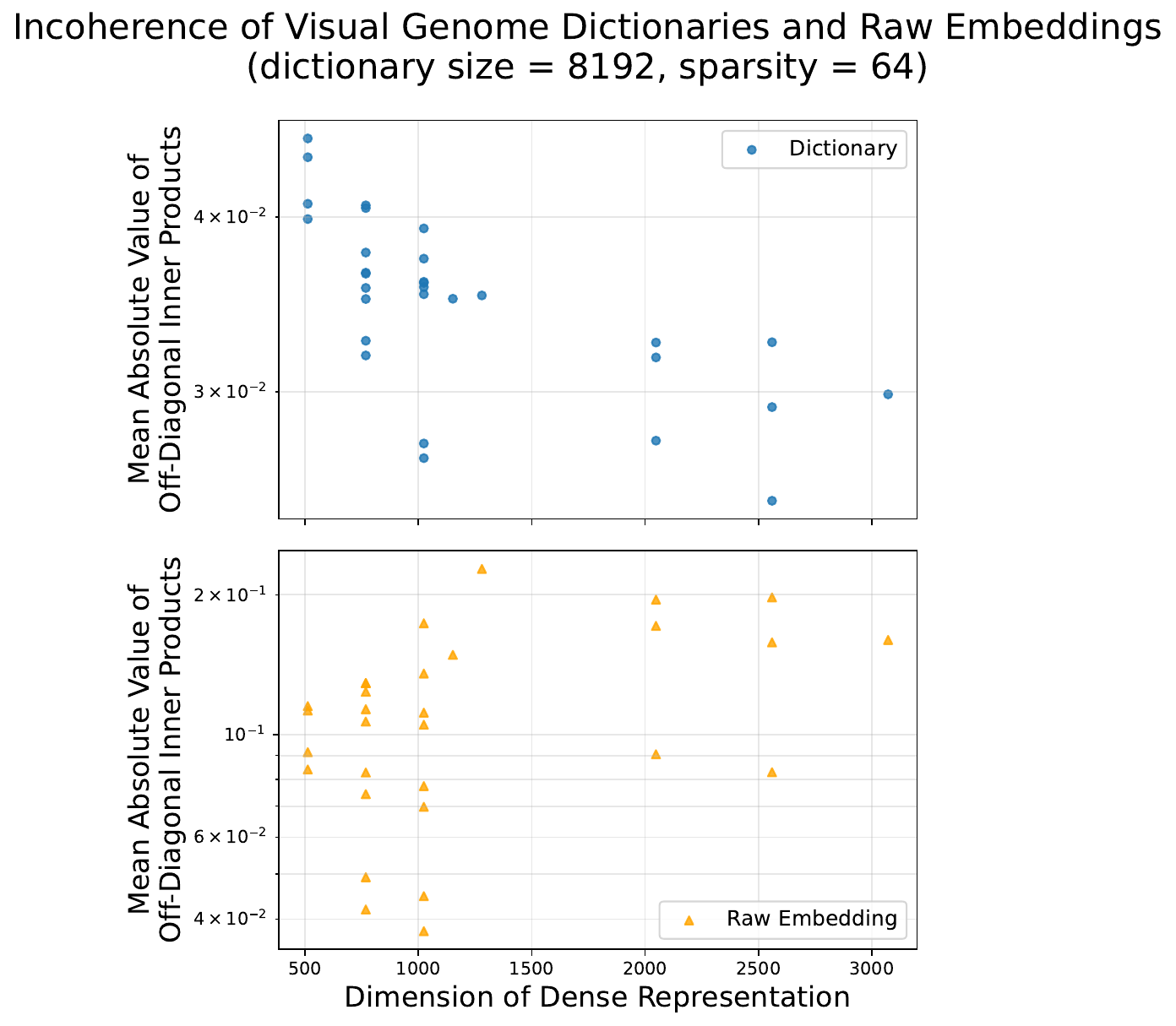}
    \end{minipage}
    \hfill
    \begin{minipage}[t]{0.48\textwidth}
        \centering
        \includegraphics[width = .8\linewidth]{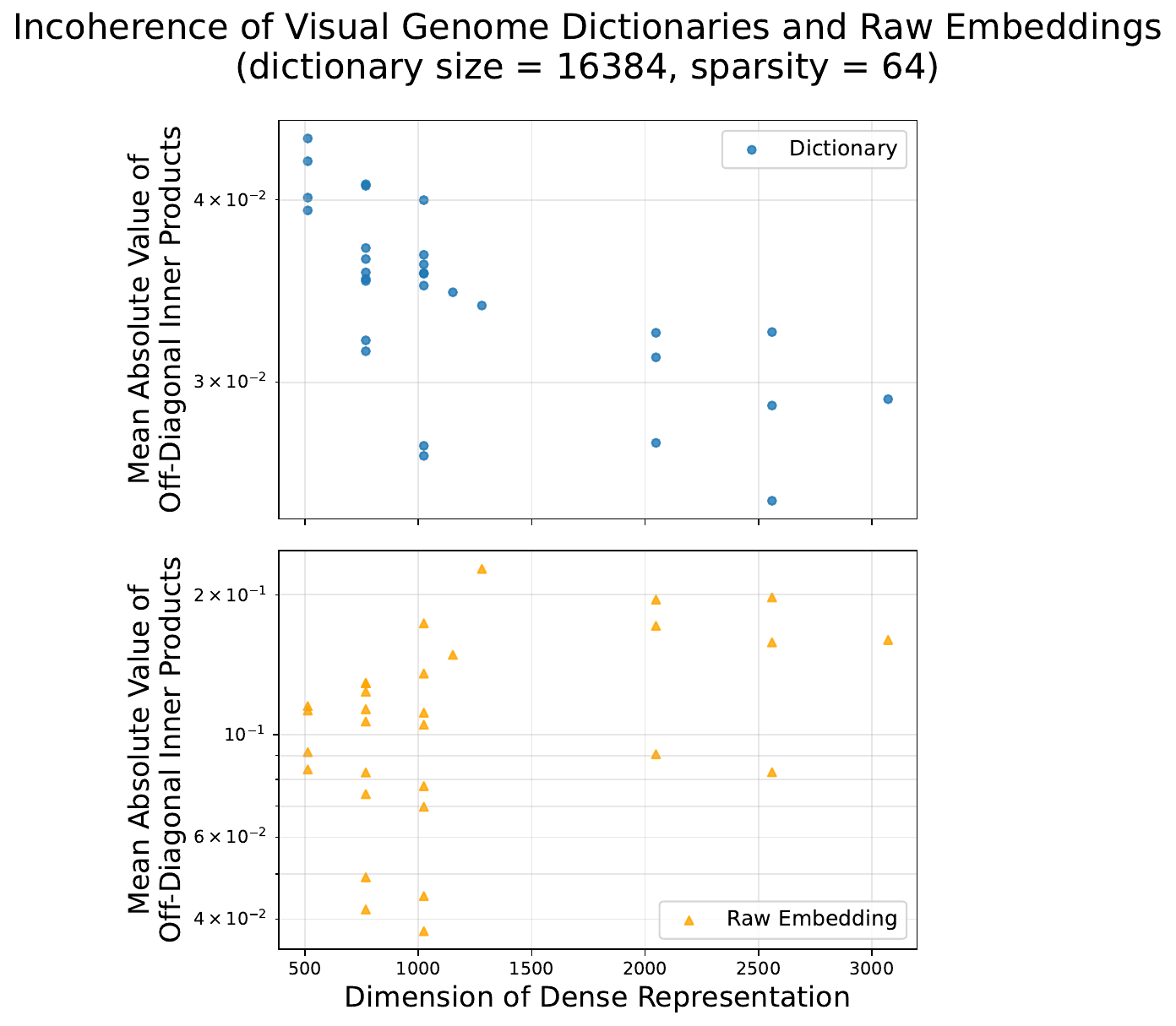}
    \end{minipage}

    \vspace{0.4cm}

    \begin{minipage}[t]{0.48\textwidth}
        \centering
        \includegraphics[width = .8\linewidth]{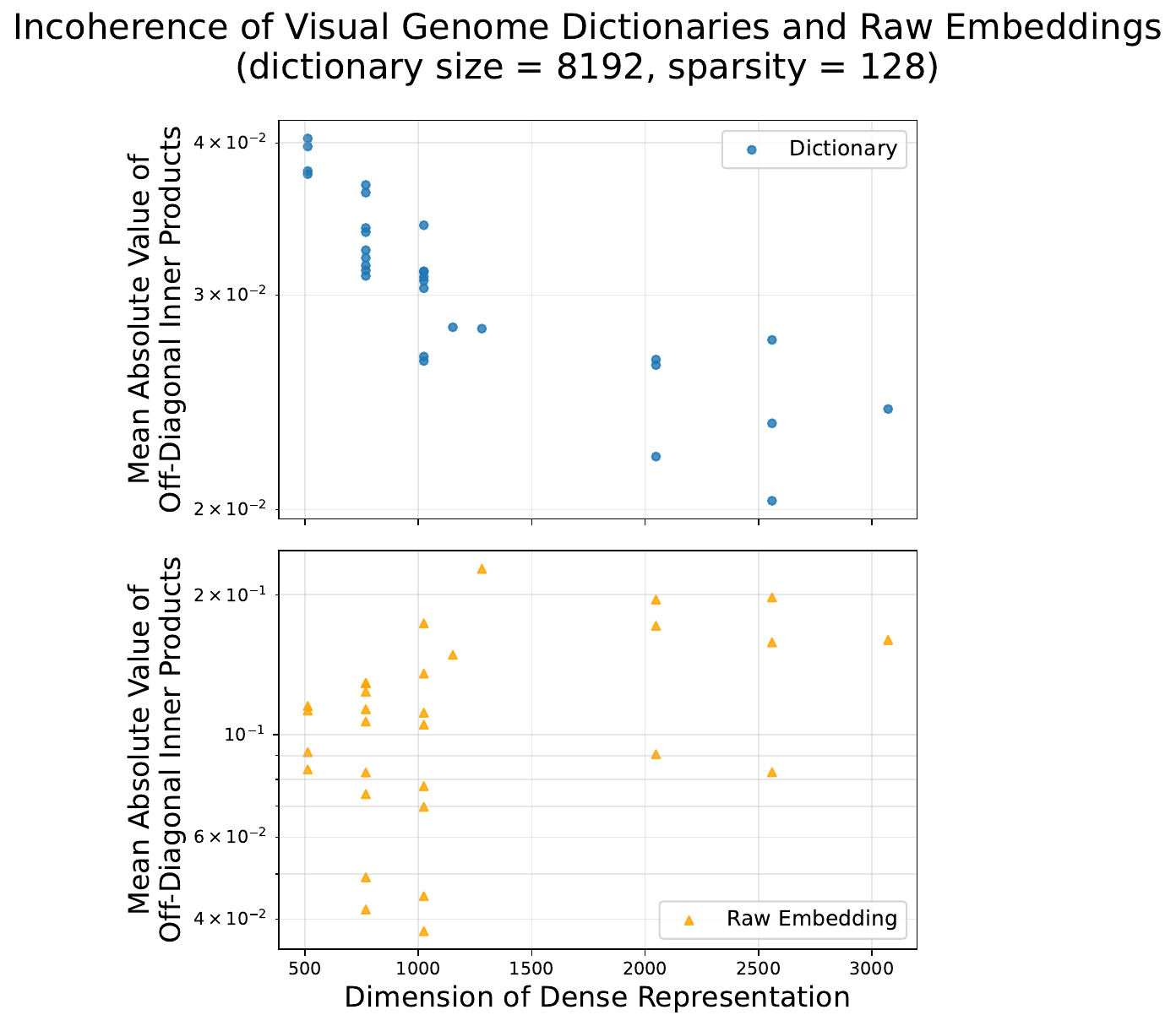}
    \end{minipage}
    \hfill
    \begin{minipage}[t]{0.48\textwidth}
        \centering
        \includegraphics[width = .8\linewidth]{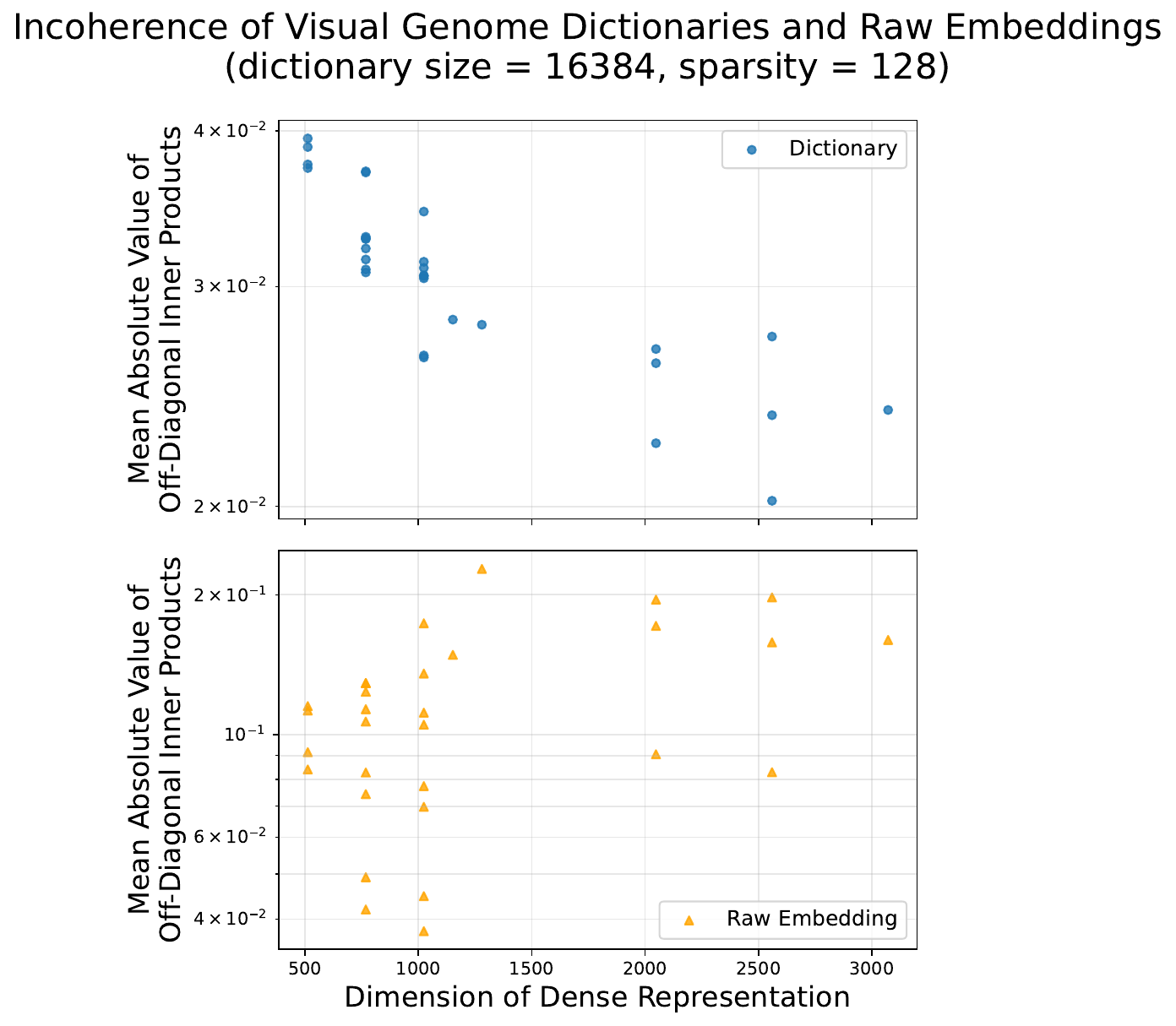}
    \end{minipage}

    \caption{The same figure as Figure~\ref{fig:appendixincoherencecoco} but over the Visual Genome dataset. Qualitatively, all trends are similar.}
    \label{fig:appendixincoherencevisual_genome}
\end{figure}

\clearpage

\subsection{Magnitudes of Sparse Features}
\label{appendix:magntiduesofsparsefeaturs}
Here, we plot statistics for the non-zero values of $\sqrt{k}\times M(X, \theta, f)$ for the sparse features we trained over COCO, CC3M, Visual Genome. We give the 5th and 95th percentiles of the values and their ratio. This shows that most entries are on the order of $\Theta(1/\sqrt{k})$ in support of Assumption A.3). In the case of COCO, we ablate the dimension by showing results for $d = 8192, 16384.$ For the other datasets we only show 
$d =  16384$ for brevity.

\subsubsection{Experiments on COCO}
\input{sparse_feature_percentile_latex_tables/sparse_feature_entries_coco_truncated_group-pre_truncation_d8192_k32}

\clearpage

\input{sparse_feature_percentile_latex_tables/sparse_feature_entries_coco_truncated_group-pre_truncation_d8192_k64}

\clearpage

\input{sparse_feature_percentile_latex_tables/sparse_feature_entries_coco_truncated_group-pre_truncation_d8192_k128}
\clearpage

\input{sparse_feature_percentile_latex_tables/sparse_feature_entries_coco_truncated_group-pre_truncation_d16384_k32}
\clearpage

\input{sparse_feature_percentile_latex_tables/sparse_feature_entries_coco_truncated_group-pre_truncation_d16384_k64}
\clearpage

\input{sparse_feature_percentile_latex_tables/sparse_feature_entries_coco_truncated_group-pre_truncation_d16384_k128}

\clearpage

\subsubsection{Experiments on CC3M}
\input{sparse_feature_percentile_latex_tables/sparse_feature_entries_cc3m_truncated_group-pre_truncation_d16384_k32}

\clearpage

\input{sparse_feature_percentile_latex_tables/sparse_feature_entries_cc3m_truncated_group-pre_truncation_d16384_k64}
\clearpage

\input{sparse_feature_percentile_latex_tables/sparse_feature_entries_cc3m_truncated_group-pre_truncation_d16384_k128}
\clearpage

\subsubsection{Experiments on Visual Genome}
\input{sparse_feature_percentile_latex_tables/sparse_feature_entries_visual_genome_truncated_group-pre_truncation_d16384_k32}

\clearpage

\input{sparse_feature_percentile_latex_tables/sparse_feature_entries_visual_genome_truncated_group-pre_truncation_d16384_k64}

\clearpage

\input{sparse_feature_percentile_latex_tables/sparse_feature_entries_visual_genome_truncated_group-pre_truncation_d16384_k128}

\clearpage

\subsection{Residuals of Sparse Autoencoders}
\label{appendix:saeresiduals}
\subsubsection{Experiments on COCO}
Here, we show the residuals for trained autoencoders subject to different sparsity and dimension. The main goal of this section is to demonstrate the small residual errors, thus justifying the sparsity assumption A.1) in Section~\ref{sec:statmodel}.

\input{residual_statistics/residual_statistics_coco_d_8192}

\clearpage

\input{residual_statistics/residual_statistics_coco_d_16384}

\clearpage

\subsubsection{Experiments on CC3M}

\input{residual_statistics/residual_statistics_cc3m_d_8192}

\clearpage

\input{residual_statistics/residual_statistics_cc3m_d_16384}

\clearpage

\subsubsection{Experiments on Visual Genome}

\input{residual_statistics/residual_statistics_visual_genome_d_8192}

\clearpage

\input{residual_statistics/residual_statistics_visual_genome_d_16384}

\clearpage

%% file: tikz/modelspecifics1.tex
\begin{longtable}{l c c c}
\caption{Specifications of Used Models: Architecture}\label{table:modelspecs_architecture}\\
\toprule
\textbf{Model Identifier} & \textbf{Params (B)} & \textbf{Depth} & \textbf{Width} \\
\midrule
\endfirsthead

\toprule
\textbf{Model Identifier} & \textbf{Params (B)} & \textbf{Depth} & \textbf{Width} \\
\midrule
\endhead

\bottomrule
\endfoot

Llama-3.2-1B & 1.0 & 16 & 2048 \\
Llama-3.2-3B & 3.0 & 28 & 3072 \\
Qwen3-1.7B & 1.7 & 28 & 2048 \\
Qwen3-4B & 4.0 & 36 & 2560 \\
Gemma-3-1B & 1.0 & 26 & 1152 \\
Gemma-3-4B & 4.0 & 32 & 2560 \\

\midrule
BGE-Base-en-v1.5 & 0.11 & 12 & 768 \\
BGE-Large-en-v1.5 & 0.33 & 24 & 1024 \\
F2LLM-1.7B & 1.7 & 28 & 2048 \\
F2LLM-4B & 4.0 & 36 & 2560 \\
Nomic-Embed-v1.5 & 0.14 & 12 & 768 \\
Nomic-Embed-v2-MoE & 0.47 & 12 & 768 \\

\midrule
SigLIP2-Base-T & 0.086 & 12 & 768 \\
SigLIP2-Large-T & 0.43 & 24 & 1024 \\
LAION-CLIP-B/32-T & 0.15 & 12 & 512 \\
LAION-CLIP-H/14-T & 0.98 & 24 & 1024 \\
OpenAI-CLIP-B/32-T & 0.15 & 12 & 512 \\
OpenAI-CLIP-L/14-T & 0.43 & 12 & 768 \\

\midrule
SigLIP2-Base-I & 0.086 & 12 & 768 \\
SigLIP2-Large-I & 0.43 & 24 & 1024 \\
LAION-CLIP-B/32-I & 0.15 & 12 & 768 \\
LAION-CLIP-H/14-I & 0.98 & 32 & 1280 \\
OpenAI-CLIP-B/32-I & 0.15 & 12 & 768 \\
OpenAI-CLIP-L/14-I & 0.43 & 24 & 1024 \\

\midrule
BEiT-Base & 0.086 & 12 & 768 \\
BEiT-Large & 0.30 & 24 & 1024 \\
DINOv2-Base & 0.086 & 12 & 768 \\
DINOv2-Large & 0.30 & 24 & 1024 \\
ViT-MAE-Huge & 0.63 & 32 & 1280 \\
ViT-MAE-Large & 0.30 & 24 & 1024 \\

\end{longtable}

%% file: tikz/modelspecifics2.tex
\begin{longtable}{l c c l c}
\caption{Specifications of Used Models: Data and Modality; MM = Multimodal}\label{table:modelspecs_data_modality}\\
\toprule
\textbf{Model Identifier} & \textbf{Text Tok.} & \textbf{Img Tok.} & \textbf{Modality} & \textbf{Year} \\
\midrule
\endfirsthead

\toprule
\textbf{Model Identifier} & \textbf{Text Tok.} & \textbf{Img Tok.} & \textbf{Modality} & \textbf{Year} \\
\midrule
\endhead

\bottomrule
\endfoot

Llama-3.2-1B & 9.00E+12 & 0 & LLM & 2024 \\
Llama-3.2-3B & 9.00E+12 & 0 & LLM & 2024 \\
Qwen3-1.7B & 3.60E+13 & 0 & LLM & 2025 \\
Qwen3-4B & 3.60E+13 & 0 & LLM & 2025 \\
Gemma-3-1B & 2.00E+12 & 0 & LLM & 2025 \\
Gemma-3-4B & 4.00E+12 & 4.00E+11 & LLM & 2025 \\

\midrule
BGE-Base-en-v1.5 & 4.00E+11 & 0 & Text Emb. & 2023 \\
BGE-Large-en-v1.5 & 4.00E+11 & 0 & Text Emb. & 2023 \\
F2LLM-1.7B & 3.60E+13 & 0 & Text Emb. & 2025 \\
F2LLM-4B & 3.60E+13 & 0 & Text Emb. & 2025 \\
Nomic-Embed-v1.5 & 6.00E+10 & 0 & Text Emb. & 2024 \\
Nomic-Embed-v2-MoE & 1.40E+12 & 0 & Text Emb. & 2025 \\

\midrule
SigLIP2-Base-T & 2.16E+11 & 2.16E+11 & MM, Text & 2025 \\
SigLIP2-Large-T & 2.16E+11 & 2.16E+11 & MM, Text & 2025 \\
LAION-CLIP-B/32-T & 6.12E+11 & 6.12E+11 & MM, Text & 2022 \\
LAION-CLIP-H/14-T & 5.76E+11 & 5.76E+11 & MM, Text & 2023 \\
OpenAI-CLIP-B/32-T & 7.20E+09 & 7.20E+09 & MM, Text & 2021 \\
OpenAI-CLIP-L/14-T & 2.34E+11 & 2.34E+11 & MM, Text & 2021 \\

\midrule
SigLIP2-Base-I & 2.16E+11 & 2.16E+11 & MM, Image & 2025 \\
SigLIP2-Large-I & 2.16E+11 & 2.16E+11 & MM, Image & 2025 \\
LAION-CLIP-B/32-I & 6.12E+11 & 6.12E+11 & MM, Image & 2022 \\
LAION-CLIP-H/14-I & 5.76E+11 & 5.76E+11 & MM, Image & 2023 \\
OpenAI-CLIP-B/32-I & 7.20E+09 & 7.20E+09 & MM, Image & 2021 \\
OpenAI-CLIP-L/14-I & 2.34E+11 & 2.34E+11 & MM, Image & 2021 \\

\midrule
BEiT-Base & 0 & 2.20E+12 & Image Found. & 2021 \\
BEiT-Large & 0 & 2.20E+12 & Image Found. & 2021 \\
DINOv2-Base & 0 & 1.42E+08 & Image Found. & 2023 \\
DINOv2-Large & 0 & 1.42E+08 & Image Found. & 2023 \\
ViT-MAE-Huge & 0 & 5.26E+11 & Image Found. & 2021 \\
ViT-MAE-Large & 0 & 4.03E+11 & Image Found. & 2021 \\

\end{longtable}

%% file: sparse_feature_percentile_latex_tables/sparse_feature_entries_coco_truncated_group-pre_truncation_d8192_k32.tex
\begin{table}[htbp]
\centering
\small
\resizebox{\textwidth}{!}{%
\begin{tabular}{lrrrr}
\toprule
Model & $d_{\mathrm{trunc}}$ & 5\% & 95\% & 95\%/5\% \\
\midrule
\texttt{BAAI\_\_bge-base-en-v1.5\_text} & 6501 & 0.0204259 & 0.334828 & 16.3923 \\
\texttt{BAAI\_\_bge-large-en-v1.5\_text} & 6703 & 0.0204214 & 0.331682 & 16.2419 \\
\texttt{codefuse-ai\_\_F2LLM-1.7B\_text} & 8128 & 0.025685 & 0.345818 & 13.4638 \\
\texttt{codefuse-ai\_\_F2LLM-4B\_text} & 8162 & 0.0270543 & 0.331568 & 12.2557 \\
\texttt{facebook\_\_dinov2-base\_img} & 8182 & 0.036548 & 0.254256 & 6.95678 \\
\texttt{facebook\_\_dinov2-large\_img} & 8173 & 0.0307982 & 0.269656 & 8.75558 \\
\texttt{facebook\_\_vit-mae-huge\_img} & 7064 & 0.0304735 & 0.153104 & 5.02419 \\
\texttt{facebook\_\_vit-mae-large\_img} & 5320 & 0.0371731 & 0.17026 & 4.58019 \\
\texttt{google\_\_gemma-3-1b-it\_text} & 7281 & 0.0326729 & 0.261633 & 8.00764 \\
\texttt{google\_\_gemma-3-4b-it\_text} & 7923 & 0.0311795 & 0.287769 & 9.22945 \\
\texttt{google\_\_siglip2-base-patch16-256\_img} & 7245 & 0.0403547 & 0.252237 & 6.25049 \\
\texttt{google\_\_siglip2-base-patch16-256\_text} & 6953 & 0.023171 & 0.348481 & 15.0395 \\
\texttt{google\_\_siglip2-large-patch16-256\_img} & 7104 & 0.0406788 & 0.259245 & 6.37297 \\
\texttt{google\_\_siglip2-large-patch16-256\_text} & 7395 & 0.0248135 & 0.346706 & 13.9724 \\
\texttt{laion\_\_CLIP-ViT-B-32-laion2B-s34B-b79K\_img} & 7354 & 0.0433992 & 0.232559 & 5.3586 \\
\texttt{laion\_\_CLIP-ViT-B-32-laion2B-s34B-b79K\_text} & 6894 & 0.019666 & 0.346289 & 17.6085 \\
\texttt{laion\_\_CLIP-ViT-H-14-laion2B-s32B-b79K\_img} & 7784 & 0.0422671 & 0.250111 & 5.91739 \\
\texttt{laion\_\_CLIP-ViT-H-14-laion2B-s32B-b79K\_text} & 6800 & 0.0213895 & 0.346613 & 16.2048 \\
\texttt{meta-llama\_\_Llama-3.2-1B-Instruct\_text} & 7557 & 0.0331361 & 0.283795 & 8.56452 \\
\texttt{meta-llama\_\_Llama-3.2-3B-Instruct\_text} & 7656 & 0.0315294 & 0.305695 & 9.69556 \\
\texttt{microsoft\_\_beit-base-patch16-224\_img} & 8188 & 0.038307 & 0.219244 & 5.72335 \\
\texttt{microsoft\_\_beit-large-patch16-224\_img} & 8185 & 0.0365769 & 0.22952 & 6.27498 \\
\texttt{nomic-ai\_\_nomic-embed-text-v1.5\_text} & 6088 & 0.017752 & 0.332005 & 18.7024 \\
\texttt{nomic-ai\_\_nomic-embed-text-v2-moe\_text} & 5997 & 0.0220674 & 0.379232 & 17.1852 \\
\texttt{openai\_\_clip-vit-base-patch32\_img} & 7507 & 0.0395703 & 0.226191 & 5.71618 \\
\texttt{openai\_\_clip-vit-base-patch32\_text} & 7259 & 0.0213545 & 0.336768 & 15.7704 \\
\texttt{openai\_\_clip-vit-large-patch14\_img} & 7984 & 0.0400789 & 0.242927 & 6.06122 \\
\texttt{openai\_\_clip-vit-large-patch14\_text} & 7305 & 0.0228015 & 0.332887 & 14.5994 \\
\texttt{Qwen\_\_Qwen3-1.7B-Base\_text} & 8029 & 0.0316816 & 0.263245 & 8.30907 \\
\texttt{Qwen\_\_Qwen3-4B-Base\_text} & 7962 & 0.0317245 & 0.274417 & 8.65001 \\
\bottomrule
\end{tabular}
}%
\caption{Statistics of nonzero sparse feature entries for COCO (truncated sparse features), pre-truncation sparse dimension $d=8192$, and pre-truncation sparsity $k=32$.}
\label{tab:sparse-feature-entries-coco-truncated-d8192-k32}
\end{table}

%% file: sparse_feature_percentile_latex_tables/sparse_feature_entries_coco_truncated_group-pre_truncation_d8192_k64.tex
\begin{table}[htbp]
\centering
\small
\resizebox{\textwidth}{!}{%
\begin{tabular}{lrrrr}
\toprule
Model & $d_{\mathrm{trunc}}$ & 5\% & 95\% & 95\%/5\% \\
\midrule
\texttt{BAAI\_\_bge-base-en-v1.5\_text} & 6862 & 0.0122533 & 0.229056 & 18.6934 \\
\texttt{BAAI\_\_bge-large-en-v1.5\_text} & 7009 & 0.0123748 & 0.223808 & 18.0858 \\
\texttt{codefuse-ai\_\_F2LLM-1.7B\_text} & 7735 & 0.0155105 & 0.24688 & 15.9169 \\
\texttt{codefuse-ai\_\_F2LLM-4B\_text} & 7858 & 0.0166596 & 0.2394 & 14.3701 \\
\texttt{facebook\_\_dinov2-base\_img} & 8179 & 0.0245707 & 0.187224 & 7.6198 \\
\texttt{facebook\_\_dinov2-large\_img} & 8179 & 0.0207208 & 0.197674 & 9.53989 \\
\texttt{facebook\_\_vit-mae-huge\_img} & 8019 & 0.0190106 & 0.103233 & 5.4303 \\
\texttt{facebook\_\_vit-mae-large\_img} & 7973 & 0.0231411 & 0.115512 & 4.99162 \\
\texttt{google\_\_gemma-3-1b-it\_text} & 7831 & 0.0196218 & 0.17284 & 8.80854 \\
\texttt{google\_\_gemma-3-4b-it\_text} & 8001 & 0.0189279 & 0.186128 & 9.83351 \\
\texttt{google\_\_siglip2-base-patch16-256\_img} & 7974 & 0.0257828 & 0.185305 & 7.18714 \\
\texttt{google\_\_siglip2-base-patch16-256\_text} & 7290 & 0.0143483 & 0.242921 & 16.9303 \\
\texttt{google\_\_siglip2-large-patch16-256\_img} & 7862 & 0.0256438 & 0.19303 & 7.52735 \\
\texttt{google\_\_siglip2-large-patch16-256\_text} & 7444 & 0.0156715 & 0.24251 & 15.4746 \\
\texttt{laion\_\_CLIP-ViT-B-32-laion2B-s34B-b79K\_img} & 8084 & 0.0286987 & 0.170161 & 5.92922 \\
\texttt{laion\_\_CLIP-ViT-B-32-laion2B-s34B-b79K\_text} & 7544 & 0.0122092 & 0.23524 & 19.2675 \\
\texttt{laion\_\_CLIP-ViT-H-14-laion2B-s32B-b79K\_img} & 7987 & 0.0277527 & 0.180794 & 6.51445 \\
\texttt{laion\_\_CLIP-ViT-H-14-laion2B-s32B-b79K\_text} & 7276 & 0.0132696 & 0.232559 & 17.5257 \\
\texttt{meta-llama\_\_Llama-3.2-1B-Instruct\_text} & 8023 & 0.0193464 & 0.189485 & 9.79434 \\
\texttt{meta-llama\_\_Llama-3.2-3B-Instruct\_text} & 8025 & 0.0181734 & 0.201008 & 11.0605 \\
\texttt{microsoft\_\_beit-base-patch16-224\_img} & 8183 & 0.0266547 & 0.158994 & 5.96496 \\
\texttt{microsoft\_\_beit-large-patch16-224\_img} & 8182 & 0.0260675 & 0.164443 & 6.30836 \\
\texttt{nomic-ai\_\_nomic-embed-text-v1.5\_text} & 6300 & 0.0108577 & 0.226129 & 20.8265 \\
\texttt{nomic-ai\_\_nomic-embed-text-v2-moe\_text} & 6709 & 0.0120219 & 0.263876 & 21.9496 \\
\texttt{openai\_\_clip-vit-base-patch32\_img} & 8080 & 0.0260672 & 0.173387 & 6.65154 \\
\texttt{openai\_\_clip-vit-base-patch32\_text} & 7675 & 0.0133075 & 0.218397 & 16.4116 \\
\texttt{openai\_\_clip-vit-large-patch14\_img} & 8075 & 0.0267408 & 0.179849 & 6.72564 \\
\texttt{openai\_\_clip-vit-large-patch14\_text} & 7459 & 0.0143854 & 0.226057 & 15.7143 \\
\texttt{Qwen\_\_Qwen3-1.7B-Base\_text} & 7981 & 0.0191093 & 0.179467 & 9.3916 \\
\texttt{Qwen\_\_Qwen3-4B-Base\_text} & 8043 & 0.0189182 & 0.185603 & 9.81081 \\
\bottomrule
\end{tabular}
}%
\caption{Statistics of nonzero sparse feature entries for COCO (truncated sparse features), pre-truncation sparse dimension $d=8192$, and pre-truncation sparsity $k=64$.}
\label{tab:sparse-feature-entries-coco-truncated-d8192-k64}
\end{table}

%% file: sparse_feature_percentile_latex_tables/sparse_feature_entries_coco_truncated_group-pre_truncation_d8192_k128.tex
\begin{table}[htbp]
\centering
\small
\resizebox{\textwidth}{!}{%
\begin{tabular}{lrrrr}
\toprule
Model & $d_{\mathrm{trunc}}$ & 5\% & 95\% & 95\%/5\% \\
\midrule
\texttt{BAAI\_\_bge-base-en-v1.5\_text} & 6576 & 0.00947677 & 0.159422 & 16.8224 \\
\texttt{BAAI\_\_bge-large-en-v1.5\_text} & 6692 & 0.00952511 & 0.160723 & 16.8736 \\
\texttt{codefuse-ai\_\_F2LLM-1.7B\_text} & 6711 & 0.0109136 & 0.191012 & 17.5021 \\
\texttt{codefuse-ai\_\_F2LLM-4B\_text} & 6878 & 0.0120174 & 0.190678 & 15.8667 \\
\texttt{facebook\_\_dinov2-base\_img} & 8145 & 0.0167647 & 0.133462 & 7.96089 \\
\texttt{facebook\_\_dinov2-large\_img} & 8159 & 0.0141138 & 0.13977 & 9.90304 \\
\texttt{facebook\_\_vit-mae-huge\_img} & 7508 & 0.0129893 & 0.0728865 & 5.61129 \\
\texttt{facebook\_\_vit-mae-large\_img} & 7477 & 0.0158122 & 0.0890399 & 5.63108 \\
\texttt{google\_\_gemma-3-1b-it\_text} & 7139 & 0.0131372 & 0.121057 & 9.2148 \\
\texttt{google\_\_gemma-3-4b-it\_text} & 7520 & 0.0125978 & 0.129427 & 10.2738 \\
\texttt{google\_\_siglip2-base-patch16-256\_img} & 7596 & 0.0183 & 0.147516 & 8.06099 \\
\texttt{google\_\_siglip2-base-patch16-256\_text} & 7091 & 0.0102159 & 0.175918 & 17.22 \\
\texttt{google\_\_siglip2-large-patch16-256\_img} & 7273 & 0.0178537 & 0.157 & 8.79372 \\
\texttt{google\_\_siglip2-large-patch16-256\_text} & 6836 & 0.0112409 & 0.184169 & 16.3838 \\
\texttt{laion\_\_CLIP-ViT-B-32-laion2B-s34B-b79K\_img} & 7631 & 0.0212935 & 0.134371 & 6.31041 \\
\texttt{laion\_\_CLIP-ViT-B-32-laion2B-s34B-b79K\_text} & 7226 & 0.00875272 & 0.157099 & 17.9486 \\
\texttt{laion\_\_CLIP-ViT-H-14-laion2B-s32B-b79K\_img} & 7525 & 0.0199689 & 0.134361 & 6.72851 \\
\texttt{laion\_\_CLIP-ViT-H-14-laion2B-s32B-b79K\_text} & 6789 & 0.00932585 & 0.159463 & 17.0991 \\
\texttt{meta-llama\_\_Llama-3.2-1B-Instruct\_text} & 7231 & 0.0123421 & 0.13002 & 10.5347 \\
\texttt{meta-llama\_\_Llama-3.2-3B-Instruct\_text} & 7458 & 0.0117533 & 0.143917 & 12.2448 \\
\texttt{microsoft\_\_beit-base-patch16-224\_img} & 8073 & 0.0186828 & 0.112873 & 6.04155 \\
\texttt{microsoft\_\_beit-large-patch16-224\_img} & 8123 & 0.0190029 & 0.114191 & 6.00911 \\
\texttt{nomic-ai\_\_nomic-embed-text-v1.5\_text} & 6132 & 0.0082975 & 0.158105 & 19.0545 \\
\texttt{nomic-ai\_\_nomic-embed-text-v2-moe\_text} & 6411 & 0.00847488 & 0.192248 & 22.6845 \\
\texttt{openai\_\_clip-vit-base-patch32\_img} & 7753 & 0.0184021 & 0.128132 & 6.9629 \\
\texttt{openai\_\_clip-vit-base-patch32\_text} & 7344 & 0.00973346 & 0.147666 & 15.171 \\
\texttt{openai\_\_clip-vit-large-patch14\_img} & 7800 & 0.0189537 & 0.130877 & 6.9051 \\
\texttt{openai\_\_clip-vit-large-patch14\_text} & 7212 & 0.0104516 & 0.157168 & 15.0377 \\
\texttt{Qwen\_\_Qwen3-1.7B-Base\_text} & 7394 & 0.0127484 & 0.124095 & 9.73415 \\
\texttt{Qwen\_\_Qwen3-4B-Base\_text} & 7447 & 0.0126204 & 0.13062 & 10.3499 \\
\bottomrule
\end{tabular}
}%
\caption{Statistics of nonzero sparse feature entries for COCO (truncated sparse features), pre-truncation sparse dimension $d=8192$, and pre-truncation sparsity $k=128$.}
\label{tab:sparse-feature-entries-coco-truncated-d8192-k128}
\end{table}

%% file: sparse_feature_percentile_latex_tables/sparse_feature_entries_coco_truncated_group-pre_truncation_d16384_k32.tex
\begin{table}[htbp]
\centering
\small
\resizebox{\textwidth}{!}{%
\begin{tabular}{lrrrr}
\toprule
Model & $d_{\mathrm{trunc}}$ & 5\% & 95\% & 95\%/5\% \\
\midrule
\texttt{BAAI\_\_bge-base-en-v1.5\_text} & 12156 & 0.0209062 & 0.334887 & 16.0185 \\
\texttt{BAAI\_\_bge-large-en-v1.5\_text} & 12634 & 0.0209635 & 0.32843 & 15.6667 \\
\texttt{codefuse-ai\_\_F2LLM-1.7B\_text} & 16232 & 0.0263172 & 0.342903 & 13.0296 \\
\texttt{codefuse-ai\_\_F2LLM-4B\_text} & 16336 & 0.027954 & 0.330471 & 11.822 \\
\texttt{facebook\_\_dinov2-base\_img} & 16316 & 0.0370923 & 0.248351 & 6.69549 \\
\texttt{facebook\_\_dinov2-large\_img} & 16302 & 0.0320621 & 0.263345 & 8.21361 \\
\texttt{facebook\_\_vit-mae-huge\_img} & 12167 & 0.0300413 & 0.151626 & 5.04725 \\
\texttt{facebook\_\_vit-mae-large\_img} & 8429 & 0.0367916 & 0.170366 & 4.63058 \\
\texttt{google\_\_gemma-3-1b-it\_text} & 13509 & 0.0329931 & 0.269656 & 8.17308 \\
\texttt{google\_\_gemma-3-4b-it\_text} & 15421 & 0.031239 & 0.280447 & 8.97747 \\
\texttt{google\_\_siglip2-base-patch16-256\_img} & 12600 & 0.0398165 & 0.248102 & 6.23113 \\
\texttt{google\_\_siglip2-base-patch16-256\_text} & 13421 & 0.0233936 & 0.343155 & 14.6688 \\
\texttt{google\_\_siglip2-large-patch16-256\_img} & 11990 & 0.0402536 & 0.254713 & 6.32772 \\
\texttt{google\_\_siglip2-large-patch16-256\_text} & 14299 & 0.0248937 & 0.340488 & 13.6777 \\
\texttt{laion\_\_CLIP-ViT-B-32-laion2B-s34B-b79K\_img} & 13138 & 0.0432908 & 0.227514 & 5.25549 \\
\texttt{laion\_\_CLIP-ViT-B-32-laion2B-s34B-b79K\_text} & 12445 & 0.0198198 & 0.340883 & 17.1991 \\
\texttt{laion\_\_CLIP-ViT-H-14-laion2B-s32B-b79K\_img} & 13751 & 0.0421843 & 0.24899 & 5.90244 \\
\texttt{laion\_\_CLIP-ViT-H-14-laion2B-s32B-b79K\_text} & 12834 & 0.0218466 & 0.345366 & 15.8087 \\
\texttt{meta-llama\_\_Llama-3.2-1B-Instruct\_text} & 13775 & 0.0336688 & 0.286113 & 8.49786 \\
\texttt{meta-llama\_\_Llama-3.2-3B-Instruct\_text} & 14243 & 0.0317138 & 0.296067 & 9.33558 \\
\texttt{microsoft\_\_beit-base-patch16-224\_img} & 16331 & 0.0389659 & 0.220325 & 5.65431 \\
\texttt{microsoft\_\_beit-large-patch16-224\_img} & 16348 & 0.0373687 & 0.227565 & 6.08971 \\
\texttt{nomic-ai\_\_nomic-embed-text-v1.5\_text} & 11309 & 0.018072 & 0.329122 & 18.2117 \\
\texttt{nomic-ai\_\_nomic-embed-text-v2-moe\_text} & 11172 & 0.0221846 & 0.358485 & 16.1592 \\
\texttt{openai\_\_clip-vit-base-patch32\_img} & 13599 & 0.0395022 & 0.228092 & 5.77415 \\
\texttt{openai\_\_clip-vit-base-patch32\_text} & 13370 & 0.021659 & 0.330874 & 15.2765 \\
\texttt{openai\_\_clip-vit-large-patch14\_img} & 14855 & 0.0399646 & 0.240326 & 6.01348 \\
\texttt{openai\_\_clip-vit-large-patch14\_text} & 14012 & 0.0229268 & 0.32795 & 14.3042 \\
\texttt{Qwen\_\_Qwen3-1.7B-Base\_text} & 15622 & 0.0318452 & 0.260091 & 8.16736 \\
\texttt{Qwen\_\_Qwen3-4B-Base\_text} & 15127 & 0.0322912 & 0.277167 & 8.58336 \\
\bottomrule
\end{tabular}
}%
\caption{Statistics of nonzero sparse feature entries for COCO (truncated sparse features), pre-truncation sparse dimension $d=16384$, and pre-truncation sparsity $k=32$.}
\label{tab:sparse-feature-entries-coco-truncated-d16384-k32}
\end{table}

%% file: sparse_feature_percentile_latex_tables/sparse_feature_entries_coco_truncated_group-pre_truncation_d16384_k64.tex
\begin{table}[htbp]
\centering
\small
\resizebox{\textwidth}{!}{%
\begin{tabular}{lrrrr}
\toprule
Model & $d_{\mathrm{trunc}}$ & 5\% & 95\% & 95\%/5\% \\
\midrule
\texttt{BAAI\_\_bge-base-en-v1.5\_text} & 13425 & 0.0121456 & 0.221162 & 18.2093 \\
\texttt{BAAI\_\_bge-large-en-v1.5\_text} & 14080 & 0.0124886 & 0.215718 & 17.2732 \\
\texttt{codefuse-ai\_\_F2LLM-1.7B\_text} & 15641 & 0.015605 & 0.238566 & 15.2877 \\
\texttt{codefuse-ai\_\_F2LLM-4B\_text} & 15951 & 0.0169069 & 0.233267 & 13.7971 \\
\texttt{facebook\_\_dinov2-base\_img} & 16349 & 0.0251458 & 0.183733 & 7.30671 \\
\texttt{facebook\_\_dinov2-large\_img} & 16300 & 0.0218632 & 0.19413 & 8.87933 \\
\texttt{facebook\_\_vit-mae-huge\_img} & 16135 & 0.0186105 & 0.098865 & 5.31231 \\
\texttt{facebook\_\_vit-mae-large\_img} & 16064 & 0.0228066 & 0.113366 & 4.97077 \\
\texttt{google\_\_gemma-3-1b-it\_text} & 15833 & 0.0193333 & 0.172215 & 8.9077 \\
\texttt{google\_\_gemma-3-4b-it\_text} & 16197 & 0.0190315 & 0.187962 & 9.87633 \\
\texttt{google\_\_siglip2-base-patch16-256\_img} & 15904 & 0.0253582 & 0.180948 & 7.13568 \\
\texttt{google\_\_siglip2-base-patch16-256\_text} & 14425 & 0.0141629 & 0.238086 & 16.8106 \\
\texttt{google\_\_siglip2-large-patch16-256\_img} & 15625 & 0.0250484 & 0.18953 & 7.56654 \\
\texttt{google\_\_siglip2-large-patch16-256\_text} & 14830 & 0.0154696 & 0.234929 & 15.1864 \\
\texttt{laion\_\_CLIP-ViT-B-32-laion2B-s34B-b79K\_img} & 16248 & 0.0284984 & 0.166281 & 5.83476 \\
\texttt{laion\_\_CLIP-ViT-B-32-laion2B-s34B-b79K\_text} & 14551 & 0.0121581 & 0.225188 & 18.5216 \\
\texttt{laion\_\_CLIP-ViT-H-14-laion2B-s32B-b79K\_img} & 15906 & 0.0273307 & 0.175952 & 6.43791 \\
\texttt{laion\_\_CLIP-ViT-H-14-laion2B-s32B-b79K\_text} & 14355 & 0.013256 & 0.231981 & 17.5001 \\
\texttt{meta-llama\_\_Llama-3.2-1B-Instruct\_text} & 16062 & 0.0195725 & 0.191495 & 9.78384 \\
\texttt{meta-llama\_\_Llama-3.2-3B-Instruct\_text} & 16196 & 0.0182568 & 0.196727 & 10.7755 \\
\texttt{microsoft\_\_beit-base-patch16-224\_img} & 16364 & 0.0271664 & 0.155521 & 5.72477 \\
\texttt{microsoft\_\_beit-large-patch16-224\_img} & 16340 & 0.0267062 & 0.159824 & 5.98453 \\
\texttt{nomic-ai\_\_nomic-embed-text-v1.5\_text} & 11943 & 0.0108978 & 0.223007 & 20.4635 \\
\texttt{nomic-ai\_\_nomic-embed-text-v2-moe\_text} & 13154 & 0.0118922 & 0.256222 & 21.5455 \\
\texttt{openai\_\_clip-vit-base-patch32\_img} & 16206 & 0.0256195 & 0.168311 & 6.56963 \\
\texttt{openai\_\_clip-vit-base-patch32\_text} & 14997 & 0.0135262 & 0.218082 & 16.1229 \\
\texttt{openai\_\_clip-vit-large-patch14\_img} & 16190 & 0.0265251 & 0.176135 & 6.64031 \\
\texttt{openai\_\_clip-vit-large-patch14\_text} & 14616 & 0.0144087 & 0.225442 & 15.6462 \\
\texttt{Qwen\_\_Qwen3-1.7B-Base\_text} & 16147 & 0.0191057 & 0.176288 & 9.22695 \\
\texttt{Qwen\_\_Qwen3-4B-Base\_text} & 16203 & 0.0192802 & 0.187183 & 9.70855 \\
\bottomrule
\end{tabular}
}%
\caption{Statistics of nonzero sparse feature entries for COCO (truncated sparse features), pre-truncation sparse dimension $d=16384$, and pre-truncation sparsity $k=64$.}
\label{tab:sparse-feature-entries-coco-truncated-d16384-k64}
\end{table}

%% file: sparse_feature_percentile_latex_tables/sparse_feature_entries_coco_truncated_group-pre_truncation_d16384_k128.tex
\begin{table}[htbp]
\centering
\small
\resizebox{\textwidth}{!}{%
\begin{tabular}{lrrrr}
\toprule
Model & $d_{\mathrm{trunc}}$ & 5\% & 95\% & 95\%/5\% \\
\midrule
\texttt{BAAI\_\_bge-base-en-v1.5\_text} & 12953 & 0.00921178 & 0.148766 & 16.1495 \\
\texttt{BAAI\_\_bge-large-en-v1.5\_text} & 13322 & 0.00922429 & 0.150628 & 16.3295 \\
\texttt{codefuse-ai\_\_F2LLM-1.7B\_text} & 13374 & 0.0107951 & 0.181575 & 16.8202 \\
\texttt{codefuse-ai\_\_F2LLM-4B\_text} & 13935 & 0.0117504 & 0.177851 & 15.1358 \\
\texttt{facebook\_\_dinov2-base\_img} & 16335 & 0.0168717 & 0.131751 & 7.80896 \\
\texttt{facebook\_\_dinov2-large\_img} & 16330 & 0.0144381 & 0.138797 & 9.61319 \\
\texttt{facebook\_\_vit-mae-huge\_img} & 15186 & 0.0127273 & 0.0686851 & 5.39665 \\
\texttt{facebook\_\_vit-mae-large\_img} & 15179 & 0.0152865 & 0.0845693 & 5.53228 \\
\texttt{google\_\_gemma-3-1b-it\_text} & 14454 & 0.012964 & 0.113653 & 8.76685 \\
\texttt{google\_\_gemma-3-4b-it\_text} & 15251 & 0.0123482 & 0.123099 & 9.96902 \\
\texttt{google\_\_siglip2-base-patch16-256\_img} & 15346 & 0.0176359 & 0.140577 & 7.9711 \\
\texttt{google\_\_siglip2-base-patch16-256\_text} & 13790 & 0.00994756 & 0.172252 & 17.316 \\
\texttt{google\_\_siglip2-large-patch16-256\_img} & 14669 & 0.0170101 & 0.147325 & 8.66105 \\
\texttt{google\_\_siglip2-large-patch16-256\_text} & 13731 & 0.0108356 & 0.174262 & 16.0824 \\
\texttt{laion\_\_CLIP-ViT-B-32-laion2B-s34B-b79K\_img} & 15571 & 0.0206177 & 0.130748 & 6.34155 \\
\texttt{laion\_\_CLIP-ViT-B-32-laion2B-s34B-b79K\_text} & 13924 & 0.00858519 & 0.150882 & 17.5747 \\
\texttt{laion\_\_CLIP-ViT-H-14-laion2B-s32B-b79K\_img} & 15182 & 0.0190137 & 0.127393 & 6.70005 \\
\texttt{laion\_\_CLIP-ViT-H-14-laion2B-s32B-b79K\_text} & 13568 & 0.00908236 & 0.156768 & 17.2608 \\
\texttt{meta-llama\_\_Llama-3.2-1B-Instruct\_text} & 14908 & 0.0122443 & 0.125104 & 10.2173 \\
\texttt{meta-llama\_\_Llama-3.2-3B-Instruct\_text} & 15092 & 0.0115528 & 0.135239 & 11.7062 \\
\texttt{microsoft\_\_beit-base-patch16-224\_img} & 16247 & 0.0187374 & 0.109942 & 5.86751 \\
\texttt{microsoft\_\_beit-large-patch16-224\_img} & 16305 & 0.0190576 & 0.114827 & 6.02529 \\
\texttt{nomic-ai\_\_nomic-embed-text-v1.5\_text} & 11706 & 0.00806524 & 0.1498 & 18.5736 \\
\texttt{nomic-ai\_\_nomic-embed-text-v2-moe\_text} & 12364 & 0.00816995 & 0.179353 & 21.9528 \\
\texttt{openai\_\_clip-vit-base-patch32\_img} & 15716 & 0.0176926 & 0.127141 & 7.18613 \\
\texttt{openai\_\_clip-vit-base-patch32\_text} & 14336 & 0.00962948 & 0.1523 & 15.816 \\
\texttt{openai\_\_clip-vit-large-patch14\_img} & 15775 & 0.0183329 & 0.132362 & 7.2199 \\
\texttt{openai\_\_clip-vit-large-patch14\_text} & 14018 & 0.0102086 & 0.154717 & 15.1555 \\
\texttt{Qwen\_\_Qwen3-1.7B-Base\_text} & 15177 & 0.0125856 & 0.118007 & 9.37631 \\
\texttt{Qwen\_\_Qwen3-4B-Base\_text} & 15230 & 0.0125451 & 0.125952 & 10.04 \\
\bottomrule
\end{tabular}
}%
\caption{Statistics of nonzero sparse feature entries for COCO (truncated sparse features), pre-truncation sparse dimension $d=16384$, and pre-truncation sparsity $k=128$.}
\label{tab:sparse-feature-entries-coco-truncated-d16384-k128}
\end{table}

%% file: sparse_feature_percentile_latex_tables/sparse_feature_entries_cc3m_truncated_group-pre_truncation_d16384_k32.tex
\begin{table}[htbp]
\centering
\small
\resizebox{\textwidth}{!}{%
\begin{tabular}{lrrrr}
\toprule
Model & $d_{\mathrm{trunc}}$ & 5\% & 95\% & 95\%/5\% \\
\midrule
\texttt{BAAI\_\_bge-base-en-v1.5\_text} & 13920 & 0.0254295 & 0.311401 & 12.2457 \\
\texttt{BAAI\_\_bge-large-en-v1.5\_text} & 13831 & 0.026407 & 0.303213 & 11.4823 \\
\texttt{codefuse-ai\_\_F2LLM-1.7B\_text} & 16310 & 0.0306691 & 0.284788 & 9.28584 \\
\texttt{codefuse-ai\_\_F2LLM-4B\_text} & 16345 & 0.0327587 & 0.269746 & 8.23433 \\
\texttt{facebook\_\_dinov2-base\_img} & 16214 & 0.0380363 & 0.247877 & 6.51686 \\
\texttt{facebook\_\_dinov2-large\_img} & 16219 & 0.0330927 & 0.262779 & 7.9407 \\
\texttt{facebook\_\_vit-mae-huge\_img} & 13601 & 0.0263153 & 0.130909 & 4.97464 \\
\texttt{facebook\_\_vit-mae-large\_img} & 10839 & 0.0327 & 0.148462 & 4.54012 \\
\texttt{google\_\_gemma-3-1b-it\_text} & 14404 & 0.0345314 & 0.217325 & 6.29354 \\
\texttt{google\_\_gemma-3-4b-it\_text} & 15952 & 0.0323284 & 0.203945 & 6.30854 \\
\texttt{google\_\_siglip2-base-patch16-256\_img} & 13684 & 0.0396712 & 0.228396 & 5.75724 \\
\texttt{google\_\_siglip2-base-patch16-256\_text} & 14513 & 0.0292539 & 0.282951 & 9.67224 \\
\texttt{google\_\_siglip2-large-patch16-256\_img} & 13259 & 0.0411318 & 0.231333 & 5.62419 \\
\texttt{google\_\_siglip2-large-patch16-256\_text} & 14814 & 0.0329464 & 0.277085 & 8.41017 \\
\texttt{laion\_\_CLIP-ViT-B-32-laion2B-s34B-b79K\_img} & 14661 & 0.0408718 & 0.213936 & 5.2343 \\
\texttt{laion\_\_CLIP-ViT-B-32-laion2B-s34B-b79K\_text} & 15403 & 0.0270586 & 0.285249 & 10.5419 \\
\texttt{laion\_\_CLIP-ViT-H-14-laion2B-s32B-b79K\_img} & 15240 & 0.0415339 & 0.221113 & 5.32368 \\
\texttt{laion\_\_CLIP-ViT-H-14-laion2B-s32B-b79K\_text} & 14969 & 0.0305111 & 0.284787 & 9.33387 \\
\texttt{meta-llama\_\_Llama-3.2-1B-Instruct\_text} & 15554 & 0.0250882 & 0.207663 & 8.27732 \\
\texttt{meta-llama\_\_Llama-3.2-3B-Instruct\_text} & 15899 & 0.0252918 & 0.214321 & 8.47392 \\
\texttt{microsoft\_\_beit-base-patch16-224\_img} & 16234 & 0.0411063 & 0.208686 & 5.07673 \\
\texttt{microsoft\_\_beit-large-patch16-224\_img} & 16283 & 0.039738 & 0.210243 & 5.29074 \\
\texttt{nomic-ai\_\_nomic-embed-text-v1.5\_text} & 13240 & 0.0225723 & 0.320595 & 14.203 \\
\texttt{nomic-ai\_\_nomic-embed-text-v2-moe\_text} & 13178 & 0.0237316 & 0.337105 & 14.2049 \\
\texttt{openai\_\_clip-vit-base-patch32\_img} & 14149 & 0.0408863 & 0.214074 & 5.23584 \\
\texttt{openai\_\_clip-vit-base-patch32\_text} & 15307 & 0.029697 & 0.272682 & 9.18212 \\
\texttt{openai\_\_clip-vit-large-patch14\_img} & 15072 & 0.0408191 & 0.215579 & 5.28132 \\
\texttt{openai\_\_clip-vit-large-patch14\_text} & 15134 & 0.0312728 & 0.277359 & 8.86902 \\
\texttt{Qwen\_\_Qwen3-1.7B-Base\_text} & 15721 & 0.0222427 & 0.185882 & 8.35699 \\
\texttt{Qwen\_\_Qwen3-4B-Base\_text} & 15813 & 0.0232599 & 0.201925 & 8.68125 \\
\bottomrule
\end{tabular}
}%
\caption{Statistics of nonzero sparse feature entries for CC3M (truncated sparse features), pre-truncation sparse dimension $d=16384$, and pre-truncation sparsity $k=32$.}
\label{tab:sparse-feature-entries-cc3m-truncated-d16384-k32}
\end{table}

%% file: sparse_feature_percentile_latex_tables/sparse_feature_entries_cc3m_truncated_group-pre_truncation_d16384_k64.tex
\begin{table}[htbp]
\centering
\small
\resizebox{\textwidth}{!}{%
\begin{tabular}{lrrrr}
\toprule
Model & $d_{\mathrm{trunc}}$ & 5\% & 95\% & 95\%/5\% \\
\midrule
\texttt{BAAI\_\_bge-base-en-v1.5\_text} & 14852 & 0.016976 & 0.232482 & 13.6947 \\
\texttt{BAAI\_\_bge-large-en-v1.5\_text} & 14482 & 0.0172617 & 0.219226 & 12.7002 \\
\texttt{codefuse-ai\_\_F2LLM-1.7B\_text} & 15930 & 0.0204981 & 0.215745 & 10.5251 \\
\texttt{codefuse-ai\_\_F2LLM-4B\_text} & 16269 & 0.0218171 & 0.197833 & 9.06778 \\
\texttt{facebook\_\_dinov2-base\_img} & 16348 & 0.0257843 & 0.180466 & 6.99908 \\
\texttt{facebook\_\_dinov2-large\_img} & 16262 & 0.0230792 & 0.191679 & 8.30527 \\
\texttt{facebook\_\_vit-mae-huge\_img} & 16196 & 0.0165786 & 0.0880321 & 5.31 \\
\texttt{facebook\_\_vit-mae-large\_img} & 16092 & 0.020561 & 0.0975483 & 4.74433 \\
\texttt{google\_\_gemma-3-1b-it\_text} & 15631 & 0.0222217 & 0.150841 & 6.78801 \\
\texttt{google\_\_gemma-3-4b-it\_text} & 16164 & 0.0197148 & 0.144143 & 7.31142 \\
\texttt{google\_\_siglip2-base-patch16-256\_img} & 15612 & 0.0249098 & 0.166845 & 6.69794 \\
\texttt{google\_\_siglip2-base-patch16-256\_text} & 15529 & 0.019116 & 0.20296 & 10.6173 \\
\texttt{google\_\_siglip2-large-patch16-256\_img} & 15459 & 0.0254063 & 0.169271 & 6.66256 \\
\texttt{google\_\_siglip2-large-patch16-256\_text} & 15214 & 0.0217111 & 0.200358 & 9.2284 \\
\texttt{laion\_\_CLIP-ViT-B-32-laion2B-s34B-b79K\_img} & 16172 & 0.0267453 & 0.156716 & 5.85957 \\
\texttt{laion\_\_CLIP-ViT-B-32-laion2B-s34B-b79K\_text} & 16190 & 0.0177759 & 0.198928 & 11.1909 \\
\texttt{laion\_\_CLIP-ViT-H-14-laion2B-s32B-b79K\_img} & 15818 & 0.02685 & 0.159632 & 5.94533 \\
\texttt{laion\_\_CLIP-ViT-H-14-laion2B-s32B-b79K\_text} & 15349 & 0.0199849 & 0.199043 & 9.95967 \\
\texttt{meta-llama\_\_Llama-3.2-1B-Instruct\_text} & 16136 & 0.0151176 & 0.143841 & 9.51482 \\
\texttt{meta-llama\_\_Llama-3.2-3B-Instruct\_text} & 15936 & 0.0152475 & 0.151104 & 9.91012 \\
\texttt{microsoft\_\_beit-base-patch16-224\_img} & 16325 & 0.0285476 & 0.151634 & 5.3116 \\
\texttt{microsoft\_\_beit-large-patch16-224\_img} & 16217 & 0.0282752 & 0.150382 & 5.31852 \\
\texttt{nomic-ai\_\_nomic-embed-text-v1.5\_text} & 14371 & 0.0146403 & 0.231188 & 15.7913 \\
\texttt{nomic-ai\_\_nomic-embed-text-v2-moe\_text} & 14431 & 0.0148379 & 0.258115 & 17.3956 \\
\texttt{openai\_\_clip-vit-base-patch32\_img} & 16153 & 0.026202 & 0.155622 & 5.93932 \\
\texttt{openai\_\_clip-vit-base-patch32\_text} & 16222 & 0.019699 & 0.189125 & 9.60072 \\
\texttt{openai\_\_clip-vit-large-patch14\_img} & 16016 & 0.0271577 & 0.157692 & 5.80654 \\
\texttt{openai\_\_clip-vit-large-patch14\_text} & 15835 & 0.0208797 & 0.188855 & 9.04492 \\
\texttt{Qwen\_\_Qwen3-1.7B-Base\_text} & 16243 & 0.0135206 & 0.1262 & 9.33389 \\
\texttt{Qwen\_\_Qwen3-4B-Base\_text} & 16131 & 0.014687 & 0.131423 & 8.94826 \\
\bottomrule
\end{tabular}
}%
\caption{Statistics of nonzero sparse feature entries for CC3M (truncated sparse features), pre-truncation sparse dimension $d=16384$, and pre-truncation sparsity $k=64$.}
\label{tab:sparse-feature-entries-cc3m-truncated-d16384-k64}
\end{table}

%% file: sparse_feature_percentile_latex_tables/sparse_feature_entries_cc3m_truncated_group-pre_truncation_d16384_k128.tex
\begin{table}[htbp]
\centering
\small
\resizebox{\textwidth}{!}{%
\begin{tabular}{lrrrr}
\toprule
Model & $d_{\mathrm{trunc}}$ & 5\% & 95\% & 95\%/5\% \\
\midrule
\texttt{BAAI\_\_bge-base-en-v1.5\_text} & 14303 & 0.0146377 & 0.182631 & 12.4767 \\
\texttt{BAAI\_\_bge-large-en-v1.5\_text} & 14068 & 0.0139354 & 0.169947 & 12.1953 \\
\texttt{codefuse-ai\_\_F2LLM-1.7B\_text} & 14125 & 0.0153332 & 0.169556 & 11.0581 \\
\texttt{codefuse-ai\_\_F2LLM-4B\_text} & 15147 & 0.0159894 & 0.140653 & 8.79667 \\
\texttt{facebook\_\_dinov2-base\_img} & 16344 & 0.0176766 & 0.123655 & 6.9954 \\
\texttt{facebook\_\_dinov2-large\_img} & 16311 & 0.0165585 & 0.128942 & 7.78708 \\
\texttt{facebook\_\_vit-mae-huge\_img} & 15552 & 0.0111773 & 0.055826 & 4.9946 \\
\texttt{facebook\_\_vit-mae-large\_img} & 15165 & 0.0139581 & 0.0644786 & 4.61943 \\
\texttt{google\_\_gemma-3-1b-it\_text} & 15014 & 0.0162736 & 0.1093 & 6.71638 \\
\texttt{google\_\_gemma-3-4b-it\_text} & 15163 & 0.0137327 & 0.101828 & 7.415 \\
\texttt{google\_\_siglip2-base-patch16-256\_img} & 14792 & 0.0179614 & 0.13477 & 7.50329 \\
\texttt{google\_\_siglip2-base-patch16-256\_text} & 15329 & 0.0146317 & 0.1537 & 10.5046 \\
\texttt{google\_\_siglip2-large-patch16-256\_img} & 14066 & 0.0174327 & 0.128601 & 7.37699 \\
\texttt{google\_\_siglip2-large-patch16-256\_text} & 14486 & 0.0164128 & 0.143507 & 8.7436 \\
\texttt{laion\_\_CLIP-ViT-B-32-laion2B-s34B-b79K\_img} & 14939 & 0.0211209 & 0.131191 & 6.21146 \\
\texttt{laion\_\_CLIP-ViT-B-32-laion2B-s34B-b79K\_text} & 15804 & 0.0147165 & 0.152747 & 10.3793 \\
\texttt{laion\_\_CLIP-ViT-H-14-laion2B-s32B-b79K\_img} & 14682 & 0.0197453 & 0.115771 & 5.86322 \\
\texttt{laion\_\_CLIP-ViT-H-14-laion2B-s32B-b79K\_text} & 14663 & 0.0157474 & 0.137498 & 8.73149 \\
\texttt{meta-llama\_\_Llama-3.2-1B-Instruct\_text} & 14912 & 0.0110526 & 0.101962 & 9.22514 \\
\texttt{meta-llama\_\_Llama-3.2-3B-Instruct\_text} & 15181 & 0.0107867 & 0.104912 & 9.72611 \\
\texttt{microsoft\_\_beit-base-patch16-224\_img} & 16255 & 0.0200309 & 0.105119 & 5.24783 \\
\texttt{microsoft\_\_beit-large-patch16-224\_img} & 16211 & 0.0207458 & 0.105733 & 5.09659 \\
\texttt{nomic-ai\_\_nomic-embed-text-v1.5\_text} & 13653 & 0.0129699 & 0.181755 & 14.0137 \\
\texttt{nomic-ai\_\_nomic-embed-text-v2-moe\_text} & 14216 & 0.0117553 & 0.206009 & 17.5248 \\
\texttt{openai\_\_clip-vit-base-patch32\_img} & 15319 & 0.0193033 & 0.120389 & 6.23671 \\
\texttt{openai\_\_clip-vit-base-patch32\_text} & 15813 & 0.016171 & 0.144277 & 8.92197 \\
\texttt{openai\_\_clip-vit-large-patch14\_img} & 15455 & 0.0201092 & 0.118739 & 5.90469 \\
\texttt{openai\_\_clip-vit-large-patch14\_text} & 15490 & 0.0168784 & 0.14567 & 8.63054 \\
\texttt{Qwen\_\_Qwen3-1.7B-Base\_text} & 15321 & 0.010005 & 0.0818743 & 8.18334 \\
\texttt{Qwen\_\_Qwen3-4B-Base\_text} & 15425 & 0.0108446 & 0.0882932 & 8.14165 \\
\bottomrule
\end{tabular}
}%
\caption{Statistics of nonzero sparse feature entries for CC3M (truncated sparse features), pre-truncation sparse dimension $d=16384$, and pre-truncation sparsity $k=128$.}
\label{tab:sparse-feature-entries-cc3m-truncated-d16384-k128}
\end{table}

%% file: sparse_feature_percentile_latex_tables/sparse_feature_entries_visual_genome_truncated_group-pre_truncation_d16384_k32.tex
\begin{table}[htbp]
\centering
\small
\resizebox{\textwidth}{!}{%
\begin{tabular}{lrrrr}
\toprule
Model & $d_{\mathrm{trunc}}$ & 5\% & 95\% & 95\%/5\% \\
\midrule
\texttt{BAAI\_\_bge-base-en-v1.5\_text} & 9587 & 0.0258465 & 0.294767 & 11.4045 \\
\texttt{BAAI\_\_bge-large-en-v1.5\_text} & 9458 & 0.0251976 & 0.294414 & 11.6842 \\
\texttt{codefuse-ai\_\_F2LLM-1.7B\_text} & 12253 & 0.0314145 & 0.292297 & 9.30452 \\
\texttt{codefuse-ai\_\_F2LLM-4B\_text} & 13250 & 0.0327303 & 0.28757 & 8.78605 \\
\texttt{facebook\_\_dinov2-base\_img} & 16312 & 0.0369236 & 0.252441 & 6.83685 \\
\texttt{facebook\_\_dinov2-large\_img} & 16295 & 0.0318728 & 0.267687 & 8.39862 \\
\texttt{facebook\_\_vit-mae-huge\_img} & 11802 & 0.0303962 & 0.155136 & 5.10379 \\
\texttt{facebook\_\_vit-mae-large\_img} & 8466 & 0.0371367 & 0.173793 & 4.67981 \\
\texttt{google\_\_gemma-3-1b-it\_text} & 9008 & 0.0333885 & 0.233623 & 6.99711 \\
\texttt{google\_\_gemma-3-4b-it\_text} & 11425 & 0.031951 & 0.231986 & 7.26069 \\
\texttt{google\_\_siglip2-base-patch16-256\_img} & 12248 & 0.0399248 & 0.253171 & 6.34119 \\
\texttt{google\_\_siglip2-base-patch16-256\_text} & 9888 & 0.0289573 & 0.293752 & 10.1443 \\
\texttt{google\_\_siglip2-large-patch16-256\_img} & 11782 & 0.0402489 & 0.257826 & 6.40578 \\
\texttt{google\_\_siglip2-large-patch16-256\_text} & 10414 & 0.0307455 & 0.282401 & 9.1851 \\
\texttt{laion\_\_CLIP-ViT-B-32-laion2B-s34B-b79K\_img} & 12854 & 0.0428923 & 0.231121 & 5.38841 \\
\texttt{laion\_\_CLIP-ViT-B-32-laion2B-s34B-b79K\_text} & 11322 & 0.0269072 & 0.307126 & 11.4143 \\
\texttt{laion\_\_CLIP-ViT-H-14-laion2B-s32B-b79K\_img} & 13682 & 0.041509 & 0.249676 & 6.015 \\
\texttt{laion\_\_CLIP-ViT-H-14-laion2B-s32B-b79K\_text} & 10699 & 0.029232 & 0.293988 & 10.0571 \\
\texttt{meta-llama\_\_Llama-3.2-1B-Instruct\_text} & 8295 & 0.0305928 & 0.225879 & 7.38339 \\
\texttt{meta-llama\_\_Llama-3.2-3B-Instruct\_text} & 8918 & 0.0314317 & 0.239794 & 7.62903 \\
\texttt{microsoft\_\_beit-base-patch16-224\_img} & 16309 & 0.0388014 & 0.225075 & 5.80069 \\
\texttt{microsoft\_\_beit-large-patch16-224\_img} & 16358 & 0.0371547 & 0.236853 & 6.37478 \\
\texttt{nomic-ai\_\_nomic-embed-text-v1.5\_text} & 9404 & 0.0242776 & 0.304992 & 12.5627 \\
\texttt{nomic-ai\_\_nomic-embed-text-v2-moe\_text} & 8001 & 0.0289007 & 0.298165 & 10.3169 \\
\texttt{openai\_\_clip-vit-base-patch32\_img} & 13494 & 0.0391038 & 0.228169 & 5.83494 \\
\texttt{openai\_\_clip-vit-base-patch32\_text} & 11921 & 0.0275645 & 0.294967 & 10.701 \\
\texttt{openai\_\_clip-vit-large-patch14\_img} & 14821 & 0.039517 & 0.241423 & 6.10934 \\
\texttt{openai\_\_clip-vit-large-patch14\_text} & 11422 & 0.0282427 & 0.290792 & 10.2962 \\
\texttt{Qwen\_\_Qwen3-1.7B-Base\_text} & 10365 & 0.028953 & 0.221531 & 7.6514 \\
\texttt{Qwen\_\_Qwen3-4B-Base\_text} & 9710 & 0.0290642 & 0.205751 & 7.0792 \\
\bottomrule
\end{tabular}
}%
\caption{Statistics of nonzero sparse feature entries for Visual Genome (truncated sparse features), pre-truncation sparse dimension $d=16384$, and pre-truncation sparsity $k=32$.}
\label{tab:sparse-feature-entries-visual-genome-truncated-d16384-k32}
\end{table}

%% file: sparse_feature_percentile_latex_tables/sparse_feature_entries_visual_genome_truncated_group-pre_truncation_d16384_k64.tex
\begin{table}[htbp]
\centering
\small
\resizebox{\textwidth}{!}{%
\begin{tabular}{lrrrr}
\toprule
Model & $d_{\mathrm{trunc}}$ & 5\% & 95\% & 95\%/5\% \\
\midrule
\texttt{BAAI\_\_bge-base-en-v1.5\_text} & 13218 & 0.015027 & 0.193868 & 12.9013 \\
\texttt{BAAI\_\_bge-large-en-v1.5\_text} & 13397 & 0.0144733 & 0.187733 & 12.971 \\
\texttt{codefuse-ai\_\_F2LLM-1.7B\_text} & 15704 & 0.0181161 & 0.202969 & 11.2038 \\
\texttt{codefuse-ai\_\_F2LLM-4B\_text} & 15987 & 0.0188635 & 0.198664 & 10.5317 \\
\texttt{facebook\_\_dinov2-base\_img} & 16364 & 0.025118 & 0.185073 & 7.36813 \\
\texttt{facebook\_\_dinov2-large\_img} & 16331 & 0.0217118 & 0.194673 & 8.96623 \\
\texttt{facebook\_\_vit-mae-huge\_img} & 16115 & 0.0189728 & 0.104986 & 5.53349 \\
\texttt{facebook\_\_vit-mae-large\_img} & 16029 & 0.0228774 & 0.118996 & 5.20144 \\
\texttt{google\_\_gemma-3-1b-it\_text} & 15698 & 0.0197309 & 0.143592 & 7.27752 \\
\texttt{google\_\_gemma-3-4b-it\_text} & 15873 & 0.0187409 & 0.145958 & 7.78824 \\
\texttt{google\_\_siglip2-base-patch16-256\_img} & 15983 & 0.0253152 & 0.183988 & 7.26788 \\
\texttt{google\_\_siglip2-base-patch16-256\_text} & 14472 & 0.0164526 & 0.203958 & 12.3967 \\
\texttt{google\_\_siglip2-large-patch16-256\_img} & 15693 & 0.0249241 & 0.187682 & 7.53016 \\
\texttt{google\_\_siglip2-large-patch16-256\_text} & 15001 & 0.0180389 & 0.19326 & 10.7135 \\
\texttt{laion\_\_CLIP-ViT-B-32-laion2B-s34B-b79K\_img} & 16275 & 0.0282605 & 0.169612 & 6.00173 \\
\texttt{laion\_\_CLIP-ViT-B-32-laion2B-s34B-b79K\_text} & 15058 & 0.0156479 & 0.197197 & 12.6022 \\
\texttt{laion\_\_CLIP-ViT-H-14-laion2B-s32B-b79K\_img} & 16039 & 0.0269098 & 0.179611 & 6.67456 \\
\texttt{laion\_\_CLIP-ViT-H-14-laion2B-s32B-b79K\_text} & 14669 & 0.0167427 & 0.203237 & 12.1389 \\
\texttt{meta-llama\_\_Llama-3.2-1B-Instruct\_text} & 15515 & 0.0177032 & 0.143266 & 8.09264 \\
\texttt{meta-llama\_\_Llama-3.2-3B-Instruct\_text} & 15509 & 0.0177851 & 0.157042 & 8.82998 \\
\texttt{microsoft\_\_beit-base-patch16-224\_img} & 16365 & 0.0272404 & 0.156832 & 5.75734 \\
\texttt{microsoft\_\_beit-large-patch16-224\_img} & 16355 & 0.0267243 & 0.162864 & 6.0942 \\
\texttt{nomic-ai\_\_nomic-embed-text-v1.5\_text} & 12464 & 0.0137996 & 0.202276 & 14.6581 \\
\texttt{nomic-ai\_\_nomic-embed-text-v2-moe\_text} & 13234 & 0.0148908 & 0.202929 & 13.6278 \\
\texttt{openai\_\_clip-vit-base-patch32\_img} & 16216 & 0.0255316 & 0.1704 & 6.67408 \\
\texttt{openai\_\_clip-vit-base-patch32\_text} & 15272 & 0.0163975 & 0.198987 & 12.1353 \\
\texttt{openai\_\_clip-vit-large-patch14\_img} & 16206 & 0.0260699 & 0.176949 & 6.78749 \\
\texttt{openai\_\_clip-vit-large-patch14\_text} & 14799 & 0.0168074 & 0.195592 & 11.6372 \\
\texttt{Qwen\_\_Qwen3-1.7B-Base\_text} & 15791 & 0.0172485 & 0.13732 & 7.96129 \\
\texttt{Qwen\_\_Qwen3-4B-Base\_text} & 15711 & 0.0171786 & 0.129968 & 7.5657 \\
\bottomrule
\end{tabular}
}%
\caption{Statistics of nonzero sparse feature entries for Visual Genome (truncated sparse features), pre-truncation sparse dimension $d=16384$, and pre-truncation sparsity $k=64$.}
\label{tab:sparse-feature-entries-visual-genome-truncated-d16384-k64}
\end{table}

%% file: sparse_feature_percentile_latex_tables/sparse_feature_entries_visual_genome_truncated_group-pre_truncation_d16384_k128.tex
\begin{table}[htbp]
\centering
\small
\resizebox{\textwidth}{!}{%
\begin{tabular}{lrrrr}
\toprule
Model & $d_{\mathrm{trunc}}$ & 5\% & 95\% & 95\%/5\% \\
\midrule
\texttt{BAAI\_\_bge-base-en-v1.5\_text} & 13282 & 0.0107109 & 0.123221 & 11.5042 \\
\texttt{BAAI\_\_bge-large-en-v1.5\_text} & 13727 & 0.0100894 & 0.126666 & 12.5543 \\
\texttt{codefuse-ai\_\_F2LLM-1.7B\_text} & 14317 & 0.0116631 & 0.143572 & 12.3099 \\
\texttt{codefuse-ai\_\_F2LLM-4B\_text} & 14049 & 0.0122185 & 0.140684 & 11.514 \\
\texttt{facebook\_\_dinov2-base\_img} & 16341 & 0.0167531 & 0.130932 & 7.8154 \\
\texttt{facebook\_\_dinov2-large\_img} & 16346 & 0.0143212 & 0.137849 & 9.62553 \\
\texttt{facebook\_\_vit-mae-huge\_img} & 15237 & 0.012855 & 0.0738263 & 5.74299 \\
\texttt{facebook\_\_vit-mae-large\_img} & 15270 & 0.0155369 & 0.0901772 & 5.80407 \\
\texttt{google\_\_gemma-3-1b-it\_text} & 14503 & 0.0124713 & 0.090303 & 7.24089 \\
\texttt{google\_\_gemma-3-4b-it\_text} & 15135 & 0.0118836 & 0.098183 & 8.26206 \\
\texttt{google\_\_siglip2-base-patch16-256\_img} & 15435 & 0.0175791 & 0.140541 & 7.99478 \\
\texttt{google\_\_siglip2-base-patch16-256\_text} & 13879 & 0.0110849 & 0.142973 & 12.898 \\
\texttt{google\_\_siglip2-large-patch16-256\_img} & 14796 & 0.0168805 & 0.150169 & 8.896 \\
\texttt{google\_\_siglip2-large-patch16-256\_text} & 14343 & 0.012156 & 0.134844 & 11.0928 \\
\texttt{laion\_\_CLIP-ViT-B-32-laion2B-s34B-b79K\_img} & 15658 & 0.0203661 & 0.133905 & 6.57488 \\
\texttt{laion\_\_CLIP-ViT-B-32-laion2B-s34B-b79K\_text} & 14446 & 0.0109357 & 0.125573 & 11.4829 \\
\texttt{laion\_\_CLIP-ViT-H-14-laion2B-s32B-b79K\_img} & 15349 & 0.0188075 & 0.129284 & 6.87407 \\
\texttt{laion\_\_CLIP-ViT-H-14-laion2B-s32B-b79K\_text} & 13865 & 0.0113247 & 0.130727 & 11.5435 \\
\texttt{meta-llama\_\_Llama-3.2-1B-Instruct\_text} & 14982 & 0.0110692 & 0.0934486 & 8.44225 \\
\texttt{meta-llama\_\_Llama-3.2-3B-Instruct\_text} & 15068 & 0.0108947 & 0.105156 & 9.65201 \\
\texttt{microsoft\_\_beit-base-patch16-224\_img} & 16287 & 0.0188108 & 0.112288 & 5.96935 \\
\texttt{microsoft\_\_beit-large-patch16-224\_img} & 16325 & 0.0191548 & 0.114931 & 6.00014 \\
\texttt{nomic-ai\_\_nomic-embed-text-v1.5\_text} & 13087 & 0.00968237 & 0.128162 & 13.2367 \\
\texttt{nomic-ai\_\_nomic-embed-text-v2-moe\_text} & 13168 & 0.00933232 & 0.133336 & 14.2876 \\
\texttt{openai\_\_clip-vit-base-patch32\_img} & 15817 & 0.0176415 & 0.128041 & 7.25797 \\
\texttt{openai\_\_clip-vit-base-patch32\_text} & 14699 & 0.011835 & 0.123648 & 10.4477 \\
\texttt{openai\_\_clip-vit-large-patch14\_img} & 15864 & 0.0181518 & 0.132014 & 7.27279 \\
\texttt{openai\_\_clip-vit-large-patch14\_text} & 14456 & 0.0120083 & 0.132492 & 11.0334 \\
\texttt{Qwen\_\_Qwen3-1.7B-Base\_text} & 15126 & 0.0111363 & 0.0887757 & 7.97173 \\
\texttt{Qwen\_\_Qwen3-4B-Base\_text} & 15385 & 0.0112631 & 0.0877425 & 7.79026 \\
\bottomrule
\end{tabular}
}%
\caption{Statistics of nonzero sparse feature entries for Visual Genome (truncated sparse features), pre-truncation sparse dimension $d=16384$, and pre-truncation sparsity $k=128$.}
\label{tab:sparse-feature-entries-visual-genome-truncated-d16384-k128}
\end{table}

%% file: residual_statistics/residual_statistics_coco_d_8192.tex
\begin{table}[ht]
\centering
\begin{tabular}{lccc}
\hline
Model & $k=32$ & $k=64$ & $k=128$ \\
\hline
BAAI/bge-base-en-v1.5\_text & 0.1875 & 0.1619 & 0.1349 \\
BAAI/bge-large-en-v1.5\_text & 0.1922 & 0.1669 & 0.1425 \\
codefuse-ai/F2LLM-1.7B\_text & 0.2631 & 0.2413 & 0.2261 \\
codefuse-ai/F2LLM-4B\_text & 0.3041 & 0.2848 & 0.2742 \\
facebook/dinov2-base\_img & 0.3919 & 0.3476 & 0.3008 \\
facebook/dinov2-large\_img & 0.395 & 0.36 & 0.3219 \\
facebook/vit-mae-huge\_img & 0.318 & 0.2717 & 0.2333 \\
facebook/vit-mae-large\_img & 0.3392 & 0.278 & 0.2332 \\
google/gemma-3-1b-it\_text & 0.2771 & 0.2337 & 0.2009 \\
google/gemma-3-4b-it\_text & 0.2798 & 0.2424 & 0.2151 \\
google/siglip2-base-patch16-256\_img & 0.3411 & 0.2869 & 0.2401 \\
google/siglip2-base-patch16-256\_text & 0.2077 & 0.1786 & 0.1546 \\
google/siglip2-large-patch16-256\_img & 0.3424 & 0.2879 & 0.2442 \\
google/siglip2-large-patch16-256\_text & 0.2251 & 0.195 & 0.1686 \\
laion/CLIP-ViT-B-32-laion2B-s34B-b79K\_img & 0.3569 & 0.2947 & 0.2331 \\
laion/CLIP-ViT-B-32-laion2B-s34B-b79K\_text & 0.1808 & 0.1537 & 0.1292 \\
laion/CLIP-ViT-H-14-laion2B-s32B-b79K\_img & 0.3777 & 0.327 & 0.2796 \\
laion/CLIP-ViT-H-14-laion2B-s32B-b79K\_text & 0.1997 & 0.1718 & 0.1488 \\
meta-llama/Llama-3.2-1B-Instruct\_text & 0.2827 & 0.239 & 0.209 \\
meta-llama/Llama-3.2-3B-Instruct\_text & 0.2729 & 0.233 & 0.2064 \\
microsoft/beit-base-patch16-224\_img & 0.4253 & 0.3786 & 0.3287 \\
microsoft/beit-large-patch16-224\_img & 0.4536 & 0.4148 & 0.3703 \\
nomic-ai/nomic-embed-text-v1.5\_text & 0.1673 & 0.1444 & 0.1183 \\
nomic-ai/nomic-embed-text-v2-moe\_text & 0.1888 & 0.1581 & 0.1343 \\
openai/clip-vit-base-patch32\_img & 0.33 & 0.274 & 0.2222 \\
openai/clip-vit-base-patch32\_text & 0.1953 & 0.166 & 0.1376 \\
openai/clip-vit-large-patch14\_img & 0.3687 & 0.3201 & 0.2723 \\
openai/clip-vit-large-patch14\_text & 0.2135 & 0.1861 & 0.1596 \\
Qwen/Qwen3-1.7B-Base\_text & 0.2942 & 0.2571 & 0.2311 \\
Qwen/Qwen3-4B-Base\_text & 0.286 & 0.248 & 0.2208 \\
\hline
\end{tabular}
\caption{Residuals for dataset coco, sparse dimension d=8192, and k in \{32, 64, 128\}.}
\label{tab:residuals_coco_d_8192}
\end{table}

%% file: residual_statistics/residual_statistics_coco_d_16384.tex
\begin{table}[ht]
\centering
\begin{tabular}{lccc}
\hline
Model & $k=32$ & $k=64$ & $k=128$ \\
\hline
BAAI/bge-base-en-v1.5\_text & 0.1775 & 0.1523 & 0.1282 \\
BAAI/bge-large-en-v1.5\_text & 0.1812 & 0.1566 & 0.1342 \\
codefuse-ai/F2LLM-1.7B\_text & 0.2449 & 0.2227 & 0.2109 \\
codefuse-ai/F2LLM-4B\_text & 0.2817 & 0.2616 & 0.2532 \\
facebook/dinov2-base\_img & 0.3551 & 0.3135 & 0.2682 \\
facebook/dinov2-large\_img & 0.3578 & 0.3249 & 0.2877 \\
facebook/vit-mae-huge\_img & 0.3039 & 0.253 & 0.2182 \\
facebook/vit-mae-large\_img & 0.3278 & 0.259 & 0.219 \\
google/gemma-3-1b-it\_text & 0.2626 & 0.2188 & 0.1886 \\
google/gemma-3-4b-it\_text & 0.2657 & 0.2299 & 0.203 \\
google/siglip2-base-patch16-256\_img & 0.3191 & 0.2622 & 0.2221 \\
google/siglip2-base-patch16-256\_text & 0.1955 & 0.1685 & 0.1465 \\
google/siglip2-large-patch16-256\_img & 0.3224 & 0.2635 & 0.2262 \\
google/siglip2-large-patch16-256\_text & 0.212 & 0.1823 & 0.1589 \\
laion/CLIP-ViT-B-32-laion2B-s34B-b79K\_img & 0.3307 & 0.267 & 0.2176 \\
laion/CLIP-ViT-B-32-laion2B-s34B-b79K\_text & 0.1695 & 0.144 & 0.1227 \\
laion/CLIP-ViT-H-14-laion2B-s32B-b79K\_img & 0.3518 & 0.2973 & 0.2593 \\
laion/CLIP-ViT-H-14-laion2B-s32B-b79K\_text & 0.1882 & 0.1612 & 0.1405 \\
meta-llama/Llama-3.2-1B-Instruct\_text & 0.2692 & 0.2257 & 0.197 \\
meta-llama/Llama-3.2-3B-Instruct\_text & 0.2619 & 0.2225 & 0.1959 \\
microsoft/beit-base-patch16-224\_img & 0.3877 & 0.3418 & 0.2946 \\
microsoft/beit-large-patch16-224\_img & 0.4141 & 0.3753 & 0.3336 \\
nomic-ai/nomic-embed-text-v1.5\_text & 0.1581 & 0.1351 & 0.1113 \\
nomic-ai/nomic-embed-text-v2-moe\_text & 0.1789 & 0.1486 & 0.1275 \\
openai/clip-vit-base-patch32\_img & 0.3054 & 0.2497 & 0.2069 \\
openai/clip-vit-base-patch32\_text & 0.1833 & 0.1555 & 0.1306 \\
openai/clip-vit-large-patch14\_img & 0.3396 & 0.2915 & 0.2515 \\
openai/clip-vit-large-patch14\_text & 0.1994 & 0.1744 & 0.1513 \\
Qwen/Qwen3-1.7B-Base\_text & 0.2784 & 0.2426 & 0.2172 \\
Qwen/Qwen3-4B-Base\_text & 0.2719 & 0.2347 & 0.2086 \\
\hline
\end{tabular}
\caption{Residuals for dataset coco, sparse dimension d=16384, and k in \{32, 64, 128\}.}
\label{tab:residuals_coco_d_16384}
\end{table}

%% file: residual_statistics/residual_statistics_cc3m_d_8192.tex
\begin{table}[ht]
\centering
\begin{tabular}{lccc}
\hline
Model & $k=32$ & $k=64$ & $k=128$ \\
\hline
BAAI/bge-base-en-v1.5\_text & 0.294 & 0.2581 & 0.2207 \\
BAAI/bge-large-en-v1.5\_text & 0.305 & 0.2717 & 0.2379 \\
codefuse-ai/F2LLM-1.7B\_text & 0.3925 & 0.3703 & 0.3536 \\
codefuse-ai/F2LLM-4B\_text & 0.4361 & 0.4098 & 0.3927 \\
facebook/dinov2-base\_img & 0.4345 & 0.386 & 0.338 \\
facebook/dinov2-large\_img & 0.451 & 0.4106 & 0.3704 \\
facebook/vit-mae-huge\_img & 0.3056 & 0.2633 & 0.2236 \\
facebook/vit-mae-large\_img & 0.3292 & 0.2747 & 0.2275 \\
google/gemma-3-1b-it\_text & 0.3862 & 0.3447 & 0.3081 \\
google/gemma-3-4b-it\_text & 0.3787 & 0.3444 & 0.3175 \\
google/siglip2-base-patch16-256\_img & 0.3581 & 0.3085 & 0.2599 \\
google/siglip2-base-patch16-256\_text & 0.3132 & 0.2789 & 0.2422 \\
google/siglip2-large-patch16-256\_img & 0.3657 & 0.3149 & 0.2686 \\
google/siglip2-large-patch16-256\_text & 0.346 & 0.3102 & 0.2722 \\
laion/CLIP-ViT-B-32-laion2B-s34B-b79K\_img & 0.3696 & 0.312 & 0.2476 \\
laion/CLIP-ViT-B-32-laion2B-s34B-b79K\_text & 0.3021 & 0.2638 & 0.2261 \\
laion/CLIP-ViT-H-14-laion2B-s32B-b79K\_img & 0.407 & 0.36 & 0.3075 \\
laion/CLIP-ViT-H-14-laion2B-s32B-b79K\_text & 0.3483 & 0.3162 & 0.2775 \\
meta-llama/Llama-3.2-1B-Instruct\_text & 0.3323 & 0.2975 & 0.2715 \\
meta-llama/Llama-3.2-3B-Instruct\_text & 0.3399 & 0.307 & 0.2813 \\
microsoft/beit-base-patch16-224\_img & 0.4645 & 0.4152 & 0.3631 \\
microsoft/beit-large-patch16-224\_img & 0.498 & 0.4586 & 0.4115 \\
nomic-ai/nomic-embed-text-v1.5\_text & 0.2695 & 0.2355 & 0.1983 \\
nomic-ai/nomic-embed-text-v2-moe\_text & 0.276 & 0.2409 & 0.2103 \\
openai/clip-vit-base-patch32\_img & 0.3583 & 0.3013 & 0.2431 \\
openai/clip-vit-base-patch32\_text & 0.3167 & 0.276 & 0.2317 \\
openai/clip-vit-large-patch14\_img & 0.4035 & 0.3552 & 0.3026 \\
openai/clip-vit-large-patch14\_text & 0.3508 & 0.3145 & 0.2725 \\
Qwen/Qwen3-1.7B-Base\_text & 0.3252 & 0.2935 & 0.2681 \\
Qwen/Qwen3-4B-Base\_text & 0.3256 & 0.2934 & 0.2688 \\
\hline
\end{tabular}
\caption{Residuals for dataset cc3m, sparse dimension d=8192, and k in \{32, 64, 128\}.}
\label{tab:residuals_cc3m_d_8192}
\end{table}

%% file: residual_statistics/residual_statistics_cc3m_d_16384.tex
\begin{table}[ht]
\centering
\begin{tabular}{lccc}
\hline
Model & $k=32$ & $k=64$ & $k=128$ \\
\hline
BAAI/bge-base-en-v1.5\_text & 0.2689 & 0.2369 & 0.2039 \\
BAAI/bge-large-en-v1.5\_text & 0.2809 & 0.2496 & 0.219 \\
codefuse-ai/F2LLM-1.7B\_text & 0.3595 & 0.3358 & 0.3226 \\
codefuse-ai/F2LLM-4B\_text & 0.4031 & 0.3751 & 0.3586 \\
facebook/dinov2-base\_img & 0.4042 & 0.3594 & 0.3122 \\
facebook/dinov2-large\_img & 0.4169 & 0.3797 & 0.3404 \\
facebook/vit-mae-huge\_img & 0.2945 & 0.251 & 0.2135 \\
facebook/vit-mae-large\_img & 0.3194 & 0.2617 & 0.2177 \\
google/gemma-3-1b-it\_text & 0.3602 & 0.3175 & 0.2861 \\
google/gemma-3-4b-it\_text & 0.3545 & 0.3196 & 0.294 \\
google/siglip2-base-patch16-256\_img & 0.3387 & 0.2894 & 0.2465 \\
google/siglip2-base-patch16-256\_text & 0.2887 & 0.2582 & 0.2283 \\
google/siglip2-large-patch16-256\_img & 0.3459 & 0.2923 & 0.2533 \\
google/siglip2-large-patch16-256\_text & 0.3208 & 0.2857 & 0.2547 \\
laion/CLIP-ViT-B-32-laion2B-s34B-b79K\_img & 0.3475 & 0.2922 & 0.2366 \\
laion/CLIP-ViT-B-32-laion2B-s34B-b79K\_text & 0.2759 & 0.2413 & 0.2105 \\
laion/CLIP-ViT-H-14-laion2B-s32B-b79K\_img & 0.3816 & 0.3348 & 0.291 \\
laion/CLIP-ViT-H-14-laion2B-s32B-b79K\_text & 0.3218 & 0.2906 & 0.2598 \\
meta-llama/Llama-3.2-1B-Instruct\_text & 0.3104 & 0.2749 & 0.2514 \\
meta-llama/Llama-3.2-3B-Instruct\_text & 0.3183 & 0.2853 & 0.2607 \\
microsoft/beit-base-patch16-224\_img & 0.4346 & 0.3879 & 0.3389 \\
microsoft/beit-large-patch16-224\_img & 0.4649 & 0.4286 & 0.3868 \\
nomic-ai/nomic-embed-text-v1.5\_text & 0.2454 & 0.2153 & 0.1813 \\
nomic-ai/nomic-embed-text-v2-moe\_text & 0.2498 & 0.2205 & 0.1946 \\
openai/clip-vit-base-patch32\_img & 0.3381 & 0.2838 & 0.2336 \\
openai/clip-vit-base-patch32\_text & 0.2907 & 0.2529 & 0.2184 \\
openai/clip-vit-large-patch14\_img & 0.3774 & 0.3324 & 0.2877 \\
openai/clip-vit-large-patch14\_text & 0.3232 & 0.291 & 0.2579 \\
Qwen/Qwen3-1.7B-Base\_text & 0.3044 & 0.2725 & 0.2496 \\
Qwen/Qwen3-4B-Base\_text & 0.3053 & 0.274 & 0.2503 \\
\hline
\end{tabular}
\caption{Residuals for dataset cc3m, sparse dimension d=16384, and k in \{32, 64, 128\}.}
\label{tab:residuals_cc3m_d_16384}
\end{table}

%% file: residual_statistics/residual_statistics_visual_genome_d_8192.tex
\begin{table}[ht]
\centering
\begin{tabular}{lccc}
\hline
Model & $k=32$ & $k=64$ & $k=128$ \\
\hline
BAAI/bge-base-en-v1.5\_text & 0.2294 & 0.1795 & 0.1412 \\
BAAI/bge-large-en-v1.5\_text & 0.2268 & 0.1796 & 0.1431 \\
codefuse-ai/F2LLM-1.7B\_text & 0.3082 & 0.2586 & 0.2302 \\
codefuse-ai/F2LLM-4B\_text & 0.3321 & 0.2852 & 0.2589 \\
facebook/dinov2-base\_img & 0.3875 & 0.3431 & 0.2962 \\
facebook/dinov2-large\_img & 0.3908 & 0.356 & 0.3179 \\
facebook/vit-mae-huge\_img & 0.3182 & 0.2718 & 0.2335 \\
facebook/vit-mae-large\_img & 0.3385 & 0.2777 & 0.2333 \\
google/gemma-3-1b-it\_text & 0.2994 & 0.2401 & 0.2033 \\
google/gemma-3-4b-it\_text & 0.2907 & 0.243 & 0.2097 \\
google/siglip2-base-patch16-256\_img & 0.3372 & 0.2823 & 0.2365 \\
google/siglip2-base-patch16-256\_text & 0.2601 & 0.2085 & 0.1727 \\
google/siglip2-large-patch16-256\_img & 0.3392 & 0.2837 & 0.2399 \\
google/siglip2-large-patch16-256\_text & 0.2829 & 0.2296 & 0.1919 \\
laion/CLIP-ViT-B-32-laion2B-s34B-b79K\_img & 0.3516 & 0.2896 & 0.2296 \\
laion/CLIP-ViT-B-32-laion2B-s34B-b79K\_text & 0.2426 & 0.1938 & 0.1586 \\
laion/CLIP-ViT-H-14-laion2B-s32B-b79K\_img & 0.3716 & 0.3208 & 0.2744 \\
laion/CLIP-ViT-H-14-laion2B-s32B-b79K\_text & 0.2678 & 0.2172 & 0.1815 \\
meta-llama/Llama-3.2-1B-Instruct\_text & 0.2692 & 0.215 & 0.1814 \\
meta-llama/Llama-3.2-3B-Instruct\_text & 0.2748 & 0.2221 & 0.1881 \\
microsoft/beit-base-patch16-224\_img & 0.4204 & 0.3736 & 0.3237 \\
microsoft/beit-large-patch16-224\_img & 0.4504 & 0.411 & 0.3662 \\
nomic-ai/nomic-embed-text-v1.5\_text & 0.217 & 0.1696 & 0.1309 \\
nomic-ai/nomic-embed-text-v2-moe\_text & 0.2395 & 0.1834 & 0.1479 \\
openai/clip-vit-base-patch32\_img & 0.3248 & 0.269 & 0.2188 \\
openai/clip-vit-base-patch32\_text & 0.2491 & 0.2004 & 0.1618 \\
openai/clip-vit-large-patch14\_img & 0.3626 & 0.3144 & 0.2667 \\
openai/clip-vit-large-patch14\_text & 0.2664 & 0.2187 & 0.182 \\
Qwen/Qwen3-1.7B-Base\_text & 0.2799 & 0.23 & 0.1988 \\
Qwen/Qwen3-4B-Base\_text & 0.2681 & 0.2183 & 0.1863 \\
\hline
\end{tabular}
\caption{Residuals for dataset visual\_genome, sparse dimension d=8192, and k in \{32, 64, 128\}.}
\label{tab:residuals_visual_genome_d_8192}
\end{table}

%% file: residual_statistics/residual_statistics_visual_genome_d_16384.tex
\begin{table}[ht]
\centering
\begin{tabular}{lccc}
\hline
Model & $k=32$ & $k=64$ & $k=128$ \\
\hline
BAAI/bge-base-en-v1.5\_text & 0.2211 & 0.171 & 0.1341 \\
BAAI/bge-large-en-v1.5\_text & 0.219 & 0.17 & 0.1354 \\
codefuse-ai/F2LLM-1.7B\_text & 0.294 & 0.2425 & 0.2142 \\
codefuse-ai/F2LLM-4B\_text & 0.3156 & 0.2667 & 0.2407 \\
facebook/dinov2-base\_img & 0.3478 & 0.3063 & 0.2608 \\
facebook/dinov2-large\_img & 0.3508 & 0.318 & 0.2804 \\
facebook/vit-mae-huge\_img & 0.3015 & 0.2512 & 0.2177 \\
facebook/vit-mae-large\_img & 0.327 & 0.2568 & 0.2196 \\
google/gemma-3-1b-it\_text & 0.2863 & 0.2244 & 0.1899 \\
google/gemma-3-4b-it\_text & 0.2797 & 0.2298 & 0.1985 \\
google/siglip2-base-patch16-256\_img & 0.3147 & 0.257 & 0.219 \\
google/siglip2-base-patch16-256\_text & 0.2496 & 0.1959 & 0.1641 \\
google/siglip2-large-patch16-256\_img & 0.318 & 0.258 & 0.2223 \\
google/siglip2-large-patch16-256\_text & 0.2706 & 0.2147 & 0.1809 \\
laion/CLIP-ViT-B-32-laion2B-s34B-b79K\_img & 0.3247 & 0.2601 & 0.2141 \\
laion/CLIP-ViT-B-32-laion2B-s34B-b79K\_text & 0.2319 & 0.1829 & 0.1499 \\
laion/CLIP-ViT-H-14-laion2B-s32B-b79K\_img & 0.3437 & 0.2901 & 0.2535 \\
laion/CLIP-ViT-H-14-laion2B-s32B-b79K\_text & 0.2564 & 0.2042 & 0.1714 \\
meta-llama/Llama-3.2-1B-Instruct\_text & 0.2606 & 0.2025 & 0.1715 \\
meta-llama/Llama-3.2-3B-Instruct\_text & 0.2655 & 0.211 & 0.1799 \\
microsoft/beit-base-patch16-224\_img & 0.3803 & 0.3344 & 0.2864 \\
microsoft/beit-large-patch16-224\_img & 0.4077 & 0.3686 & 0.3261 \\
nomic-ai/nomic-embed-text-v1.5\_text & 0.2092 & 0.1608 & 0.1247 \\
nomic-ai/nomic-embed-text-v2-moe\_text & 0.2323 & 0.1738 & 0.1399 \\
openai/clip-vit-base-patch32\_img & 0.299 & 0.2433 & 0.2013 \\
openai/clip-vit-base-patch32\_text & 0.2377 & 0.1876 & 0.1531 \\
openai/clip-vit-large-patch14\_img & 0.3318 & 0.284 & 0.2443 \\
openai/clip-vit-large-patch14\_text & 0.2542 & 0.2055 & 0.172 \\
Qwen/Qwen3-1.7B-Base\_text & 0.2689 & 0.2172 & 0.1879 \\
Qwen/Qwen3-4B-Base\_text & 0.2583 & 0.2077 & 0.1775 \\
\hline
\end{tabular}
\caption{Residuals for dataset visual\_genome, sparse dimension d=16384, and k in \{32, 64, 128\}.}
\label{tab:residuals_visual_genome_d_16384}
\end{table}